\documentclass{svmono}
\usepackage{float}
\usepackage{microtype}

\usepackage{newtxtext}
\usepackage{newtxmath}
\usepackage[bottom]{footmisc}

\usepackage{nicefrac}
\usepackage{listings}
\usepackage{multicol}
\PassOptionsToPackage{hyphens}{url}
\usepackage[pagebackref=true]{hyperref}
\usepackage{graphicx}
\usepackage{rotating}
\usepackage{pdflscape}
\usepackage{tikz}
\usetikzlibrary{positioning}
\usetikzlibrary{arrows}
\usetikzlibrary{chains}
\usetikzlibrary{shapes.geometric}
\usepackage{smartdiagram}
\usepackage{makeidx}
\makeindex

  \title{{\Huge\bf \ \\ \ \\
      Learning to Play}\\ \ \\
    \Large\em  Reinforcement Learning and Games\\ \ \\ \ \\
        \includegraphics[width=7cm]{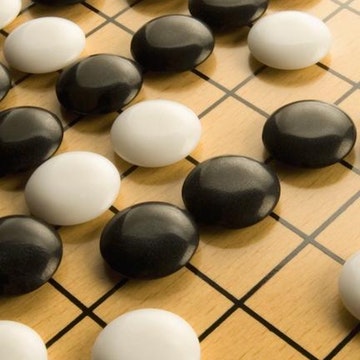}} 
\author{Aske Plaat}
\date{
  \today}
\begin{document}
\maketitle
\frontmatter
\begin{dedication}
To my students
\end{dedication}
\renewcommand*{\lstlistlistingname}{List of Algorithms}

\setcounter{tocdepth}{1}
\tableofcontents

\lstset{language=Python}
\lstset{escapeinside={(*}{*)}}
\lstset{frame=lines,captionpos=b}
\lstset{basicstyle=\footnotesize\ttfamily,breaklines=true}


\preface

Amazing breakthroughs in reinforcement learning  have taken
place. Computers teach themselves  to play Chess and Go and beat world
champions.
There is  talk about expanding application
areas towards  general  artificial  intelligence (AI).
The breakthroughs in Backgammon, Checkers, Chess, Atari, Go, Poker,
and StarCraft have shown that we can build machines that exhibit
intelligent game playing of 
the highest level. These successes have been widely publicized in the
media, and inspire AI entrepreneurs and scientists alike. Reinforcement learning in games has become a mainstream AI research
topic.
It is a broad topic, and the 
successes build on a  range of diverse techniques, from exact planning
algorithms, to adaptive sampling,  deep function approximation, and
ingenious self-play methods.


Perhaps because of the breadth of these
technologies, or 
because of the recency of the breakthroughs, there are few books that
explain these  methods in  depth. This book covers  all methods in
one comprehensive volume,
explaining the latest research, bringing you as close as possible to
working implementations, with many references to the original
research papers. 

The
programming examples  in this book are in Python, the language in which most
current reinforcement learning research is
conducted. We help you to get started with  machine learning frameworks
such as  Gym, TensorFlow,
and Keras, and  provide exercises to help 
understand how AI is 
learning to play. 

\ \\

\noindent
This is not a typical  reinforcement learning textbook.
Most books on reinforcement learning take the single-agent
perspective, of path finding and robot planning. 
We take as inspiration the  breakthroughs in game
playing, and use two-agent  games to explain the full power of deep reinforcement
learning.

Board games have always
been associated with reasoning and intelligence. Our games perspective
allows us to make connections with artificial intelligence and 
general intelligence, giving a  philosophical flavor to an
otherwise technical field.

\subsubsection*{Artificial Intelligence}
Ever since my early days as a  student I have
been captivated by artificial intelligence, by machines
that  behave in seemingly intelligent ways. Initially I had been
taught that, because
computers were deterministic  machines, they could never do
something new. 
Yet in AI these  machines do complicated things such as recognize
patterns, and play  
Chess games. Actions  emerged from these machines,
behavior that appeared not to have been programmed into them.  The actions
seemed new, and even creative, at times. 

For my thesis I got to work  on game playing programs for combinatorial games such
as Chess, Checkers, and Othello. The paradox became even more
apparent. These game playing programs all followed an elegant
architecture, consisting of a search function and an evaluation
function.\footnote{The search function  simulates the
  kind of look-ahead that many human game players do in their head, and the evaluation
function assigns a numeric score to a board position indicating how
good the position is.}  These two
 functions together could  find good moves all by
themselves. Could intelligence be so simple?

The search-evaluation architecture has been around since 
the earliest days of computer Chess. Together with minimax, it was proposed in a 1952 paper
by Alan 
Turing, mathematician, code-breaking war hero, and  one of the
fathers of computer science and artificial intelligence. 
The search-evaluation architecture is also used in Deep Blue, the
Chess program that beat World Champion Garry Kasparov in
1997 in New York.

After that historic moment, the
attention of the AI community shifted to a new  game with which to further develop
ideas for 
intelligent game play. It was the East Asian game of Go that emerged as
the new grand test of intelligence. Simple, elegant, and
mind-bogglingly complex. 

This new game spawned the creation of important new algorithms, and
not one, but two, paradigm shifts. The first algorithm to upset the worldview of games
researchers was  Monte Carlo
Tree Search, in 2006.  Since the 1950s generations of game playing researchers, myself included, were brought
up with minimax. The essence
of minimax is to look ahead as far as you can, to then choose the best
move, and to make sure that all moves are tried (since behind
seemingly harmless moves deep attacks may hide that you can only
uncover if you search all moves). And now Monte Carlo Tree
Search introduced randomness into the search, and
sampling,  deliberately missing moves. Yet it worked in
Go, and much better than minimax.

Monte Carlo Tree Search  caused a strong
increase in playing strength, although not yet enough to beat world
champions.
For that, we had to wait another ten years.

In 2013 our worldview was in for a new
shock, because again a new paradigm shook up the conventional wisdom. Neural
networks were widely viewed to 
be too slow and too inaccurate to be useful in games. Many Master's theses
of stubborn students had sadly confirmed this to be the case. Yet in
2013 GPU power allowed the use of a simple neural net to learn to
play Atari video games 
just from looking at the video pixels, using a method called deep
reinforcement learning. Two years and much hard work later, deep
reinforcement learning was combined with Monte Carlo Tree
Search in the program AlphaGo. The level of play was improved  so much
that a year later finally world champions in Go were
beaten, many years before experts had expected that this would
happen. And in other games, such as StarCraft and Poker, self-play
reinforcement learning also caused breakthroughs.
%


The AlphaGo wins were widely publicized. They have had a large impact,
on science, on the public perception of AI, and on
society. AI researchers everywhere were invited to give  lectures. Audiences  wanted to know what
had happened, whether computers finally had become intelligent, what more could be expected
from AI, and what all this would mean for the future of the human
race. Many start-ups were created, and existing technology companies started
researching what AI could do for them.
\ \\

\noindent
The modern history of computer games spans some 70 years.
There has been much excitement. Many ideas were tried, some with success. 
Games research in reinforcement learning has witnessed multiple paradigm shifts, going from
heuristic planning, to adaptive sampling, to
deep learning, to self-play. The achievements are large, and so is the range of
techniques that are used. We are now at a point where the techniques
have matured somewhat, and  achievements can be documented and put
into perspective. 


In explaining the technologies, I will tell the story of how one
kind of  intelligence 
works,  the intelligence needed to play two-person games of tactics and
strategy.
(As to knowing the future of the human race, surely  more
is needed than an understanding of heuristics, deep reinforcement learning, and
game playing programs.) It will be a story involving many scientists,
programmers, and game enthusiasts, all fascinated by the same goal:
creating artificial intelligence. Come and join this fascinating ride. 

\extrachap{Acknowledgments}

This book would not have been possible without the help of many
friends. 
First of all, I want to thank everybody at LIACS, for creating such a
wonderful and vibrant CS \& AI department to work in. So many people
 make this place such a joy to do research; I love you all!

Many people have read and commented on drafts of this book. Thank you
 Vedran Dunjko, Mike Huisman, Walter Kosters,  Wojtek Kowalczyk, Alfons Laarman, Thomas Moerland,
and Mike Preu\ss\ 
 for your very sharp eyes and  criticism. Thank you to Andrius Bernatavicius,
Hui Wang, Zhao Yang, Michael Emmerich, Wojtek Kowalczyk, Mike Preu\ss, Thomas Moerland, Xuhan Lu,
Joost Broekens, Matthias M\"uller-Brockhausen,  my friends and fellow
conspirators in the reinforcement learning group. In addition, thank
you to Lise Stork and Wojtek Kowalczyk for discussions on deep
learning and zero-shot learning. Thanks to Michael Emmerich for
curriculum learning. Thank you to  Walter Kosters, Frank Takes, Jan
van Rijn, and Jonathan Vis for discussions on teaching, MCTS, and combinatorial
games. Thanks to Mike Preu\ss\ for countless discussions and for running
the games research at Leiden. Thanks to Max van Duijn for inspiring
discussions on theory of mind and creativity.
%
%
Thank you Walter Kosters for a shared love of combinatorics and a sense of humor.
%
%
%
A very special thank you to Thomas B\"ack, with whom I have had so many discussions on science, the universe, and everything. Without you, this book would not have been possible.

I thank all students of the reinforcement learning course in Leiden,
past, present, and future, 
for their wonderful enthusiasm, sharp questions, and many suggestions. 



Finally, I thank Saskia, Isabel, Rosalin, and little Lily, for being who they
are, for knowing no limits, for playing games, for enduring me, and for their boundless love.

\vspace{\baselineskip}
\begin{flushright}\noindent
Leiden,\hfill {\it ~ }\\
April 2020\hfill {\it Aske Plaat}\\
\end{flushright}

\mainmatter


\chapter{Introduction}\label{chap:intro}


\section{Tuesday March 15, 2016}
Tuesday  March 15, 2016, is an important day in the history of
artificial intelligence, and not just in the history of artificial intelligence, but in the
history of intelligence, period. That 
day was the final day of a five-game match in which a computer
challenged one of the strongest human Go players. Go is an ancient game of strategy, played by millions around
the world, originating in East Asia. Playing the game well
requires  years of training and a considerable amount of intelligence. Dedicated
Go schools exist in China, Korea, and Japan, and professional Go
players enjoy star status in their countries and can live well from
their prize money. Go is considered to be a hallmark of human
intelligence and creativity, and the game was thought to be safe from
machine intelligence for at 
least another decade.

\begin{figure}[t]
\begin{center}
\includegraphics[width=\textwidth]{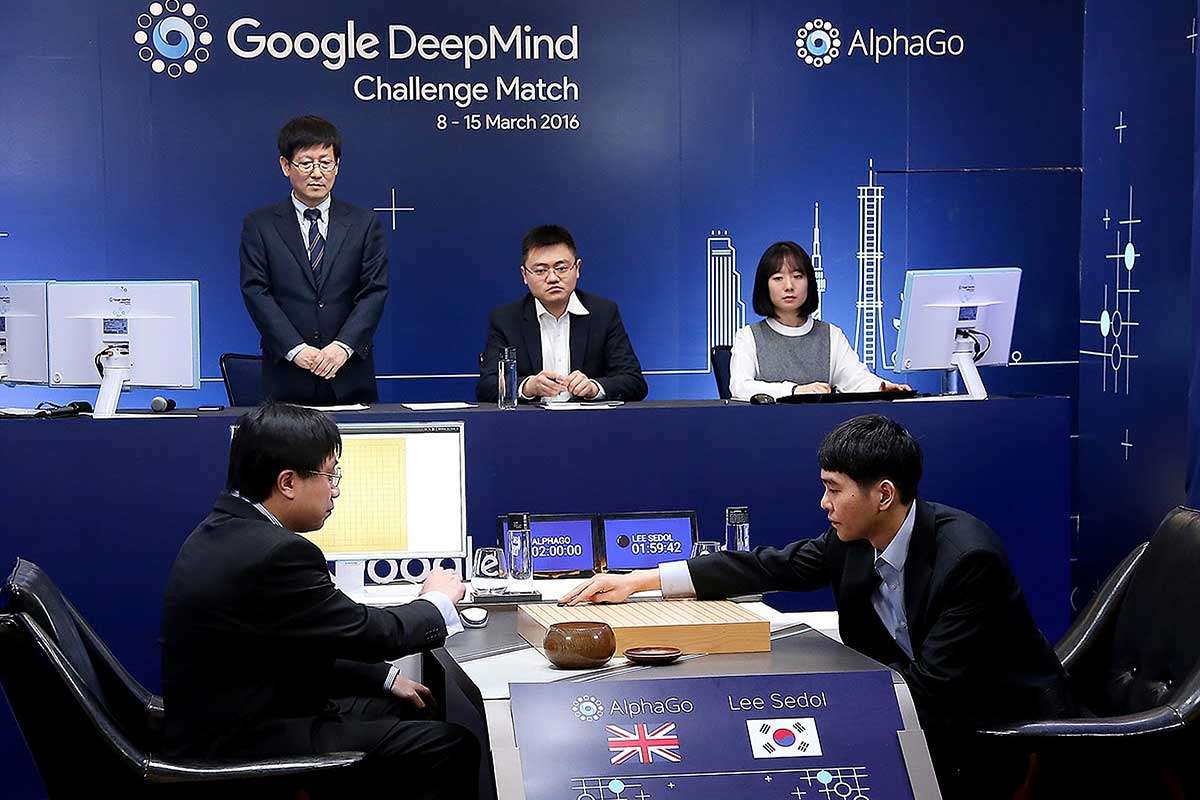}
\caption{AlphaGo versus Lee Sedol}\label{fig:fourseasons}
\end{center}
\end{figure}

Yet on March 15, 2016, at the luxurious Four Seasons Hotel in Seoul, the computer program
AlphaGo beat top Go professional Lee Sedol (Fig.~\ref{fig:fourseasons}). AlphaGo had been developed by a team of
scientists of the AI company DeepMind. The prize money, 1 million dollars, was donated 
to charities. Lee Sedol received 170,000 dollars for his
efforts.  The match consisted  of 5 games, and the human
champion was beaten 4--1. A subsequent match in 2017 in China against
champion Ke Jie was
also won by the computer. Commentators described AlphaGo's level of
play as very strong, and the style of play as refreshing, with a few highly
surprising moves, that some  described as  beautiful and  creative. AlphaGo was awarded a 9 dan professional rank, the
highest rank possible, by the Chinese Weiqi association, and the  journal
\emph{Science} chose the event as one of the {\em breakthrough of the year\/} runners
up. Today, human Go players are taking inspiration from the games
played by AlphaGo, trying to learn from its deep insights and unusual moves~\cite{zhou2018}.


The breakthrough performance of AlphaGo came as a surprise to the
world and to the research community. The level
of play of computer Go programs had been stagnant around the strong
amateur level for years. Experts in computer games had expected grandmaster level play to be at least ten years away, let alone beating the
world's strongest players.

The techniques used in AlphaGo are the result of many
years of research, and cover a surprisingly wide range of topics. It is a culmination of decades of research by many
researchers. A main motivation for writing this book is to provide a
comprehensive overview of the technologies that  
led to the creation of AlphaGo.

This brings us to the  problem statement of this book,  to describe
the core of what this book is about.

\subsubsection*{Problem Statement}
The main question that we are concerned with in this book is
the following:
\begin{quotation}
{\em   What are the machine learning methods that are
used in Chess and Go to achieve a level of play stronger than the strongest humans?}
\end{quotation}
Among the technologies used are (1) heuristic planning methods, (2) adaptive sampling,
(3) deep reinforcement learning, and (4) self-play.  We
provide an in-depth introduction to  all four technologies, each in their
own chapter. We will explain how
they work with example 
code. 
We use the formalism of reinforcement learning as the common language
to describe all the technologies that are covered
in this book, both for planning and learning. We describe the
algorithms and architectures.

Our 
style is  hands-on, providing the intuition behind the
algorithms to apply them in practice. References to the 
literature are provided to find more details and  the relevant
mathematics behind the methods.

\section{Intended Audience}
Now that we know what this book is about, we can see who this book is
for.

This book is intended for you if you are excited about finding out how
computers play games, and which  techniques  are used to
achieve such amazing displays of intelligent behavior. We aim to be
comprehensive, to provide all necessary details. Basic introductions to
Markov decision processes, reinforcement learning, and deep learning are
provided. We also aim to make
these details more accessible than the scientific 
papers on which the content is based, translating formulas in
research papers into  intuitive concepts. The level of difficulty of
this book is targeted at a graduate-level university course in AI. We
explain the algorithms, and we provide opportunity for practice.

Our approach to artificial intelligence is hands-on. After each new method is
described, we will show  how the concepts can be implemented in 
examples, exercises, and small  programs. After basic methods are
explained, move advanced enhancements are discussed, enhancements that
are necessary to achieve high performance in practice.

Games AI is an open
 field. Researchers have released 
their  algorithms and code allowing replication of results. This is a
sharing-oriented research 
 field. We point you to code bases at GitHub throughout the
 book. Even when  you 
 do not have access to Google-scale compute power, at the end of
this book, you should be able to create a fully functioning self-learning
program, and you can  join the active community
of reinforcement learning researchers. For that purpose,
Appendix~\ref{ch:env} contains pointers to software environments and
open-source code frameworks
for machine learning, deep reinforcement learning, and games.

As you may have noticed, there is great excitement about  
reinforcement learning, excitement that also prompted the writing of this
book. The algorithms and the pointers to code  also serve this purpose: to
make it easy for  you to join the excitement and to stimulate you to join
the people already active in the field.

\subsubsection*{Graduate Course}
This book is designed for a 
graduate course in Reinforcement Learning with Games. The material
is organized in chapters that  roughly correspond to a
full semester 
course. The availability of source code provides for practical
assignments where you can get your hands dirty and immerse
yourself in this fascinating topic. Pieces of working 
code   can be extended to your liking for further study and research.

Each chapter starts with an introduction in which the core problems
and core solution concepts of the chapter are summarized. Chapters end
with questions 
to refresh your memory of the material that was covered, and with
larger programming exercises. Chapters are concluded with a summary
and with more pointers to the literature.

\begin{figure}
\begin{center}
\begin{tikzpicture}[>=triangle 45,
  desc/.style={
		scale=1.0,
		rectangle,
		rounded corners,
		draw=black, 
		}]
  \node [desc] (intro) at    (0,8.5) {1. Introduction};
  \node [desc] (games) at    (0,7.5) {2. Intelligence and Games};
  \node [desc] (rl) at           (0,6.5) {3. Reinforcement Learning};
  \node [desc] (heur)  at       (0,5.5) {4. \bf Heuristic Planning};
  \node [desc] (mcts) at        (0,4.5) {5. \bf Adaptive Sampling};
  \node [desc] (super) at       (0,3.5) {6. \bf Function Approximation};
  \node [desc] (self) at        (0,2.5) {7. \bf Self-Play};
  \node [desc] (agi) at      (0,1.5) {8. Conclusion};
  \draw [semithick,->] (intro) -- (games);
  \draw [semithick,->] (games) -- (rl);
  \draw [semithick,->] (rl) -- (heur);
  \draw [semithick,->] (heur) -- (mcts);
  \draw [semithick,->] (mcts) -- (super);
  \draw [semithick,->] (super) -- (self);
  \draw [semithick,->] (self) -- (agi);

\end{tikzpicture}
\caption[Chapter overview]{Overview of the chapters of this book}\label{fig:overview}
  \end{center}
\end{figure}
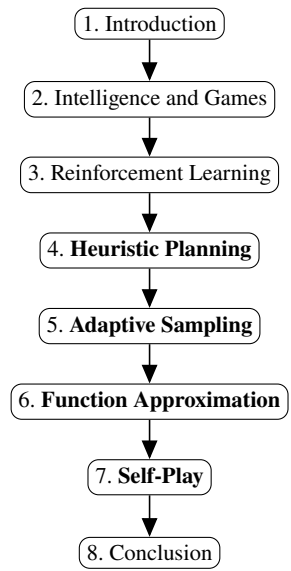

Figure~\ref{fig:overview} shows the structure of the book. There is a
logical progression in the chapters; they build
on each other.  It should
be possible to cover all chapters in a one-semester course, possibly
skipping the starred sections. Normally the chapters would be taught
in sequence. Chapters~\ref{chap:games} and
\ref{chap:agi} are less technical, and provide some reflection on
intelligence and artificial intelligence. They can
be combined towards the end,  putting the technical material first.

All links to web pages were correct at the time of writing and have
been stable for some time. The 
companion website for this book  \url{https://learningtoplay.net} 
contains free updates, slides, and other course material that you are
welcome to explore and use in your course.

\subsubsection*{Prerequisite Knowledge}
In order to build up an understanding and to experiment with the
algorithms, a general
computer science background at undergraduate level is 
required. Some  proficiency in programming is needed,
preferably in Python, since that is the language that is currently
used in the field of reinforcement learning, and the language that we
use in this book. Also some
background in undergraduate AI is beneficial, although we strive to be
comprehensive and self-contained.

\subsubsection*{Experiment and Theory}
As in many fields, progress in reinforcement learning is driven by an
interaction between experiment and theory. Experiments are designed to better
understand theory, and new theory is developed to explain experimental
results. Reinforcement learning rests on a strong theoretical
foundation in computational complexity and decision theory, some of
which is introduced in Chap.~\ref{chap:rl}.

However, many of the 
eye-catching advances in the field (Deep Blue, TD-Gammon, and most recently
Atari and AlphaGo) are driven by experimental and engineering
advances. There are some intuitions to explain results, but many
theoretical questions remain. 
This state of affairs is reflected in the style of this book.
In the later chapters we mention experiments and
intuitions, but few formulas, lemmas, or proofs.  Establishing a 
comprehensive theory of deep reinforcement learning is still future
work.

\begin{table}[t]
  \begin{center}

\begin{tabular}{lcccc}
  {\bf Chapter} & {\bf Algorithm} & {\bf Select} & {\bf Backup} & {\bf
                                                                  Eval}\\ \hline\hline
   Chap.~\ref{chap:search} & alpha-beta & left to right & minimax
                                   & heuristic \\ 
  Chap.~\ref{chap:trial} & MCTS & adaptive & average & sample \\ 
   Chap.~\ref{chap:dqn} & DQN & - & - & generalize \\ 
   Chap.~\ref{chap:plan} & self-play & adaptive & average &
                                                              generalize
  \\ \hline
\end{tabular}

  \caption{Heuristic-Sample-Generalize}\label{tab:heursampgen}
  \end{center}
\end{table}

\section{Outline}\label{sec:heurlearn}
We have almost come to the end of the introduction. Let us have a
closer look at what we can expect in the remainder of this book.

This book covers the technologies behind the major AI breakthroughs in
reinforcement learning in
games;  see
Table~\ref{tab:heursampgen}.
Chess, Atari, and Go required
surprisingly diverse techniques. The methods in this book go from
heuristic planning to self-play.  First we describe 
methods that need domain-specific heuristics, and we 
end with a method that is able to generalize features in many
domains, and to teach itself to do so. We will describe in detail
heuristic methods, sampling, and 
generalization. Together the chapters tell a story of how AI progressed from
artificial specific (or narrow)
intelligence towards somewhat more general intelligence.

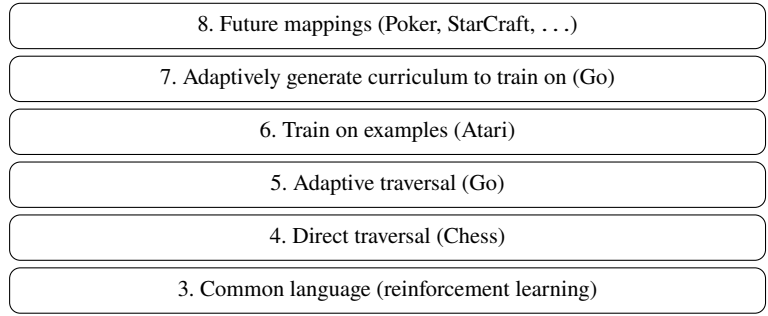
\begin{figure}[t]
\begin{center}
\begin{tikzpicture}[>=triangle 45,
  desc/.style={
		scale=1.0,
		rectangle,
		rounded corners,
		draw=black, 
		}]

  \node [desc,minimum width=10cm,minimum height=0.6cm] (rl) at
  (0,0.7) {3. Common language (reinforcement learning)};
  \node [desc,minimum width=10cm,minimum height=0.6cm] (mm) at
  (0,1.4) {4. Direct traversal (Chess)};
  \node [desc,minimum width=10cm,minimum height=0.6cm] (mcts) at
  (0,2.1) {5. Adaptive traversal (Go)};
  \node [desc,minimum width=10cm,minimum height=0.6cm] (dqn) at
  (0,2.8) {6. Train on examples (Atari)};
  \node [desc,minimum width=10cm,minimum height=0.6cm] (self) at
  (0,3.5) {7. Adaptively generate curriculum to train on (Go)};
  \node [desc,minimum width=10cm,minimum height=0.6cm] (future) at
  (0,4.2) {8. Future  mappings (Poker, StarCraft, $\ldots$)};

\end{tikzpicture}
\caption[Stack of chapters]{Stack of chapters, with  (from bottom to
  top) increasingly advanced mappings from
problems to methods}\label{fig:layersch}
  \end{center}
\end{figure}

Figure~\ref{fig:layersch} gives a related overview of the topics of this
book, in terms of mappings between problems and methods that go from
specific to general. The first chapters are specific in the 
sense that the methods make  direct use of the structure of the problem
to  traverse board positions. The later chapters
are more abstract, and the mapping between method and problem becomes
less direct. Methods become adaptive, and learn features from examples of
the problem domain,
instead of directly traversing  the problem itself. Self-play methods that teach
themselves which examples to learn from, are described. 

The story of games research is the story of AI researchers who were
learning how to 
make algorithms that  learn to learn.

\subsubsection*{*Starred Sections}
The chapters are organized from planning to learning to
self-play. Figure~\ref{fig:overview} gives an overview  of the 
chapters. The names of the four core technical chapters are printed in bold.

Sections and exercises whose name starts
with a *star are  advanced material, which is essential for understanding
high-performing implementations, but may be skipped when in a hurry, 
without missing the main thread.

We will now describe the main topics of this book. Please
 refer to Table~\ref{tab:heursampgen} and
Fig.~\ref{fig:layersch}, since the concepts
Heuristic-Sample-Generalize and (in)direct mappings between methods
and problem spaces are important threads.

\subsubsection*{The Chapters}
Let us now discuss the chapters in more detail.
We will
start in Chap.~\ref{chap:games} with early thoughts on intelligence
and intelligent
machines. We then provide a brief historical overview of some of the
important  games and game 
playing programs, starting with the early designs of Claude Shannon,
Christopher Strachey, and Alan Turing.

In the next chapter, Chap.~\ref{chap:rl}, we introduce reinforcement
learning. The concepts and formalizations of reinforcement learning
are at the basis of the historic advances in artificial intelligence
that we have recently witnessed. Concepts such as value function,
policy function, and
Q-learning are discussed in this chapter.

In Chap.~\ref{chap:search}, we 
introduce heuristic planning.
In this chapter we discuss the  principles of the search methods
that are at the basis of  AI breakthroughs in
Checkers and Chess. Heuristic planning in games
uses domain-specific knowledge to reduce the size of the state space, and then
uses minimax planning to traverse that reduced state space.
The planning methods of heuristic planning are simple and rigid, and it turns out that many
enhancements are needed to achieve good performance in real game
playing programs. We introduce
enhancements that  overcome some of the rigidity of the minimax
approach. 

In the next chapter, Chap.~\ref{chap:trial}, we delve deeper into  the
principles of learning by trial and error. Major topics in this
chapter are the 
exploration/exploitation trade-off and the concept of evaluation by
random playout, leading to the Monte Carlo Tree Search (MCTS)
algorithm. The algorithm in this  
chapter is variable-depth and variable-width by design, and great performance improvements, especially in $9\times 9$ Go,
are achieved with MCTS. The chapter ends with a  discussion of enhancements of MCTS.

Then, we switch paradigms from planning to learning. Instead of {\em traversing\/} the
states directly and finding good states, we now try to {\em generalize\/} features from the states. We start in
Chap.~\ref{chap:super} with a  review of deep learning
methods for supervised learning. Among many other things, convolutional
networks are discussed, and the problem of overfitting. In 2012 deep learning methods caused a
breakthrough in image recognition performance. Much of the
current interest in artificial intelligence is caused by this
breakthrough for which Bengio, Hinton, and LeCun received the 2018
Turing award. 
Section~\ref{sec:dqn} looks at how these methods can be used in a
reinforcement learning setting:
generalization of features in deep neural networks for learning by
trial and error, of which  Deep Q-Networks (DQN) is the best known
algorithm. Q-learning with function approximation and bootstrapping is
in principle unstable, 
and  important enhancements had to be developed to achieve stable
training. With these enhancements, we find that deep reinforcement learning can be an effective way to improve
on heuristic evaluation functions. 

Chapter~\ref{chap:plan} combines planning and training 
to create self-learning systems. This is a long chapter in which
the methods behind AlphaGo are explained, and AlphaGo's variants. AlphaGo Zero teaches itself
to play Go from scratch. This
kind of learning  resembles human learning, creating much interest in
the research community, and some self-play and self-learning variants
and other related research are discussed.

Finally,
we conclude in Chap.~\ref{chap:conclusions} by reflecting on 
artificial intelligence progress, and on the relation to natural
intelligence. We discuss possible future research directions.

The appendices contain an overview of open-source deep learning environments
that are suitable for experimentation, details on the AlphaGo implementations that were too detailed
for the main text, the tournaments of AlphaGo against Fan Hui, Lee Sedol,
and Ke Jie, pointers about learning to play and program Go and Chess,
and some pointers to the Python programming
language.
\\

\noindent
After this overview, let us begin our journey through the world of
reinforcement learning and games.

\chapter{Intelligence and Games}\label{chap:games}



As long as we have been around, humans have had dreams about
artificial forms of intelligence, wondering if it would be possible to
make intelligent machines. It is  wonderful 
to live in a time where we are able to see the first realizations of these dreams
appearing.

This chapter introduces the field of study of this book,  artificial
intelligence and games.

We will briefly touch upon the nature of intelligence, human and
artificial. We will see that most AI researchers take a pragmatic,
behavioristic approach to intelligence. Board games such as
Chess and Checkers were used by the earliest computer science
researchers to study reasoning processes. We will list the
defining features of these games.

To give a broader understanding of the history of the field, we will then
review some of the better known game playing programs, both old and
new.
The technical material with the algorithms that achieved all these
accomplishments starts in the next chapter.

\section*{Core Problems}
\begin{itemize}
\item What is intelligence?
\item How can we measure intelligence?
\end{itemize}

\section*{Core Concepts}
\begin{itemize}
\item Intelligence: embodied/computational; AI: symbolic/numerical 
\item Early work: Shannon, Turing, and Samuel
\item Modern work: TD-Gammon,
  Chinook, Deep Blue, AlphaGo, and Libratus
\end{itemize}  

\section{Intelligence}
What is intelligence? Psychology provides various definitions of
intelligence. Most definitions  include recognition,
reasoning,  memory,  learning, 
problem  solving, and adapting to your environment. These
elements are  related to cognition.

In
psychology most definitions implicitly assume that intelligence is
human, and therefore embodied. Related to intelligence are intuition,
emotion, self-awareness and volition, or {\em free will}. Humans have
all of these capacities. However, artificial
intelligence typically studies a  limited set of capacities.
Essential elements currently studied in AI are related to cognition,
such as recognition,
reasoning, memory, and learning~\cite{russell2016artificial}.
Intelligent decision making is at the core of current
AI research.

An important difference between human intelligence and artificial
intelligence is that humans are intelligent beings. We are, we have an
identity, and we are 
self-aware. Computers are not. Artificial intelligence research mostly studies
intelligent behavior. Machines that exhibit intelligent behavior are
generally not considered to {\em be\/} intelligent, just to {\em
  behave\/}
intelligently, even programs that jokingly suggest that they have an identity by printing texts such
as {\em Hello, World}. As Shakespeare wrote: ``To be or not to be, that is the question.''

We will now discuss  cognition and intelligence in more depth.

\subsection{Human Intelligence}
%
Psychology teaches us  that intelligence includes  cognitive
abilities, the abilities to learn, form
concepts, and reason. These abilities allow us to recognize patterns, comprehend
ideas,   solve problems, plan, and use language to
communicate. Cognitive abilities enable intelligent
behavior. They  also enable other capacities, such as conscious
thought. There are two schools of thought on cognitive intelligence.

\subsubsection*{Embodied and Computationalist Intelligence}
Experience and conscious thought are elements of 
embodied intelligence.
The theory of embodied intelligence stresses our ability to {\em
  perceive\/} information, 
to use it as knowledge in adapting behaviors to an environment~\cite{francisco1991embodied}.

A rival theory is the theory of 
computationalism. Computationalism states that
thoughts are a form of computation following a set of laws for the relations
among representations. Computationalism regards the brain as  a big
computer~\cite{mccorduck2009machines}. Clearly, artificial
intelligence, as part of computer science, studies  intelligence
through computations, and thus builds
on computationalism.

Interestingly, the field of reinforcement
learning has elements of both computationalism and
embodiment.  Reinforcement 
learning is all about learning in response to stimuli from an
environment, both in AI and in psychology.  It studies how agents act in an
environment and learn from responses to alter their behavior, as in
the embodied view of intelligence. 
Reinforcement learning is related to both schools, the school of
embodiment and the school of computationalism.

We will now have a look at two similar approaches to artificial
intelligence, a mathematical  approach and a biological approach.

\subsubsection*{Mathematics and
  Biology}\label{sec:schools}\index{symbolic AI}\index{connectionist AI}
Artificial intelligence is the 
study of machines that exhibit behavior for which humans need 
intelligence.  Main topics in the study of AI are cognitive functions
such as learning and problem 
solving~\cite{russell2016artificial}.

%
%
Artificial intelligence research is based on two fields of science:
mathematical logic and biology~\cite{nilsson2009quest}. One of the
early ways of modeling thought 
processes is the top-down deductive approach that has come to be known
as {\em symbolic AI}. Symbolic AI
is inspired by mathematical logic. In symbolic AI logic, {\em reasoning\/} is used to draw
inferences and conclusions from underlying facts and assumptions about
the world. In this view, intelligence is considered to be equivalent to a top-down reasoning
process. This school has yielded progress in expert systems, 
reasoning systems, and theorem proving systems. Well-known outcomes of
this line of research 
are the  STRIPS planner~\cite{fikes1971strips}, the Mathematica and
MATLAB computer algebra systems~\cite{buchberger1982computer}, 
the programming language PROLOG~\cite{clocksin2012programming}, and
also semantic (web)
reasoning~\cite{berners2001semantic,antoniou2004semantic}. The
material covered in the chapter on heuristic planning has grown out of the symbolic approach. 

A second school of thought is the {\em connectionist\/} approach to AI.
It purports that intelligence
emerges bottom up, out of interaction processes between many small
elements. Connectionism is inspired by
biology. Examples of the
successes of this approach are embodied intelligence
(robotics)~\cite{brooks1991intelligence}, Nature-inspired algorithms
such as Ant colony optimization~\cite{dorigo1997ant}, swarm
intelligence~\cite{kennedy2006swarm,bonabeau1999swarm}, genetic
algorithms and  evolutionary
strategies~\cite{fogel1994introduction,holland1992genetic,back1993overview}, and, last but not least, neural
networks~\cite{haykin1994neural,lecun2015deep}. This approach is also called the numerical, data driven, approach,
because of the many numerical parameters of artificial neural
networks. The material covered in the chapter  on learning has
grown out of the connectionist approach.

\subsubsection*{Thinking, Fast and Slow}
The philosophical implications of the AlphaGo achievement are
large. 
The realization
that we are now able to create a machine that can  learn to play such a
tremendously challenging and rich game of strategy, beating the
smartest professional players, is baffling. Indeed, for many researchers the
goal of artificial intelligence research is to understand
intelligence, and here we  have made a
tremendous 
step. It is now certainly the case, 
at least for the narrow domain of  combinatorial board games,
that we have succeeded in achieving learning and reasoning
behavior  beyond the  human level, although it is not known how similar the
artificial and human thought processes are. In fact, it has often been
argued that humans and computers think in very different ways.

Daniel Kahneman is a psychologist and winner of the Nobel Prize in
Economics. In his works he  combines
psychology and economics.  He has helped  found the field of behavioral
economics. 
In his book
{\em Thinking Fast and Slow}~\cite{kahneman2011thinking}  he presents  topics of his
most famous papers. Central in Kahneman's book are system 1 
thinking, or fast, approximate, intuitive human thinking, and system 2
thinking, or  slow, exact,
reasoned human thinking. These two concepts fit well with our AI
schools: learning/connectionist is related to system 1 thinking, and
planning/symbolic is related to system 2 thinking.

It is interesting to note that Kahneman's system  1 and 2 suggests at least a crude  similarity between
artificial and human thought processes.
At a few places we will highlight links between the  methods described
here and the 
concepts in Kahneman's book. 
%
Kahneman's book describes fast and slow thinking in human
intelligence. In this book we focus on  artificial intelligence
(see also Chap.~\ref{chap:fast}). 

\subsection{Artificial Intelligence}
Let us now have a closer look at the
main topic of this book, artificial intelligence.
Artificial intelligence is old. There are indications that the first
ideas on mechanical thought are
as old as human 
thought, and are related to materialistic worldviews. The first recorded dreams of intelligent machines, which we
now call robots, go back to Indian
philosophies 1500 BC of Charvaka~\cite{mccorduck2009machines,wujastyk2003roots}. In Greek the word
``automaton'' means {\em acting on one's own will}. In Greek mythology,
Homer's \emph{Iliad} describes automatons,  self-operating machines, that
open doors automatically. There are many more examples of automata in
Greek mythology,
such as an artificial person made of bronze, or watchdogs made of gold and
silver. In Chinese texts mechanical engineers, or artificers, are
described, as well as artificial wooden flying birds~\cite{needham1965science,koetsier2001prehistory}.

By attempting to describe human thought as the mechanical manipulation
of symbols, classical philosophers were the first to describe ideas of
modern AI. This work on mathematical reasoning culminated in
the invention of the  digital computer in the 1940s. The advent of the
first computers  and the ideas 
behind it prompted  scientists to start serious
discussions on the possibility of creating an electronic brain.


\begin{figure}[t]
\begin{center}
\includegraphics[width=\textwidth]{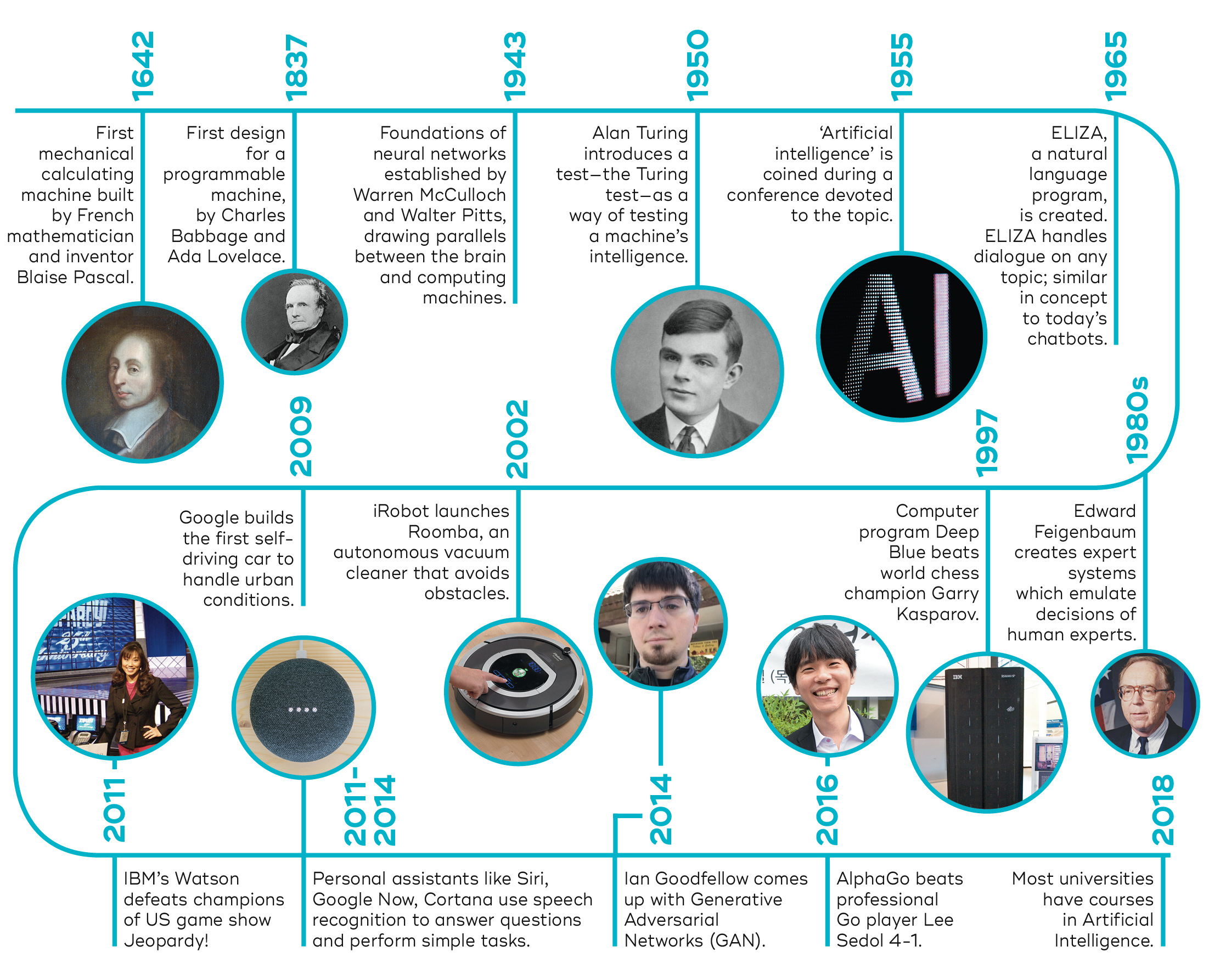}
\caption{A history of AI \cite{queensland2018brain}}\label{fig:dartmouth}
\end{center}
\end{figure}

\begin{figure}[t]
\begin{center}
\includegraphics[width=9cm]{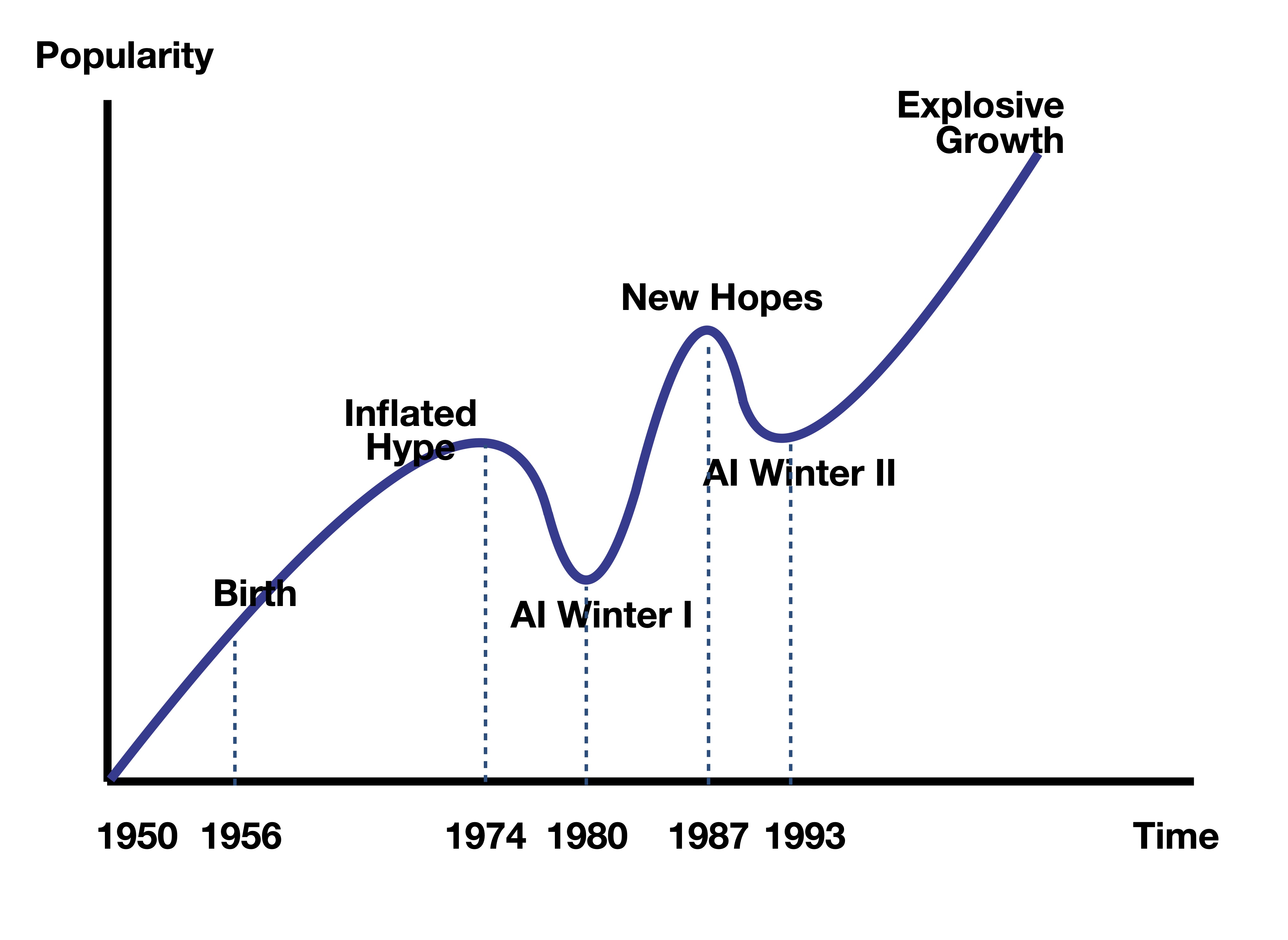}
\caption{AI winters~\cite{lim2018}}\label{fig:winter}
\end{center}
\end{figure}

The modern field of AI research was subsequently founded in the summer of 1956 at a
workshop at Dartmouth College, New Hampshire (Fig.~\ref{fig:dartmouth}). Among those present were John McCarthy, Marvin
Minsky, Nathaniel Rochester, Claude Shannon, Allen Newell, Herbert
Simon, Arthur Samuel, John Nash, and Warren McCulloch. These would all
become  leaders of AI research in the following decades.\footnote{Some would become known for other fields, such as information
  theory and economics.} There
was  much optimism then,  and many predicted that human-level  machine
intelligence would be achieved within 20
years (this was in 1956)~\cite{moor2006dartmouth,mccorduck2009machines}.  
Eventually, however, it became clear that they had  underestimated the
difficulty of this project. An AI winter followed in which funding for
AI research all but dried up (Fig.~\ref{fig:winter}~\cite{lim2018}). Despite
the funding drought, progress was made, and another period of inflated
expectations followed (about knowledge representation and expert
systems), followed by another winter. Achievements in the
1990s (Deep Blue, deep learning, Watson) heralded the current AI summer that we are still enjoying. After 2000 interest in AI has been high
again, because of 
machine learning successes in search, vision, speech, and robotics. In fact, many other
unexpected inventions did happen in industry, such as in search engines (Google
and Baidu), in social networks (Facebook and Tencent), in recommender
systems (Amazon, Alibaba, and Netflix), and in
multi-touchscreen communicators (Apple, Samsung, and Huawei).

\begin{figure}[t]
\begin{center}
\includegraphics[width=8cm]{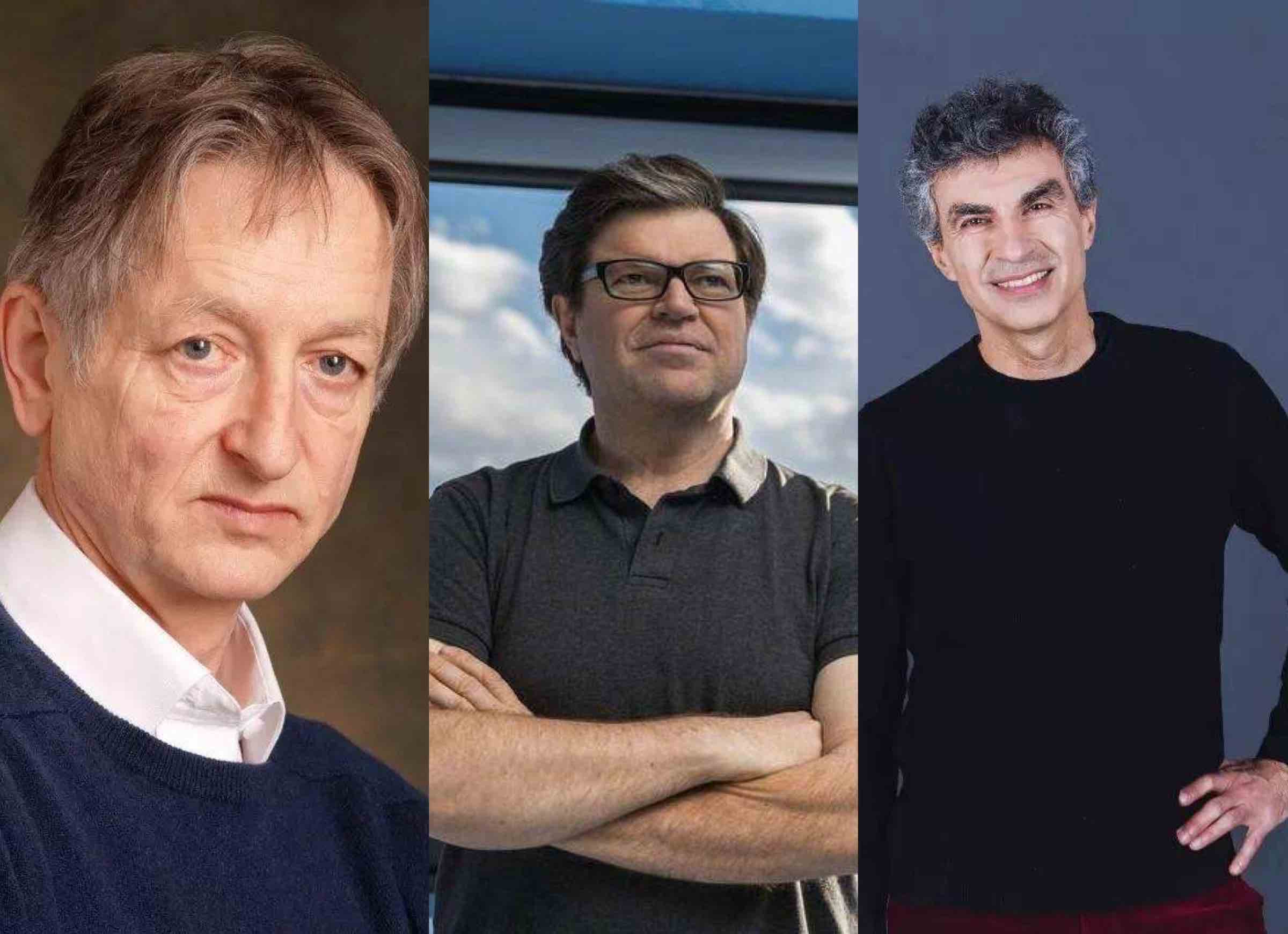}
\caption{Geoffrey Hinton,  Yann LeCun, and Yoshua Bengio}\label{fig:bhl}
\end{center}
\end{figure}

In 2018 the highest recognition in computer science, the Turing award,
was awarded to three key researchers in deep learning: Bengio, Hinton,
and LeCun (Fig.~\ref{fig:bhl}).\index{Yoshua Bengio}\index{Yann LeCun}\index{Geoffrey Hinton}

\subsection{Intelligent Behavior}

In contrast to some parts of psychology, AI is traditionally less interested
in defining the essence of 
intelligence (``being intelligent'') as it is in behaving intelligent, with a strong focus on
creating intelligent systems. The AI approach is a pragmatic approach.

AI researchers are interested in  machines that
exhibit  reasoning, knowledge, planning, learning,
communication, perception, and the ability to interact with
the physical environment and with humans.  The focus is on behavior; by which methods the machines
exhibit this behavior is considered less relevant. Consciousness and self-awareness
are not (yet) on the mainstream AI research agenda, and simulating the
human brain is not part of typical AI research.

Note that the strict separation between behavior and being is
becoming somewhat less strict. The focus on behavior of AI fits
  the  school of symbolic AI especially well, where intelligence is based on
  human logic, and not on human hardware. The other school,
  connectionism, is emphatically inspired by biology and 
  neurology, i.e., a simulation of human hardware to achieve
  intelligent behavior.

\begin{figure}[t]
\begin{center}
\includegraphics[width=7cm]{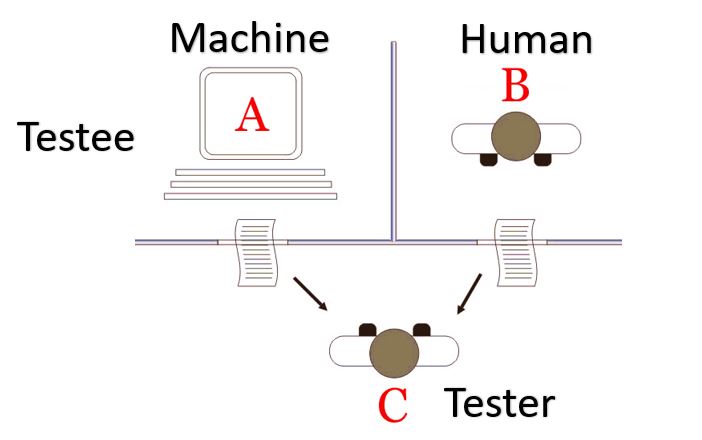}
\caption{Turing test~\cite{turing1950computing}}\label{fig:turingtest}
\end{center}
\end{figure}

\subsubsection*{Behavior: Turing Test}\index{Turing test}
In 1950 Alan Turing proposed a test for how to determine if a machine
exhibits  intelligence. The machine passes the test when its behavior
is indistinguishable from that of a human. In the Turing test a human
evaluator engages in a natural language conversation with a human
and with a machine (Fig.~\ref{fig:turingtest}). The evaluator
is aware that one of the two participants  is a
machine, and both participants are separated from one another by a
screen.
Participants are restricted to text only,  such as through
keyboard and screen (speech technology is deemed too hard).
If the evaluator cannot  tell which conversation is machine-generated
and which is human, then
the machine is considered to have passed the test. Note that the
machine may make mistakes as long as they are mistakes that humans
would also make. The Turing test is an imitation game. It might even
be the case 
that making deliberate mistakes is a good idea, for example, when the
domain would be a game of Chess, then a Chess machine could be
identified by its being much stronger than a human would be.

The paper that introduced the Turing test was titled: ``Computing
Machinery and Intelligence''~\cite{turing1950computing}.  It opens
with the words: ``I 
propose to consider the question, `Can machines
think?' '' and is a classic in AI.

The Turing test has been highly influential, but is also widely criticized.
One of the first criticisms was exemplified by the Eliza
experiment. Eliza was a computer program developed by
Joseph Weizenbaum in 1966~\cite{weizenbaum1966eliza}. The program had a list of keywords for which it
had responses in which it transformed the user's input in a certain
way. If no keyword was used, a
generic ``witty'' reply was given, sometimes reusing part the user's
input. Eliza was developed to appear like a psychotherapist, allowing
the program to have little knowledge of the real world.
The program was able to fool some people into
believing that they were talking to a real person. However, its simple
rule-based structure, and its use of superficial tricks, only
succeeded in suggesting 
that intelligent behavior can be created with rather unintelligent
means.\footnote{Some would go so far to say that some unintelligent
  humans behave in a more human way than  intelligent humans,
  confusing the discussion even more.} Eliza's success showed the limitations of the Turing test as a
mechanism for identifying true artificial intelligence.\footnote{The
  genealogy of Eliza and a list of programs that implement it can be
  found \href{http://elizagen.org/index.html}{here} \url{http://elizagen.org/index.html}.}

Another noteworthy approach to Turing's question became known as the Loebner Prize. The
Loebner Prize provided an annual platform for practical Turing
tests. It was named after Hugh Loebner, who provided the prize money.

The first Loebner Prize competition in 1991 was again won by a mindless
program  that misled  interrogators into making the
wrong identification by such means as imitating human typing
errors. It generated much controversy in which the term {\em artificial
  stupidity\/} was used ~\cite{shapiro1992turing}.  

A problem of the Turing test is that it does not directly test if the computer behaves
intelligently, but rather if humans think it behaves like another human
being. Turing tests have since taught us that human behavior and intelligent behavior are not
the same thing. The Turing test can fail to identify intelligent
behavior correctly because some human behavior is unintelligent, and
because some intelligent behavior is inhuman (such as performing
complicated mathematical sums quickly and correctly).
The Turing test  requires deception on the part of the computer;
it does not require highly intelligent behavior, in fact, it
discourages it, to prevent the computer from appearing to be inhumanly
smart.

For these reasons, other tests of intelligence have been devised, such
as tests for the ability to play games.

Before we take a look at those games, we will  look at what kinds of
artificial intelligence have been created.

\subsubsection*{Specialized and General Intelligence}
Humans are good at many different things. Our intelligence is a
general kind of intelligence; it allows us to cook, read, argue, do
mathematics, and 
enjoy a movie. Studying artificial ways of being intelligent in all
these domains is a daunting task. Machine intelligence researchers
typically limit themselves to a single domain, and
artificial intelligence is almost always narrow or  special
intelligence. Ever since the start of AI, games have played an important
role. The rules of a game limit its scope, making it  better
suited as a first object of study than the full complexity of reality.

Artificial
{\em general\/} intelligence (AGI) would be the intelligence of a machine
that can  perform any intellectual task that a human can.
AGI is the
holy grail of  AI research, and a favorite  topic in science fiction and future
studies. AGI is sometimes called {\em strong AI}, although that term
is used by some other researchers to mean machines that experience
consciousness~\cite{searle2008mind}; see also Sect.~\ref{sec:strong}.  
As of 2017, one study found that over forty organizations worldwide are doing active
research on AGI~\cite{baum2017survey}.

It
  should be noted, of course, that most humans also have special
  intelligence abilities. Some humans are better at playing Chess and
   weaker at Tennis, and very few humans are equally good in all
  kinds of intelligence (if such a thing could even be
  measured). In terms of definitions, AGI is even less well defined than AI.



\subsubsection*{\em Drosophila melanogaster}
In 1965, the Soviet mathematician Aleksandr Kronrod called Chess the
\emph{Drosophila melanogaster} of artificial
intelligence~\cite{landis2001aleksandr}. \emph{Drosophila melanogaster}  is a type of
fruit fly that is used by developmental biologists in genetics
experiments as a model organism; because of its fast reproduction cycle it allows quick
experiments. Also, by using \emph{Drosophila}, biologists hope to learn about
biological processes in general, beyond the particular
species. Similarly, by using Chess, AI researchers hope to learn
about processes of intelligence in general, beyond the particular game.
\index{\emph{Drosophila Melanogaster}}


Such an experimenter's  tool with quick
turnaround times is  useful in AI research. Chess was a convenient domain
that was well suited for such experimentation. It is clearly a game
for which humans need 
intelligence, requiring many years of  dedication and study to achieve
proficiency.

In addition to having a limited scope, a second advantage of games is the
clarity of the rules.  Possible
actions are well defined by the rules, and the end of a game is also
well defined. In this, games differ from  
real-world activities,  such as preparing a meal, performing a dance, or
falling in love, where goals and rules are often implicit and
less well defined. Combinatorial games are thought to abstract from
nonessential 
elements. Games allow researchers to focus on  reasoning
processes, processes that are assumed
to capture the essence of intelligence. Thus, in games it is 
easier to achieve good performance than in other real-world tasks, a
clear advantage for researchers~\cite{risi2020chess}.


\subsubsection*{Clear Performance}\label{sec:elo}
Games also allow for a clear and undisputed way to measure
progress. Game rules
define win and loss, and the intelligence of programs can be measured by
letting them play against each other or against humans. Many  games have
rating systems, such as in Tennis, where the best of the world are
ranked based on how many wins have been scored against other
players. An often used measure is the  Elo rating,
a system devised by the Hungarian-American physicist  Arpad Elo. 
The
basis of the Elo rating is pairwise
comparison~\cite{elo1978rating}.\index{Arpad Elo}\index{Elo} In Chess,
a beginner starts around 800, a mid-level player is around 1600, and a
professional, around 2400.

\subsubsection*{Anthropomorphization}
In  the original Turing test {\em any\/} question
can be asked of the machine. It is a  test for general
intelligence, the  kind of intelligence that humans have. As with the
\emph{Drosophila}, limiting 
the scope of our domain to a single game clearly cannot be used for
the general Turing test; all we can hope for is to achieve special 
intelligence, the kind of intelligence needed to win a certain
game. Still, achievements in specific intelligence, such as by Deep
Blue, Watson, and AlphaGo, attract large publicity. This may be in part
because the human mind generalizes easily, and
anthropomorphizes\footnote{Treat as if human.}
easily. When we see a device do something in a particular domain for
which a human must be incredibly smart, then we automatically assume that the device
must {\em be\/} smart, and will do  as well in another domain {\em
because that is what a human would do}. It is easy for  us to forget that the machine is
not a human, something that writers of science fiction books and
Hollywood movies eagerly exploit. Current machine intelligence is almost
always highly specialized, which is quite unhuman like. 

In
Chap.~\ref{chap:fast} the issues of artificial and human
intelligence will be revisited, including the topics of learning, and
specialized and general intelligence.  By then, we will have acquired a
deep understanding of how to create specialized artificial intelligence in games.



\subsection{Machine Learning}
Machine learning is the field of research that studies algorithms
that learn relations  that were not explicitly
preprogrammed, such as training an algorithm with a
 set of images in order to classify them. Machine learning encompasses both symbolic and connectionist
methods. Machine learning algorithms find patterns, categorize
data, or  make decisions.  Machine learning consists of three groups
of learning problems: 
supervised learning, reinforcement learning, and unsupervised
learning~\cite{bishop2006pattern}.

In supervised learning a (typically large) database of labeled
examples exists, from which the type of relation between example and label is
inferred. The labels have to be provided beforehand, by humans, or by 
other means. Chapter~\ref{chap:super} covers supervised learning in
detail.

In reinforcement learning there is no database of labeled examples;
interaction is the key. In reinforcement learning an agent can
interact with an environment, which returns a new state and a reward value
after each action. The reward value can be positive or negative. By probing the environment, the
reinforcement learning agent can learn relationships that hold in the
environment. Reinforcement learning problems are often modeled as
Markov decision processes~\cite{sutton2018introduction}; see Chap.~\ref{chap:rl}. Most games can be modeled as reinforcement
learning problems. In addition to Chap.~\ref{chap:rl},
Chap.~\ref{chap:dqn} also covers reinforcement learning in
detail.

Unsupervised learning algorithms find patterns in data without the
need for human labeling. Typical unsupervised algorithms cluster data
based on inherent properties, such as distance.

\subsubsection*{Reinforcement Learning in Games}
In this book we study reinforcement learning algorithms in
games. Reinforcement learning problems are interesting in that some
problems are best solved by symbolic methods, while for others
connectionist methods work best. In the chapter on self-play, we see how the two
methods in AI join to achieve great success.

Thus, the work in this book draws on both schools of
thought. Planning is closely related to symbolic AI, and learning to
connectionism. Reinforcement learning is related to both computationalism
and embodied intelligence.
%
The breakthroughs of Chess are related to symbolic
AI. Breakthroughs in Atari are related to connectionism. Breakthroughs in
AlphaGo (and other self-play programs) 
in Chap.~\ref{chap:selfplay} rely on both schools of thought,
and their success is a success of an integrated approach to AI,
expressed in a reinforcement learning framework. For other combined
approaches, see~\cite{valiant2008knowledge,garcez2015neural}.

After having looked at  artificial intelligence in some depth, it is
time to look at games.

\section{Games of Strategy}
Games come in many shapes and sizes. Some are easy, some are
hard. Before we will look at game playing programs, we first describe
the combinatorial games that are used frequently to study artificial
intelligence.  In 
the literature on games the characteristics of games are described
in a fairly standard taxonomy.

\subsection{Characteristics of Games}
Games are used in artificial intelligence because they provide a
challenging environment to play them well. The complexity of games is
determined by a number of characteristics.
Important characteristics of games are: the number of players,
whether the game is zero-sum or non-zero-sum,  perfect or imperfect information,
turn based or simultaneous action, what the
complexity of taking decisions is, and what the state space complexity
is. We will discuss
these characteristics now in more detail.

\begin{figure}[t]
\begin{center}
\includegraphics[width=4cm]{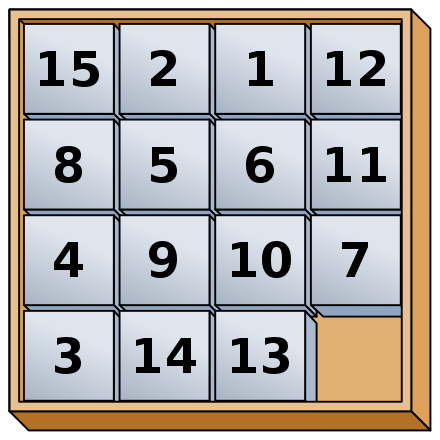}
\caption{15 Puzzle~\cite{hordern1986sliding}}\label{fig:15p}
\end{center}
\end{figure}
\index{15-puzzle}

\subsubsection*{Number of Players}
One of the most important elements of a game is the number of players.
One-player games are usually called puzzles. The goal of a puzzle is
to find a solution while following the rules of the puzzle. Well-known
examples are mazes, Sudoku, mathematical problems such as the traveling
salesperson problem (TSP), and the 15-puzzle~\cite{hordern1986sliding,korf1996finding} (Fig.~\ref{fig:15p}). Typical optimization problems are to find the shortest
(lowest cost) solution of a problem instance.

Two-player games are ``real'' games. Most games in this book
are two-player games; these have been studied the most in AI. The
game play in two-player games is often competitive, zero-sum: my win
is your loss. Keen
reasoning and  calculation often play a large role in these games,
providing satisfaction to the players when a hard problem is
solved. Quite a number of two-player 
games exist that provide a nice balance between being too easy and
too hard for players, and for
computer programmers. Many  multiplayer
games  are still highly challenging for AI.
Examples of two-player games that are popular in AI are Chess,
Checkers, Go, Othello, and Shogi. These games will all be described in
a little more detail shortly.

Multiplayer games are played by three or more players. Psychology,
collaboration, and hidden information (and bluffing) often play a large role in these
games. Much of the fun of playing these games may be related to social
aspects. The game play
can be competitive and collaborative, since players may collude in
different phases to obtain their goals.  Many card games are
multiplayer games. Well-known examples of
multiplayer games are the card games Bridge and Poker, and
strategy games such as Risk, Diplomacy, StarCraft, and Defense of the Ancients (DOTA).

\subsubsection*{Zero-Sum vs.\ Non-Zero-Sum}
An important aspect of a game is whether it is competitive or
collaborative. Many two-player games are competitive: the win ($+1$) of
player A is the loss ($-1$) of player B. These games are called {\em
  zero-sum\/} because the sum of the wins for the players remains a
constant zero. Competition is an important element in the real world, and these games provide a useful model for the
study of strategy and conflict, in practice, and especially in theory.  The field of game theory has
important links to economics and political science. The classic work
is Von Neumann and Morgenstern's 1944 book \emph{Theory of games and
economic behavior}~\cite{von1944theory,von2007theory}, which laid the
foundation for economic game theory. Many more modern and
accessible works have been published since then, such
as~\cite{davis2012game}. Classical game theory analyzes strictly
rational behavior; the classical {\em homo economicus\/} is assumed to
always act rationally. Many years later behavioral economics married
psychological insight with economics. Amongst others, Kahneman and Thaler have
written popular and accessible
accounts~\cite{kahneman2011thinking,thaler2008nudge}.

In contrast, in collaborative games the players win if they can create win/win
situations. Again, the real world is full of opportunities for
collaboration, and these games can be studied to hone our
understanding of
collaboration intelligence. Examples of collaboration games are Bridge,
negotiation   games such as Diplomacy~\cite{kraus1994negotiation,de2018challenge}, management simulations such
as Hexagon~\cite{duke2004policy}, or board games such as Magic
Maze. Collaboration also occurs frequently in  multiplayer games such as
Poker and Risk.

\subsubsection*{Perfect vs.\ Imperfect Information}
In perfect information games all relevant information is known to all
players. This is the case in typical board games such as Chess and
Checkers. In imperfect information games some information may be hidden for
some players. This is the case in typical card games such as Bridge
and Poker, where not all cards are known to all players.

A special form of (im)perfect information games are games of chance, such as
Backgammon and Monopoly, in which dice play an important role.
There is no hidden information in these games, and these games are
therefore sometimes considered to be perfect information games,
despite the uncertainty present at move time.

\subsubsection*{Turn-Based or Simultaneous-Action}
In turn-based games players move in sequence. In simultaneous-action
games, players make their moves at the same time. Simultaneous-action
games are considered imperfect information, since  players
hold secret information that is relevant for a player at move time.

Chess, Checkers, and Monopoly are examples of turn-based games. StarCraft and
Diplomacy are simultaneous-action games.

\subsubsection*{Decision Complexity}
The difficulty of playing a game depends on the complexity of the
game. The decision complexity is the number of end positions that define the
value (win, draw, or loss) of the initial game position (see
Chap.~\ref{chap:search} for a theoretical explanation of {\em critical tree}). The larger the
number of actions in a position, the larger the decision
complexity. Games with small board sizes such as Tic-Tac-Toe ($3\times
3$) have a smaller complexity than games with larger boards, such as
Gomoku ($19\times 19$). When the action space is very large it can be
treated as a continuous action space. In Poker, for example, the
monetary bets can be of any size, defining an action size that is
practically continuous.

\subsubsection*{State Space Complexity}
The state space complexity of a game is the number of legal positions
reachable from the initial position of a game. State space and
decision complexity are normally positively correlated, since games
with high decision complexity typically have high state space complexity. As we shall see
in Chap.~\ref{chap:search}, determining the exact state space
complexity of a game is a nontrivial task. For most games
approximations of the state space have been calculated.

In general, games with a larger state space complexity are harder to
play (``require more intelligence'') for humans and computers.

\subsubsection*{Zero-Sum Perfect-Information Games}\index{zero-sum game}
Now that we have seen several important aspects of games, let us look
at concrete examples of zero-sum perfect-information games.

Chess and Go are two player, zero sum, perfect information, turn
based, discreet action 
games.

To study strategic reasoning in AI,
two-player zero-sum perfect-information games are frequently used.
Strategies,
or policies, determine the outcome of these kinds of games.  This was
the first class of games tried by AI researchers.

\subsection{Examples of Games}
We will now give examples of games that have played an important
role in artificial intelligence research.
Table~\ref{tab:games} summarizes the games and their characteristics.

\begin{table}[t]
  \begin{center}
\begin{tabular}{lccccc}
  {\bf Name}&{\bf board}&{\bf state space}&{\bf zero-sum}&{\bf
                                                           information}&{\bf
                                                                         turn}\\
  \hline\hline
  Chess&$8\times 8$&$10^{47}$&zero-sum&perfect&turn\\ 
  Checkers&$8\times 8$&$10^{18}$&zero-sum&perfect&turn\\ 
  Othello&$8\times 8$&$10^{28}$&zero-sum&perfect&turn\\ 
  Backgammon&24 &$10^{20}$&zero-sum&chance&turn\\ 
  Go&$19\times 19$&$10^{170}$&zero-sum&perfect&turn\\ 
  Shogi&$9\times 9$&$10^{71}$&zero-sum&perfect&turn\\ 
  Poker&card&$10^{161}$&non-zero&imperfect&turn\\ 
  StarCraft&real-time strategy&$10^{1685}$&non-zero&imperfect&simultaneous\\ \hline
\end{tabular}
  
  \caption{Characteristics of games}\label{tab:games}
  \end{center}
\end{table}

\begin{figure}[t]
  \begin{center}
\includegraphics[width=8cm]{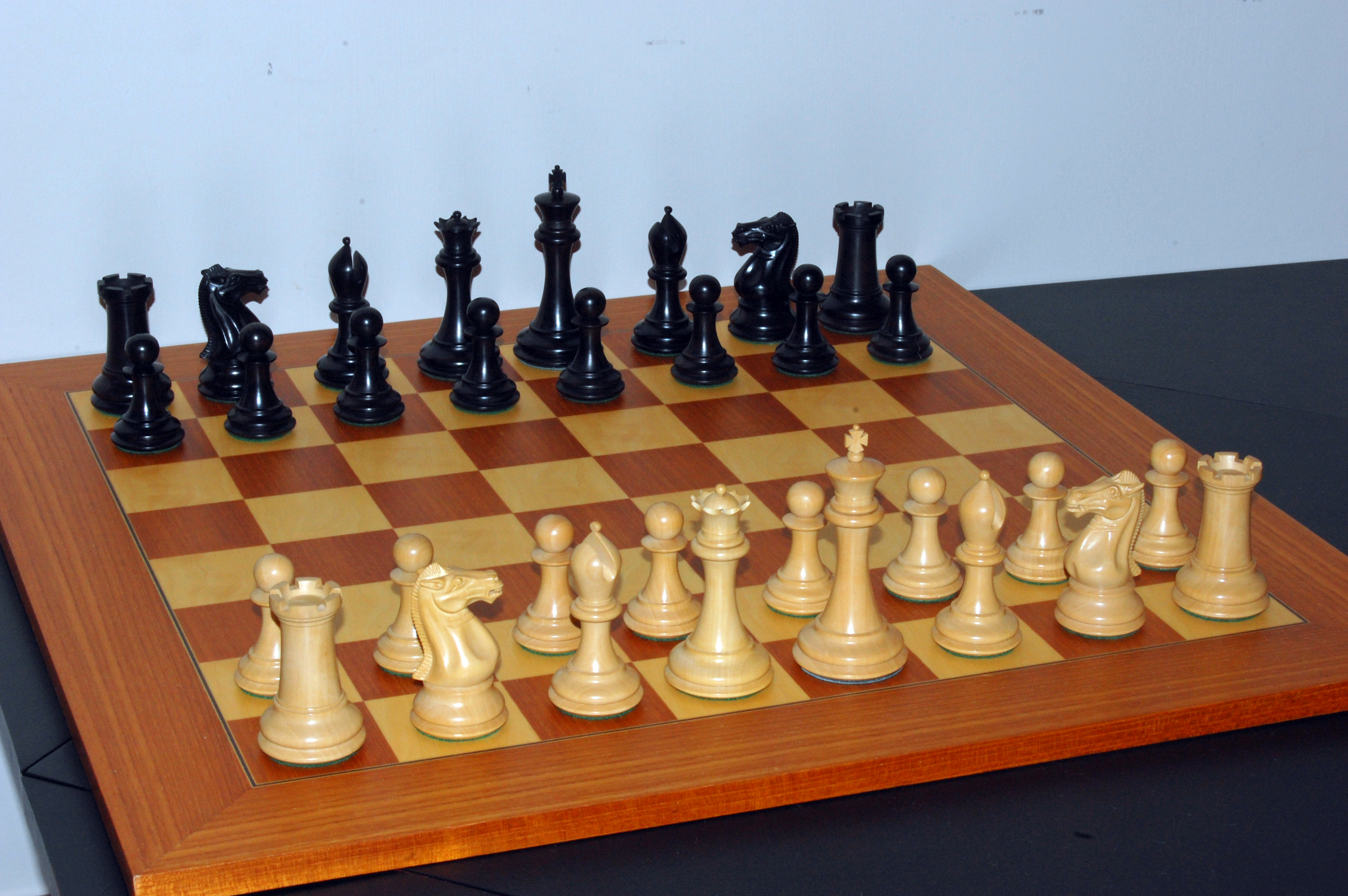}
  \caption{Chess}\label{fig:chess}
  \end{center}
\end{figure}

\subsubsection*{Chess}
Chess (Fig.~\ref{fig:chess}) is a two-player zero-sum perfect-information turn-based game, played on an
$8\times 8$ board with 64 squares and 32 pieces, many with different move rules. The pieces move and can be
captured. The goal of Chess is to capture the king. It has a 
state space complexity estimated to be $10^{47}$. The number of
actions in a typical board state is around 35. It is a game where
material balance (the number and importance of the pieces that 
the sides have) is quite important for the outcome of the
game. The ability to reason over deep tactical lines in order to
capture a piece is important. Chess has been used as a testbed  of AI
research  since the early days of AI.  

Computer players are typically based on heuristic planning and consist
of a search function (that looks ahead) and an evaluation
function (that scores the board position using heuristics).  A
heuristic is a rule of thumb that captures domain knowledge. In Chess a
useful heuristic is to count material balance (which side has more
pieces) and mobility (which side has more legal moves) and center
control.

\index{tactical game}
Chess is a tactical game in the sense that substantial changes in the
static score of a position are present in the state space. Such sudden changes occur for
example when an important piece, such as the queen, is captured. Chess
also has ``sudden death:'' the game can end abruptly when the
king is captured (checkmate). Tactical situations (i.e., large changes between
static evaluation between positions) can typically be dealt with by
searching, or looking ahead.

\begin{figure}[t]
  \begin{center}
\includegraphics[width=6cm]{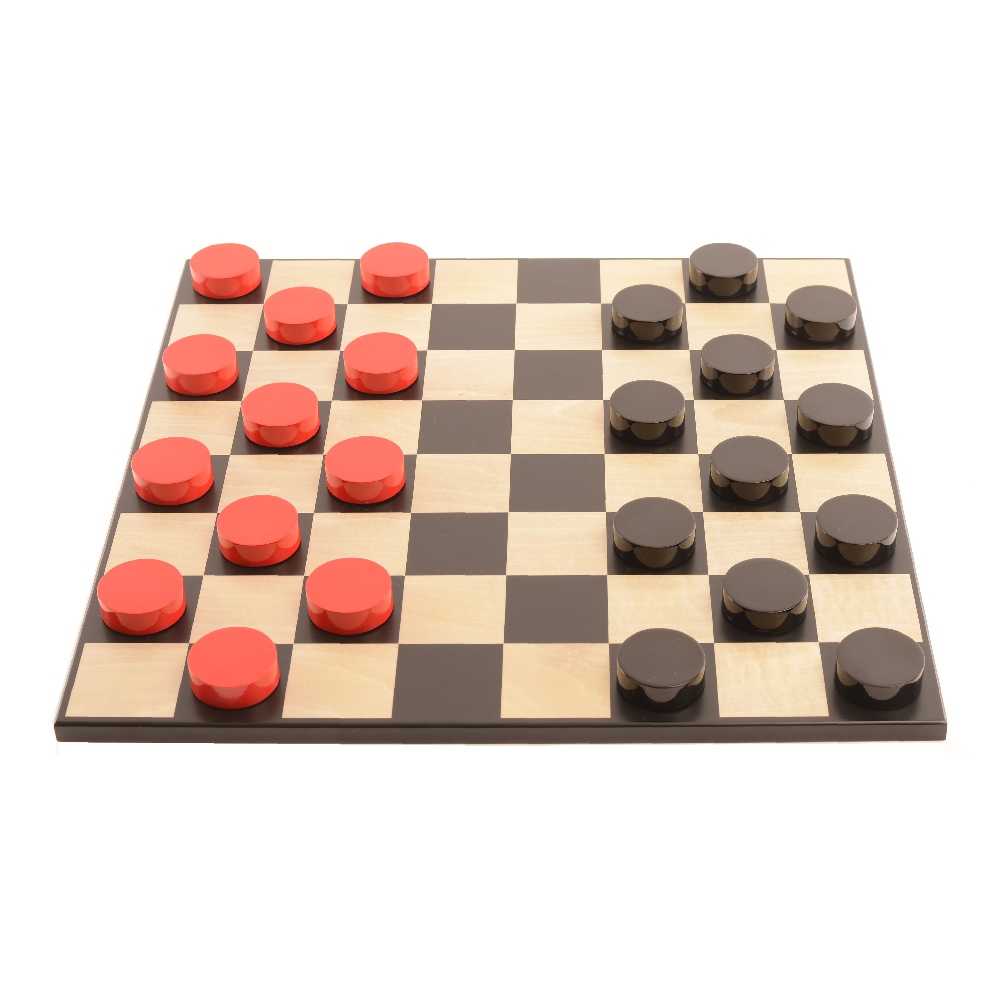}
  \caption{Checkers}\label{fig:checkers}
  \end{center}
\end{figure}

\subsubsection*{Checkers}\label{sec:checkers}
Checkers (Fig.~\ref{fig:checkers}) is also a zero-sum
perfect-information turn-based game, played on an  
$8\times 8$ board with 24 pieces. (North American and British Checkers
is played on an $8\times 8$ board. In most other countries the board
size is $10\times 10$. In British English the game is called Draughts.) The pieces move diagonally (only on
the dark squares) and can be
captured. All pieces move in the same fashion. The number of
actions in a typical board position is around 3. The goal of Checkers is to capture all the opponent's
pieces. Ordinary pieces move only forward, until they reach the other
side of the board, where they are promoted to kings, which can move
backward as well. 
As in Chess, material balance
is a good indicator (heuristic) for the chance to win, with mobility a
good second indicator.

 The state space of
Checkers is estimated to be $10^{18}$. This state space is just small
enough for the game to have been mathematically solved in recent years: it has been proven by
traversing all relevant lines of play that perfect play by both sides yields a
draw~\cite{schaeffer2007checkers}.  This computation took months of
computer time by a state-of-the-art research team. 

Computer players for Checkers and Chess have a highly related
design. Many of the techniques for search and evaluation functions
that work on Checkers also work in Chess.

\begin{figure}[t]
\begin{center}
\includegraphics[width=6cm]{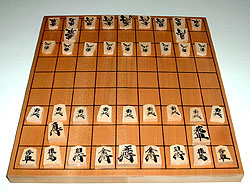}
\caption{Shogi}\label{fig:shogi}
\end{center}
\end{figure}

\subsubsection*{Shogi}
Shogi (Fig.~\ref{fig:shogi}) is a zero-sum perfect-information
turn-based game, sometimes described as Japanese Chess. Shogi is
played on a $9\times 9$ board, with 40 pieces. Pieces can be captured,
and when they are, they may be returned to the board as piece for the
capturing side. The number of
actions in a typical board position is around 92. The state space complexity of Shogi is significantly larger than
Chess, $10^{71}$, because of the reintroduction of pieces, and the larger board
size, among others.

Important heuristics for Shogi are material balance and
mobility. Techniques that work in Chess and Checkers playing programs
often work in Shogi as well.

\begin{figure}[t]
\begin{center}
\includegraphics[width=6cm]{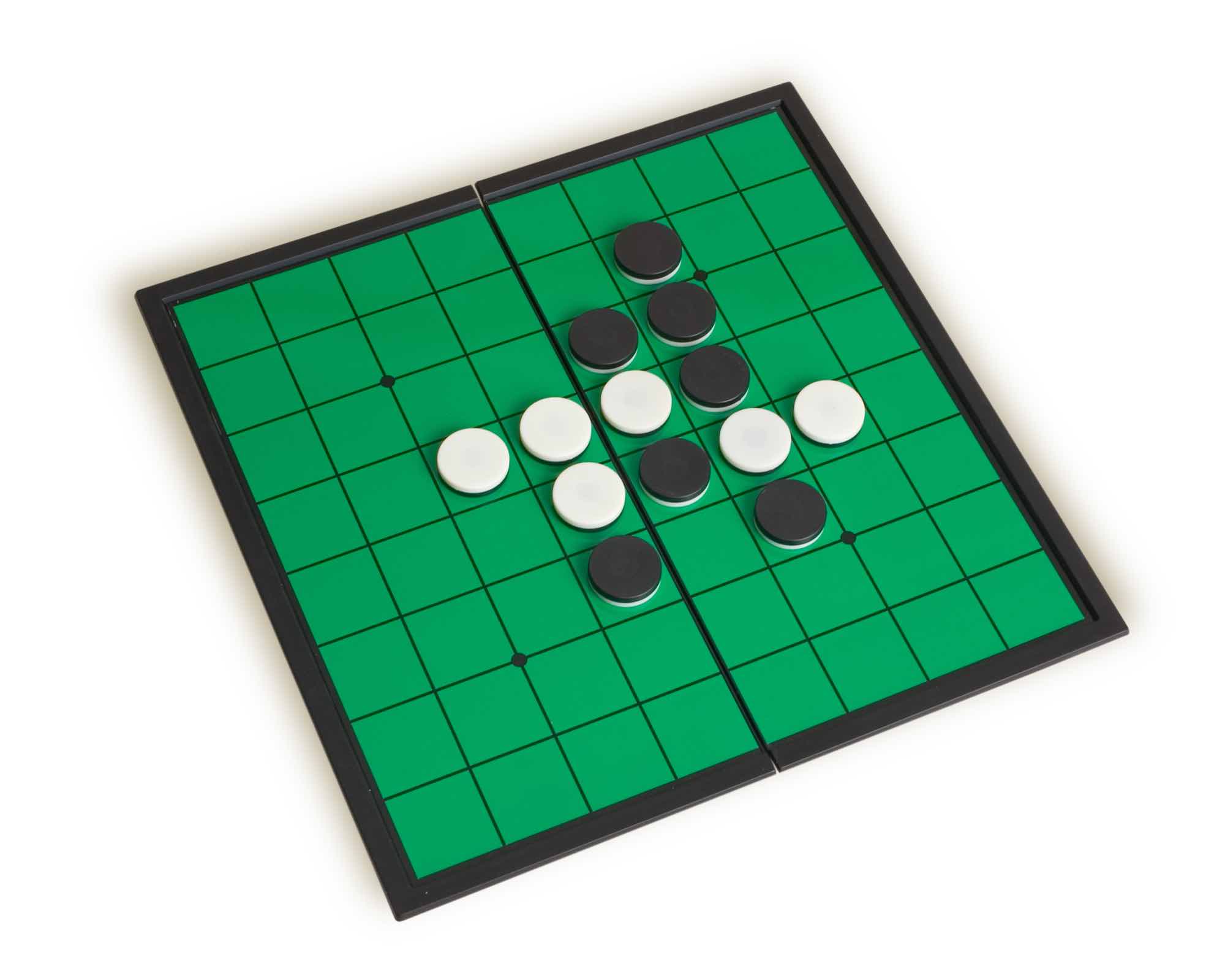}
\caption{Othello}\label{fig:othello}
\end{center}
\end{figure}

\subsubsection*{Othello}
Othello (Fig.~\ref{fig:othello}) is a zero-sum perfect-information
turn-based game, played on an  
$8\times 8$ board. Othello  is also known as Reversi. Othello is a
disc flipping game. The number of
actions in a typical board position is around 10. When a disc is placed adjacent to an opponent's
disc, and there is a straight line to another friendly disc, the
opponent's discs change color to the friendly side.  In each move one
disc is placed. Discs are never taken off the board. When the board is
full the game stops, and the side with the most discs wins.

The state space complexity of Othello is estimated to be $10^{28}$. Traditional
heuristics such as material and mobility work to a limited
extent. More advanced statistical and machine learning methods were
needed for breakthrough performance~\cite{buro1995logistello} (Sect.~\ref{sec:logistello}).

\begin{figure}[t]
\begin{center}
\includegraphics[width=8cm]{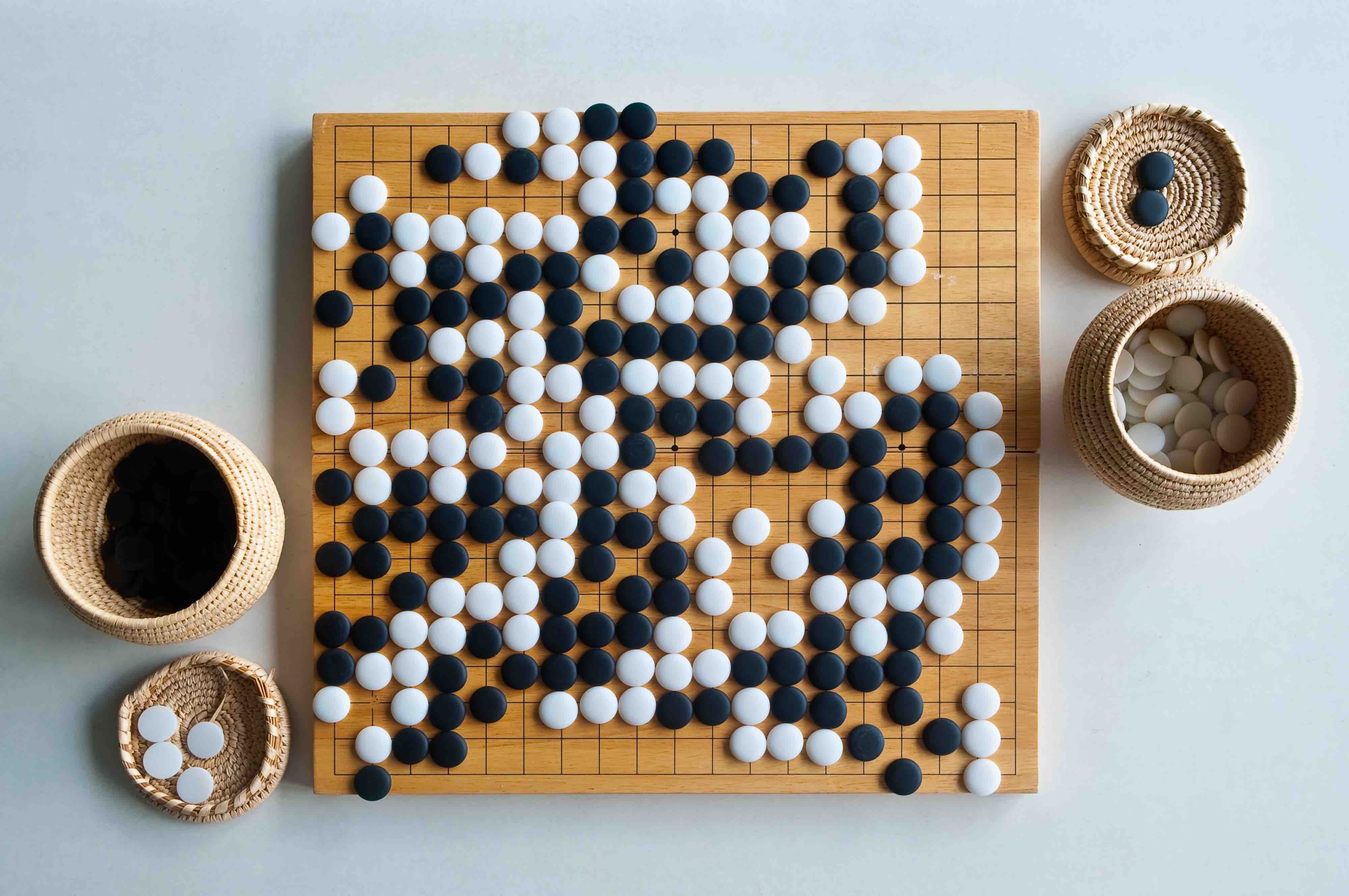}
\caption{Go}\label{fig:go}
\end{center}
\end{figure}

\subsubsection*{Go}
Go (Fig.~\ref{fig:go})  is a zero-sum perfect-information turn-based game, played on a
$19\times 19$ board. It is also played on smaller boards, for quicker
and easier games for teaching, $9\times 9$ and $13\times 13$. Go
originated in East Asia, and is popular in Korea, Japan, and
China. In Go, stones are placed on intersections on the board. The number of
actions in a typical board position is around 250. Surrounded
stones are captured, although this occurs infrequently among strong
players.  The board is large, and the
state space complexity
of Go is also large: $10^{170}$. 

In Go, the objective in the game is to acquire territory, intersections
that the opponent cannot claim. No effective  heuristics
have been devised for Go to calculate this territory
efficiently for use in a real-time game playing program. Approaches as in Chess and Checkers failed to produce
anything but amateur-level play in Go. Sampling-based
methods (Monte Carlo Tree Search) and deep reinforcement learning
caused  a breakthrough (Chaps.~\ref{chap:trial} and \ref{chap:plan}).

\index{strategic game}
Capturing stones occurs less frequently in Go than in Chess. Go is a 
more strategic game: stones, once placed, do not move, and typically radiate
influence all the way to the end of the game. Strategically placing 
stones to later work together is an important 
concept in Go. Sudden death (checkmate in Chess) does not exist in
Go. Strategic implications of a move are long term, and are typically
beyond the horizon of a search or look ahead. Long-term effects
are typically dealt with by the evaluation function.

\begin{figure}[t]
\begin{center}
\includegraphics[width=6cm]{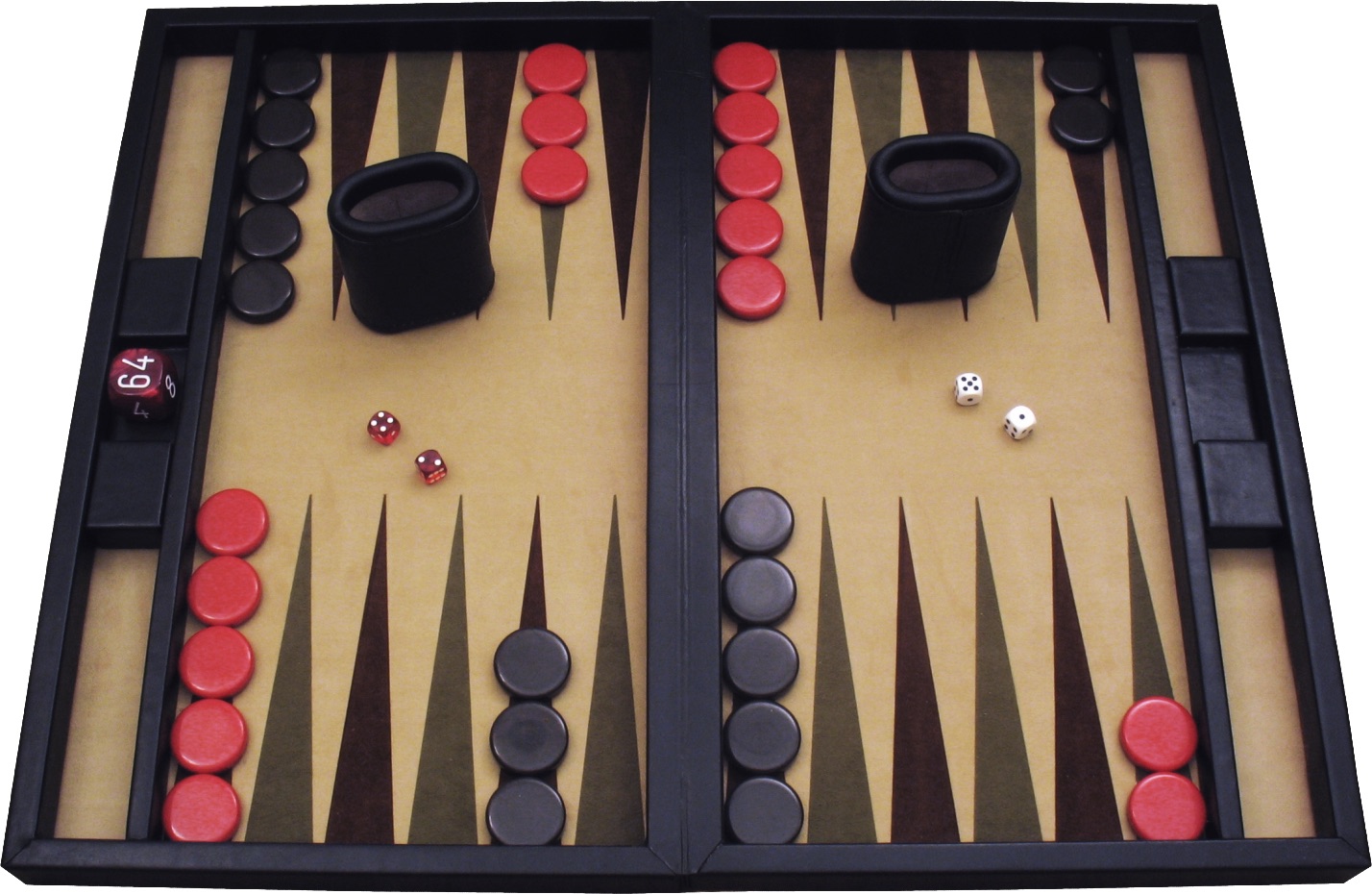}
\caption{Backgammon}\label{fig:bg}
\end{center}
\end{figure}

\subsubsection*{Backgammon}\index{Backgammon}
Backgammon (Fig.~\ref{fig:bg})  is a zero-sum turn-based game of
chance. Backgammon is one of the oldest and most popular board
games, often played for money. It is played on a board as in the figure. Dice are used to
determine the number of moves that a player can make. Each side has 15
pieces. The objective is to be the first to move all pieces off the
board. The number of
actions in a typical board position is around 250.

Backgammon is a game of skill and luck. The state space complexity of
Backgammon is estimated at $10^{20}$. Computer player 
BKG achieved some success in 1970--1980 with a heuristic planning
approach~\cite{berliner1977experiences,berliner1980backgammon}. Later
TD-Gammon used a small neural network for temporal difference
learning, and achieved world champion level 
play~\cite{tesauro1995td} (see the next section).

\begin{figure}[t]
\begin{center}
\includegraphics[width=10cm]{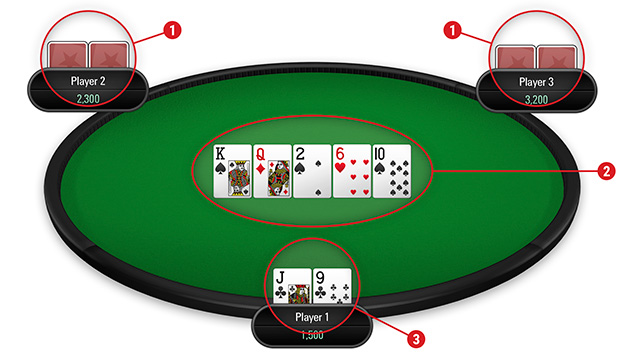}
\caption{Poker}\label{fig:poker}
\end{center}
\end{figure}

\subsubsection*{Poker}
Poker (Fig.~\ref{fig:poker})  is a non-zero-sum imperfect-information turn-based card game of chance. Betting is an integral
part of the game play. It is used to signal the quality of a hand of
cards to the other players (or to mislead them). Poker is a multiplayer game. The action space
is large due to the betting, and varies greatly for different versions
of the game.  The difficulty of
play is in estimating the quality of a hand with respect to others,
and has psychological aspects, to ascertain if players are bluffing:
placing high bets on weak hands, trying to scare other players.

The state space of No-Limit Texas Hold'em, a popular Poker variant, has been
estimated between $10^{18}$~\cite{billings2003approximating} and
$10^{161}$~\cite{brown2018superhuman,johanson2013measuring}. There
are many Poker variants, and this number can  differ from variant to variant.

Poker is an imperfect-information game (the
hidden hands 1 in the figure). 
Because of this it is closer to some decision
making situations that are common in the real world, than games such as
Chess and Checkers.

Most computer Poker players are based on a detailed analysis of the
quality of the hands using a methods called counterfactual regret
analysis~\cite{zinkevich2008regret,brown2019superhuman}. 

\begin{figure}[t]
\begin{center}
\includegraphics[width=10cm]{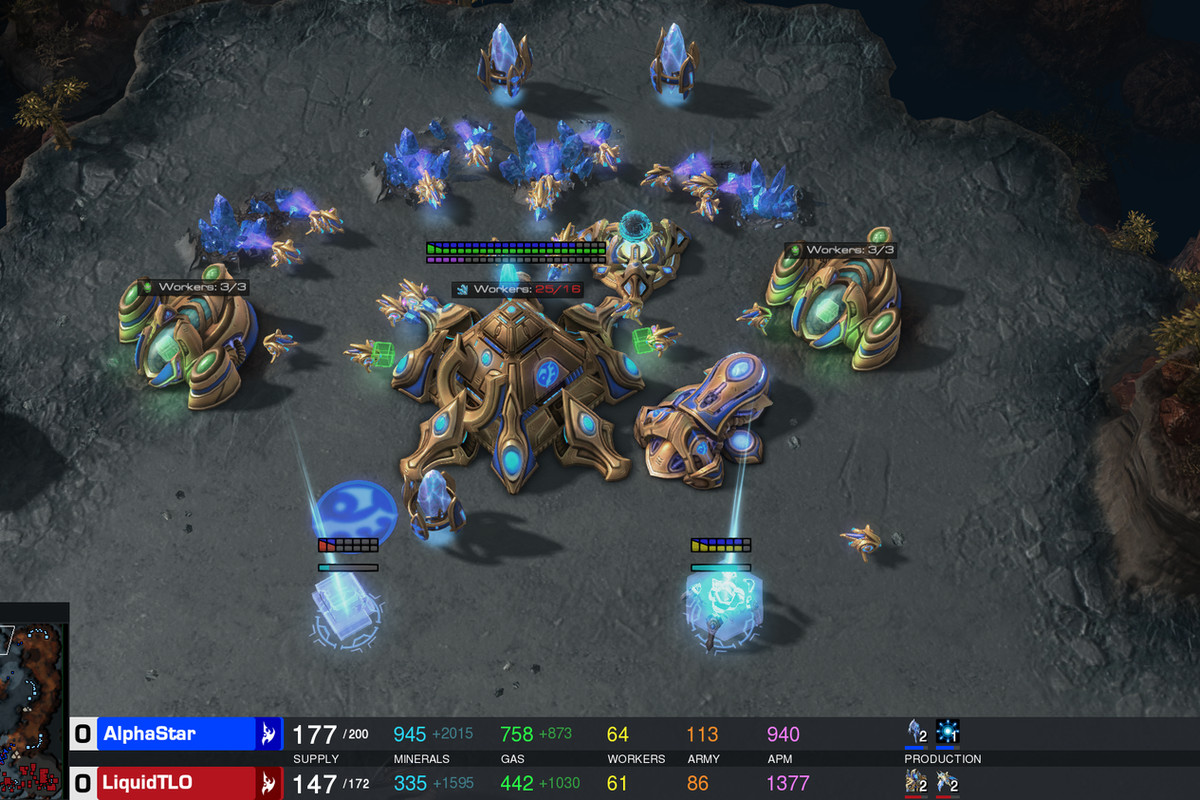}
\caption{StarCraft~\cite{vinyals2019grandmaster}}\label{fig:starcraft}
\end{center}
\end{figure}\index{StarCraft}

\subsubsection*{StarCraft}
StarCraft (Fig.~\ref{fig:starcraft})  is a non-zero-sum imperfect-information simultaneous action real-time strategy game. It is a
multiplayer game. The goal of StarCraft is to gather resources,
create buildings, develop new technologies, and train attack units,
which then must prepare for and fight battles.
StarCraft combines imperfect
information with multiplayer aspects,  long-term strategy, and short-term
tactics. Armies have to be built up (strategy), and battles have to be
played (tactics). The imperfect-information aspect is that the game map is only
partially visible, and the actions of the other players are not all
known. 

It is a highly popular,
highly complex game. The number of actions in  StarCraft positions has
been estimated to be between $10^{50}$ and
$10^{200}$~\cite{ontanon2013survey}. The state space complexity of
StarCraft has been conservatively estimated by these authors at
$10^{1685}$, which is a very, very large number.

Of all the games that are used in AI, the complexity of the actions
in StarCraft is one of the highest, and the decision making comes closer to real-world decision making than in other games.  As AI methods are
improving, the interest in  StarCraft is growing. 

\subsection{Early  Game Playing Programs}
%
Now that we have studied the characteristics of some combinatorial games, it is time
to have a look at the research field of computer game play.

In order to get a good
understanding of the field, let us start with a short overview  of some
of the better known game playing programs in the history of AI. Even before computers were powerful enough to run  game playing
programs, there were paper designs of game playing programs.  First we discuss two
hypothetical  game players, and
then we continue with the first real game playing programs.

As an aside, please note that there is a common approach in all these attempts. A
search function does  look ahead, and  a 
heuristic evaluation function  scores the board positions. Together the
two functions form the search-eval architecture, which we will study
in detail in Chap.~\ref{chap:search}.\index{search-eval architecture}


\subsubsection*{Claude Shannon 1949}
Claude  Shannon (1916--2001) 
was an
American mathematician, 
electrical engineer, and cryptographer. Shannon is primarily known as
the father of information theory. 

In 1949 Shannon presented a paper titled ``Programming a Computer for
playing Chess.'' The paper describes a design of how to program a
computer to play Chess based 
on position evaluation and move selection. The paper describes  strategies
for restricting the number of possibilities to be
considered. Shannon's paper is  one of the first articles
published on the topic~\cite{shannon1988programming}.

\begin{figure}[t]
\begin{center}
\includegraphics[width=11cm]{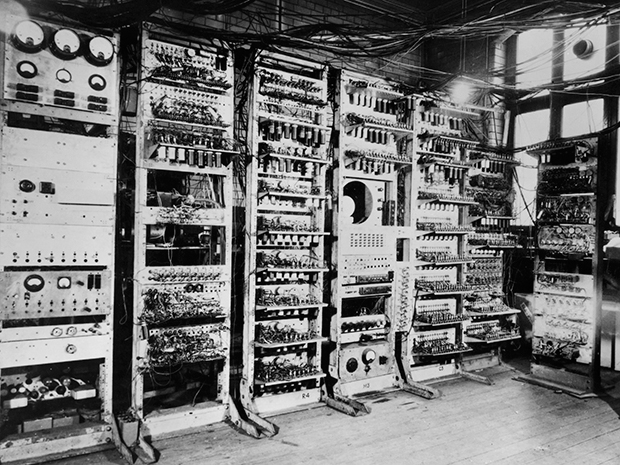}
\caption{Ferranti Mark 1}\label{fig:ferranti}
\end{center}
\end{figure}
\index{Ferranti Mark 1 computer}

\begin{figure}[t]
\begin{center}
\includegraphics[width=4cm]{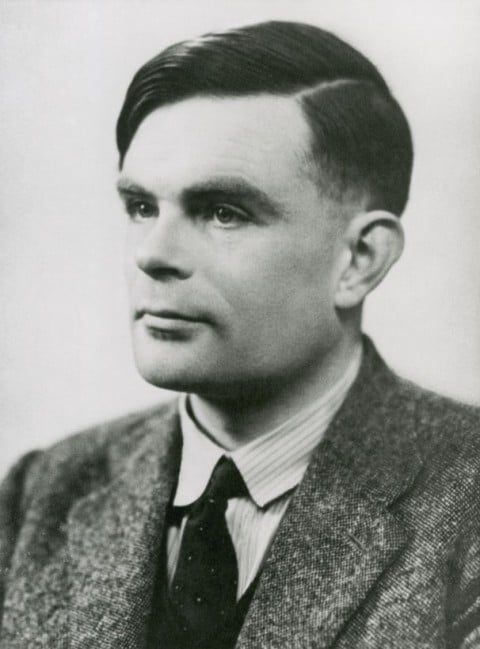}
\caption{Alan Turing}\label{fig:turing}
\end{center}
\end{figure}
\index{Alan Turing}

\subsubsection*{Alan Turing 1950}
In 1948 Alan Turing (1912--1954,  Fig.~\ref{fig:turing}) worked with David Champernowne on  a Chess
program for a hypothetical computer. They named their
program Turbochamp. In 1950, the program was finished. In 1952, Turing
tried to implement it on a Ferranti Mark 1,  Fig.~\ref{fig:ferranti}, but the computer was
unable to execute the program, so he ended up
executing the program by hand, flipping through the pages of the
algorithm and carrying out its instructions on a 
chessboard. Reportedly, this took about half an hour per move.
The program is the first to have played a full Chess game, even though it was
executed on paper and by hand~\cite{turing1953digital}. Turbochamp had an
evaluation function and a search function, the design that is still used
in today's programs.


\subsubsection*{Christopher Strachey 1951}
Christopher Strachey (1916--1975) 
was a preeminent computer scientist
at Oxford, well known
for his work on denotational semantics and programming language
design. In 1951, Christopher Strachey wrote a Checkers program for the
Ferranti Mark 1 computer of the University of 
Manchester. It could play a complete game of Draughts (Checkers) at
reasonable speed~\cite{strachey1952logical}.

\begin{figure}[t]
\begin{center}
\includegraphics[width=8cm]{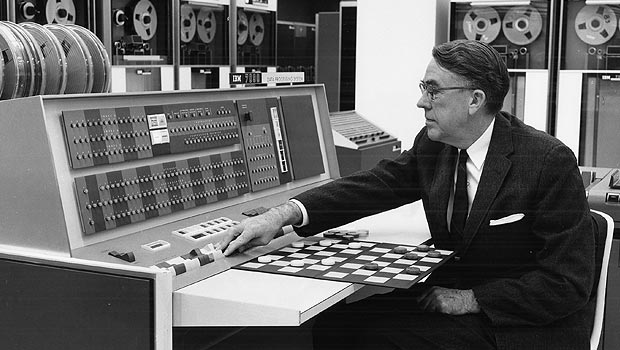}
\caption{Arthur Samuel}\label{fig:samuel}
\end{center}
\end{figure}
\index{Arthur Samuel}

\subsubsection*{Arthur Samuel 1959}
Arthur Samuel (1901--1990,  Fig.~\ref{fig:samuel}) started in 1949 to write a Checkers
 program 
for the IBM 701 computer. It was among the world's first successful
self-learning programs, adapting coefficients in the evaluation function
based on the outcome of the search~\cite{samuel1959some}. 

The main driver of the program was a
search tree of the board positions reachable from the current
state. To reduce memory consumption Samuel implemented a version of what is now called alpha-beta pruning.
He used a heuristic evaluation function based on the position on the
board to prevent having to search to the end of the game. He also used
a method to memorize search results, and later versions used
supervised learning from professional games and from self-play. Around
the 1970s the program had progressed to amateur level~\cite{schaeffer2008one}.


\section{Game Playing Programs}
The playing strength of game playing programs has increased
steadily since the early days. Computer hardware has become much more powerful, and better
algorithms have been found. Most of these algorithms are described in
this book.

For the pioneers the challenge was to create a program that could
successfully play
a game to the end. We will now skip a few years.
The level of play has increased greatly, and now the
challenge has become to beat human champions. We will pick up around
1990, where some programs started  playing at world champion level.

\subsection{TD-Gammon 1992}\index{TD-Gammon}
At the end of the 1980s Gerald Tesauro
worked on Backgammon
programs. (Figure~\ref{fig:bg} showed a Backgammon board.)

His programs were based on neural networks that learned good patterns of
play. His first program, Neurogammon, was trained using supervised
learning. It achieved an intermediate level of
play~\cite{tesauro1989neurogammon}. His second program,
TD-Gammon, was based on temporal difference learning and self-play, a form of
reinforcement learning (see Chaps.~\ref{chap:rl} and \ref{chap:dqn}). Combined with
hand-crafted heuristics, in 1992 it played at world-class human
championship 
level, becoming the first computer program to do so in a game of
skill~\cite{tesauro2002programming}. 

TD-Gammon is named after temporal difference because it updates  its
neural net after each move, reducing the difference between the
evaluation of previous and current positions. 
%
In Tesauro's previous program  Neurogammon an expert trained the program by
supplying the ``correct'' evaluation of each position. The neural
network used a single hidden layer with 80 units. In contrast,
TD-Gammon initially learned knowledge-free, or {\em tabula rasa}.
Tesauro describes TD-Gammon's tabula rasa self-play as follows: ``The move that is selected is  the move with maximum expected
outcome for the side making the move. In other words, the neural
network is learning from the results of playing against itself. This
self-play training paradigm is used even at the start of learning,
when the network's weights are random, and hence its initial strategy
is a random strategy''~\cite{tesauro1995temporal}.

The self-play version used the
raw board encoding directly,  without any hand-crafted heuristic
features. It reached a level of play comparable to Neurogammon: that of an
intermediate-level human Backgammon player. After adding a simple
heuristic search it reached world-class level.

TD-Gammon's
success  inspired many other researchers to try 
neural networks and self-play approaches, culminating eventually in recent high-profile results in  Atari~\cite{mnih2013playing} and
AlphaGo~\cite{silver2016mastering,silver2017mastering},  
Chaps.~\ref{chap:dqn} and \ref{chap:plan}). 

A modern
reimplementation of TD-Gammon in TensorFlow is available on GitHub
\href{https://github.com/fomorians/td-gammon}{TD-Gammon}.\footnote{\url{https://github.com/fomorians/td-gammon}}



\subsection{Chinook 1994}
The Checkers program Chinook was the first computer program to win the
world champion title 
in a competition against humans. It was developed by Canadian computer
scientist and Chess player Jonathan
Schaeffer. 
Based on his experience in Chess,
 Schaeffer decided to write a Checkers program, 
believing that the technology existed to achieve world
championship level. 

Chinook's architecture follows the standard search-eval approach of computer Chess, based
on alpha-beta search and a hand-crafted heuristic evaluation
function (see Chap.~\ref{chap:idtt}). Both functions were highly
enhanced, well tested, and highly
tuned~\cite{schaeffer1992world,schaeffer1996chinook}. Apart 
from meticulous 
testing and tuning, Chinook derived much of its strength from its 
endgame databases. These endgame databases  contained perfect knowledge for any board
position with 8 or fewer pieces on the board.
\index{Chinook}
\index{Checkers}

Schaeffer started development in 1989. At first, progress came
swift. In 1990 Chinook won the right to play in the 
human World Championship by being second to Marion Tinsley in the US
Nationals. 
Tinsley, the current World Champion, was an exceptionally
strong player who had reigned supreme for  a long time in
Checkers.

In  1994  Chinook was declared
the Man--Machine World Champion. In 1995, Chinook defended its
Man--Machine title 
in a 32-game match. The final score was 1–0 with 31 draws for Chinook. At
the time Chinook was rated at a very high 2814 Elo. After the match,
Chinook retired from competitive 
playing. Later, Schaeffer and his team used Chinook to solve Checkers,
proving  in 2007 that with perfect play by both sides, Checkers is a
draw~\cite{schaeffer2007checkers}.

Chinook's win is made all the more impressive because of the
opponent it faced: Dr. Marion Tinsley. Tinsley,
a professor of mathematics, played Checkers at a level that was
unheard of. He was
world champion from 1955--1958 and 1975--1991 and never lost a world
championship match, and lost only seven games  in his 45-year career,
two of which to  
Chinook~\cite{schaeffer2008one}. He withdrew from
championship play during the years 1958--1975, relinquishing the title
during that time. It was said that Tinsley was to Checkers what
Leonardo da Vinci was to science, what Michelangelo was to art, and
what Beethoven was to music~\cite{madrigalcheckers}. Schaeffer wrote
a book about his experience, in which he
recalls  one particular event~\cite{schaeffer2008one}. In one game from
their match in 1990, Chinook made a
mistake on the tenth move. Tinsley instantly remarked, ``You're going to regret
that.'' And, yes, after they played on, Chinook resigned after move
36, fully 26 moves later, when it saw its loss.  Schaeffer looked back
into the database and 
discovered that Tinsley picked the only strategy that could have
defeated Chinook from that point and that Tinsley was able to see his
win an implausible
64 moves into the future~\cite{schaeffer2008one}.



\begin{figure}[t]
\begin{center}
\includegraphics[width=6cm]{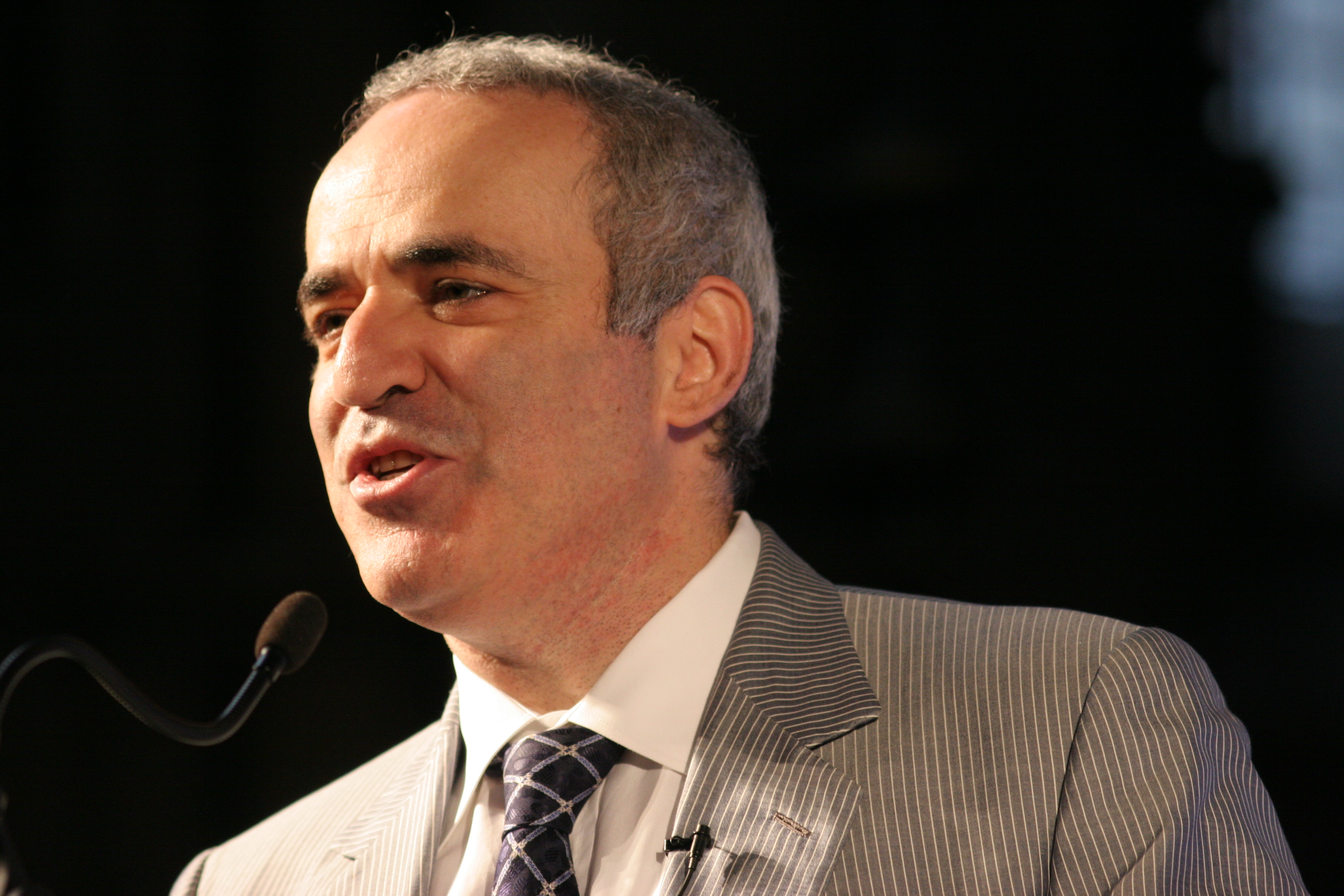}
\caption{Garry Kasparov}\label{fig:garrykasparov}
\end{center}
\end{figure}


\begin{figure}[t]
\begin{center}
\includegraphics[width=11cm]{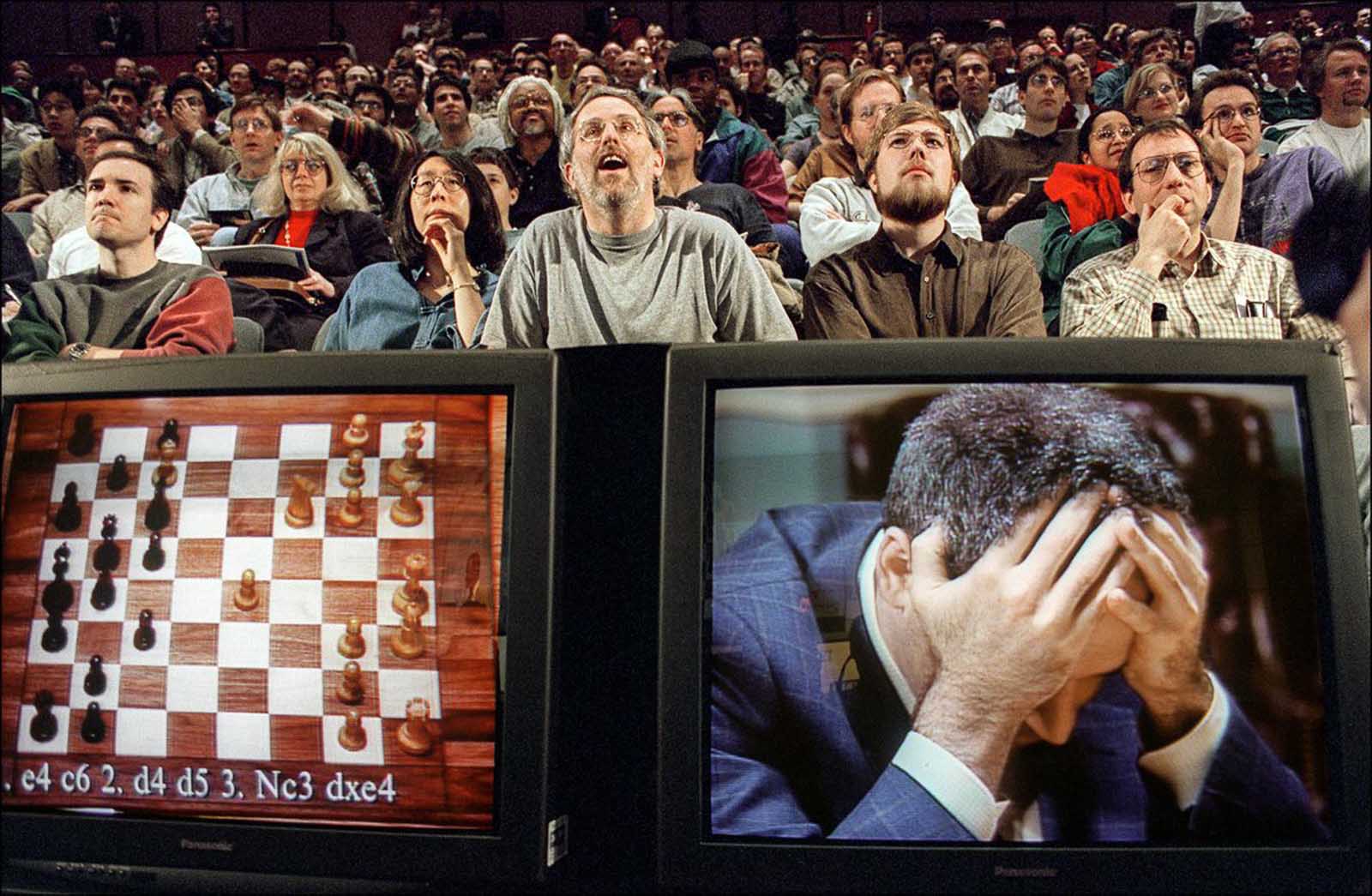}
\caption[Game 6 Caro Kahn]{Excitement in 1997. Game 6, Caro Kahn. Kasparov resigned
  after 19 moves}\label{fig:kasparov}
\end{center}
\end{figure}

\subsection{Deep Blue 1997}\label{sec:deepblue}\index{Deep Blue}
If the Checkers events caused excitement, then Chess did even more so. In 1985
Feng-hsiung Hsu, 
a PhD student at Carnegie
Mellon University, started work on a Chess program that, after much
development, showed enough promise for IBM to become involved.
After
quite some more years of development, in 1996 performance had improved enough
for IBM to sponsor a match against the human world Chess 
champion Garry Kasparov
(Fig.~\ref{fig:garrykasparov}).\footnote{Pictured speaking at the Turing centennial conference in Manchester. Image credit: Creative Commons.} The match
took place at the ACM Computer Science Conference in
Philadelphia. 
On February 10, 1996, Deep
Blue won its first game against human champion Kasparov, making Chess
history. However, Kasparov went on to win three games and draw two,
winning the six-game match by 4--2.

A year later, in May 1997, the rematch was played in New
York at the 35th floor of the Equitable Center (Fig.~\ref{fig:kasparov}).\index{Garry Kasparov} An upgraded version of 
Deep Blue played Kasparov again, this time capable of evaluating 200
million positions per second, 
twice as fast as the 1996 version. Kasparov was defeated $3\frac{1}{2}$--$2\frac{1}{2}$
in a highly publicized match. Deep Blue became the first computer system to
defeat a reigning world Chess champion 
in a match under standard Chess tournament time controls. Several 
accounts of the match, its historical significance, and the technology
behind it, have been
written~\cite{schaeffer1997kasparov,campbell2002deep,hsu1990grandmaster,hsu2004behind,levy1991computers,newborn2013deep}. 

Deep Blue started out as a hardware design project, when Feng-hsiung
Hsu wanted to see if he could create a
special-purpose VLSI Chess move generator chip. This project evolved into a
full-fledged Chess program, which was named Deep Thought, after the 
fictional computer in Douglas Adams' {\em The Hitchhiker's Guide to the
Galaxy,} that computed the answer to life, the universe, and
everything.\footnote{Which, as you may know, is
  42.}
\index{Deep Blue}\index{Garry Kasparov}\index{Chess}\index{ChipTest}

The design of Deep Blue followed the standard search-eval architecture, with quite a
number of special enhancements. To start with, the evaluation function used 480
specially developed hardware chips, allowing it to evaluate 200 million
positions per second in parallel. The search algorithm was also heavily
parallelized, running on a 30-node IBM RS/6000 system.
Deep Blue  would
typically search to a depth of 6--8 moves and to a
maximum of more than 20 moves in some situations.\footnote{To be
  precise, 6--8 moves for a side. In official Chess terminology a
  move is one half-move by white plus one half-move by black. So what
  we call 6--8
  moves should really be called 3--4 moves if we would follow official
  Chess terminology.}
The opening library was provided by grandmasters Miguel Illescas, John
Fedorowicz, and Nick de Firmian. Deep Blue had an endgame database of
positions with 5--6 pieces.

Deep Blue's evaluation chips were highly customizable, containing
8000 individual parts, and a database of master games was used to tune their
weights~\cite{hsu2004behind}.  
The success of Deep Blue was rightfully billed as the success of
brute force, due to the impressive number of 200 million evaluations
per second. However, the fact that the evaluation weights were
initially tuned by supervised learning, and not by hand, would equally
allow a claim to a victory  due to machine learning (although it did
not use neural nets, nor did  Deep Blue use any  deep
learning).



In its time, Deep Blue, with its capability of evaluating 200 million positions per
second, was the fastest computer to face a world Chess
champion. Since then, the focus has shifted to
software, rather than using dedicated Chess
hardware. Chess algorithms have progressed, and modern Chess programs
such as Stockfish, Houdini, and Deep Fritz are more efficient than the
programs of Deep Blue's 
era. Today less hardware is needed to achieve the same performance. In a
November 2006 match between Deep Fritz and World Chess 
Champion Vladimir Kramnik, the program ran on a computer system
containing one dual-core Intel Xeon 5160 CPU, capable of evaluating only
8 million positions per second, but searching to an average depth of
17 to 18 moves in the middle game, thanks to fast heuristics and
aggressive search
extensions (Chap.~\ref{chap:idtt}).\index{Stockfish}
Stockfish is currently one of the strongest Chess programs. It is open
source, available on GitHub 
\href{https://github.com/official-stockfish/Stockfish}{here}.\footnote{\url{https://github.com/official-stockfish/Stockfish}} 


\subsection{Logistello 1997}
\index{Logistello}\index{Othello}\label{sec:logistello}
After the massive team effort for Chess, a year later a one-person effort
achieved impressive results in the game of Othello.
The program is named Logistello. It was written by Michael Buro
and was  a very strong player, having beaten the human world champion
Takeshi Murakami with a score of 6--0 in 1997. The best Othello programs
are now much stronger than any human player.

Logistello's evaluation
function is based on patterns of discs, and has over a million
numerical parameters which were tuned using  advanced logistic linear regression~\cite{buro1995logistello,buro1997othello,buro2003evolution}.


\subsection{AlphaGo 2016}\index{AlphaGo}\index{Go}\label{sec:alphago2}

After the 1997 defeat of the  Chess world champion, the game of Go
(Fig.~\ref{fig:go}) became the next   benchmark 
game, the \emph{Drosophila} of AI,  and research interest in Go intensified significantly.

Traditionally, computer Go programs followed the conventional Chess
design of a minimax search with a heuristic evaluation function that
was influence based. The GNU Go program is a good example of this 
approach~\cite{millen1981programming}\index{GNU Go}. This Chess
approach, however, did not work for Go, or at least not well enough. The level of play
was stuck at mid-amateur level for quite some time.

In a sense, the game
of Go worked very well as \emph{Drosophila}, in that new AI approaches were
developed, and many researchers  produced  interesting findings. New
algorithms have been developed, such as Monte Carlo Tree Search, 
and impressive progress has been made
in deep reinforcement learning.

\begin{figure}[t]
 \begin{center}
 \begin{tabular}{ccc}
\includegraphics[width=3.5cm]{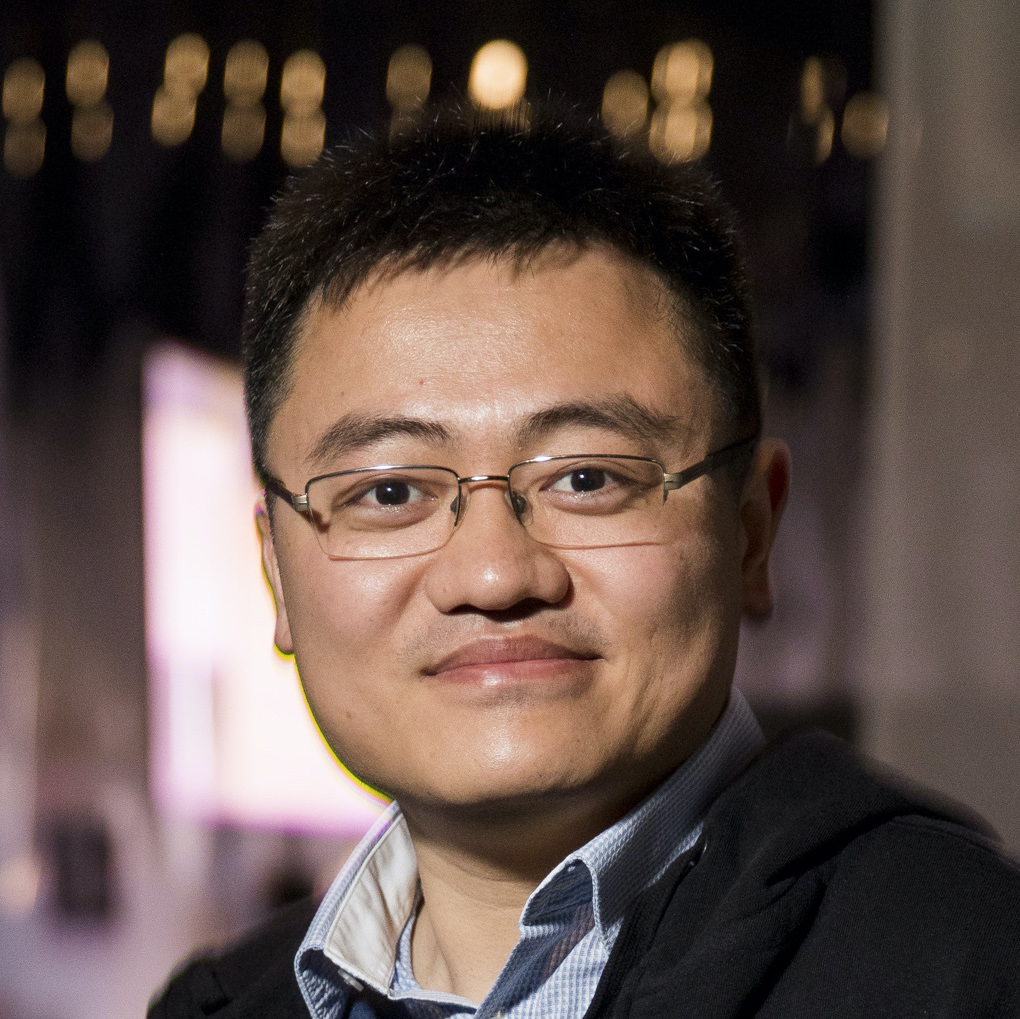} &
\includegraphics[width=3.5cm]{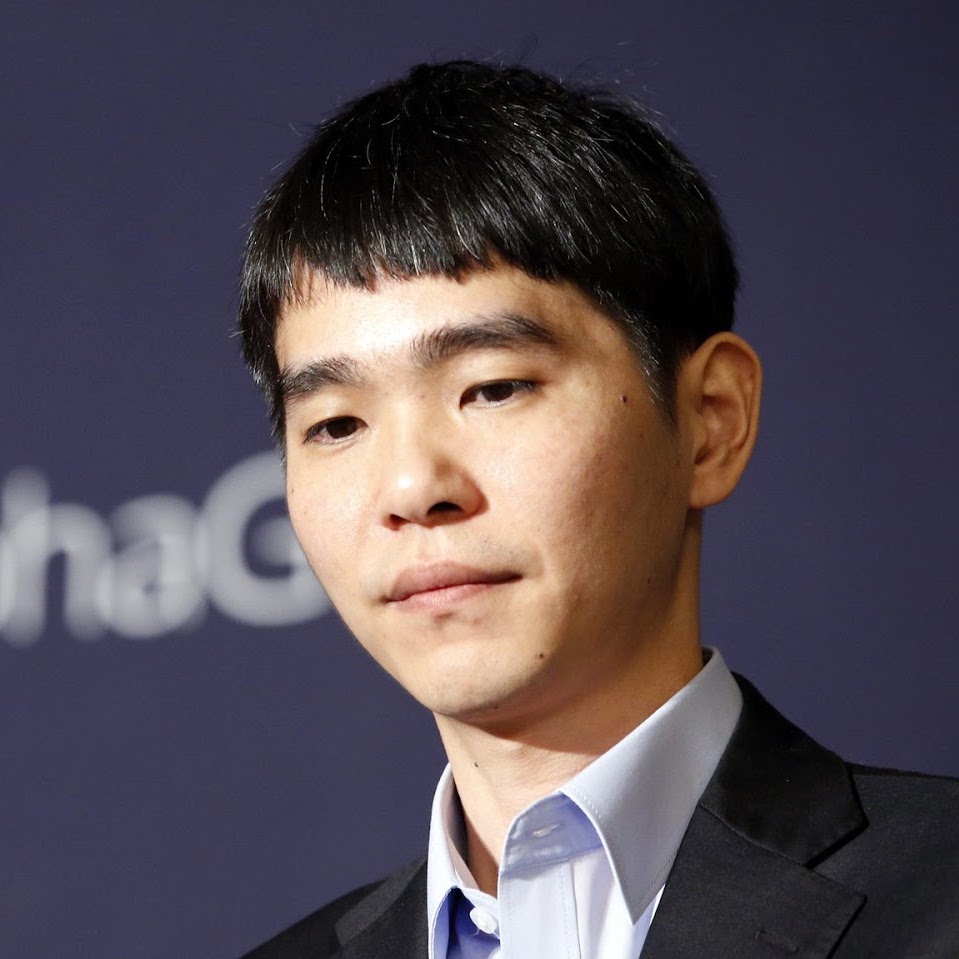} &
\includegraphics[width=3.5cm]{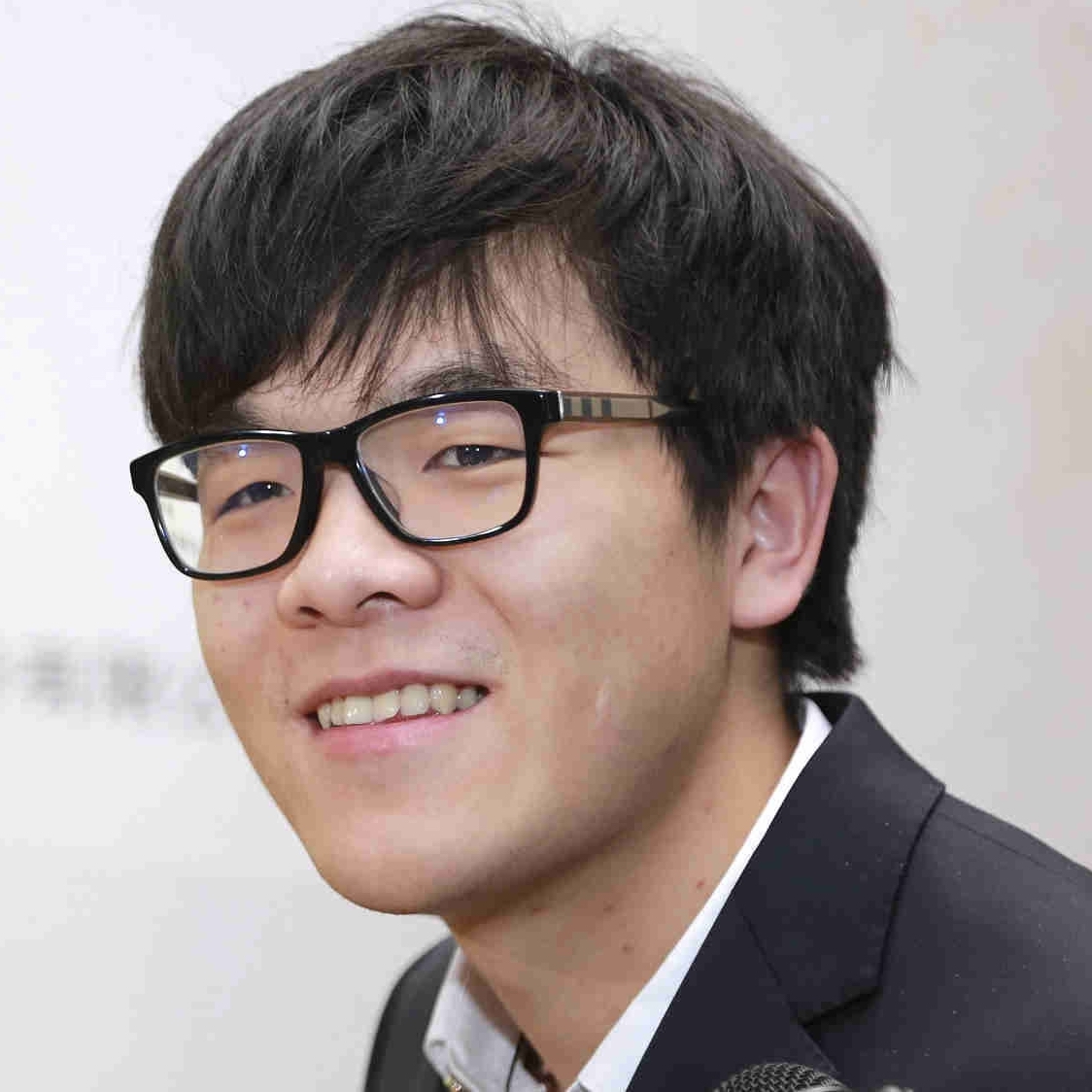} \\
\includegraphics[width=3.5cm]{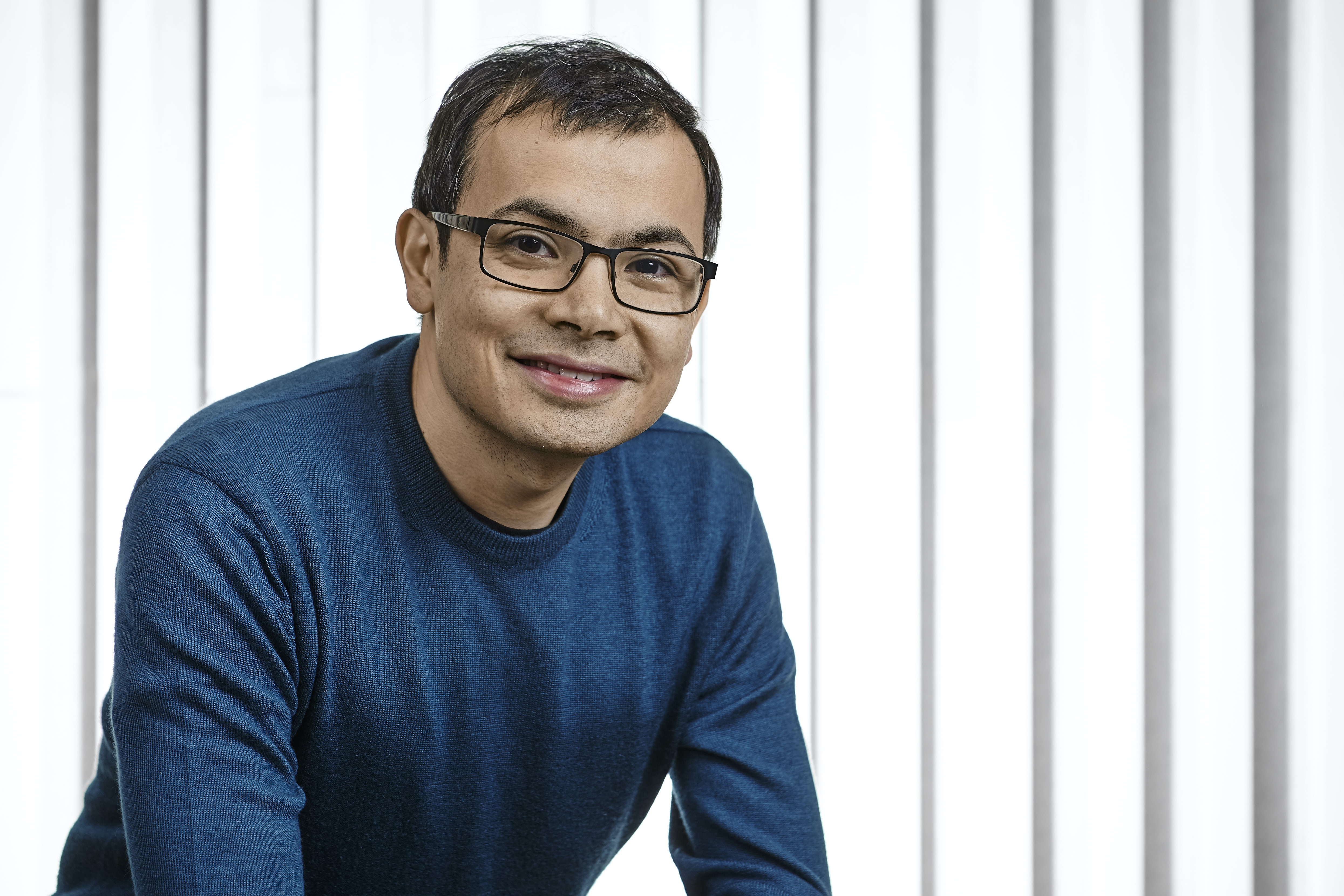} &
\includegraphics[width=3.5cm]{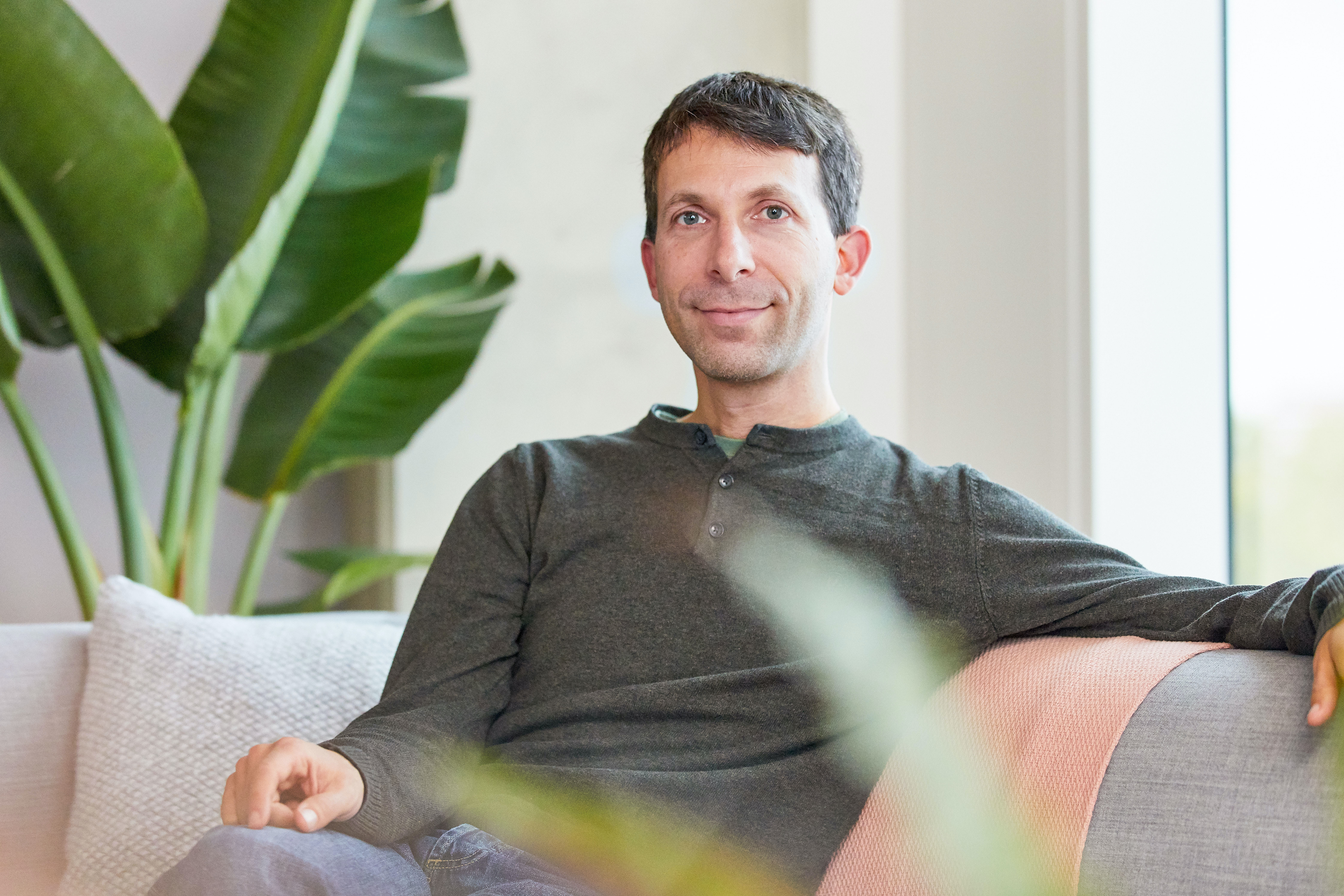} &
\includegraphics[width=3.5cm]{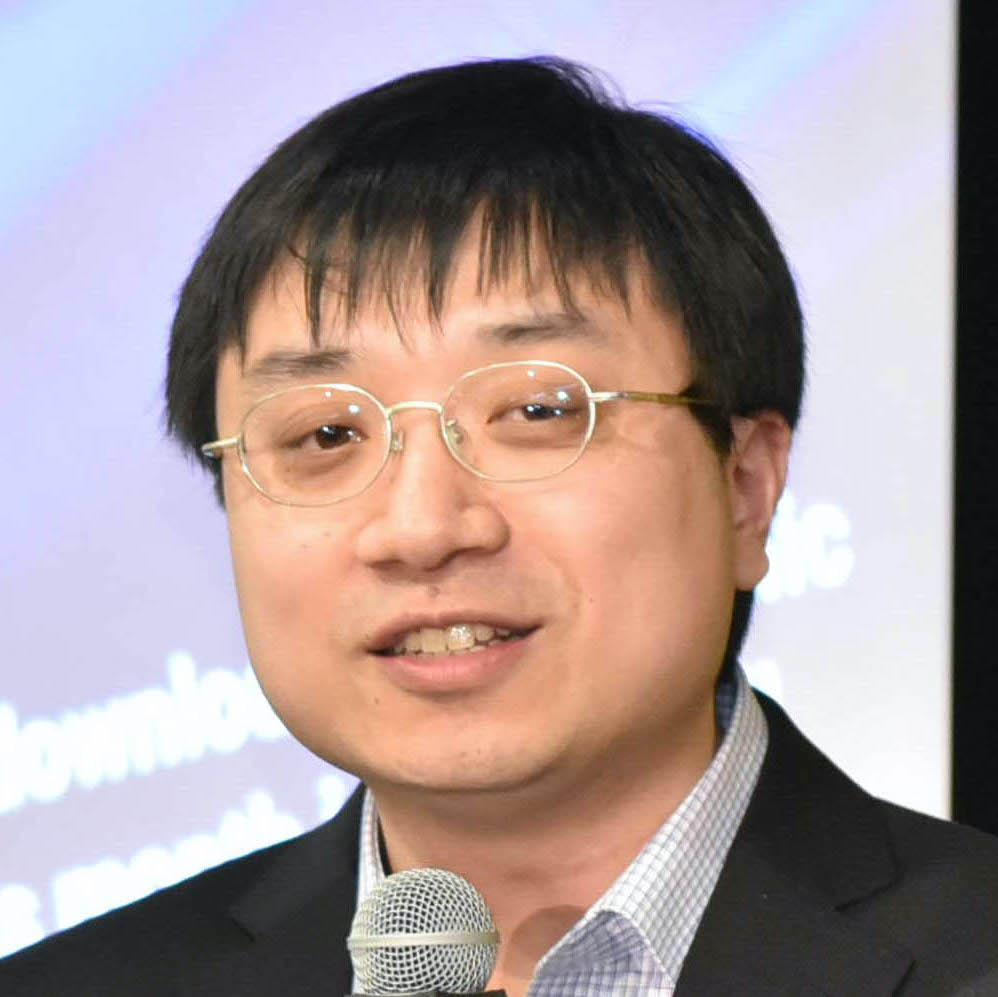}
\end{tabular}

 \caption[DeepMind team and Go champions]{Fan Hui, Lee Sedol, Ke Jie;
   Demis Hassabis, David Silver and Aja
 Huang (left-to-right, top-to-bottom)}\label{fig:fanhui}
 \end{center}
 \end{figure}

In 2015--2017 the DeepMind AlphaGo team, headed by
David Silver, 
played three matches
in which it beat all human champions that it played. 
There was even a book written analyzing the Go-theoretical innovations that AlphaGo has
unlocked for human Go masters~\cite{zhou2018}.
%
%
%
To get an appreciation of the achievement of AlphaGo, and to
introduce the program, we will now have a closer look at the matches
that it played, quoting some of the commentary that was written about
the matches.
Please go to Appendix~\ref{chap:matches} for a record of all games played in the Fan Hui
tournament, the Lee Sedol tournament, and the Ke Jie tournament. First
we discuss the Fan Hui games, then the Lee Sedol games, and finally
the Ke Jie games. Fig.~\ref{fig:fanhui} shows the players and some of
the AlphaGo team members.\footnote{Image credit Hassabis and Silver:
  DeepMind; Hui: Tristan Fewings/Getty Images.} 
\index{Fan Hui}\index{Lee Sedol}\index{Ke Jie}

\subsubsection*{Fan Hui}
The games against Fan Hui were   played in October 2015 in London
as part of the development effort of AlphaGo. Fan Hui is the 2013, 2014, and
2015 European Go Champion, then rated at 2p dan. He described the program as very
strong and stable. ``It seems like a wall. I know AlphaGo is a
computer, but if no one told me, maybe I would think the player was a
little strange, but a very strong player, a real person.''

\subsubsection*{Lee Sedol}
The games against Lee Sedol were played in May 2016 in Seoul as part
of a highly televised media circus. Although there is no official
worldwide ranking in international Go, 
in 2016 Lee Sedol was widely considered  one of the four best
players in the world. Lee Sedol has been a 9p dan professional Go 
player since 2003. When he played AlphaGo, he had 18
international titles, and was confident that he would win against the
computer.

To his surprise, the match ended in a 4--1 win for AlphaGo, with the
program playing very strong games.
The games against Lee Sedol attracted a large amount of media
attention, comparable to the Kasparov--Deep Blue matches two decades
earlier.  A movie was made of AlphaGo, which can be found  \href{https://www.alphagomovie.com}{here}.\footnote{\url{https://www.alphagomovie.com}}   

\begin{figure}[t]
    \centering{\includegraphics[width=7cm]{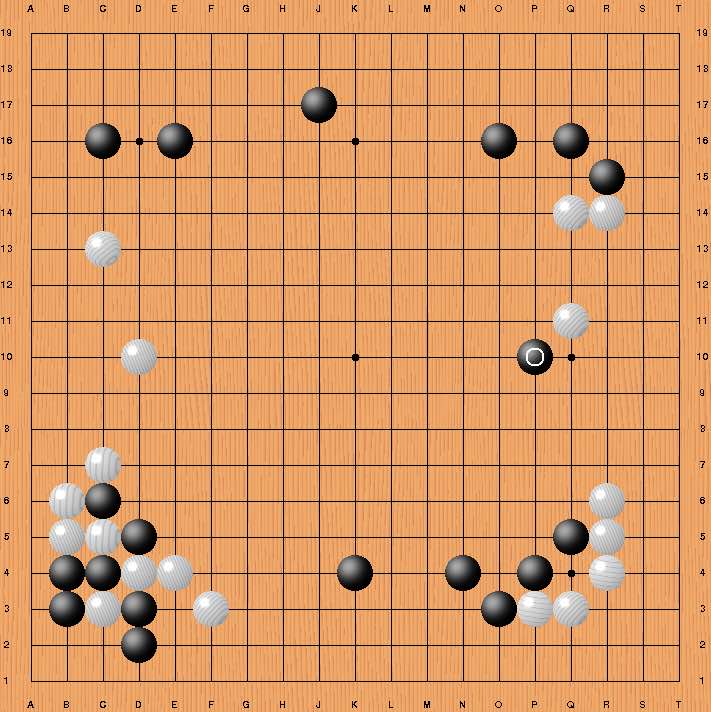}
     \caption{``Speechless'' move 37 by AlphaGo  in game 2 (played on
       P 10)}\label{fig:leesedolmove37}}
\end{figure}

The match started with game 1, which was a win for AlphaGo.
Then, in game 2, AlphaGo played a  move on the right-hand side
of the board that shocked spectators (Fig.~\ref{fig:leesedolmove37}). Commentators said
that it was a very strange move; many thought that it was a mistake. Lee
Sedol took nearly fifteen minutes to
formulate a response. Fan Hui explained that, at
first, he also could not believe his eyes. But then he saw the beauty in
this rather unusual move.
Indeed, the move changed the course of the game, and  AlphaGo went on
to win.
At the post-game press conference Lee Sedol was in
shock. He said: ``Yesterday, I was surprised,'' referring to his loss
in game 1.\label{sec:beauty}
``But today I am speechless. If you
look at the way the game was played, I admit, it was a very clear loss
on my part. From the very beginning of the game, there was not a
moment in time when I felt that I was leading.'' 

Fan Hui gives more commentary, showing why this move looks so profound
from a human perspective.
He writes: ``Black 37 is one of the two moves from this match sure to
go down in Go history. This move proved so stunning that, when it
appeared on the screen, many players thought the stone had been put
down in the wrong place.''  


``On seeing Black 37, I wrote down the following: `Here?! This goes
beyond my understanding. Globally, there's nothing wrong with it, it's
going in the right direction [$\dots$] and AlphaGo always pays special
attention to coordinating the stones. It seems anything is possible in
Go! Everyone will be talking about this move! A human would never dare
play it, it's too difficult to estimate. But AlphaGo can. Perhaps this
move is a sign of its confidence.' '' 

``This move made a deep impression on me during the game. I
experienced first confusion, then shock, and finally delight. It
reminded me of an old Chinese saying: `A beginner plays the corners,
an average player the sides; but a master controls the center.' These
days, due to the convergence of strengths and the pressure of
competition, something close to the opposite is true, with most
players focusing on the corners and sides. In contrast, AlphaGo's
talent for central control is second to none. Perhaps, through
AlphaGo, we too can become the `masters' of which the proverb speaks.'' 

``Returning to the game, we may say that Black 37 casts an invisible
net across the board. Together with the lower side, Black's shoulder
hit creates potential all across the center. Although it helps White
make territory on the right, the presence of White  means that a
Black invasion there would not have been valuable anyway. Of course,
Black should be reluctant to give away fourth-line territory too
easily, but one must give to get.''

``After the match, when I examined the data back at DeepMind, I saw
that AlphaGo had not even been thinking about 37 only a few moves
before. It had been expecting a different move, and its data indicated that a
human player would hardly consider the shoulder hit a possibility. It
was only when White played 36 that AlphaGo discovered 37, and boldly
decided that this move would work even better.''

``The pace of the game was much slower than the previous day, so Lee had
already gone out to smoke before 37. The minute he caught sight of
AlphaGo's reply, he stared blankly at the board. Then he smiled, sat
down, and started thinking. The longer he thought, the more serious
his expression became, while the clock ran down little by
little''~\cite{fanhui2016deepmind}. And indeed, this game was also won by AlphaGo.

Lee continued to play three more games in the match, winning  game 4
by a beautiful move, but
losing the match 1--4, the first time in history that a Go champion of
his caliber had lost a match to a computer program.



\subsubsection*{Ke Jie}
A year later another match was played, this time in China, against the Chinese
champion. Ke Jie was ranked  number one in
the Korean, Japanese, and Chinese ranking systems at the time of the match.  He was also first among all human
players worldwide  under R\'emi Coulom's ranking system~\cite{goratings}.

On 23--27 May 2017 a three-game match was played in Wuzhen, China. The
match was won
3--0 by AlphaGo.
AlphaGo was subsequently awarded a professional 9-dan title by the
Chinese Weiqi Association. 

As preparation for this match the team had organized an online 60-game
tournament against 
top pros, from 29 December 2016 to 4 January 2017. This version of
AlphaGo ran on  a strong machine that used four
specially developed tensor processing units (TPU). TPUs are a kind of GPU, specially
developed by Google for fast and efficient
tensor processing. They are
described in Sect.~\ref{sec:tpu}. This version of AlphaGo achieved
a very high Elo rating of 
4858 against the online pros~\cite{silver2017mastering}; it showed
that the program was 3 stone stronger than the version
that defeated Lee Sedol.\index{TPU} They named this match version {\em
AlphaGo Master}. It was a refined and further developed version of
AlphaGo, although the basic architecture was unchanged, with a
combination of  supervised and reinforcement learning (even though the researchers
were, at the time, already working on a redesign, AlphaGo Zero). 

In the rest of the book we will have a look at the methods that were used to achieve these
results.  
Chapter~\ref{chap:trial} covers the search algorithm,
Chap.~\ref{chap:dqn} covers deep learning, and 
Chap.~\ref{chap:plan} goes in depth into self-play.

Appendix~\ref{chap:resources} contains  links to tutorials on the
rules of Go and on learning to play.

\subsection{Poker 2018}\index{Poker}
In contrast to games of perfect information such as Chess, Checkers,
and Go, Poker is a game of imperfect information. It is a card game in
which some of the cards are hidden from the player. The set of
possible actions is large, posing a challenge for search-based
AI approaches. This has held back the
development of strong Poker programs for some time.



Recently, however, impressive progress has been made by groups from
the University  of
Alberta~\cite{bowling2015heads,bowling2009demonstration,moravvcik2017deepstack}
and Carnegie Mellon University~\cite{sandholm2010state,brown2018superhuman}.
Programs DeepStack and Libratus have defeated some
of the strongest human Poker players in one-on-one
play~\cite{brown2018depth}.\index{DeepStack}\index{Libratus}\index{Pluribus}
A year later they published equally strong results for
multiplayer Poker, with the Pluribus
program~\cite{brown2019superhuman} defeating top players in
six-player Poker. 


Research in computer Poker is an active field, with versions of
research programs  being licensed for commercial entertainment
purposes. Section~\ref{sec:poker} provides  more information on the
methods for Poker.

\subsection{StarCraft 2018}\index{StarCraft}\index{AlphaStar}
StarCraft is a popular real-time strategy  game with
strong human 
champions. Researchers have become interested in
using the game as a test bed for AI. Since 2010 academic competitions
have been organized that stimulated
research~\cite{togelius2010multiobjective,ontanon2013survey}.  In
2017 a Python interface to the game was created to facilitate its
use in AI research~\cite{vinyals2017starcraft}.

In 2018 DeepMind's player AlphaStar played two-person test matches
against some of the world's strongest players, which it
won~\cite{vinyals2019grandmaster}. Research into multiplayer methods
for StarCraft is becoming quite an active field.
Section~\ref{sec:starcraft} provides more information on the
challenges that StarCraft poses for AI and looks into further research.

\section*{Summary}
We have covered a lot of ground in this chapter. We have discussed elements of
intelligence, and mentioned that most AI researchers take a
behavioristic approach to intelligence. For AI researchers ``to be or not to be'' is not
the question; most  take the pragmatic
 approach of duck-typing: ``If it walks like a duck and
quacks like a duck, then it must be a duck.''

We have looked at
characteristics of games, and discussed the important
concepts {\em zero sum\/} and {\em perfect information}. We have also
had a look at historical approaches to game playing programs, and we
have listed modern successes, where computer programs have achieved a
level of play beyond that of the strongest human players.

This chapter discussed key aspects of AI:
recognition, reasoning, memory, and learning. AI takes a behavioristic
approach to intelligence, which is operationalized with the Turing
test, introduced by Alan Turing in 1950. We then proceeded with a
brief history of AI, starting with the 1956
Dartmouth College workshop. A list of ambitious goals for AI were
formulated, which turned out to be overly ambitious, causing  AI funding
winters. However, after research picked up by the turn of the
millennium, a number of  AI 
technologies (such as search, social networks, deep learning, and recommender
systems) had deeply affected society, and AI summer arrived.

Games have been a favorite test-bed for AI research. Games such as
Chess and Go have been called the \emph{Drosophila} of AI, after
the fruit fly from genetics research. 
An important difference between human intelligence and artificial 
 intelligence is that human intelligence is general, or broad, while
artificial intelligence is special, or narrow. Human intelligence can solve many
problems; an AI program can solve only one, such as playing good
Chess. Research is under way into artificial general intelligence. 

The chapter provided a short overview of important designs and
programs. We listed the early designs by Shannon, Strachey, Turing, and Samuel. The
programs featured were TD-Gammon (Backgammon), Chinook (Checkers),
Deep Blue (Chess),  Logistello (Othello), AlphaGo (Go), Libratus and
DeepStack (Poker), and AlphaStar (StarCraft). 

Games are a favorite experimenter's tool for AI.
Many breakthroughs in intelligence and learning have taken place in a
variety of games, each with their own characteristics.

The early designs established some of the basic algorithms and
approaches, such as the search-eval architecture, minimax, alpha-beta,
heuristics, and learning. Computers were not yet powerful enough to
reach strong levels of play. The later programs all did, using highly
enhanced versions of the original blueprints, as well as new
techniques, such as deep learning.

It is now time to prepare ourselves for the four core chapters of this
book that
look in detail at the methods that
have been developed by researchers. We will do so by first discussing the
reinforcement learning paradigm, as a common language for the four
following chapters.

\section*{Historical and Bibliographical Notes}
We conclude each chapter with a brief summary of entry points to the
scientific literature. The works mentioned here may be more accessible
than some of the more technical papers. 

In this chapter we now list  a few  such works.

An excellent, popular, and
all-encompassing standard textbook of AI is by Russell and
Norvig~\cite{russell2016artificial}. If you have taken an undergraduate
course in AI, chances are that you already own a copy.

Artificial intelligence and search are also used to generate content
for computer games~\cite{togelius2010search}. 
An excellent text on how to use artificial intelligence for creating
entertaining and believable computer games is Yannakakis and
Togelius~\cite{yannakakis2018artificial}, which also covers procedural
content generation and game design.

Insightful and entertaining accounts have been written about the human
and scientific side of writing world champion-level game playing
programs. Schaeffer's account of how he
and his team created Chinook is  memorable, not just
because of his inside story, but also because of the picture that
it paints of Marion Tinsley, the amazing Checkers
champion~\cite{schaeffer2008one}. 

Much has been written about the Chess match between Deep Blue and Garry
Kasparov. Feng-hsiung Hsu's account gives the story behind their ``quest
to build the mother of all Chess machines.'' It is an awe-inspiring
story of this moment in the history of Chess-playing intelligence
versus computer programming intelligence~\cite{hsu2004behind}.

A good reference on how computers play Chess is
\cite{levy1991computers}, and Chap.~\ref{chap:search} covers many of
the techniques that are used in Chess programs.

OpenSpiel is a library for reinforcement learning in games. It
provides a wealth of  high-quality implementations of
algorithms~\cite{lanctot2019openspiel}. A link to OpenSpiel is
\href{https://deepmind.com/research/open-source/openspiel}{here}.\footnote{\url{https://deepmind.com/research/open-source/openspiel}}\index{OpenSpiel}
Appendix~\ref{ch:env} contains more  environments relevant
to research in  reinforcement learning and games.

A popular psychology book is \emph{Thinking Fast and Slow} by
Kahneman~\cite{kahneman2011thinking}. Kahneman is one of the founders of
the field of behavioral economics, and his approach to intelligence
fits well with the search-eval architecture that is prevalent in
artificial intelligence in games.

\chapter{Reinforcement Learning}\label{chap:rl}

The field of reinforcement learning studies the behavior of agents that
learn through interaction
with their environment. Reinforcement learning is a general paradigm,
with links to trial-and-error methods and behavioral conditioning
studies.
In this chapter we will introduce  basic concepts and algorithms that will be
used in the rest of the book.

It is customary to use Markov decision processes (MDPs)
as the  formalism to model reinforcement learning. MDPs are  a suitable paradigm  to model the games that we use as benchmarks of intelligent behavior. 
Reinforcement learning and Markov decision processes allow solution
methods for games to be formalized concisely  and precisely.
Our language will be 
mathematical in nature, but   links to agent behavior, and
behavioral studies, are never
far away.

To start, let us summarize the core problem and  concepts that
will be covered in this chapter.

\section*{Core Problem}
\begin{itemize}
\item How can an agent learn from interaction with an environment?
\end{itemize}
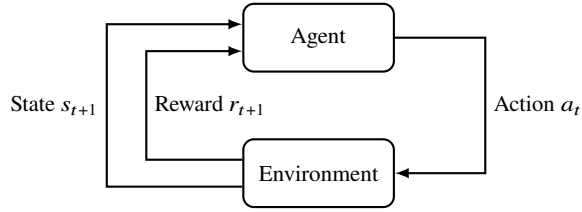
\begin{figure}[t]
\begin{center}
\tikzstyle{block} = [rectangle, draw, 
    text width=6em, text centered, rounded corners, minimum height=3em]
    
\tikzstyle{line} = [draw, -latex]

\begin{tikzpicture}[node distance = 6em, auto, thick]
    \node [block] (Agent) {Agent};
    \node [block, below of=Agent] (Environment) {Environment};
    
     \path [line] (Agent.0) --++ (4em,0em) |- node [near start]{Action $a_t$} (Environment.0);
     \path [line] (Environment.190) --++ (-6em,0em) |- node [near start] {State  $s_{t+1}$} (Agent.170);
     \path [line] (Environment.170) --++ (-4.25em,0em) |- node [near start, right] {Reward $r_{t+1}$} (Agent.190);
\end{tikzpicture}
\caption{Reinforcement learning: agent and environment~\cite{sutton2018introduction}}\label{fig:agent}
\end{center}
\end{figure}
\index{reinforcement learning}\index{agent}\index{environment}\label{sec:rl}


\section*{Core Concepts}
\begin{itemize}
\item Agent and environment
\item State,  action, policy, and reward
\item Value function, credit assignment, exploration/exploitation, and
  exact/approximate
\end{itemize}  
The reinforcement learning paradigm consists
of an agent and an environment (Fig.~\ref{fig:agent}). The
environment is in a certain state $s$. At time $t$, the agent performs
action $a$,
resulting in a new state in the environment. Along with this new state
comes  reward value $r$ (which may be positive or negative). The goal of
the agent is to learn  how to maximize the rewards that the
environment returns. To this end the agent learns a function, called
policy $\pi$. It can improve (reinforce) this policy by interacting
repeatedly with the environment
(note that taking an action---sampling from the environment---may be
costly).

\section{Markov Decision Processes}\index{Markov}
Reinforcement learning problems can be 
modeled as Markov decision processes (MDPs).
Markov decision problems are an
important class of problems that have the Markov property: the next
state  depends only on the current state and the actions available in it (no
memory of previous states or other information is necessary)~\cite{howard1964dynamic}.
The no-memory property is important
because it makes reasoning about future states possible using only the
information present in the current state. If previous
histories would all influence the current state, then reasoning about
the current state would be much harder or impossible.
MDPs are named after Russian
mathematician Andrey Markov (1856--1922) best known for his work on stochastic
processes.

Where board games use the terms {\em board position\/} and {\em move,} the
mathematical formalisms of MDP and
reinforcement learning   talk about {\em state\/} and {\em action}.

\begin{figure}[t]
\begin{center}
\tikzset{
  treenode/.style = {align=center, inner sep=0pt, text centered,
    font=\sffamily},
  arn_n/.style = {treenode, circle, white, draw=black,
    fill=black, text width=2mm},
  arn_r/.style = {treenode, circle, black, draw=black, 
    text width=3mm, thick}
}

\begin{tikzpicture}[->,>=stealth',level/.style={sibling distance = 1.2cm/#1,
  level distance = 1cm}] 
\node [arn_r,label=above:{$s$}] {}
    child{ node [arn_n] {} 
            child{ node [arn_r] {} 
            }
            child{ node [arn_r] {}
            }                            
    }
    child{ node [arn_n] {} 
            child{ node [arn_r] {}
            }            
            child{ node [arn_r] {}
            }
            edge from parent node[right] {$\pi$} 
    }
    child{ node [arn_n,label=above:{$a$}] {}
            child{ node [arn_r] {} 
            }
            child{ node [arn_r,label=right:{$s'$}] {}  edge from parent node[right] {$r$} 
            }
    }
;  
\end{tikzpicture}
\caption{Backup diagram with MDP tuple~\cite{sutton2018introduction}}\label{fig:rltree}
\end{center}
\end{figure}
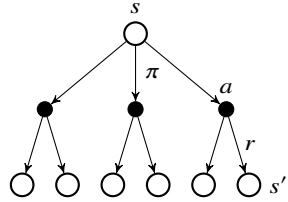

A Markov decision process is a 5-tuple $(S,A,P,R,\gamma)$:
\begin{itemize}
\item $S$ is a finite set of legal states of the environment; the initial state is denoted as $s_0$.
\item $A$ is a finite set of actions (if the set of actions differs
  per state, then $A_s$ is the finite set of
  actions at state $s$).
\item $P_a(s, s^\prime) = \Pr(s_{t+1} = s^\prime | s_t = s, a_t = a)$
  is the probability that action $a$ in state $s$ at time $t$ will
  lead to state $s^\prime$ at time $t+1$ in the environment.
\item $R_a(s,s^\prime)$ is the reward received after action $a$
  transitions state $s$ to state $s^\prime$.
\item $\gamma \in[0,1]$ is the discount factor representing the
  difference between future and present rewards.
\end{itemize}
Figure~\ref{fig:rltree} shows the MDP tuple $s, a, p, r$ as
well as $s^\prime, \pi$, and how the value can be calculated. The root node
at the top is state $s$, where policy $\pi$ allows the agent to choose between three
actions $a$, which, following distribution $p$, can each end up in two
states $s^\prime$, with their reward $r$. In this way the value of the
root can be calculated, by a backup procedure~\cite{sutton2018introduction}. (The value of $\gamma$
is 1, in this tiny example.) 

Deterministic board games such as Chess and Go can easily be
formulated as MDPs with $\gamma=1$ (future outcomes  (lookahead) are as important as present outcomes)
and  $P=1$  (actions  predictably lead to one future
state).\footnote{To be precise, only versions of Chess and Go that do
  not use rules to prevent repetition, such as the 50-move rule and
  the ko-rule. The repetition-prevention rules require memory and thus
  violate the Markov property.}

\subsubsection*{Terminology}\index{state}\index{action}\index{reward}
Reinforcement learning and MDPs introduce a new formal  terminology for board
games. The board
configuration is called  state $s$, a move is  action $a$,
and the score of an end position is the  reward $R$ (typically win, loss, or
draw). In reinforcement learning a  strategy for a player (all
the moves from positions leading to a certain end position and
outcome) is called a {\em policy.}  
In  games such as Chess and Go, actions lead deterministically to one
state (in Poker and Backgammon they can
stochastically lead to different states,
see~\cite{sutton2018introduction}).
The state space of a game is
the set of all states that are reachable from the initial board position or
state.

\begin{figure}[t]
\begin{center}
\includegraphics[width=8cm]{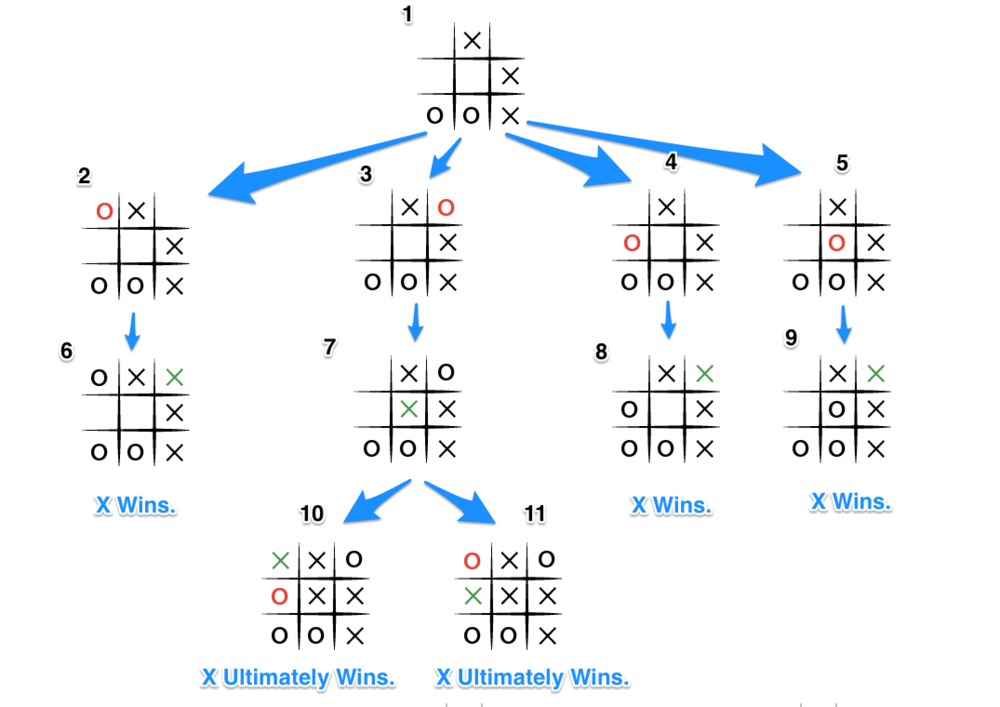}
\caption{Tic Tac Toe game tree}\label{fig:ttt-tree}
\end{center}
\end{figure}

The state space is easily visualized as a directed
graph.  Figure~\ref{fig:ttt-tree} shows a partial graph of a position in the well-known game
of Tic Tac Toe (also known as Noughts and Crosses, or Three in a
Row). This new diagram is a simplification of the diagram in
Fig.~\ref{fig:rltree}, since $p=1$ actions directly lead to a single
successor state $s^\prime$. Most trees in this book are of the new,
simpler, type.

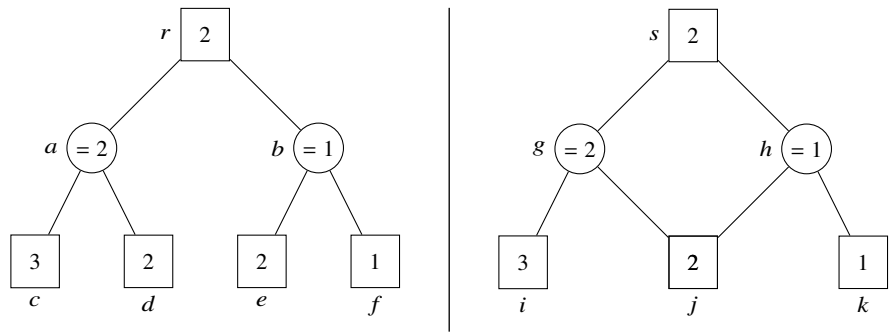
\begin{figure}[t]
  \begin{center}
    \begin{tabular}{c|c}
\begin{tikzpicture}
\tikzstyle{max node}=[rectangle,draw,inner sep=7]
\tikzstyle{min node}=[circle,draw,inner sep=2]
\node[max node,label=left:{$r$}](0){$2$}
          child[sibling distance=30mm]{node[min node,label=left:{$a$}]{$=2$}
                         child[sibling distance=15mm]{node[max node,label=below:{$c$}]{$3$}}
                         child[sibling distance=15mm]{node[max node,label=below:{$d$}]{$2$}}}
          child[sibling distance=30mm]{node[min node,label=left:{$b$}]{$=1$}
                         child[sibling distance=15mm]{node[max node,label=below:{$e$}]{$2$}}
                         child[sibling distance=15mm]{node[max node,label=below:{$f$}]{$1$}}};
                     \end{tikzpicture} \ \ \ \ \ \ \ \ 
      &
\ \ \ \ \ \ \ \ \begin{tikzpicture}
\tikzstyle{max node}=[rectangle,draw,inner sep=7]
\tikzstyle{min node}=[circle,draw,inner sep=2]
\node[max node,label=left:{$s$}](0){$2$}
          child[sibling distance=30mm]{node[min node,label=left:{$g$}]{$=2$}
                         child[sibling distance=15mm]{node[max node,label=below:{$i$}]{$3$}}
                         child[sibling distance=30mm]{node[max node,label=below:{$j$}]{$2$}}}
          child[sibling distance=30mm]{node[min node,label=left:{$h$}]{$=1$}
                         child[sibling distance=30mm]{node[max node]{$2$}}
                         child[sibling distance=15mm]{node[max node,label=below:{$k$}]{$1$}}};
                     \end{tikzpicture}
        \end{tabular}
\caption{A tree and a graph; node $j$ is a transposition}\label{fig:treetrans}
\end{center}
\end{figure}

Graphs introduce yet another terminology. States/boards are nodes in the
graph, actions/moves are links, and terminal states are leaves. Note
that the graph in  Fig.~\ref{fig:ttt-tree} does not have cycles, and is a tree
(artificial intelligence is full of ``trees,'' all growing upside
down, with the root at 
the top). Such a graph of the state space of a game is  often called a
game tree. In general, MDPs and game graphs may contain cycles, although most
games have rules to prevent never-ending game
situations. Another difference between graphs and trees are
transpositions: nodes that have more than one parent. Transpositions are identical
board positions that can be reached by different move
sequences. Section~\ref{sec:tt} discusses transpositions in more
detail. A tree does not contain transpositions: in a tree
only one path exists between any two nodes, and all children have exactly
one parent. Graphs can have transpositions: in a graph, child nodes may
be shared between parents, and
there are more ways than one to reach a certain position from the root. See
Fig.~\ref{fig:treetrans} for an example of a tree and a graph with a
transposition. Nodes in Fig.~\ref{fig:treetrans} are either squares
or circles. Squares indicate the max (first) player is to move, and circles
indicate the min (second) player is to move. The value of a max node is the
maximum of its children, while the value of a min node is the minimum of its
children (see Sect.~\ref{sec:minmax}).

A score function is defined for the leaves of the game tree/graph  to tell whether
the game has been won, lost, or drawn. In reinforcement learning
terminology we say that the reward is known. If we look at the graph
in Fig.~\ref{fig:ttt-tree} again, and we want to find the value of a
nonterminal board position, all we have to do is work upwards from
the leaves of the tree, and determine the  values of the parent
nodes, carefully taking the  max or min, until finally the  value of all nodes in the tree is known. At
that point we know if the initial position of the game is won or lost
if both players play perfectly.

Now that we have discussed basic terminology we are ready to look at
two central elements in reinforcement learning: the policy function
and the value function.

\section{Policy Function, State-Value Function, Action-Value Function}\index{value function}
In reinforcement learning two functions are usually associated with the
states: the policy function and the value function. Furthermore,  this value function
exists both for states and for actions.

\subsection{Policy Function}\label{sec:policy}\index{policy function}
Of central importance in reinforcement learning is the
concept of {\em policy,} or $\pi$. The policy function $\pi$
associates states with actions. A 
deterministic policy $\pi \colon s\mapsto a$ returns one action for each
state. A stochastic policy $\pi(a|s)$ gives the probability of taking
action $a$ in state $s$.

For two-person games the goal of reinforcement learning and MDPs is
for the agent to find an optimal
policy $\pi^\star$ that gives an action $a=\pi^\star(s)$ that should
be chosen in each state $s$  that
leads to a win for the first player.
Finding the optimal policy function is also known in engineering as
solving the {\em optimal 
control problem}~\cite{sutton2018introduction}.

\subsection{State-Value Function}\index{value function}
The state value function $V(s)$ returns the expected value of state
$s$; it is the discounted value of the expected future rewards of a state. In
leaves, the ``expected'' value  is equal to the reward $V(s)=R(s)$. 
The expected value $V(s)$ of a state $s$  is determined by the current policy, and
should therefore be written $V^\pi(s)$. Many solution methods exist to
 efficiently determine value functions and policy functions. Finding the
value function solves the {\em estimation problem}.

In the next chapter, Chap.~\ref{chap:search}, we will discuss
methods that efficiently determine this root value, and, by extension,
the optimal policy $\pi^\star$, the tree of actions that leads to it.

\subsection{Action-Value Function}\index{Q-function}
The action value $Q(s,a)$ is the value of taking action $a$ in state
$s$ under policy $\pi$. Writing it more precisely as $Q^\pi(s, a)$,
the function takes action $a$ and then
takes policy $\pi$ for the remainder. The value of the $Q^\pi(s,a)$ function is equal to the
value function when taking action $a$ in state $s$:
$$Q^\pi(s,a) = V^\pi_a(s).$$
The final value of state $s$ is given by the value function
$V^\pi$, after the fact, when all computations have been done. The
$Q^\pi(s,a)$ function gives the value of a single action $a$, and is
important during the calculation. The Q-value can be computed for intermediate
single action values. Important algorithms such as Q-learning and DQN
are named after this action-value function.

\subsection{Implementing Functions}
It is common to discuss value and policy problems in terms of 
functions. Functions map input values to output values, and are
sometimes called mappings.

At this point it is useful to consider that value and policy functions
can be implemented in different ways. 
The most basic and easy to implement situation is when for a function
$f(x) \mapsto y$  a closed-form 
mathematical formula is known, such as $f(x) = x+5$. The formula can be implemented
straightforwardly as a procedure that 
takes its input arguments and runs some code to calculate the return
value: \verb|def f(x): return x+5|.  This is the {\em code\/} approach.

A slightly more complicated situation is the following.
There are nonlinear functions $f(x) \mapsto y$ that can not be expressed as a closed-form
mathematical formula, or for which no such formula is known. However, when a
database of input/output pairs is available, then we can impement the
mapping as an array, list, or dictionary in most programming languages:
\verb|def f(x): return table[x]|.  This is
typically the case in machine learning problems, where the resulting
function is to be learned by looking at examples. When the number of
examples is manageable, they can be stored in memory. The function is
then implemented as  a lookup
table. 
This is the {\em database\/} approach. 

For most real-world machine learning situations we have an even more
complicated situation, where the database is too large to fit in
memory, or the training 
set must generalize to cover unseen examples from the test set that
are not present in the training database. In this case
the function can be learned  with  machine learning methods, although the
function values will necessarily be approximated. The function
mapping can  be implemented as a parameterized (neural) network that takes as input a
state, and produces  as output a value.  This is  what is  used in
many real-world
machine learning situations. Since the function is approximated using a
set of parameters $\theta$, it is often written as $f_\theta(x)
\mapsto y$.
This is the {\em function approximation\/} approach. 
Chapter~\ref{chap:super} is dedicated to function approximation.

\section{Solution Methods}
Now that we have introduced the basic elements and terminology of
reinforcement learning, it is time to discuss solution methods,
methods that help us find the optimal values and policies.

In this section we will  introduce  many building blocks of reinforcement learning:
the Bellman equation to recursively compute root values when the
transition function is known, the
exploration/exploitation trade-off to find optimal paths,
temporal-difference learning to find the policy by sampling, the value
iteration and policy iteration solution methods,  and on- and
off-policy methods for learning the optimal policy.

The methods will be
introduced one by one. At the end of the chapter full algorithms are described
that combine all these fundamental concepts.

We will now start with two
essential elements of reinforcement learning: Bellman recursion and the exploration/exploitation trade-off.


\subsection{Bellman Recursion}
\begin{figure}[t]
\begin{center}
\includegraphics[width=6cm]{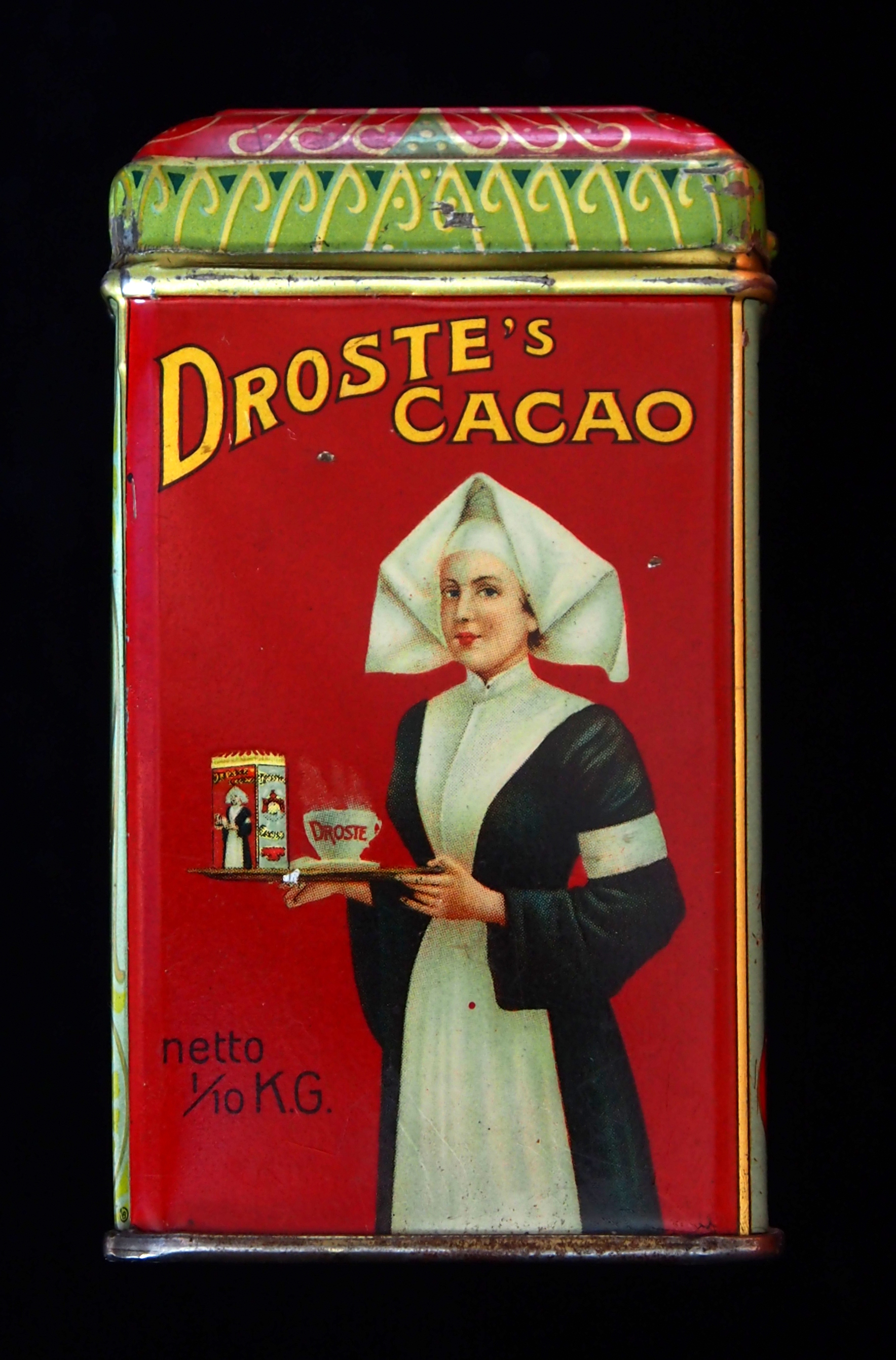} 
\caption{Recursion: Droste effect}\label{fig:dragon}
\end{center}
\end{figure}

Games consist of actions by  players. For each action that the
player makes, the optimal policy of the board position must
be found. (Or, equivalently, the optimal value must be
found.) In the previous section we discussed Tic Tac Toe and how the
optimal policy and the value of a position depend on the rewards, or leaf
values. We  mentioned how the value of the root 
can be found by recursively traversing the tree upwards, working
backwards, computing the values of
inner nodes. This  recursive solution method described informally in the previous
section can be described formally with the Bellman equation. Figure~\ref{fig:dragon} gives an artistic impression of the
concept of recursion, known as the Droste effect: a picture of a
picture of a picture within a picture etc. Richard Bellman
showed that
discrete optimization problems can be described as a recursive backward
induction problem~\cite{bellman1957dynamic}, using what he called
dynamic programming. The Bellman equation shows the
relationship between the 
value function in one state and the next state. 

The Bellman equation of the expected value for being in state $s$ and following  policy $\pi$ is
$$ V^\pi(s)=R(s, \pi(s)) + \gamma\sum_{s^\prime}P(s^\prime|s,
\pi(s))V^\pi(s^\prime) ,$$
where $R$ is the reward function and $\gamma$ is the discount
rate. The value of a state is equal to the reward of that state,
following policy $\pi$, and the discounted sum of the values of its successors. Before a node has been searched, it is not yet known which child is
the best successor. The Bellman equation defines a recursive relation on how to compute
the expected value of the initial state $s_0$.
Note that for games such as Chess and Checkers  the search space is defined
implicitly by the rules of the game and  the initial board position, of which we desire to know the
optimal policy  and  the expected value (win, draw, or loss).
 For common two-person zero-sum perfect-information games the Bellman
 equation simplifies greatly:
$\gamma=1$, and $R$ is commonly chosen to be $+1$ for win, $-1$ for
loss, and $0$ for draw, and also for intermediate nodes. The expected value
therefore  depends recursively on the rewards at  leaf nodes. The probability of transitioning from state $s$
to $s^\prime$ under perfect information in the optimal policy is
$1$. In  perfect-information game playing algorithms the Bellman equation reduces to a simple backup function.

\subsubsection*{Dynamic Programming}\index{dynamic programming}\index{Bellman equation}
Dynamic programming was
introduced by  Bellman in the 1950s. It recursively
breaks down problems into smaller subproblems that are then 
solved. Dynamic programming can be used to solve problems whose
structure follows the Bellman equation. Dynamic programming works when
the transition function $P$ is known; it is a model-based planning method  (Sect.~\ref{sec:modelbased}). The key idea of dynamic
programming is to use the value function to guide the
search for good policies~\cite{sutton2018introduction}. We will
discuss two dynamic programming algorithms for solving reinforcement
learning problems in Sect.~\ref{sec:sarsa}.
In the next chapter, we will see more  methods.

Two dynamic programming methods to iteratively traverse the state
space are value iteration and policy iteration. 

\subsubsection*{Value Iteration}\index{value iteration}\label{sec:valuepolicy}

Value iteration methods directly improve the estimate of the optimal
value function by traversing the state space.  Pseudocode for a basic version of value iteration is
shown in Listing~\ref{lst:val-it}.
Value iteration converges to the optimal  value function by
iteratively improving the estimate of $V(s)$. The value function $V(s)$ is first
initialized to random values. Value iteration repeatedly updates $Q(s,a)$ and 
$V(s)$ values until convergence occurs (when values of $V(s)$ stop
changing much). Value iteration has been proven  to
converge to the optimal values, but, as we can see in the pseudocode
in Listing~\ref{lst:val-it},
it does so quite inefficiently by essentially enumerating the entire state space.

\lstset{label={lst:val-it}}
\lstset{caption={Value iteration pseudocode (based on~\cite{alpaydin2009introduction})}}
\begin{lstlisting}[language=Python,float]
def value_iteration():
    initialize(V) 
    while not convergence(V):
        for s in range(S):
            for a in range(A):
                Q[s,a] = (*$\sum_{s'\in S} P_a(s,s')(R_a(s,s') + \gamma V[s'])$*)
            V[s] = max_a(Q[s,a])
    return V

\end{lstlisting}

Value iteration methods help find the best action  for each state
by following the action that yields the optimal value function. This
can work with a finite set of actions.  However, convergence in large
spaces is
slow~\cite{alpaydin2009introduction,sutton2018introduction}.

\subsubsection*{Policy Iteration}\index{policy iteration}

Value iteration algorithms keep improving the value function at each
iteration until the value function converges. However, most
agents only care about  finding the optimal control policy (the best move) and
not its value.

In games with a low branching factor, such as Checkers, there are only
few actions in a state. When there are also many different state
values (which is the case in most game playing programs) then the
optimal policy may converge before the value function. Directly
updating the policy function will then converge faster than taking the
route via the value function. 
Policy iteration methods directly improve the policy function.
This approach is also
useful when the action space 
is continuous (or stochastic). 

Both value iteration and policy iteration  can be used when there is prior knowledge
about the model/environment. For both methods the MDP 5-tuple,
especially the transition and reward function, must be  known. When that is
not the case, then other methods are needed, such as Q-learning (to be
discussed shortly).

\subsubsection*{Approximate Solutions}\label{sec:approx}
An important choice in implementing reinforcement learning
algorithms is which storage concept to use for the states, the
actions, and the policy and value functions.
For small state spaces, all states can be stored individually in a
table, array, or  tree. Small state spaces can be enumerated, and solutions can be exact.

Many reinforcement learning problems have state spaces that are arbitrarily large. For example, the number of possible images of even a small
black and white picture is very large (the number of gray scale values raised to the
power of the number of pixels). The number of instances
encountered during training will be small in relation to the possible
number of instances, and the trained policy should work well on unseen
problems. 
The number of possible states is too large to be stored, so some form
of generalization has to be used.
Where exact methods store the function values in a table with an
entry for each state/action pair, for
large state spaces  function approximation must be used. Artificial
neural networks are popular methods.

Neural networks, or function approximators, are parameterized by a
vector of weights $\theta$. The length of the vector  is much less
than the number of  possible states. States are {\em represented\/}
by the weight vector. This has two advantages. First, the examples
from the very large
number of possible states can be stored efficiently in limited
memory. Second, the example states are generalized according to some
method, recording the generalization into the weight parameters,
allowing unseen states to be recognized as belonging to some
(hopefully appropriate) value or class.

In classification or other statistical tasks where unseen states must
be classified or evaluated, generalization is  a goal by
itself. Through generalization the ``essence'' of the states is to be
found. Generalization transforms a high-dimensional space to a lower-dimensional space.

In Chap.~\ref{chap:dqn} approximation methods will be introduced. Deep
Q-networks (DQN) will be discussed. DQN consists of a deep convolutional
network for function approximation, which is  trained by reinforcement
learning  using Q-learning (see page~\pageref{qlearning}).

\subsection{Exploration and
  Exploitation}\index{exploration}\index{exploitation}\label{sec:epsilon}
Now that we know how to calculate the expected value of a state, we
should look at how to do so efficiently.
For each action to be taken in a state, let us maintain a table of numbers that
represent the probability of winning for that action. As actions
are performed, the value in the table begins to reflect more reliably the
expectation of winning when taking that action.

Now the question is: Which action should we take? One approach is
to always choose the action that has accrued  the highest winning
probability so far. This
approach is called a {\em greedy\/} approach. Another approach is to sometimes try
another action, temporarily ignoring the one with currently the highest winning probability,
exploring a new successor state, in the hope of finding an untried action with an
even higher overall winning probability.

The $\epsilon$-greedy approach is to mostly try  the best (greedy) action
except to explore an $\epsilon$ fraction of times a randomly selected
other action.\index{$\epsilon$-greedy} If $\epsilon = 0.1$ then 90\%
of the times the best action is taken, and 10\% of the times a random
other action. 
The $\epsilon$-greedy
policy is an example of a {\em soft\/} policy: a policy with a finite
probability of selecting any of the possible actions. 

This choice between greedily exploiting known information and
exploring unknown nodes
to gain new information is called the exploration/exploitation
trade-off. It is a central concept in reinforcement
learning.

Let us look at the exploration/exploitation trade-off in more
intuitive and concrete
terms.
Assume you have moved into a new neighborhood, which you do not yet know,
and you want to find the route to the supermarket. After some trial
and error, you have found a route, which you  meticulously
memorize. The greedy approach is then, when you need to go to the
supermarket again, to always follow this same route. You exploit the
knowledge that you have to the fullest, without ever investing the
cost of trying out a
possible new route (exploring). Trying out a different route will take
some extra time and effort, but might pay off in finding a shorter
route, which you can then use many times for as long as you live in
the neighborhood.

Clearly, always exploiting might be the safe way, but it is likely to not
find an optimal solution. How much to explore, and in what situations, is a
fundamental topic in reinforcement learning that has been studied in
depth in the optimization
literature~\cite{holland1975adaptation,witten1976apparent} and in
multi-armed bandit 
problems~\cite{auer2002using,lai1985asymptotically} (see also Chapter~\ref{chap:trial}).\index{bandit theory} 

Smart use of exploitation and exploration is at the basis of the
breakthroughs in Go.
In Chap.~\ref{chap:trial} we will discuss in depth the Monte Carlo Tree Search
algorithm, MCTS, whose selection rule UCT uses 
bandit theory to find a good exploration/exploitation trade-off.

\subsubsection*{Temporal Difference}\index{temporal difference}\index{TD}
We  have now discussed how to calculate the values of states when the
policy function is known and how to then find the optimal policy for
states.
We are building up our tools to construct
solution methods with which we can estimate the optimal policy and
value of a game state, allowing us to construct a game playing program
that can compute the best moves so that it can play a full game.

The Bellman equation calculates the value of states when a model $P$
of the problem is known. For model-free problems (Sect.~\ref{sec:modelbased}) we need another
solution method, that can work by sampling.

Perhaps the best known solution method for estimating  policy functions
is temporal difference
learning~\cite{sutton1988learning}. This method is  the basis of
TD-Gammon~\cite{tesauro1995temporal}.  The temporal difference in the
name refers to the difference that it calculates between two time
steps, which it uses to calculate the  value of the new time step.

 The Bellman equation describes how  $V(s)$ can
  be calculated  using dynamic programming when the transition function
  $P$ is known. It is a model-based planning method. TD is different: under the
  policy $\pi$ it samples from the environment to
  adjust the current $V(s)$ estimate using a learning rate. TD does
  not need to know the transition function $P$ and  finds $V(s)$  by sampling.
  
TD works by updating the  current estimate of the state $V(s)$  with an
error value based on the estimate of the next state. It is, therefore,
a
bootstrapping method, where the update takes into account previous
estimates: 

$$V(s)\leftarrow V(s)+\alpha[R^\prime+\gamma V(s^\prime)-V(s)] .$$

Here $s$ is the current state, $s^\prime$ the new state, and
$R^\prime$ the reward of the new state. Note the introduction of $\alpha$, the learning rate. The $\gamma$
parameter is the discount rate.\index{$\alpha$, TD learning
  rate}\index{$\gamma$, TD discount rate} The last term $-V(s)$
subtracts the value of the current state, so that TD computes the
temporal difference. Another way to write the update rule is $V(s)
\leftarrow \alpha[R^\prime+\gamma V(s^\prime)] + (1-\alpha)V(s)$ as
the difference between the new TD target and the old value. Note the
absence of $P$ in the formula; TD is model-free.

Now, in order to arrive at a full reinforcement learning algorithm, we
also need a control method. TD gives us a learning
rule for value updates, the error part of trial and error. For a full
reinforcement learning 
algorithm  we also need a control method, which
tells us which nodes to choose for expansion, the trial part of trial
and error. Two such control methods
are  SARSA,\footnote{The name of the SARSA algorithm is a play on the MDP symbols as
  they occur in the action value update formula: $s, a, r, s, a$.}
yielding on-policy TD control, and Q-learning, yielding off-policy TD
control. SARSA and Q-learning are discussed more fully  in
Sect.~\ref{sec:sarsa}.



\subsection{Actor-Critic}\index{actor-critic}\label{sec:actor-critic}
SARSA and Q-learning learn the value function of
a state in order to find the optimal policy. Value function methods
work well when there is a finite set of actions. They work less well
in continuous or stochastic environments, such as robotics, where arm
movements or distance to drive are continuous choices.

In addition to finding the value function, to then find the best action, it is also possible to try
to compute the optimal policy directly, without first going through the
intermediate step of calculating
values. Williams~\cite{williams1992simple} introduced the REINFORCE
policy gradient algorithm. Direct policy evaluation can be more
efficient in continuous or stochastic environments and when there are many
different reward values. Converging to the value may then take a long
time.

A first challenge for policy gradient algorithms is
how to find reliable information on which action is best. Of course
there is 
the reward function, which is known in end states.  However, when it is not
known in intermediate states,  the more
intermediate steps have to be taken between 
an action and a reward, and the more unreliable this end signal
is. Statistics can be called to the rescue, by taking many samples,
which is computationally problematic. In large
problems, policy gradient methods may suffer from low sample efficiency,
since many samples are needed for reliable reward
information~\cite{sutton2018introduction,grondman2012survey}.

At this point you may ask if it would be possible to combine the
value-based approach and the policy-based approach. After all, value
functions can  backup  reward functions reliably. If reward
values are unreliable in large problems, might  value
function methods be used to provide a more reliable signal for the
value of the actions? Indeed this is possible, and this approach is
called
Actor-Critic~\cite{sutton1984temporal,sutton2000policy}. The
actor is the policy method; the critic is the 
value method. Each uses its own separate method, typically a
  neural
network. The critic network calculates the action value function for
each state, and
the actor the policy
function for each state. In Actor-Critic, value functions (not unlike
SARSA or Q-learning) are used 
to calculate more reliable values of states or actions more efficiently than sampling
many rewards, in large problems.

 \begin{figure}[t]
 \begin{center}
   \begin{tikzpicture}
  \begin{scope}[blend group = soft light]
    \draw[black,very thick] (210:1.2) circle (2);
    \draw[black,very thick]  (330:1.2) circle (2);
  \end{scope}
  \node at ( 410:0.9)   {Policy};
  \node at ( 150:1.4)   {Value function};
  \node at ( 220:2.2)   {Value-based};
  \node at ( 320:2.2)   {Policy-based};
  \node at ( 40:0) [font=\large] {Actor };
  \node at ( 270:1) [font=\large] { Critic};
\end{tikzpicture}
 \caption{Value and policy in Actor-Critic}\label{fig:actor-critic}
 \end{center}
 \end{figure}
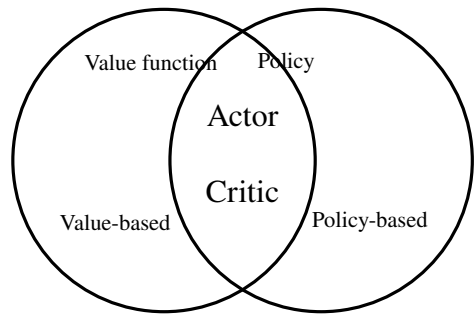


Actor-Critic 
interleaves a policy function (actor) and a
value function (critic). These algorithms have been studied widely~\cite{konda2000actor,kimura2000analysis,sutton2018introduction,vamvoudakis2010online,konda1999actor,o2016combining}. 
This combination of the value/policy idea is shown in Fig.~\ref{fig:actor-critic}.

SARSA and Q-learning do not have an explicit policy representation;
they are critic-only algorithms. 
%
Actor-Critic alternates policy evaluation with policy improvement
steps.
Actor-Critic methods combine the best of both worlds: better
convergence than value methods, and better policies due to value
learning and reduced variance. Actor-Critic methods are quite
popular. Two well-known methods, asynchronous advantage Actor-Critic
(A3C) and proximal policy optimiation (PPO), will be introduced in
Sect.~\ref{sec:a3c}, where we will discuss  advanced actor-critic
methods in the 
context of function approximation---A3C is an improvement over the
popular  Deep
Q-Network (DQN) algorithm.


 \begin{figure}[t]
 \begin{center}
 \begin{tikzpicture}[->,scale=1]
   \node (i) at (2,2)  {\bf Value/Policy};
   \node (j) at (3,0) {\bf Experience};
   \node (k) at (0,0) {\bf Model};


   \draw[bend left,->,very thick]  (i) to node [auto] {acting} (j);
   \draw[bend left,->,very thick]  (j) to node [auto] {model learning} (k);
   \draw[bend left=60,->,very thick,in=135]  (k) to node [auto] {planning} (i);
   \draw[bend left,->,very thick]  (j) to node [auto] {direct RL} (i);
\end{tikzpicture}
 \caption{Model-free and model-based methods~\cite{sutton2018introduction}}\label{fig:modelfreemodelbased}
 \end{center}
 \end{figure}
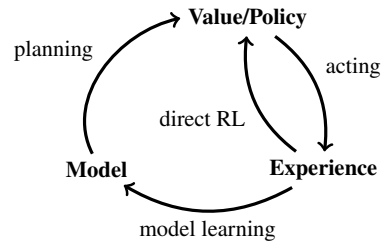
\subsection{Model-Free and
  Model-Based Methods}\index{model-based}\index{model-free}\label{sec:modelbased}
A basic distinction in reinforcement learning  is between
model-free and model-based methods (such as temporal difference and
value iteration)~\cite{kaelbling1996reinforcement}. Model-free methods learn a value
and policy function directly from the observations (Fig.~\ref{fig:modelfreemodelbased}, the inner loop). Model-based methods, on the
other hand, first need an intermediate model of the transition
probabilities. This model can be given by the rules of the problem, or
it can be learned. To use the model then  another reinforcement learning algorithm is  used
to find the value and policy function, often an exact planning
algorithm~\cite{kober2013reinforcement,wang2019benchmarking,plaat2020model}.  Model-based methods follow the outer loop in the
figure. Model-free methods are also known as direct methods. 

Some reinforcement learning methods are model-based,
 such as dynamic
programming, and others are model-free, such as
TD. Model-based methods can be used when a
model of the environment exists, when transition and reward
functions are known. This can be a statistical model, a
simulator, or the rules for the dynamics of the environment, such as
we have in games.

When a model is available, it can be used to do
look-ahead planning, as we shall do in the next chapters on
planning. When no model is available,  methods must be used to first
learn a model of the environment, often using sampling. The chapters on learning discuss
deep reinforcement learning methods that start with no model and
create a model
. 
Finally, combinations of
model-based and model-free methods, of planning and training, are
discussed. AlphaGo is perhaps the best known example of this combined 
approach~\cite{silver2016mastering}. Section~\ref{sec:further} further discusses
model-free and model-based reinforcement learning.

Let us now look at another central concept in reinforcement learning:
that of on-policy and off-policy learning methods for model-free learning.

\subsection{On-Policy Learning and Off-Policy Learning}\index{off-policy
  learning}\index{on-policy learning}\label{sec:off-policy}
We will now discuss two model-free solution methods, methods that 
work when the transition function of the MDP
is not known.

Reinforcement learning is concerned with learning a policy from
actions and rewards. The
agent selects an action, and learns from the reward that it gets back
from the environment; or, in  terms of the Tic Tac Toe tree of
Fig.~\ref{fig:ttt-tree}, the value of the root is learned by
downward selecting an action, and by backing up the value that is
returned. Note that the policy determines the downward action
selection, and the subsequent learning takes place in the backup of
the reward value.

Normally, the learning  takes place by backing
up the value of the action that was selected by the policy: the
learning is {\em on-policy}. It turns out that there is an alternative. If the learning takes
place by backing up values of another action, not the one selected by
the policy, than this is known as {\em off-policy learning}. This makes
sense when the policy explores by selecting a non-optimal 
action (i.e., it does not perform greedy exploitation). On-policy learning would then have to backup the value of the
non-optimal exploration action (since otherwise it would not learn on-policy). Off-policy learning,
however, is free to backup the value of the best action instead, and
not the inferior one
selected by the exploration policy. Thus, in the case of exploration,
off-policy learning can be more 
efficient, by not stubbornly backing up the value of the action
selected by the policy, but of the best action instead. (In the case
of an exploiting step by the policy, on-policy learning and off-policy
learning are the same.) 

All learning methods face a trade-off: they try to learn the best
target action
from current behavior that is known so far (exploitation), and they  choose
new (most likely non-optimal) behavior in order to explore and be able to
find actions that 
are  better than the current best actions. Thus, in principle, there
are two policies: first, the behavior policy that is used for
actual action selection behavior (sometimes exploring), and second, the target policy that is learned by backing
up values. The first policy performs \emph{downward} selection to expand states, the second
\emph{upward} value propagation to update the target policy. 

Recall that the $Q(s,a)$
function computes the expected value of (downward) selecting action $a$ in state
$s$.
On-policy learning and off-policy learning update the $Q(s,a)$
function by backing up values (upward) from their successor actions (which can be implemented as an array, or lookup table). When
the algorithms sample the state/action pairs often enough, the algorithms are
proven to converge to the correct values.

\subsubsection*{On-Policy}
In on-policy learning a single policy function is used for (downward) action
selection and for (upward) value backup towards the learning target.
On-policy learning updates values directly on this single 
policy. The same policy array is used for exploration behavior
and for the incumbent optimal target policy.

The update formula is
$$Q(s_t,a_t)\leftarrow Q(s_t,a_t)+\alpha[r_{t+1}+\gamma
Q(s_{t+1},a_{t+1})-Q(s_t,a_t)] .$$

On-policy learning selects an action, evaluates it, and moves on to
better actions. A well-known on-policy approach is SARSA. On-policy
learning starts with a starting  policy, 
samples the state space with this policy, and improves the
policy by backing up values of the selected actions. Note that the
term $Q(s_{t+1}, a_{t+1})$ can also be written as
$Q(s_{t+1},\pi(s_{t+1}))$, highlighting the difference with off-policy
learning.
Since SARSA is on-policy,  it updates its Q-values
using the Q-value of the next state $s$ and the current policy's
action. 
In a short while we will look at SARSA example code, and
compare its behavior with off-policy learning.

The primary advantage of on-policy learning is that it directly
optimizes the target of interest, and achieves stable learning. The
biggest drawback is sample 
inefficiency, since policies are estimated from the full rollouts. This may further cause large variance---if it goes in the wrong
direction, there is no 
stabilizing other information to get it
out.\footnote{See~\cite{nachum2017bridging} for a discussion on variance reduction.
To avoid large bias, SARSA can use on-policy
data, or updates can be  slowed.} 

\subsubsection*{Off-Policy}
Off-policy learning is more complicated. It uses two separate
policy arrays: one for exploratory downward selection behavior,
and one to update as the current target backup policy. Learning (backing up) is from data
{\em off\/} the downward selection policy, and the 
whole method is therefore called off-policy learning.

The update formula is
$$Q(s_t,a_t)\leftarrow Q(s_t,a_t)+\alpha[r_{t+1}+\gamma \max_a
Q(s_{t+1},a)-Q(s_t,a_t)] .$$

The difference from on-policy learning is that the $\gamma
Q(s_{t+1},a_{t+1})$ term has been replaced by $\gamma \max_a
Q(s_{t+1},a)$. The learning is from backup values of the best,
not the actual, action.

The reason that Q-learning is off-policy is that it updates its
Q-values using the Q-value of the next state $s$ and the greedy action
(not necessarily the selection policy's action---it is learning off the
selection policy).

In this sense, off-policy learning collects all available information and uses it
simultaneously to construct the best target policy. The best known off-policy
algorithm is Q-learning~\cite{watkins1989learning}.\label{qlearning}
It gathers information from (partially) 
exploring moves, it evaluates states as if a greedy policy was used, and
it slowly reduces random exploration. 

\subsubsection*{Differentiating On-Policy and Off-Policy Learning}
To understand the difference between on-policy and off-policy learning,
let us look at the update formulas. A way to differentiate on-policy and
off-policy learning is to look how the backup formula depends on the
next action 
$a_{t+1}$, which can be written as $\pi(s_{t+1})$. Q-learning is off-policy learning since it does not depend
on the next action $a_{t+1}$, i.e., it does not depend on the policy $\pi$. No
matter what $a_{t+1}$ is, the $\max_a Q(s_{t+1}, a)$ part does not
depend on $a_{t+1}$.   $Q(s_t, a_t)$ is updated regardless of
the future policy $\pi$ starting from $s_{t+1}$. On the other hand, SARSA is on-policy
learning because the term $Q(s_{t+1}, a_{t+1})$ in $Q(s_t, a_t)$ depends
on the next actual action $a_{t+1}$, the one taken by following the
policy $\pi(s_{t+1})$. The backpropagation learning in SARSA  depends on the behavior policy.

To recapitulate, off-policy approaches can learn the value of the
optimal target policy
regardless of the selection policy, while in on-policy methods the
agent learns the value of the policy whose actions it is following.

Off-policy learning is especially important when there is a database
of previously stored trajectories (i.e., data in the form of tuples
$(s,a,r',s')$). This data has been collected by previously applying some
policy, and cannot be changed. This is a common case, for example for
medical problems. To use this kind of data, only off-policy methods can be
used. 

Off-policy methods are more flexible in the type of problems they can
be used for. Their theoretical  properties, however,
  are different. If we compare Q-learning with SARSA, the
difference  is in the $\max$ operator used in the Q-learning
update rule. The $\max$ operator is  nonlinear, which can make it more
difficult to combine the algorithm with function approximators. With
function approximators (neural nets), on-policy methods are usually
more stable. 

Off-policy learning can be problematic since the Q-values tend to
overestimate (because of the max operator). Off-policy learning that 
strays too far off the selection policy may get lost (or rather,
become inefficient).
In Chap.~\ref{chap:dqn} we will cover learning
stability of off-policy methods in more depth, including methods that
limit the amount of off-policy learning.

We will now look at an underlying topic that permeates all of machine
learning and that has been mentioned a few times already:
sample efficiency.

\subsection{Sample Efficiency}\index{sample efficiency}\label{sec:sample}
An algorithm is sample efficient if it gets the most out of every
sample and efficiently learns a function with only few
samples of the environment.

Model-free methods often  require many environment samples to converge
on the policy. A problem of model-free methods is that once a sample
has been used, it is thrown away, and a new sample is
sought.

Model-based algorithms use the samples of
the environment more efficiently. They use the samples to build a
transition model of the environment, which is then used internally to 
find the policy. Model-based methods are often more sample efficient, using
the model to converge quickly to an optimum with few samples, provided
that the predictions from the transition model are
good~\cite{wang2019benchmarking,moerland2020model,plaat2020model}. Constructing
accurate transitions models can be difficult in practice,  and most
good results so far have been achieved with model-free methods (unless a
transition model is provided  by the application domain).

SARSA and Q-learning, two model-free methods, are effectively performing random
search in the environment, not using information from previous samples
to guide the search.
A drawback of on-policy learning is that it samples from its own
target policy, which may lead to myopia, or tunnel vision, sampling  a local
part of the search space.  When an
algorithm learns only a part of the search space, learning can be unstable.

In off-policy learning old samples may be used, not related to the
target policy. This may prevent myopia, but on the other hand not using all samples of
the behavior policy may
also lead to low sample efficiency.

Several methods have been proposed to improve sample efficiency of
model-free methods. One
such  method is importance
sampling~\cite{glynn1989importance,munos2016safe,wang2015dueling},
which  samples from a distribution that overweights
the important region. In this way the target is sampled heavily (as in
on-policy) but the rest is not neglected completely (as in off-policy,
preventing myopia).

\section{Conclusion}
We have now discussed in depth many elements of the reinforcement learning
paradigm, one of the major paradigms of machine learning. To conclude
this chapter, let us 
discuss some differences from  supervised learning.

\subsubsection*{Supervised Learning}
Machine learning is the part of artificial intelligence
that studies mechanical learning principles.
Machine learning algorithms are algorithms  to
learn  patterns from data~\cite{bishop2006pattern,russell2016artificial}.  

In its most basic classification form, machine learning works by
looking at pairs of
examples and labels $(E, L)$.  The task of the learning
algorithm is then to associate 
the correct label with the correct class of examples, so that when it is shown a new example,
it classifies it with the correct label. A well-known machine learning
problem is to learn the classification of a large sequence of pictures
(examples $E$) as either dogs
or cats (labels $L$).  Such a problem  is an example of supervised
learning where  the labels play the role of supervisor in  the
learning process. If not all
examples have labels, then supervised learning is only
possible for the examples with labels. Supervised learning needs a
database of (example, label) pairs. It will be treated in depth in Chap.~\ref{chap:super}.

\subsubsection*{Interaction}
Note that trial and error (or {\em interaction\/}) is an essential aspect of learning by
reinforcement. Like supervised learning, reinforcement learning is a
kind of machine learning. But unlike supervised learning, in reinforcement
learning there is no preexisting database with labeled states. All
states to learn from will be generated during the learning process by
interaction with the environment. By
smartly choosing the agent's actions, the sequence of states may be
generated to learn efficiently, without the need for  a large
database of examples.

Furthermore,  in supervised learning we
can
never learn more than the examples in the supervisory database provide; the teacher
provides an upper limit to what the pupil can learn. In
reinforcement learning there is no teacher holding us back; only the
world to explore.

\subsubsection*{Credit Assignment}
Now, in supervised learning all examples are labeled. However, in
reinforcement learning some actions may not  return a reward, but future
rewards must be propagated to previous states.  In other words,  state spaces in
reinforcement learning may have a {\em sparse\/} reward
structure. This is the case in most games, where only terminal states
have a reward $\{ -1, 0, +1\}$, and for all other states the
reward value must be propagated backward from the terminal
states. This is known
as the credit assignment problem. Long-range credit assignment
 is a challenging problem. This too will be discussed in Chap.~\ref{chap:dqn}.

\subsubsection*{Dependency}
There are more differences between supervised learning and
reinforcement learning.
Supervised learning has a database from which it
draws examples to learn from; reinforcement learning generates its own learning examples
through interaction with the environment. In reinforcement learning
there is therefore a dependency between subsequent samples, since the learning algorithm
influences its own learning examples. This may cause learning
anomalies, cycles, and local maxima. Special care must be taken in
reinforcement learning to not get stuck in local maxima, as we will
see in Chap.~\ref{chap:dqn}.

\begin{table}[t]
  \begin{center}
\begin{tabular}{lccl}
  {\bf Chapter} & {\bf Name} & {\bf MDP-tuple} & {\bf Reinforcement learning}\\ \hline\hline
    Chap.~\ref{chap:search} & alpha-beta & $(S, A, 1, R, 1)$ &
                                                                   policy, backup \\ 
   Chap.~\ref{chap:trial} & MCTS & $(S, A, P, R, 1)$
  & pol, b/u, exploration/exploitation\\ 
   Chap.~\ref{chap:dqn} & DQN  &  $(S, A, P, R, \gamma$)& pol, b/u,
                                                             expl/expl,
                                                             discount, off-policy
  \\ 
   Chap.~\ref{chap:plan} & self-play & $(S, A, P, R, \gamma)$&pol, b/u,
                                                             expl/expl,
                                                                discount,
                                                                off-pol, self-play \\ \hline
\end{tabular}

  \caption{MDP tuple and reinforcement learning in the chapters}\label{tab:mdp}
  \end{center}
\end{table}

\subsubsection*{MDP Tuple and Reinforcement Learning}
We are approaching the end of this chapter. Let us look at how the
principles that were covered apply to the different chapters further
on in this book.

Reinforcement learning can formally be described by the 5-tuple of
Markov decision processes.
The remainder of this book will discuss many different reinforcement learning
methods in depth, from heuristic planning, adaptive sampling, to
generalization (as we saw in Table~\ref{tab:heursampgen}). These
methods  will start simple, with some of the MDP tuples being
constant at 1, in the early chapters. As the methods become more elaborate, the full
tuple is used, and more reinforcement learning elements apply. 
Table~\ref{tab:mdp} gives an overview of how the full generality of
reinforcement learning develops throughout the chapters of this book, as the
methods progress from simple search to advanced generalization methods with adaptive self-learning.

\section{Practice}\index{SARSA}\index{Q-learning}\label{sec:sarsa}\index{OpenAI
  Gym}\index{Gym}
It is time to try the insights that we have discussed in practice with
programming exercises. 
First  are some questions to check your understanding of this
chapter and the previous chapter. Each question is a closed question where a simple, one 
sentence answer is possible.
\subsubsection*{Questions}
\begin{enumerate}
\item Name three important elements of intelligence.
\item What element of intelligence does computer game playing focus
  on?
\item Describe the symbolic approach to AI. Describe the
  connectionist approach to AI.
\item Who wrote the first Chess program?
\item When did Deep Blue beat Kasparov, in 1996 in Philadelphia or
  in 1997 in New York?
\item Describe the difference between strategy and tactics in a
  board game. Give an example game of each.
\item Describe the reinforcement learning model.
\item Give the 5-tuple of a Markov decision process.
\item What is a value function?
\item What is a policy function?
\item Describe in words the intuition of Bellman's equation.
\item Describe the exploration/exploitation dilemma. Can you give one
  simple algorithm for trading off exploration and exploitation?
\item Describe in words the intuition behind temporal difference learning.
\item Why is it sometimes necessary to approximate a solution?
\item What is the difference between on-policy learning and
  off-policy learning? Can you give one example algorithm for each?
\item Why is sample efficiency important?
\end{enumerate}


\subsection{Algorithms}
Next, we will  look at on-policy and off-policy
algorithms and  their behavior.
How do on-policy learning and off-policy learning 
learn optimal policies for the Taxi environment? We will use SARSA as
the on-policy learner, and Q-learning as the off-policy learner.

As we saw in the
previous chapter, benchmarks are of great importance
for progress in AI. To support AI progress, the organization OpenAI has provided an
easy-to-use suite of benchmarks for scientists and students to use. It
is called Gym, and has a Python interface. One of the
classic examples in reinforcement learning is the Taxi problem,  introduced by 
Dietterich~\cite{dietterich2000hierarchical}. We will use the Taxi example from OpenAI Gym.

Python is a programming language that is popular in artificial
intelligence. It supports quick development, rich and flexible 
data structures, and many third-party packages have been written, ranging from numerical
simulation, graphics, to machine learning. If Python is not present on your
computer, please go to
Appendix~\ref{chap:python} to learn more about Python and how to
install it on your computer. 

Now please install Gym. Go to the webpage and the
GitHub page of OpenAI Gym and  have a look around to see what is
there. OpenAI Gym can be found
\href{https://gym.openai.com}{here},\footnote{\url{https://gym.openai.com}}
and the Gym GitHub page can be found  \href{https://github.com/openai/gym}{here}.\footnote{\url{https://github.com/openai/gym}}
You will see different sets of environments, from easy to
advanced. There are the classics, such as Cartpole and Mountain
Car. There are also small text environments. Taxi is there, and there is
the Atari Learning Environment~\cite{bellemare2013arcade}, which was
used in the paper that introduced DQN~\cite{mnih2013playing}. 
MuJoCo is also available, an environment for experimentation with simulated robotics.
(Our Taxi environment can be found in \verb|gym/envs/toy_text|, in case
you would like to have a look at how the environment is written.)
Installing Gym on your computer is  easy. Type
\verb|pip install gym| (or \verb|pip3 install gym|) at a command
prompt.  You should also install numpy and matplotlib.
Go to Spinning Up to get started with Gym.

\begin{figure}[t]
\begin{center}
\includegraphics[width=5cm]{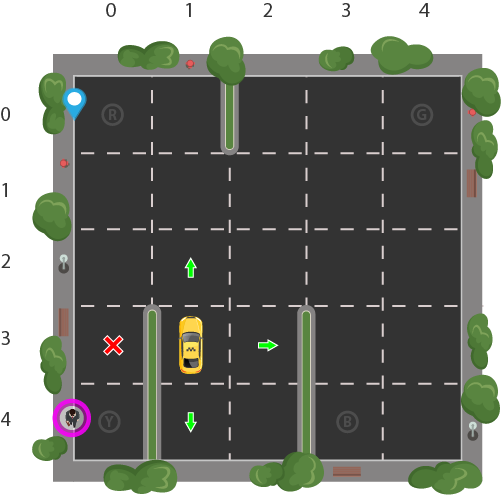}
\caption{Taxi world~\cite{learn}}\label{fig:taxi}
\end{center}
\end{figure}

The Taxi example 
(Fig.~\ref{fig:taxi}) is 
an environment where taxis move up, down, left, and right, and can
pickup and drop off passengers. 
The Gym documentation describes the Taxi world as follows.
There are four designated locations in the grid world indicated by 
R(ed), B(lue), G(reen), and Y(ellow). When the episode starts, the
taxi starts off at a random square and the passenger is at a 
random location. The taxi drives to the passenger's location, picks
up the passenger, drives to the passenger's destination (another
one of the four specified locations), and then drops off the
passenger. Once the passenger is dropped off, the episode ends. 

There are 500 discrete states, since there are 25 taxi positions, five
possible locations of the passenger (including the case when the
passenger is in the taxi), and 4 destination locations $(25\times
5\times 4)$.  

The environment returns  a new result tuple at each step.  There are six
discrete deterministic actions for 
the Taxi driver:
\begin{itemize}
\item[0:] Move south
\item[1:] Move north
\item[2:] Move east 
\item[3:] Move west 
\item[4:] Pick up passenger
\item[5:] Drop off passenger
\end{itemize}    

There is a reward of $-1$ for each action and an additional reward of
$+20$ for delivering the passenger. There is a reward of $-10$ for
executing actions {\em pickup\/} and {\em dropoff\/} illegally.
Please refer to
Listing~\ref{lst:evaltaxi}.

OpenAI not only provides environments, but
also the other half of reinforcement learning, the agent algorithms. It provides baseline
algorithms for learning policies of agent behavior.  Code of all well-known reinforcement
learning algorithms is present
\href{https://github.com/openai/baselines}{here}\footnote{\url{https://github.com/openai/baselines}}
and
\href{https://stable-baselines.readthedocs.io/en/master/}{here}.\footnote{\url{https://stable-baselines.readthedocs.io/en/master/}}

\subsubsection*{SARSA}\index{SARSA}
SARSA is the on-policy learning algorithm that we will use first. Its name comes
from the tuple $(s, a, r^\prime, s^\prime, a^\prime)$ that determines the
next value of the $Q(s,a)$-function. We recall the on-policy update formula: 

$$Q(s_t,a_t)\leftarrow Q(s_t,a_t)+\alpha[r_{t+1}+\gamma
Q(s_{t+1},a_{t+1})-Q(s_t,a_t)] .$$

\lstset{label={lst:sarsa}}
\lstset{caption={SARSA Taxi example, after~\cite{learn}}}
\begin{lstlisting}[language=Python,float]
# SARSA for OpenAI Gym Taxi environment
import gym
import numpy as np
import random
#Environment Setup
env = gym.make("Taxi-v2")
env.reset()
env.render()
# Q[state,action] table implementation
Q = np.zeros([env.observation_space.n, env.action_space.n])
gamma = 0.7   # discount factor
alpha = 0.2   # learning rate
epsilon = 0.1 # epsilon greedy
for episode in range(1000):
    done = False
    total_reward = 0
    current_state = env.reset()
    if random.uniform(0, 1) < epsilon:
        current_action = env.action_space.sample() # Explore state space
    else:
        current_action = np.argmax(Q[current_state]) # Exploit learned values        
    while not done:
        next_state, reward, done, info = env.step(current_action) # invoke Gym
        if random.uniform(0, 1) < epsilon:
            next_action = env.action_space.sample() # Explore state space
        else:
            next_action = np.argmax(Q[next_state]) # Exploit learned values        
        sarsa_value = Q[next_state,next_action]
        old_value = Q[current_state,current_action]
       
        new_value = old_value + alpha * (reward + gamma * sarsa_value - old_value) 

        Q[current_state,current_action] = new_value
        total_reward += reward
        current_state = next_state    
        current_action = next_action    
    if episode % 100 == 0:
        print("Episode {} Total Reward: {}".format(episode,total_reward))
\end{lstlisting}

Listing~\ref{lst:sarsa} shows Python code for improving a 
policy in the Taxi world with the SARSA algorithm, adapted from~\cite{learn}.
The best policy found by SARSA can now be used by following the best action
values in the Q-table. 
When we  evaluate the performance of our SARSA agent, we do not need
to explore  any more, since the best action is right there, in the Q-table.

\lstset{label={lst:evaltaxi}}
\lstset{caption={Evaluate the optimal SARSA Taxi result, after~\cite{learn}}}
\begin{lstlisting}[language=Python,float]
total_epochs, total_penalties = 0, 0
ep = 100
for _ in range(ep):
    state = env.reset()
    epochs, penalties, reward = 0, 0, 0
    done = False
    while not done:
        action = np.argmax(Q[state])
        state, reward, done, info = env.step(action)
        if reward == -10:
            penalties += 1
        epochs += 1
    total_penalties += penalties
    total_epochs += epochs
print(f"Results after {ep} episodes:")
print(f"Average timesteps per episode: {total_epochs / ep}")
print(f"Average penalties per episode: {total_penalties / ep}")
\end{lstlisting}

Listing~\ref{lst:evaltaxi} shows the simple code to evaluate the
SARSA Taxi policy. The number of illegal pickups/dropoffs is shown
as penalty.

\subsubsection*{Q-Learning}\index{Q-learning}
Q-learning performs off-policy learning. We recall its update
formula:

$$Q(s_t,a_t)\leftarrow Q(s_t,a_t)+\alpha[r_{t+1}+\gamma \max_a
Q(s_{t+1},a)-Q(s_t,a_t)] .$$

\lstset{label={lst:q}}
\lstset{caption={Q-learning Taxi example, after~\cite{learn}}}
\begin{lstlisting}[language=Python,float]
# Q learning for OpenAI Gym Taxi environment
import gym
import numpy as np
import random
#Environment Setup
env = gym.make("Taxi-v2")
env.reset()
env.render()
# Q[state,action] table implementation
Q = np.zeros([env.observation_space.n, env.action_space.n])
gamma = 0.7   # discount factor
alpha = 0.2   # learning rate
epsilon = 0.1 # epsilon greedy
for episode in range(1000):
    done = False
    total_reward = 0
    state = env.reset()
    while not done:
        if random.uniform(0, 1) < epsilon:
            action = env.action_space.sample() # Explore state space
        else:
            action = np.argmax(Q[state]) # Exploit learned values        
        next_state, reward, done, info = env.step(action) # invoke Gym
        next_max = np.max(Q[next_state])
        old_value = Q[state,action]
       
        new_value = old_value + alpha * (reward + gamma * next_max - old_value) 

        Q[state,action] = new_value
        total_reward += reward
        state = next_state    
    if episode % 100 == 0:
        print("Episode {} Total Reward: {}".format(episode,total_reward))

        
\end{lstlisting}

Listing~\ref{lst:q} shows Python code for finding a 
policy in the Taxi world with the Q-learning algorithm.

Again, Listing~\ref{lst:evaltaxi} can be used to evaluate the
Q-learning Taxi policy. 
On average, Q-learning tends to find better policies than SARSA~\cite{sutton2018introduction}.

\subsubsection*{Exercises}
\begin{enumerate}
\item Experiment with Taxi world. Download Gym, and perform the
 SARSA and Q-learning steps described before. Running ready-made SARSA and
Q-learning code is nice, 
but you get a better idea of how the Taxi world works by tracing a few steps
of the simulator by yourself. Do so and add code to allow you to print
(render) each step, step by step. Watch the taxi move in its world.

\item In Taxi world, which is better: SARSA or Q-learning? How do you
know? On which 
criteria do you measure algorithm quality?

\item The environment is initialized randomly each time. This makes it
hard to compare the two  algorithms if each run is different,
and it complicates debugging.
Make a deterministic version of the algorithm comparison, where each
starts  at the same configuration.

\item How many runs do you need to do to get statistically
significant results for $p<0.05$ (if necessary, see a basic text on
statistics or 
experiment design such as~\cite{bartz2010experimental})?

\item Experiment with different values for $\epsilon$, $\alpha$, and $\gamma$. Which are
better?  Experiment with decaying values for
$\epsilon$, $\alpha$, and $\gamma$.

\item Go to Gym, and look for the Mountain Car example. 
Compare SARSA and Q-learning. Which learns faster, and what about  the solution
quality?
\end{enumerate}

\section*{Summary}

This has been a challenging chapter, let us summmarize what we have seen. We have introduced many abstract
formal concepts from reinforcement learning, to lay the ground work 
for understanding the algorithms in the following chapters. At the end
you have been invited to experiment with the algorithms. 

We started with the introduction of the concepts of agent and
environment, and introduced 
states, actions, and rewards, which are the basis of the
agent-environment interaction. In reinforcement learning the agent learns
through interaction with an environment, by trial and error.
Markov decision processes are a powerful and popular
formalism. In processes with the Markov property, the distribution of
future states depends solely on the current state, not on the states
preceding it, i.e., there is no memory, which simplifies the
mathematical analysis of Markov processes.

Two related central elements in reinforcement learning are the value
of a state and the policy: the list of actions to be taken when in a state,
defining a future behavior (strategy). Policies can be implemented as
arrays, functions, mappings, or lists, and all these terms are used in
the literature. The Bellman equation is a central concept in
reinforcement learning. It recursively defines the value of a
state based on the 
reward of its successor states. Bellman introduced the concept of
dynamic programming for model-based reinforcement learning (planning). Other important algorithms that we
covered are  Temporal Difference learning, SARSA, and Q-learning for
model-free reinforcement learning.

Reinforcement learning algorithms typically gradually build up
information about the best actions to be taken in a state. Greedy
algorithms always exploit the available information by following the
action with the highest expected reward. Other algorithms may explore
new or under explored actions, with a larger uncertainty, to possibly
find better answers. A well-known simplistic  approach is the
$\epsilon$-greedy approach, which chooses to exploit in $1-\epsilon$
occurrences and
chooses to explore in $\epsilon$   times.

On-policy learning backs up state values for the current downward
behavior selection 
policy, off-policy learning backs up state reward values for the best policy
independent of the downward behavior policy it follows. Q-learning is a model-free
off-policy method.

We discussed  exact and approximate
methods. Exact methods are suitable for small state spaces;
approximate methods are used when the state space is very
large. In fact, the state space may be so large that almost all of the
states that we see at test time are new, and cannot have been trained
on. Approximation methods must be able to 
generalize well. Lookup tables   are well-known  
exact data structures; neural networks are well-known approximation
data structures. 



\section*{Historical and Bibliographical Notes}
The literature on reinforcement learning and games is rich and
varied. This section will provide more pointers to interesting works.

Game playing programs solve reinforcement learning problems. 
Reinforcement learning is an old and large and rapidly expanding field. This
chapter has only introduced the bare minimum. Two excellent
comprehensive treatments of the field are the books by Sutton and
Barto~\cite{sutton2018introduction} 
and Bertsekas and 
Tsitsiklis~\cite{bertsekas1996neuro}.  Sutton and
Barto divide reinforcement learning into exact tabular methods and approximate
methods. This distinction maps nicely to Kahneman's categories of
thinking slow and thinking fast, and Shannon and Turing's search-eval
architecture.



Q-learning was introduced by Watkins~\cite{watkins1989learning}.
%
Kaelbling provides a widely cited review of reinforcement
learning~\cite{kaelbling1996reinforcement}. Kaelbling has also contributed
greatly to robotics, planning, and hierarchical reinforcement learning
(Sect.~\ref{sec:hier}). She received the 1997 IJCAI Computers and
Thought award. 
For a review of
reinforcement learning in robotics see~\cite{kober2013reinforcement}.

A popular lecture series on reinforcement learning is by David Silver
\href{http://www0.cs.ucl.ac.uk/staff/d.silver/web/Teaching.html}{here}.\footnote{\url{http://www0.cs.ucl.ac.uk/staff/d.silver/web/Teaching.html}}
Lecture notes of Sergey Levine's course on deep reinforcement learning
at Berkeley are available \href{http://rail.eecs.berkeley.edu/deeprlcourse/}{here}.\footnote{\url{http://rail.eecs.berkeley.edu/deeprlcourse/}}

\chapter{Heuristic Planning}\label{chap:search}\label{chap:idtt}


Combinatorial games have been used in AI to study
reasoning and  decision making since the early days of AI. An important
challenge in decision making is
how to search  large state spaces
efficiently. The methods in this book are organized around four main
paradigms (and chapters). The first paradigm that we study for  decision making is
heuristic planning. It is a model-based paradigm.  Heuristic planning  uses a
human-inspired approach: it is believed that when
playing a board game such as Chess, most human players (1) try 
to look ahead a few moves in their mind, and then (2) try  to assess if they like the situation that
they arrive at, and play the move they liked best. Heuristic planning  formalizes this concept, using
a search function and a heuristic evaluation function. 

A heuristic is a domain-specific rule of thumb. As a rule of thumb, it works
most of the time, but not always. Being domain specific, a heuristic  is not
general; it only works  in certain games, often exploiting a
game-specific feature. Scientists, who
strive for general methods, have
a love-hate relationship with heuristics. They are often too
successful to ignore for a specific problem, but the domain specificity limits their general
applicability. Much of the efforts for generalization and feature discovery
 that have been so successful in deep learning (see
 Chap.~\ref{chap:dqn}) are driven by
the desire to transcend domain-specific heuristics.

A first implementation of the look-ahead idea is the minimax
algorithm. Minimax has been highly successful, especially since many
more or less general enhancements have been developed over the
years. We will cover some of these enhancements in depth, such as alpha-beta,
iterative deepening, and transposition tables.

Heuristic planning has been quite successful in  tactical games,
such as Chess, Checkers, and Othello. Planning and heuristics are basic
concepts of AI. Future chapters will introduce different  paradigms,
but often elements of good old heuristic planning  will appear.


\section*{Core Problems}
\begin{itemize}
\item How can we search a large state space efficiently?
\item How can search effort be directed to promising areas of the state space?
\end{itemize}

\section*{Core Concepts}
\begin{itemize}
\item Search function
\item Heuristic evaluation function
\item Critical tree, alpha-beta pruning, and move ordering
\end{itemize}
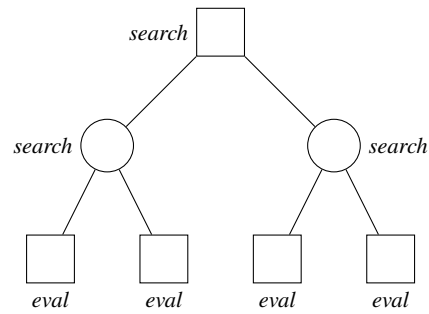
\begin{figure}[t]
\begin{center}
\begin{tikzpicture}
\tikzstyle{max node}=[rectangle,draw,inner sep=9]
\tikzstyle{min node}=[circle,draw,inner sep=7]
\tikzstyle{cut node}=[rectangle,draw,fill,inner sep=5]
\node[max node,label=left:{\em search}](0){}
          child[sibling distance=30mm]{node[min node,label=left:{\em search}]{}
                         child[sibling distance=15mm]{node[max
                             node,label=below:{\em eval}]{}}
                         child[sibling distance=15mm]{node[max
                             node,label=below:{\em eval}]{}}}
          child[sibling distance=30mm]{node[min node,label=right:{\em search}]{}
                         child[sibling distance=15mm]{node[max
                             node,label=below:{\em eval}]{}}
                         child[sibling distance=15mm]{node[max
                             node,label=below:{\em eval}]{}}};
\end{tikzpicture}
\caption{Search-Eval architecture in a tree}\label{fig:se-arch}
\end{center}
\end{figure}
%
First we will discuss the basis of heuristic planning: the search-eval
architecture. Then we delve deeper into computing the size of the state space,
and discuss basic search and evaluation functions. We continue with an
analysis to find the smallest part of the state space  must be traversed
to find the optimal policy. Finally, we discuss search enhancements,
such as alpha-beta and transposition tables, and evaluation function
enhancements, such as end-game databases.   

\section{The Search-Eval Architecture}
Let us start with the basic
architecture of heuristic planning.
Traditional Chess-style game playing programs consist of  a search function and an
evaluation function. Together these two functions form the search-eval
architecture. The transition function $P$ is known in these
games; all successor states can be generated in each board position by following
the game rules. Finding the optimal policy  is done by generating all
successor states and performing a look-ahead {\em search}. Searching the state space exhaustively is
infeasible. The size of the state space of Chess is $10^{47}$, too
large to search fully.
Therefore, after the initial state 
has been searched to a certain search depth,  the search tree is
chopped off. A heuristic evaluation function is
called before the end of the game at a fixed distance from the
root. See Fig.~\ref{fig:se-arch} for an illustration, where the root
node is shown at the top.

The root represents the initial state
$s_0$,  the  board position for which we wish to find the optimal policy, and
the value. One level deeper we see two inner nodes, the children of
the root, and at depth 2 we see four leaf nodes, which are not searched
further, but where the evaluation function is called to determine a heuristic
leaf value. In the square nodes it is the player's turn to move, and in the
circle nodes the opponent's. Square nodes are max nodes; circle nodes
are min nodes.

The search-eval architecture is a simple architecture. It consists of 
two functions: a search function and an evaluation function. The search
function traverses the states one by one. It is
an exact function. The evaluation function returns a heuristic
value of the state on which it is called. It is an approximate function.

Note that the introduction of a heuristic evaluation function changes
the range of values of the value function. The values for the full state 
space are determined by the terminal states, whose value can be
$\{-1,0,+1\}$ for loss, draw, and win  (as viewed by the first
player). A heuristic  returns more values, typically in a more
fine-grained range such as 
$[-16$,$000, +16$,$000]$. Positive values indicate an advantage for the
first player; negative values indicate a disadvantage. 
Section~\ref{sec:eval} provides more details on evaluation functions.

\subsubsection*{Exact Search and Approximate Eval}

Finding the optimal policy in a game of skill requires searching
through   the state space. For small state spaces, 
methods that traverse the states one by one are suitable; they give
precise answers. They are unsuitable for large 
state spaces, since traversal would take too long. The alternative, 
approximation methods, do not traverse the state space at
all. Instead, they   approximate the value of a single state
directly. Approximation methods  use combinations of {\em features\/} of the
state. Examples of features are heuristics such as material 
balance---which side has more and better pieces---or
mobility (Sect.~\ref{sec:eval}). In later chapters we will encounter other approximation
methods, such as sampling in Chap.~\ref{chap:trial} and
generalization by neural nets
in Chap.~\ref{chap:dqn}.

Exact methods are precise but slow.
Approximation methods are imprecise but fast.
The
search-eval architecture provides a framework for the two approaches
to work together and achieve the best of both worlds: exact search methods traverse a part of
the state space, calling the approximation methods to evaluate states
at their search horizon. Together they find a good approximation of the value
function without taking too long.  

The goal of exact and approximate methods is the same: to determine
the value function (and  the optimal policy).
Researchers have developed search and eval functions that together
are  strong enough to achieve a level of play that beats the best humans
in Chess and Checkers. The first ideas for the search-eval
architecture go back to Shannon~\cite{shannon1988programming} and
Turing~\cite{turing1953digital}.\index{search-eval architecture}

%
One could ask the question why search has to be used,  what is it that
the heuristics miss that search has to be used at all?
The reason that the search
function is necessary is that the evaluation
function misses game dynamics. An evaluation function provides a
static assessment of a state, but cannot foresee the effects of
dynamics caused by captures and other tactical play. Evaluation
functions only work well in quiet (stable) positions (see also
quiescence search in Sect.~\ref{sec:quiescence}). The exercises at
the end of this chapter  will provide an opportunity to experiment
with the search and the evaluation function to develop an
understanding of the search-eval
interplay.

On the other hand, the evaluation
function is necessary because the search space is too large to search
completely.  
Let us now look in more detail at the size of the state space.

\subsection{State Space}

Searching for the best action of a state is hard because the state
space of possible successors (and successors of successors of
$\ldots$) is so large. Let us try to see how large it is. An often
used measure to bound the size of the state space is to 
calculate how
many legal  states would be traversed if all successors are
evaluated. 
In general, finding 
the exact state space size of games is 
a surprisingly challenging problem that has generated many
interesting research papers~\cite{shannon1988programming,schaeffer2007checkers,bonnet2013complexity,allis1994searching,fraenkel1981computing,iwata1994othello,hsieh2007fairness,iida2002computer,tromp2006combinatorics}.

\begin{figure}[t]
\begin{center}
\includegraphics[width=8cm]{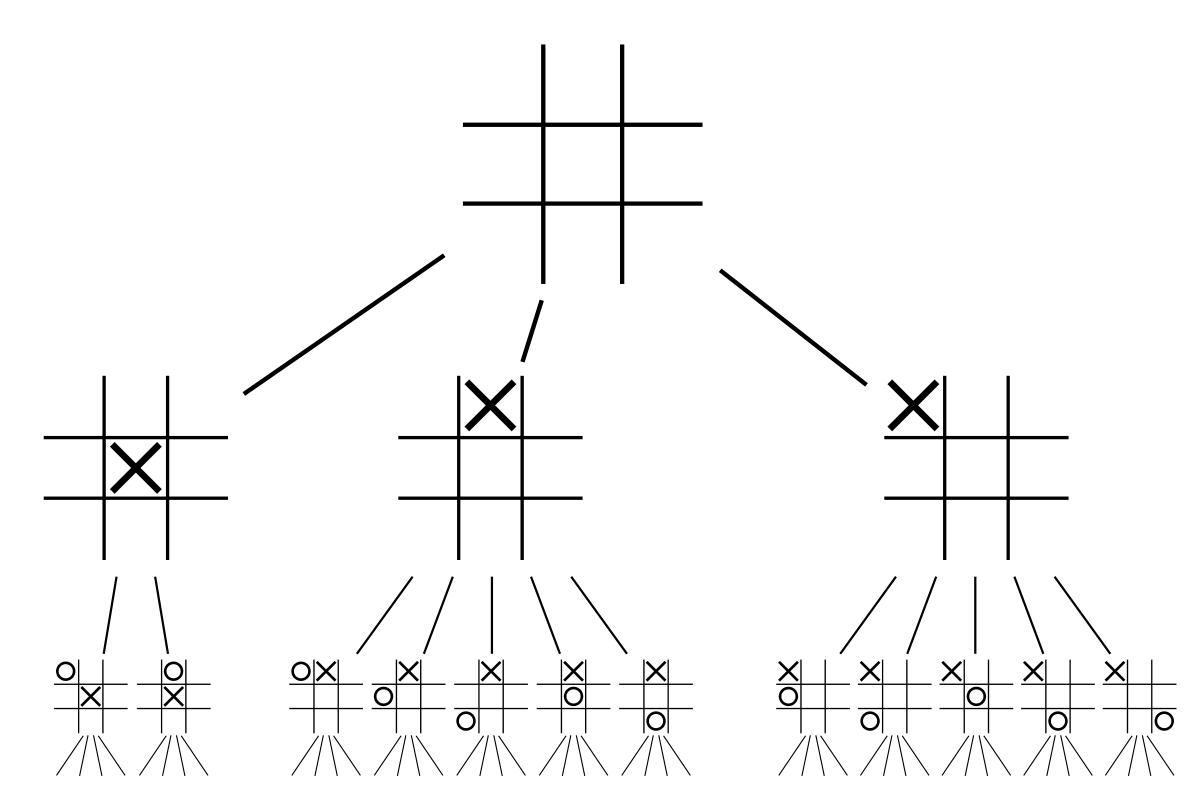}
\caption{Tic Tac Toe game tree (part)}\label{fig:ttt-tree2}
\end{center}
\end{figure}
\index{Tic Tac Toe}

For small games, such as Tic Tac Toe, the state space is small enough
to enumerate with a computer.
See Fig.~\ref{fig:ttt-tree2} for an impression of a
depth-2 tree of how the possible states are enumerated. 
To compute the size of the Tic Tac Toe state space, we note that there are 9 squares
on the board that can be either empty, cross, or circle. Thus there are $3\times
3\times\ldots\times 3= 3^9=19$,$683$ possible board configurations. However,
this is an overestimation of the reachable state space, since it
contains unreachable and illegal positions.
When we remove illegal positions we come to $5,478$ states. When we
remove symmetric positions (rotations and reflections) there are only
$765$ positions left~\cite{schaefer2002}. 

Beyond Tic Tac Toe, for more challenging games such as Chess or Go, calculating the
precise number of legal and reachable game positions is harder.
For Chess the size of the state space is often approximated  as $10^{47}$, and for
Go on the standard $19\times 19$ board this number is
$10^{170}$~\cite{allis1994searching,walraet2016googolplex,tromp2006combinatorics}.\index{state
  space complexity}\footnote{John Tromp  published in 2016 that the exact number of
  legal Go states is
  208168199 381979984 69947863 33448627 7028652 24538845 3054842 56394568
  209274196 127380153 78525648 45169851 9643907 25991601 5628128 54608988
  831442712 971531931 75577366 20397247 064840935. Tromp used advanced
  combinatorial methods in his calculations~\cite{tromp2016}.}

Note that the state space size is different from the game tree
size. The former is the game {\em graph}; the latter is the
game {\em tree}.    The same position from the graph can occur in many
different lines of play  that constitute the
tree.  These shared
positions are  called transpositions. The difference between the
game graph and the game tree is
large~\cite{shannon1988programming,allis1994searching,plaat1996exploiting,plaat1996exploiting}.
For Tic Tac Toe an upper bound for the former is
$3^9=19$,$683$ (pieces on squares), while for the latter it is $9!=362$,$880$
(different full game sequences
of moves).

An
essential element of Chess and Checkers programs is the transposition
table, a cache of positions, preventing the generation of states that
have already been searched before. In
Sect.~\ref{sec:tt} we will go 
deeper into transposition tables and their advantages.

\subsubsection*{Code}
Listing~\ref{lst:smalltree} gives an example in Python code of a very
small tree, with one max node (the root), three min nodes, and nine
leaves. This code is meant to be easy to understand; clearly,
more efficient ways of coding a tree are possible. 
In a short while we will provide an algorithm to compute the value of the tree by
looking at the leaf values, and later on by looking at the material (pieces).
This tree is shown in Fig.~\ref{fig:minimax}. The values at
the leaves $(6, 1, 3, 3, \ldots)$ stand for the heuristic value of the
leaves, where 6 is better than 3. The values themselves have no special
meaning; they were 
chosen for explanatory purposes. Again, squares are max nodes, and circles
are min nodes.

\lstset{label={lst:smalltree}}
\lstset{caption={Small tree code}}
\begin{lstlisting}[language=Python,float]
# a small tree, to be searched by the minimax algorithm
leaf_node1 = {'type': 'LEAF', 'value': 1, 'material': 3}
leaf_node2 = {'type': 'LEAF', 'value': 2, 'material': 4}
leaf_node3 = {'type': 'LEAF', 'value': 3, 'material': 2}
leaf_node4 = {'type': 'LEAF', 'value': 4, 'material': 9}
leaf_node5 = {'type': 'LEAF', 'value': 5, 'material': 1}
leaf_node6 = {'type': 'LEAF', 'value': 6, 'material': 7}
min_node1 = {'type': 'MIN', 'children' : [leaf_node6, leaf_node1, leaf_node3], 'material': 2}
min_node2 = {'type': 'MIN', 'children' : [leaf_node3, leaf_node4, leaf_node2], 'material': 7}
min_node3 = {'type': 'MIN', 'children' : [leaf_node1, leaf_node6, leaf_node5], 'material': 1}
max_node1 = {'type': 'MAX', 'children' : [min_node1, min_node2, min_node3], 'material': 3}

root = max_node1
\end{lstlisting}

\subsection{Search}\label{sec:minmax}
How can we determine if a position is won, lost, or drawn, and how can
we find the best move that leads to this outcome? In a two-player
zero-sum game players are always assumed to choose moves 
that lead to the best successor positions. Let us assume
that all values of states in the tree are viewed from the first
player.
If node $n$ represents the  position of which 
we wish to determine the value, then the value of node $n$ is the
maximum of the value of its children $C_n$.
The value of these child
nodes is determined analogously, although here the moves are made by our
opponent, who chooses positions that will minimize the outcome for
us. Next it is our move again, and we choose the position that
maximizes our winning probability. This alternating of maximizing and
minimizing is the reason the procedure is called {\em minimax}.\footnote{Note
that minimaxing is a kind of self-play. Since our opponent uses the
same algorithm as we do, in minimax we play against ourselves; our
opponent is just as smart as we are. This kind of minimax self-play is
a different kind of self-play than the one in Chap.~\ref{chap:plan}, where a self-play loop is used to train
a neural network evaluation function to learn to play a game from scratch.}


\lstset{label={lst:minimax}}
\lstset{caption={Minimax code}}
\begin{lstlisting}[language=Python,float]
INF = 99999

def eval(n):
    return n['value']


def minimax(n):
    if n['type'] == 'LEAF':
        return eval(n)
    elif n['type'] == 'MAX':
        g = -INF
        for c in n['children']:
            g = max(g, minimax(c))
    elif n['type'] == 'MIN':
        g = INF
        for c in n['children']:
            g = min(g, minimax(c))
    else:
        error("Wrong node type")
    return g

print("Minimax value: ", minimax(root))

            

        
\end{lstlisting}

\begin{figure}[t]
\begin{center}
\includegraphics[width=8cm]{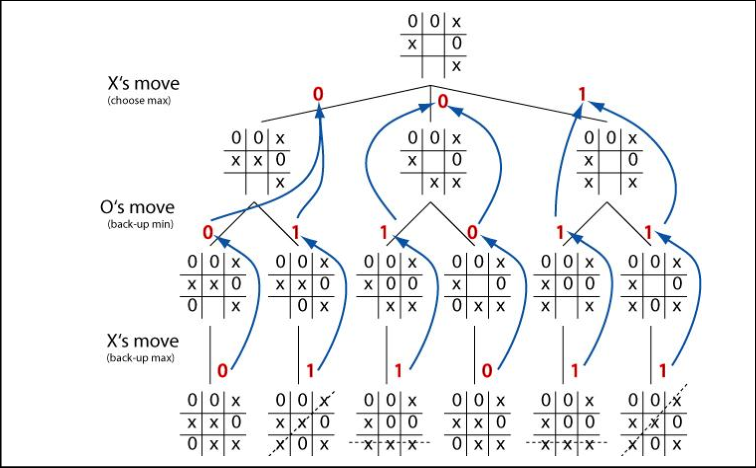}
\caption{Tic Tac Toe game tree with values~\cite{elnaggar2014comparative}}\label{fig:TTTminimax}
\end{center}
\end{figure}

\begin{figure}[t]
  \begin{center}
    \begin{tikzpicture}
\tikzstyle{max node}=[rectangle,draw,inner sep=6]
\tikzstyle{min node}=[circle,draw,inner sep=4]

\node[max node](0){2}
          child[sibling distance=30mm]{node[min node]{1}
                  child[sibling distance=10mm]{node[max node]{6}}
                  child[sibling distance=10mm]{node[max node]{1}}
                  child[sibling distance=10mm]{node[max node]{3}}}
          child[sibling distance=30mm]{node[min node]{2}
                  child[sibling distance=10mm]{node[max node]{3}}
                  child[sibling distance=10mm]{node[max node]{4}}
                  child[sibling distance=10mm]{node[max node]{2}}}
          child[sibling distance=30mm]{node[min node]{1}
                  child[sibling distance=10mm]{node[max node]{1}}
                  child[sibling distance=10mm]{node[max node]{6}}
                  child[sibling distance=10mm]{node[max node]{5}}};
\end{tikzpicture}
\caption{Minimax tree of Listing~\ref{lst:minimax}}\label{fig:minimax}
\end{center}
\end{figure}
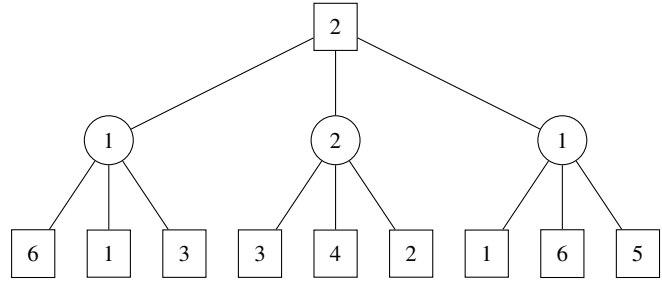

In Fig.~\ref{fig:TTTminimax} a Tic Tac Toe tree is drawn with
values.   Figure~\ref{fig:minimax} shows a more abstract minimax tree,
where nodes are drawn as squares (max) and circles (min). 
By following the values we can see how the values of the
nodes are determined. 

Listing~\ref{lst:minimax} gives example code for the minimax
algorithm. 
The minimax function will determine the value of a node by taking the minimum
of the value of the children of the min nodes, and the maximum of the
values of the children of the max nodes. In the case of the tree in Listing~\ref{lst:smalltree}, the value of
nodes \verb|min_node1| and \verb|min_node3| is both 1, and the value
of \verb|min_node2| is 2. The maximum of 1 and 2 is 2, so the minimax
value of the root is 2.\index{minimax}

\subsubsection*{Search Tree}\index{backtrack}
The minimax procedure recursively traverses the entire game
tree (Listing~\ref{lst:minimax}). Minimax is a {\em backtracking\/} procedure. It starts at the
root, and does a depth-first traversal of the tree by calling itself
recursively for each child node. When a child has been searched, it
backtracks up (returning its value) and searches down the next child,
going down-up-down-up through the tree. Minimax is a trial-and-error
procedure, as all reinforcement learning procedures are. The trial
element is trivial: try all successors. The error 
part is the backup rule. Minimax does not explore; all learning is
on-policy. The values that are backed up are those of the nodes that
were selected.

The nodes that are
actually traversed during a search procedure are referred to as the {\em
search tree}. The tree version of the full state space is also known
as the game tree. Since minimax recursively traverses all nodes in the
state space,  the search tree is identical
to the game tree. In the rest of this book we will see 
many examples of procedures whose search tree is significantly smaller
than the game tree while they still are able to find the  value of the root. Minimax does a full-depth  full-width expansion of
the game tree.  For regular trees, where all nodes have the same
number of children, the number of leaves of the game tree is $w^d$, where $w$ is
the width of the nodes (the number of children of a node) and $d$ is
the depth of the tree. For the tree in Figure~\ref{fig:minimax} the
width is $3$ and the depth is $2$, so the number of leaves is
$3^2=9$. The size of the game tree is exponential in the depth
parameter, and is dominated by the number of leaves. A useful
approximation of the size of the tree is
therefore  $w^d$, and also of the running time of the minimax
procedure, since it visits every node.
\index{search tree}

\lstset{label={lst:depthmm}}
\lstset{caption={Depth-limited minimax}}
\begin{lstlisting}[language=Python,float]
INF = 99999

def heuristic_eval(n):
    return n['material']

def minimax(n, d):
    if d <= 0:
        return heuristic_eval(n)
    elif n['type'] == 'MAX':
        g = -INF
        for c in n['children']:
            g = max(g, minimax(c, d-1))
    elif n['type'] == 'MIN':
        g = INF
        for c in n['children']:
            g = min(g, minimax(c, d-1))
    else:
        error("Wrong node type")
    return g

print("Minimax value: ", minimax(root, 2))

            

        
\end{lstlisting}


Now that we have looked at the search function, it is time to look at
the heuristic evaluation function.

\subsection{Eval}\label{sec:eval}

Calling the minimax function on a Chess position, whose game tree is
of size $10^{47}$, would take too long to compute its value.\footnote{Even if node evaluation
would take a very fast one nanosecond, it would still take $10^{38}$ seconds,
or $3\times 10^{30}$ years, which is about $10^{20}$ times the estimated age of the
known universe.} In order to reduce the size of the
search tree, the search depth can be reduced.\footnote{Also the width
of the tree can be reduced. With forward pruning fewer  children are
expanded at inner nodes. See Sect.~\ref{forward}.} To do so, even the
earliest Chess and   
Checkers programs (Turing, Samuel) used a heuristic evaluation
function as an approximation of the true value of a position. The heuristic is based
on domain knowledge, provided by domain experts. In Chess and Checkers,
the most important heuristic is material balance (the number and the
importance of the pieces on the board).

Instead of traversing the tree to the  depth where the
game ends due to an end-game position, the heuristics create
artificial leaves, at a
much shallower search depth. The deeper search is artificially limited,
and a heuristic is called to statically  evaluate whatever position happens to
occur at that depth.

A basic assumption of heuristic planning is that deeper searches
provide better approximations of the optimal policy and the  value of
the root.
Listing~\ref{lst:depthmm} adds depth limiting to our Python code of
minimax. New is
parameter \verb|d|. The parameter is decremented in the recursive call of the
function. The \verb|eval| function no longer uses the score
at the leaves, but takes a heuristic value of the node whenever the
depth parameter is 0. The heuristic eval in the example
code is quite straightforward; it uses the \verb|material| value of the
node (see Listing~\ref{lst:smalltree}). 

\subsubsection*{Domain Knowledge}
A heuristic is a function that encodes domain-specific
knowledge. To be useful in practice, a heuristic must be such that it
can be computed efficiently. For 
example, in Chess, having two pawns is better than having  one pawn,
and  the queen is more valuable than the rook. For many years heuristic evaluation
functions have been a central piece of combinatorial search. In fact,
they are so important that the field 
has been called heuristic
planning. (See, for example, the title of Judea Pearl's seminal
work~\cite{pearl1984heuristics}.)  

A heuristic evaluation function in general takes the form
$$h(s) = c_1\times f_1(s) + c_2\times f_2(s) + c_3\times f_3(s) +
\ldots + c_n\times f_n(s) ,$$
where $h(s)$ is the heuristic function of the state, $f_i(s)$ are the
feature terms such as material balance, mobility, or center
control, and $c_i$ are the coefficients, the weights that signal the
importance that a feature has in the total evaluation function.

In Chess and Checkers  material balance is the dominant factor in
the strength of a position. For this reason, the heuristic
evaluation function of Listing~\ref{lst:depthmm} 
uses the \verb|material| field of the nodes defined in
Listing~\ref{lst:smalltree}.
\index{heuristic evaluation}

Heuristic evaluation functions are linear combinations of features.
A typical Chess evaluation function consists of the following
features: material, mobility, king safety, and center control. The
evaluation function is static (it applies to a single
position). Tactical aspects such as captures are taken care of by the
search. To search as many positions as possible, speed of evaluation is
important. For that reason heuristics are
often optimized for speed, not for accuracy.

\subsubsection*{Heuristic Approximations}
Heuristic evaluations are, in almost all games, crude
approximations of the  value of the states. They capture static
elements of a position, and lack insight into the dynamics. Achieving
good play requires  extra work either on heuristics or on understanding
dynamics with search algorithms. In   reinforcement learning theory the two
approaches  (exact planning and function approximation) are often 
treated  as mutually exclusive approaches.
In most
board games, however, they are studied together out of necessity, due to
the low quality of play of heuristic evaluation 
functions without search, and conversely the low quality of play of
search without heuristic evaluation functions. 

A further problem with the heuristic values is that their unreliability
is not smoothed in any way by the minimaxing. Minimax applies a
sequence of max operations, which tends to exacerbate noise or
errors in the heuristic estimates. This effect, where deep minimax
searches of noisy values enlarge the noise, is called search
pathology (since deeper searches are supposed to give more reliable
results). Nau, Pearl, and others have studied search
pathology~\cite{nau1980pathology,pearl1983nature,beal1983recent,wilson2009error,sadikov2005bias}. In
practice, search pathology does not occur frequently. In any case, 
alternatives to minimax have been proposed that back up probability
distributions instead of single point values (Sect.~\ref{sec:expectimax}).

In tournament Chess and Checkers programs the heuristic evaluation function
is a highly advanced piece of code, using many features. The features
are typically  manually chosen, and manually optimized
and tuned. This feature engineering is quite labor intensive. The 
search space is large, and finding the precise state at which  unwanted
behavior occurs may be difficult. It may even be unclear if
solving a problem in one part of the play does not introduce a weakness in
another part.  To achieve world championship level,
the teams behind Chinook and Deep Blue spent years  refining their
heuristics.

A further  downside of a heuristic is, of course, that it is  domain
specific; e.g., a piece of Chess code that assigns the queen 900 points is of no 
use in Checkers, which does not have queens. 

\subsubsection*{Heuristics in Practice}
Today, many  Chess programs are open source, including some of the top
programs, such as
Stockfish~\cite{stockfish}.\footnote{The name Stockfish may seem
  strange for a Chess program. The authors Romstad and Costalba state that it reflects
  their two countries 
: ``produced in Norway
and cooked in Italy.''} To get an idea of just how extensive current
state-of-the-art heuristic 
evaluation functions have become, you may want to have a look at the Stockfish
source at GitHub, which can be found  \href{https://github.com/official-stockfish/Stockfish}{here}.\footnote{\url{https://github.com/official-stockfish/Stockfish}}

Stockfish has a conventional  evaluation function with manually designed features and manually
tuned weights. In other games evaluation functions based on machine
learning have been successful, such as
Othello (Logistello~\cite{buro1995statistical}),  Backgammon
(Neurogammon~\cite{tesauro1989neurogammon}), and of course Go. Despite some effort to
automate tuning of weights by Samuel~\cite{samuel1959some} and the Deep
Blue team~\cite{hsu2004behind}, in Chess and Checkers most programs use manual evaluation   
functions. Some papers describe success in
evolving coefficients of manual features~\cite{david2014genetic}, and
in AlphaGo both the features themselves and the coefficients are
learned~\cite{silver2016mastering,silver2017mastering,silver2018general}. (Table~\ref{tab:programs}
in Sect.~\ref{sec:history} lists  game playing programs that use
machine learning.)

Machine learning evaluation functions have the advantage of being
more general and easier to debug. At the end of this chapter we will
revisit machine learning evaluation enhancements. 
But first, we will explore in more depth how  the minimax function
works.


\section{State Space Complexity}\index{state space}
The state space of two-agent zero-sum perfect-information games has a
minimax structure. This  structure follows directly from the fact that
the value of  a state is determined by the value of
the best successor state (since we assume that rational players always
play the best move). The minimax state space is redundant; only a small part of the
state space (the optimal policy) defines the value of the initial state.  
We will now do a theoretical analysis  to
find the size of   the optimal policy. This analysis will
show us  that  there is
an exponential amount of redundancy in the tree, which allows us to create  much more
efficient search algorithms.


\subsubsection*{Width and Depth}
Let us start with a regular fixed-width fixed-depth tree. The size of this
tree is dominated by the number of leaf 
nodes, which is $w^d$ for width $w$  and depth  $d$. In Fig.~\ref{fig:minimax} we see a tree with $w=3$ and $d=2$
(the root node is depth 0).

\subsubsection*{Critical Path}
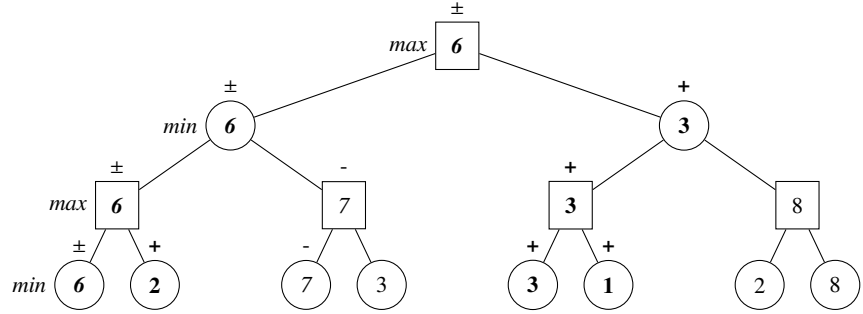
\begin{figure}[t]
\begin{center}
\begin{tikzpicture}
\tikzstyle{max node}=[rectangle,draw,inner sep=6]
\tikzstyle{min node}=[circle,draw,inner sep=4]
\tikzset{level distance=30pt}

\node[max node,label=left:{\em max},label=above:{\bf\em $\pm$}](0){\bf\em 6}
          child[sibling distance=60mm]{node[min node,label=left:{\em
                min},label=above:{\bf\em $\pm$}]{\bf\em 6}
                  child[sibling distance=30mm]{node[max
                      node,label=left:{\em max},label=above:{\bf\em $\pm$}]{\bf\em 6}
                            child[sibling distance=10mm]{node[min
                                node,label=left:{\em
                                  min},label=above:{\bf\em $\pm$}]{\bf\em 6}}
                            child[sibling distance=10mm]{node[min
                                node,label=above:{\bf +}]{\bf 2}}}
                  child[sibling distance=30mm]{node[max node,label=above:{-}]{\em 7}
                            child[sibling distance=10mm]{node[min node,label=above:{-}]{\em 7}}
                            child[sibling distance=10mm]{node[min node]{3}}}}
          child[sibling distance=60mm]{node[min node,label=above:{\bf +}]{\bf 3}
                  child[sibling distance=30mm]{node[max
                      node,label=above:{\bf +}]{\bf 3}
                            child[sibling distance=10mm]{node[min
                                node,label=above:{\bf +}]{\bf 3}}
                            child[sibling distance=10mm]{node[min
                                node,label=above:{\bf +}]{\bf 1}}}
                  child[sibling distance=30mm]{node[max node]{8}
                            child[sibling distance=10mm]{node[min node]{2}}
                            child[sibling distance=10mm]{node[min node]{8}}}};
\end{tikzpicture}
\caption[Critical tree]{{\bf\em $\pm$ Critical path}, {\bf + max
solution tree}, {\em - min solution tree}}\label{fig:prooftree}
\end{center}
\end{figure}

\index{critical path}
Figure~\ref{fig:prooftree} shows a different minimax tree, $w=2$, $d=3$,
with leaf values, and values at the inner nodes. 
Recall that the leaf values are typically
integer values from a heuristic score, for example in the range $[-16$,$000,
+16$,$000]$. As usual, the root is a max
node; its value is the maximum of its child values, and the children of
the root are min nodes, whose value  is the minimum of their child
values, as can be seen in the example tree.  The path of bold, italic
number 6s is called
the {\em critical path,} the path of nodes whose value is equal to the value of the
root.

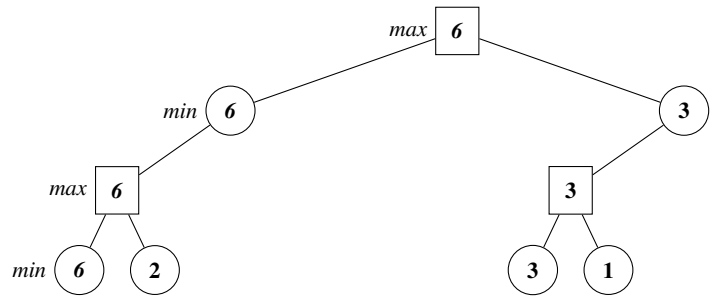
\begin{figure}[t]
\begin{center}
\begin{tikzpicture}
\tikzstyle{max node}=[rectangle,draw,inner sep=6]
\tikzstyle{min node}=[circle,draw,inner sep=4]
\tikzset{level distance=30pt}

\node[max node,label=left:{\em max}](0){\bf\em 6}
          child[sibling distance=60mm]{node[min node,label=left:{\em
                min}]{\bf\em 6}
                  child[sibling distance=30mm]{node[max
                      node,label=left:{\em max}]{\bf\em 6}
                            child[sibling distance=10mm]{node[min
                                node,label=left:{\em
                                  min}]{\bf\em 6}}
                            child[sibling distance=10mm]{node[min
                                node]{\bf 2}}}
                  child[sibling distance=30mm,transparent]{node[transparent]{\em 7}
                            child[sibling distance=10mm]{node[min node,transparent]{\em 7}}
                            child[sibling distance=10mm]{node[min node,transparent]{3}}}}
          child[sibling distance=60mm]{node[min node]{\bf 3}
                  child[sibling distance=30mm]{node[max node]{\bf 3}
                            child[sibling distance=10mm]{node[min
                                node]{\bf 3}}
                            child[sibling distance=10mm]{node[min
                                node]{\bf 1}}}
                  child[sibling distance=30mm,transparent]{node[transparent]{8}
                            child[sibling distance=10mm]{node[min node,transparent]{2}}
                            child[sibling distance=10mm]{node[min node,transparent]{8}}}};
\end{tikzpicture}
\caption[Max solution tree]{Max solution tree: upper bound determined by all children at
  max nodes, a single child at min nodes}\label{fig:maxsoltree}
\end{center}
\end{figure}

\begin{figure}[t]
  \begin{center}
\begin{tikzpicture}
\tikzstyle{max node}=[rectangle,draw,inner sep=6]
\tikzstyle{min node}=[circle,draw,inner sep=4]
\tikzset{level distance=30pt}

\node[max node,label=left:{\em max}](0){\bf\em 6}
          child[sibling distance=60mm]{node[min node,label=left:{\em
                min}]{\bf\em 6}
                  child[sibling distance=30mm]{node[max
                      node,label=left:{\em max}]{\bf\em 6}
                            child[sibling distance=10mm]{node[min
                                node,label=left:{\em
                                  min}]{\bf\em 6}}
                            child[sibling distance=10mm,transparent]{node[min
                                node,transparent]{\bf 2}}}
                  child[sibling distance=30mm]{node[max node]{\em 7}
                            child[sibling distance=10mm]{node[min node]{\em 7}}
                            child[sibling distance=10mm,transparent]{node[min node,transparent]{3}}}}
          child[sibling distance=60mm,transparent]{node[min node,transparent]{\bf 3}
                  child[sibling distance=30mm]{node[max node,transparent]{\bf 3}
                            child[sibling distance=10mm]{node[min node,transparent]{\bf 3}}
                            child[sibling distance=10mm]{node[min node,transparent]{\bf 1}}}
                  child[sibling distance=30mm]{node[max node,transparent]{8}
                            child[sibling distance=10mm]{node[min node,transparent]{2}}
                            child[sibling distance=10mm]{node[min node,transparent]{8}}}};
\end{tikzpicture}
\caption[Min solution tree]{Min solution tree: lower bound determined by all children at
  min nodes, a single child at max nodes}\label{fig:minsoltree}
\end{center}
\end{figure}
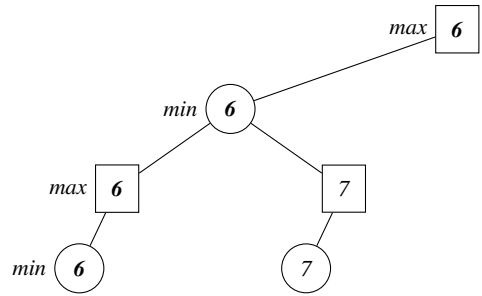

\begin{table}[t]
\begin{center}
\begin{tabular}{cccc}
{\bf Solution tree} & {\bf Value} & {\bf \# children at max
                                            node} & {\bf \# children at min node}\\ \hline\hline
Max & upper bound & all & $1$\\ \
Min & lower bound & $1$ & all\\ \hline
\end{tabular}
\caption{Solution trees and bounds}\label{tab:solutiontree}
\end{center} 
\end{table}

\begin{table}[t]
\begin{center}
\begin{tabular}{c}
{\bf Critical/solution trees}\\ \hline\hline
Critical tree = max solution tree $\bigcup$ min solution tree\\ 
Critical path = max solution tree $\bigcap$ min solution tree\\ \hline
\end{tabular}
\caption{Critical tree and path}\label{tab:crittree}
\end{center}
\end{table}

\subsubsection*{Critical Tree}
We will now  analyze  the lower bound of the size
of the tree that needs to be expanded to determine the value of the root.

As we shall see,  not all nodes in the minimax tree play a role in 
determining the value at the root. In fact, only a small fraction of the tree
nodes  do so. Let us try to construct the
smallest possible subtree that defines the 
minimax value. By the definition of the maximum function that takes
the highest of its inputs, at max nodes, the highest child 
determines the value of the parent node. All other children are
unnecessary; they are noncritical nodes that are ``dominated'' by their critical
sibling. Likewise, at min nodes, only the lowest-valued child determines the value of
the node.

So, it would seem that we only need to expand a single child to know
the value of a node. There is a complication, however. If we only expand
one child, then we do not know for certain what the value of a node
is. Even if it is the critical child, since we have only seen one child
so far we do not yet know if there is another, better child. At a max parent, the
single expanded child creates a 
{\em lower bound\/} on the minimax value, since expansions of subsequent
children can only increase the value upwards. Likewise,
a min node whose children have been partly expanded has an upper bound
on its final value, since subsequent expansion can only lower the value
further.

A partly expanded max node defines a lower bound on the minimax value. Lower bounds
of {\em min\/} nodes, however, require all children of that min node
to be expanded (since, if it were partly expanded, then expansion of
another child could lower the value further, and the lower bound
evidently was not a lower bound).

Likewise, a partly expanded min node is an upper bound on the min node
value. Upper bounds of {\em max\/} nodes, however, require all
children of that max node to be expanded.

Now we have enough information to construct the smallest tree that
defines the minimax value. If we wish to have an upper bound at the
root node,  then we need a tree where at least {\em one\/} child of
the min nodes is present, and {\em all\/}
children of all max nodes are present. Figure~\ref{fig:maxsoltree}
shows such a tree. This tree is called a max solution tree in the
literature, the tree determining an upper
bound~\cite{de1994solution,pijls1990another}. In a max solution tree,
all children at the max nodes are included, and one child at the min nodes.

Conversely, if we wish to have a lower bound at the root node, then we have to
construct a tree where at least one child of all max nodes is present,
and all children of all min nodes. Figure~\ref{fig:minsoltree} shows 
such a tree, which is called a min solution tree, a tree determining a
lower bound in a minimax tree.\index{min solution tree} In a min solution tree,
all children at the min nodes are included, and one child at the max
nodes (Table~\ref{tab:solutiontree}).

The minimax value of a position is, of course, defined by  two  solution
trees of equal value: one min solution tree defining a lower bound of
the same value as the upper bound defined by its max solution tree.
The intersection of the 
min solution tree and the max solution tree is the critical path. 
Figure~\ref{fig:prooftree}  contains such an intersection of a
min solution tree and a max solution tree. The union of the two
solution trees is called the
critical tree, or the proof tree (see Table~\ref{tab:crittree}). It is the smallest tree that proves
the value of the position.

The critical path is the optimal policy of optimal actions in each
state, the critical tree proves its value. An
optimal algorithm to find the optimal policy would search no more than the critical tree.
The critical tree  proves that the minimax value is not less than the
value of the min solution tree, and not larger than the value of the
max solution tree. Every algorithm that wishes to find the minimax
value has to traverse at least this critical tree, otherwise it has
not seen enough of the tree to determine its value for certain~\cite{knuth1975analysis}.\index{critical tree}

The size (or rather, the number of leaf nodes) of a max rooted max solution tree
is $w^{\lceil d/2\rceil}$, or the square root of the size of the
minimax tree, where $\lceil x\rceil$ rounds up $x$ to the nearest integer. The number of leaves of a max rooted min solution
tree is $w^{\lfloor d/2\rfloor}$.  The reason that the size of the
solution trees is determined by  {\em half\/} of the search depth
is interesting to see. As the minimax tree grows down from the root
the number of nodes is multiplied at each depth by $w$, but in a solution tree it is
intermittently multiplied by $1$, and then by $w$, and by $1$, and by
$w$, and so on. If $w_d$ is the branching factor at depth $d$, then
the number of leaves of the depth-$3$ minimax tree of
Fig.~\ref{fig:prooftree} is $2_1\times 2_2\times 2_3=8$. The
number of leaves of the max solution tree of
Fig.~\ref{fig:maxsoltree} is $2_1\times 1_2\times 2_3=4$. The
number of leaves of the min solution tree of
Fig.~\ref{fig:minsoltree} is $1_1\times 2_2\times 1_3=2$. 

Likewise, the number of leaves  of the critical tree (the union of the two solution
trees) is $w^{\lceil d/2\rceil}+w^{\lfloor d/2\rfloor}-1$, the square root of the size of the
minimax tree, $\sqrt{w^d}$. Thus, the size of the critical tree that defines
the minimax value is the square root of the size of the minimax
tree. 

In order to determine the value of a
minimax tree with $1$,$000$,$000$ leaves, at least $1000$ leaves have to be
examined. A bad algorithm would expand a million nodes, an optimally
good algorithm a thousand. In the same time that minimax would search to depth $5$, an
optimal algorithm would search to depth $10$, double the depth.

In Listing~\ref{lst:minimax} we have seen an $O(w^d)$ algorithm for
traversing the tree and finding the minimax value.

\subsubsection*{Conclusion}
We have now looked in depth at the part of the state space that
defines the minimax value.
It is now time to look at enhancements to the basic minimax algorithm.

In Sect.~\ref{sec:alphabeta} we will
describe alpha-beta, a best-case $\sqrt{w^d}$ algorithm for 
traversing the tree to find the minimax value.

\section{Search Enhancements}
The previous sections provided us with the basic building blocks for heuristic
planning in games, with the search-eval architecture and the minimax
algorithm. We have seen that theoretically it is possible to find the best move with a
thousand node expansions, where minimax would do a million. We will
now discuss search enhancement methods with which we can approach optimal efficiency.

Minimax is a fixed-width fixed-depth  algorithm that is too
inefficient to achieve high performance. The heuristic
function is domain dependent. In the enhancements in
this section and the next, we will see a trend towards variable-depth search and
generalization of the evaluation function (foreshadowing a bridge to
the next chapter on adaptive sampling).

An important role in search enhancements is played by the alpha-beta
algorithm, which uses a powerful exponential pruning mechanism.

\begin{table}[t]
  \begin{center}

\begin{tabular}{lccc}
   {\bf Name} & {\bf Principle} & {\bf Applicability} & {\bf Effectiveness}  \\ \hline\hline
  alpha-beta & backward pruning & all minimax games & $w^d\rightarrow\sqrt{w^d}$  \\ 
  iterative deepening & iterate search depth & with TT: move ordering
                                    & helps alpha-beta   \\ 
  transposition table & cache of states & with ID: move ordering &
                                                                   helps
                                                                   alpha-beta
  \\ 
  null window & alpha-beta cutoffs & all minimax games & more cutoffs  \\ 
  forward pruning & variable width & domain specific  & variable  \\ 
  capture moves & ordering & games with captures  & good  \\ 
  killer moves, HH & ordering & domain specific  & good  \\ 
  backup rule & probability &imperfect info  & imperfect info feasible    
  \\ \hline
\end{tabular}

  \caption{Search enhancements}\label{tab:enhancements}
  \end{center}
\end{table}

Table~\ref{tab:enhancements} gives an overview of the search
enhancements that are presented in this section, to help you find your way in
the myriad of terms and approaches. The effectiveness of the
approaches depends to a large  extent on the domain, hence the qualitative terms.
The enhancements are presented in the order shown in the table.

Most
search enhancements are related to alpha-beta, and improve its performance.
%
%
%
The alpha-beta algorithm is the core of decades of
high-efficiency game playing search. 
The difference between the
best case and the worst case performance of 
alpha-beta is large, $\sqrt{w^d}$ versus $w^d$. This best case is
achieved on well-ordered trees,  where the
best successor positions are the first that are expanded. On well-ordered trees alpha-beta
can find many cutoffs. (On badly ordered trees it performs as badly as
minimax.) Many of the search enhancements that we will now discuss
have been developed to improve the successor ordering, in order to
improve the performance of alpha-beta.

We will now start with the alpha-beta algorithm.

\subsection{Alpha-Beta}\index{alpha-beta}\label{sec:alphabeta}
Listing~\ref{lst:alphabeta} shows the alpha-beta algorithm
.
Let us see  how alpha-beta works, to understand how it finds the true
minimax value without searching the entirety of the tree.

\subsubsection*{Alpha-Beta Window}\index{alpha-beta window}
Alpha-beta maintains a {\em window\/} on the value of a
position. The window consists of two bounds, named $\alpha$ and
$\beta$. The $\alpha$ and $\beta$ bounds together form lower and
upper bounds on the possible value of the position, or node, that the algorithm
searches.

At the root node, alpha-beta starts out with an $\langle\alpha,\beta\rangle$-window
with the widest possible window, which we write as
$\langle-\infty,+\infty\rangle$.\footnote{Note that $\infty$ is
  shorthand for a number outside the range of numbers returned by the
  evaluation function. Also note that evaluation functions are assumed
  to return integer values.} Alpha-beta is a recursive algorithm to
find the minimax value of a node. In
order to find this value of a node, it calls itself recursively,
to find the values of the children of the node in order to compute its
own value. Each  call to alpha-beta
uses the latest (tightest) version of the window. The window for
recursive calls is updated 
with the results of the child nodes. When, at a  node $n$, the
window collapses (the bounds  become equal) the subtree below $n$
does not have to be explored since then the value of 
node $n$ is known, and thus the search of $n$ is
stopped. Alpha-beta is  then said to {\em cut off\/} the search of the
remaining children of $n$.\index{alpha-beta cutoff} The alpha-beta
cutoff is the central mechanism to achieve the  $w^d\rightarrow
\sqrt{w^d}$ best-case potential.

\lstset{label={lst:alphabeta}}
\lstset{caption={Alpha-beta}}
\begin{lstlisting}[language=Python,float]
INF = 99999

def alphabeta(n, a, b, d):
    if d <= 0:
        return eval(n) # return n['value']
    elif n['type'] == 'MAX':
        g = -INF
        for c in n['children']:
            g = max(g, alphabeta(c, a, b, d-1))
            a = max(a, g) # update alpha
            if a >= b:
                break # beta cutoff, g>=b
    elif n['type'] == 'MIN':
        g = INF
        for c in n['children']:
            g = min(g, alphabeta(c, a, b, d-1))
            b = min(b, g) # update beta
            if a >= b:
                break # alpha cutoff, a>=g
    return g

print("Minimax value: ", alphabeta(root, -INF, INF, depth))

            

        
\end{lstlisting}

\subsubsection*{Alpha-Beta Example}
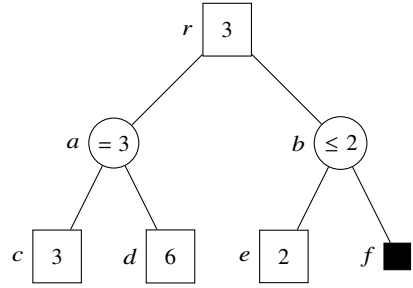
\begin{figure}[t]
\begin{center}
\begin{tikzpicture}
\tikzstyle{max node}=[rectangle,draw,inner sep=7]
\tikzstyle{min node}=[circle,draw,inner sep=2]
\tikzstyle{cut node}=[rectangle,draw,fill,inner sep=5]
\node[max node,label=left:{$r$}](0){$3$}
          child[sibling distance=30mm]{node[min node,label=left:{$a$}]{$=3$}
                         child[sibling distance=15mm]{node[max node,label=left:{$c$}]{$3$}}
                         child[sibling distance=15mm]{node[max node,label=left:{$d$}]{$6$}}}
          child[sibling distance=30mm]{node[min node,label=left:{$b$}]{$\leq2$}
                         child[sibling distance=15mm]{node[max node,label=left:{$e$}]{$2$}}
                         child[sibling distance=15mm]{node[cut node,label=left:{$f$}]{}}};
\end{tikzpicture}
\caption[Alpha-beta example]{
Alpha-beta starts at the root, with a window
of $\langle-\infty,+\infty\rangle$.  In other words, alpha-beta is called with
$n=r, a=-\infty,b=+\infty$. At the root it expands the max node's first child
$a$. 
Node $a$ is of type min, and  min node $a$'s first
child, leaf node $c$ is visited. Node $c$ is a leaf node with value $3$. Coming out of the recursion back in node
$a$ alpha-beta updates variables $g=\min(+\infty,3)$ and
$b=\min(+\infty,3)$ to $3$. It expands the second child of min
node $a$, leaf node $d$. This node $d$ is expanded with window
$\langle-\infty,3\rangle$, a tighter beta bound. Its value is $6$, coming 
back in $a$, alpha-beta updates the variables  $g=\min(3,6)$ and
$b=\min(3,6)$ to $3$, finds no other children at $a$, and returns
$g=3$ to $r$. Back  at the root $r$, alpha-beta
updates  $g=\max(-\infty,3)$ and
$a=\max(-\infty,3)$ to $3$. There is another child $b$,
which it expands with window $\langle 3,+\infty\rangle$; it
expands its first child $b$, finding 
leaf node $e$. The value of $2$ is returned to parent $b$, who updates
the variables $g=\min(+\infty,2)$ and $b=\min(+\infty,2)$ to $2$. Now
the if statement is tested and $3\geq2$ is true, so the break statement
breaks out of the loop, returning the value of $g=2$ to parent $r$,
cutting off the expansion of child $f$. At parent $r$, the root, the variables are
updated to  $g=\max(3,2)$ and
$a=\max(3,2)$ to $3$. Alpha-beta is ready and returns $g=3$ to the
caller as the minimax value $n$ {\em without having
expanded all nodes of the tree}.}\label{fig:abtree}
\end{center}
\end{figure}

We will now look in detail at the code,\footnote{You will note that the
code for the max and the min node is quite similar. A clever
reformulation exists, NegaMax, that switches the point of view each
time that it is called~\cite{fishburn1982analysis}. NegaMax has no code duplication and is
functionally equivalent. The switching of sides complicates reasoning
about trees, bounds, and values somewhat, and we do not use it in this
book.} to see how alpha-beta traverses the
same example tree of Fig.~\ref{fig:abtree}.\footnote{The code shown is the so-called fail-soft version of
alpha-beta, a version that may return values outside the
$\langle\alpha,\beta\rangle$-window. This may seem odd, but is
actually quite useful later on, when we introduce narrow-window
searches. The fail-hard version always returns a value inside the
$\langle\alpha,\beta\rangle$-window, or equal to $\alpha$ or $\beta$.}
The root $r$ is a max
node, the level below the root are
two min nodes, representing the two child positions of the player. Each
of these has two child positions (again max nodes) for which the
opponent moves. These are leaf positions (it is a very small
tree). The leaves have
values $3, 6, 2, \cdot$. Figure~\ref{fig:abtree} shows an
example tree and a detailed example run. Working through the example
in the caption of the figure will show you how an alpha-beta cutoff works.

In this tiny example tree, only one node was cut off, but in larger
trees, the maximum gain of alpha-beta is to cut off
$w^d-\sqrt{w^d}$ leaf nodes. (So, in a $w=10,d=6$ tree with $1$,$000$,$000$ leaves, 999,000
leaves can be cut off, searching only a critical tree of 1000 leaves. If
minimax can see important capture moves three moves in advance, alpha-beta, in the best
case, can see them six moves in advance, with the same computational
effort.) Typical high-performance game playing programs come close to the best
case; at the majority of the nodes (more than 90\%) the first child is also the best child~\cite{levy1991computers,schaeffer1986experiments}.

\subsubsection*{Alpha-Beta Cutoff}
In addition to the walk-through in the figure we will explain how
alpha-beta cut-offs work intuitively. 
The concept of {\em cutoff\/} can be understood intuitively as
follows. Let us assume we are the max
player searching the value of a max node (as in Fig.~\ref{fig:abtree}). 

As we perform our look-ahead,
we encounter child positions. Until all children have been searched,
the values of the children seen so far provide a bound
on the final value. In our case this intermediate value is a lower
bound (since we are in a max node our value can 
only increase with each child value that comes in). In alpha-beta,
this lower bound is recorded in the $\alpha$ value of the
$\langle\alpha,\beta\rangle$-window as soon as the first subtree of
the root has been searched. The $\alpha$-value thus increases
from $-\infty$ to the value of the best child $c_\alpha$ that we have seen
so far. In the example tree, this $c_\alpha$ is node $a$, the first
child of the root $r$. The alpha-beta window at root $r$, after $a$
has been searched, is $\langle 3,+\infty\rangle$. So far, so good.

Now, we continue the search with child $b$  with window $\langle
3,+\infty\rangle$. The positions 
that our opponent searches can only lower the values it finds, since our
opponent is the minimizer. Thus, in min nodes, the intermediate
search results are upper bounds. In min nodes, the search results
update $\beta$ of the $\langle\alpha,\beta\rangle$-window. As soon as
the opponent finds child positions $c_\beta$ equal
to or below
that good move $c_\alpha$ that we found in the previous paragraph,
the  search at our opponent's node can be stopped, since its remaining children cannot influence
{\em our\/} value anymore. In the example this child is node $e$,
whose value 2 causes the $\beta$ bound to fall below $\alpha$. For us, the remaining 
child positions have become irrelevant as soon as it has been proven
that their min parent's value will be lower than the good move $c_\alpha$. {\em The subsequent children of the min node are said to be cut
off by the good child $c_\alpha = a$ of the max node that we found earlier
and the strong reply move $c_\beta = e$ that the opponent found.}

After I have found a strong move $c_\alpha=a$, then a strong opponent's
move $c_\beta=e$ can
cut off the search for further opponent's moves. The reason is that my
opponent's move $c_\beta=e$  is so strong (bad for the first player) that searching
further for other children of $b$ that are even worse for me is
useless, since I will no longer play $b$, but choose instead to play
my own strong move $c_\alpha=a$.

In considering my own move, as soon as I have identified a strong
enough reply by 
my opponent to one of my options, I move on, cutting off the search for
other replies to that option. If I have not yet found a strong
counter-reply by my opponent, I continue searching for one in this
option.  If 
there is no strong reply, then I am happy, since I have found a
position where my opponent has no strong replies to my move.

\subsubsection*{Move Ordering}
Cutoffs  are possible due to the intermeshing of
maximization and minimization. 
%
%
The impact of alpha-beta on the development of early Chess and Checkers
programs was large. Through this simple algorithmic trick programs could
search twice as deep as before, achieving a much better quality of
play.\index{alpha-beta} 

It should be noted, however, that there is a catch. Alpha-beta is a
left-to-right algorithm, and it can
perform cutoffs only if the trees are well ordered. The example tree
in Fig.~\ref{fig:abtree} is well ordered: the root has two subtrees, one with value $3$ and one with value $\leq2$. It is well
ordered since  the highest (best) child occurs first. If the subtrees had
been ordered the other way around, then node $b$ would have been
expanded first, and leaf $f$ would have been expanded, along with all
other leaves, arriving at the same minimax value of $3$. The ordering
of successor positions does not influence the value of the position,
but it does influence the efficiency with which alpha-beta can compute
the value.
Alpha-beta's worst case time complexity is still $w^d$.

There is always an ordering of the leaves on which alpha-beta achieves the best-case
bound; this tree is called a perfectly ordered tree. In a
perfectly ordered tree all children of max nodes are ordered
high to low, and all children of min nodes are ordered low to high
(or to be more precise, all that is needed is that the first child is the best child).
Consequently, much research in move ordering heuristics has been
performed. Two of the best-known heuristics are iterative deepening
and transposition tables, which we will discuss next.

\subsection{Iterative Deepening}\index{iterative deepening}
Iterative deepening is a search enhancement that, together with
transposition tables,  can achieve excellent  move 
ordering, allowing alpha-beta to achieve high efficiency. A basic
assumption of the search-eval architecture is that looking 
deeper improves the quality of the answer.  A shallow search provides
a rough estimate of 
value and best move; a deeper search provides a better
estimate. Likewise, a shallow search provides an approximation of the
ordering for a deeper search.

In all practical applications of reinforcement learning, there is a
time limit to the search. In typical Chess tournaments a move has to be
produced  every three minutes on average.
The more time we have, the deeper we can
search, and the better the playing strength will be.  Thus it would be
useful if we could stop our search at {\em any time\/} and produce a good
answer~\cite{hansen2007anytime}. Iterative deepening provides a
solution. Instead of performing one large search to a
certain fixed search depth $d$, iterative deepening performs a
sequence of quick
searches, starting with depth $0$, increasing to depth $d$. It may
seem wasteful to perform these $d-1$ shallow searches, but due to the
exponential nature of the search, the shallow searches take negligible
time.

Iterative deepening is a technique to transform
a standard fixed-depth depth-first search algorithm into an anytime
algorithm, that can be stopped at any time and still have a useful
answer available.\footnote{Note that for anytime behavior also the
  alpha-beta searcher must now be interruptible, and check for
  key-pressed or time-up conditions---not shown in the alpha-beta code
  in the listing.}

In addition, when the search results (especially the best moves) of
all the shallow searches are
retained in a large cache (the transposition table), the move ordering
of each subsequent $d+1$ search will benefit from the shallower $d$ searches.

Thus, iterative deepening has  two advantages: (1) it transforms
alpha-beta into an anytime algorithm, and, together with transposition
tables, (2) it improves the move ordering for alpha-beta, increasing its efficiency.

The code of iterative deepening is  simple; see Listing~\ref{lst:id}.
A disadvantage of iterative deepening is that it expands more nodes:
all node searches before the final interation are
unnecessary. However, because of the exponential nature of the search
space, this number is small compared with the nodes searched in the
final iteration. Also, the search time of each iteration is difficult
to predict in practice, since the size of subtrees is unknown
beforehand. When an anytime algorithm is necessary, such as in
 a time-controlled tournament, the negligible extra time for the
earlier iterations is  traded off for the certainty of having a best move,
when one is needed.

\lstset{label={lst:id}}
\lstset{caption={Iterative deepening}}
\begin{lstlisting}[language=Python,float]
def iterativedeepening(n):
    d = 1
    while not keypressed() and not time_is_up():
        f = minimax(n, d)
        d = d + 1
    return f

\end{lstlisting}

Though useful by itself, iterative deepening really shines when
combined with transposition tables, since then it can improve move
ordering for alpha-beta significantly.

\subsection{Transposition Tables}\label{sec:tt}
\begin{figure}[t]
\begin{center}
\includegraphics[width=6cm]{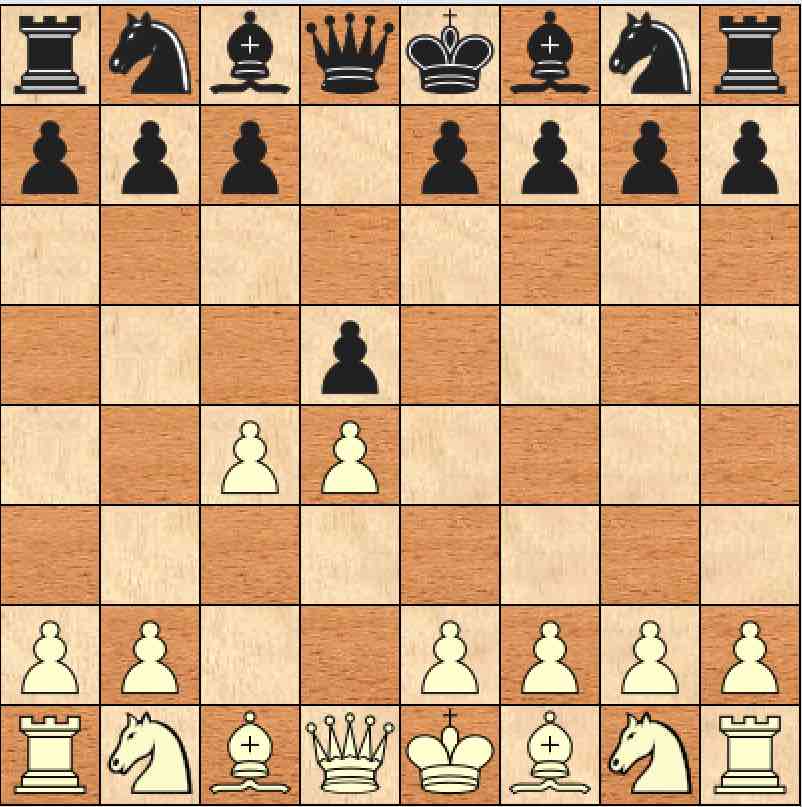}
\caption{Transposition: c4,d5,d4 or d4,d5,c4?}\label{fig:transposition}
\end{center}
\end{figure}

\index{transposition table}
In many games there are positions that can be reached through multiple parent
positions. For example, the position in Fig.~\ref{fig:transposition}
can be reached through the move sequence d4,d5,c4 but also through the
move sequence c4,d5,d4. The position has two parent positions (d4,d5) and
(c4,d5). Arriving at the same position through a different move
sequence is called a  transposition. So (c4,d5,d4) is a
transposition for the state that we get by playing (d4,d5,c4). It is
an identical state with a different move history.\footnote{To transpose means
to exchange.}

In Chess, the state space is not a tree but a graph. Searching
a transposition is wasteful, and exponentially so.
Tree traversal algorithms such as minimax and alpha-beta can
be transformed to  algorithms that traverse a graph efficiently through the addition of a
cache or 
table. Such a cache or table is called a transposition table. The
table  stores positions and their search results, in order to prevent
transpositions from being searched.
Before a new node is searched, the graph algorithm  checks if the
search result is already known in the table. Search results can be stored in
lookup tables (hash tables or dictionaries). For board games often the
fast Zobrist hash function is used~\cite{zobrist1970new}. In Python,
the Dictionary type is convenient to implement a transposition table.

The table entries typically contain an identification key, the search
depth, the value, a best move, and an indication of whether the value is a bound or a full
value. Entries are typically less than 100 bytes. A
program that searches $100$,$000$ nodes per second would need to store 1.8
GB to store the entire search tree for a three minute search. Typical
cache replacement schemes such as least-recently-used or depth-first are often used
when collisions occur.

Listing~\ref{lst:abtt} shows how transposition table results
are used to transform a tree search algorithms into a graph search algorithm. Two italic lines
of code are added to the code.  The two lines show basic versions of  the
use of lookup and 
store functions, returning the search value $g$, and the best move. In a real
program more elements would be added. For one,  real programs would
use a null window variant of alpha-beta (to be discussed soon). The search
results that are stored would be upper bounds, lower bounds, or minimax values.
Note that  in real programs not only the values of the best successor state 
would be stored, but also the actions  leading to them. In this
way, information on the {\em best action\/} in a shallower search depth is acquired, which will
greatly help with move ordering to produce many alpha-beta
cutoffs. Thus, even when a transpostion lookup for the necessary
search depth fails to find an entry of the required depth, and a
search ensues, this search will still use the best-move information stored in
the table for the shallower search depth, in order to improve the
tree ordering of the deeper alpha-beta search. For this reason, the
Listing shows how best-move \verb|ttbm| is inserted into the list of
children, so that it is searched first.

\lstset{label={lst:abtt}}
\begin{lstlisting}[language=Python,float,caption=Use of transposition
table]
INF = 99999

def ttalphabeta(n, a, b, d):
    (*\itshape (hit, g, ttbm) = lookup(n, d) *) # find in transposition table for depth d
    if hit: # if not hit at this depth, still use tt-bestmove
        return g
    if d <= 0:
        g = eval(n) # return n['value']
        bm = ()
    elif n['type'] == 'MAX':
        g = -INF
        for c in ttbm+n['children']: # search tt-bestmove first
            gc = ttalphabeta(c, a, b, d-1)
            if gc > g:
                bm = c # save best-move information
                g = gc
            a = max(a, g) # update alpha
            if a >= b:
                break # beta cutoff, g>=b
    elif n['type'] == 'MIN':
        g = INF
        for c in ttbm+n['children']:  # search tt-bestmove first
            gc = ttalphabeta(c, a, b, d-1)
            if gc < g:
                bm = c # save best-move information
                g = gc
            b = min(b, g) # update beta
            if a >= b:
                break # alpha cutoff, a>=g
    (*\itshape store(n, g, d, bm) *) # store in transpostition table
    return g

print("Minimax value: ", ttalphabeta(root, -INF, INF, depth))

            

        
\end{lstlisting}

Apart from the efficiency improvement due to the prevention of
duplicate search of transposed states, it is these best actions of
shallower searches that make
transposition tables so important. In an iterative deepening search,
at all inner nodes best action information will be present in the
transposition table. In most situations  search values and
best moves are moderately stable between iterations: the previous
search depth is a good approximation of the current search depth. The
consequence is that the  best action information from the transposition
table from a previous search depth will frequently be the
best action for a deeper search depth and therefore cause an alpha-beta cutoff at
inner nodes.\footnote{The opposite situation, where shallow searches
are not good predictors for deeper searches, is considered to be
pathological.\index{pathology} See the works of Nau, Pearl, and others~\cite{nau1980pathology,pearl1983nature,beal1983recent,wilson2009error,sadikov2005bias}.}

Together, iterative deepening and transposition tables allow for the
search of a
well-ordered tree and an efficient alpha-beta search, approaching the
best case of alpha-beta in practice~\cite{schaeffer1996new}. Note that
this is an
empirical finding, as pathological counter-examples can be
constructed, but do not appear to occur frequently in practice.

\subsection{*Null Window Search}\index{null window}
Before we go on to more heuristics in the next section, we will look further into
enhancements based on the alpha-beta window. The window is a powerful
concept, with which more 
efficiency can be gained. The  move ordering can be improved in order
to approach a perfectly ordered tree, improving the number of cutoffs
that alpha-beta can find. There is another way to achieve more cutoffs, and that is through null windows.

This is a starred section (*) that covers advanced material. If you
only want to follow the big picture then this  can  be skipped (or skimmed). For an understanding of how heuristic planning can
achieve good performance,  to get an
appreciation of the amount of 
complexity and effort that went into current state-of-the-art game
playing programs, please read on.

\subsubsection*{Narrow Windows, Bounds, and Solution Trees}
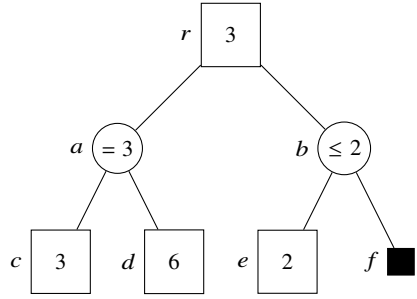
\begin{figure}[t]
\begin{center}
\begin{tikzpicture}
\tikzstyle{max node}=[rectangle,draw,inner sep=9]
\tikzstyle{min node}=[circle,draw,inner sep=2]
\tikzstyle{cut node}=[rectangle,draw,fill,inner sep=5]
\node[max node,label=left:{$r$}](0){$3$}
          child[sibling distance=30mm]{node[min node,label=left:{$a$}]{$=3$}
            child[sibling distance=15mm]{node[max node,label=left:{$c$}]{$3$}}
                         child[sibling distance=15mm]{node[max node,label=left:{$d$}]{$6$}}}
          child[sibling distance=30mm]{node[min node,label=left:{$b$}]{$\leq2$}
                         child[sibling distance=15mm]{node[max node,label=left:{$e$}]{$2$}}
                         child[sibling distance=15mm]{node[cut node,label=left:{$f$}]{}}};
\end{tikzpicture}
\caption[Narrow-window example]{
Alpha-beta starts at the root with narrow window
 $\langle-\infty,2\rangle$.  In other words, alpha-beta is called with
$n=r, a=-\infty,b=2$. At the root it expands  first child
$a$.
Node $a$'s first
child, leaf node $c$, is visited. Coming out of the recursion back in node
$a$, alpha-beta updates variables $g=\min(+\infty,3)$ to $3$ and
$b=\min(2,3)$ to $2$. No cutoff happens since $-\infty \not\geq
3$. We then expand the second child of min
node $a$, leaf node $d$.  Its value is $6$; coming 
back in $a$, alpha-beta updates the variables  $g=\min(3,6)$ and
$b=\min(3,6)$ to $3$, finds no other children at $a$, and returns
$g=3$ to $r$. Back  at the root $r$, alpha-beta
updates  $g=\max(-\infty,3)$ and
$a=\max(-\infty,3)$ to $3$. Now a cutoff happens, since $3 \geq
2$. The root returns the value 3, a lower bound since the max node has
unexpanded children. Let us also try a second narrow window, this
time with a high window: $\langle 4,+\infty\rangle$. At the root, node
$a$ and leaf $c$ are expanded, which returns $3$ at node $a$. There a cutoff
happens, since $4\geq 3$. Min node $a$ returns upper bound  $3$ to the
root. There, no  cutoff happens since $3\not\geq +\infty$ and node $b$
is expanded, with window $\langle 4,+\infty\rangle$. Node $e$ is
expanded and returns 2, which causes a cutoff in $b$ since $4\geq 2$,
and upper bound 2 is returned to the root, which returns $\max(3,2)=3$. This value
is an upper bound, since the two children returned upper bounds. After
two narrow window passes we now have an upper bound and a lower bound of
equal value 3.}\label{fig:narrowwindow}
\end{center}
\end{figure}

Alpha-beta searches with a wide window, which causes cutoffs at
internal nodes, returning the minimax value at the root node. The
tighter the window, the more cutoffs. So far we have only considered
alpha-beta as a routine that searches with a 
wide window at the root. At most inner nodes, alpha-beta
is called with a tighter search window. Let us have a closer look at
what happens: how much of the tree is searched, and how  should we
interpret a return value that falls outside the alpha-beta window
with which it was called? Figure~\ref{fig:narrowwindow} shows the
familiar tree, whose minimax value is $3$. In the figure we will
traverse the tree now two times, first  with a narrow alpha-beta  
window $\langle-\infty,2\rangle$ just below that minimax value, and
second with a narrow window $\langle 4,+\infty\rangle$ just above the minimax value, to show
what happens when the return value falls outside the alpha-beta window. Please
follow the example in the figure, and refer to the alpha-beta code in Listing~\ref{lst:alphabeta}.

The lower narrow window $\langle-\infty,2\rangle$ yields as return
value a lower bound $3 \leq v$ on the
minimax value (future child
expansion can still increase the value at max nodes, hence the value so
far is a lower bound on the final return value). As we can see, the
subtree expanded so far is a min solution tree (all children of min
nodes, not all of max nodes). The higher
narrow window $\langle 4,+\infty\rangle$ yields as return value an
upper bound $v \leq 3$ since the min children of the root returned upper
bounds, as only some of their children were expanded. The tree
traversed in this second pass for this upper bound is a max solution
tree (all children of max nodes, some of min nodes). The first (low) window
yielded lower bound $3 \leq v$, and the second  (high) window yielded
upper bound $v \leq 3$.  Since, in this
case, the lower and the upper bound are equal, we have found the
minimax value $3 \leq v \leq 3$ and the proof tree is the union of the
min solution tree and the max solution tree.

\subsubsection*{Post-condition and Bounds}
A narrow window causes more cutoffs, but returns a bound if the
return value is outside the window. Can we construct useful search
algorithms using only bounds?

We will now formalize these notions in  the alpha-beta post-condition, in
order to see if we can create a useful and efficient algorithm based
on the narrow window idea.
 We know that alpha-beta returns the minimax value if
it is called with the window $\langle-\infty,+\infty\rangle$.  When
the return value falls outside the window, then the return value
bounds the minimax value.  
\index{alpha-beta post-condition}

The post-condition of $g = \mbox{alpha-beta}(n, \alpha,\beta)$ is as
follows~\cite{knuth1975analysis,plaat1996best}:\label{postcondition}
\begin{enumerate}
\item $\alpha<g<\beta$: the window was right, $g$ is the minimax value of the tree
rooted in $n$. Alpha-beta has traversed at least the nodes in a critical tree, the
union of a max and a min solution tree.
\item $g\leq\alpha$: the window was too high, $g$ is an upper bound on the minimax value of the tree
rooted in $n$. Alpha-beta has traversed at least the nodes in a max solution tree
whose value is an upper bound $g$ on the minimax value.
\item $\beta\leq g$: the window was too low, $g$ is a lower bound on the minimax value of the tree
rooted in $n$. Alpha-beta has traversed at least the nodes in a min solution tree
whose value is a lower bound $g$ on the minimax value.
\end{enumerate}
This post-condition allows the construction of the ultimate in efficient
decision procedures: the null window search, or NWS. A null window of
$\alpha =\beta-1$ guarantees\footnote{We assume
that all values are integer values, so with $\alpha =\beta-1$ there is no
room for an integer value to fit in the window. A null window of
$\alpha =\beta$ would not work, since alpha-beta would return
immediately and not traverse any nodes. The term ``one window'' would
perhaps have been
better, but history has decided otherwise.}  that alpha-beta performs
the most efficient search possible, at the cost of returning only a
Boolean answer: the minimax value is either above or below the input
window. Finding the minimax value with the NWS decision procedure
requires a sequence of NWS calls. There are two main algorithms in the
literature, Scout, which uses NWS recursively, and MTD($f$), which 
uses NWS iteratively from the root.\footnote{The efficiency gain for null windows
  versus wide windows depends to a large degree on the quality of the
  move ordering. Experiments with tournament quality programs yielded
  differences of around 20\% in leaf count, increasing with search
  depth~\cite{plaat1994new,plaat1996best}. For badly ordered trees the
difference is much larger.} Scout is also known as
principal variation search~\cite{fishburn1982analysis}.


\subsubsection*{Scout}
Scout~\cite{pearl1980scout,reinefeld1989spielbaum}
works on the assumption that the first child of a node is the best
child. This first child is called the {\em principal
  variation}. Assuming such a well-ordered tree, Scout searches the first child with a 
wide window, and then uses NWS 
to check efficiently if the other children are indeed inferior. If not,
they are searched with a wide window, to find the value.
Listing~\ref{lst:scout} shows the code.\index{Scout}

\lstset{label={lst:scout}}
\lstset{caption={Scout~\cite{pearl1980scout,reinefeld1989spielbaum}}}
\begin{lstlisting}[language=Python,float]
INF = 99999

def eval(n):
    return n['value']

def scout(n, a, b):
    if n['type'] == 'LEAF':
        return eval(n)
    elif n['type'] == 'MAX':
        g = -INF
        for c in n['children']:
            if c == n['children'][0]:
                g = max(g, scout(c, a, b)) # first child wide window
            else:
                g = max(g, scout(c, a, a+1)) # other children nws
                if g >= a+1:
                    g = max(g, scout(c, g, b)) # if better, wide window 
            a = max(a, g) # update alpha
            if a >= b:
                break # beta cutoff, a>=b
    elif n['type'] == 'MIN':
        g = INF
        for c in n['children']:
            if c == n['children'][0]:
                g = min(g, scout(c, a, b)) # first child wide window
            else:
                g = min(g, scout(c, b-1, b)) # other children nws
                if g <= b-1:
                    g = min(g, scout(c, a, g)) # if better, wide window 
            b = min(b, g) # update beta
            if a >= b:
                break # alpha cutoff, a>=b
    else:
        error("Wrong node type")
    return g

print("Minimax value: ", scout(root, -INF, INF))

            

        
\end{lstlisting}

\subsubsection*{Transposition Table and Null Window Search}
The re-searches may seem inefficient, and they would be, if alpha-beta
would recompute search results for each search anew. (Node $a$ and $c$ in
the example of Figure~\ref{fig:narrowwindow} are re-expanded.)
Actual game playing versions of alpha-beta
use transposition tables, large lookup tables in which all nodes and their
search results are efficiently stored. Re-searching is therefore not
more expensive, since no new  nodes are expanded. The code in the
listing omits the lookup and store calls, and the depth parameter.

\subsubsection*{MTD($f$) and MTD-bi}\label{sec:mtdbi}
MTD($f$)~\cite{plaat1996best} takes as input a first
guess to the minimax value, and then 
iteratively calls a sequence of NWS decision procedures to home in on
the minimax value. If the first guess is too high, then NWS will
return a lower value, an upper bound, which is fed into the next NWS
call, and a sequence of lower return values (upper bounds) follows, until the
minimax value is found and NWS returns a higher return value (a lower
bound equal to the previous upper bound). If the first
guess is too low, the converse occurs.

In contrast to
Scout, which uses some wide window searches, MTD($f$) uses only
NWS. Experimental 
comparisons showed indeed that MTD($f$) expands even fewer nodes than
Scout on average, although trees can be constructed in which either
algorithm performs best~\cite{plaat1996best,plaat1996research}. Even more so than Scout,
MTD($f$) relies  on a well-functioning lookup table to store
intermediate search results in, 
which requires care with inconsistencies in the tree.

\lstset{label={lst:mtdf}}
\lstset{caption={MTD($f$) \cite{plaat1996best}}}
\begin{lstlisting}[language=Python,float]
def mtdf(n, f, d):
    ub = +INF
    lb = -INF
    while lb < ub:
        if f == lb:
            b = f+1
        else:
            b = f
        f = ttalphabeta(n, b-1, b, d)
        if f < b:
            ub = f
        else:
            lb = f
    return f

\end{lstlisting}

Listing~\ref{lst:mtdf} shows the code for MTD($f$).\index{MTD($f$)}
MTD($f$) starts the search for the minimax value at a heuristic first
guess, and then iterates in the direction of the minimax value based
on the outcome of the null window search. When the granularity of the
evaluation function is fine, the number of iterations can become large,
and the overhead of traversing through the tree also increases.

An alternative approach is to use a binary search~\cite{knuth1997art}
to interpolate between the upper and the lower bound. That
value can then be used as the pivot for the null window search, which will produce a
new bound, and to repeat the binary search. This may result in fewer
iterations of null window search to find the minimax value. This
algorithm is called MTD-bi~\cite{plaat1996best}. The idea of bisecting
the search goes back to\index{MTD-bi}
C*~\cite{coplan1982special,weill1992negac} where it is used in
end-game search. MTD-bi is used in Sunfish, 
 a tiny fully functional Chess program in 111 lines of Python code
 (Sect.~\ref{sec:sunfish}).

After this elaborate discussion of alpha-beta and related move
ordering enhancements, it is time to discuss other enhancements that
have been developed to improve the performance of heuristic planning.

\subsection{*Heuristic Forward Pruning}\label{forward}\index{forward
  pruning}
If we want to search deeper,
the first idea that often springs to mind  is to only search
 ``important'' nodes. This method is called forward pruning, in contrast to
 alpha-beta's backward pruning. In forward pruning,  the idea is to
 use domain-dependent heuristics to 
determine  these important nodes. If for certain moves we can immediately see
that they are not good, then  they should not be
searched. In early game playing
programs this was used since compute power was so limited relative
to the large number of subtrees to be searched.  This is the
approach taken in Go playing programs such as GNU Go, where the move
generator generates only the ``sensible'' moves such as connect, corner,
jump, and territory making moves. 
The disadvantage of heuristic forward
pruning is that the static heuristics may be wrong, and valuable moves are
missed, that only a search would reveal to be valuable. In practice it turned out
that static board heuristics can all too easily miss deeper tactical
game dynamics. As soon as compute power allowed, programmers of Chess and Checkers programs
started to use full-width move lists.

Plain heuristic forward pruning misses too many dynamics in most
games. In Section~\ref{sec:nullmove}  further developments of forward
pruning are introduced, named null move pruning~\cite{donninger1993null} and
Prob-cut~\cite{buro1997experiments}.

\subsection{*Capture Moves}
Another ordering mechanism that is highly domain specific, but is often highly
effective,  is capture move ordering. For alpha-beta 
it is important to search an ordered sequence of moves.
An obvious move ordering enhancement is to prioritize  capture moves,
since in
many games the dominant factor of the heuristic evaluation function
is material balance, e.g., how many pieces of a color are still on the
board. It stands to reason that, in positions where among the available actions
there are capture moves, those capture moves
lead to a better position, and should be tried first.

Many Chess programs therefore employ a move generator that first
generates capture moves, and if no cutoff has occurred, then the noncapture moves.

Note that this enhancement is heuristic; sometimes it fails, and a
non-capture move is best. It uses domain-dependent knowledge,
and only works in domains where captures are important and
frequent. In Go, for example, captures are less frequent, and the
capture enhancement is of no importance in this domain.


\subsection{*Killer Moves and History Heuristic}\index{killer
  move}\index{history heuristic}
Another well-known and effective enhancement is called the {\em killer move.} In many domains
actions that are good actions in one position are also good  in
another position. For example, if I am in a position where I find a
strong action, say by moving
my pawn from e4 to f5 I can capture my opponent's queen, then the move
pawn e4-f5 is likely to be a good move in many other positions,
provided the board has not changed too much. This move is called a
killer move. The enhancement works by recording moves that are so
strong as to cause an alpha-beta cutoff, and then first trying these moves in subsequent 
positions~\cite{huberman1968program}.

A generalization of the killer move is the history heuristic, which
adds  counters to the moves that cause cutoffs, making the heuristic
more refined.
The history heuristic maintains a table indexed by {\em from\/} and
{\em to\/}
squares, and when there is a cutoff, the entry in the table is
incremented, possibly weighted by search
depth~\cite{schaeffer1989history}.  The history heuristic came out of
basic experiments on the interplay of search and knowledge in Chess,
one of the earlier works with a principled empirical approach to this
question~\cite{schaeffer1986experiments}. 

Killer moves are even easier to implement than the history
heuristic,  are  more general than capture moves, and work in
many domains. They are present in most Chess and Checkers playing
programs. Killer moves were first described by Barbara Liskov, who
would (much later) win the 2008 Turing award for work on object-oriented and type-safe
programming. 
\index{Barbara Liskov} The
history heuristic was developed by Jonathan Schaeffer, who would later
lead the Chinook Checkers effort.

Most search enhancements focus on  selection of moves as a way to
improve efficiency.

To conclude our discussion of enhancements of the search function, we
will look at a  different element of the search, the backup rule.

\subsection{*Backup
  Rules Other than Minimax}\index{Expectimax}\index{*-minimax}\index{Proof number search}\label{sec:expectimax}
Most enhancements in this chapter are applicable to two-person
zero-sum perfect-information games, games that can be modeled well in
the minimax 
paradigm. For other games, different backup rules are
needed. Interestingly, even for some zero-sum games a non-minimax
backup rule works best, as we shall see in the next chapter on
MCTS. In this section we will discuss alternative backup rules.

The  minimax backup rule has dominated game playing research for a long
time (1950--2006). For
two-person zero-sum games with a single heuristic reward value it is intuitively appealing, since it
follows the idea of how humans play a game in which they try to pick
the best move.

For imperfect-information
games or games of chance, such as card games or dice games, minimax is not an obvious
choice. Researchers have tried to find
work-arounds to force these games into the minimax paradigm, with
limited success.

Statistical sampling can
be used to cope with the randomness  of the game. A disadvantage is
that in order to get reliable 
outcomes, the number of samples must be high. Alternatives to minimax
have been developed. Already in 1966 Michie published the Expectimax
algorithm~\cite{michie1966game} for computing  probabilities with the
product rule. In 1983 Ballard published the *-minimax rule, 
which augmented the alpha-beta algorithm with chance nodes, allowing
distributions to be propagated through the product rule (by multiplication) while still
allowing the efficiency of
cutoffs~\cite{ballard1983minimax,reibman1983non}. Subsequent
experiments are reported
in~\cite{chi1988comparison,nau1986experiments,hauk2004rediscovering}. Expectimax
and *-minimax did not become popular after these experiments, although
in 2013 Lanctot et al.~\cite{lanctot2013monte} reported good results
with a combination of Expectimax and statistical sampling.  

In 1988 McAllester~\cite{mcallester1988conspiracy} introduced conspiracy number search as an
alternative to alpha-beta, which is based on backing up the size of
the subtree in addition to the value (how many nodes {\em conspire\/} to
change a node's value). It thus explicitly searches for
small subtrees. It did not become popular, but it did lead to the
development of proof number
search~\cite{allis1994proof,allis1994searching}, which is based on the
search for 
Boolean values such as win/loss. Proof number search is often used for mathematically
solving games, to determine the exact win/loss/draw value of a
position---not its heuristic approximation. To solve a game, the state
space is fully searched to the terminal states~\cite{romein2003solving,schaeffer2007checkers}.

The next chapter, on Monte Carlo Tree Search,  will introduce the
{\em averaging of statistical samples\/} as backup rule. MCTS and its
backup rule did become quite popular and successful.

To summarize the different backup rules, minimax is an efficient
algorithm for backing up scalar values. Expectimax 
works with probability distributions. Averaging works well for statistical
sampling. For Boolean values such as win/loss, proof number search is
an efficient algorithm.

\subsection{Trends in Search}
We are coming to the end of the section on search enhancements and
will close with some conclusions.

Please refer to Table~\ref{tab:enhancements} on page~\pageref{tab:enhancements}. Minimax is a rigid fixed-width, fixed-depth algorithm, distributing
the search effort equally over all parts of the tree, good or bad. The
enhancements reported in this section all point to a trend of
variable-width, variable-depth, using forward or backward
pruning,  and move ordering, to
devote  most of the search effort to the most promising part of the
tree, to find the best move, disregarding bad moves as early as
possible.

Furthermore, there is a trend towards generality, away from
domain-specific heuristics, towards mechanisms such as the history
heuristic, 
iterative deepening, and transposition tables, methods that work in
many domains.

\begin{table}[t]
  \begin{center}

\begin{tabular}{lccc}
   {\bf Name} & {\bf Principle} & {\bf Applicability} & {\bf Effectiveness}  \\ \hline\hline
  search extensions & variable depth & domain specific  & good  \\ 
  quiescence search  & variable depth & domain specific  & good  \\
  null-move search  & variable depth & no Zugzwang  & good  \\ 
  odd-even effect  & move advantage & minimax games  & unstable values
  \\ 
  piece-square tables  & evaluation coding & heuristic evaluation  &
                                                                     efficiency  \\ 
  coefficient learning  & machine learning & evaluation tuning  &
                                                                  automated learning  \\ 
  end-game database  & database & endgame  & perfect knowledge  \\ 
    opponent modeling  & different opponent & teaching  & explore weaknesses   \\ \hline
\end{tabular}

  \caption{Eval enhancements}\label{tab:enhancements2}
  \end{center}
\end{table}

\section{Evaluation Enhancements}\label{sec:evalenhance}
After elaborating on the many search enhancements that have been
invented, it is time to see how 
heuristic evaluation can be enhanced.  Table~\ref{tab:enhancements2}
gives an overview of the evaluation function enhancements. The
enhancements try to ameliorate the limitations of static heuristics
with search extensions, and improve the evaluation function by smart
implementations and databases. (The term Zugzwang, or forced moving, is
explained in Sect.~\ref{sec:nullmove}.) 

\subsection{*Search Extensions}
The heuristic evaluation function computes a score of the static board
position. These scores are then used in the search. Potentially important
tactical dynamics in the position cannot be seen by a standard static
evaluator, since the tactical problem is over the {\em  search horizon.} It would
help if this situation could be detected, and the 
search be extended somewhat to see the effect of the tactical dynamics. 
Singular extensions were devised by the Deep Thought
team (the predecessors of Deep Blue) for Chess~~\cite{anantharaman1990singular,hsu2004behind}. There are many variations of search
extension; all are based on domain-specific heuristics. Two of the
best-known search extensions for Chess are check extensions and 
capture extensions. Positions that are in check or where some kind of  capture
happened are searched one level deeper. A problem with capture
extensions can be that in positions where many capture opportunities
exist, the search expands too much. For this reason partial depth
extensions have been devised.  Tuning the amount and type of
search extensions is an important part of the tuning effort of game
playing programs, especially in 
Chess, which is a highly tactical
game.

Singular extensions are related in some sense to forward pruning. Forward
pruning uses heuristics to restrict search to certain ``good''
moves. Search extensions do a full-width search up to a certain depth,
and then use heuristics to extend certain positions for deeper search,
combining  full-width and selective search. 

Multi-cut introduced search
reductions~\cite{bjornsson2001multi}. Here a shallow search is used to determine if a
cutoff is taken early in a node. Related to multi-cut are
Prob-cut~\cite{buro1997experiments} and a pruning technique in
Chinook~\cite{schaeffer1992world} which uses a shallow search to
identify nodes to terminate the search.
Finally, search extensions are reported to  reduce the problem of search
pathology~\cite{kaindl1988minimaxing,althofer1988root,schrufer1986presence}.

\subsection{*Quiescence Search}\index{quiescence search}\label{sec:quiescence}
Related to search extensions is quiescence search. Quiescence search~\cite{beal1990generalised} and search extensions
both try to reduce the impact  of the search horizon. It tries to distinguish between
so-called {\em   quiet\/} and {\em noisy\/} positions. A noisy
position is a position whose value cannot be reliably estimated with
static heuristics, for example, because of the presence of threats or
other dynamics. Noisy
positions should be searched deeper until a quiet position arises,
that can more reliably be evaluated. Quiescence search
is most often implemented
by looking at differences between positions in the heuristic
evaluation function. As with search extensions, getting
quiescence search to work requires much experimentation to build up
understanding of the effects of the static evaluation heuristics that are used in one's
program~\cite{greenblatt1967greenblatt,kaindl1982dynamic,beal1990generalised,schadd2009quiescence}.

\subsection{*Null-Move Heuristic}\index{null move}\index{Zugzwang}\label{sec:nullmove}
We will now discuss the null-move heuristic. 
A null move is a pass, not making a move,  a position where the only difference with the
successor position is which side is to
move~\cite{beal1989experiments}. In some games passing is 
legal, such as in Go; in some games it is not, such as in Chess.
Being allowed to move typically provides opportunities for
improvement.  If it is not, and all possible moves make a position
worse, it is called {\em Zugzwang}, or forced moving. In this case
the introduction of a pass move causes an unfair advantage to occur.

The null-move heuristic tries to get alpha-beta cutoffs
cheaply by seeing if a pass creates a situation that is strong enough to
cause a cutoff before the regular moves
are tried.
If a position is so strong that not moving already provides a cutoff, then
it is a good idea to take that cutoff. In that case the cut off has been
found without an expensive search.

Null moves typically work well,  at least in Chess. (In other games, such as Othello,
Zugzwang plays a bigger role, and null moves are not
used, or in a different way~\cite{buro1997experiments}.)
The null-move idea has been developed further by extending  
zero-depth cutoff searches to small-depth cutoff searches, with
success~\cite{goetsch1990experiments,donninger1993null,heinz1999adaptive,buro1997experiments,brockington1997keyano}.
The ProbCut algorithm~\cite{buro1997experiments}
in particular has been successful in Othello.

\subsection{*Odd/Even Effect}\index{odd/even effect}
As was noted before, in most positions having the opportunity
to make a move is  advantageous. If we are white, then positions with
white to move are typically better than black-to-move positions. This
fact causes pronounced swings in 
the score between search depths. The odd/even
effect is caused by a simple statistical fact, that the expected value
of the maximum of a distribution is greater than the expected value of
the minimum of a distribution:
$E(\max(\mbox{random}[-999,+999]))>E(\min(\mbox{random}[-999,+999]))$.

The odd/even effect is a statistical effect;
it is possible to construct counter-example trees where the value at
the node is stable as the search depth increases.  This  effect is
also related to the search horizon. Many game trees have a pronounced
odd/even effect, especially when the trees are fixed depth.   As iterative
deepening searches deeper (Listing~\ref{lst:id}) the scores at the
root node go up and down as the 
leaf values switch between being minimized and maximized.

The odd/even effect
occurs in ``normal'' trees, for non-Zugzwang positions, and without
search pathology.\index{Zugzwang}\index{pathology} The odd/even
effect is not caused by search instability, but is a consequence of
the min-max structure of the minimax tree~\cite{edwards1961alpha}.

The value swings do
complicate empirical analysis and finding errors in programs, and
comparing search results from different search depths becomes less
reliable.

The odd/even effect can be reduced  by
search extensions, that transform the fixed-depth search into a
variable-depth search.
Another solution is to prevent the effect, using search extensions
and shallow searches to extend and reduce depth in multiples of two.


\subsection{*Piece-Square Tables}\index{piece-square
  tables}\index{Fruit (Chess program)}\index{Fabien Letouzey}\label{sec:piece}
After discussing enhancements that are concerned with the search
horizon, we will now look at the core of the evaluation function, the
efficient computation of the heuristic value.

Piece-square tables are a popular method for creating efficient
evaluation functions. They are used in many strong Chess
programs~\cite{good1968five}.
A piece-square table is a table for each piece on each square for each
color. Values for the piece $\times$ square $\times$ color combination
indicate the value of the piece being on this square. Piece-square
tables are efficient since moving a piece from a square to another
amounts to subtracting the source value and adding the destination
value to the heuristic evaluation score.

Piece-square tables are quite popular because of their efficiency. A
well-known implementation of piece-square tables is by Fabien 
Letouzey 
in his open source
program Fruit.\footnote{The source
code is available here:  
\href{http://arctrix.com/nas/chess/fruit/}{Fruit}. \url{http://arctrix.com/nas/chess/fruit/}}

Furthermore, piece-square tables are well suited for
learning the evaluation
function~\cite{beal1999learning,droste2008learning}, sometimes leading
to surprising differences between manually tuned and machine-learned values.
The creation of an evaluation function that encodes heuristics in a
correct way and combines their values efficiently is a challenging
problem. Let us look at how this can be automated.

\subsection{Coefficient Learning}
As has been noted before, tuning the coefficients of evaluation
function terms (Sect.~\ref{sec:eval}) by hand is a 
difficult and error-prone process. Recall the general form of the
evaluation function:
$$h(s) = c_1\times f_1(s) + c_2\times f_2(s) + c_3\times f_3(s) +
\ldots + c_n\times f_n(s) .$$

An alternative is to train the
coefficients by supervised learning. This works by creating a large
database of test positions from grandmaster games, of which the value
or the best
move is known. The difference between the evaluation of the state and
the correct answer constitutes an error value. Then an iterative optimization process is used to
adjust the  evaluation function coefficients such that the sum of the
error values is minimized.

Many authors report on machine
learning methods that train the coefficients against databases of
grandmaster games or in self-play~\cite{samuel1959some,baxter1999knightcap,campbell2002deep,quinlan1983learning,thrun1995learning,fogel2005further,nitsche1982learning,furnkranz1996machine,fogel2006blondie25,david2008genetic,tesauro1995td}.
Frequently, it is reported that the coefficients deviate considerably
from hand-tuned
coefficients (the machine-learned coefficients ``work, but don't
make sense''). Many Chess programmers have tried to find a middle
ground by adjusting the  coefficients to values that make intuitive
sense, and also have low error scores, after they have been optimized
by machine learning.

Most work in automated Chess evaluation optimization is in the automated tuning of
the coefficient of the features. There is also some work that goes a
step further, and tries end-to-end learning of evaluation 
functions
directly from board positions, without the intermediate step of hand-crafted heuristic features~\cite{david2016deepchess,silver2017mastering,lai2015giraffe}. This work is quite recent and often
related to the work reported in Chap.~\ref{chap:plan}. 

We are coming to the end of this overview of evaluation function
enhancements. There is one more to discuss: the endgame.


\subsection{End-Game Database}\index{end-game
  database}\index{tablebase}\index{pattern database}
In Chess and Checkers the endgame is an important phase of the game.
Inaccurate play can often result in loss of the game. It is also a
phase of the game where few pieces are on the board, allowing the
exact minimax value to be computed (without heuristic
approximation). Sometimes highly ``artificial'' lines of play turn out
to be the correct line of play, with  nontrivial and counter-intuitive
moves. Work on end-game databases was proposed early, e.g., by Bellman
\cite{bellman1965application}, and
Thompson~\cite{thompson1986retrograde}.
\index{Ken Thompson}

Much of the strength of world champion Checkers program Chinook came
from the 8-piece
end-game database~\cite{schaeffer1996chinook}. Chess program Deep
Blue used a database that contained 5-piece endings and some 6-piece
endings.

End-game databases are often computed by retrograde analysis, working
backwards from the
end~\cite{fang2000construction,lake1993solving,thompson1986retrograde}. Retrograde
analysis is used frequently in solving games (determining the outcome
in the case of perfect play by both sides, by full enumeration of
relevant parts of the
state space). Games that have been solved are Nine Men's
Morris~\cite{gasser1996solving}, Awari~\cite{romein2003solving}, and
Checkers~\cite{schaeffer2007checkers}.  

Culberson and Schaeffer generalized end-game databases to pattern databases, to
be used in all stages of a search~\cite{culberson1998pattern,korf2000recent}. Pattern
databases  play an important role in the playout phase of MCTS-based
Go programs (Sect.~\ref{sec:playout}).

\subsection{Opponent Modeling}\index{opponent modeling}
The minimax procedure assumes that our opponent uses the same
knowledge as we do. For perfect
(non-heuristic)  play, this is appropriate. When the values in the
state space are heuristic evaluations, i.e., imperfect approximations
of perfect play, then it may be reasonable to assume that our opponent
has a different play style than we do. Also for teaching purposes it
may be better to play other moves than the optimal policy. This {\em opponent
modeling\/} can  be used to exploit perceived weaknesses in the
opponent's play. Opponent modeling is often relevant in multiplayer
imperfect-information
games~\cite{billings1998opponent,ganzfried2011game}.  

\subsection{Trends in Eval }
We are now at the end of our discussion of heuristic evaluation
enhancements. Let us see if we can find a common trend in the enhancements.

Heuristic evaluation functions are domain specific. One would think
that enhancements of evaluation functions trend towards more specific
knowledge. This is not the case. The techniques that are used to
enhance evaluation functions trend towards finding automated methods for
feature creation: retrograde
analysis for creating end-game databases, machine learning of
coefficients or piece-square tables, and some variations trying to
overcome the search horizon problem. In many cases, the
heuristic knowledge is compartmentalized in features that are created
with automated methods. Furthermore, many
game programmers use automated testing whenever a change is made to
the code base. Even in search extensions
and quiescence search, whose  implementations
typically involve domain specifics, there have been efforts towards
general methods~\cite{beal1990generalised}.

The overall trend in evaluation function enhancements is that
enhancements start as domain specific, and then work is being undertaken
towards automated methods and automated learning.

\section{Conclusion}
\index{Arpad Elo}\index{Elo}
Determining the strength of Chess programs is of great importance. 
The playing strength of a program can be
determined by playing tournaments, and by computing a rating based on the
outcome and the strength of the opponent. A well-known internet tournament is
the Top Chess Engine Championship (TCEC), where strong computer Chess programs are compared and a
rating is calculated. TCEC can be found
 \href{https://tcec.chessdom.com/live.html}{here}.\footnote{\url{https://tcec.chessdom.com/live.html}}
In computer Chess the Elo rating is frequently used to compare  playing
strength.

The Elo rating was introduced by
Arpad Elo (Sect.~\ref{sec:elo}). It is calculated using tournament outcomes. For each game,
the winner's rating increases, and the loser's rating decreases. The
change depends on the difference between the original ratings. A win by a
stronger program increases its rating by less than a surprise win by a
weaker program. If player 1 has a rating $R_1$ and player 2
has a rating $R_2$ then the expected score of player 1 is
$E_1=\frac{1}{1+10^{(R_2-R_1)/400}}$. If the actual score of player 1
was $S_1$ then the new rating of player 1 is
$R_1^\prime=R_1+K(S_1-E_1)$, where $K$ is a constant that determines the
maximum adjustment per game. It is typically set from 16
for master players to 32 for weak players.\index{Elo rating}

Code for computing the Elo rating can be found at \href{https://github.com/ddugovic/BayesianElo}{BayesElo}.\footnote{\url{https://github.com/ddugovic/BayesianElo}}

\subsubsection*{Search-Eval and System 1 and System 2}
We have come to the end of this chapter on heuristic planning. The
search-eval architecture was originally inspired by the way in which
humans are thought to play a game. Can we
see a relation between the heuristic planning methods and human
thinking as it is described by Kahneman?




Since the earliest days of game playing research, the search-eval
architecture has been the cornerstone of game playing programs. The
architecture is simple. 
The search function traverses
possible game positions, and the evaluation function scores the
position with a heuristic value function. 
%
%
%
In the early days of Turing's and Shannon's papers, when search-eval was
conceived~\cite{turing1953digital,shannon1988programming}, the search
and eval functions were quite straightforward. The heuristic function
was the most basic  function that worked,  and the search function
was a minimax(-like) function to
traverse the nodes in standard depth-first order. This simple model proved quite
versatile and amenable to successive refinements. Over the years many
enhancements were  introduced, and  performance has reached world
champion level  in games such as Chess and
Checkers. These enhancements are very important---they elevate the level
of performance from being barely able to follow the rules to World Champion beating programs.



\begin{table}[t]
  \begin{center}

\begin{tabular}{lcccc}
  {\bf Chapter} & {\bf Algorithm} & {\bf Select} & {\bf Backup} & {\bf
                                                                  Eval}\\ \hline\hline
   Chap.~\ref{chap:search} & alpha-beta & left to right & minimax
                                   & heuristic \\ 
  Chap.~\ref{chap:trial} & MCTS & adaptive & average & sample \\ 
   Chap.~\ref{chap:dqn} & DQN & - & - & generalize \\ 
   Chap.~\ref{chap:plan} & self-play & adaptive & average &
                                                              generalize
  \\ \hline
\end{tabular}

  \caption{Heuristic-Sample-Generalize}\label{tab:heursampgen4}
  \end{center}
\end{table}


Table~\ref{tab:heursampgen4} gives an overview of the
different paradigms in this book. The entry for
Chap.~\ref{chap:search}   shows the heuristic search approach in 
reinforcement learning. 
The focus of  this chapter has been on  full-width search and manually crafted
heuristic evaluation functions, and value backup via the minimax
rule.
 
As the table indicates, in the following chapters different, more
adaptive, techniques will be introduced. In Chap.~\ref{chap:plan}
this will culminate in functions 
that are able to self-learn to play a game as complicated as Go with
 only the rules of the game known to the program.

At the time when early AI researchers such as Turing, Shannon, and Samuel discussed
Chess playing computers, it was believed that by developing better
Chess programs we would learn how
human thought worked, and we would learn  about human
intelligence. When 40 to 50 years later Deep 
Blue finally beat Kasparov, this belief had changed. It was said that brute force had won over
human intellect. We had not learned about how humans think, but we had
learned how to make a fast computer search many positions.

Kahneman's view on human thinking as consisting of fast and slow
points in another direction. Perhaps Chess computers do operate in a way that
is related to the way that humans think. Humans have a fast, heuristic, system 1, and a slow,
deliberative, system 2. This  corresponds  to Chess computers
that have a fast heuristic
evaluation function, and a slow planning function that painstakingly
reasons through some of the possible positions. In this sense there is a
correspondence to how humans and computers
think or play Chess. The search-eval architecture is
not only an architecture to organize the functions of a Chess
program in a clean way; it  also fits how Kahneman describes that humans play
and think. Perhaps, at a high level, Turing and
Shannon's dream of creating a mechanical model of thought has
succeeded after all.

Of course the ``brute force'' argument is not without merit. Computers
calculate fast and deep. It is
 true that computers always blindly follow their program, and do
not deviate. Human system 2 thinking is not so perfect, and is easily
distracted by system 1 associations. Indeed, computers can search many more
positions than humans can.

Computer Chess has also taught
us that  writing error-free heuristic programs is hard. It has taken four to five
decades of research by a dedicated community of researchers for
artificial intelligence  to rise to the challenge. Too much of this time has
been spent in frustration removing errors from and improving heuristic evaluation functions that
were hard to understand and hard to debug.

Conceptually the way in which computers and humans play Chess may be
highly related. In the details, there are important differences. 


Let us now review this chapter, and get our hands dirty with exercises
about a concrete Chess program.

\section{Practice}


\subsubsection*{Questions}
Below are some questions to check your understanding of this
chapter. Each question is a closed question where a simple, one
sentence answer is possible.

\begin{enumerate}
\item What is a heuristic?
\item Why is an exponential state space problematic?
\item How can you search an exponential search space efficiently?
\item What is the role of a search function?
\item What is the role of an evaluation function?
\item Why is minimax a form of self-play?
\item Why is minimax recursive?
\item What is the number of leaves of a regular tree of width $w$
  and depth $d$?
\item What is the critical tree?
\item What is the number of leaves of a regular tree that proves the
  value at the root?
\item What is a solution tree?
\item What is material balance?
\item What is Elo?

\item What is forward pruning?
\item What is backward pruning?
\item Draw a tree with an alpha-beta cutoff.
\item What causes the performance of alpha-beta to go from worst case
  to best case?
\item What is a transposition?
\item How do iterative deepening and transposition tables help the
  performance of alpha-beta?
\item What is a null window?
\item What is a null move?
\item What is a killer move?
\item What is a search extension?
\item How can you use supervised learning in a Chess program?
\end{enumerate}

\subsection{Sunfish---Chess in 111 lines of Python}\label{sec:sunfish}
Chess programs are often large pieces of code.  For Chess, there
are many open-source programs available that allow study of the code. However,
the size of the codes typically requires a large investment of study time. There
are also a few short  Chess programs. For example, the program
Micro-Max, by Harm Geert Muller, is a minimalist Chess program in 133
lines of C code, in less than 2000 characters (not counting
comments). Another example is Sunfish, a Chess
program in 111 lines of Python. Sunfish was written by Thomas
Ahle. Although the name is similar to Stockfish, Sunfish was
inspired by Micro-Max. You can download it from
GitHub \href{https://github.com/thomasahle/sunfish}{here}.\footnote{\url{https://github.com/thomasahle/sunfish}} The
entire program code fits in a single file, \verb|sunfish.py|.
\index{Sunfish}

The program consists of a heuristic evaluation function based on piece
square tables (Sect.~\ref{sec:piece}). It uses null window search
algorithm MTD-bi (Sect.~\ref{sec:mtdbi}) enhanced with iterative 
deepening, transposition tables, null moves, and killer moves.
Sunfish is small
enough to play around with and study, to learn how a real working
Chess program functions.

The small size and organization of the code also allow easy
modification. Things to try are, for example, changes to the
piece-square table, to see the effect on playing style of a different 
evaluation function (see the Exercises).



\subsection{Implementation: New or
  Make/Undo}\index{makemove}\index{undomove}
You may have noticed that the alpha-beta code in the figures lacks
implementation details such as board manipulation and move
making. These ``details'' are actually quite important for creating a
working program.
Recursive game playing programs call the search routine with the
current board state, often indicated with parameter \verb|n| for the
new node. This board can be created and allocated anew in each search
node, in a clean value-passing style (local variable). Another option is to pass a
reference to the board, and to apply a \verb|makemove| operation
placing the stone on the board
before the recursive call, and an \verb|undomove| operation removing
the stone from the board when it comes out of the recursion (global variable).

This reference-passing style may be quicker
if allocating the memory for a new board is an expensive operation. It may also be
more difficult to implement correctly, if capture moves cause many
changes to the board, which must be remembered for the subsequent
undo. Sometimes the incremental evaluation calculation is also incorporated into
the makemove/undomove routines, for even more efficiency.

For parallel implementations in a shared memory at least all parallel threads must have
their own copy of a value-passing style board. (On a distributed memory
cluster the separate machines will have their own copy of the board by
virtue of the distributed memory.)

\subsubsection*{Exercises}\label{sec:assignments}
\begin{enumerate}
\item Download and install Sunfish and BayesElo for Python. Play your first
game against Sunfish. Write a script that performs a tournament of two
identical instances with the same time controls of
Sunfish. Suggestion: 10 seconds per move. Look at the games. Do they
make sense? [Only if you know how to play Chess] Does Sunfish at this
time setting look like a strong program to you? Try different settings.


\item Setup BayesElo to allow you to run a tournament between two
  programs to determine their playing strength.

\item Running tournaments costs compute time. How long should a
tournament be? How many games are needed for results to be  statistically
significant? Add confidence intervals or error bars  to the graph
showing the standard deviation of the Elo rating.  

\item Make a change: halve the playing time of the second player. How
does the Elo rating change?

\item Read part 1 of Kahneman~\cite{kahneman2011thinking}, on
the two systems (less than a hundred pages of  reading. Part 2 is about
heuristics, but not the kind we have used here in our evaluation
functions). How is the thinking of a computer chess program related to
human thinking? Is minimax like system 1 or system 2? Is heuristic
evaluation like system 1 or system 2?
\end{enumerate}
Sunfish implements null-window search, null moves, iterative
deepening, transposition tables, and killer moves. Throughout this
chapter it was often claimed that certain enhancements work well.
Using the Sunfish and Elo code, the
following exercises are to 
    create a graph showing the effect of individual search
    enhancements.  In the exercises we study the influence of search enhancements and evaluation
heuristics.  The
baseline is standard Sunfish with all enhancements and all
heuristics. Keep in mind   how many runs of a program you need to get
statistically significant results~\cite{bartz2010experimental}.

\begin{enumerate}

\item Try the following: Replace MTD-bi by MTD$(f)$ (see Listing~\ref{lst:mtdf}). How is performance influenced?
If the tournaments take too long to run, what can you do to get
quicker results?\index{MTD($f$)} 

\item
Replace the null window search by wide-window
alpha-beta. How is performance influenced?

\item Replace the move ordering heuristics: put killer moves last
in the move list that is generated by the move
generator.

\item Show the Elo rating of the program without enhancements. Remove
  killer moves, then remove null moves, then try removing the
  transposition table and iterative deepening. Are the enhancements
  related?
  
  \item Study the code of the
 piece-square tables. Make a change to one of the squares, describe in words the intended
meaning, and compare the difference in play.

\item Create test positions that contain Zugzwang (every move makes
  your position worse; not moving is
  advantageous). Check if turning off null moves improves performance.
\end{enumerate}

\section*{Summary}
In this chapter we introduced  heuristic plannning reinforcement
learning algorithms. The main problem for thesse algorithms is the size of the
state space. We have calculated the size of the  state space, and found
that for most games a standard minimax enumeration of the full state space is
infeasible. Heuristic methods are introduced that can 
artificially reduce the state space. Static board evaluation functions do not capture game
dynamics, and search remains necessary. Thus we arrived at a solution
architecture consisting of two functions: the search-eval
architecture. 

The search functions covered in this chapter
have been conceptually simple, in the sense that they are directly
based on   maximization
and minimization of the reward values: I try to maximize my outcome;
my opponent tries to minimize my outcome.
We have discussed the minimax
procedure, and have looked at the concept of the critical tree, the smallest
set of nodes that has to be searched to be certain of the value of the
state space. This smallest tree concept has then been used to arrive at
efficient search algorithms.
The efficiency of both the search and the evaluation function can be
greatly enhanced with general and domain-specific enhancements. 
We covered a wide range of 
enhancements, both search enhancements and eval enhancements.
The
first search enhancement is alpha-beta, an algorithm that can search
twice as deep as minimax, if the tree is optimally ordered.

Achieving world champion level of performance with alpha-beta required
researchers and game programmers to invent many search
enhancements. For alpha-beta cutoffs to work effectively, the tree
must be ordered. Perhaps the most important enhancements for
alpha-beta are iterative
deepening and transposition tables. They  work
together by storing best-move information of previous searches. In
this way, alpha-beta searches a tree that is reasonably to
well ordered by the previous iterations. Null windows improve the
effectiveness of alpha-beta even further.

Another important area is relaxing  full-width search. A basic
approach is heuristic forward pruning. For Chess, it was found that a
better approach is to use a  mix of full-width and selective
search (extensions and reductions).

On the eval side, we discussed quiescence, search extensions,
piece-square tables, null moves,  end-game databases, and evaluation
function learning.
The first few levels of the tree are searched full width, so
that no move is left untried, and after a certain depth extensions (or
reductions) are used.

A popular heuristic is null-move pruning,
where before the legal moves an extra  pass move is
searched to a shallow depth. If a position is so strong that even
passing yields a cutoff, then it is safe to take the cutoff.

Researchers have a large imagination, and many ideas have been
tried. Not all have been successful in the minimax framework and in
the games in which they 
have been tried. Among them are forward pruning (in Chess and in Go),
different backup rules (such as star-minimax), Monte Carlo evaluation
(such as in Go), and neural network evaluation (in Chess and in
Go). Interestingly, as we will see in the upcoming chapters, although
these did not work in Chess all these
ideas have been used eventually to achieve a high level of play in different frameworks or games.

We noted a trend towards variable-depth search and automation in
evaluation. Enhancements typically start with an idea
to exploit a domain-specific feature (such as the killer move, or hand-tuned coefficients). After
some more research, often a more general version arises, that performs
even better (such as the history heuristic, or machine-optimized
evaluation coefficients).

In the next chapter we will continue the trend towards more
generally applicable methods and more selective search. We will see how Monte Carlo
evaluation, a 
different backup rule, and selective search together form an algorithm
that works well in  domains in which
minimax and heuristics do not work.

\section*{Historical and Bibliographical Notes}

One of the earlier works in heuristic planning is Pearl's book
\emph{Heuristics}~\cite{pearl1984heuristics} (he would later co-author {\em
  The Book of Why},  on his  work on causation~\cite{pearl2018book}). Minimax is described in
most standard artificial intelligence undergraduate texts, such as
Russell and Norvig~\cite{russell2016artificial}, although without
solution trees. 
There is an extensive literature on solution trees, see 
~\cite{kumar1988general,pijls1990another,stockman1979minimax,pearl1984heuristics,plaat1996best}.\index{solution
  tree}\index{max solution tree}

One of the first open-source Chess programs is Crafty~\cite{hyatt1997crafty}.
Chess programs have always used the most compute power that was available, and
many have stimulated research into parallel programs, where multiple
processors search the state space collectively. Deep Blue used special
hardware and software for parallel search. Parallel Chess has also
stimulated the development of research into parallel programming
environments, such as Cilk~\cite{leiserson1998programming}. The parallel
Cilkchess program achieved some success in its days. 
Fruit is an influential open source program by Fabien Letouzey; the
well-known open-source program GNU
Chess is now based on Fruit~\cite{fruit}.



%
%
%
%



The \href{https://www.chessprogramming.org/Main\_Page}{Chess
programming wiki} contains a wealth of information on null moves,
search extensions,  transposition
tables, and many, many, other things.\footnote{\url{https://www.chessprogramming.org/Main_Page}}

Alpha-beta is reported to have been reinvented
independently by several authors, among them 
McCarthy, 
Samuel,  Brudno, Richard,
Hart, and Levine~\cite{newellcomputer}. John McCarthy won the 1971
Turing award for work on AI.\index{John McCarthy}
A seminal analysis of the alpha-beta 
algorithm is Knuth and Moore~\cite{knuth1975analysis}. This analysis
was  without the benefit of the concept of
solution trees, which were introduced in later publications on 
Stockman's SSS*
algorithm~\cite{stockman1979minimax,pearl1984heuristics,pijls1990another}. The
relation between SSS*, solution trees, critical trees, and alpha-beta
is described in~\cite{plaat1996best,plaat1996research}.\index{SSS*}

Killer moves were described by Barbara Liskov (Huberman) in her 1968
thesis~\cite{huberman1968program}.
The history heuristic is described in Jonathan Schaeffer's thesis and
subsequent paper~\cite{schaeffer1989history}.
An overview paper of search enhancements
is~\cite{schaeffer2001unifying}. A  paper on search versus heuristics
is~\cite{junghanns1997search}.
Null moves are described by Donninger~\cite{donninger1993null}. Buro
introduced Probcut~\cite{buro1997experiments}.
Search pathologies, where searching deeper gives worse quality
answers, were introduced by Nau. They have been further 
studied by Pearl, Beal, and others~\cite{nau1980pathology,pearl1983nature,beal1983recent,wilson2009error,sadikov2005bias}.

\chapter{Adaptive Sampling}\label{chap:trial}
The previous chapter discussed heuristic planning. We saw the
basic search-eval architecture, and the  many enhancements that it needed for high performance.
The search enhancements were necessary
to overcome limitations of fixed-depth fixed-width minimax
search. Many of the
evaluation enhancements focused on alpha-beta cutoffs and the search horizon.

The two main challenges that are addressed in this chapter are (1) adaptive node selection
and (2) evaluation of states for which there is no efficient
heuristic evaluation function.

In minimax, all parts
of the tree are searched equally, no matter how unpromising. The
enhancements attempted to fix this, to allow the search to focus on promising
parts of the state space.  This approach  worked very well for games in
which tactical play 
is important, such as in Chess, Checkers, and Othello, where a single capture
move can often change the static score significantly. For other games,
notably Go, this approach did not work. Go is  more strategic in
nature, moves can have hidden long-term effects that are not easily found
with a short tactical search as in Chess. For Go, researchers  did not
find efficient heuristics for 
variable-depth and variable-width search, nor  for value estimation of states. 

Addressing these two problems led to a new algorithm, based on a new paradigm.
In this chapter the new algorithm, Monte Carlo Tree
Search (MCTS)~\cite{coulom2006efficient,browne2012survey}, is introduced. MCTS
combines three techniques: (1) it  performs
 adaptive exploration (it is not fixed-width fixed-depth), (2) it  uses sampling to determine an
evaluation (it does not use heuristics), and (3) it
  uses averaging as backup function of the statistics to aid node
selection (it does not  minimax). MCTS works well for Go and many other
applications, although for games such as Chess, Checkers, and Othello,
alpha-beta with a heuristic evaluation function performs better~\cite{browne2012survey}.

\section*{Core Problems}
\begin{itemize}
\item How to focus search effort where it is needed?
\item What if there is no efficient heuristic function?
\end{itemize}

\section*{Core Concepts}

\begin{itemize}
\item UCT adaptive node selection
\item Monte Carlo sampling
\item Average as  backup rule
\end{itemize}
First we will discuss the history and the challenges that led to the creation of Monte
Carlo Tree Search and how, after  decades of minimax, a different
paradigm emerged. Then we discuss the algorithm, including the UCT
selection rule, that governs the exploration/exploitation
trade-off. Finally, we discuss MCTS enhancements such as RAVE and
playout pattern databases. 

\section{Monte Carlo Evaluation}

When Kasparov was defeated in 1997 by Deep Blue, many computer game researchers went
looking for  their new \emph{Drosophila melanogaster}, a new test bed to help
make  progress in understanding intelligent reasoning.

Another game of strategy, the game of Go, became the focus of
much  research effort.
Creating a player for Go seemed a daunting task. The branching factor of Chess
is around 35 in the middle game; for Go it is around 200. The average
length of a Chess game is 80  moves by a player; for Go it is closer
to 250. The state space of Chess is estimated to be $10^{47}$;
for Go it is $10^{170}$.

Go programs of around that time played at amateur
level. They were designed like Chess programs but with forward pruning
because of the high branching factor: a highly
selective alpha-beta search, 
  and heuristic evaluations based on
the concept of territory, as humans are assumed to
do~\cite{muller2002computer}. The program GNU Go is a 
well-known example of this approach; typical designs of that time are described by
M\"uller, Chen and Chen, Boon, and Kierulf et al.~\cite{muller2002computer,chen1999static,boon1991overzicht,kierulf1990smart}

Since a full-width look-ahead search is infeasible due to the
branching factor, most early Go
programs used aggressive forward pruning-based heuristics: (1) generate a limited
number of heuristically likely candidate moves,\footnote{Such as: play
  in corner, connect via 2-point jump, and break ladder.} and (2) search
for the optimal policy in this reduced state space~\cite{muller2002computer}. A problem
with this approach is that too many good candidates are missed. For
evaluation, territory was calculated for each leaf. Such a territory
computation used slow flooding-style
algorithms and life/death calculations, much slower than the
piece-square tables in Chess, 
necessitating the highly selective search. A different 
approach was needed. 


In 1993, inspired by work on randomized solutions for the traveling
salesperson problem, physicist  Br\"ugmann wrote a $9\times9$ Go program
based on simulated annealing~\cite{brugmann1993monte}, and in 1990
Abramson explored  random evaluations in
Othello~\cite{abramson1990expected}. These programs did not have
a heuristic evaluation function, but played a series of so-called
{\em Monte Carlo playouts\/}:
random moves until the end of the game was reached, where the
position was scored as win or loss. The playouts were repeated many times, and results were
averaged for each top-level move.

\begin{figure}[t]
  \begin{center}
\begin{tabular}{c|c}
  \begin{tikzpicture}
\tikzstyle{max node}=[rectangle,draw,inner sep=6,very thick]
\tikzstyle{min node}=[circle,draw,inner sep=5,very thick]
\tikzstyle{cut node}=[rectangle,draw,fill,inner sep=2,very thick]
\node[max node](0){}
          child[sibling distance=20mm,very thick]{node[min node]{}
                         child[sibling distance=6mm,very thick]{node[max node]{}}
                         child[sibling distance=6mm,very thick]{node[max node]{}}
                         child[sibling distance=6mm,very thick]{node[max node]{}}}
          child[sibling distance=20mm,very thick]{node[min node]{}
                         child[sibling distance=6mm,very thick]{node[max node]{}}
                         child[sibling distance=6mm,very thick]{node[max node]{}}
                         child[sibling distance=6mm,very thick]{node[max node]{}}}
          child[sibling distance=20mm,very thick]{node[min node]{}
                         child[sibling distance=6mm,very thick]{node[max node]{}}
                         child[sibling distance=6mm,very thick]{node[max node]{}}
                         child[sibling distance=6mm,very thick]{node[max node]{}}};
                     \end{tikzpicture} 
\ \ \ \ \ & \ \ \ \ \
  \begin{tikzpicture}
\tikzstyle{max node}=[rectangle,draw,inner sep=6]
\tikzstyle{min node}=[circle,draw,inner sep=5]
\tikzstyle{cut node}=[rectangle,draw,fill,inner sep=1]
\node[max node,very thick](0){}
          child[sibling distance=20mm,thin]{node[cut node]{}
                         child[sibling distance=6mm,thin]{node[cut node]{}}
                         child[sibling distance=6mm,thin]{node[cut node]{}}
                         child[sibling distance=6mm,thin]{node[cut node]{}}}
          child[sibling distance=20mm,very thick]{node[min node]{}
                         child[sibling distance=6mm,thin]{node[cut node]{}}
                         child[sibling distance=6mm,thin]{node[cut node]{}}
                         child[sibling distance=6mm,very thick]{node[max node]{}}}
          child[sibling distance=20mm,thin]{node[cut node]{}
                         child[sibling distance=6mm,thin]{node[cut node]{}}
                         child[sibling distance=6mm,thin]{node[cut node]{}}
                         child[sibling distance=6mm,thin]{node[cut node]{}}};
                     \end{tikzpicture}
\end{tabular}  
  
  \caption{Searching a tree vs.\ searching a path}\label{fig:treepath}
  \end{center}
\end{figure}
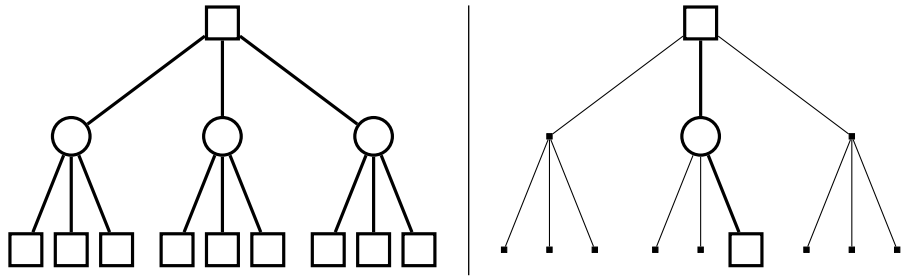

This approach differs fundamentally from the minimax
approach, where all moves at each node are searched, resulting in an exponential
search of a {\em tree\/} of size   $w^d$,  from which the best child is chosen. In Br\"ugmann's program, instead, 
the program searched  a number of linear {\em path\/}s of size $d$
(Fig.~\ref{fig:treepath}). Instead of using the minimax 
function, the program took the average of the scores. In  contrast to the
heuristic knowledge-based programs of the time, this program had no
heuristic knowledge. Although the program did not play great Go,  it played better
than 
random. The field of computer Go generally considered this attempt at
connecting the sciences of statistical mechanics and
artificial intelligence to be a curiosity. The hand-crafted
knowledge-based programs of the time performed  better. For  ten years not
much happened with Monte Carlo Go.



Then, in 2003 Bouzy and Helmstetter experimented again
with Monte Carlo playouts, again finding that a
program can play somewhat reasonable Go moves without a heuristic evaluation
function~\cite{bouzy2004monte}.

%
Three years later, with the introduction of Monte Carlo Tree Search
(MCTS),  a
breakthrough occurred: a new kind of recursive tree search 
was added to the Monte Carlo playouts. R\'emi Coulom 
introduced MCTS as a
tree-based algorithm~\cite{coulom2007monte,coulom2009monte,coulom2006efficient}.\index{R\'emi Coulom}\footnote{Following work by Chang et
al.~\cite{chang2005adaptive}, Auer et al.~\cite{auer2002finite}, and
Cazenave and Helmstetter~\cite{cazenave2005combining}.} 
Rules for node selection, expansion, playout,
and backup were specified.  

The early works by
Br\"ugmann~\cite{brugmann1993monte} and Bouzy and 
Helmstetter~\cite{bouzy2004monte} showed that averaged random playouts
do provide at the least an indication of the quality of a position. A ``flat''
algorithm, with playouts at the root only, did not provide great
results, but a recursive, tree based version, combined with a
smart exploration/exploitation selection rule, did.

Two of the first Monte Carlo programs were Sylvain
Gelly's MoGo~\cite{gelly2006modification} 
and R\'emi Coulom's
Crazy Stone~\cite{coulom2006efficient}.\index{MoGo}\index{Crazy Stone}\index{Sylvain Gelly}
%
Crazy Stone and MoGo were
instantly successful, and many other new Go playing programs were written
based on
MCTS~\cite{coulom2007monte,chaslot2008monte,chaslot2010monte,muller2009fuego,enzenberger2010fuego,baudivs2011pachi,gelly2012grand,rimmel2010current,lee2010special}.\index{MCTS}
Since the introduction of MCTS the playing strength of programs improved rapidly to the
level of weak master (2-3 dan) and stronger on the small $9\times 9$ board.

MCTS is also successful in other domains, and there has been great
research interest. 
Browne et al.~\cite{browne2012survey} provide an extensive survey, referencing 240 publications.

\begin{table}[t]
  \begin{center}

\begin{tabular}{lcccc}
  {\bf Chapter} & {\bf Algorithm} & {\bf Select} & {\bf Backup} & {\bf
                                                                  Eval}\\ \hline\hline
   Chap.~\ref{chap:search} & alpha-beta & left to right & minimax
                                   & heuristic \\ 
  Chap.~\ref{chap:trial} & MCTS & adaptive & average & sample \\ 
   Chap.~\ref{chap:dqn} & DQN & - & - & generalize \\ 
   Chap.~\ref{chap:plan} & self-play & adaptive & average &
                                                              generalize
  \\ \hline
\end{tabular}

  \caption{Heuristic-Sample-Generalize}\label{tab:heursampgen5}
  \end{center}
\end{table}

\subsubsection*{Reinforcement Learning}
In this chapter we look deeper into MCTS.
First, let us see how MCTS fits into the general picture of reinforcement
learning methods for games (Table~\ref{tab:heursampgen5}).
MCTS is a reinforcement learning algorithm, as are heuristic planning
algorithms. In the Introduction we mentioned three basic paradigms:
heuristics, sampling, and generalization. MCTS represents the sampling
method. All three methods are reinforcement learning methods, and with
MCTS we see the second main paradigm of reinforcement learning.

\begin{table}[t]
  \begin{center}
\begin{tabular}{lccl}
  {\bf Chapter} & {\bf Name} & {\bf MDP-tuple} & {\bf Reinforcement learning}\\ \hline\hline
    Chap.~\ref{chap:search} & alpha-beta & $(S, A, 1, R, 1)$ &
                                                                   policy, backup \\ 
   Chap.~\ref{chap:trial} & MCTS & $(S, A, P, R, 1)$
  & pol, b/u, exploration/exploitation\\ 
   Chap.~\ref{chap:dqn} & DQN  &  $(S, A, P, R, \gamma$)& pol, b/u,
                                                             expl/expl,
                                                             discount, off-policy
  \\ 
   Chap.~\ref{chap:plan} & self-play & $(S, A, P, R, \gamma)$&pol, b/u,
                                                             expl/expl,
                                                                discount,
                                                                off-pol, self-play \\ \hline
\end{tabular}

  \caption{MDP tuple and reinforcement learning in the chapters}\label{tab:mdp2}
  \end{center}
\end{table}

Likewise, in Chap.~\ref{chap:rl}, the Markov decision process 5-tuple was
introduced. Table~\ref{tab:mdp2} repeats the table from that chapter. We
see how the exploration/exploitation trade-off is an integral part of
MCTS, in contrast to minimax, where domain-specific enhancements were
needed to achieve variable-depth search. With MCTS, we are using more
elements of the MDP tuple in the core of the paradigm.

\section{Monte Carlo Tree Search}
\begin{figure}[t]
    \centering{
    \includegraphics[width=12cm]{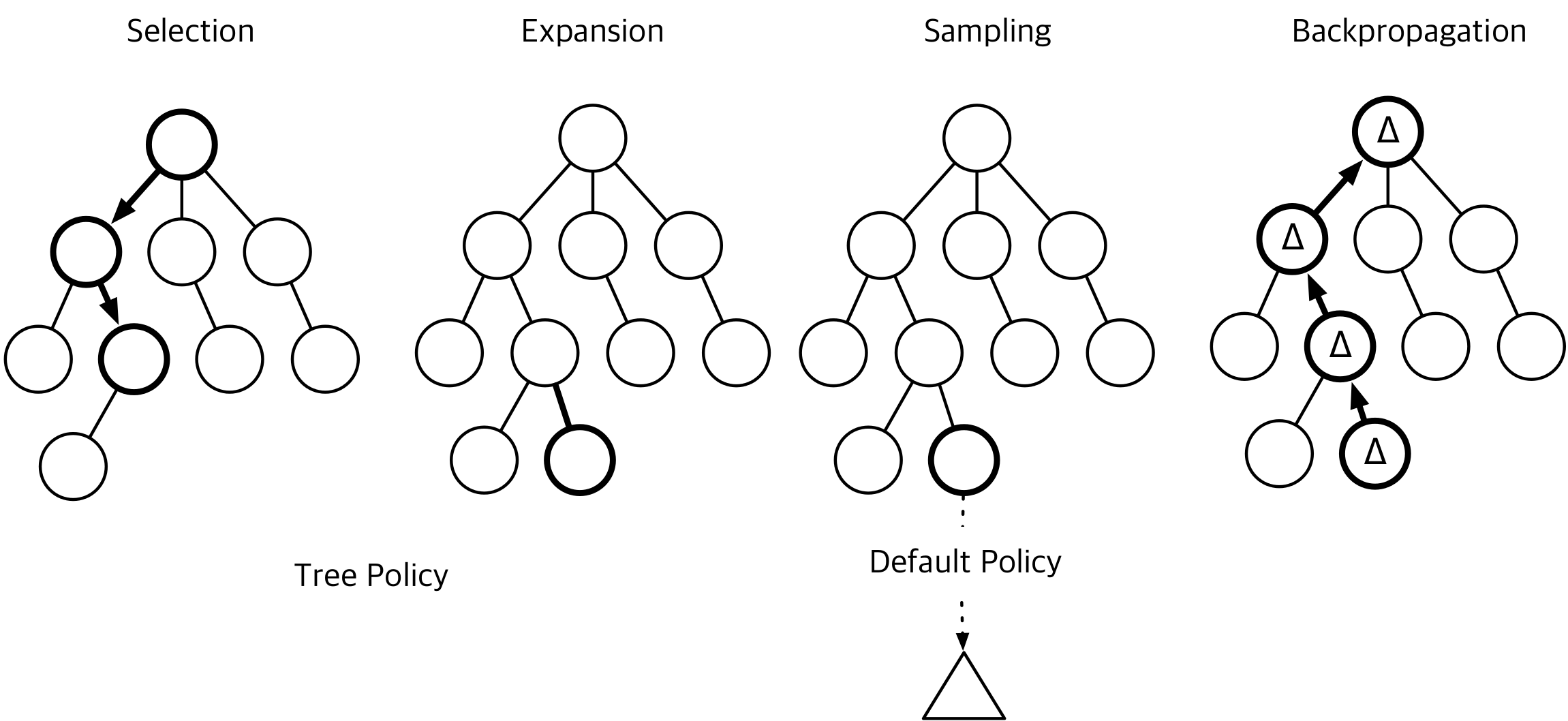}
     \caption{Monte Carlo Tree Search \cite{browne2012survey}}\label{fig:mcts}}
\end{figure}

\index{Monte Carlo Tree
  Search}\index{MCTS}\label{sec:mcts}
We will now look at the MCTS algorithm in depth. MCTS consists of four
operations: select, expand, playout, and backpropagate (Fig.~\ref{fig:mcts}). Note that the third operation (playout)
sometimes goes by the names {\em rollout}, {\em simulation}, and {\em sampling}. Back-propagation
is sometimes called {\em backup}. Select is the downward policy trial
part, backup the upward
error/learning part of the algorithm. Soon, we will discuss the operations
in order.

\subsubsection*{Iteration vs.\ Recursion}
As in heuristic planning, in MCTS the state space is tree shaped,
starting from the initial state $s_0$, using the rules of the game to
generate successor states. Unlike in heuristic planning, where the
tree is built by recursively adding subtrees, in MCTS the state
space is traversed iteratively, and the tree data structure is built in a
step by step fashion, node by node. MCTS is a mix between a tree and a
path algorithm. A typical size of an MCTS search
is to do 1000--10,000 iterations. In MCTS each iteration starts at
the root $s_0$, traversing the tree down to the leaves using a
selection rule, expanding an extra node, and performing a playout. The
result of the playout is then propagated back to the root. During the
backpropagation, statistics at all internal nodes are updated. These
statistics are then used in future iterations by the selection rule
to go to the currently ``most interesting'' part of the tree.

The statistics consist of two counters: the win count $w$ and the visit
count $v$. During backpropagation, the visit count $v$ at all nodes
that are on the path back from the leaf to the root are
incremented. When the result of the playout was a win, then the win
count $w$ of those nodes is also incremented. If the result was a loss,
then the win count is left unchanged.

The selection rule uses the win rate $w/v$ and the visit count $v$ to decide
whether to exploit high-win-rate parts of the tree or to explore
low-visit-counts parts. An often used selection rule is UCT
(Sect.~\ref{uct}). It is this selection rule that governs  
the exploration/exploitation trade-off in MCTS.

\subsection{The Four MCTS Operations}
Let us now look in more detail at the four operations. Please refer to
Listing~\ref{lst:mcts} and Fig.~\ref{fig:mcts}.
%
%
As we see in the figure and the listing, the main 
steps are repeated as long as there is time left. Per step, the activities 
are as follows~\cite{browne2012survey}.

\lstset{label={lst:mcts}}
\lstset{caption={MCTS pseudo-Python \cite{browne2012survey,czarnogorski2018monte}}}
\begin{lstlisting}[language=Python,float]
def monte_carlo_tree_search(root):
    while resources_left(time, computational power):
        leaf = select(root) # leaf = unvisited node 
        simulation_result = rollout(leaf)
        backpropagate(leaf, simulation_result)
    return best_child(root) # or: child with highest visit count

def select(node):
    while fully_expanded(node):
        node = best_child(node)    # traverse down path of best UCT nodes
    return expand(node.children) or node # no children/node is terminal 

def rollout(node):
    while non_terminal(node):
        node = rollout_policy(node)
    return result(node) 

def rollout_policy(node):
    return pick_random(node.children)

def backpropagate(node, result):
    if is_root(node) return 
    node.stats = update_stats(node, result) 
    backpropagate(node.parent)
   
def best_child(node, c_param=1.0):
    choices_weights = [
        (c.q / c.n) + c_param * np.sqrt((np.log(node.n) / c.n))    # UCT
        for c in node.children
    ]
    return node.children[np.argmax(choices_weights)]
\end{lstlisting}

\subsubsection*{Select}\index{MCTS!select}
In the selection step the tree is traversed from the root node down
until a leaf is reached where a new child is selected that is not part of
the tree yet. At each internal state the selection rule is
followed to determine which action to take and thus which state to
explore next. The UCT rule is the best-known  
selection rule~\cite{kocsis2006bandit}.

The selections at these states constitute the policy $\pi$ of
actions in the look-ahead simulation leading to the state that currently is the best. 

\subsubsection*{Expand}\index{MCTS!expand}
Then, in the expansion step a child is added to the tree. In most
cases only one child is added. In some MCTS versions all successors of a leaf
are added to the tree~\cite{browne2012survey}.

\subsubsection*{Playout}\index{MCTS!playout}\label{sec:playout}\index{pattern database}
Subsequently, during the playout  step moves are played
in self-play 
until the end of the game is reached.\footnote{Note that this self-play is like the self-play in Chap.~\ref{chap:search} where in
  minimax and alpha-beta the opponent moves for finding the best reply are
  found  by the same algorithm. Unlike minimax, here only one,
  randomly chosen, successor is played out: a path, not a tree. In Chap.~\ref{chap:plan}  we will see
  a different kind of self-play, where AlphaZero uses self-play loops to
  train a neural network evaluation function from scratch.} (These
node are not added to the MCTS tree, but their search result is, in
the backpropagation operation.) The reward $r$ of 
this simulated game is $+1$ in case of a win for the first 
player, $0$ in case of a draw, and $-1$ in case of a win for the
opponent. Originally, playouts were random (the Monte Carlo part in
the name of MCTS) following
Br\"ugmann's~\cite{brugmann1993monte} and Bouzy and
Helmstetter's~\cite{bouzy2004monte} original approach. In practice,
most Go playing programs improve on the random playouts by using
databases of small $3\times 3$ patterns with best
replies and other fast heuristics~\cite{gelly2006modification,coulom2007monte,chaslot2010monte,silver2009reinforcement,culberson1998pattern}. 

Small amounts of domain knowledge are used after all, albeit not in the form of a heuristic evaluation function.

\subsubsection*{Backpropagation}\index{MCTS!backup}
In the backpropagation step, reward $r$ is propagated back upwards in
the tree, through the 
nodes that were traversed down previously. Note that two counts are
updated: the visit count, for all nodes, and the win count, depending
on the reward value.


MCTS is on-policy: the values that are backed up are those of the
nodes that were selected (except for RAVE, at the end of this chapter). 



\subsubsection*{Code}

Many websites
 contain
useful 
resources on MCTS, including  example
\href{https://int8.io/monte-carlo-tree-search-beginners-guide/}{code}
(see Listing~\ref{lst:mcts}).\footnote{\url{https://int8.io/monte-carlo-tree-search-beginners-guide/}} The
 pseudocode is  
from an example Tic Tac Toe program (only the main methods
are shown). The MCTS algorithm can be coded in many different
ways. Listing~\ref{lst:mcts}~\cite{czarnogorski2018monte} follows the survey paper of Browne et al.~\cite{browne2012survey}.  The
four main operations of MCTS are clearly indicated. 

\subsection{Policies}
At the end of the search, after the required number of iterations has been
performed, or when time is up, MCTS returns the value and the action
with the highest visit count.\footnote{An alternative would be to return
  the action with the highest win rate. However, the visit count takes
  into account the win rate (through UCT) and the number of simulations on which it is based. A high
  win rate may be based on a low number of simulations, and will thus
  be high variance. High visit count will be low variance. High
  visit count implies high win-rate with high confidence, high
  win rate may be low confidence~\cite{browne2012survey}.}  The action of this initial state $s_0$
constitutes the policy $\pi(s_0)$  (Sect.~\ref{sec:policy}).

It should be pointed out that there are multiple policies in MCTS, and
the word
``policy'' is used to describe these different kinds. During the
MCTS look-ahead simulation a second policy was used. The actions selected by
the selection rule in the tree form a selection
policy. Additionally, the actions in the playout phase are also a
policy. This third type of policy is sometimes called the {\em default policy}.

\subsection{Example}
\begin{figure}[t]
\begin{center}
\begin{tikzpicture}
\tikzstyle{max node}=[rectangle,draw,inner sep=4,minimum size=0.6cm]
\tikzstyle{min node}=[circle,draw,inner sep=2]
\node[max node,label=left:{$r$}](0){$\ ^{w:1,1,2,3}_{v:1,2,3,4}$}
          child[sibling distance=30mm]{node[min
            node,label=left:{$a$}]{$\ ^{w:1,2,3}_{v:1,2,3}$}
                         child[sibling distance=15mm]{node[max node,label=left:{$c$}]{+}}
                         child[sibling distance=15mm]{node[max node,label=left:{$d$}]{+}}}
          child[sibling distance=30mm]{node[min
            node,label=left:{$b$}]{$\ ^{w:0}_{v:1}$}
                         child[sibling distance=15mm]{node[max node,label=left:{$e$}]{$-$}}
                         child[sibling distance=15mm]{node[max node,label=left:{$f$}]{\quad}}};
\end{tikzpicture}
\caption[MCTS example]{
MCTS starts at the root, does not select a move because there are no
children in the tree, but expands a random child. Assume $a$ is
expanded. Then a random playout is performed at node $a$. Assume
node $d$ is the first node in the playout. In this
tiny state space, the random path that is played out consists
only of node $d$ and is a terminal node. The value at node $d$ is a
win, so backpropagation increments all win values of the path to the
root by 1 (nodes $a,r$) and also the visit counts. The second
iteration starts at  root $r$, where there are unexpanded children, so
it expands a node, which is node $b$. At $b$ a playout is
performed, which ends up at, say, $e$, whose value is a loss. This
gets propagated (no win values change, and the visit counts  
of $b, r$ are incremented). Next, the third iteration starts. At the
root all children are expanded, so we select a child. We select $a$
since its UCT value is $1/1+1\times\sqrt{\ln 2/1}=1.83$, and $b$'s value is
$0/1+1\times\sqrt{\ln 2/1}=0.83$. There are unexpanded children so it
expands a node, assume it picks node $c$, a win. The values are updated ($a$'s win and visit
count are incremented to 2 and 2, $r$'s to 2 and 3). Next, the fourth iteration selects
node $a$ since the UCT value of $a$ is $2/2+1\times\sqrt{\ln
  3/2}=1.63$ and $b$ is $0/1+1\times\sqrt{\ln 3/1}=1.04$. It finds
that there are unexpanded children of $a$ in the tree, so it expands
child $d$, a win, and updates the win and visit
values of $d, a,$ and $r$ accordingly. It continues with more iterations
until time is up.}\label{fig:mctsex} 
\end{center}
\end{figure}
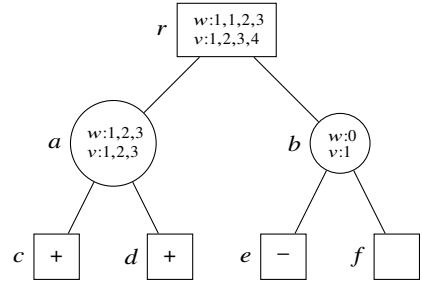

Let us illustrate the working of MCTS with an
example. Figure~\ref{fig:mctsex} gives an example of a few MCTS
iterations. (The UCT formula is described in the next subsection.)
In the example we see how MCTS traverses the state space in
iterations, how the UCT formula adapts between exploiting high win
values and exploring unseen parts of the tree, and how the statistics
are averaged in the backpropagation phase.

\subsection{UCT Selection Rule}
The adaptive exploration/exploitation behavior of MCTS is governed by the
selection rule. One of the most popular selection rules is the UCT formula.


UCT was introduced in 2006 by  Kocsis and
Szepesv\'ari~\cite{kocsis2006bandit}.\index{Levente Kocsis}\index{Csaba Szepesv\'ari}
The paper provides a theoretical guarantee of
eventual convergence to the minimax
value. The selection rule was named UCT,  for Upper Confidence
bounds for multi-armed bandits applied to Trees.
\index{UCT}\label{uct}

The
exploration/exploitation trade-off is central in many reinforcement
learning algorithms. The selection rule determines the way in which the
current values of the children influence which parts of the tree will
be explored more.   The UCT formula
is 
$$
 \mbox{UCT}(j)=\frac{w_{j}}{n_{j}}+C_{p}\sqrt{\frac{\ln n}{n_{j}}} ,
$$
where $w_{j}$ is the number
of wins in child $j$, $n_{j}$ is the number of times child $j$ has
been visited, $n$ is the number of times the parent node has been
visited, and $C_{p}\geq0$ is a constant (the tunable
exploration/exploitation parameter). The first term in the UCT
equation, the win rate, is the exploitation term. A child with a high
win rate receives  through this term an exploitation bonus. The second term is for
exploration. A child that has been visited infrequently has an exploration bonus. The level of exploration can be
adjusted by the $C_{p}$ constant.  A low $C_p$ does little exploration;
a high $C_p$ has more exploration. Section~\ref{sec:explexpl} goes
deeper into the choice for values for $C_p$. The selection rule then is to
select the child with the highest UCT sum (the standard argmax function).

The UCT formula balances {\em win rate\/} ($\frac{w_{j}}{n_{j}}$) and
{\em ``newness''\/} ($\sqrt{\frac{\ln n}{n_{j}}}$) in the selection of
nodes to expand.\footnote{The square-root term is a measure of the
  variance (uncertainty) of the action value. The use of the natural
  logarithm ensures that, since increases get smaller over
  time, old actions are selected less frequently. However, since logarithm values
  are unbounded, eventually all actions will be 
  selected~\cite{sutton2018introduction}.} Many alternative selection rules have been
proposed. Alternatives for UCT are variants such as 
Auer's UCB1~\cite{auer2002using,auer2002finite,auer2010ucb} and
P-UCT~\cite{rosin2011multi,matsuzaki2018empirical}. Most rules are
based on bandit theory, which we will briefly touch upon next.

\begin{figure}[t]
    \centering{\includegraphics[width=4cm]{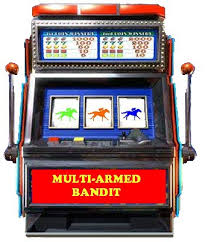}
     \caption{A multi-armed bandit}\label{fig:bandit}} 
\end{figure}

\begin{figure}[t]
    \centering{\includegraphics[width=6cm]{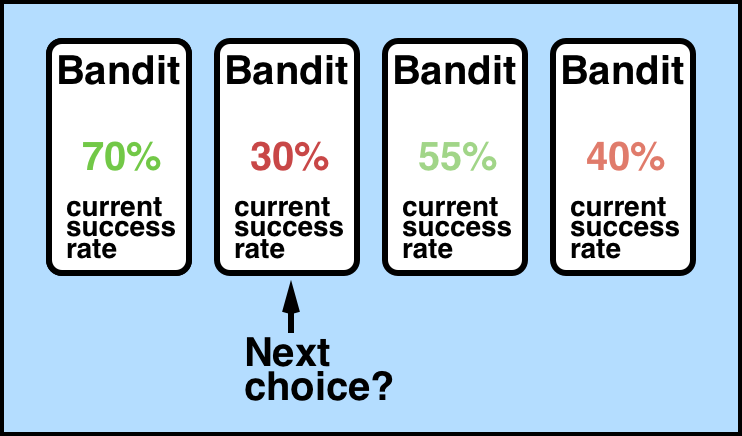}
     \caption{Explore or exploit? \cite{wong2017}}\label{fig:choice}}
\end{figure}

\subsubsection*{Multi-armed Bandit Problem}\index{multi-armed bandit}
The work on UCT and upper confidence bounds (UCB) is based on bandit theory, a topic well studied in the
field of stochastic
scheduling and decision theory~\cite{auer2002finite,katehakis1987multi,chang2005adaptive}.
The theory provides a systematic means to optimize the search
strategies to explore more promising parts of the tree, while still
ensuring exploration. Seminal work has
been done by Lai~\cite{lai1985asymptotically,lai1995machine}, proving
important optimality results. 

A  
multi-armed bandit is a mathematical model whose name refers to a
casino slot machine, but with more 
than one arm (Fig.~\ref{fig:bandit}). Each arm can have a different payout. The probability
distribution of the payout of the arms is also not known. As arms are
played, the player gets more information as to their payout
probability. However, each play also costs money. The multi-armed
bandit problem then is to select the best arm in as few tries as
possible (Fig.~\ref{fig:choice}). Multi-armed bandit theory thus
analyzes situations of
short-term cost versus long-term gain. 

Consider the following situation: One arm gives 10 dollars with
probability 0.1, while 
another gives 1000 dollars with probability 0.002.
After 50 tries, the first arm probably yielded some win, whereas the
second probably did not. You may then want to play mostly
the first arm, but this is not optimal in the long run, since the
expected reward of the second one is actually twice as high. But to
learn this, you will need to sample the second arm for quite a while. 

Multi-armed bandits have a classic reinforcement learning
exploitation/exploration trade-off: should I play the arm that I know
to give  high payout so far, or should I play arms whose payout I do
not know but may have higher payout? In a  paper Lai and
Robbins~\cite{lai1985asymptotically} construct an optimal  selection policy
(optimal in the sense that it has maximum convergence rate to the arm with the highest
mean) for a case where the  reward distributions of the arms
is known. An active field of research subsequently studied relaxations
of this condition.

In
MCTS the multi-armed bandit problem is applied in a recursive setting,
in a tree of decisions. Each state is a bandit, and each action
(child) is an arm.



\subsection{The Impact of MCTS}
Before 2006, mainstream game 
programming research was based on heuristic planning. The minimax
backup rule was the main paradigm, spawning a large effort to formalize and
code heuristic human
knowledge. The conventional wisdom was that human knowledge should be
captured in heuristics, and that no move should be missed (due to the
tactical nature of the popular perfect-information games that were
used at the time). There was
no overarching principle behind the many heuristic enhancements,
except that they should be efficient to code, and a means to make
the search more selective should be tried, in order to overcome the
horizon problem.

When MCTS was introduced, it provided an
elegant framework for selective search, amenable to theoretical
analysis and successful in practice. MCTS follows a different,
non-exponential paradigm. Minimax is based on searching all successor
states and is an inherently exponential tree searcher, with a
$\sqrt{w^d}$ best-case time complexity for alpha-beta. MCTS is a sampling algorithm. It searches 
paths from the root to the terminals, not trees. This non-exponential principle circumvents
the need for the plethora of heuristic search enhancements trying to
curtail the  exponential time complexity of minimax. Equally important, the sampling also removed the need for
heuristic evaluation.

The success of the new approach of selective playout sampling
created  a paradigm shift in the world of computer game playing.  In
Go, and in many other games, it turned 
out that some moves 
{\em can\/} be missed, or rather,  that {\em most\/}  moves can be missed, as long
as on average a reliable 
picture of the state space emerges. The  paradigm changed 
from an exponential $O(w^d)$ to a polynomial  $O(d)$
complexity. Exponential worked for Chess; polynomial worked for
Go.\footnote{In practice, both minimax and MCTS are made into anytime
  algorithms that produce a best move when stopped. For minimax,
  iterative deepening transforms it into an anytime algorithm; for
  MCTS, the algorithm is already an anytime algorithm that performs an iteration of playouts.} The new paradigm worked
so well that no human knowledge in the form of heuristics was
needed (or at least, initially). 

The dominant
language of the field changed with the introduction of MCTS. The
graph-based language of nodes, leaves, and strategies changed to state, action, and
policy.


\subsubsection*{Applications of MCTS}
The introduction of MCTS improved performance of Go programs
considerably, from medium amateur to strong amateur. Where
heuristics-based GNU Go played around 10 kyu, Monte Carlo programs
progressed to 2-3 dan\footnote{Absolute beginners
  in Go start at 30 kyu, progressing to 10 kyu, and advancing to 1
  kyu. Stronger amateur players then
  achieve 1 dan, progressing to 7 dan, the highest amateur rating for
  Go. Professional Go players have a rating from 1 dan to 9 dan,
  written as 1p--9p.}\index{kyu (Go rank)}\index{dan (Go rank)}  in a few years' time.

\begin{figure}[t]
    \centering{\includegraphics[width=8cm]{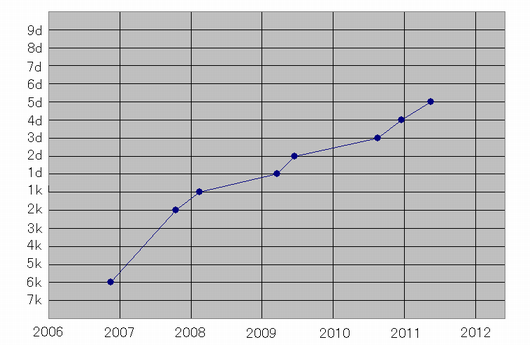}
     \caption{Performance of MCTS programs~\cite{gelly2012grand}}\label{fig:mctsperf}}
\end{figure}

On the small board, $9\times 9$, Go
programs achieved very strong play. 
Figure~\ref{fig:mctsperf} shows a graph of the improvements in the
performance of Go programs on the large $19\times 19$
board~\cite{gelly2012grand}. Performance did not improve much beyond 
the 4-5 dan level, despite
much effort by researchers. It was thought that perhaps the large action space
of the $19\times 19$ board was too hard for MCTS. Many enhancements
were considered, for the playout phase, and for the selection. One of
the enhancements that was tried was the introduction of neural
nets. As we will see in Chap.~\ref{chap:plan}, in 2015 this became successful.

MCTS revolutionized the world of heuristic search. Previously, in order to achieve best-first search, one had
to find a domain specific heuristic to guide the search in a smart
way. With MCTS this was no longer necessary. Now a general method
existed that could find the promising parts of the search without a
domain-specific heuristic, just by using statistics of the search
itself.

MCTS quickly proved successful in other games, both two agent and
single agent: for video games~\cite{chaslot2008monte}, for single
player applications~\cite{schadd2008single}, and for many other games
such as EinStein
w\"urfelt nicht~\cite{lorentz2011mcts},
Settlers of Catan~\cite{szita2009monte},
Havannah~\cite{lorentz2011experiments},
Amazons~\cite{kloetzer2010monte}, and mathematical games such as the
Voronoi game~\cite{bouzy2011mcts}.

\begin{table}[t]
  \begin{center}

\begin{tabular}{lccc}
  {\bf  Name} & {\bf Principle} & {\bf Applicability} & {\bf Effectiveness}  \\ \hline\hline
  expl/expl & expl/expl/N & small/large searches & all MCTS  \\ 
  RAVE & statistics sharing & early phase & all MCTS   \\ 
  play-out db & small patterns & domain specific & domain dependent \\ 
  adaptive expl/expl & variable $C_p$ & general  & all MCTS  \\   
  parallelism & parallel search & general  & all MCTS  
  \\ \hline
\end{tabular}

  \caption{MCTS enhancements}\label{tab:mcts-enh}
  \end{center}
\end{table}

\section{*MCTS Enhancements}
Basic MCTS already showed much promise, and researchers  immediately found
ways to improve performance of the basic algorithm further. They 
focused on action selection and on playout. We will discuss some of
these enhancements.
Table~\ref{tab:mcts-enh} shows the MCTS enhancements that we will discuss.

First we will discuss tuning
of $C_p$, the exploration/exploitation parameter, and then we will
discuss RAVE, a method to speed up the distribution of statistics
at the early stage of an MCTS search. After that, we will discuss
playout databases, adaptive $C_p$, and parallelism.

\subsection{Exploration/Exploitation}\label{sec:explexpl}\index{$C_p$}

\begin{figure}[t]
    \centering{\includegraphics[width=12cm]{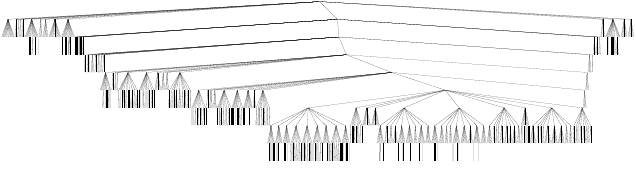}
     \caption{Adaptive MCTS tree~\cite{kuipers2013improving}}\label{fig:mctstree}}
\end{figure}

The search process of MCTS is guided by the statistics values in the tree. MCTS
discovers during the search where the promising parts of the tree
are. The tree expansion of MCTS is an inherently variable-depth and
variable-width search process. To illustrate this we show a picture of
a snapshot of the search tree of an MCTS 
optimization from~\cite{kuipers2013improving,vermaseren2014investigations}. In
Fig.~\ref{fig:mctstree} 
we see that some parts of the tree are searched more deeply than
others. The tree in the illustration is part of an expression optimization problem
in a symbolic algebra system~\cite{vermaseren2000new}. 

An important element of MCTS is the exploration/exploitation trade-off that
can be tuned with the $C_p$ parameter. The effectiveness of MCTS 
depends  on the choice of this parameter. Typical initial choices for
Go programs are $C_p = 1$ or $C_p=0.1$; programs that use RAVE found
$C_p=0$ to works best~\cite{browne2012survey}, although in AlphaGo we
will see highly explorative choices such as $C_p=5$. In general, when
compute power is low,  $C_p$ should be low, and when more compute
power is available, more exploration (higher $C_p$) is advisable.

\begin{figure}[t]
    \centering{\includegraphics[width=12cm]{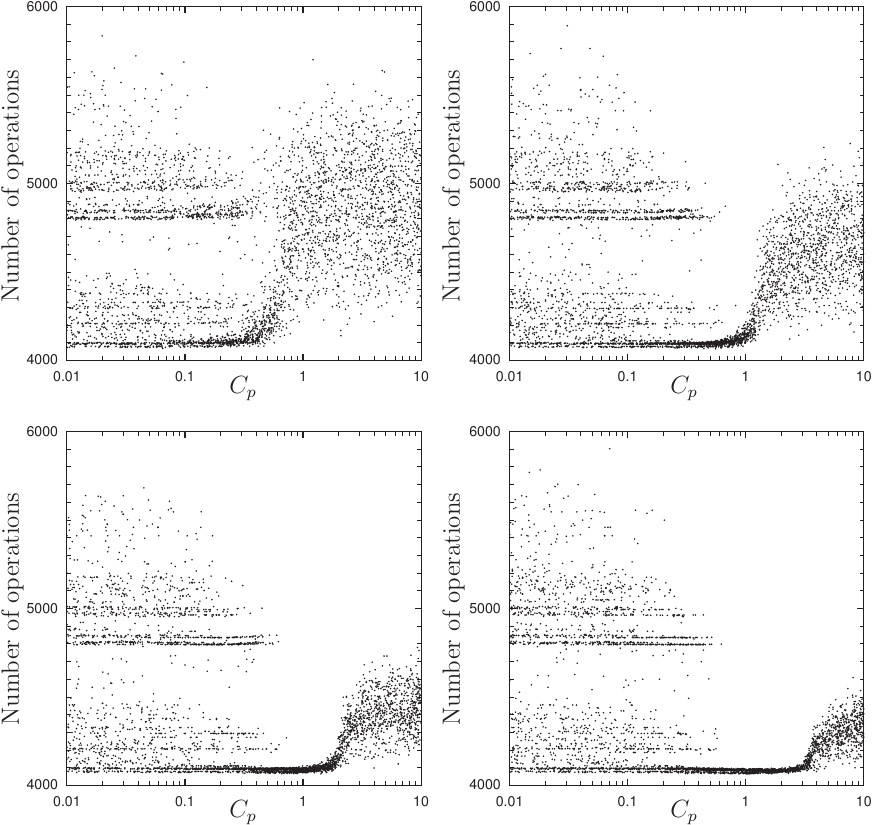}
     \caption[Four scatter plots]{Scatter plots of $C_p$
       for $N$ = 300,
       1000, 3000, and 10,000 node expansions per MCTS run. Each dot is one
       MCTS run to optimize a large polynomial. Lower is
       better. Exploitation (small $C_p$) works best in small trees ($N=300$),
       exploration in large trees ($N=10,000$)~\cite{kuipers2013improving}}\label{fig:mctscp}}
\end{figure}

Kuipers et
al.~\cite{kuipers2013improving,vermaseren2014investigations} have performed experiments
to plot the relation between $C_p$ and the number of MCTS node expansions
$N$. When $C_p$ is small, MCTS favors parts of the tree that have been
visited before and are known to be good; when $C_p$ is large,
unvisited parts of the tree are favored. Figure~\ref{fig:mctscp} shows
four plots of an optimization for a large  symbolic algebra
polynomial. The plots
show results from   a large number of optimization runs,  all starting
with a different random seed. On the $y$-axis the number of operations of the optimized
polynomial is shown; the lower this number is, the better.  The lowest
number found is close to $4000$. This minimum is achieved in the case
of $N = 3000$ iterations for a value of $C_p$ with $0.7 \leq C_p \leq
1.2$. Dots above this minimum represent a suboptimal search result.  

For  300  iterations per data
point (left upper panel), some  structure  is visible, with a minimum
emerging at $C_p \approx 0.4$. With more tree expansions  
(see the other three panels) the picture becomes clearer,
and the value for $C_p$ for which the best answers are found becomes
higher; the picture appears to shift to the right. For low
numbers of tree expansions (see again upper left panel)
there is almost no discernible advantage of setting the
exploration/exploitation parameter at a certain value. For 1000
iterations (see upper right panel) 
MCTS works best when exploitative (the left part of the plot achieves
the lowest number of operations). As the number
of iterations $N$ is larger (the two lower panels) MCTS
achieves better results when its selection policy is more
explorative, to try to get out of local minima. For the graphs with
iterations of 3000 and 10000 the range of good results for $C_p$
becomes wider, indicating that the choice between
exploration/exploitation becomes less critical.

For small values of $C_p$, such that MCTS behaves exploitatively, the
method gets trapped in one of the local minima as can be seen from
scattered dots that form “lines” in the left-hand sides of the four
panels. For large values of $C_p$, such that MCTS is more
explorative, many of the searches do not lead to the global minimum
as can be seen from the cloud of points on the right-hand side
of the four panels. For intermediate values of $C_p \approx 1$, MCTS balances
well between exploitation and exploration and finds almost always an
ordering  that is very close to the best one
known~\cite{kuipers2013improving,vermaseren2014investigations}.

\subsection{Selection: RAVE}\index{RAVE}\label{sec:rave}
We will now look at another enhancement. 
One of the best known enhancements to MCTS is RAVE, or rapid action
value estimation. RAVE addresses the problem that initially, at the
start of the search, all visit and win rate counts of 
the nodes are zero and MCTS is expanding with little guidance of where
to go. The early part of MCTS thus unfolds in the dark.  The sooner search
information is available throughout the tree, the better the children that will be selected and the
more efficient the search will be. 
Two closely related techniques have
been developed: all moves
as first (AMAF) and rapid action value estimation (RAVE). They allow for the
node statistics (win count and visit count) to be updated more quickly.

Already in~\cite{brugmann1993monte} Br\"ugmann gives the first
description of AMAF in his Monte Carlo Go. Gelly and
Silver~\cite{gelly2007combining} report on the use of AMAF in MCTS, combining it
with tree search and UCT, in a Go
program. Helmbold and Parker-Wood~\cite{helmbold2009all} also find AMAF to
be effective in Go. An AMAF update of
the win and visit counts updates not only the counts of nodes on the
playout path to the root of the MCTS tree, but also  siblings of those nodes that
occurred at {\em any\/} position in the playout. Statistics are
updated for all actions of a state that occurred in the playout.  Since playouts tend to touch
about half the board for a side, this means that many extra nodes in
the MCTS tree are updated by
AMAF. AMAF
works in the backup operation, in order to influence node
selection in the UCT formula. Through AMAF many more nodes receive the
statistics, allowing UCT to make a more informed
exploration/exploitation decision. AMAF is learning off-policy.

Being off-policy, the cost of AMAF is a loss of precision. Branching out the win and visit statistics
to nodes that are not exactly on the path, but are loosely related,
reduces the precision of the algorithm. A better situation would be to
only use AMAF at the start of the search, as warm up, and then to not
use AMAF anymore when MCTS has sufficient statistics to work
well. This is the approach that  RAVE follows. At the start it use AMAF-like
spreading of updates, whose use is governed by a parameter that
decreases as nodes are expanded. See \cite{helmbold2009all,chaslot2010monte,browne2012survey} for a more in-depth discussion of AMAF and its variants,
and \cite{gelly2007combining,gelly2011monte} for their variant RAVE. Gelly and Silver
further extend the UCT formula to take advantage of prior knowledge,
making use of pattern databases that are trained from grandmaster
games.

In RAVE the nodes whose statistics are updated are not just the nodes
that were selected by the policy. RAVE causes limited off-policy learning to
occur in MCTS.

\subsection{Playout: Pattern Database}
Another enhancement concerns the playout phase.
Sinnce the playout phase chooses moves randomly, playouts are quick
 and no heuristic has to be devised. However, should such
quick heuristics exist, they will almost surely achieve better play than
random moves.

Most Go programs replace  random playouts by
heuristics or databases of small (often $3\times 3$) patterns. The
patterns are  
learned off-line by supervised machine learning from grandmaster
games.
These patterns check each position in the $3\times 3$ pattern around
where the last move was played. 
 
Gelly et
al.~\cite{gelly2006modification,gelly2008achieving} describe how
pattern databases are used for better playouts in MoGo, achieving
grandmaster performance on the small $9\times 9$ Go board. 

\subsection{Adaptive Exploitation/Exploration}\index{$C_p$}
The $C_p$ parameter of MCTS governs the exploration/exploitation
trade-off. Ruijl et al.~\cite{ruijl2013combining} describe a scheme in
which 
$C_p$ decreases towards zero over the course of the search, starting
out favoring exploration, and becoming more exploitative towards the
end of the search. This focuses search effort towards building deeper
trees towards the end. This is beneficial
in an application where  large formulas are optimized with
MCTS~\cite{ruijl2013combining,ruijl2014hepgame,Ruijl_2018}.

Such adaptive  parameters are also used in simulated annealing,
$\epsilon$-greedy exploration, and other reinforcement learning methods.

\subsection{Ensembles and Parallelism}\label{sec:par}
One way to improve performance without actually having to change the
algorithm is to parallelize it. A parallel algorithm can run faster
when parallel hardware is available, such as a multi-core or  cluster
computer.

MCTS consists of many search iterations starting at the root. These
searches are semi-independent. Through the win and visit counts the
iterations 
communicate search results. The searches can in principle be run
independently, altough the individual iterations miss out on some of the
accumulated win and visit statistics in the other
iterations. Nevertheless, MCTS appears
quite well suited for speedup through parallelization. Indeed, most large
Go programs run   the iterations in parallel.

Traditionally, the literature identifies three
approaches to parallelization: root parallelism, tree
parallelism, and leaf parallelism~\cite{chaslot2008parallel,mirsoleimani2015scaling}. In root
parallelism, processors perform  MCTS searches independently,
not sharing a tree (thus reducing synchronization and communication,
at the cost of less efficient searches because of less win/visit statistics).

In tree parallelism the tree is shared, at the cost
of more synchronization and communication, but also leading to a more efficient
search~\cite{segal2010scalability}.

Leaf parallelism again shares no information, and performs
independent playouts in parallel~\cite{rocki2011large}.

Mirsoleimani et
al.~\cite{mirsoleimani2017structured,mirsoleimani2018pipeline} provide a unified view on tree
and root parallelism, solving some of its problems, and allowing
hybrid versions   of root and tree parallelism to be created with
pipeline parallel MCTS.

Although root parallelism may seem inefficient due to the lack of
sharing of information, there is an interesting relation with ensemble
search. Fern
and Lewis~\cite{fern2011ensemble} and Mirsoleimani et
al.~\cite{mirsoleimani2015ensemble}
study ensembles of independent
MCTS searches, in which 
the tree is searched independently,  in relation to
exploration. Ensembles of MCTS searches explore
more of the tree than a single MCTS search does, because the searches
are independent. Root parallel 
searches are also independent; they share no information, just
as ensemble search. Thus, if one
would like to achieve a certain exploration/exploitation balance in a
parallel setting, all one has to do is to use root parallelism with a
reduced $C_p$ parameter. The reduced $C_p$ dials back the exploration of each
individual search, and one gets
more exploration inherent in the independent (ensemble or parallel) searches.

\subsection{Rollout Algorithms}\index{rollout}\index{alpha-beta}\index{SSS*}
We are at the end of this section on MCTS enhancements. There is a
final observation to be made, which is not an enhancement, but
more  a philosophical curiosity.

Alpha-beta and MCTS are usually regarded as algorithms from different
classes, of an almost incomparable nature. Alpha-beta is a recursive depth-first 
tree traversal algorithm. MCTS is a path-traversal algorithm,
iteratively invoked until the search budget is exhausted, starting
from the root and 
returning to it after each pass. Alpha-beta is rigid left-to-right; MCTS
is inherently adaptive and selective, using UCT.

MCTS is part of a family of {\em rollout\/} 
algorithms~\cite{bertsekas2013rollout}. In rollout algorithms the
starting point is a  base policy, whose performance is
evaluated in some way, possibly by simulation. Based on that
evaluation, an improved policy is obtained by one-step
look-ahead.

In 1995, Rivest~\cite{rivest1987game} published an iterative approach
for approximation of the minimax value, with good results.
Twenty years later, inspired by MCTS,
Huang~\cite{huang2015pruning} presented a rollout formulation
(path-based) of alpha-beta. In this formulation the algorithmic structure 
of the rollout version of alpha-beta and MCTS look strikingly
similar. The two differences are the selection rule (trial) and
the type of 
values that are backed up (error). Alpha-beta has a left-first selection rule;
MCTS selects the child with the highest UCT value. Alpha-beta backs up
upper bounds and lower bounds; MCTS backs up node counts and
win counts.

Listing~\ref{lst:mctsrollout} shows MCTS in compact pseudo-Python
(after~\cite{huang2015pruning}).
Listing~\ref{lst:abrollout} shows alpha-beta rollout code in
pseudo-Python (after~\cite{huang2015pruning}).\footnote{Note that the code is pseudo-Python. It looks like Python, but, for
the sake of compactness, some
essential functions are missing, such as initializations of counters
and bounds, tree manipulation code, and max and min functions for all
children. The listings contain some code duplication to highlight
similarity of algorithmic structure.}

\lstset{label={lst:mctsrollout}}
\lstset{caption={Rollout MCTS in pseudo-Python~\cite{huang2015pruning}}}
\begin{lstlisting}[language=Python,float]
# python-like pseudo-code of MCTS. After [Huang 2015]
def mcts(root, budget):
  while root.n < budget: 
    g = rollout(root)
  return g

def rollout(s):
  if s.isleaf():
    g = playout(s)
  elif s.ismax():
    c_select = s.argmax_c(c.avg + cp * sqrt(log(s.n)/(2*c.n)))
    g = rollout(c_select)
    s.avg = s.n/(s.n+1)*s.avg + 1/(s.n+1)*g
    s.n   = s.n + 1
  elif s.ismin():
    c_select = s.argmin_c(c.avg - cp * sqrt(log(s.n)/(2*c.n)))
    g = rollout(c_select)
    s.avg = s.n/(s.n+1)*s.avg + 1/(s.n+1)*g
    s.n   = s.n + 1
  return g
\end{lstlisting}

\lstset{label={lst:abrollout}}
\lstset{caption={Rollout alpha-beta in pseudo-Python~\cite{huang2015pruning}}}
\begin{lstlisting}[language=Python,float]
# python-like pseudo-code of alpha-beta. After [Huang 2015]
def alphabeta(root):
  # all bounds should be initialized to (-inf, +inf) (not shown)
  while root.lb < root.ub: 
    rollout(root, root.lb, root.ub)
  return root.lower

def rollout(s, alpha, beta):
  if s.isleaf():
    s.lb = s.playout()
    s.ub = s.playout()
  elif s.ismax():
    c_select = first([c for c in s.child() if max(alpha, c.lb) < min(beta, c.ub)])
    rollout(c_select, max(alpha, c_select.lb), min(beta, c_select.ub))
    s.lb = max([c.lb for c in s.child()])
    s.ub = max([c.ub for c in s.child()])
  elif s.ismin():
    c_select = first([c for c in s.child() if max(alpha, c.lb) < min(beta, c.ub)])
    rollout(c_select, max(alpha, c_select.lb), min(beta, c_select.ub))
    s.lb = min([c.lb for c in s.child()])
    s.ub = min([c.ub for c in s.child()])
  return 
\end{lstlisting}

Huang's formulation is important because it elegantly captures the
difference between these two important algorithms in a single
algorithmic framework. His formulation also highlights that the two
algorithms have more in common 
than previously assumed, even though they come from different paradigms.







\section{Practice}
Below are some questions to check your understanding of this
chapter. Each question is a closed question where a simple, one
sentence answer is possible based on the text in this chapter.
\subsubsection*{Questions}
\begin{enumerate}
  \item How are Go players ranked?
  \item What is the difference between tactics and strategy in games?
  \item What is the difference between exponential and polynomial time
    complexity? Can you name an algorithm for each?
    \item What are the four steps of MCTS?
    \item Describe two advantages of MCTS over rigid heuristic
      planning.
      \item How is MCTS recursive?
      \item Describe the function of each of the four operations of MCTS.
        \item Give the UCT formula.
  \item How does UCT achieve trading off exploration and exploitation?
    Which inputs does it use?
    \item When $C_p$ is small, does MCTS explore more or exploit more?
  \item For small numbers of node expansions, would you prefer more
    exploration or more exploitation?
  \item What does RAVE do?
    \item What is the role of pattern databases in the playout phase?
  \item Give three ways in which the performance of MCTS can be
    enhanced.
    \item What is a rollout algorithm?

\end{enumerate}
\subsubsection*{Exercises}

For the exercises in this chapter we will use two code bases:
Browne et al.~\cite{browne2012survey} and Michi. The MCTS implementation of Browne is simple, clear, and
well suited to learn MCTS with. The example code implements Othello and Nim, but not
Go. The Michi implementation does implement Go. It is a minimalist
implementation of an MCTS Go program.

\begin{enumerate}
\item Implement MCTS, for example from
  \href{https://int8.io/monte-carlo-tree-search-beginners-guide/}{here}.\footnote{\url{https://int8.io/monte-carlo-tree-search-beginners-guide/}}
 Familiarize
  yourself with the code, and implement a game of Othello. 
\item Study the impact of search effort. Make a graph of Elo rating
   on the
  $y$-axis and $N$ the number of node expansions on the $x$-axis (play
  different $N$ numbers of node expansions against each other).
\item Study the impact of exploration and exploitation. For a
  fixed search budget, plot Elo rating for four different $C_p$
  values: $\{0.1, 0.7, 1.0, 3.0\}$.
\item Implement a Hex game player with MCTS.  See the
  \href{https://www.maths.ed.ac.uk/~csangwin/hex/index.html}{Hex page}\footnote{\url{https://www.maths.ed.ac.uk/~csangwin/hex/index.html}}
  for information. See also~\cite{browne2000hex,hayward2019hex,huang2013mohex}.
\item Go to \href{https://github.com/pasky/michi}{Michi}\footnote{\url{https://github.com/pasky/michi}} and
  download the code.  Play a game against the computer on $9\times 9$
  and on $19\times 19$. Who won? Is your computer fast enough for the
  big board?
\item Do the preceding Othello exercises (1--3) also for $9\times 9$ Go.
  
\end{enumerate}


\section*{Summary}


In this chapter the dual challenges of (1) rigid search and (2) lack of an efficient
heuristic evaluation function were addressed. We introduced a new
paradigm, the adaptive sampling paradigm, 
with the MCTS 
algorithm. Where minimax needed enhancements and tweaks to focus the
search effort on the important part of the state space, MCTS is
inherently selective. The most popular selection rule is UCT, which
provides a good exploration/exploitation trade-off. UCT is based on
 a large body of bandit theory.

The development of MCTS was driven by Go, which emerged as the new
\emph{Drosophila} of AI, after 
Deep Blue beat Kasparov in Chess. 

The fact that MCTS does not need a heuristic evaluation function
allows it to work for Go, and for other domains for which no efficient
heuristic 
function was found. MCTS caused a breakthrough in Go performance,
achieving master level play on the small $9\times 9$ board.

Traditionally, MCTS has four operations: select, expand, playout, and
backup (although we also showed a more compact rollout formulation,
providing a link to alpha-beta \cite{huang2015pruning}).

As before, enhancements are important to achieve the highest possible performance. Two well-known
enhancements are RAVE, to speed up the propagation of node statistics
in the first part of the search, and databases of small patterns in
the playout operation.

The previous chapters focused on model-based reinforcement
learning. Heuristic planning and adaptive sampling need a simulator
with the rules of the game as the transition function.

\section*{Historical and Bibliographical Notes}

MCTS was first developed for Go. Soon other games and optimization applications 
were also found to benefit  from the flexible
adaptivity of MCTS and especially from the fact that no domain-specific
heuristic evaluation function was needed. See for some
examples~\cite{browne2012survey,arneson2010monte,chaslot2008monte,silver2010monte,kuipers2013improving,chaslot2010monte,moerland2018a0c,moerland2018monte}.

A highly cited
survey on MCTS is~Browne et al.~\cite{browne2012survey}. It has  more
than 200 references to papers on MCTS and provided a wealth of
knowledge on MCTS, highly recommended.
Coulom introduced the  MCTS algorithm~\cite{coulom2006efficient}, and
he implemented it in his program Crazy Stone.
Gelly et al.\ wrote another  Monte Carlo-based program,
MoGo~\cite{gelly2006modification,gelly2012grand}, which was successful
in tournaments against other programs, and against human Go players.
Gelly and Silver published an influential paper on pattern
databases and AMAF/RAVE~\cite{gelly2007combining,gelly2011monte}.

Of prime importance in MCTS is the selection rule, which gives it
adaptive selectivity. 
UCB1 and UCT are selection rules based on bandit theory, derived
in~\cite{kocsis2006bandit,auer2002using}. 
In their seminal 2006 paper, Kocsis and Szepesv\'ari introduce
UCT~\cite{kocsis2006bandit}.
Most Go programs use some form of parallel search. There has been quite some research into parallelization of MCTS; see,
e.g., Chaslot et al.\ and Mirsoleimani et
al.~\cite{chaslot2008parallel,mirsoleimani2018pipeline}. Ensemble MCTS
is related to parallel MCTS and exploration; see~\cite{fern2011ensemble}.

On-line there are many resources to be found for programming a
computer Go program. 
At \href{https://senseis.xmp.net/?ComputerGoProgramming}{Senseis} there is much information on how to write your own
Go program.\footnote{\url{https://senseis.xmp.net/?ComputerGoProgramming}}

\chapter{Function Approximation}\label{chap:super}\label{chap:dqn}\label{chap:conv}


The previous chapter featured evaluation by random playouts. In this
chapter we try to improve upon random playouts, using function
approximation methods. The aim of this chapter is  ambitious:  we focus on large, high-dimensional,
\emph{end-to-end}, problems; actions 
must be found for states without  hand-crafted intermediate heuristic features.

Deep learning is a method to approximate an objective function by
automated discovery of  features.  
This objective function can be a regression function, a classifier, a
value function, or a 
policy function. Classification functions are used in supervised image
recognition tasks, and value and policy functions are used in
reinforcement learning in games.

Deep learning has been quite succesful in image recognition, yielding
methods  that classify
images better than humans. In games, deep learning has found policy
functions that play vintage Atari arcade games as good or better than humans. Deep learning  is a
driving force of the  AI revolution.

End-to-end learning means that we learn output labels and
actions  directly from raw pixel inputs, un-preprocessed, without intermediate
hand-crafted heuristics. End-to-end learning is
computationally intensive.
Progress  has been made possible by advances in compute
power in CPUs and especially in GPUs  around the turn
of the millennium. Furthermore, around that time
large datasets of hand-labeled  data were created, providing the
necessary training data. Finally,
algorithmic advances solved key problems in deep learning. Together
with the increase
in compute power and big data these developments enabled the deep learning revolution.

To quote one of the central works of this area, Mnih et al.~\cite{mnih2013playing}:
``Recent breakthroughs in computer vision and speech recognition have
relied on efficiently training deep neural networks on very large
training sets. The most successful approaches are trained directly
from the raw inputs, using lightweight updates based on stochastic
gradient descent. By feeding sufficient data into deep neural
networks, it is often possible to learn better representations than
hand crafted features.''\index{Volodymyr Mnih}






\section*{Core Problem}
\begin{itemize}
\item Can we find end-to-end value/policy functions for
  high-dimensional problems?
\end{itemize}

\section*{Core Concepts}
\begin{itemize}
\item Automated feature discovery through deep learning
\item Convolutional neural network (CNN) and deep Q-network (DQN)
\item Overfitting  and stable learning
\end{itemize}
%
%
%
%
Many of the
advances in deep reinforcement learning started in supervised learning. Therefore the
first part of this chapter is devoted to an introduction to the advances in deep
supervised learning. A good understanding of deep learning is
important in modern reinforcement learning; we will  spend quite
some effort on this topic.

\section{Deep Supervised Learning}\index{neural
  network}\index{function approximation}
Towards the end of Chap.~\ref{chap:rl}, in Sect.~\ref{sec:approx}
we discussed ways to approximate functions when the state space
becomes so large that it is unlikely that states, when tested, will have been seen
before in training. Statistical learning theory provides a framework for machine
learning methods such as supervised learning and reinforcement
learning~\cite{russell2016artificial,friedman2001elements}.

Supervised
learning is the task of learning a regression or classification function that maps an input to an
output based on example input-output
pairs~\cite{russell2016artificial,goodfellow2016deep,bishop2006pattern}. Our
goal is to solve the inverse function problem: find a function to fit
its  outputs.\index{inverse function problem} For example, the training set may be a set of pictures, labeled
with a description, such as {\em CAT\/} or {\em DOG}. Because  the learning process learns a function from
an explicit set of labeled training
data it is called {\em supervised}, since the labels supervise the
learning process.

\subsection{Generalization and Features}\index{supervised learning}
Supervised learning is a basic and widely studied problem in
artificial intelligence. A wide range of algorithms exist for
supervised learning: support vector machines, linear regression,
Bayesian classification, decision trees, random forests, nearest
neighbor, and artificial neural networks, to name just a
few~\cite{russell2016artificial,bishop2006pattern,hall2009weka}.

Since around 2010 artificial neural networks have achieved
impressive results in image recognition, speech recognition, and game
playing. We will focus on deep neural networks in this chapter.

We will first discuss learning as a generalization task, and
then  how generalization can be viewed as a feature discovery task.

\begin{table}[t]
  \begin{center}

\begin{tabular}{lcccc}
  {\bf Chapter} & {\bf Algorithm} & {\bf Select} & {\bf Backup} & {\bf
                                                                  Eval}\\ \hline\hline
   Chap.~\ref{chap:search} & alpha-beta & left to right & minimax
                                   & heuristic \\ 
  Chap.~\ref{chap:trial} & MCTS & adaptive & average & sample \\ 
   Chap.~\ref{chap:dqn} & DQN & - & - & generalize \\ 
   Chap.~\ref{chap:plan} & self-play & adaptive & average &
                                                              generalize
  \\ \hline
\end{tabular}

  \caption{Heuristic-Sample-Generalize}\label{tab:heursampgen3}
  \end{center}
\end{table}

\subsubsection*{Learning as Generalization from Training Set to Test Set}\index{overfitting}\index{underfitting}\index{training
  set}\index{test set}\label{sec:trainset}\label{sec:generalization}
Learning is the adaptation of behavior from experience. 
Supervised
learning is closely related to generalization, which is the
formulation of common properties in instances of
data. Table~\ref{tab:heursampgen3} shows (again) the main paradigms of
this book, to show how generalization is contrasted
to the other paradigms. 

Supervised learning learns a continuous regression or a discrete classification
function. Regression  finds the relation between examples and a
continuous variable; classification  finds the relation
between examples and a set of discrete classes.\footnote{Also in continuous regression
problems the test examples will not be seen at training time, and
approximation is necessary.}  

A standard method in machine learning to assess the quality of a categorization 
algorithm is to divide the dataset into a
training part and a test part ($k$-fold cross validation)~\cite{bishop2006pattern}. A typical
split is 90\% of the data for the training set, and the remaining 10\%
for the test set. The algorithm
is trained on examples from  the training part. The
quality of the resulting approximation function is then tested on the
remaining (unseen) 10\%, 
to see how well it generalizes. If the percentage of correct
predictions on the test set is about the same as on the training set,
then the approximation  generalizes well to the test set. If not, the
training process
may have underfitted or overfitted (captured too little of the signal
or too much of the noise, see Sect.~\ref{sec:overfitting}).

Deep learning
provides methods to generalize over data to approximate regression and
classification functions. It does so by discovering common
features. Let us see  how this is achieved. 

\subsubsection*{Generalization as Feature Discovery}\index{feature
  discovery}\index{end-to-end learning}
Generalization can be achieved through the discovery of common
features.
Traditionally, feature discovery is a labor-intensive task. 
Common features in a dataset are painstakingly identified by hand, by
one by one
looking at the examples in the dataset, and by writing small algorithms
that recognize the features. For image data,
such hand-identified features are typically
simple features such as lines, squares, circles, and
angles. These features can then be used in decision trees to construct
recognizers to classify an image.

As we shall see in the next subsections, deep learning has allowed
this manual process to be automated.  We have just
described machine learning as a generalization task.
By finding  commonalities we
can group  the instances into a smaller number of  higher-level concepts. These
concepts are more abstract, more general, than the many lower-level
instances.\footnote{Feature discovery is  related to dimensionality
reduction, a technique that can be used for visualization of complex
data~\cite{roweis2000nonlinear,tenenbaum2000global,maaten2008visualizing}.} 

In the context of image recognition the commonalities are 
called features, and the process of learning  higher-level abstractions is  called 
representation learning~\cite{lecun2015deep}.
Deep nets can perform feature recognition through a
layered hierarchy of increasingly abstract 
representations~\cite{hinton2006fast,lecun2015deep}.
Today deep learning is one of the 
core technologies that is used in automated function approximation.

From a
scientific point of view, the hand
crafting and hand tuning is undesirable as it is hard to 
reproduce. Hand crafting depends on expert knowledge, in contrast to  the training
of approximators by an optimization algorithm such as stochastic
gradient descent, which can be reproduced easily by running the
algorithm again.

Furthermore, and  
most importantly, the hand-crafted heuristics do not generalize well
to other domains, whereas deep  feature recognizers have been
shown to work well in many domains (although they typically have to be
retrained for each domain).

Finally, there may be problem domains  where our domain intuition fails to 
come up with efficient features. We may not understand our own human
interpretations well enough to come up with efficient algorithms. This
was the case in Go, as we saw in the  previous chapter.

\subsubsection*{End-to-End Learning}
End-to-end learning, or  classification directly from  high-dimensional raw
pixel data, is computationally quite demanding, as opposed  to learning
based on intermediate (lower dimensional) hand-crafted
features. You may recall that in Chap.~\ref{chap:search} heuristic
value functions were used to 
search the large state spaces of Chess and
Checkers (Sect.~\ref{sec:eval}). Automated tuning of features was
discussed as a hybrid method between hand-crafted
heuristic functions and end-to-end feature learning. 
The features themselves are still hand crafted, but the coefficients
(weights) $c_i$ of the features
in the polynomial evaluation function $h(s) = c_1\times f_1(s) + c_2\times f_2(s) + c_3\times f_3(s) +
\ldots + c_n\times f_n(s)$ are learned by supervised learning.
The difference is that the
features in the Chess evaluation function are manually designed
heuristics, combined in a linear function whose coefficients are
manually tuned, whereas the features in a deep net are trained
(generalized) on a large set of examples in an automatic training process.

We will now see in more detail how neural networks can perform automated feature
discovery. First we will describe the general architecture for shallow
artificial neural networks.

\begin{figure}[t]
\begin{center}
\includegraphics[width=7cm]{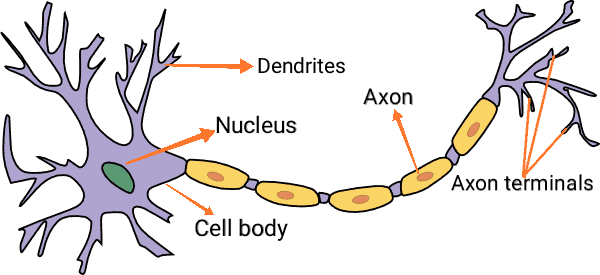}
\caption{Biological neuron \cite{quoc2018}}\label{fig:bnn}
\end{center}
\end{figure}

\subsection{Shallow Neural  Network}
The architecture of artificial neural networks is inspired by the
architecture of biological neural networks, such as the human
brain. Neural networks consist of neural core cells that are connected by
 nerve cells (see, for example, \cite{bear2007neuroscience}). Figure~\ref{fig:bnn} shows a
drawing of a biological neuron, with two nuclei, an axon, and dendrites~\cite{quoc2018}.

\begin{figure}[t]
\begin{center}
\includegraphics[width=5cm]{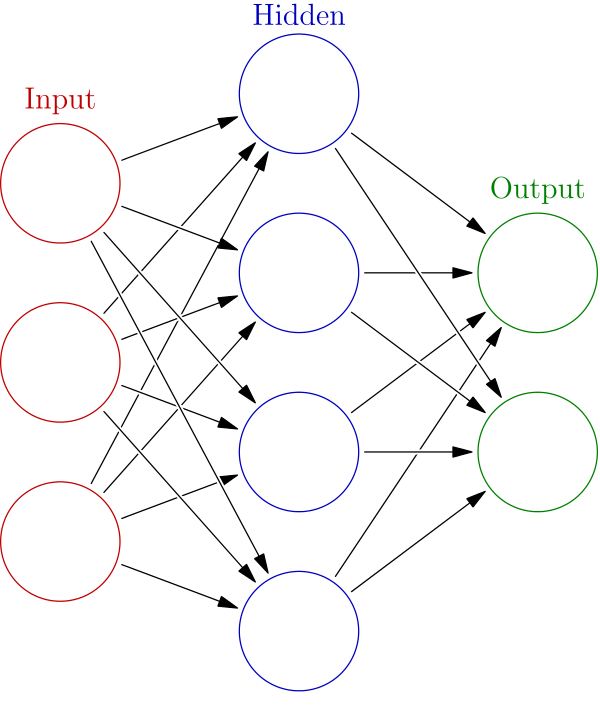}
\caption{A shallow neural network}\label{fig:nn}
\end{center}
\end{figure}

Figure~\ref{fig:nn} shows a simple fully connnected artificial neural
network, with an input layer, 
an output layer, and a single hidden layer. A neural network with a single hidden layer
(or a few, 2--3)
is called shallow. When the network has more  hidden layers,
it is called deep.


Just as computer Chess, neural networks research has a long history,
and much work was done before they were successful. Already in 1943 McCulloch and
Pitts~\cite{mcculloch1943logical} provided a computational model for
neural networks.
The history of the development of neural networks is quite interesting and
rich, with high ups and deep downs. Training neural networks is
a problem that has been studied by many researchers. The 2018 Turing award was awarded to
Bengio, Hinton, and LeCun for their work in this area (Fig.~\ref{fig:bhl}). Section~\ref{sec:bib:neural} provides
 pointers to some of the literature.

The most popular training method is based on gradient
descent, of which the stochastic version performs very well in deep
neural networks.   Much
research has been done on 
algorithms for efficient  gradient descent~\cite{werbos1974beyond}. Stochastic gradient descent (SGD)
and backpropagation have allowed the training of
multi-layer networks, paving the way for deep learning's success.


\subsubsection*{Stochastic Gradient Descent and Backpropagation}\label{sec:algo}\index{backpropagation}\index{SGD}\index{epoch}
We will now discuss the networks and their training algorithms in more detail.

Neural networks consist of neurons and connections, typically
organized in layers (as in Fig.~\ref{fig:nn}). The neurons
act as functions that process their input signal as a weighted
combination of inputs, producing an output
signal.  This function is called  an activation function. Popular
activation functions are the rectified linear unit (ReLU: partly
linear, partly zero), the hyperbolic tangent, and the
sigmoid or logistic function $\frac{1}{1+e^{-x}}$ (these will be
discussed in more detail shortly). The neurons are connected
by  weights. At each neuron $j$ the incoming weights $ij$ are summed $\sum$
and then processed  by
the activation function $\sigma$. The output $o$ of neuron $j$ is therefore:  $o_j = \sigma(\sum_i x_iw_{ij})$  for weight $ij$ of
predecessor neuron $x_i$. The outputs of this layer of neurons are
fed to the
inputs for the weights of the next layer.

\begin{figure}[t]
  \centering{\includegraphics[width=8cm]{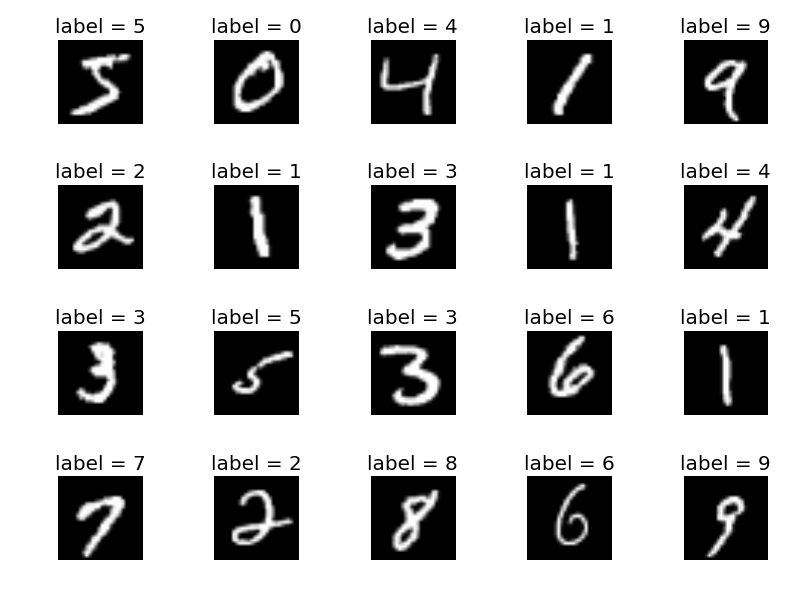}\caption{Some MNIST images}\label{fig:mnist}}
\end{figure}

In supervised learning, a network is trained on a set of example pairs $(x,z)$: inputs $x$
and labels, or targets, $z$. Typical examples are images of pictures or digits,
in which case the labels are the correct digit (Fig.~\ref{fig:mnist}).

The neural network is a parameterized  function $f_\theta(x)\rightarrow y$ that
converts input to output. The behavior depends on the parameters
$\theta$. Training the network is adjusting the weights such that the
required input-output relation is achieved. This is done by
minimizing the error (or loss) function that calculates the
difference between the output $y$ and the network target $z$.

The training process consists of training {\em epochs,} individual passes in
which the network weights are optimized towards the goal. When training starts, the weights of the
network are initialized to small random numbers. 
Each epoch consists of a forward
(recognition, usage) pass and a backward (training, adjustment) pass. The forward pass
is just the regular recognition operation for which the network is designed.
The input layer is exposed to the input
(e.g., the image), which is then propagated through the network to the output
layers, using the weights and activation functions. The output layer provides the answer, by
having a high value at the neuron corresponding to the right label
(such as CAT or DOG, or the correct number).

The backward pass uses the difference between the
forward recognition outcome and the true label.
At the output
layer the propagated value $y$ is compared with the other part of the 
example pair, the label $z$. The difference with the label is calculated, yielding
the {\em error}.
Two common error functions are the mean squared error $\frac{1}{n}\sum_{i}^n(z_i-y_i)^2$ (for
regression) and the cross-entropy error $-\sum_i^M z_i\log y_i$ (for classification
of $M$ classes). See, for example, Goodfellow et
al.~\cite{goodfellow2016deep} for much more on error functions.  

Training stops when the error function has been reduced below a
certain threshold for a single example, or when the loss on an entire validation set has dropped
sufficiently. More elaborate stopping criteria will be discussed later
in relation to overfitting.

Then, in the backward phase, this error is propagated back
to the input layer, adjusting the weights in the direction so that the
error becomes smaller. This method uses the gradient over the weights,
and is called gradient descent.

Most neural nets are trained using a stochastic
version of   gradient descent, or
SGD~\cite{song2018mean}, that samples  the gradients. SGD speeds up the process and introduces some noise,
reducing overfitting.  Goodfellow et
al.~\cite{goodfellow2016deep} provide more details.

Especially in image and speech recognition impressive results have
been reported by deep
networks~\cite{bengio2009learning}. Popular
variants of SGD are AdaGrad~\cite{duchi2011adaptive} and
Adam~\cite{kinga2015method}, methods that adaptively change the
learning rate and momentum of the backpropagation.
%
%
The theory and practice of gradient descent  algorithms is quite
elaborate and subtle.
Many articles and books have been written on
the subject of gradient descent; see, for example
~\cite{goodfellow2016deep,haykin2009neural,hinton2006reducing,bishop1995neural,lecun2015deep}.

Now that we have discussed essential concepts of shallow neural
networks, we will turn to deep learning.

\begin{figure}[t]
    \centering{\includegraphics[width=6cm]{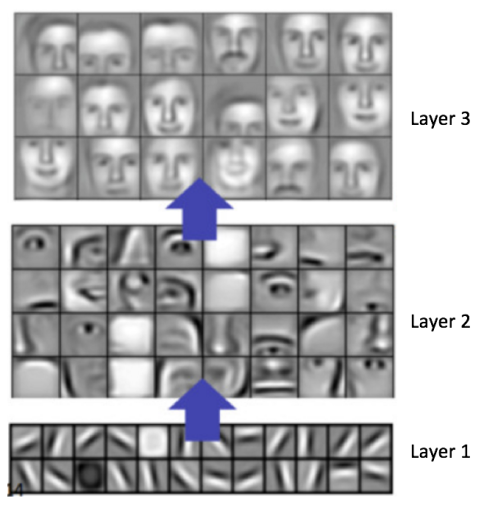}
     \caption{Layers of features of increasing complexity~\cite{lee2009convolutional}}\label{fig:leelayers}}
\end{figure}

\subsection{Deep Learning}\index{deep learning}\label{sec:deepnetwork}
There has been quite some theoretical work on the question of which
functions can be represented with neural networks.\footnote{The
universal approximation theorem states that a  feed-forward network
with a single hidden layer containing a finite number of neurons can
approximate a wide variety of continuous
functions~\cite{cybenko1989approximations}. Although it states that 
simple neural networks can represent a  variety of interesting
functions when given appropriate parameters, the theorem does not
say how those parameters can be learned. Learning may take very long,
and may take an impractical number of neurons in the hidden layer.}
Some functions can
only be represented  with an architecture of at least a
certain
depth~\cite{bengio2009curriculum,bengio2007greedy,haastad1991power}.
For complex tasks such as image recognition  single-layer
networks do not perform well, or need preprocessing with hand-coded
feature recognizers.  A breakthrough in end-to-end (direct image) recognition was
achieved only when multi-layer  networks  were introduced and when ways 
were found to train them efficiently~\cite{schmidhuber2015deep,lecun2015deep}.



\index{Yann LeCun}\index{Yoshua Bengio}\index{Geoffrey Hinton}
LeCun, Bengio, and Hinton~\cite{lecun2015deep},
explain  how multiple hidden layers in networks learn increasingly
abstract representations. In deep learning, each layer  transforms its input data into a
 more abstract  representation.
The  hierarchy of network layers together can 
recognize a hierarchy of low-to-high level
concepts~\cite{lecun2015deep,lee2009convolutional}. For example, in
face recognition (Fig.~\ref{fig:leelayers}) the first hidden layer may 
encode edges; the second layer then composes and 
encodes simple structures of edges; the third layer may encode higher-level concepts such as noses or
eyes; and the fourth layer may work at the abstraction level of a
face. Deep feature learning finds what to abstract at
which level on its own~\cite{bengio2013representation}, and can come
up with classes of intermediate concepts, that
work, but look   counterintuitive upon 
inspection by humans.


For deep learning architectures to perform recognition tasks at a
level comparable to human performance, a
number of problems had to be solved, that we will now discuss.

\subsection{Scaling Up}

The development of deep learning benefitted greatly  from efforts in
handwriting recognition. The standard test set for handwriting
recognition was MNIST (for Modified
National Institute of Standards and Technology).
Standard  MNIST images are low-resolution $x\times y=32\times 32$ pixel images
of single handwritten digits (Fig.~\ref{fig:mnist}). Of course,
researchers wanted to process more complex scenes than single digits,
and higher-resolution images.
To achieve higher accuracy, and to process more complex scenes,
networks needed to grow in size and complexity.

There are  three ways in which networks can grow. We assume that the
layers have the same width as the input size. First, the 
width $x\times y=n$ of
the network can be increased, by 
having more neurons in each layer. Second, the depth $d$ of the network
can be increased, by having more layers. And third, the complexity of
the network can be increased, by using a more complex type of
processing at the neurons, and a different interconnection structure.

As you might  expect, all three were necessary,  in unexpected and
nontrivial ways.

\subsubsection*{Increasing Width: Slow Learning and Overfitting}

\begin{figure}[t]
  \centering{\includegraphics[width=10cm]{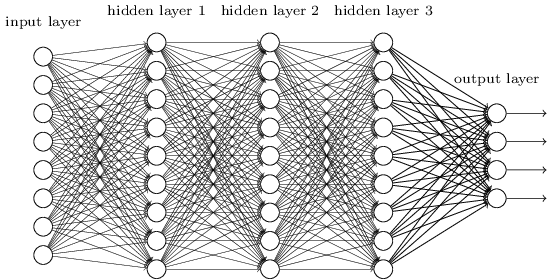}\caption{Three
      fully connected hidden layers~\cite{saha2018}}\label{fig:layers}}
\end{figure}

\begin{figure}[t]
  \centering{\includegraphics[width=8cm]{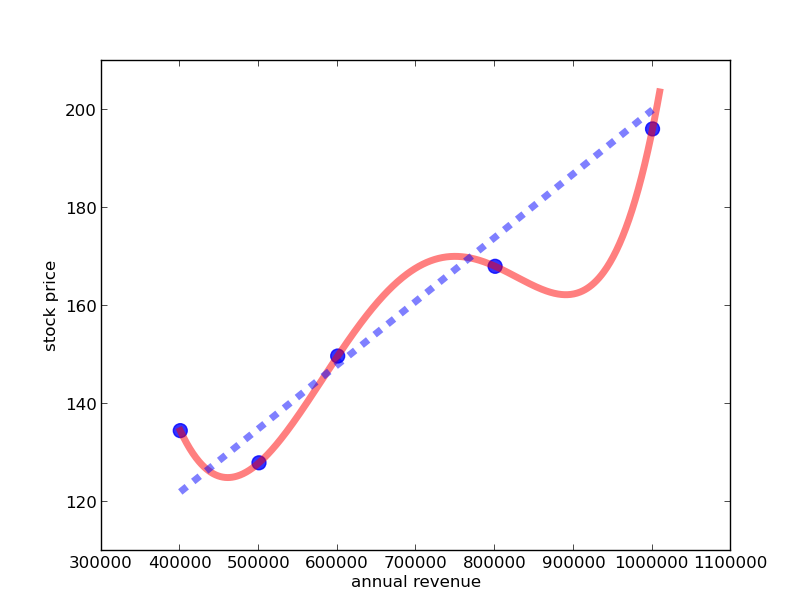}\caption[Overfitting]{Overfitting:
    does the curvy red line or the straight dashed blue line generalize the information in the data points best?}\label{fig:overfitting}}
\end{figure}

Simply increasing the width $n$  of a fully connected network of depth
$d$ increases the
network size $s=n^2\times d$. This is
not a scalable solution (Fig.~\ref{fig:layers}). More width creates
three problems. First, in 
a fully connected network it
causes a quadratic increase in weights and 
the same holds for the training time.  Second,  overfitting becomes
a problem, since after increasing the capacity of the network, the
(quadratically higher)
number of parameters $n^2\times d$ may become 
greater than the number of observations, in which case  the network will
overfit: it will perfectly memorize training examples, and be unable
to correctly generalize to different test examples. Figure~\ref{fig:overfitting} illustrates overfitting, which will
be described more fully in Sect.~\ref{sec:overfitting}.
Third, only increasing the width does not address the problem that some
complex functions can only be approximated with a network of 
a certain depth~\cite{bengio2009curriculum,bengio2007greedy,haastad1991power}.

\subsubsection*{Increasing Depth to Prevent Weights Increase}
Increasing the width of a fully connected network runs into problems
since the number of weights increases quadratically. 
Let us look what happens when we increase the depth $d$ of the network.

As 
the depth increases, the number of weights $s=n^2\times d$ increases
linearly (Fig.~\ref{fig:layers})  and so does the 
training time. The overfitting problem also increases
linearly. And, complicated functions  can be approximated better by
the deeper network.

Therefore,  increasing the  depth of the network is a
way forward. However, it does not solve all our
problems. 
If we want to approximate a function, especially one with a high-dimensional input such as for  higher-definition images, then at least our input layer has
to have a correspondingly high number of inputs. So,  we still have not completely solved the problem of the increasing
number of weights of fully connected networks. The number of weights just keeps
increasing with high-resolution input, causing long training times, overfitting, and
exploding and
vanishing gradients (to be discussed shortly).

For our solution we need the third element: a  different connection
structure. Therefore, instead of fully connected networks, we will now look at
convolutional networks, and related improvements.

\subsection{Convolutional Neural Nets}\index{convolutional network}\index{CNN}
As we wish to process higher-resolution images, increasing $n$ or $d$
increases the number of weights too rapidly, leading to unacceptably long training times and
overfitting when the network is fully connected.

The solution lies in the third element: using  a different (sparse) interconnection structure. 
Convolutional neural nets (CNNs) take their inspiration from biology. 
Early work
on monkey and cat visual cortexes indicated  that neurons respond to
small regions in the visual
field~\cite{hubel1963shape,hubel1968receptive}. The
visual cortex in animals and humans is not fully connected, but
locally connected, in a receptive field.

\begin{figure}[t]
  \centering{\includegraphics[width=8cm]{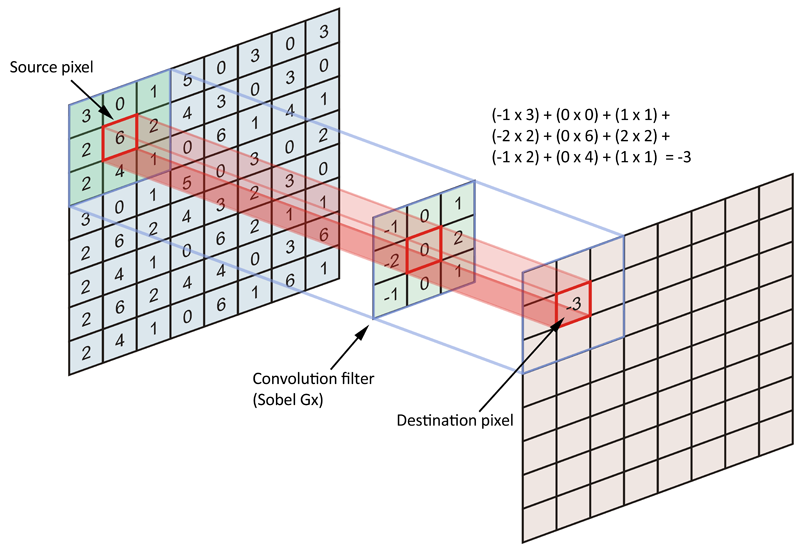}\caption{Convolutional filters~\cite{goodfellow2016deep}}\label{fig:filters}}
\end{figure}

The connectivity pattern of CNNs resembles the animal
visual cortex~\cite{hubel1963shape,matsugu2003subject}. A CNN consists
of convolutional operators or filters. A typical convolution operator
has a small {\em receptive field\/} (it only connects to a limited number of
neurons, say $5\times 5$), whereas a fully connected neuron connects to
all neurons in the layer below. Convolutional filters detect the
presence of local patterns, or features. The next layer therefore acts as a feature
{\em map}.  A CNN layer can be
seen as a set of learnable 
filters (or kernels), invariant for local
transformations~\cite{goodfellow2016deep}.



\subsubsection*{Convolutional Filter}
In Fig.~\ref{fig:filters} we see an example filter.
Filters can be used to identify certain features. Features are basic
elements such as edges,  straight lines, round lines, curves, and 
colors. To work as a curve detector, the  filter should have a pixel
structure with high values  indicating a shape of a curve. By
then multiplying and adding these filter values with the pixel values,
we can detect if the shape is present.  The sum of the 
multiplications in the input
image  will be large if there is a shape that resembles the curve in the filter.

Now this filter can only detect a certain shape of curve. Other
filters can detect other shapes. A filter layer can be visualized in
an activation map, showing where a specific filter (layer)
fires. Larger activation maps can 
recognize more elements in the input image. Adding more filters 
increases the size of the network, which effectively is a large
activation map. 
The filters in the first network layer process (``convolve'') the
input image and fire (have high values) when a specific feature it is
built to detect is in the input image. Training a convolutional net is
training a filter that consists of layers of subfilters.

By going through the convolutional layers of the network, increasingly
complex features can be represented in the activation maps.

Before the CNN starts, the weights or filter values are initialized to 
random values. The filters in the lower layers are ``empty''; they are not trained to look
for edges and curves. The filters in the higher layers are not yet trained
 for eyes and mouths. During training, the filters in these layers
 take on their task as recognizers for respective representations.

 Once they are trained, they can be used for as many recognition tasks
 as needed. A recognition task consists of a single quick forward pass
 through the network.

Let us spend some more time on understanding these filters.

\subsubsection*{Shared Weights}
\index{weight sharing}\index{shared weights}
In CNNs the filter parameters are shared in a layer. Each layer thus
defines a filter operation. A filter is
defined by few parameters but is applied to many pixels of the image; 
each filter is replicated across the entire
visual field. These replicated units share the same parameterization
(weight vector and bias) and form a feature map. This means that all
the neurons in a given convolutional layer respond to the same feature
within their specific response field. Replicating units in this way
allows for features to be detected regardless of their position in the
visual field, thus constituting the property of translation invariance.

This weight sharing is also
important to prevent an increase in the number of weights in deep and
wide nets, and to 
prevent overfitting,  as we shall see later on.

Real-world images  consist of repetitions of many smaller elements. Due to this
so-called {\em translation invariance}, the same patterns reappear
throughout an image. 
CNNs can take advantage of this. The weights of 
the links are 
shared, resulting in a large reduction in the number of weights that
have to be trained. Mathematically CNNs are
constraints on what the weight values can be (some zero, some have to
be equal). This is a significant advantage of CNNs, since the
computational requirements of training the weights of many fully
connected layers would be prohibitive. In addition, statistical
strength is gained, since the effective data per weight increases.

Deep CNNs work well in image recognition tasks, for visual filtering
operations in spatial dependencies, and for feature recognition
(edges, shapes)~\cite{lecun1989backpropagation}.\footnote{Interestingly, this paper
was already published in 1989. The deep learning revolution happened
twenty years later, when publicly available datasets, more efficient
algorithms, and more compute power in the form of GPUs were available.}


\subsubsection*{CNN Architecture}
Convolutions recognize features---the deeper the network, the more
complex the features. A typical
CNN architecture consists of a number of stacked convolutional
layers. In the
final layers, fully connected layers are used to then classify the inputs.

In the convolutional layers, by connecting only locally, the number of
weights is dramatically 
reduced in comparison with a fully connected net. The ability  of a
single  neuron in a CNN to
recognize different features, however, is less than that of a fully connected
neuron. 

By stacking many such locally connected layers on top of each other we
can achieve the desired nonlinear 
filters whose joint effect becomes increasingly global, as more layers are
added.\footnote{Nonlinearity is essential. If all neurons performed
  linearly, then there would be no need for layers.
  Linear  recognition functions cannot discriminate between  cats and dogs.} The neurons become responsive to a larger
region of pixel space, so that the network first creates
representations of small parts of the input, and  from these
representations create
larger areas. By
stacking convolutional layers on top of each other, they can 
recognize and represent increasingly complex concepts without an
explosion of weights.

\begin{figure}[t]
  \centering{\includegraphics[width=6cm]{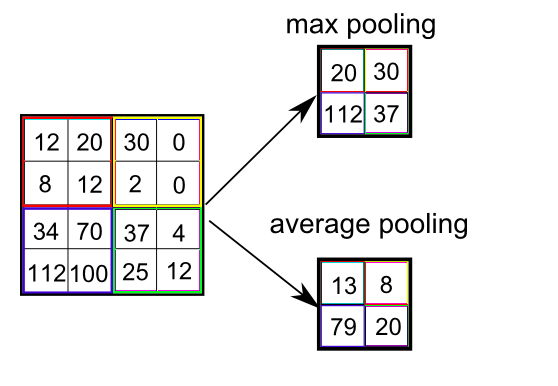}\caption{Max
      and average $2\times 2$ pooling~\cite{goodfellow2016deep}}\label{fig:maxpooling}}
\end{figure}

\subsubsection*{Max Pooling}\index{max pooling}
A further method for reducing the number of weights is pooling.
Pooling is an  operation related to convolving. Pooling is a kind of
nonlinear downsampling (expressing the information in lower
resolution with fewer bits).  Typically, a $2\times 2$ block is down sampled to a scalar
value (Fig.~\ref{fig:maxpooling}). Pooling reduces the dimension of the
network. The most frequently used form is max pooling. It
is an important component for object
detection~\cite{cirecsan2012multi} and is an integral part of most CNN
architectures.

Max pooling also allows small translations, such as shifting the cat
by a few pixels, or scaling, such as putting the cat closer to the camera.

\begin{figure}[t]
  \centering{\includegraphics[width=12cm]{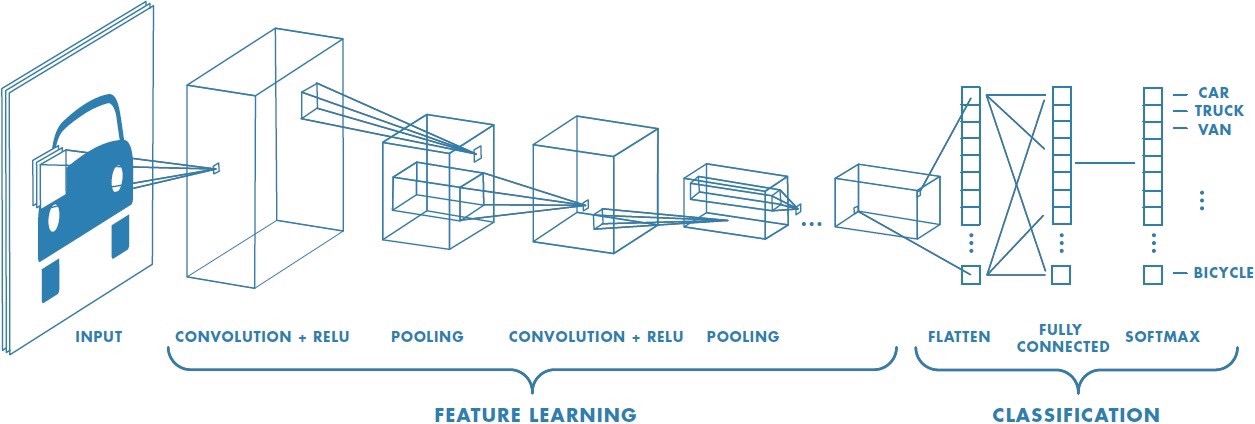}\caption{Convolutional
      network example architecture~\cite{saha2018}}\label{fig:cnn}}
\end{figure}

A typical CNN architecture consists of an architectures of multiple
layers of convolution, max pooling, and ReLU layers, topped off by a
fully connected layer (Fig.~\ref{fig:cnn}).\footnote{Often with a soft-max function. The
  soft-max function normalizes an input vector of real numbers to a
  probability distribution $[0,1]$.}\index{soft-max}
  Section~\ref{sec:conv} will discuss more
concrete examples of well-known CNN architectures.

\subsection{Exploding and Vanishing Gradients}\label{sec:vanishing}
The deep convolutional architectures solved  the problem of
processing higher-resolution and more complex scenes by controlling
the number of weights organized in layers of filters. However, the deep architectures brought a new problem. The popular training algorithm, backpropagation, in
combination with the conventional sigmoid activation function and deep
networks, suffers
from the vanishing gradient problem. 

\subsubsection*{Gradients}\index{vanishing gradients}
The vanishing gradient problem was one of the problems that  held
back progress in multi-level neural 
networks for some time.


The backpropagation algorithm computes the gradient of the error
function  by the
chain rule of composite functions. Originally, most neural networks used the sigmoid function
as their activation function. The
problem is that the derivative of the sigmoid function 
$\frac{1}{1+e^{-x}}$ on the domain $[0,1]$ is a value between $0$
and $0.25$.  Backpropagation multiplies these values with the weights for each
layer, starting from the output layer and working backwards to the input
layer. Each time that a value between $0$ and $0.25$ is multiplied by
the weights,
the derivative gets smaller, approaching zero  for the deepest
layers. Long  products of many layers thus yield  a low gradient value at
the deep layers.
This means that neurons in the earlier layers learn
much more slowly than neurons in later layers~\cite{hochreiter2001gradient,hochreiter1991untersuchungen}.
 
The result is that gradient descent with a traditional activation
function (sigmoid or tanh) cannot learn over many layers. For the deepest layers the gradients are vanishing to zero,
negating the use that an extra layer would give.

The gradient in deep neural
 networks is unstable, tending to either explode or vanish in earlier
 layers. This instability was a fundamental problem for gradient-based
 learning in deep neural networks.
 
Multiple solutions to the vanishing gradient problem have been
found.
One solution  is to model the network as a
multi-level hierarchy of networks pretrained one level at a
time~\cite{schmidhuber1992learning}.   Hinton's work on training deep
networks addresses this problem~\cite{hinton2006fast}.

Let us look at two more solutions.

\begin{figure}[t]
  \centering{\includegraphics[width=8cm]{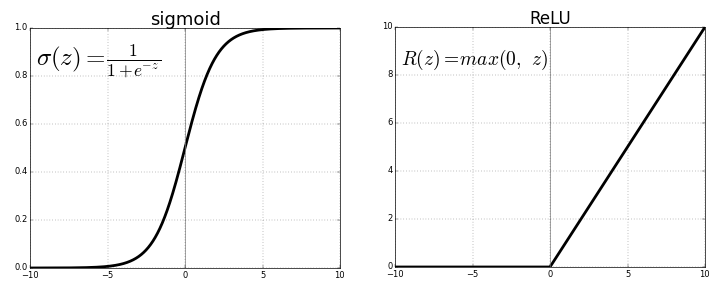}\caption{Sigmoid
      and ReLU}\label{fig:relu}}
\end{figure}

\subsubsection*{ReLU}\index{ReLU}
Since the problem of vanishing gradients is caused by the derivative
of the activation function, alternative functions were proposed.
Traditionally the activation function of neurons is a
well-differentiable function such as $\tanh$ or the sigmoid
function. A 
simpler activation functions is the
ReLU function (short for
rectified linear unit). The definition of ReLU is
$f(x)=\max(0,x)$. Over the positive part of 
its input ReLU is linear, otherwise it is zero. See Fig.~\ref{fig:relu}. ReLU is a simple
function that is quick to calculate, is
nonlinear, and removes negative values from the
activations~\cite{krizhevsky2012imagenet}. The major advantage of ReLU
is that in deep networks it suffers much less from the vanishing
gradient problem. Deep ReLU networks are much easier to train than
sigmoid networks.\footnote{The derivative of the ReLU function has a
  discontinuity. This discontinuity has
  led to the introduction of soft-ReLU, which replaces the sharp
  transition from 0 to linear with a small smooth part.}

\subsubsection*{Batch Normalization}\index{batch normalization}
Another solution to the vanishing gradient problem is batch
normalization. Batch normalization periodically rescales the data. It
performs a batch standardization and  assures that each batch is in a safe
area, taking the gradients away from the zero, the one, or
infinity. This technique was introduced 
in~\cite{ioffe2017batch}. At the cost of some extra computation, the effect of 
periodical normalization  is manyfold. 
Training is sped up, overfitting is reduced, and, finally, the
vanishing gradient problem is reduced substantially, since gradient values
are normalized. 

Batch normalization is at the moment the most powerful method to
prevent overfitting.

\subsection{Overfitting}\index{dropouts}\index{regularization}\index{overfitting}\label{sec:dropouts}\label{sec:overfitting}
Overfitting is a major problem in large neural networks.
The intuition behind  overfitting (and underfitting) is as follows. Let
us assume that the training data represents a noisy
signal, as most data generated by measurements of natural processes does. The learning
process performs well, when the signal from the training data is learned, and the
noise is disregarded. Only then will it generalize well to the test data.
Underfitting occurs when the capacity of the network is too
small to learn, model, or represent the signal, and a crude
approximation results. Overfitting occurs
when the capacity is so large that it learns even the noise in the
training data (Fig.~\ref{fig:overfitting}). Both underfitting
and overfitting result in limited accuracy at test time.

Since the number of weights in
deep networks is often in the millions, it is easily greater than the
number of observations. In such a situation overfitting is a problem
that reduces the generalization performance  of the network.
Overfitting is said to have happened when a trained model corresponds
too closely to a particular 
dataset, and  fails to reliably predict a future
observation. The network has been fit to the particular training
set, the signal and the noise, but not to the underlying
structure of the data.  Overfitting occurs frequently when a model has
more parameters than the training dataset has examples. 


Figure~\ref{fig:overfitting}
illustrates this problem. The red curvy line perfectly fits all data
points, but is unlikely to perform well on a separate test set. The
dashed straight line is more likely to have caught the essence of the
training domain, even though it misses some of the data
points.



Overfitting can be prevented in a number of ways, some of which are
aimed at restoring the balance between the number of network
parameters and the number of training examples. We will discuss data
augmentation and capacity reduction.

Another approach is to look at
the the training process. Examples are regularization, early stopping,
 dropouts, and  batch normalization.

\subsubsection*{Data Augmentation}\index{data augmentation}
Overfitting occurs when there are more parameters in the network than
examples to train on. One possible solution is  to increase the amount
of data. This method is called data augmentation. The training dataset
is increased through manipulations such as rotations, reflections, 
noise, rescaling, etc. A disadvantage of this method is that the
computational cost of training increases. 
 


\subsubsection*{Capacity Reduction}
Another easy solution to overfitting lies in the realization that
overfitting is a result of the network having too large a capacity; the
network has too many parameters. A cheap way of preventing this
situation is to reduce the capacity of the network, by reducing the
width and depth of the network.

\subsubsection*{L1 and L2 Regularization}
A standard  method to try  when we suspect that the network is overfitting, is
regularization. Regularization involves adding an extra term to the loss
function. The regularization term forces the network to not be too
complex. The term penalizes the model for  using too high weight
values.  This limits  flexibility, but also encourages
 building solutions based on multiple features. Two popular versions
 of this method are L1 and L2
 regularization~\cite{ng2004feature,goodfellow2016deep}. For different situations
 different preferred methods exist.

\subsubsection*{Early Stopping}\index{early stopping}\index{holdout validation set}

Perhaps the easiest solution to overfitting is the early stopping solution. Early stopping is
based on the observation that overfitting can be regarded as a
consequence of so-called overtraining (training that progresses beyond
the signal, into the noise). By terminating the training
process earlier, for example 
by using a higher stopping threshold for the error function, we can
prevent overfitting from occurring~\cite{caruana2001overfitting}.

Finding the right moment to stop
may take some
experimenting~\cite{prechelt1998automatic,prechelt1998early}.
A convenient and popular way is to add a third set to the training set/test set
duo which then becomes a training set, a test set, and a holdout validation set. The role
of the training set and the test set remains the same. However, after
each training epoch, the network is evaluated 
against the holdout validation set, to see  if under- or
overfitting occurs, and if we should stop training. Finally, the test set is used in the familiar
former role to see how well the trained network
generalizes to unseen instances. In this way, overfitting can be
prevented dynamically during
training~\cite{prechelt1998early,bishop2006pattern,goodfellow2016deep}.

\subsubsection*{Dropout}

A popular method to reduce overfitting is to introduce dropout layers
into the networks. Dropout reduces the effective capacity of the
network by stochastically dropping a certain percentage of neurons
from the backpropagation
process~\cite{hinton2012improving,srivastava2014dropout}. Dropout is
an effective and computationally efficient  method to reduce
overfitting~\cite{goodfellow2016deep}.


\subsubsection*{Batch Normalization}
Batch normalization periodically normalizes the
data~\cite{ioffe2017batch}, as we just mentioned. This has many benefits, including a
reduction of overfitting. 


\subsection{Datasets and Networks}\index{Fei-Fei Li}
Now that we have discussed in depth the deep learning algorithms and
their challenges and solutions, it is time to look at datasets and
methods that have been instrumental in the progress of the field.

The algorithmic advances that were discussed in
Sect.~\ref{sec:algo} were facilitated by the availability of large
training sets  of labeled  examples. Deep learning has benefited from
a number of \emph{Drosophilas}.
A well-known  repository of  datasets is
the University of California at Irvine Machine Learning
repository~\href{https://archive.ics.uci.edu/ml/index.php}{UCI}.\footnote{\url{https://archive.ics.uci.edu/ml/index.php}}\index{UCI}

The MNIST database may well be the best known of these datasets. MNIST
is a database of 60,000 handwritten digits provided by the National
Institute of Standards and Technology. The original MNIST database is
available at
\href{http://yann.lecun.com/exdb/mnist/}{MNIST}.\footnote{\url{http://yann.lecun.com/exdb/mnist/}}\index{MNIST}
CIFAR is one of the most widely used set of images  in machine learning
and vision~\cite{krizhevsky2009learning}.\index{CIFAR} CIFAR-10 has also 60,000 images in 10 classes: airplanes, cars, birds, cats, deer, dogs, frogs, horses, ships, and trucks. 
Another major data set of central importance to deep learning is ImageNet~\cite{fei2009imagenet,deng2009imagenet}.\footnote{\url{http://www.image-net.org}}\index{ImageNet}  
It is a collection of more than 14 million URLs of images that have
been hand annotated with the objects that are in the picture. It contains
more than 20,000 categories. A typical category contains several
hundred training images.

The importance of ImageNet for the progress in AI is
large. The availability of labeled images allowed algorithms to learn,
and new algorithms to be created. ImageNet was conceived by Fei-Fei Li
in 2006, and in later years she developed it further. Since 2010 ImageNet runs an annual software
contest, the ImageNet Large Scale Visual Recognition Challenge
(ILSVRC)~\cite{deng2009imagenet}. Since 2012 ILSVRC has been won by deep
networks, starting the deep learning boom in science and
industry.

We will now mention some influential networks architectures. We will
start with Yann LeCun's LeNet-5.



\subsubsection*{LeNet-5}
Towards the end of the 1990s the work on neural networks moved into
deep learning, a term coined by 
Dechter in~\cite{dechter1986learning}.
\index{Rina Dechter} Some
twenty years after the
introduction of deep convolutional neural nets by LeCun et
al.~\cite{lecun1998gradient} CNNs became highly popular. This paper
introduced the architecture LeNet-5.\index{LeNet-5} LeNet-5 is a
7-layer convolutional neural net trained to classify handwritten MNIST
digits from $32\times 32$ pixel images. It is a successful network that was
used commercially to recognize digits in banking checks. The paper
 has a thorough comparison of LeNet-5 with other methods such as
principal component analysis (PCA).\index{CNN}\index{convolutional network}\index{MNIST}\index{Keras}

At GitHub a modern Keras implementation of LeNet  is available (and
also of AlexNet  and
VGG).\footnote{\url{https://github.com/eweill/keras-deepcv/tree/master/models/classification}}
The code straightforwardly lists the layer definitions, shown in 
Listing~\ref{lst:lenet}.

\lstset{label={lst:lenet}}
\lstset{caption={LeNet-5 code in Keras~\cite{lecun1998gradient,weill}}}
\begin{lstlisting}[language=Python,float]

def lenet_model(img_shape=(28, 28, 1), n_classes=10, l2_reg=0.,
	weights=None):

	# Initialize model
	lenet = Sequential()

	# 2 sets of CRP (Convolution, RELU, Pooling)
	lenet.add(Conv2D(20, (5, 5), padding="same",
		input_shape=img_shape, kernel_regularizer=l2(l2_reg)))
	lenet.add(Activation("relu"))
	lenet.add(MaxPooling2D(pool_size=(2, 2), strides=(2, 2)))

	lenet.add(Conv2D(50, (5, 5), padding="same",
		kernel_regularizer=l2(l2_reg)))
	lenet.add(Activation("relu"))
	lenet.add(MaxPooling2D(pool_size=(2, 2), strides=(2, 2)))

	# Fully connected layers (w/ RELU)
	lenet.add(Flatten())
	lenet.add(Dense(500, kernel_regularizer=l2(l2_reg)))
	lenet.add(Activation("relu"))

	# Softmax (for classification)
	lenet.add(Dense(n_classes, kernel_regularizer=l2(l2_reg)))
	lenet.add(Activation("softmax"))

	if weights is not None:
		lenet.load_weights(weights)

	# Return the constructed network
	return lenet


\end{lstlisting}


\subsubsection*{AlexNet}\index{AlexNet}\index{GPU}
In 2012 Krizhevsky et al.\ 
published
the AlexNet architecture~\cite{krizhevsky2012imagenet}\index{Alex
  Krizhevsky}\index{Geoffrey Hinton} with 
breakthrough performance on the ImageNet dataset of 85\% accuracy. They  used a deep
network of 8 layers. The first 5 layers were convolutional layers,
some with max pooling, and 3 layers were fully connected
layers. Krizhevsky et al.\ overcame problems of vanishing 
gradients and overfitting with ReLU, data augmentation, and L2 regularization, and effectively used the power of
GPUs to train a large network. Their work caused many follow-up works 
showing even better performance, and attracted much attention, setting
off the general interest in deep learning.

As we shall see shortly, deep networks have since grown further in size, with  more layers, and
different learning algorithms and connection structures. The number of weights in
such networks is often in the millions, greater than the number of observations on
which the network is trained. This makes overfitting an important
problem, and many of the methods that we discussed previously have been
developed because of these larger networks.


\subsubsection*{Crafting Neural Architectures}
There is a flip side to the advantages of automated feature discovery
with deep learning, since for many
domains much time has to be 
spent finding a neural network
architecture for the domain that works, before the training process results in
features that perform well. Some of the time previously 
spent hand crafting heuristic features is now spent hand crafting network
architectures to automatically learn the features.\footnote{Of course, research
 has now focused on ways to automate the hand crafting of network
architectures. Section~\ref{sec:nas} describes ways to
automatically generate appropriate network architectures for different
domains in neural architecture search.}

\subsubsection*{TensorFlow}\index{TensorFlow}\index{GPU}\index{Torch}\index{Theano}\index{Caffe}\index{Keras}
The deep learning breakthrough was caused by the co-occurrence of
three major developments: (1)
algorithmic advances that solved key problems in deep learning, (2) the
availability of large datasets of labeled training data, and (3)  the
availability of computational power (GPU).

The most expensive operations in  image processing and neural
network training are essentially operations on matrices. Matrix
operations are some of the most well-studied problems in computer science. Their
algorithmic structure
is well understood, and for  basic linear algebra
operations  high-performance parallel implementations
for CPUs exist, such as the BLAS~\cite{dongarra1988extended,choi1996pb}.

Graphical processing units, or GPUs, were originally
developed for fast processing of image and video data, largely driven
by the video gaming industry. Modern GPUs consist of 
thousands of processing units optimized to process linear algebra matrix operations
 in
 parallel~\cite{sanders2010cuda,liu2012sparse,song2014scaling},
 offering matrix performance that is orders of magnitude faster than
 CPUs~\cite{oh2004gpu,cirecsan2010deep,sze2017efficient}. 

Neural network training packages support
GPU parallelism. Well-known dedicated deep learning packages are Berkeley's
Caffe~\cite{jia2014caffe}, Facebook's
PyTorch~\cite{paszke2019pytorch}, Theano~\cite{bergstra2011theano}
which has been subsumed by Google's TensorFlow~\cite{tensorflow2015whitepaper,abadi2016tensorflow}, and
its user-friendly add-on Keras~\cite{chollet2017deep}. Most packages
have Python interfaces. Some machine learning and mathematical 
packages also offer deep learning tools, such as MATLAB and
R.

TensorFlow provides high-quality
implementations of many machine learning and neural network algorithms
and operations.  (A
tensor is a  multi-dimensional array, often used for transformations.)
The programming concept of
TensorFlow takes some getting used to.  Programs are constructed as a data-flow
graph in which the sequence of tensors defines the operations. A
higher level, easier, interface is provided by Keras. Keras comes with
TensorFlow, and is recommended for learning the packages.

Implementing a full working deep learning algorithm that performs well
on a new problem is a challenging task. Many problems of
backpropagation, gradient descent, overfitting, vanishing gradients,
and numerical stability have to be solved. 
It is because of the free availability of high-quality deep learning
methods that so much progress has been made in  recent years. Advances in
image recognition, speech recognition, game playing, automated
translation, and autonomous vehicles are made possible in  large part by
these software suites.  The free availability of high-quality
implementations may be the fourth reason for the deep learning
breakthrough.  Appendix~\ref{ch:env} contains overviews of  machine learning packages.

\subsubsection*{Conclusion}
We have discussed in depth methods for deep supervised learning. This
has been a long section; many
problems had to be overcome to achieve end-to-end learning.
The deep learning breakthrough in image
recognition was made possible only by
the convergence of better algorithms,  computational power, and
large labeled datasets.

\section{*Advanced Network Architectures}

The breakthroughs in deep learning have prompted much further
research in fascinating areas. 
Let us now look deeper at more advanced deep learning methods. Table~\ref{tab:sl-enh} shows advanced network
architectures that we will discuss.
%
This is a starred section, with more advanced material, which  may
be skipped when in a hurry.

\begin{table}[t]
  \begin{center}

\begin{tabular}{lccc}
  {\bf  Name} & {\bf Principle} & {\bf Applicability} & {\bf Effectiveness}  \\ \hline\hline
  AlexNet & CNN & image & improve  \\ 
  inception & net inside net & image & improve   \\ 
  residual & skip links & image & improve \\ 
  GAN      & generative-adversarial & image   & robustness  \\ 
  RNN/LSTM & state & sequential proc & sequences      \\ \hline
\end{tabular}

  \caption{Advanced network architectures}\label{tab:sl-enh}
  \end{center}
\end{table}

Much of the work in supervised learning is driven by image recognition
tasks. Just as in game playing, the availability of clear benchmarks and competitions  has
facilitated  progress. The ImageNet
database~\cite{deng2009imagenet,russakovsky2015imagenet} 
has been at the center of this field. Guo et al.\ provide a
review of the state of the art of visual
recognition~\cite{guo2016deep}.


\subsection{Examples of Concrete Convolutional Networks}\label{sec:conv}

As we saw, after the turn of the millennium two important
developments  happened: large databases of (hand) labeled images 
became available, and  powerful GPUs became  available and
were used for training neural networks.
Together, these two developments
allowed major improvements in training speed and accuracy.
Let us start by having a closer look at AlexNet.

\subsubsection*{AlexNet}\index{AlexNet}
The 2012 ImageNet database as used by AlexNet has  14 million labeled
images. The network featured a highly optimized 2D two-GPU implementation of 5
convolutional layers and 3 fully connected layers. The filters in the
convolutional layers are $11\times 11$ in size.

The neurons use a ReLU activation function. In AlexNet
images were scaled to $256\times 256$ RGB pixels.
The size of the
network was large, with 60 million parameters. This causes
considerable overfitting. AlexNet used data augmentation and dropouts
to prevent overfitting.

Krizhevsky et al.\ won the 2012 ImageNet competition with an error rate of
15\%, significantly better than the number two, who achieved 26\%.  Although there were earlier reports of
CNNs that were successful in applications such as bioinformatics and
Chinese handwriting recognition, it was this convincing win of the 2012 ImageNet
competition for which  AlexNet has become well known.

AlexNet has become  an important network architecture for research. Many resources are available for AlexNet.
The original AlexNet code is available on GitHub at
\href{https://github.com/akrizhevsky/cuda-convnet2}{AlexNet}.\footnote{\url{https://github.com/akrizhevsky/cuda-convnet2}} Berkeley's
Caffe project maintains a ``model zoo'' where multiple trained models
are available, complete with  Caffe code that can be studied. The
AlexNet Caffe model is
available on GitHub at
\href{https://github.com/BVLC/caffe/tree/master/models/bvlc\_alexnet}{BVLC
  AlexNet}.\footnote{\url{https://github.com/BVLC/caffe/tree/master/models/bvlc\_alexnet}}
Also, as mentioned before, a Keras implementation of AlexNet, LeNet,  and VGG is available at
\href{https://github.com/eweill/keras-deepcv/tree/master/models/classification}{Models
  Keras}.\footnote{\url{https://github.com/eweill/keras-deepcv/tree/master/models/classification}} 

\subsubsection*{ZFnet}\label{sec:zfnet}
\begin{figure}[t]
    \centering{\includegraphics[width=12cm]{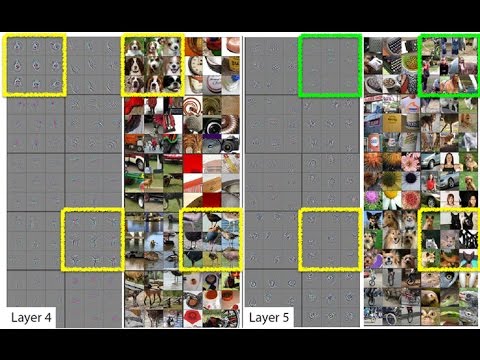}
     \caption{ZFnet layers~\cite{zeiler2014visualizing}}\label{fig:zfnetlayers9}}
\end{figure}

A year later a paper by Zeiler and Fergus~\cite{zeiler2014visualizing}
 improved upon AlexNet. Their network has become known as ZFnet. The work also
provided an enlightening explanation about how convolutional nets
work, including a
visualization of the inner layers (Fig.~\ref{fig:zfnetlayers9}). In a sense, this allows a look into
the brain of the neural net, to see what features it recognizes at
which layer.\index{ZFnet} Afterwards many papers have provided these
kinds of insights into the hidden layers of their network.

\label{sec:zflayers}
Deep network layers learn  a hierarchy of abstract  features. The
layers learn concepts of increasing 
generality~\cite{bengio2013representation,lecun2015deep}, creating a
hierarchy of abstract concepts. The look into the ``brains'' of the
net does not yield pictures that are easy to interpret. How general are the concepts learned by self-play?
What do they ``look'' like, can they be represented as a
two-dimensional picture for us to interpret? A deeper understanding of 
these concepts might enable us to learn more complex representations,
and thus achieve better local minima in the error
function~\cite{pezzotti2017deepeyes,wang2020cnn}. 
Montavon et al.~\cite{montavon2018methods} provides an tutorial on
methods for interpreting deep nets.

\subsubsection*{VGG}\index{VGG}
Simonyan  and Zisserman~\cite{simonyan2014very} followed a different
philosophy than AlexNet and ZFnet, with small filters ($3\times 3$) and a deep hierarchy of 16-19
layers. Their VGG net was also successful. VGG
faithfully implements the idea behind deep nets as needing many layers (16)
in order to build a hierarchical representation of visual data~\cite{lecun2015deep}.

\subsubsection*{GoogLeNet} 
A further notable network is GoogLeNet~\cite{szegedy2015going}, which
is even deeper, with 22 layers. The
new architecture and algorithms perform better than AlexNet, yet have
12 times fewer parameters, showing how strong the new architecture and
algorithms are. The GoogLeNet architecture is about sparse networks,
allowing a deeper network without increasing the total number of
parameters. The network is built around  {\em inception\/}
modules.\index{inception}\index{GoogLeNet} In previous architectures,
convolutional layers were stacked sequentially on top of each
other. Inception modules are small 
networks inside a network, where different sized convolutions are placed in parallel
to each other. Through the inception modules  sparse network
structures were achieved, reducing training time. In addition to the
inception modules, batch 
normalization\index{batch normalization} was used
to prevent vanishing
gradients~\cite{szegedy2016rethinking} \index{GoogLeNet} and dropouts
for regularization.

\begin{figure}[t]
    \centering{\includegraphics[width=7cm]{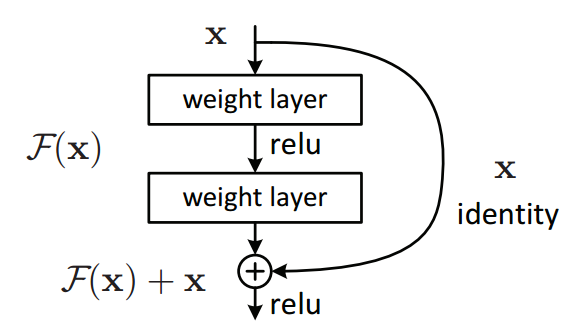}
     \caption{Residual cell~\cite{he2016deep}}\label{fig:rescell}}
\end{figure}

\begin{figure}[t]
    \centering{\includegraphics[width=13cm]{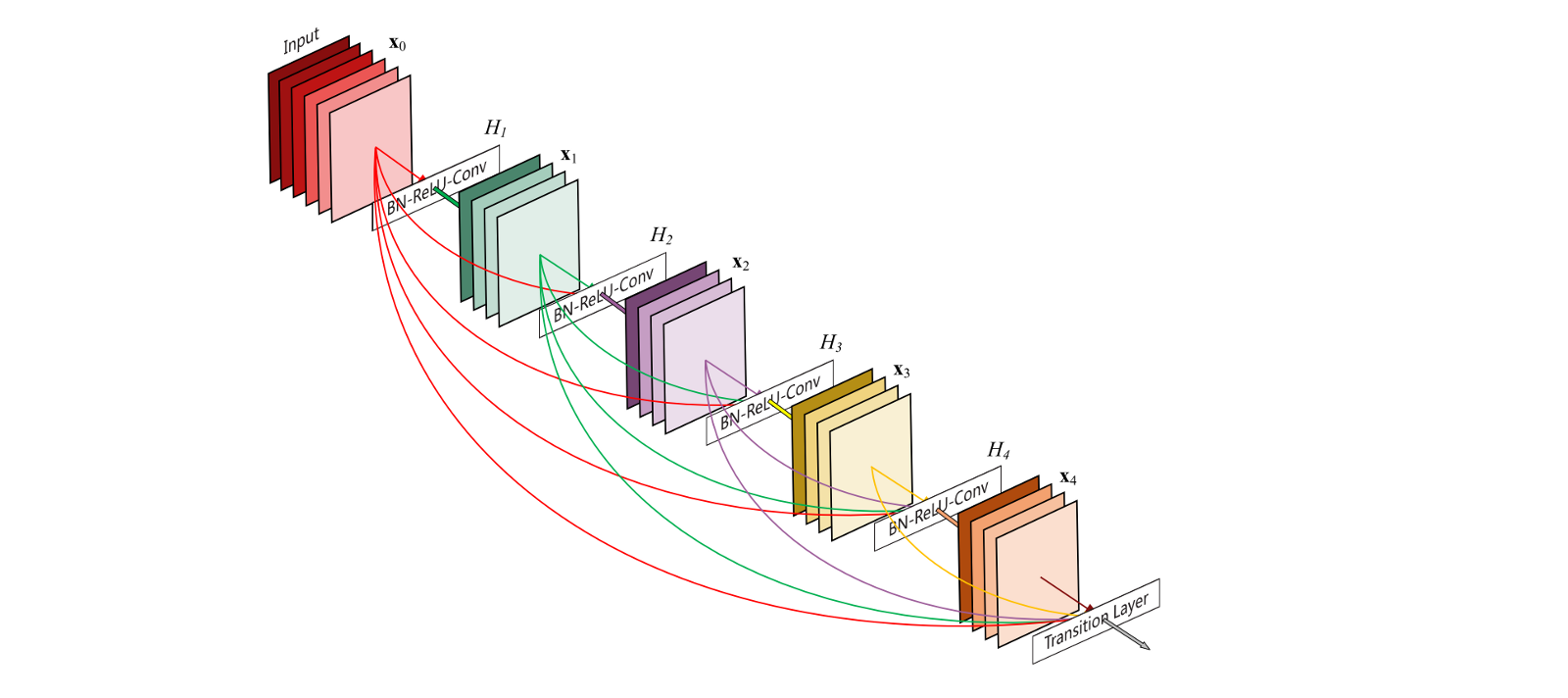}
     \caption{Residual net~\cite{he2016deep}}\label{fig:resnet}}
\end{figure}

\subsubsection*{Residual Networks}\label{sec:resnet}
After inception modules, another important innovation is the residual network architecture,
or {\em ResNets}. This
was introduced in the 2015 ImageNet challenge, which He et al.\ won with a
very low error rate of 3.57\%. This error rate is actually lower than
what most humans achieve:
5-10\%~\cite{he2016deep}.\index{ResNet}\index{residual net} ResNet has
no fewer than 152 layers, and, as 
before, this is achieved through a {\em simpler\/} architecture with
fewer connections (fewer
parameters than VGG). The problem of vanishing gradients, which deep networks suffer
from, is addressed with intermediate normalization steps.

The main contribution of ResNets is based on the following
observation. The authors found that when adding more layers to
their net, they were obtaining {\em lower\/} training accuracy.  This
is counterintuitive, since more parameters should allow the net to
achieve at least the accuracy of the shallower net. 
The deep network suffered from
degrading training accuracy due to reasons related to
overfitting~\cite{he2016deep}.

Residual nets  introduce {\em skip links\/} to 
cure this degradation (Fig.~\ref{fig:rescell}). Skip links are connections skipping one or
more layers, allowing the training to go directly to other layers
reducing the effective depth of the network (Fig.~\ref{fig:resnet}). Skip links create a mixture of a shallow and a deep
network, preventing the accuracy degradation and vanishing gradients
of deep networks. Furthermore, He et al.\ note that normal
layers have a harder time  learning of an identity function, than
learning a mapping to 0, and the  skip layers allow the net to more
easily learn an identity function.

Following up on ResNets, DenseNets were devised, building on the main
insight behind residual nets, and achieving even more
impressive results~\cite{huang2017densely}.

These architectural
innovations have again caused important  performance improvements. As a result, image
recognition tasks  now frequently exceed human
performance~\cite{guo2016deep,dodge2017study}. This is an active area
of research, and better architectures are presented at the major
machine learning conferences each year.

\subsection{Generative and Sequential Architectures}
In addition to mainstream image recognition architectures, there are
 other architectural innovations of different kinds of networks.


\begin{figure}[t]
  \centering{\includegraphics[width=\textwidth]{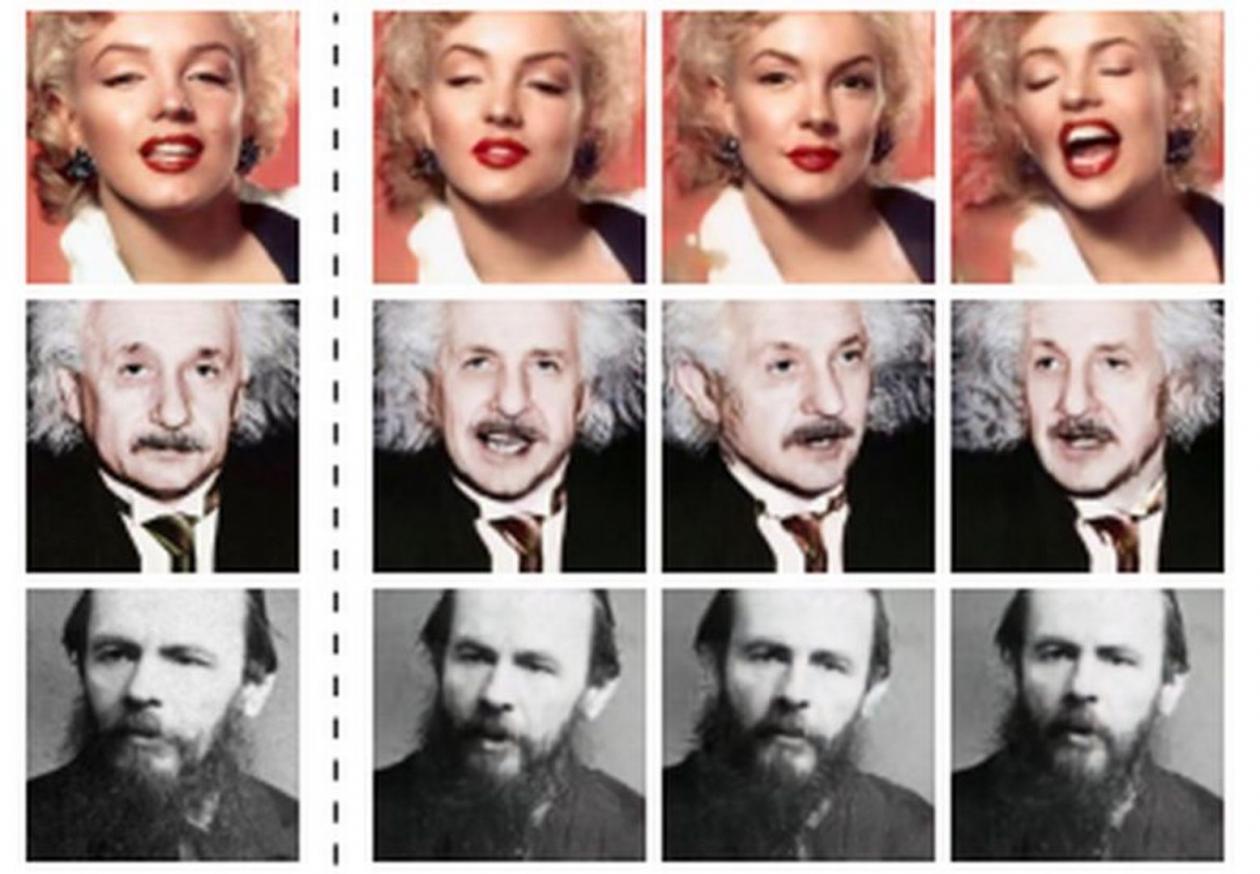}
     \caption{Deep fake~\cite{yang2019exposing}}\label{fig:deepfake}}
\end{figure}

\begin{figure}[t]
    \centering{\includegraphics[width=\textwidth]{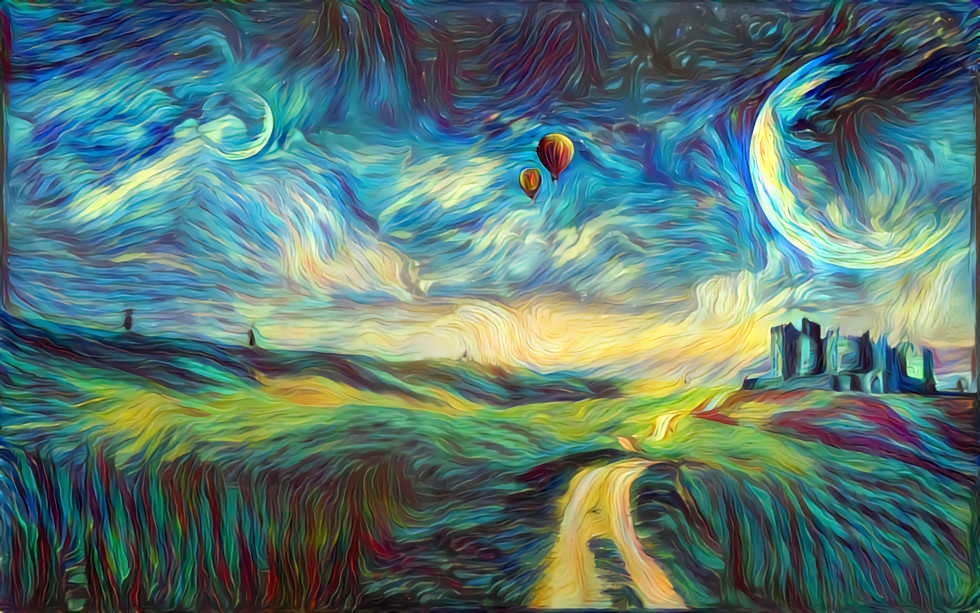}
     \caption{Deep dream~\cite{karras2017progressive}}\label{fig:deepdream}}
\end{figure}

\subsubsection*{Generative Adversarial Networks}\index{deep fake}\label{sec:onepixel}
An active research area is deep generative modeling. Generative
models are models from which a new example can be sampled. (A contrast
to generative models are the regular discriminative models, which output
a label, for classification. In a sense, a generative  model is a
discriminative model, but backwards.) There are several classes of generative
models. An important and popular class that has made quite an impact
is the generative 
adversarial
network, or
GAN~\cite{goodfellow2014generative}.\index{GAN}\index{generative adversarial net}

It was found that if an image is slightly perturbed,
and imperceptibly to the human 
eye, deep networks can easily be fooled to characterize an image as the wrong
category~\cite{szegedy2013intriguing}.\index{adversarial attack}\index{one-pixel problem} This brittleness is known as the
one-pixel problem~\cite{su2019one,szegedy2013intriguing}.  Deep networks are susceptible
to adversarial attacks. The
one-pixel problem   has spawned an active area of research to
understand this problem, and to make image recognizers more robust.
(Adversarial attacks are also relevant to deep reinforcement learning;
Sect.~\ref{sec:rladversary}.)

GANs are generative models that generate
adversarial examples. The purpose of adversarial examples  
is to fool the discriminator (recognizer). They are {\em designed\/} to cause a wrong
classification.
In GANs one network, the generator, generates an image, and a second network, the
discriminator, tries to recognize the image. The goal for the generator is
to mislead the discriminator, in order to make better discriminators.

In addition to making recognizers more
robust against the one-pixel problem, one of the other uses of
generative networks is to generate
artificial photo-realistic images such as {\em deep fake\/}
images~\cite{yang2019exposing}; see Fig.~\ref{fig:deepfake} and {\em deep dreaming},
Fig.~\ref{fig:deepdream}\footnote{\href{https://deepdreamgenerator.com}{Deep Dream Generator}
  \url{https://deepdreamgenerator.com}}~\cite{karras2017progressive}.
GANs have significantly increased our theoretical
understanding of supervised training.\index{deep fake}\index{deep dreaming}

\subsection{Sequential Processing}\index{RNN}\index{recurrent
  neural network}\index{captioning challenge}\index{LSTM}
Image recognition  has had a large
impact on network architectures, leading to  innovations in
network architectures such as convolutional nets.

Other applications, such as  time series analysis and speech
recognition,  have also caused new architectures to
be created. Image recognition is a single-time-step modeling task. In contrast,
speech recognition, text mining, and time series analysis concern sequential data that must
be modeled. Such sequences can be
modeled by recurrent neural nets. Some of the better known recurrent
neural nets are Hopfield
networks~\cite{hopfield1982neural}, recurrent neural
nets (RNN)~\cite{bertolami2009novel,fernandez2007application},  and long
short-term memory (LSTM)~\cite{hochreiter1997long}.

\begin{figure}[t]
\begin{center}
\includegraphics[width=2cm]{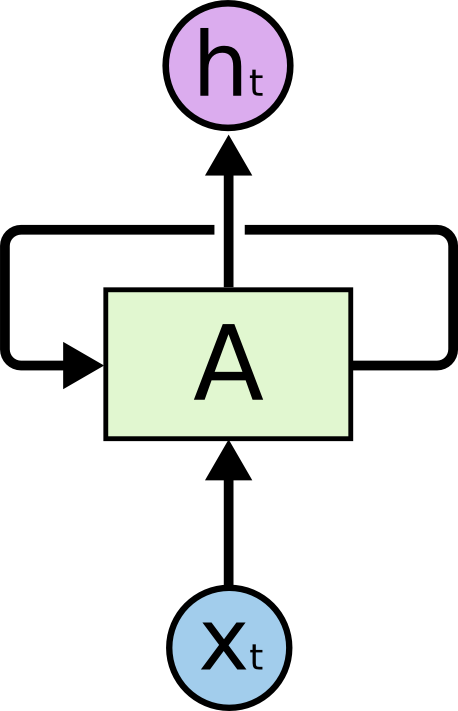}
\caption[RNN]{RNN $x_t$ is an input vector, $h_t$ is the
  output/prediction, and $A$ is the RNN \cite{colah2015}}\label{fig:RNN-rolled}
\end{center}
\end{figure}

\begin{figure}[t]
\begin{center}
\includegraphics[width=6cm]{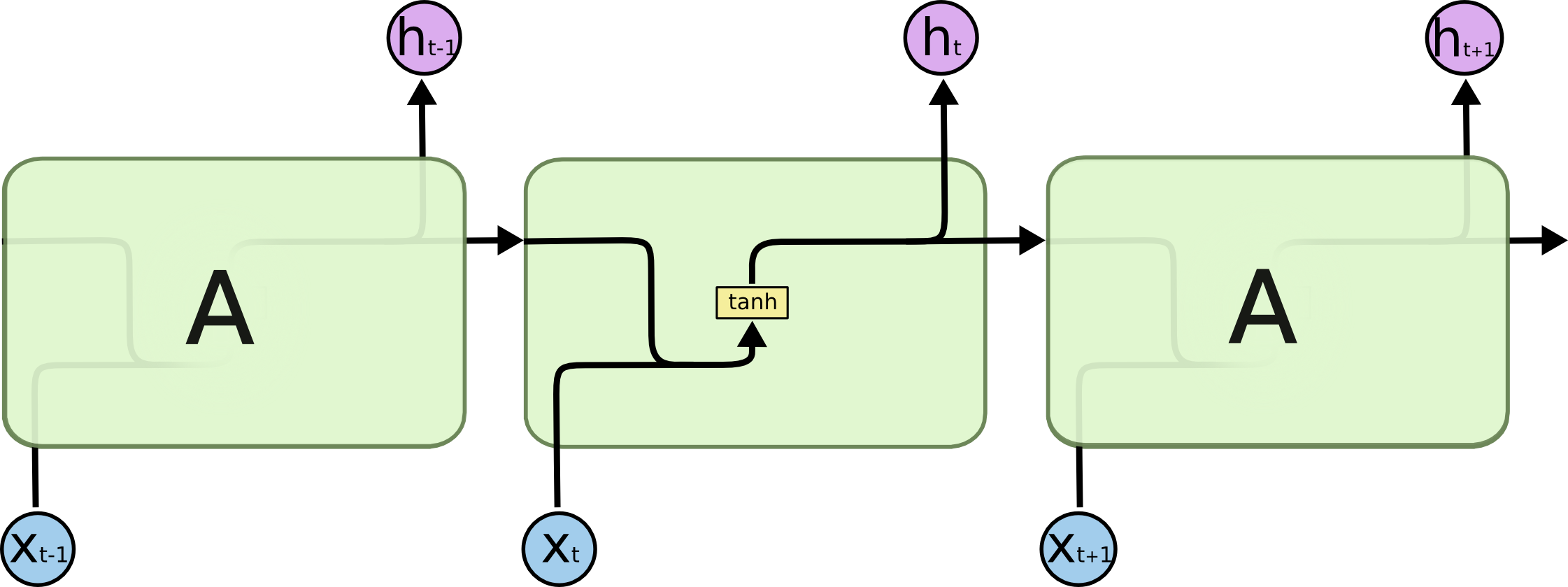}
\caption[Time-unrolled RNN]{Time-unrolled RNN neuron, with a $\tanh$ activation function;
  shown are the previous and next time steps as well \cite{colah2015}}\label{fig:simplernn}
\end{center}
\end{figure}

Figure~\ref{fig:RNN-rolled} shows a basic recurrent neural network. An
RNN neuron is the same as a normal neuron, with input, output, and
activation function. However, RNN neurons have an extra pair of looping input/output
connections. Figure~\ref{fig:simplernn} shows the internal structure
of an RNN neuron (a $\tanh$ activation  function) with their loop connections unrolled as previous and
next time steps. Through this structure, the values of the parameters
in an RNN can evolve. In effect, RNNs have a kind of variable-like state.

\begin{figure}[t]
\begin{center}
\includegraphics[width=10cm]{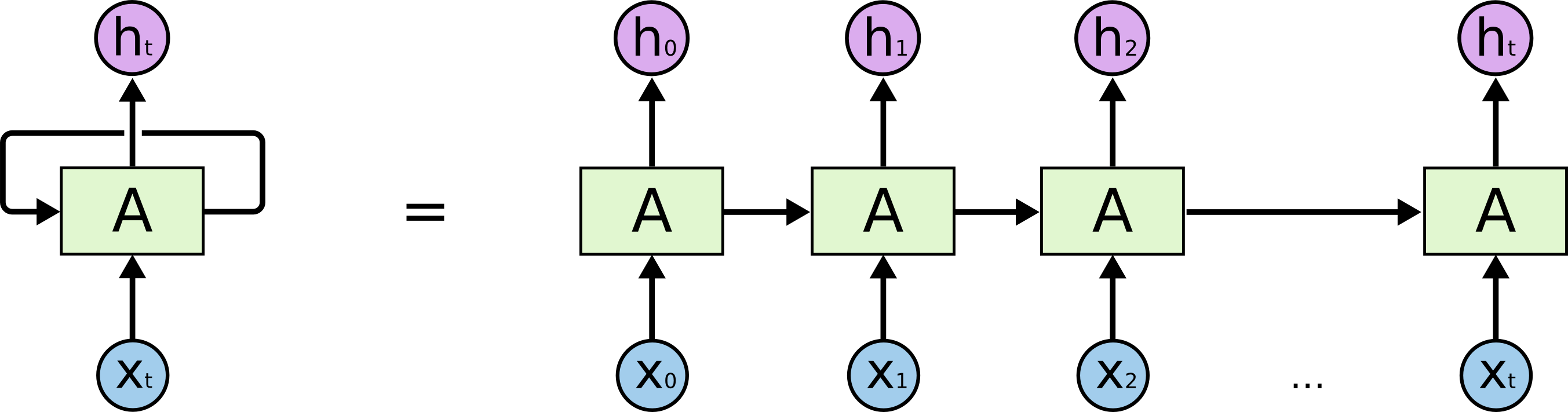}
\caption{RNN unrolled in time \cite{colah2015}}\label{fig:RNN-unrolled}
\end{center}
\end{figure}

\begin{figure}[t]
\begin{center}
\includegraphics[width=10cm]{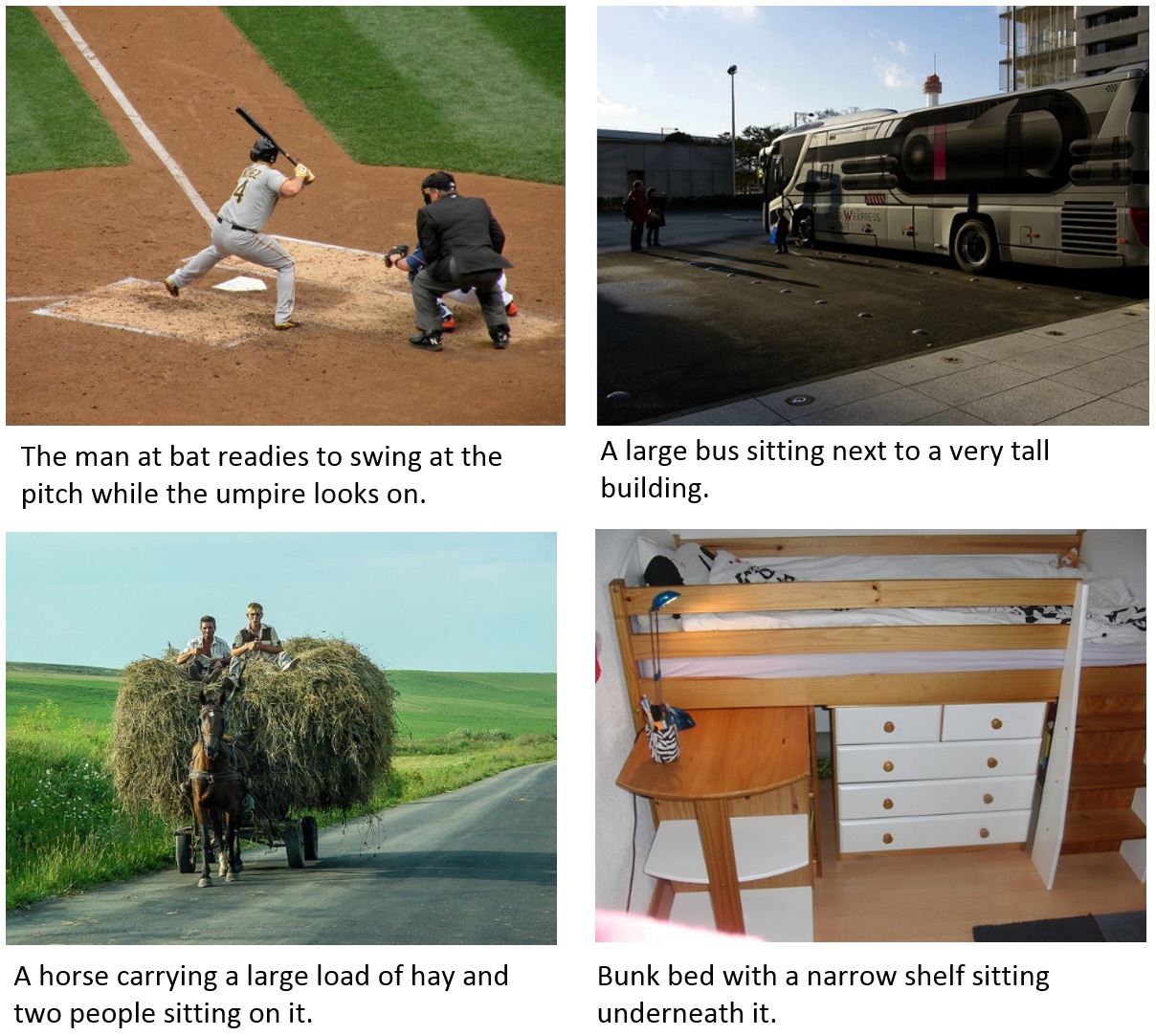}
\caption{Captioning tasks \cite{chen2015microsoft}}\label{fig:coco}
\end{center}
\end{figure}

To understand how RNNs work, it helps to unroll
the network, as has been done in Figs.~\ref{fig:simplernn} and
\ref{fig:RNN-unrolled}. The
recurrent neuron loops have 
been drawn as a straight line to show the network in a deep layered style, with 
connections between the layers. In reality the layers are  time steps in
the processing of the recurrent connections.
In a sense, an RNN is a deep neural net folded into a single layer
of recurrent neurons. 

Where deep convolutional networks are successful
in image classification, RNNs are used for more demanding tasks, such
as captioning challenges. In a captioning task the network is shown a
picture, and then has to come up with a textual description that makes
sense~\cite{vinyals2015show}. Figure~\ref{fig:coco} gives examples of captioning tasks, from
Chen et al.~\cite{chen2015microsoft}.

Captioning tasks are hard, and a computer succeeding in ``seeing'' an image and ``interpreting'' it  with a 
caption that makes sense, describing what can be seen on the picture, can be a
startling experience. Seeing one's own RNN learning  to generate the first
appropriate descriptions of images can be a satisfying experience for
 machine learning researchers (see the exercises at the end of this
 chapter). Interestingly, simple models can deliver 
surprisingly good results.


A step up from image captioning is video captioning, where a
description has to be generated for a sequence of
images~\cite{venugopalan2015sequence}. 


\begin{figure}[t]
\begin{center}
\includegraphics[width=12cm]{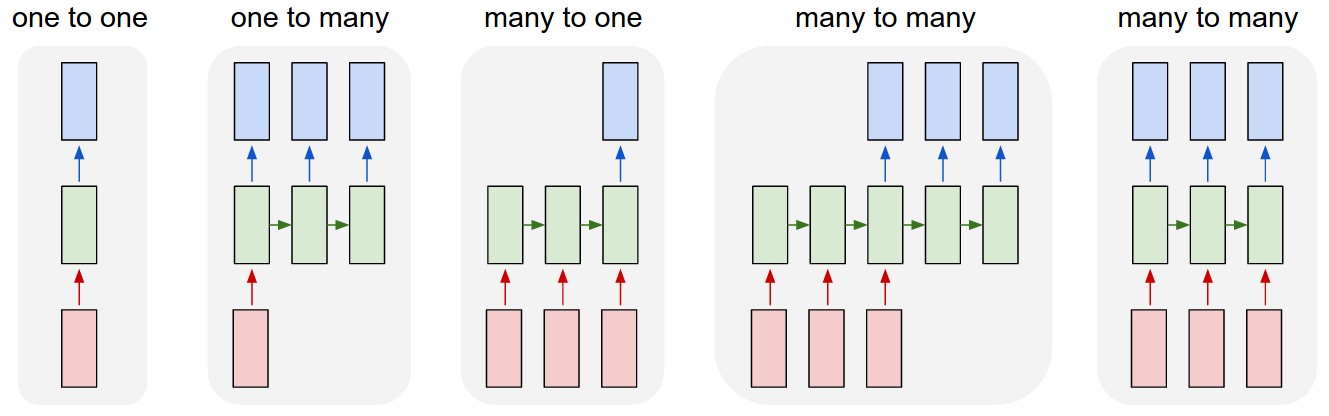}
\caption{RNN configurations \cite{karpathy2015}}\label{fig:RNN-many}
\end{center}
\end{figure}


The main innovation of  recurrent nets is that they allow us to work
with sequences of vectors. Figure~\ref{fig:RNN-many} shows
different 
combinations of sequences that we will discuss now. There can be sequences in the input, in
the output, or in both.  Karpathy has written an accessible and
well-illustrated blog on
the different RNN
configurations~\cite{karpathy2015}.
The figure shows different rectangles. Each rectangle is a vector. 
Arrows represent computations, such as matrix
multiply. Input vectors are in red, output vectors are in blue, and
green vectors hold the state. Following Karpathy~\cite{karpathy2015}, from left to
right we see:
\begin{enumerate}
  \item  {\em One to one\/}, the standard network  without RNN. This network
maps a fixed-sized input to fixed-sized output, such as an image
classification task (picture in/class out).
\item {\em One to many\/} adds a sequence in the output. This can be an
image captioning task that takes an image and outputs a sentence of 
words.
\item {\em Many to one\/} is the opposite, with a sequence in the input. Think for example of
sentiment analysis (a sentence is classified for words with negative
or positive emotional meaning).
\item {\em Many to many\/} has both a sequence for input and
a sequence for output. This can be the case in machine
translation, where a sentence in English is read and then a sentence
in French is produced.
\item {\em Many to many\/} is a related but different situation, with synchronized
input and output sequences. This can be the case in video 
classification where  each frame of the video should be labeled.
\end{enumerate}

\subsubsection*{State}
Time series prediction  is  a difficult  
 problem, for which  sequential processing  is  well
 suited. Sequential networks are more powerful
than  fixed networks, which can only process  a predetermined
number of  steps.  Moreover, where  normal deep networks
represent a single  stateless mathematical function mapping from input to output,
RNNs combine the 
input vector together with the state from the previous time step to produce a new state
vector, to be fed into the next time step. In this particular sense
RNNs have state. The fact that RNNs have such state in addition to a learned
function brings them closer to a conventional imperative computer
program (such as a Python program with variables)~\cite{van2004concepts,abelson1996structure}. An RNN can be interpreted as 
a fixed program with certain inputs and some internal variables. In
this way, RNNs are  simple computer programs that do not
have to be programmed, but learn their functionality by training. RNNs
are   appealing for
building intelligent systems~\cite{geron2017hands}. This is an active
area of research.

Many different successful applications for recurrent neural networks
have been found. Graves et al.\ describe speech to text
processing~\cite{graves2014towards}. Machine translation is described
by Sutskever et al.~\cite{sutskever2014sequence}. Earlier work
described text generation~\cite{sutskever2011generating}.
RNNs for video classification are described by Donahue et
al.~\cite{donahue2015long}. Mnih et al.\ describe RNNs for visual
attention~\cite{mnih2014recurrent}.


\begin{figure}[t]
\begin{center}
\includegraphics[width=10cm]{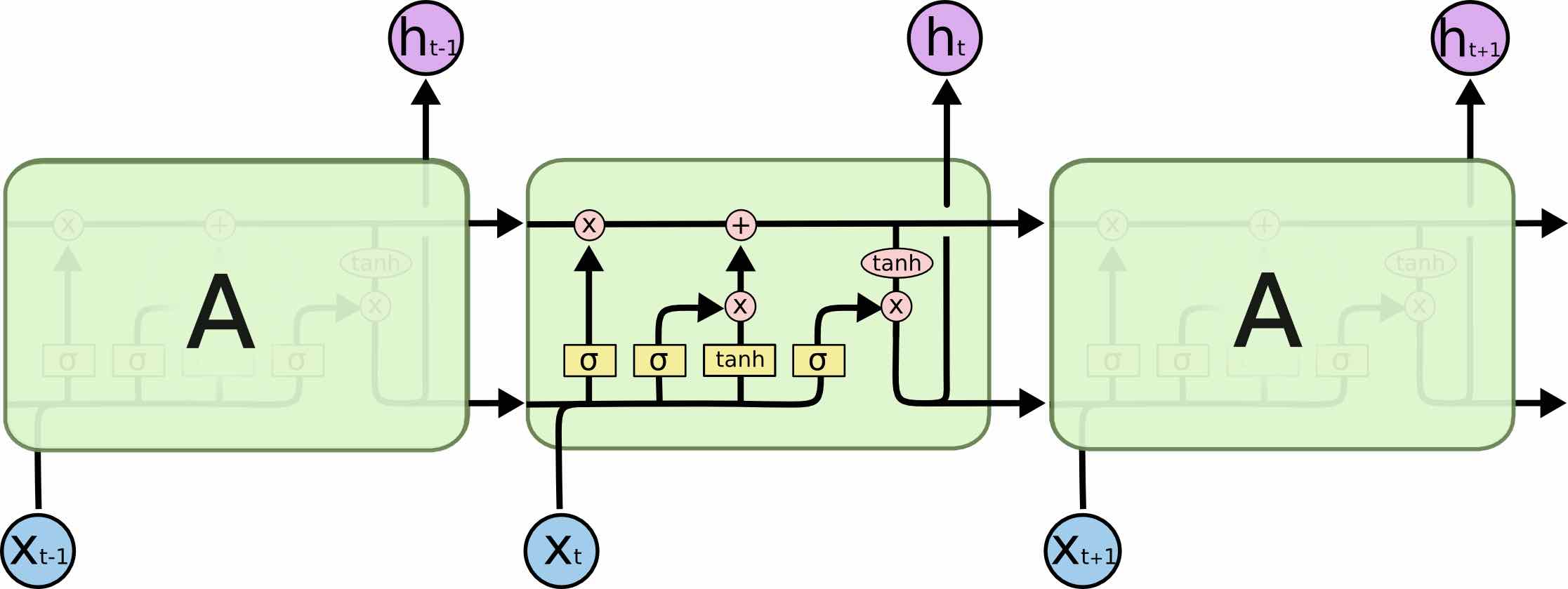}
\caption{LSTM \cite{karpathy2015}}\label{fig:LSTM3}
\end{center}
\end{figure}

\subsubsection*{Long Short-Term Memory}\index{LSTM}
Time series prediction is more complex than  conventional regression
or classification. It adds  the complexity of a sequence dependence among the input
variables.
 
LSTM  (long short-term memory) is a more  powerful type of neuron designed to handle
sequences. Figure~\ref{fig:LSTM3} shows the LSTM module,  allowing
comparison with the simple RNN of Fig.~\ref{fig:simplernn}. 
LSTMs are designed for sequential problems, such as time series, and
planning.
LSTMs were introduces by Hochreiter and
Schmidhuber~\cite{hochreiter1997long}.

RNN training suffers from the vanishing gradient problem. For
short-term  sequences this problem may be controllable by the same
methods 
as for deep CNN (Sect.~\ref{sec:vanishing}). For  
long-term remembering LSTM are better suited.

LSTM building blocks avoid the 
vanishing gradient problem by introducing
skip connections  that insert unchanged gradients between layers, just
like residual nets.
The skip links 
make LSTMS less susceptible to the vanishing gradient problem; this
makes LSTMs better suited for  large architectures.

LSTMs are frequently used to solve diverse problems~\cite{schmidhuber2002learning,graves2005bidirectional,greff2017lstm,gers1999learning}.
 Researchers have experimented with  LSTMs for creating
hybrid architectures that combine planning 
and learning, with some initial success~\cite{silver2017predictron}; see also
Sect.~\ref{sec:explicit}.




\subsubsection*{Conclusion}
There has been a wealth of research into deep neural networks for image
recognition (and beyond). Many complex deep architectures were
created,  and  achieved impressive recognition results.
Research on generative adversarial networks and sequence
processing is a highly active area of supervised learning research.

Let us now go on to the main topic of this book, deep reinforcement
learning, to see how deep learning methods work there.

\section{Deep Reinforcement Learning}\index{DQN}
\label{sec:dqn}
Deep reinforcement learning has shown inspiring results in
artificial intelligence. Before the successes, there were quite
some obstacles that had to be overcome. Methods from  supervised
learning were not as easy to transfer to reinforcement learning as some
 had hoped. There are challenges in generalization and 
 learning convergence. In this section we will
discuss these problems and the methods that have been found to address
them. Table~\ref{tab:suprl} summarizes some of the differences between
supervised learning and reinforcement learning.

The need for deep reinforcement learning came out of the wish  for
improving performance beyond heuristics and sampling, beyond what can be
achieved with (automated tuning of) hand-crafted features.
End-to-end learning requires training in
high-dimensional state spaces, far exceeding 
the capabilities of exact or heuristic planning methods, or of shallow
networks or other conventional machine learning methods. 


Let us now start at the
beginning, with the need for end-to-end  learning in a reinforcement
learning context.

\subsection{End-to-End Reinforcement Learning}\label{sec:rlfa}
The state space of problems for which hand-crafted features have been
created is typically of manageable size. Much of the
complexity of the problem domain is abstracted away by the
hand-crafted heuristics. The heuristics translate raw board features 
into a small number of heuristic parameters, converting a
high-dimensional state space  to a low-dimensional state
space. End-to-end learning 
does away with the hand-crafted heuristics, and takes the problem of
learning from the raw inputs head on. The computational complexity of
end-to-end learning problems is correspondingly  higher.

In the previous section we saw the problems that had to be
overcome in supervised learning to achieve end-to-end learning. Only
after convolutional neural networks had been invented, and after
solutions for overfitting and vanishing gradients had been found, did
deep end-to-end supervised learning
work.

In reinforcement learning a similar situation exists. Here the
problem was the quality of the generalization and the convergence of the training process.
Let us now see why stable deep reinforcement learning is such a difficult problem.


Reinforcement learning is a form of machine learning, where just as in
supervised learning, a function is learned from examples through generalization. The way in
which these examples are presented to the learning algorithm, however, is different.\footnote{Also note
  that the functions to be learned are different. In 
supervised learning  a classification or a regression function is typically learned. In
reinforcement learning  a policy or a value function is
learned.}


\begin{table}[t]
  \begin{center}
  \begin{tabular}{ll}
    {\bf Supervised learning} & {\bf Reinforcement learning} \\
    \hline\hline
    (example-label) or (example-number) pair & (state-action) or (state-reward) pair \\ 
    database of ground truths &  environment gives reward \\ 
    static learning process & interactive learning process \\
    fixed set of examples & changing and interdependent set of examples \\ 
    dense: ground truth for all examples & sparse: rewards for some
                                           examples \\ \hline
  \end{tabular}
  \caption{Supervised vs.\ reinforcement learning}\label{tab:suprl}
\end{center}
\end{table}

First, supervised learning  infers  functions from a database of 
example-label pairs, where the ground truth is given. Reinforcement
learning infers a function for state-action pairs (for the policy
function) or state-reward
pairs (for the value function). It has no fixed database of
examples with ground truths. Positive or negative rewards are only
provided after interaction with an environment. Actions by the agent
determine the response from the environment. This interaction
influences the learning process. The set of examples is
not fixed, and may change from learning process to learning
process. See Table~\ref{tab:suprl} for a summary.

The second difference is that in supervised learning ground truths are
present for {\em all\/}
example-label pairs. The state space is dense. In reinforcement
learning problems the state space is often sparse,  the outcome of an action may only be known after many
further steps, and the rewards are
only available for some of the state-action pairs. For other
state-action pairs the value or policy 
has to be inferred, propagated over a long distance of intermediate states
before the reward of a state-action pair is known. This gives rise to
the credit assignment 
problem, and, as we shall see next, to generalization problems and to
learning instability. 


\label{sec:stable}
Much research into the stability of reinforcement learning has been
performed. We will discuss generalization to test sets, correlation
between states,  fluctuations in the data distribution, and the deadly triad.


 \begin{figure}[t]
 \begin{center}
 \includegraphics[width=8cm]{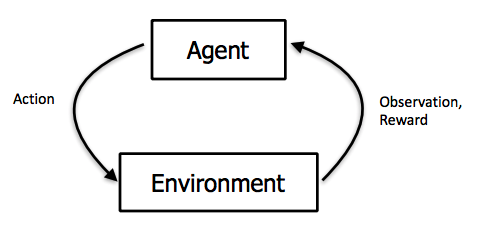}
 \caption{Agent/environment loop~\cite{sutton2018introduction}}\label{fig:action}
 \end{center}
 \end{figure}

\subsubsection*{Generalization to Test Set}
The first problem we will look into is generalization. This problem
concerns the ability to generalize training results to the {\em test} 
set.

 \begin{figure}[t]
 \begin{center}
 \includegraphics[width=8cm]{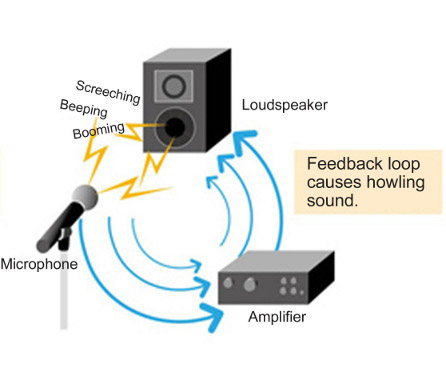}
 \caption{Feedback loop}\label{fig:feedback}
 \end{center}
 \end{figure}

Recall that a basic assumption for
convergence in Q-learning and other reinforcement learning algorithms
is that state-action pairs are sampled often enough. Clearly, this
is not the case in large state spaces with sparse rewards, especially
not when test samples are not part of the training set. The coverage of
the training set can be problem. Even worse,
the state-action pairs are not randomly distributed.

In supervised learning, data samples (states) are  independent and are
assumed to be distributed evenly over the state space. In the database of static images, there is no relation
between subsequent images, and examples are independently
sampled. Each image is separate (in contrast to a
movie, where many inter-temporal relations between frames are
present).

In model-based reinforcement learning a sequence of states is generated
by a simulator or game playing program in an
agent/environment loop (Fig.~\ref{fig:action}).  The features of the states are often
correlated, differing only by a single action, one move or one
stone. Subsequent
samples are correlated. Such a correlated training set may be myopic, covering only a
part of the state space. The training set will not work well with
standard supervised  training algorithms. The algorithms may get stuck
in local minima, and cycles may occur, which may turn into a
feedback loop when state-action pairs and values are mixed due to
function approximation (Fig.~\ref{fig:feedback} illustrates a feedback
loop in another domain).  

To illustrate myopic learning, let us consider an example of
correlated training  in  Chess. Imagine a program  in which training
got stuck in a loop with center control problems. Such a program knows much about center control but 
little about king safety, and therefore plays badly if it encounters a differently
trained agent that can outplay it in the king safety domain. We can
try to counter this effect, as we 
will see soon.

The low sample
rate and  correlations between subsequent states cause
generalization problems.
The end result
is that deep reinforcement learning algorithms that are trained on
sparse, correlated
end-to-end states may not 
generalize well when they are tested on new problem instances.

In addition to the problems of loops and local optima, learning  from
correlated consecutive samples is inefficient. Little new information
is contained in two  
states that are almost the same. The low learning efficiency translates
into a smaller effective dataset; much training is done, yet little
is learned.
Randomizing the samples breaks these
correlations and therefore reduces the probability of getting stuck.

\subsubsection*{Fluctuating Data Distribution}
A  further consequence of how samples are generated is that in
reinforcement learning the data
distribution may be fluctuating. 
In supervised learning the data set is fixed. In reinforcement learning 
output signals from the environment are infrequent or derived from
other states, and may thus be
``polluted'' by old values, and are calculated using values that may
be approximations.

Since in reinforcement learning training examples are generated
in interaction with the environment, unstable training processes
have consequences for the set of training examples. In 
reinforcement learning the  size of the
training set may vary, or may be infinite, as the training process
itself influences which examples are generated. 

This  has two consequences, one undesired, one desired. In
the short term, learning new behaviors may cause myopia or may get 
stuck locally  by feeding on its own behavior as described in
on-policy learning. This short-term training behavior can be smoothed
with a replay buffer by providing data from a larger
landscape~\cite{mnih2013playing}. 

In the longer term, learning new behaviors may
actually be desirable. Forgetting bad behaviors can also be a way to
escape ineffective parts of the state space. Thus, the replay buffer
should not be too large. In Chap.~\ref{chap:plan} we will
learn more about this  in the context of self-play.


Function approximation  may also cause problems with divergent training processes. 
For this, let us have a closer look at the so-called deadly triad. 

\subsubsection*{Convergent Training: Deadly Triad}\index{deadly triad (approximation,
bootstrapping, off-policy learning)}\label{sec:triad}
The
learning process of deep reinforcement learning itself can be
divergent. 

The
learning of Q-values may diverge, even if we assume that state-action pairs are sampled
frequently enough.
Tsitsiklis and Van Roy~\cite{tsitsiklis1997analysis}
showed in 1997 that combining off-policy reinforcement 
learning with nonlinear function
approximaters (such as neural networks)  could cause
Q-values to diverge. Sutton and
Barto~\cite{sutton2018introduction}  summarize three main
reasons for divergent training: function
approximation, bootstrapping, and  off-policy
learning. Together, these form the  so-called  {\em deadly triad\/} of
instability.
When these three techniques are combined, training can diverge,
and value estimates can become unbounded.\footnote{Actually, as
  Tsitsiklis and Van Roy~\cite{tsitsiklis1997analysis} showed, already two of the three may  be enough for
  instability. All three together increases the chance of divergent
  training more, unless special measures are taken.} 

{\em Function approximation\/} may   attribute values to states inaccurately. Whereas exact methods are designed to
recognize individual states, neural networks are
 designed to recognize individual features of states. These features may be
shared by different states, and values attributed to those features
are shared also by other states. Function approximation may cause
confusion or misidentification
of states. In a reinforcement learning process
where new states are generated on the fly, this sharing of values among states may cause 
loops or 
other forms of instability. If the accuracy of the approximation of
the true function values is good enough, then states may be identified
well enough to prevent most loops or divergent training processes. 

{\em Bootstrapping\/} of values increases the efficiency of the training because
values do not have to be calculated from the start, since previously
calculated values are reused. Bootstrapping is at the basis of the 
recursive Bellman equation, of temporal difference, and of Q-learning. However, errors or
biases in initial values may persist, and even spill over to other states
as values are propagated, and error values become
correlated between states. With function
approximation, inaccuracies in values are almost
guaranteed. Bootstrapping and function approximation can thus lead to divergent learning,
and to training loops getting stuck in one area of the state space.

{\em Off-policy\/}  learning uses a  behavior policy that is different from
the target policy that we are optimizing (Sect.~\ref{sec:off-policy}). When the behavior 
policy is improved, the off-policy 
values may not be improved, and the algorithm may not converge. Off-policy learning converges independently from the behavior
policy, and converges generally less well than
on-policy learning, especially when combined with function
approximation~\cite{sutton2018introduction}.

The consequence of the deadly
triad is that divergence may  occur.
In a divergent learning process  Q-values do not
converge, but diverge.\footnote{The deadly triad suggests us to use  on-policy instead of off-policy
learning, to improve 
the training convergence. However, when learning on-policy the target
network determines the next data sample to be trained on. Mnih et
al.~\cite{mnih2013playing} give an example of how on-policy learning
also leads to training divergence. Assume that  
the maximizing action is to move left. With on-policy learning the
training samples will be dominated by samples from the left-hand side;
if, for some reason, the maximizing action then switches to the right,
then the training distribution will 
also switch. In this way unwanted feedback loops may arise
and the network will get stuck in self-reinforcing features.}

For some time, most further research  in reinforcement
learning focused on linear function approximators, which have 
better convergence guarantees. Later, after the first positive results
of DQN~\cite{mnih2013playing}, experimental studies were performed
to find out under which circumstances convergence in the face of the deadly triad
can be achieved, causing many
further techniques  to be developed
(Sect.~\ref{sec:rainbow}).

\subsubsection*{Conclusion}
\index{TD-Gammon}
The preceding challenges to achieve stable deep reinforcement learning
might be summarized as problems of coverage, correlation, and
convergence: a lack of coverage of the state space, too much
correlation between subsequent states, and problematic convergence in
training due to the deadly triad.\index{coverage, correlation, convergence}

Although the theory suggests  reasons why function approximation may
preclude stable reinforcement learning, there were early indications
to the contrary
that stable training 
was possible in practice. TD-Gammon~\cite{tesauro1995td} used self-play
reinforcement learning, achieving stable learning in a shallow
network. Perhaps some form of stable reinforcement
learning was possible, at least in a shallow network?
TD-Gammon's training used a temporal difference
algorithm similar to Q-learning, approximating the value 
function with a  network with one hidden layer, based on
heuristic features.

TD-Gammon's success prompted attempts with TD learning in
Checkers~\cite{chellapilla1999evolving} and
Go~\cite{sutskever2008mimicking,clark2014teaching}. Unfortunately the 
success could not be replicated in these games, and it was believed
for some time that Backgammon was a special
case well suited for reinforcement learning and
self-play~\cite{pollack1997did,schraudolph1994temporal}.  One
suggestion was that perhaps the randomness of the dice 
rolls helped exploration 
and smoothing of the value function of the state space.

However, there were further reports of early successful
applications of deep neural networks  in a reinforcement learning
setting,  suggesting that stable deep reinforcement learning is
possible~\cite{heess2012actor,sallans2004reinforcement}, prompting
more work. The  
results in Atari (2013) and Go (2016) as well as further
studies~\cite{van2018deep} have now provided clear indications
that both stable training
and generalizing reinforcement learning are indeed possible, and
why.



So far, we have discussed reasons why deep reinforcement learning
processes may suffer from instability. Deep  reinforcement learning is a  process where the training examples are influenced by the training process itself, introducing the possibility
of divergent or self-reinforcing learning. We have also seen that, despite
these reasons, some papers reported stable learning (in addition to
TD-Gammon's success with shallow reinforcement learning). 

As noted before, attempts to follow up on TD-gammon's success with
neural nets and 
self-play in  Checkers and Go were not
successful at first. We will now
discuss results where
stable deep reinforcement learning was achieved on end-to-end reinforcement
learning of Atari 2600 games on the Atari Learning Environment
benchmark set (ALE)~\cite{bellemare2013arcade}.

\subsection{Atari 2600 Games}

Let us  look in more detail into the Atari 2600 experiments.



Learning actions
directly from high-dimensional sensory inputs such as sound and vision
is one of the long-standing challenges of artificial intelligence. In 2013, one year after
the AlexNet 
success in supervised learning, a paper
was published that took  reinforcement learning   a major step
further. Mnih et al.~\cite{mnih2013playing} 
published a  work on {\em end-to-end\/} reinforcement
policy learning in Atari games. Their approach is end-to-end in the sense that they 
learn joystick actions directly from the  visual state, the raw
pixels, without an 
intermediate step of hand-crafted heuristic features. The network
architecture and training algorithm were named DQN, for
Deep Q-Network.

\begin{figure}[t]
\begin{center}
\includegraphics[width=8cm]{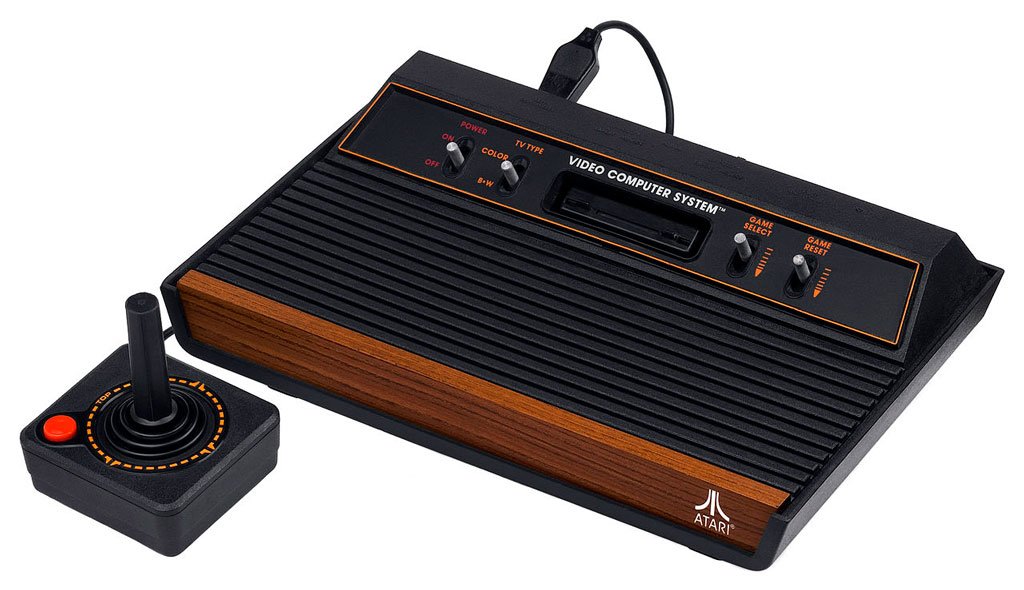}
\caption{Atari 2600 console}\label{fig:console}
\end{center}
\end{figure}

Mnih et al.\ applied deep learning to play 1980s arcade
Atari 2600 games, such as Space Invaders, Pong, and Breakout, directly
from the raw television screen
pixels~\cite{bellemare2013arcade}. Figure~\ref{fig:console} shows 
a picture of an Atari 2600 console. 

Being able to close the loop
from  pixels to policy is a major achievement. In their original 2013 workshop paper
Mnih et al.~\cite{mnih2013playing} were able to
achieve human-level play for six games. In a follow-up article in 2015 
they improved on their  work and were able to achieve a level of play
equal to humans for 49 of the games that were in the test
set~\cite{mnih2015human}.

It should be noted that there are a few Atari games that proved 
difficult to achieve strong play at. Notably  games
that required more long-range planning because long stretches of the
game do not give rewards, such as Montezuma's Revenge, were a
problem. Delayed credit assignment over long periods is still
hard.\index{Montezuma's Revenge}\label{sec:montezuma} In 2020 a paper
was published with success in 57 games using a meta-controller for
learning in the different games~\cite{badia2020agent57}.

In summary, a single architecture was able to successfully learn
control policies for many different games. Minimal prior knowledge was
used, and the neural network only processed the pixels and the
game score. The same network architecture and training procedure was
used on each game, although a network trained for one game could not
play another game well.

\subsubsection*{Arcade Learning
  Environment}\index{Atari}\index{ALE}\index{Arcade Learning Environment}
The  results  by Mnih et al.\ stimulated much further work in the area of
deep reinforcement learning. Successful end-to-end reinforcement
learning created much interest among  researchers, and many related
algorithms were developed. Let us   look into their methods and 
experiments.

\begin{figure}[t]
\begin{center}
\includegraphics[width=10cm]{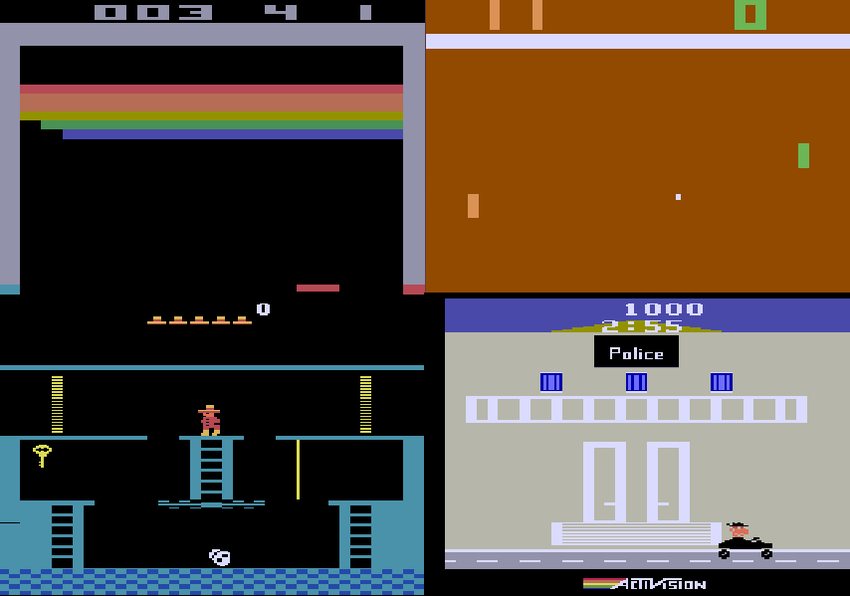}
\caption[Atari games]{Screenshots of 4 Atari Games (Breakout, Pong,
  Montezuma's Revenge, and Private Eye)}\label{fig:atarigames}
\end{center}
\end{figure}

The games that were used for DQN are from a standard benchmark set, the
Arcade Learning Environment (ALE)~\cite{bellemare2013arcade}. ALE  is a 
test-bed designed to provide challenging reinforcement learning
tasks. Among other things it contains an emulator of the Atari 2600
console. ALE presents agents with a high-dimensional\footnote{That is,
  high dimensional for machine 
  learning. $210\times 160$ pixels is not exactly high-definition
  video quality.} visual input
($210\times 160$ RGB video at $60$ Hz) of tasks that were designed to
be interesting and challenging for human players (Fig.~\ref{fig:atarigames}). The game cartridge ROM typically holds
the game code (2-4 kB), while the console memory is small, just 128
bytes. This is not a typo, the console memory really is only 128 {\em bytes}.
The actions can be input via a joystick, and a fire button (18 actions).

The original experiments were performed with seven Atari games. Training was performed on  50 million frames in total.

In their 2013 work~\cite{mnih2013playing} the neural network performed better than an expert human
player on Breakout, Enduro, and Pong. On Seaqest, Q*Bert, and Space
Invaders performance was far below that of a human. In these games a
strategy must be found that extends over longer time periods. In their
2015 work~\cite{mnih2015human} the net performed better than
human-level play in 29 of 49  
Atari games. Again, games with longer credit delay were more
difficult. 

Let us look in more detail at the network architecture of the 2013
 experiments.

\begin{figure}[t]
\begin{center}
\includegraphics[width=12cm]{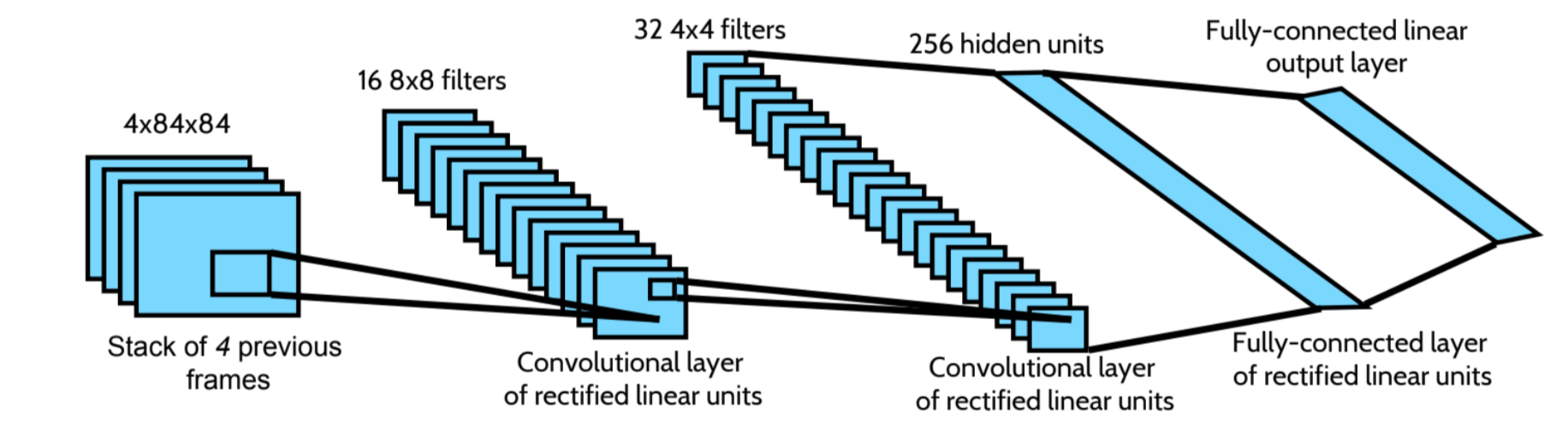}
\caption{DQN architecture \cite{hui2018rl}}\label{fig:dqnarch}
\end{center}
\end{figure}
 
\subsubsection*{Network Architecture}
The Atari task is a control task: the network trains a behavior policy directly from
pixel frame input. The task of processing raw frames involves a high
computational load. Therefore, the 2013 training architecture contains a
number of reduction steps. 
The network that is used consists of three hidden layers, which is simpler
than what is used in most supervised learning
tasks. Figure~\ref{fig:dqnarch} shows the architecture of DQN.

The images are high-dimensional data. Since working with the full resolution
of $210\times 160$ with 128 colors at 60 frames per second would be
 computationally too
intensive,  the images are preprocessed. The 
$210\times 160$ with $128$ color palette is reduced to gray scale and
$110\times 84$ pixels cropped to $84\times 84$. The first hidden layer convolves
16 $8\times 8$ filters with stride 4 and ReLU neurons. The second hidden layer
convolves 32 $4\times 4$ filters with stride 2 and ReLU neurons. The third
hidden layer is fully connected and consists of 256 ReLU neurons. The
output layer is also fully connected with one output per action (18
actions for the joystick). The outputs correspond to the  Q-values of
the individual action. The network receives the change in game score
as a number from the emulator, and derivative updates are reprocessed to
$\{-1,0,+1\}$ to indicate 
decrease, no change, or improvement of the score (Huber loss~\cite{openaibaseline2017}).

To reduce
computational demands further, frame skipping was employed. Only one in
every 3--4 frames was used, depending on the game.
For history, the net takes as input the last four resulting frames,
allowing movement to be seen by the net.
As optimizer RMSprop was used.
A variant of $\epsilon$-greedy is used, that starts with an $\epsilon$
of $1.0$ (exploring) going down to $0.1$ (90\% exploiting).

The network architecture consists of convolutional layers and a fully
connected layer. Recall that convolutional nets are  shift-invariant
feature recognizers. 
Fully connected nets allow $n\times m$ mapping. The feature
recognizers together with the $n\times m$ mapping allow mapping of
any feature (shape) to any action.

\subsection{Decorrelating States}\index{decorrelation}

Let us now look in more detail at how stable learning was
achieved. The focus of DQN is on breaking  correlations between states.
%
%
The DQN algorithm has two methods to achieve this: experience replay
and infrequent weight updates.  We will first look at experience replay.




\subsubsection*{Experience Replay}\index{experience replay}
Recall that in reinforcement learning training samples are created in
a sequence of  interactions with an environment, and that subsequent training states are
correlated. Thus the network is trained on too
many samples in a certain area, and other parts of the state
space remain unexplored. Furthermore, through function approximation,
some behavior may be 
forgotten.  When an agent  reaches a new level
in a game that is different from previous levels, the agent may forget how to
play the other level. 

DQN uses  experience replay, with a replay buffer~\cite{lin1992self}, a cache of
previously explored states.\footnote{Originally experience replay  is
  a  biologically inspired
  mechanism~\cite{mcclelland1995there,o2010play,lin1993reinforcement}.} 
The goal is to increase the independence of
training examples, by sampling training examples from this buffer. The
next state to be trained on is no longer a direct successor of the
current state, but one in a long history of previous states. In
this way the replay buffer spreads out the learning over all seen states
by sampling a batch of states at random, breaking temporal
correlations between samples.

Experience replay 
improves on standard 
Q-learning in two respects, using randomization of the order in which samples
are used to (1) break the correlations between consecutive samples
; (2) prevent unwanted feedback 
loops, which would arise when the game makes moves in a certain area
that it does not escape from, getting stuck in local areas of the state space
. By averaging the
behavior distribution over the
previous states, experience replay smooths out training and avoids
oscillations or local minima in the parameters.



Note that training by experience replay is a form of  off-policy
learning, since the  target parameters are 
different from those used to generate the sample.\footnote{But since
  DQN uses $\epsilon$-greedy exploration, in a percentage of the
  actions the behavior
  policy is used. DQN is a mix between on-policy and off-policy
  learning.} Off-policy learning
is one of the three elements of the deadly triad. It is curious to see that this
solution to stable learning is to use more of one of the causes of
the deadly triad.

In practice, experience replay stores the last $N$ examples in
the replay memory, and samples uniformly  when
performing updates. A typical number for $N$ is $10^6$ \cite{zhang2017deeper}. A form of importance sampling might differentiate
important transitions.
Experience replay  works well in practice in
Atari~\cite{mnih2015human}. However,   further  analysis of replay
buffers has pointed to possible problems.
Zhang et
al.~\cite{zhang2017deeper} study the deadly triad  with experience
replay, and find that larger networks 
resulted in more instabilities, but also that longer multi-step
returns  yielded fewer unrealistically high reward values. Indeed, in
Sect.~\ref{sec:rainbow} we will see many further enhancements.



\subsubsection*{Infrequent Weight Updates}
The second improvement in DQN is {\em infrequent weight updates.}
Infrequent updates of the target weight values also reduce correlations
and oscillations caused by loops and self-reinforcing features. This
method works by using a 
separate network for generating the targets of the update of the $Q$
value. Every $x$ updates, the network $Q$ is cloned to
obtain a target network $\hat Q$, which is used for generating the
targets for the following $x$ updates to $Q$.

This second network improves the stability of Q learning,
where normally an update that increases $Q(s_t,a_t)$ often 
increases $Q(s_{t+1},a)$ for all $a$. This also increases the
target, quite possibly
leading to oscillations and divergence of the policy. Generating the
targets using an older set of parameters adds a delay between the time
an update to $Q$ is made and the time the update changes the targets,
breaking the feedback loop or at least making oscillations less likely.


\subsection{Conclusion}
It should be noted that the fundamental reasons for instability in
reinforcement learning with function approximation still exist; the
deadly triad has not evaporated into thin air, and states are still
generated in an agent/environment loop.  DQN and related
algorithms have achieved stable learning for some games.
Van Hasselt et al.~\cite{van2018deep} mention that
divergence can be mitigated by using separate networks for
bootstrapping, using multi-step returns, and prioritizing replay to
reduce the overestimation bias. Many further advances to DQN have been
found, as we will discuss further in
Sect.~\ref{sec:rainbow}.

The topic of this chapter is how deep learning enabled automated
feature discovery to achieve true end-to-end machine learning, from
pixels to labels and actions. We have seen how a combination of new
algorithms, large training sets, and GPU compute power has enabled
researchers to achieve breakthroughs in image recognition and game
playing. These breakthroughs  have resulted in applications that we use
in our daily lives, and  have inspired much further research
(Sect.~\ref{sec:rainbow}). 

Let us take a step  back and reflect on what has been achieved in
supervised and reinforcement learning. 

\subsubsection*{End-to-End Behavior  Training}
Both the works on Atari and ImageNet succeed in end-to-end training. It
is interesting to compare end-to-end training in supervised and in
reinforcement learning. The  breakthroughs in image recognition
 with AlexNet~\cite{krizhevsky2012imagenet}
and other ILSVRC works rely on  training
on large labeled training sets. In these and later works, networks
are trained directly from the raw inputs, using  updates by stochastic
gradient descent. The layered representations that are learned in this way are
typically better representations than hand-crafted or heuristic features.

Mnih et al.\ learned an end-to-end  behavior policy from raw visual
input, also without any heuristics or intermediate features. The Atari
2600 games are games of skill. The games are 
difficult for humans, because they require dexterity and quick
hand-eye coordination. Playing the games builds reflexes. The games
are not 
so much designed as games of strategy (such as Chess, Checkers, Go,
Backgammon, or Othello) where credit assignment is
long and small differences between states can have large consequences,
and where planning algorithms shine.
The Atari 2600 games  are reflex games (thinking fast, not
 slow) with mostly short temporal credit assignment, that could be learned with a surprisingly simple neural
 network. In this sense, the problem of playing Atari well is not
 unlike an image database categorization problem: both problems are to
 find the right answer (out of a small set) that matches an input
 consisting of a set of pixels. Apparently, mapping pixels to
 categories is not that different from mapping pixels to joystick
 actions, when the  credit assignment problem has been
 overcome (see also the observations in~\cite{karpathy2016pong}). 

The algorithmic feat of breaking through the
divergence and oscillation of reinforcement learning with feedback
loops and off-policy function approximation, is highly
imaginative, and has stimulated much subsequent research. Many blogs
have been written on replicating the result, which is still not an easy
task~\cite{openaibaseline2017}.

We will now turn to further developments of advanced deep
reinforcement learning algorithms to address these challenges.

\section{*Deep Reinforcement Learning Enhancements}\index{convergence}\label{sec:rainbow}
As we have seen in Sect.~\ref{sec:stable}, the stability of the training
process is the main challenge in deep reinforcement learning.   DQN
showed that experience replay improves the 
training stability, enabling end-to-end learning in Atari. The Atari
results have spawned much activity among reinforcement learning
researchers to improve training stability further, and many
refinements of experience replay have 
been devised.  The algorithms in this section all aim to address
the problems of coverage, correlation, and convergence, and largely
succeed in doing so.\index{coverage, correlation, convergence}

Many of the topics that are covered by the enhancements are older
ideas that work well in deep reinforcement learning. DQN applies random sampling of its replay buffer, and
one of the first enhancements was prioritized
sampling~\cite{schaul2015prioritized}. It was found that DQN, being an
off-policy algorithm, typically overestimates action values (due to
the max operation). Double DQN addresses
overestimation~\cite{van2016deep}, and dueling DDQN introduces the
advantage function to standardize action
values~\cite{wang2015dueling}. Two  effective actor-critic
algorithms were introduced: A3C~\cite{mnih2016asynchronous} and
PPO~\cite{schulman2017proximal}. The effect of random noise on
exploration was tested~\cite{fortunato2017noisy}, and distributional
DQN showed that networks that use probability distributions work
better than networks that only use expected values~\cite{bellemare2017distributional}.

\begin{table}[t]
  \begin{center}

\begin{tabular}{lccc}
   {\bf Name} & {\bf Principle} & {\bf Applicability} & {\bf Effectiveness}  \\ \hline\hline
  double DQN & de-overestimate values & DQN  & convergence  \\ 
  prioritized experience  & decorrelation & replay buffer  & convergence  \\ 
  distributional  & probability distr & deep RL  & generalization  \\  
  random noise  & parametric noise & deep RL  & more exploration \\ 
  actor-critic  & value/policy interleave & A3C, PPO, dueling  & stability, parallel    
  \\ \hline
\end{tabular}

  \caption{Deep reinforcement learning enhancements}\label{tab:rainbow}
  \end{center}
\end{table}\index{Rainbow paper}
In this section we will discuss  these
enhancements.  Table~\ref{tab:rainbow}
gives an overview of some of the  deep reinforcement learning enhancements. The
enhancements improve generalization to the test set and convergence of the
training process.
Just as we saw in Chap.~\ref{chap:search} with heuristic planning,  the basic concepts and
algorithms in deep reinforcement learning are simple. And just as in
heuristic planning, the basic ideas offer much room for further
enhancement, to achieve better performance~\cite{schaeffer2001unifying}.
%
Interestingly, many of these enhancements are
independent: the concepts do not interfere and can be used together.

\begin{figure}[t]
\begin{center}
\includegraphics[width=9cm]{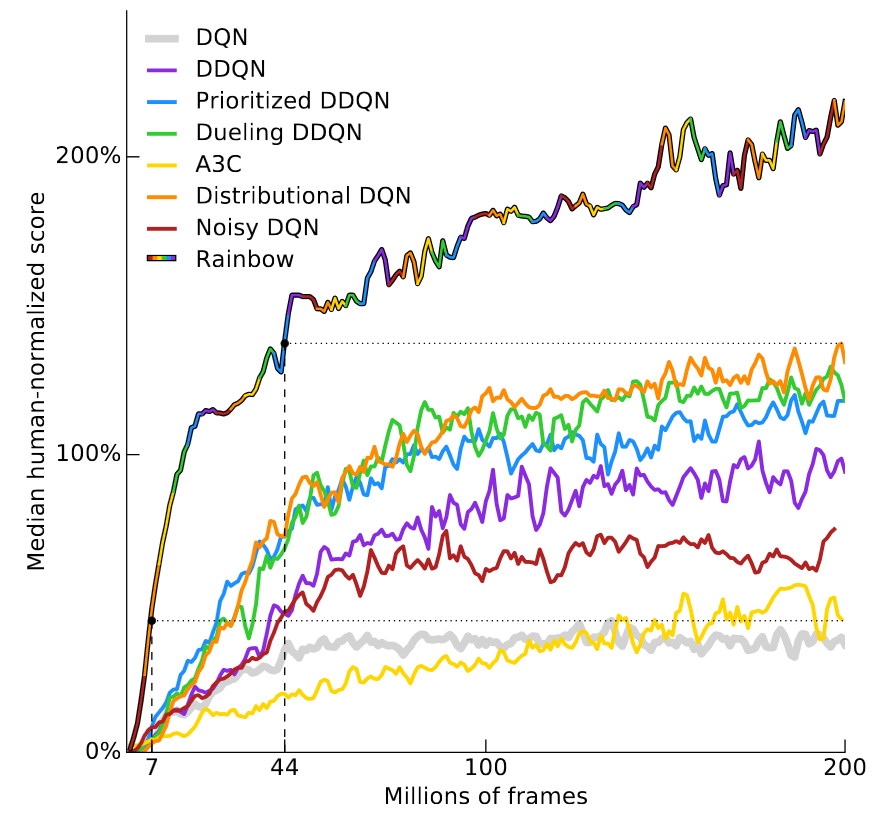}
\caption{Rainbow graph: performance over 57 Atari games~\cite{hessel2017rainbow}}\label{fig:rainbow}
\end{center}
\end{figure}

In 2017 Hessel et al. published the Rainbow
paper~\cite{hessel2017rainbow}, in which they combined seven  important enhancements. The paper is so called because the major
graph showing the cumulative performance over 57 Atari games of the
seven enhancements is 
multi-colored (Fig.~\ref{fig:rainbow}).

The Rainbow paper  summarizes
some of
them, and this section provides an overview of ideas and
enhancements.  The enhancements all
use the same or similar benchmarks (ALE or Gym), and most algorithm
implementations can be found on the OpenAI Gym \href{https://github.com/openai/baselines}{GitHub} site in the
baselines.\footnote{\url{https://github.com/openai/baselines}} This has stimulated research, resulting in the current
abundance of ideas and algorithms, of which many play an important
role in further research.

\subsubsection*{Reducing Correlation}
The baseline algorithm of the methods described 
is DQN.
The original DQN architecture is based on a simple neural network with
two hidden convolutional layers and one
    fully connected hidden layer,  a fully connected output
    layer,\footnote{The 2015 architecture uses three hidden convolutional
      layers~\cite{mnih2015human}} and $\epsilon$-greedy
    exploration. DQN  further uses a combination
of four techniques~\cite{mnih2013playing, mnih2015human}: an experience replay buffer for randomized sampling to break consecutive
 state correlations,  a separate target weight network to break target
 value correlations, clipping rewards to $\{-1, 0, +1\}$, and  skipping frames and reducing pixel resolution to reduce the computational load.

The following algorithms improve on various aspects of DQN.

\subsection{Overestimation}
\index{DDQN}
Van Hasselt et al.\ introduce double deep Q learning
(DDQN)~\cite{van2016deep}. DDQN is based on the observation that
Q-learning may overestimate action
values. They find that,
in practice, such overestimations are common. On Atari 2600 games, due to the use of a deep neural
network, DQN suffers from substantial
over-estimations. Earlier Hasselt et al.~\cite{hasselt2010double} introduced the
double Q learning algorithm in a tabular setting. The later paper
shows that this idea also works with a large deep network. This DDQN algorithm
not only reduces the 
overestimations but  also leads to much
better performance on several games.

DDQN was tested on 49 Atari games and achieved about twice the
average score of DQN with the same hyperparameters, and four times the
average DQN score with tuned
hyperparameters~\cite{van2016deep}.

\subsection{Prioritized Experience Replay}
Prioritized experience replay, or PEX,\index{prioritized DDQN} was
introduced by Schaul et al.~\cite{schaul2015prioritized}. In the
Rainbow paper its results are shown in combination with DDQN.

Recall that in DQN experience replay lets agents reuse examples
from the past. In DQN experience transitions are uniformly sampled
from a replay memory. Therefore, actions are simply replayed at the
same frequency that they were originally experienced, regardless of
their significance. Schaul et al.\ develop a framework for prioritizing
experience. Important actions are replayed more frequently, and therefore learning
efficiency is improved.

Schaul et al.\ use standard proportional prioritized replay, where the
absolute TD error is used to prioritize actions. This can be
computed in the distributional setting, using the mean action
values. In the Rainbow paper all distributional
variants prioritize actions by the Kullback-Leibler loss~\cite{hessel2017rainbow}.

\subsection{Advantage Function}\index{advantage function}
Vanilla DQN uses a single neural network as function
approximator. Dueling DDQN\index{dueling DDQN}~\cite{wang2015dueling}
improves on this architecture by using two separate estimators: a value function
and an advantage function. An advantage function computes the
difference between the value of an action and the value of the
state. In this way it standardizes on a baseline for the actions of a
state~\cite{grondman2012survey}. 
This allows better learning across actions.
Results show that the
advantage function leads to better policy evaluation when there are  many
similar valued actions.

\subsection{Distributional DQN}\label{sec:distrdqn}
DQN learns a single value, the estimated mean of the state value.\index{distributional DQN}
Distributional Q-learning~\cite{bellemare2017distributional} learns a
categorical probability distribution of discounted returns, instead of estimating
the mean. This is in contrast to most reinforcement learning algorithms that
model only the expectation of this  value. Bellemare et al.\ use the
distributional perspective to design a new algorithm which applies
Bellman's equation to the learning of approximate
distributions. Performance results of distributional DQN on Atari are
good, showing the importance of the distributional perspective. Other
relevant research into distributional perspectives is by Moerland et
al.~\cite{moerland2017efficient,moerland2018potential}.

Distributional reinforcement learning is an interesting and promising
approach. Dabney et al.~\cite{dabney2020distributional} report on
experiments with mice that showed correspondence between distribution
reinforcement learning algorithms and the dopamine levels in mice. It
is hypothesized that the brain represents possible future rewards not
as a single expected value, but as a probability distribution.

\subsection{Noisy DQN}
Noisy DQN\index{noisy DQN}~\cite{fortunato2017noisy} uses stochastic network layers
for exploration. In NoisyNet parametric noise is
added to the weights. This noise induces randomness in the 
agent's policy, increasing exploration. The
parameters of the noise are learned with gradient descent along with
the remaining network weights.  The authors replace the standard
exploration heuristics for A3C, DQN, and 
dueling agents (entropy reward and $\epsilon$-greedy) with
NoisyNet. The better exploration yields substantially higher scores for  Atari.

\subsection{R2D2}
R2D2\index{R2D2}~\cite{kapturowski2018recurrent} stands for Recurrent Replay Distributed DQN, and is built upon
prioritized distributed replay and $5$-step double Q-learning.  It
uses a dueling network architecture and an LSTM layer after the
convolutional stack, see
also~\cite{wang2016dueling,gruslys2017reactor}. The RNN uses the
recurrent state to exploit long-term temporal dependencies. They also
find that RNN allows for better representation learning. R2D2 achieved
good results on the 57 Atari games~\cite{kapturowski2018recurrent}. 

A recent benchmark  achievement has been published as Agent57, the first deep
reinforcement learning program that achieves a score higher than the
human baseline on all 57 Atari 2600 games. Agent57 uses a
meta-controller that adapts long- and short-term behavior of the agent,
training for a range of policies, from very exploitative to very
explorative~\cite{puigdomenech2020agent57}.\index{Agent57}

\subsection{Actor-Critic Approaches}\index{actor-critic}

Q-learning and value-based methods work well for discrete problem
domains such as games. For continuous problem domains, such as
robotics, policy methods are preferred. A challenge in policy methods
is to find a reliable estimator for the quality of actions. A good way to achieve an
accurate sense of the quality of actions is through the interleaving
of value and policy methods, in an actor-critic approach (see also
Sect.~\ref{sec:actor-critic}). There have 
been exciting advances in actor-critic, and some of the best
performing algorithms on both games and robotics, such as PPO, are actor-critic.

Policy gradient methods caused recent breakthroughs in
using deep neural networks for control, from video games, to 3D
movement, to board games. Getting good results with policy gradient
methods is challenging. They are sensitive to the choice of
step size---too small, and progress is  slow; too large and
the signal is overwhelmed by the noise. When they lack good value estimates of the
actions, they can have very poor sample
efficiency, taking millions  of time steps to learn simple
tasks~\cite{schulman2015trust}. 

We will describe two main actor-critic algorithms: A3C and PPO.



\subsubsection*{A3C}\index{actor-critic}\index{A3C}\label{sec:a3c}\index{decorrelation}
A3C stands for  Asynchronous Advantage Actor
Critic~\cite{mnih2016asynchronous} and is a  
parallel implementation of Advantage Actor-Critic (A2C), an
actor-critic algorithm that uses the advantage
function~\cite{sutton2018introduction,grondman2012survey}. 

In reinforcement learning, the amount of computational effort required
for training is often substantial, and can be a prohibiting factor in
achieving good results. Although
the speedup provided by GPUs enabled much progress, the training times
are still large. One way  to speed up algorithms is
to run parts of the search effort on multiple computers in parallel. Mnih et
al.~\cite{mnih2016asynchronous} present a method for parallel asynchronous gradient
descent. The method allows implementation of different
reinforcement learning variants, and related algorithms. One of these, an actor-critic
algorithm called A3C, performs  well. 

A3C uses
a multi-step bootstrap
target~\cite{sutton2018introduction,sutton1988learning}.
One of the effects of actor-critic is that
it stabilizes training, just as experience replay.
The parallel actor learners  stabilize the
controllers.  The parallelism also decorrelates the samples into a
more stationary process, since the parallel agents experience a
variety of unrelated states.
The Rainbow experiments include A3C and show its performance against
advanced DQN algorithms, and in combination.

\subsubsection*{Proximal Policy Optimization}\index{Proximal policy optimzation}
Another actor-critic algorithm is proximal policy optimization, or
PPO~\cite{schulman2017proximal}. PPO is an important algorithm that is successful
in discrete action spaces (games) and continuous action spaces
(robotics). PPO was not part of the Rainbow paper, which focuses mostly on 
value-based methods.  

As we have seen~\cite{duan2016benchmarking} policy gradient estimates
can have high variance and they can be sensitive to the settings of their
hyperparameters. After the gradient is calculated, choosing the
step size can be difficult: too small, and the algorithm is slow and
sample inefficient; too large, and the algorithm overshoots and does
not find the optimum. Several approaches have been proposed to make policy
gradient algorithms more robust. One is to constrain the updates to a
region, the so-called trust region. The trust region restricts
the amount by which any update is allowed to change the
policy. This approach is used in trust region
policy optimization (TRPO~\cite{schulman2015trust}). In every
iteration  TRPO collects a  batch of data and optimizes the 
surrogate loss subject to a constraint on
how much the policy is allowed to change, expressed in terms of the
Kullback-Leibler divergence. TRPO performs well, but uses expensive  second-order
derivative calculations, and a more efficient approach was sought.

Proximal policy optimization~\cite{schulman2017proximal}   is motivated by the same question as TRPO: how can we take the
largest possible improvement step on a policy using the data we
currently have, without stepping so far that we accidentally cause
performance collapse. Where TRPO tries to solve this problem with a
complex second-order method, PPO is a family of first-order methods
to keep new policies close to the old. PPO
methods are  simpler to implement than TRPO, while performing as well
in practice~\cite{heess2017emergence}.
PPO trains a stochastic policy  on-policy, in contrast to DQN, which
is off-policy. This means that it
explores by sampling actions according to the latest version of its
stochastic policy.

While simpler than TRPO, the
implementation of PPO is still complicated. OpenAI provides an   implementation in its  \href{https://openai.com/blog/openai-baselines-ppo/\#ppo}{baseline}.\footnote{\url{https://openai.com/blog/openai-baselines-ppo/\#ppo}}

\subsection{Enhancements in the Context of Game Playing Programs}\label{sec:history}
To close this section on reinforcement learning enhancements, and to
close this long chapter on deep learning, let us take a step back and
look from a historical perspective at learning in games. We will look at how the various
 methods have been used in the context of 
game playing programs. We will discuss  examples of database, supervised,
and reinforcement learning. Some of the programs that we discuss will
apply a form of self-play, a topic that will de explored more deeply
in the next chapter.

Note that the simple minimax procedure already performs a kind of
self-play.
In minimax we determine our
best move by calculating our opponent's best reply, which
is based on our best reply, etc. In TD-Gammon and AlphaGo Zero style self-play we use a search-eval
player to train our own
evaluation function, by playing a small tournament against ourselves.

However, it can be difficult to get self-learning to work, because of
instability in deep reinforcement learning training algorithms. When a supervised 
learning system does not learn, then we can try to find out why it is
not learning certain examples by checking them one by one. A self-learning
reinforcement learning system produces its own training examples, it
can be unstable, and it can be
hard to find out why  a system is not learning, when there is no
database of known knowledge that the system should have learned.

Therefore,  we will see
many examples of database learning, before we see self-play appear in game playing programs.

\subsubsection*{Earlier Works}\label{sec:early}
Already in 1959 Samuel~\cite{samuel1959some} introduced the idea of
learning by self-play in his  Checkers program. The coefficients of
the heuristic function were updated towards the 
value of the search after black and 
white had each played one move. Today, more and more programs are based on end-to-end learning.

We will now have a look at previous game playing programs that used
supervised and reinforcement learning.
Table~\ref{tab:programs} lists   
notable works. The
first column contains the name of the program, if given in the
paper. The second column lists which game(s) the program plays. The third column is
the planning algorithm, if applicable. The fourth column is the
training algorithm, if applicable. CNN stands for convolutional neural
net, and FCN for fully connected net. The fifth column describes the data
with which the learning took place. For supervised learning this is
typically a database of grandmaster games, and for reinforcement
learning a self-play loop is used, or databases with self-played games. The sixth column is the network
architecture. Only the hidden layers are counted. For the older papers
this is often a small number of hidden layers. After the
advent of the use of GPUs in 2012 larger nets became feasible. The seventh column
lists the evaluation method. This can be a linear combination of
heuristic features, a system where the network learns the coefficients
of heuristic features, or a full end-to-end learning system from  raw board
input. Last is a reference to a major work describing the approach of
the program. Only the name of the first author is mentioned. Some programs
were developed over many years by different authors. Please refer to
the original 
papers for the often fascinating histories.

\begin{table}[t]
\begin{center}
    {\footnotesize  \begin{tabular}{lccccccc}
    
      {\bf Program} & {\bf Game} & {\bf Plan} &{\bf Learn} & {\bf Data} & {\bf Net} & {\bf Eval} & {\bf Reference}
      \\ \hline\hline
     Samuel & Check. & AlphaB. & coeff. & self-play  & - & coeff. feat. &
    Samuel 1959 \cite{samuel1959some} \\ 
      Chinook & Check. & AlphaB. & - & endgames  & - & heur. feat. &
    Schaeffer 1992 \cite{schaeffer1992world} \\ 
      Logistello & Othello & AlphaB. & regress. &  patterns &-&coeff. pat.&    Buro 1995 \cite{buro1995logistello} \\ 
      Deep Blue & Chess & AlphaB. & coeff. & manual  & - & coeff. feat. &
    Hsu 1990 \cite{hsu1990grandmaster} \\ 
      NeuroCh. & Chess & AlphaB. & TD(0) & db/self-play  & 1 layer & coeff. feat. & Thrun 1995 \cite{thrun1995learning} \\ 
      Neurogam. & B.gam.& - &FCN & games db & 1 layer  & coeff. feat. & Tesauro 1989 \cite{tesauro1989neurogammon}      \\ 
      TD-gam. & B.gam. & AlphaB.&TD$(\lambda)$  & self-play & 1 layer & end to end & Tesauro 1995 \cite{tesauro1995td}  \\ 
      - & Go&-&TD(0) & self-play  & 2 layer & end to end & Schraudolph \cite{schraudolph1994temporal}  \\
     MoGo & Go & MCTS&TD(0)  & self-play  & - & coeff/shape & Gelly 2008 \cite{gelly2006modification}  \\
      NeuroGo & Go&-&FCN & games db  & 1 layer & coeff. feat. & Enzenberger \cite{enzenberger1996integration}  \\ 
      Blondie24 &Check.&AlphaB. &FCN & evo  play & 3 layer & end to end & Fogel 2001 \cite{fogel2001blondie24} \\ 
     Giraffe    &Chess & AlphaB. & CNN & self-play     & 3 layer  & end to end & Lai 2015 \cite{lai2015giraffe}    \\
     DeepChess & Chess&AlphaB. &FCN & games db  & 4 layer  & end to end  & David 2016 \cite{david2016deepchess}   \\ 
     - &          Atari &-& CNN  & DQN  & 3 layer & end to end & Mnih 2013 \cite{mnih2013playing}    \\ 
    ExIt &       Hex &MCTS& CNN & self-play & 15 layer & end to end  & Anthony 2017 \cite{anthony2017thinking}    \\ 
     - & Go & MCTS&CNN & games db & 5 layer & end to end & Clark 2015 \cite{clark2015training}
    \\ 
     DarkForest & Go &MCTS& CNN & games db & 12 layer & end to end & Tian 2015 \cite{tian2015better}
    \\ 
    AlphaGo & Go & MCTS&CNN  & games db & $\pi$:12C/1F& end to end & Silver 2016 \cite{silver2016mastering} \\
     &  & &  &  self-play &  $v$:13C/1F&  & 
    \\ 
    AlphaGo Z. & Go & MCTS & ResNet & self-play & 20 block   & end to end & Silver 2017 \cite{silver2017mastering}\\
         &  &  &  & &  heads:$\pi,v$ &  &
    \\ 
     & Go &  & & & 20 block  & & \\
    AlphaZero & Chess & MCTS&ResNet & self-play& heads:$\pi,v$& end to end& Silver 2018 \cite{silver2018general}  \\
     & Shogi & & & & & & 
    \\ \hline
  \end{tabular}
}

\caption[Learning in game playing programs]{Evolution of planning, learning, and self-play in game
playing programs}\label{tab:programs} 
  \end{center}
\end{table}

We will now discuss how the methods of these earlier programs
developed.

\subsubsection*{Learning Shapes and Patterns}
Game playing programs have been using many types of machine learning
to improve the quality of play. 
For this
reason, many early works focused on learning shapes and patterns. This
learning was typically off-line, using months and months of training
time. 
The Checkers program Chinook employed an
end-game database that contained all board positions
with 8 or fewer pieces. As soon as the database was reached by the
search, perfect knowledge was available. The end-game database greatly
contributed to the strength of
Chinook~\index{Chinook}\cite{schaeffer1996chinook}. In Chess, Deep
Blue had a database that was accessed as soon as 5 or fewer pieces
were on the board~\cite{campbell2002deep}.

In Go, the emphasis of databases is on small local patterns. Stoutamire used hashed sets of patterns  to
find good and bad shapes~\cite{stoutamire1991machine}. Van
der Werf used neural networks and automatic feature extraction to
predict good local moves from game
records~\cite{van2002local,van2004ai}. 

\subsubsection*{Supervised Learning}\index{supervised learning}
Many researchers have tried supervised learning with databases of
grandmaster games.
Tesauro's 1989 Neurogammon\index{Neurogammon} achieved good results in
Backgammon with supervised learning of the coefficients of an evaluation
function of handcrafted features. He used a single hidden
layer~\cite{tesauro1990neurogammon}.

In Go,  Enderton used
neural networks in the search in his  Go program Golem, for move
ordering and for forward pruning of 
selective search~\cite{enderton1991golem}. Enzenberger's
NeuroGo\index{NeuroGo} used a neural network for 
position evaluation~\cite{enzenberger1996integration}.

Much later, Maddison et al.\ published work on a 12-layer convolutional
neural net for move prediction in Go~\cite{maddison2014move}, laying
a foundation for the supervised learning of AlphaGo.
Facebook's AI Research lab created
Darkforest\index{darkforest} for end-to-end deep learning of Go
positions and MCTS~\cite{tian2015better}.

\subsubsection*{Reinforcement Learning}\index{temporal difference}

Supervised learning, as in Neurogammon, uses a database to learn  from good moves.
After the success with supervised learning in Neurogammon, Tesauro
tried temporal difference learning from the raw board in a small neural network in 1992. This
was further augmented with expert designed features, resulting in
play at  world class level. TD-Gammon's success was the first widely
published  success of
self-play~\cite{tesauro1995temporal,tesauro2002programming}.\index{TD-Gammon} 

In Go Schraudolph et al.~\cite{schraudolph1994temporal} used temporal difference to
learn position evaluation from board networks. They note that to be
successful, networks should reflect the spatial organization of the input signals.
Huang et al.~\cite{huang2003learning} used self-play of 16 different players to learn
opening strategies in Go.

Silver, in his PhD thesis~\cite{silver2009reinforcement}, applied reinforcement
learning to search in Go, and Silver et
al.~\cite{silver2007reinforcement} and
Gelly and Silver~\cite{gelly2008achieving,gelly2007combining,gelly2012grand}
studied reinforcement learning of local shape patterns.
Silver et al.~\cite{silver2012temporal} further combined temporal difference learning
in MCTS.

In Chess,
Thrun's NeuroChess\index{NeuroChess}~\cite{thrun1995learning} tried
temporal difference feature learning.
Inspired by the progress in deep learning, in
2015 Matthew Lai tried end-to-end learning in Chess: he went beyond
evaluation function weight 
tuning with Giraffe~\cite{lai2015giraffe},\index{Giraffe} using DQN, 
and self-play, including feature extraction and pattern recognition
for end-to-end reinforcement learning. Giraffe reached master-level
play. Lai later joined the AlphaZero team.

In some of the early works the small size of the networks limited
success. In others the temporal difference  learning or self-play is
based on hand-crafted or
heuristic features of the state space, possibly introducing  bias
or errors. End-to-end 
learning from the raw board or pixels requires more computational
power. When first done successfully end-to-end
(AlexNet~\cite{krizhevsky2012imagenet},
Atari~\cite{mnih2013playing,mnih2015human}) the papers generated much
research interest.

Concurrent to the work by the AlphaGo
team, Anthony et al.~\cite{anthony2017thinking}\index{ExIt} worked on a similar
idea for tabula rasa learning 
based on self-play, combining MCTS and DQN. The name of their work is
ExIt, for Expert Iteration. They did not use Go, but
Hex, a simpler game, to show their methods. They do make the link with
Kahneman's work explicit, even in the title of their paper.

The relation between planning and training is also studied in
model-based and model-free reinforcement learning (Sect.~\ref{sec:modelbased}). The interplay between planning and training,
and especially of model-based and model-free approaches, is an active area of
research~\cite{anthony2017thinking,racaniere2017imagination,pascanu2017learning,wahlstrom2015pixels,levine2013guided,watter2015embed}.

\subsection{Conclusion}


Progress has been made in deep reinforcement learning by many
researchers in a short period of time. The actor-critic algorithms and
the Rainbow paper algorithms illustrate
how much our understanding of deep reinforcement learning has
improved. As in heuristic planning and in supervised 
learning, the presence of clear benchmarks was instrumental for
progress. Researchers were able to see clearly which ideas worked, and to check
if their intuition and understanding were correct. The earlier game
playing programs, OpenAI's Gym~\cite{brockman2016openai}, and
the ALE~\cite{bellemare2013arcade} are responsible to a great extent
for enabling this progress.

The next chapter will look deeper into self-play, a kind of
reinforcement learning in which the environment is played also by the
agent, and which takes performance of deep reinforcement learning to
another level.

\section{Practice}\index{Keras}\label{sec:keras}

Below are some questions to check your understanding of this
chapter. Each question is a closed question where a simple, one
sentence answer is possible.
\subsubsection*{Questions}
\begin{enumerate}
  \item What is the difference between supervised learning and
    reinforcement learning?
  \item What is the difference between a shallow network and a deep
    network?
  \item Why is function approximation important in game playing?
  \item What phases does a learning epoch have? What happens in each phase?
  \item What is the difference between heuristic features and
    learned features?
  \item What is MNIST? What is ImageNet? What is TensorFlow?
  \item Name three factors that were essential for the deep learning
    breakthrough, and why.
  \item What is end-to-end learning? Do you know an alternative?
    What are the advantages of each?
  \item What is underfitting, and what causes it? What is overfitting, and what causes it? How can you see
    if you have overfitting?
  \item Name three ways to prevent overfitting.
  \item What is the difference between a fully connected network
    and a convolutional neural network?
  \item What is max pooling?
  \item Why are shared weights advantageous?
  \item Describe the vanishing gradient problem. What are ReLU units?

\item Why is deep reinforcement learning more susceptible to unstable
  learning than deep supervised learning?
\item What is the deadly triad? 
\item Why may reinforcement learning with function approximation be
  unstable?
\item What is the role of the replay buffer?
\item How can correlation between states lead to bad generalization?

\item How did AlexNet advance the state of the art in deep learning?
\item Name three architectural innovations/characteristics of AlexNet.
\item What is an inception modules? What is a residual network? What
  is the relation between the two?
\item What is a GAN?
\item What is an RNN, and how is it different from a deeply layered
  network?
\item What are the advantages of LSTM?
\item Why is the Rainbow paper so named, and what is the main
  message?
  \item Why is PPO an actor-critic method?
\end{enumerate}

\subsubsection*{Exercises}
Let us now start with some exercises. 
Install TensorFlow and Keras (go to the TensorFlow page).\footnote{\url{https://www.tensorflow.org}} 
The exercises
below are meant to be done in Keras.


\begin{enumerate}
\item Download and install TensorFlow and Keras. Check if everything is
  working by executing the MNIST training example. Choose the default training
  dataset. Run it on the
  training set, and then test the trained network on a testset. How
  well does it predict?
\item  Which optimizer does the MNIST example use? Try a
  different optimizer,  an adaptive optimizer such as Adam, and compare
  it with simple Backprop. Does training speed improve? 
\item \href{https://www.openml.org}{OpenML} is a research project to compare
  different machine learning algorithms with different datasets. Go to
  OpenML and download different datasets to see which problems may be
  harder to learn.\footnote{\url{https://www.openml.org}}
\item The learning rate is a crucial hyperparameter. Always tune this
  one first. You may find that learning goes slowly. Try to increase the
  training speed by increasing the different learning rates. Does
  performance increase? Does the error function drop faster or slower?
  Can you explain your observations?
\item Explore options for cloud processing. The cloud provides ample
  compute power at cheap prices, often with free student or sign up bonuses. Go
  to AWS, the Google Cloud, Colab, or Microsoft Azure, and explore ways to
  speed up your learning.
  \item One of the biggest problems of supervised learning is
    overfitting. Can you check if over- or underfitting occurred in
    your experiments? Create an experiment in which you force overfitting
    to occur. What will you do: change the size of the network, change
    the algorithm (change CNN, 
    ReLU, dropout), or change a parameter (learning rate, dropout
    percentage, training time)? How can you see if overfitting has occurred? When has
    it been reduced?
\end{enumerate}

\subsubsection*{Atari DQN Code}
The Atari papers come with source code. The original DQN code
from~\cite{mnih2015human} is  at 
\href{https://github.com/kuz/DeepMind-Atari-Deep-Q-Learner}{Atari
  DQN}.\footnote{\url{https://github.com/kuz/DeepMind-Atari-Deep-Q-Learner}}
  This code is the original code, in the programming language Lua, without new
developments. Before you start working with this code, it should be
  noted that DQN can be difficult to get working. There are many 
details to get right. In addition, the
experiments often have to run for a long time, which means that
debugging and improving is slow if you have just one GPU. 

A better start is the TensorFlow implementation in Python of DQN for the Atari game
Breakout, at \href{https://github.com/floodsung/DQN-Atari-Tensorflow}{Atari
  Breakout Code}.\footnote{\url{https://github.com/floodsung/DQN-Atari-Tensorflow}} This will get you up to speed more quickly.  The code
is readable, and follows the concepts discussed in this
chapter. However, the
code is too large to cover in detail here.

To get really started with deep reinforcement learning, we suggest to
use DQN and many
of the other algorithms that are available at the OpenAI  Keras RL library at
\href{https://github.com/keras-rl/keras-rl}{Keras RL}\footnote{\url{https://github.com/keras-rl/keras-rl}}
\cite{plappert2016kerasrl,geron2017hands}.

\subsubsection*{Cartpole}
OpenAI provides baseline implementations of important reinforcement
learning algorithms~\cite{baselines}. A good implementation of DQN for the Atari
Learning Environment is available. Installation instructions for
Linux and macOS are provided.

\begin{figure}[t]
\begin{center}
\includegraphics[width=6cm]{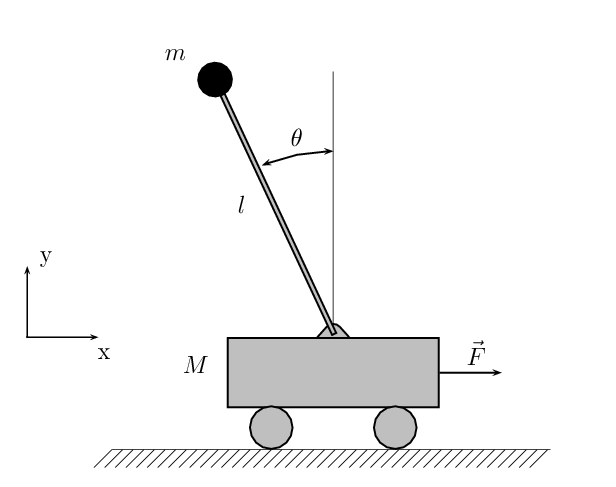}
\caption{Cartpole experiment~\cite{sutton2018introduction}}\label{fig:cartpole}
\end{center}
\end{figure}

Cartpole is a basic experiment in which an agent must learn to balance
a stick on a cart by reinforcement learning. A nice video of a
 Cartpole experiment is 
\href{https://www.youtube.com/watch?v=5Q14EjnOJZc}{Cartpole
  Video}\footnote{\url{https://www.youtube.com/watch?v=5Q14EjnOJZc}}
(this is a real-life cartpole, as opposed to the simulated cartpole
that is most often used in 
reinforcement learning experiments; Fig.~\ref{fig:cartpole}).
At \href{https://github.com/openai/baselines/tree/master/baselines/deepq}{OpenAI
  Cartpole
  baseline}\footnote{\url{https://github.com/openai/baselines/tree/master/baselines/deepq}}
  working example code for the Cartpole experiment 
is available.

The high-level code is simple, as Listing~\ref{lst:cartpole} shows. To really
understand what is going on you have to look into the implementations of
\verb|deepq.learn|, which  implements DQN in a few hundred lines of code.

\lstset{label={lst:cartpole}}
\lstset{caption={Cartpole code~\cite{baselines}}}
\begin{lstlisting}[language=Python,float]
import gym

from baselines import deepq

def callback(lcl, _glb):
    # stop training if reward exceeds 199
    is_solved = lcl['t'] > 100 and
           sum(lcl['episode_rewards'][-101:-1]) / 100 >= 199
    return is_solved

def main():
    env = gym.make("CartPole-v0")
    act = deepq.learn(
        env,
        network='mlp',
        lr=1e-3,
        total_timesteps=100000,
        buffer_size=50000,
        exploration_fraction=0.1,
        exploration_final_eps=0.02,
        print_freq=10,
        callback=callback
    )
    print("Saving model to cartpole_model.pkl")
    act.save("cartpole_model.pkl")

if __name__ == '__main__':
    main()
\end{lstlisting}




\subsubsection*{More Exercises}
For the following exercises, if you have not done so, download and install Gym.
\begin{enumerate}
  \item Go to the Mountain Car example, and run it. Experiment with
  different learning settings to understand their effect on the learning.
  \item Try the same for Cartpole.
 
\item Install ALE. Train a player for Space Invaders. Did the training
  process work well? How do you find  out if the trained player is a
  good player? Can you compute an Elo rating?

\item Identify the replay buffer in the code. Reduce the size, or
  otherwise turn it off. Does DQN still converge, what does it do to
  the quality of the player that is trained?

\item *Try a few other Atari games. Does DQN train well? Try Montezuma's
  Revenge. Does DQN work well? Have a look at the alternative deep
  reinforcement methods. How could you improve DQN to play well in
  Montezuma's Revenge? (This is a hard research
  question; see also~\cite{badia2020agent57}.)\index{Montezuma's Revenge} 
  
\end{enumerate}




\subsubsection*{Stable Baselines}\index{stable baselines}

The Gym
GitHub repository \href{https://github.com/openai/baselines}{OpenAI Baselines}\footnote{\url{https://github.com/openai/baselines}} contains
implementations of many algorithms covered 
here. You can easily download them and experiment to gain an insight
into their behavior. Stable Baselines is a fork of the OpenAI
algorithms; it has more documentation and other features. It can be
found at \href{https://github.com/hill-a/stable-baselines}{Stable
  Baselines},\footnote{\url{https://github.com/hill-a/stable-baselines}}
and the documentation is at
\href{https://stable-baselines.readthedocs.io/en/master/}{Docs}.\footnote{\url{https://stable-baselines.readthedocs.io/en/master/}}
An implementation of the Rainbow paper in Torch is
available
\href{https://github.com/Kaixhin/Rainbow}{here}.\footnote{\url{https://github.com/Kaixhin/Rainbow}}
The RL Baselines Zoo even provides a collection of pretrained agents,
at \href{https://github.com/araffin/rl-baselines-zoo}{Zoo}.\footnote{\url{https://github.com/araffin/rl-baselines-zoo}}

The open availability of TensorFlow, Keras, ImageNet, and OpenAI Gym/ALE allows
for easy replication and, importantly,
improvement of the works covered in this chapter.
Keras implementations of VGG, ResNet are available in
Keras.applications (see \href{https://keras.io/applications/}{Keras
  Applications}).\footnote{\url{https://keras.io/applications/}}
AlexNet is here:
\href{https://gist.github.com/JBed}{AlexNet},\footnote{\url{https://gist.github.com/JBed}} GoogLeNet is here:
\href{https://gist.github.com/joelouismarino}{GoogLeNet}.\footnote{\url{https://gist.github.com/joelouismarino}}

\subsubsection*{*Advanced Exercises}
Due to the advanced nature of some parts of this chapter, the following exercises
are large, project sized. These topics  and these
exercises are fascinating, but answering these questions may require a
substantial  amount of research and work (potentially thesis size).
\begin{enumerate}
  \item Reproduce the basic  AlexNet, VGG, and ResNet results on
  ImageNet in Keras. Do deeper ResNets give better results? How do deeper
  ResNet impact training time?
\item  Read  GAN papers on how to generate artificial
  images~\cite{goodfellow2014generative,karras2017progressive,tran2017disentangled}. Try
  to generate fake cat and dog images. Can you also use a
  discriminator to separate real images from fakes?

  \item Use an RNN to perform a captioning challenge in ImageNet.
  \item Read the Rainbow paper~\cite{hessel2017rainbow}. Go to the
  OpenAI baselines and download and
  run the code and perform an ablation study, leaving out the
  constituting methods one by one. 
  \end{enumerate}

\section*{Summary}
In this chapter we faced the challenge of finding function
approximators for large, high-dimensional, state spaces. Deep learning
allowed automated feature discovery for end-to-end learning.  The field of function
approximation with neural networks is large, rich, and deep. This
chapter can only give a glimpse of the work that has been
done.  

The goal of function approximation is to find a reliable value
function for large state spaces, so large that test examples
have not been  seen during training. In this way, the function approximators of
this chapter are an alternative to the heuristic feature function of
Chap.~\ref{chap:search} and the averaged random values of
Chap.~\ref{chap:trial}.


Training  deep networks is not easy. Among the problems to be
overcome are efficient training methods, overfitting, and vanishing gradients. Many
solutions have been discussed, from ReLU, batch normalization (to
prevent vanishing gradients) to dropouts (to prevent
overfitting) to convolutional layers (to both reduce the number of
weights and allow efficient translations and rotations) and we have
discussed training methods such as stochastic gradient descent.

Function approximation  blossomed when large labeled datasets
became available {\em and\/} when the necessary computational power in
the form of GPUs arrived. In addition to the datasets, easy-to-use
free software packages such as TensorFlow and Keras have lowered the
barrier of entry for research significantly. These packages have played a large
role in the success of the field.

We also discussed deep reinforcement learning, where correlations
between states cause divergent and unstable learning.

Deep supervised learning constructs complex features from example
databases.  The examples
in these databases are independent, and training converges towards an
optimum. In reinforcement learning states are correlated and the training
process easily diverges. Two techniques,  replay buffer and
infrequent weight updates, have been introduced to
decouple  states enough to allow the training to converge.

Mnih et al.\ state the following in their paper~\cite{mnih2015human}:
``a single architecture can successfully learn control policies in a
range of different environments with only very minimal prior
knowledge, receiving only the pixels and the game score as inputs, and
using the same algorithm, network architecture and hyperparameters on
each game, privy only to the inputs a human player would have.''
This architecture, 
DQN, was demonstrated on Atari games to achieve end-to-end learning
from the raw pixels on a wide
variety of games that are challenging for humans. 


For better feature discovery  new methods such as VGG and ResNet have been introduced, achieving very high levels of
performance on image recognition tasks.
Recurrent neural nets, or LSTMs, are well suited for sequence
processing, such as caption generation.

For
reinforcement learning, 
many methods have been devised to improve generalization and to reduce
divergence, and thus solve problems related to coverage, correlation,
and convergence. The methods typically use two networks to break possible
oscillations between the search and the target values, and they
use a form of replay buffers. Many variants for better
generalization and more stability have been developed since DQN, as the aptly named
Rainbow paper documents. Among the  promising methods are
actor-critic methods, such as PPO.




\section*{Historical and Bibliographical
  Notes}\label{sec:bib:neural}\index{J\"urgen Schmidhuber}

\subsubsection*{Learning in Games}
\index{Neurogammon}
In game playing, the advantages of neural networks for function approximation
were recognized early. Although the best known success of neural
networks in games is undoubtedly AlphaGo, there has also been
interesting work performed with supervised learning in games. Before
we move  deep reinforcement learning and AlphaGo, let us briefly
look at the earlier approaches that have been tried.

Twenty years before AlexNet, in 1990, Tesauro published a paper on
Neurogammon~\cite{tesauro1990neurogammon}, his program for playing
Backgammon. 
Neurogammon used supervised learning to train the coefficients of human-designed heuristic
input features, by using 400 example games played by Tesauro
himself. A later version of Neurogammon, TD-Gammon, used
reinforcement learning to train the same heuristic input
features. TD-Gammon achieved  world-class human tournament level
play~\cite{tesauro1995td,tesauro1995temporal}.
\index{Backgammon}\index{TD-Gammon}\index{Neurogammon}

In 1996 Enzenberger created the NeuroGo~\cite{enzenberger1996integration}
 Go playing program. NeuroGo uses temporal difference learning and backpropagation to
train the coefficients of manually crafted heuristics features. Its
performance did not surpass that of heuristic programs of its time.\index{NeuroGo}
In 1990 there was insufficient computer power for training of large
networks, nor were there large 
databases of example games to achieve strong play.

In 1999 Chellapilla and Fogel~\cite{chellapilla1999evolving,fogel2001blondie24} used a
fully connected network to evolve board features in Checkers, achieving  good results.
Maddison et al.~\cite{maddison2014move} show how convolutional neural
nets can be used in the context of Go to generate moves. 
Sutskever and Nair~\cite{sutskever2008mimicking} used CNNs to learn from
Go expert games.

Convolutional neural nets were also used in Go by Clark and
Storkey~\cite{clark2014teaching,clark2015training}, who had used a CNN
for supervised learning 
from a database of human professional games, showing that it
outperformed GNU Go and scored wins against Fuego.
David et al.~\cite{david2016deepchess} report on end-to-end deep
learning in Chess, achieving results on par with Falcon and Crafty,
two  alpha-beta-based programs.

\subsubsection*{Deep Learning}
Good review papers on deep learning are~\cite{schmidhuber2015deep,lecun2015deep}.
In 1958 Rosenblatt~\cite{rosenblatt1958perceptron}
introduced the perceptron, a simple neural network for pattern
recognition. 


Much work has been done to formulate this backpropagation
algorithm~\cite{werbos1974beyond}. It allowed the training of
multi-layer networks, paving the way for deep learning's success. In
the 
1980s McClelland and Rumelhart published an influential work coining
the terms parallel distributed processing and
connectionism~\cite{rumelhart1987parallel}. However, support vector
machines and other simpler classifiers became popular~\cite{boser1992training,smola1998learning}. They were easier to reason
about and performed better~\cite{rost1993prediction}. Later,
max pooling was introduced, and GPU methods were created to speed up
backpropagation. The performance of neural nets was further plagued
by the vanishing gradient problem, especially in deeper 
networks.

Among many other contributions to both the theory and practice of deep
learning, Schmidhuber
devised a hierarchy of networks that were
pretrained one network at a time, to overcome this 
problem~\cite{schmidhuber1992learning}.\index{J\"urgen Schmidhuber} Other methods were later
designed, such as ReLU threshold functions and batch
renormalization~\cite{ioffe2017batch}.

There are many books on Keras and TensorFlow; see,
e.g.,~\cite{geron2017hands}.

NeuroGo~\cite{enzenberger1996integration} and Neurogammon~\cite{tesauro1990neurogammon} used small nets to learn the coefficients of
heuristic evaluation functions of the state. They were written at a
time when little compute 
power was available, nor had the necessary advances in deep learning
training algorithms occurred, to learn features directly from the state.

Silver et al.~\cite{silver2012temporal} discuss the use of temporal
difference search in Go, comparing MCTS with TD reinforcement learning.
Matthew Lai~\cite{lai2015giraffe} experiments with DQN in Chess.


Experience replay is an important
technique for the success of deep reinforcement
learning~\cite{mnih2013playing}. Related ideas can be traced
to earlier works.
Gradient TD methods have been proven to converge
for evaluating a fixed policy with a non-linear value
approximator~\cite{bhatnagar2009convergent}.  Relaxing the fixed
control policy in  neural fitted Q learning algorithm (NFQ) has been
reported to work  for nonlinear control with
deep networks~\cite{riedmiller2005neural}. NFQ builds on work on
stable function approximation~\cite{gordon1995stable,ernst2005tree} and
experience replay~\cite{lin1992self}, and more recently on
least-squares policy iteration~\cite{lagoudakis2003least}.



Pattern recognition research has created a
wealth of   machine learning algorithms, of which
deep learning is just one. Other well-known and successful approaches
are  principal component 
analysis, decision trees, random forests, support
vector machines, $k$-means clustering, Bayesian networks, Kalman
filters, and hidden Markov models. Each of these methods works well
in different situations. Good reference works on these methods
are~\cite{bishop2006pattern,kelleher2015fundamentals}, or 
more practically oriented 
works such
as~\cite{geron2017hands,muller2016introduction}.\index{PCA}\index{principal
  component analysis}\index{$k$-means}\index{HMM}\index{hidden Markov model}\index{decision
  trees}\index{random forests}\index{SVM}\index{support vector
  machine}\index{Kalman filters}


Pointers to further work  on recurrent networks (RNN and LSTM) can be found
in~\cite{greff2017lstm,graves2005bidirectional}.  Important  work on RNNs has been done by
Sutskever~\cite{sutskever2013training},
  Mikolov~\cite{mikolov2012statistical}, and Graves~\cite{graves2012supervised}.
A more detailed explanation of LSTMs can be found
in~\cite{hochreiter1997long,karpathy2015}.


   


For more theory on deep supervised learning the book by Goodfellow et al.\ is
indispensable~\cite{goodfellow2016deep}.
Goodfellow et al.~\cite{goodfellow2014generative} introduced
generative adversarial networks, which spawned much follow-up
research 
including dreamlike (but
also curiously  realistic) fake
images.\footnote{Perhaps GANs are the answer to the question posed by Philip
  K.\ Dick in one of his science fiction stories: ``Do Androids Dream of Electric
  Sheep?'' \cite{dick1996androids}.} Huang 
et al.~\cite{huang2017adversarial} apply adversarial attacks to
reinforcement learning policies. 


\chapter{Self-Play}\label{chap:plan}\label{chap:selfplay}\label{chap:a0g}\label{chap:transfer}
This chapter is
devoted to AlphaGo-style self-play.
Self-play is an intuitively appealing AI method that has long been  used
by AI researchers in various forms,  as we saw at 
the end of the previous chapter. The 2016 
results  
showed,  many years after TD-Gammon, how self-play can achieve world
champion level play in a highly complex game. 
Self-play, as it is used in AlphaGo, combines MCTS and deep reinforcement
learning, using approaches from both symbolic AI and
connectionist AI.
\index{symbolic AI}\index{connectionist AI}

In this chapter planning and learning come together.
The previous chapter showed us how deep learning can achieve very high
performance in  image classification and Atari game play. 
In Chap.~\ref{chap:trial} we saw how the  UCT formula
provided effective adaptive node selection.
Chapter~\ref{chap:search} taught us the basics of  searching large
state spaces:  the search-eval architecture, and
enhancements.

We will now see how to combine these methods  in  a self-learning reinforcement
learning system that is able to beat the 
strongest human Go players. 
This chapter covers self-play, the AlphaGo results, and future developments.
The main problem that we discuss  is how to
achieve  stable, high performance, self-play,
 in a very challenging state space.

First, we will explain the basic principle of self-play, and the
reasons why stable self-play is   difficult to achieve. We will also
see that self-play  is faster than 
pretraining with human grandmaster games, and why.

Second, we will look in more detail at  the architecture of self-play
programs. The AlphaGo experiments actually consist of a sequence of 
three programs. We will look at the 
designs of all three programs and at the experiments.

Third, we will look at future work. The  AlphaGo 
results inspired much further work. At the end of the chapter we will
look in more detail at directions for future research.

\section*{Core Problems}
\begin{itemize}
\item How can we create training examples that allow training beyond the
  level of play of human  grandmasters?
\item How can we use  function  approximation to improve adaptive sampling?
\end{itemize}

\section*{Core Concepts}
\begin{itemize}
\item Stable self-play: coverage, correlation, and convergence
\item Curriculum learning 
\end{itemize}


\subsubsection*{A Note on Names}

In this chapter we will discuss three related programs, all with
names that confusingly start with Alpha, and all   developed by
related researchers. Each program achieved an AI milestone that was
published in a top journal. The programs  are
AlphaGo~\cite{silver2016mastering}, AlphaGo 
Zero~\cite{silver2017mastering}, and
AlphaZero~\cite{silver2018general}; see Table~\ref{tab:alphago}.\footnote{Work on a fourth program,
  AlphaStar, for StarCraft, is also under
  way~\cite{vinyals2019grandmaster}; see Sect.~\ref{sec:starcraft}. Furthermore, a related program, AlphaFold, is
  successful in protein folding~\cite{nussinov2019computational}.}
We will discuss  in  this
chapter which  methods were used and  which  advances in reinforcement
learning were achieved. These advances are mostly related to
self-play. We will also discuss further developments related to self-play.

The main publications on the three Alpha-$\star$ programs are~\cite{silver2016mastering,silver2017mastering,silver2018general}. 
Appendix~\ref{chap:godetails} contains more technical
details of the three
systems.
%
%
\begin{table}[t]
  \begin{center}
    \begin{tabular}{lcccc}
      {\bf Program} & {\bf Planning}  
      & {\bf Learning} & {\bf Input/output} & {\bf Achievement}\\ \hline\hline
      AlphaGo & MCTS & supervised \& reinforcement  & Go & beat human champions\\ 
      AlphaGo Zero & MCTS& self-play  & Go & learn tabula rasa\\ 
      AlphaZero & MCTS &self-play   &Go, Chess, Shogi &generalize
                                                        three games \\
      \hline
    \end{tabular}
    \caption{AlphaGo, AlphaGo Zero, and AlphaZero}\label{tab:alphago}
\end{center}
\end{table}
%
%

We will now  first describe self-play, as it is used in AlphaGo Zero,
 the strongest Go program on Earth. 


%
%
%
%


\section{Self-Play Architecture}
\index{AlphaGo Zero}\index{tabula rasa}\label{sec:alphagozero}

We have  seen in Sect.~\ref{sec:history} that self-play has a
long history in game playing programs
(Table~\ref{tab:programs}).
As we mentioned in the previous chapter, self-play  is quintessential
artificial intelligence: a system that can teach itself to 
become more
intelligent. By crafting a system in which
intelligence emerges out of  basic interactions, tabula rasa, from a blank slate, self-play comes
very  close to the original inspiration of the field.


\begin{figure}[t]
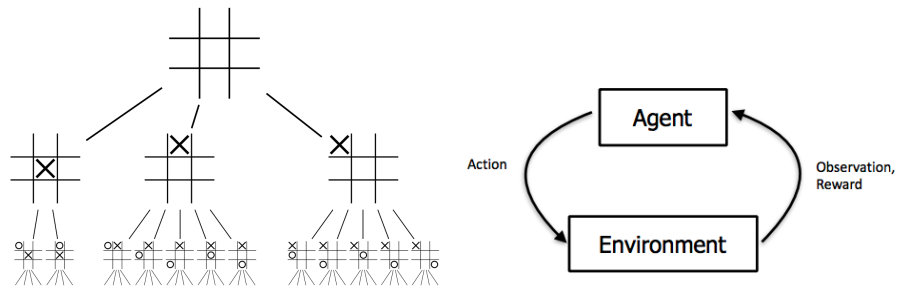

  \begin{center}
    \begin{tabular}{cc}
      \includegraphics[width=6cm]{figures/Tic-tac-toe-game-tree.png}
      &
        \includegraphics[width=6cm]{figures/action-reward}
        \end{tabular}
\caption{Minimax tree and agent/environment loop}\label{fig:minimax7}
\end{center}
\end{figure}

Interestingly, if we go back to the original search-eval
architecture, even simple minimax programs 
perform a kind of self-play, or self-lookahead
(Fig.~\ref{fig:minimax7} left panel): since minimax uses the same algorithm
for player and opponent it plays itself to find out what the best move
is.
MCTS 
also uses  self-play in this way to find the best move.


\subsection{Self-Play Learning}

The kind of self-play that we discuss in this chapter goes a step
further than finding the best move, however, and is used to create an
entire self-{\em learning} program. Let us look at the familiar reinforcement learning loop of
Fig.~\ref{fig:minimax7} (right panel). In this figure the agent learns a good
policy by querying the
environment (which actions are good in which states). In reinforcement
learning, the
environment is our opponent. If we 
perform the same trick as earlier in minimax, and replace the 
environment algorithm by the same agent algorithm then we have a 
self-learning ``{\em self-agent}'' system, a dragon biting  its own tail.

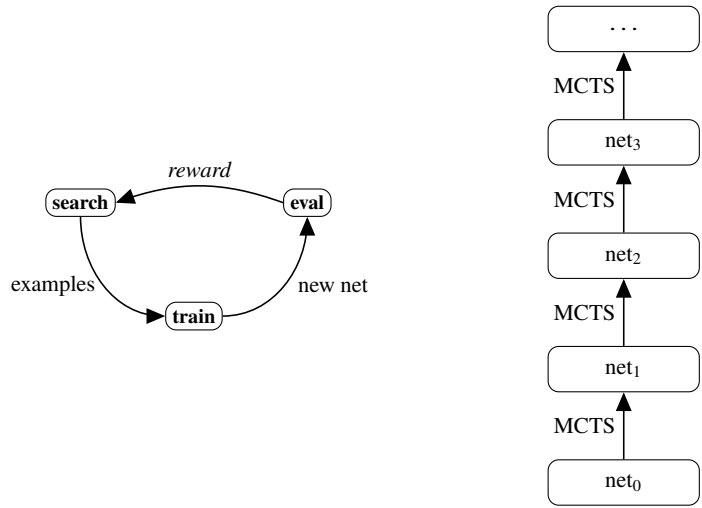
\begin{figure}[t]
  \centering{
    \vspace{2cm}
    \begin{tabular}{lr}
      \begin{minipage}{5cm}
        \vspace{-2cm}    
        \begin{tikzpicture}[>=triangle 45,
  desc/.style={
                scale=0.9,
		rectangle,
		rounded corners,
		draw=black,
		}]
  \node [desc] (search) at    (0,2) {{\bf search}};
  \node [desc] (eval) at    (3,2) { {\bf eval}};
  \node [desc] (train) at       (1.5,0.5) {{\bf train}};

  \draw (search.south) edge[semithick,->,out=270,in=180,looseness=1] node[left] {\small examples}  (train.west);
  
  \path (eval.south) edge[loop,semithick,<-,in=0,out=270,looseness=1]
  node[right] {\small  \ new  net}  (train.east);
  \path[desc] (search.east) edge[semithick,loop above,<-,min distance=7mm,in=160,out=20,looseness=1] node[above] {\small \em reward}  (eval.west);

\end{tikzpicture}
        \hspace{2cm} \ \ \ \ 
        \end{minipage}
      & \begin{minipage}{5cm}
        \vspace{-2cm}
        \hspace{2cm}
        \begin{tikzpicture}[>=triangle 45,
  desc/.style={
		scale=1.0,
		rectangle,
		rounded corners,
		draw=black, 
		}]

  \node [desc,minimum width=2cm,minimum height=0.6cm] (n0) at
  (0,1.5) {$\mbox{net}_0$};
  \node [desc,minimum width=2cm,minimum height=0.6cm] (n1) at
  (0,3) {$\mbox{net}_1$};
  \node [desc,minimum width=2cm,minimum height=0.6cm] (n2) at
  (0,4.5) {$\mbox{net}_2$};
  \node [desc,minimum width=2cm,minimum height=0.6cm] (n3) at
  (0,6) {$\mbox{net}_3$};
  \node [desc,minimum width=2cm,minimum height=0.6cm] (n4) at
  (0,7.5) {$\ldots$};

  \draw [semithick,->] (n0) -- node [left] {MCTS} (n1);
  \draw [semithick,->] (n1) -- node [left] {MCTS} (n2);
  \draw [semithick,->] (n2) -- node [left] {MCTS} (n3);
  \draw [semithick,->] (n3) -- node [left] {MCTS} (n4);

\end{tikzpicture}
        \end{minipage}
    \end{tabular}
     \caption{Self-play loop improving quality of net}\label{fig:selfplay7}}
\end{figure}


The self-play design of AlphaGo Zero  combines these two kinds of self-play:  we add
look-ahead  to agent-self-learning and get a system that
improves, through MCTS, its own training input.

We 
get a system as shown in Fig.~\ref{fig:selfplay7}: an agent system in which the
training examples  are generated by using the
same system as the opponent. In each iteration a searcher  creates
higher-quality actions.  MCTS improves the quality of the training
examples in each iteration (left panel), and the net is trained with these better
examples, improving its quality (right panel). 
By letting such a self-learning system play for a while it
will teach itself an increasingly advanced level of play all the way
to world champion.

This  kind of self-play  is used in
AlphaGo Zero. 
%
%
\index{self-learning}
%
%
The output of
MCTS is used as input for the training of the network that is used as
the evalution function in that same MCTS. An extra  loop is
wrapped around the search-eval functions to keep training the network
with the game results, creating a learning curriculum.


To summarize,  AlphaGo Zero self-play consists of the following elements:
\begin{enumerate}
\item Search: MCTS look-ahead using a policy input for selection and a
  value input for backup providing stronger examples
\item Eval: a neural network providing  a policy output and a value
  output to be used in the search
\item Train: a training algorithm to improve the neural network
  with  samples from the played games
\item[R.] Repeat the 3 steps to perform a self-play training curriculum 
\end{enumerate}



\subsubsection*{Self-Learning}
\begin{figure}[t]
    \centering{\begin{tikzpicture}[>=triangle 45,
  desc/.style={
                scale=0.9,
		rectangle,
		rounded corners,
		draw=black,
		}]
  \node [desc] (search) at    (0,2.15) {5 game\_pairs $\leftarrow$ \bf mcts};
  \node [desc] (eval) at      (6,2.15) {5 pol/val $\leftarrow$ {\bf eval}(net(state))};
  \node [desc] (train) at     (3,0.5) {7 net $\leftarrow$ {\bf train}(net, triples)};

  \draw (search.south) edge[semithick,->,out=270,in=180,looseness=1]
  node[left] {\small 6 triples}  (train.west);
  \path (eval.south) edge[loop,semithick,<-,in=0,out=270,looseness=1]
  node[right] {\small \ \ 1 tourn: {\em iterate with new\/} net}  (train.east);
  \path[desc] (search.east) edge[semithick,loop above,<-,min distance=7mm,in=160,out=20,looseness=1] node[above] {\small   policy/value}  (eval.west);
  \path[desc] (search.east) edge[semithick,loop below,->,min distance=7mm,in=200,out=340,looseness=1] node[below] {\small  state}  (eval.west);
    \path[desc] (search.north) edge[semithick,loop above,->,min
    distance=13mm,in=45,out=135] node {\small 2/4 \em game\_pairs}  (search.north);
\end{tikzpicture}
     \caption{A diagram of self-play}\label{fig:selfplay}}
\end{figure}
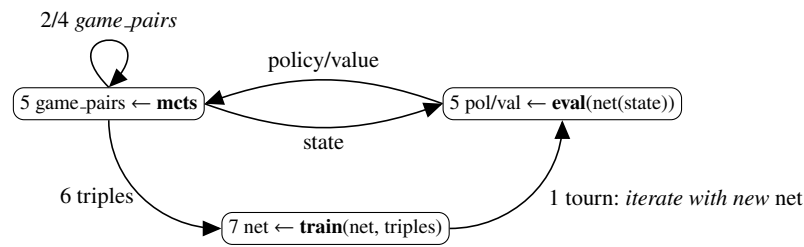

\index{self-play}\index{self-learning}
Conceptually   self-play is as simple as it is
elegant:  a  training loop around a 
standard 
search-eval player with a neural network as evaluation function.
%
%
%
%
Figure~\ref{fig:selfplay} and Listing~\ref{lst:selfplay} show the
self-play loop in more detail. The 
numbers in the diagram refer to the line numbers in the pseudocode.


\lstset{label={lst:selfplay},numbers=left}
\lstset{caption={Self-play pseudocode}}
\begin{lstlisting}[language=Python,float]
for tourn in range (1, max_tourns): # curric. of max_tourns tournaments
    for game in range(1, max_games): # play a tourn. of max_games games; 
        trim(triples) # if buffer full: replace old entries
        while not game_over(): # generate the states of one game
            game_pairs += mcts(eval(net)) # move is a (state,action) pair
        triples += add(games_pairs, game_outcome(game_pairs))     # add to buf
    net = train(net, triples) # retrain with (state,action,outc) triples
\end{lstlisting}

Let us perform an outside-in walk-through of this system using the code in
Listing~\ref{lst:selfplay}.

Line 1 is the main self-play loop. It controls how long the
execution of the curriculum of self-play tournaments will continue. 
Line 2 executes
the training episodes, consisting of tournaments of self-play games.
Line 4 plays such a game to create  ({\em state, action\/}) 
pairs for each move, and the outcome of the game. Line 5 calls MCTS to
generate an action 
in each  state. MCTS performs the simulations where it uses
the policy net in UCT selection, and the value net at the MCTS
leaves. Line 6 adds the outcome of each game to the
({\em state, action})-pairs, to make the ({\em state, action, outcome})-triples for the network to train
on. Note that since the network is a two-headed policy/value net,
both an action and an outcome are needed for network  training.
On the last
line this triples-buffer
is then used to 
train the network. The newly trained network is  used in the next
self-play iteration as the evaluation function by the searcher. 
With this net another 
tournament is played, using the searcher's look-ahead to generate a
next batch of higher-quality examples, resulting in a sequence of stronger and stronger
networks (Fig.~\ref{fig:selfplay7} right panel).

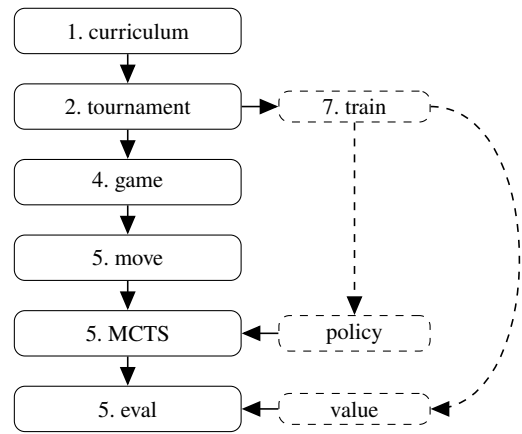
\begin{figure}[t]
    \centering{\begin{tikzpicture}[>=triangle 45,
  desc/.style={
		scale=1,
		rectangle,
		rounded corners,
		draw=black, 
		}]
  
  \node [desc,minimum width=3cm,minimum height=0.6cm] (curriculum) at
  (0,5) {1. curriculum};
  \node [desc,minimum width=3cm,minimum height=0.6cm] (tournament) at
  (0,4) {2. tournament};
  \node [desc,dashed,minimum width=2cm,minimum height=0.4cm] (train)
  at    (3,4) {7. train};
  \node [desc,minimum width=3cm,minimum height=0.6cm] (game) at
  (0,3) {4. game};
  \node [desc,minimum width=3cm,minimum height=0.6cm] (move) at
  (0,2) {5. move};
  \node [desc,minimum width=3cm,minimum height=0.6cm] (mcts) at
  (0,1) {5. MCTS};
  \node [desc,minimum width=3cm,minimum height=0.6cm] (eval) at  (0,0)
  {5. eval};
  \node [desc,dashed,minimum width=2cm,minimum height=0.4cm] (policy) at    (3,1) {policy};
  \node [desc,dashed,minimum width=2cm,minimum height=0.4cm] (value) at  (3,0) {value};

    \draw [semithick,->] (curriculum) -- (tournament);
  \draw [semithick,->] (tournament) -- (game);
  \draw [semithick,->] (tournament) -- (train);
  \draw [semithick,->] (game) -- (move);
  \draw [semithick,->] (move) -- (mcts);
  \draw [semithick,->] (mcts) -- (eval);
  \draw [semithick,<-] (mcts) -- (policy);
  \draw [semithick,<-] (eval) -- (value);
  \draw [semithick,dashed,->] (train) -- (policy);
  \draw [semithick,dashed,->] (train) to[out=0,in=0] (value);

\end{tikzpicture}
     \caption{Levels in self-play}\label{fig:onion}}
\end{figure}

\subsubsection*{Levels of Abstraction}
The self-play loop  looks elegant and simple, but contains an
ingenious interplay of learning at different levels of abstraction. 
The levels of abstraction of AlphaGo Zero-like
self-play are shown in Fig.~\ref{fig:onion}. Looking at
the figure for a moment can help us better understand self-play.

The structure of self-play is a straightforward onion-like system of
functions (or loops) wrapped around each other (Fig.~\ref{fig:onion}). The arrows in the figure point to
the lower functions that are used by the higher function to generate the
training examples. The dashed lines indicate 
the policy/value network, and its  training. The numbers in the figure
correspond to the line numbers in Listing~\ref{lst:selfplay}.

We can also peel the onion by looking outside-in at the data that is generated. 
Starting at the top, a self-play curriculum consists of tournaments. At the end of a
curriculum a
trained player is ready to be used. Each tournament consists
of games (25,000 in AlphaGo Zero); at the end of the tournament the
buffer is filled with example positions to train the network
further. Each game consists of positions and moves; at the end of a
game the {\em outcome\/} 
can be added to the moves ({\em state, action}) ready to be added to the
buffer. Each move is computed by an MCTS search; at the end of the
search a best
action is returned, to be added to the list of pairs of the game
buffer. Each MCTS search consists of simulations (1600 in AlphaGo Zero);
at the end of each simulation another node is added to the MCTS tree. 


\subsubsection*{Triple Recursion}\label{sec:triply}
The neural network of AlphaGo Zero is a single residual network with
19 hidden layers and two output layers: a value head and a policy
head. 
Compared with the original AlphaGo, which had three networks, a
striking element of AlphaGo Zero is the simplicity of 
this network architecture.


The policy and value are calculated in AlphaGo Zero using a single
network. The relationship between the two heads goes back to Chap.~\ref{chap:rl},
where policy functions and value functions were
introduced. 
Policy  and value  are intimately
related (good actions have high values and
vice versa). 


The simple and elegant network architecture allows a 
recursive, self-play training method. The training is triply
recursive: it uses its own code three times. (1) The program self-improves the player in  a curriculum of 
tournaments. (2) In each tournament the players perform
self-play games, generating examples that are used to train the
evaluation function. (3) In each game MCTS uses the same opponent model
in its look-ahead, for itself and for the opponent, just as minimax. Listing~\ref{lst:selfplay} and
Fig.~\ref{fig:selfplay} 
provide the code and the conceptual picture of this triply recursive
training structure in line 1, line 2, and line 5 (the third level of
recursion is implicit inside the MCTS look-ahead).






\subsection{Self-Play and Search-Eval}
Now that we have seen how it works conceptually, let us have a look  at which
techniques  are used in self-play.
All methods of the previous chapters come together in self-play. From
Chap.~\ref{chap:rl} we see the concepts of policy and 
value functions. From Chap.~\ref{chap:search} we see the search-eval
architecture and we see  shape
enhancements and pattern databases. From Chap.~\ref{chap:trial} we see MCTS. From
Chap.~\ref{chap:super} we see deep supervised database
learning, value function approximation, advanced generalization, exploration, and
decorrelation techniques.


Self-play programs have a
search function and 
an evaluation function just like all game playing programs. In AlphaGo Zero
the search function is 
MCTS, and the evaluation function is a deep network. They are used in
two phases: a training phase and a usage phase.

\subsubsection*{Evaluation}
\begin{figure}[t]
    \centering{\includegraphics[width=12cm]{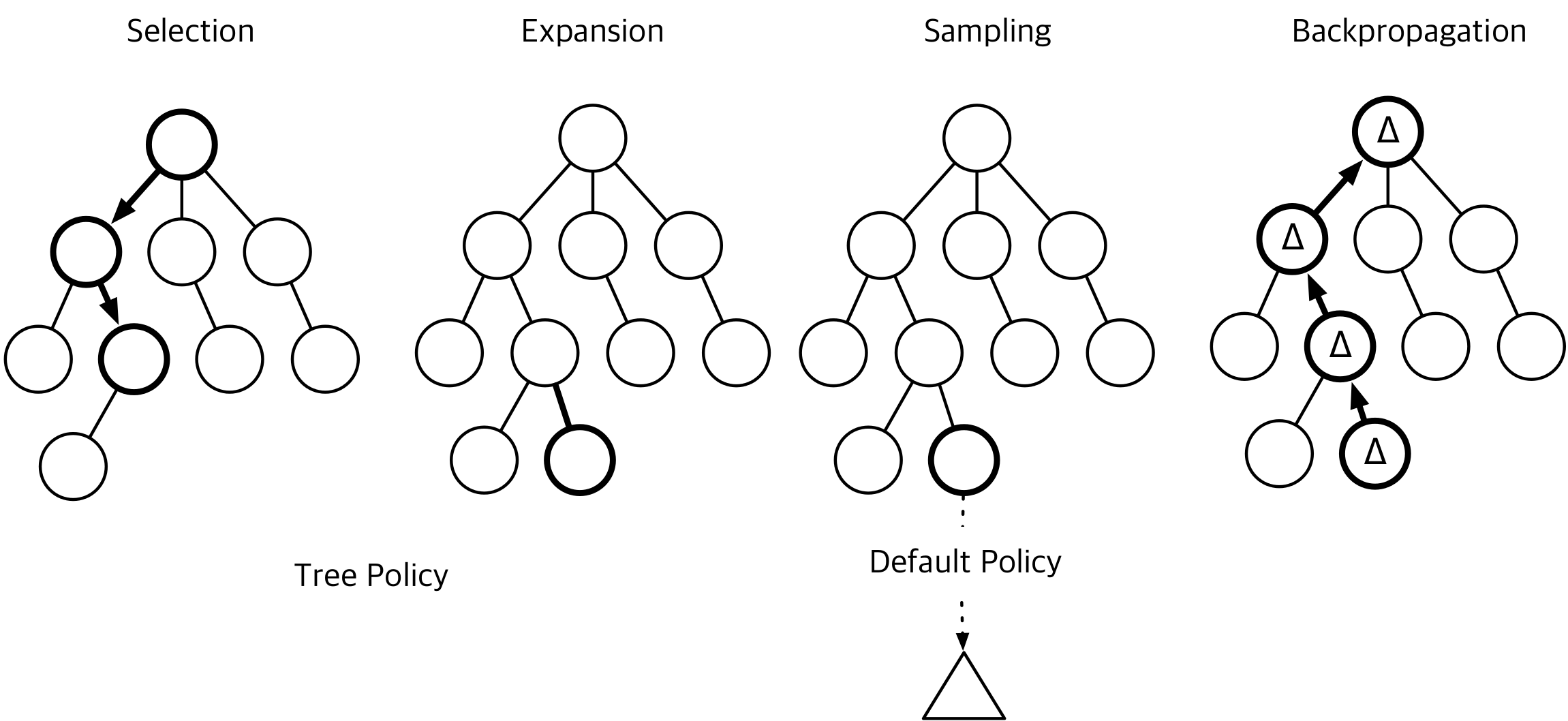}
     \caption{Monte Carlo Tree Search~\cite{browne2012survey}}\label{fig:mcts10}}
\end{figure}

The eval function in end-to-end self-play is a 
deep neural net. The function approximator is trained, 
and is then queried during play (usage phase). Thus, eval has two roles:
\begin{enumerate}
\item While in the training phase, the  eval network is the {\em learning element\/} of the self-play scheme
\item While in the usage phase, eval is a {\em policy/value function\/} to be queried for the
  policy and the value of the state
\end{enumerate}
%
MCTS is an on-policy algorithm that makes use of guidance in two places: in the downward action selection
and in the upward value backup. 
In AlphaGo Zero the function approximator  returns both elements:
a policy for the action selection and a value for the value
backup. The neural network has two output layers, or heads: a policy
head and a value head. 
The policy head of the net is used in the UCT
action
selection, and the value head of the net is used in the
backpropagation of the leaf values of the
MCTS tree. (These are the first and the last MCTS operation in Fig.~\ref{fig:mcts10}.)

\subsubsection*{Search}
In self-play the search function also has two roles: to  look ahead to
generate examples in the training phase
and to look ahead during actual play (usage phase). For both roles the look-ahead
algorithm  does exactly the same thing. It is only  its
{\em role\/} in the self-play scheme that is different during training and use:
\begin{enumerate}
\item While in the training phase, the  search function  generates the examples for the training. By looking ahead, it
generates better quality examples  for the
training.
\item While in the usage phase, when  self-play training has finished and the program is
used for normal game play, the search function
performs look-ahead using the net to find the best action in actual
play.
\end{enumerate}
The two phases are similar to the training and test phases in supervised
learning.





We should note that the
MCTS that is used in the three Alpha-$\star$ programs differs from the
MCTS from Chap.~\ref{chap:trial}.
%
MCTS is used
inside the training loop, as an integral 
part of the self-generation of training examples, to enhance the
quality of the examples for every self-play iteration.

Also, in the $\star$-Zero programs MCTS backups rely fully on the value function
approximator;
no playout is performed anymore. Only the selective
statistics-based exploration/exploitation trade-off of MCTS is left. The MC part in the
name of MCTS, which stands for the Monte Carlo playouts, really has
become a misnomer for this network-guided tree searcher.

Furthermore, selection in self-play MCTS is  different.  UCT-based
node selection now also uses the input from the policy head of the trained function
approximators, in addition to the win rate and newness. Likewise, random
playouts at the leaves are replaced by the value head of the deep
function approximator. 
What remains is that through the UCT mechanism MCTS can focus its search effort greedily
on the part with the highest win rate, while at the same time balancing
exploration of parts of the tree that are underexplored.

The  formula that is  used to incorporate input from the policy head of the
deep network is a variant of
P-UCT~\cite{silver2017mastering,moerland2018a0c,rosin2011multi,matsuzaki2018empirical}\index{P-UCT}. Let
us compare
P-UCT with UCT in order to see the
difference. The original\footnote{Substituting $a$ for the action for
  ease of comparison with the P-UCT terminology, where UCT originally uses $n_j$
  for the visit count of child $j$.} UCT formula is~\cite{kocsis2006bandit}
$$
 \mbox{UCT}(a)=\frac{w_{a}}{n_{a}}+C_{p}\sqrt{\frac{\ln n}{n_{a}}} .
 $$
 The P-UCT formula adds the policy head $\pi(s, a)$:
$$
\mbox{UCT}(a)=\frac{w_{a}}{n_{a}}+C_{p}\pi(s,a)\frac{\sqrt{n}}{1+n_{a}} .
 $$
P-UCT adds the  $\pi(s,a)$ term specifying the policy function of the
action $a$ to the exploration part of the UCT formula.\footnote{Note the small difference under the square root (no logarithm; and addition of 1
 in the denominator) also change the UCT function profile somewhat,
 ensuring correct behavior at unvisited
 actions~\cite{moerland2018a0c}.}

\subsubsection*{Dual-Headed Network}
Let us look at a few concrete facts and numbers
about the AlphaGo Zero self-play implementation, to get a  feeling
for the complexity of the problem~\cite{silver2017mastering}.

AlphaGo Zero uses a dual-headed residual network.  Policy and value loss contribute
equally to the loss function. The network is trained with SGD. L2
regularization is used to reduce overfitting. The network has 19
hidden layers, and an input layer 
and two output layers,  one for the value $v$ estimation
and one for the policy $\pi$ distribution.
The size of the mini-batch for updates is 2048. This batch is
distributed over 64 GPU workers, each with 32 data entries. The
mini-batch is sampled uniformly over the last 500,000 self-play
games (replay buffer). The learning rate started at 0.01 and went down to 0.0001
during self-play.

For search AlphaGo Zero
uses a version of on-policy MCTS that does not use random playouts
anymore,
Dirichlet noise is added to the P-UCT value at the root
node, to ensure that all moves may be tried, increasing
exploration. The $C_p$ value of MCTS in AlphaGo is 5, heavily favoring
exploration. In AlphaGo Zero
the value depends on the stage in the learning; it grows during
self-play. In each
self-play iteration 25,000 games are played. For each move, MCTS
performs 1600 simulations. In total over a three-day course of training 4.9
million games were played, after which AlphaGo Zero outperformed the
previous version, AlphaGo.

More details can be found in Appendix~\ref{sec:ag0-a}.

\subsection{Stable Self-Play through Exploration}
Self-play has a long history in AI, yet the most recent successful self-play
approach  for training a machine learning system was TD-Gammon, 30
years ago.  Now that we have seen the architecture of self-play, let 
us look at 
the performance and the challenges in achieving a
strong level of play.

Self-play uses two forms of  reinforcement learning:  MCTS  in both
downward action selection and  
value backup, and DQN-like reinforcement learning  to train the neural
network with the MCTS results. One could say  that self-play is reinforcement learning
squared. The double reliance on reinforcement learning puts the
quality of the generalization and convergence  in
the center, which
may be the reason why it took so many years  to achieve
high-performance self-learning self-play.

Since all learning is  by reinforcement, the training
process must now be even more stable. The slightest problem
in overfitting 
or correlation between states can throw off the  coverage,
correlation, and convergence. AlphaGo Zero
uses  various forms of exploration to achive stable learning. 
%
Let us summarize how stability is achieved.
\begin{itemize}
\item \emph{Coverage} of the state space is improved by playing a (very) high
  number of games. The quality of the states is further improved by
  MCTS look-ahead. MCTS
searches the state space for good training
samples, improving the quality and diversity of the  covered states.  
The exploration
part of MCTS should make sure that enough new and unexplored parts of
the state space are covered. Dirichlet noise is added  at the root
node and the $C_p$ parameter in the P-UCT formula,
that controls the level of exploration,
has been set sufficiently high, around 4-5 (see also Sect.~\ref{sec:explexpl} and
Appendix~\ref{sec:cp}). 

\item \emph{Correlation} between subsequent states is reduced
through the use of experience replay buffers and MCTS look-ahead, as
in DQN and Rainbow algorithms.

\item \emph{Convergence} of the
training is improved by using on-policy MCTS, and by taking small training
steps. Since the  learning rate is small there is little risk of
divergence, although now convergence is quite slow and requires many training
games.
\end{itemize}
By using  these measures together, stable generalization and convergence
are achieved.
Although self-play is conceptually simple, achieving stable and
high-quality self-play required slow training
with a large number of games, and quite some tuning.
There are many hyperparameters whose values must be set
correctly. Although the values are
published~\cite{silver2017mastering}, the reasoning behind the values
is not always clear.
Reproducing the AlphaGo
Zero results  is not easy, and  much time is  spent in tuning and
experimenting by efforts to reproduce the AphaGo Zero approach~\cite{prasaf2018lessons,straus2018alphazero,nair2017learning,tian2017elf,PhoenixGo2018}.

\subsection{Curriculum Learning}\label{sec:easy}
\label{sec:curriculum}
\index{curriculum learning}
As we will soon see in Sect.~\ref{sec:ag}, AlphaGo and AlphaGo Zero use different training methods. AlphaGo uses
a combination of supervised learning and reinforcement learning, while
AlphaGo Zero is purely based on self-play.
The result of  self-play is not only {\em better\/}
but also {\em faster\/} training, from weeks or even months for AlphaGo to days for AlphaGo
Zero.\label{sec:days}

Why is
self-play faster than a combination of supervised and reinforcement
learning based on games databases? Self-play is faster because it
follows a 
curriculum of examples that is ordered from easy to hard. This kind of
learning is known as curriculum learning~\cite{narvekar2020curriculum,bengio2009curriculum,silver2017mastering}. Let us see
why learning an ordered sequence of examples is
faster than learning many difficult concepts at once.

The training curriculum generated by self-players is a curriculum of
examples from easy to hard. In easy examples the training error is
small; for hard examples it is large.
Problems that generate small errors train quickly; larger errors take more
epochs to converge.
In self-play the targets from the trainer  are 
generated in lock step with the trainee. Both trainer and trainee grow
in playing quality together.

%

\label{sec:curr}\index{curriculum learning}\label{sec:curr:gen}
%





In ordinary deep reinforcement learning  the network is trained by
solving a problem in one large step, using a set of examples
that are not
sorted from easy to 
hard.
With  examples that are not sorted on error value, the program  has to
achieve the optimization step from  
no knowledge to human-level play in one big, unsorted, 
leap, by optimizing many times over the  samples. Overcoming such a
large training step (from beginner to advanced) 
costs much training time.

In contrast, in AlphaGo Zero,  the
network is trained in many small steps, starting against an opponent
that has  just as little knowledge as we have ourselves,
and will play moves that are just as easy as we do,
which goes
quickly.
Subsequently,
harder problems are generated and trained for, using  the network that
has already been pretrained with the easier examples.

Self-play naturally generates a curriculum with examples from easy to hard.
As a consequence,  training times go down and the ultimate playing
strength goes up.





\subsubsection*{Training Speed}
A curriculum strategy is faster because the learner spends less time
with    harder to predict examples.
%
%
Weinshall et al.\ show in a theoretical analysis that curriculum learning is expected to
significantly speed up learning, especially at the beginning of
training~\cite{weinshall2018curriculum}. Furthermore, they show that  generalization
is  
improved with curriculum learning, especially when the conditions
for learning are hard: when the task is difficult, when the network is small, or
when strong regularization is enforced. These theoretical result are
supported by empirical
evidence~\cite{bengio2009curriculum,matiisen2017teacher,erhan2010does,bengio2012practical}.\footnote{Also
  note the link with Hinton's unsupervised one-by-one pretraining
  of layers in a deep network, from simple concepts to
  complicated concepts~\cite{hinton2006reducing}}

Weinshall et al.\ also note that curriculum
learning is not often used  in the real 
world, since it is hard to find situations where  a teacher can rank
the examples from easy to hard.
In most supervised and reinforcement learning settings, there is 
no  ordering from easy to hard available for the
examples. This may
have precluded the widespread adoption of curriculum learning.

\subsubsection*{Correlation Revisited}\index{decorrelation}
Having disussed curriculum learning, it is interesting to note the
existence of a
paradox concerning correlation. The problem is this: in the previous
chapter we have seen  that  breaking correlations between examples
improves state space coverage, preventing cycles and local minima, yet
now curriculum learning show us that correlation between the
difficulty of states achieves curriculum learning.

So: do we want correlation, or not? There must
be enough diversity between 
states within each
self-play iteration to allow good coverage, preventing myopic
learning, to allow learning to progress in each iteration. Yet, for
curriculum learning, examples should be ordered from easy to hard.

On state features subsequent states should be uncorrelated, but on a
larger scale, on
error values, they should be 
correlated.

In
self-play,  sorting on error values occurs implicitly by the learning scheme in
which the examples are generated.
Within the iterations, MCTS, replay buffer, and other
methods are used to achieve a high level of exploration, as discussed previously.

\subsubsection*{Active Learning and Related Learning}\index{active learning}


Curriculum learning is a topic that has  been studied outside deep
learning as well. We will discuss three topics:
active learning, adversarial attacks, and generative adversarial
networks.

Active learning is a type of machine learning that is in between
supervised and reinforcement learning. Active learning is relevant
when labels are in principle available (as in supervised learning)
but at a cost. Active learning performs a kind of iterative
supervised learning, in which the agent
can choose to query which labels to reveal during the learning process. Active learning is related
to reinforcement learning and to curriculum learning, and is for
example of
interest for studies into recommender systems, where acquiring more
information may come at a cost~\cite{settles2009active,rubens2015active,das2016incorporating}.



\label{sec:rladversary}


In Sect.~\ref{sec:onepixel} we saw that deep learning is susceptible
to adversarial attacks, because of the one-pixel
problem.\index{adversarial attack} In deep reinforcement learning this
problem can also give rise to unwanted behavior~\cite{huang2017adversarial}. Gleave et
al.~\cite{gleave2019adversarial} study this problem further, showing the existence of adversarial policies in zero-sum simulated robotics.  

Interestingly, recent work in generative adversarial networks (GANs) has suggested
that the good performance of GANs may be due to curriculum learning. The
generator and discriminator work in tandem, and in a well-tuned system
generate a sequence of easy to hard training
examples~\cite{sharma2018improved,doan2019line}.\index{GAN}\index{generative adversarial net}
Since the publication of the AlphaGo Zero results, and the interest in GANs, curriculum learning has attracted more
attention. An overview of related work can be found in~\cite{weng2020curriculum}.


\subsubsection*{Human Learning is Curriculum Learning}
Curriculum
learning has been studied both in  human learning  and in deep
learning. 
%
%
Most structured, supervised, human learning  is guided by a
curriculum~\cite{weinshall2018curriculum}. When a human teacher
presents examples, the order in which the
examples are presented  is rarely random. Tasks are typically 
divided by the teacher into smaller subtasks and ordered from easy to
hard. This subtasking is
called shaping~\cite{krueger2009flexible} and  studied in
reinforcement learning, e.g., by Graves et
al.~\cite{graves2017automated}.

Krueger and Dayan~\cite{krueger2009flexible} note that in human
and animal learning, curriculum learning is able to learn
higher concepts that are beyond what can be found by trial and error. They use the term {\em
  shaping\/} for the creation of a curriculum of easy and hard targets.

In human learning children are taught easy concepts first. There is a
reason that children
start in first grade, and not in sixth. Using a neural network for
language acquisition, Elman~\cite{elman1993learning}
argues for starting small, with a small network,  and gradually increasing the difficulty of training
examples and the network size. Multi-stage curriculum strategies
improved generalization 
and show faster convergence.



Although curriculum learning has been studied in AI and in psychology
it has not been popular in AI since in 
most training situations it is difficult to find well-sorted training
curricula. However, it has been identified as a key challenge
for machine learning
throughout~\cite{mitchell1980need,mitchell2006discipline,wang2015basic}. 
%
%
Due to the self-play results, curriculum learning is
now attracting more interest in AI (see, e.g., \cite{narvekar2020curriculum,weng2020curriculum}).

\section{AlphaGo}\label{sec:ag}

We have now taken an extensive look at self-play, the learning method
that is used in AlphaGo Zero. We have looked at the
architecture, we have looked at how stable learning was achieved,
and we have
looked at curriculum learning. Self-play builds on a
long history of AI research, from Samuel's Checkers to Tesauro's
TD-Gammon to AlphaGo Zero (Table~\ref{tab:programs}).

Curiously, there was no straight line from TD-Gammon or the other
programs  to AlphaGo Zero's clean self-play
design. First another program was created, AlphaGo, that used
a more complex hybrid design with supervised learning, reinforcement
learning with a database of self-play games, and multiple neural
networks. This complex program  beat the human champions. We will now
describe the earlier AlphaGo approach in a starred section, for your
comparison and historical
insight and to see the continuity with previous research, although its
design has now been superseded by the more elegant self-play
design that achieves tabula rasa learning. Progress in science does not usually follow straight lines.


\subsection{*AlphaGo Architecture}
In 2016 AlphaGo beat human world champions. What technology was used in AlphaGo to achieve such a strong level of
play, ten years before most experts had expected this milestone to be
reached?

It turns out that in AlphaGo there was no magic bullet, no single
groundbreaking new technology that had been invented. In AlphaGo
existing approaches were combined, both supervised and
reinforcement learning, they were improved and refined, and  a way was
found for the different technologies to work
together.

\begin{figure}[t]
   \centering{\includegraphics[width=10cm]{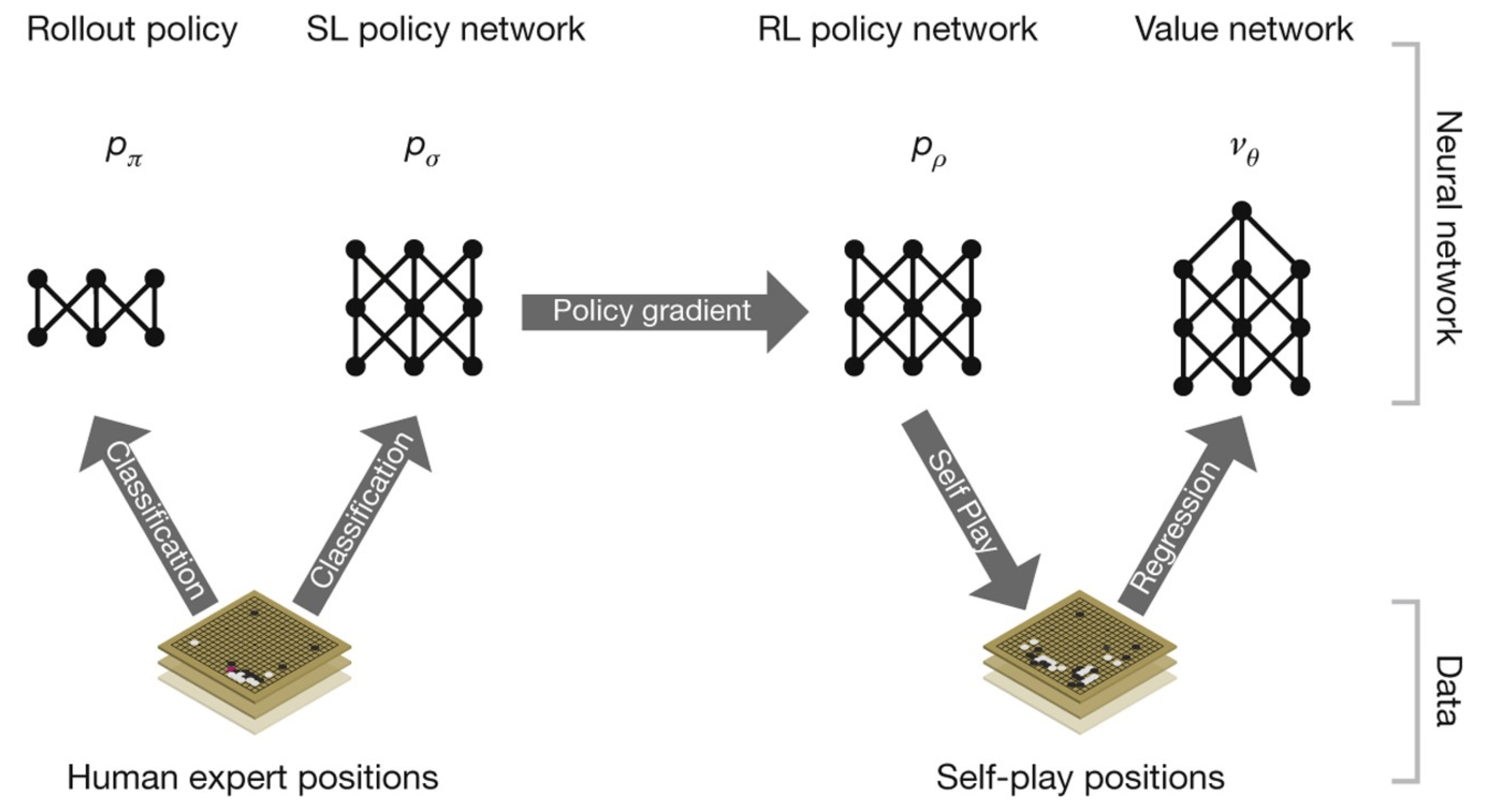}
\caption{AlphaGo networks~\cite{silver2016mastering}}\label{fig:alphago-training}}
\end{figure}

The AlphaGo design follows the search-eval architecture. For search, it employs a
version of
MCTS. For eval,
three neural nets are used: a
fast shape policy net for the MCTS playouts, a slow policy net for UCT
exploration/exploitation selection, and a value net for the leaf
values of the MCTS tree. In addition, the slow policy net is
trained by supervised and by reinforcement learning, giving a total of
four training methods for three
networks. Figure~\ref{fig:alphago-training} reproduces the familiar AlphaGo
networks picture from~\cite{silver2016mastering}.

After this high-level description, let us look in more detail into the
AlphaGo methods.




First, we will discuss the supervised learning in AlphaGo.
AlphaGo uses supervised learning to learn from grandmaster games.  
A
database of 160,000 grandmaster games  is used to train a slower selection
policy network by supervised learning (28 million positions). The
positions from grandmaster games consist of a board position and a best action (the
move played by the grandmasters). This
policy network is used in the MCTS selection
operation.
The supervised learning is used to pretrain the
network to  
a certain (high) level of play, and then positions from self-play
are used to improve it further. 

Second,  AlphaGo uses reinforcement
learning for three networks: for two 
separate policy networks and for a value network.
%
It trains  the slow policy network against an ``environment'' that really is a
database with  games that were played against a
copy of itself.
This part of the training,
against a database of self-play games, is often simply called self-play.

Third, a value network is
trained based on a database of 30 million self-play positions.
For the MCTS leaf values a mix is
used of the value network and of the outcome of  MCTS rollouts using
the fast rollout policy network. The mixing of these values is
governed by a new parameter $\lambda$. Note that in AlphaGo the random
playouts of MCTS are replaced by playouts using the fast policy net. 

Fourth, a game database
is used to train 
a fast policy network by supervised learning. The fast policy network is trained with
positions from separate  games to prevent overfitting and loops  due to
correlation between subsequent game states.

To summarize, AlphaGo uses the four training methods to train three networks:\index{AlphaGo networks}
\begin{enumerate}
  \item A ``slow'' selection policy net, first trained by supervised learning from
  grand-master games, then further refined through self-play positions
\item  A value net for position evaluation, trained by self-play
  positions 
\item A ``fast'' rollout policy net trained by supervised learning from human games,
  used for MCTS playouts
\end{enumerate}

\subsection{AlphaGo Results}

Let us now have a look in more detail at the performance that this complicated
approach has achieved.  

\index{AlphaGo}\label{sec:alphago}
An early version of AlphaGo beat
European champion Fan Hui in London in 2015 in a training match, while
AlphaGo was still being developed. A
year later, after the program had been developed further, it beat
top-ranked Lee Sedol in Seoul in March 2016. In 2017, AlphaGo beat
top-ranked Ke Jie in Wuzhen in May 2017.

\begin{figure}[t]
    \centering{\includegraphics[width=\textwidth]{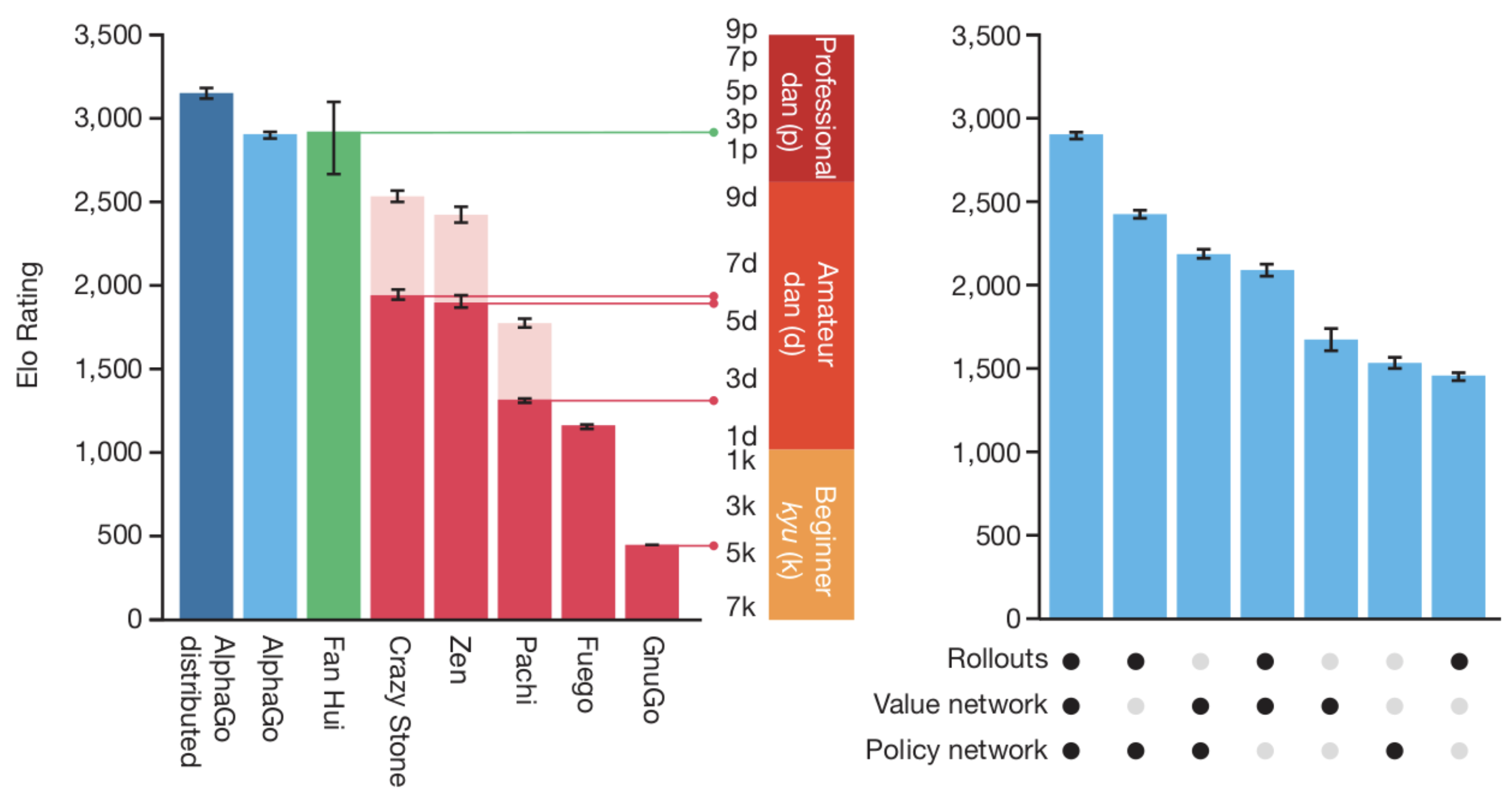}
     \caption{AlphaGo Elo rating; Effect of individual networks \cite{silver2016mastering}}\label{fig:agElo}}
\end{figure}

Figure~\ref{fig:agElo} (left panel) shows the Elo rating of AlphaGo  compared
with other programs\footnote{GnuGo is a minimax-based program. Fuego,
  Pachi, and Zen are conventional MCTS programs without self-play or a
  neural net. Pale red is 4 handicap stones.} and compared with Fan Hui. AlphaGo is clearly superior. The graph indicates a playing strength
of 5p dan. Quite some development and
experimentation were needed for AlphaGo to reach the level of play
that it did. Since publication of the
article in 2016, AlphaGo has been further refined, and the performance  has continued to improve. The version playing against Lee Sedol and  Ke Jie
was significantly stronger, at 9p dan.

We have seen that AlphaGo has three
networks. Therefore, it is instructive to see how each network 
influences performance. Figure~\ref{fig:agElo} (right panel) shows the impact
of turning some of the 
three networks off. We see that each network has a significant
contribution, and that the contribution of the networks is independent
in the sense that they all complement and 
improve each other. 

\begin{figure}[t]
    \centering{\includegraphics[width=6cm]{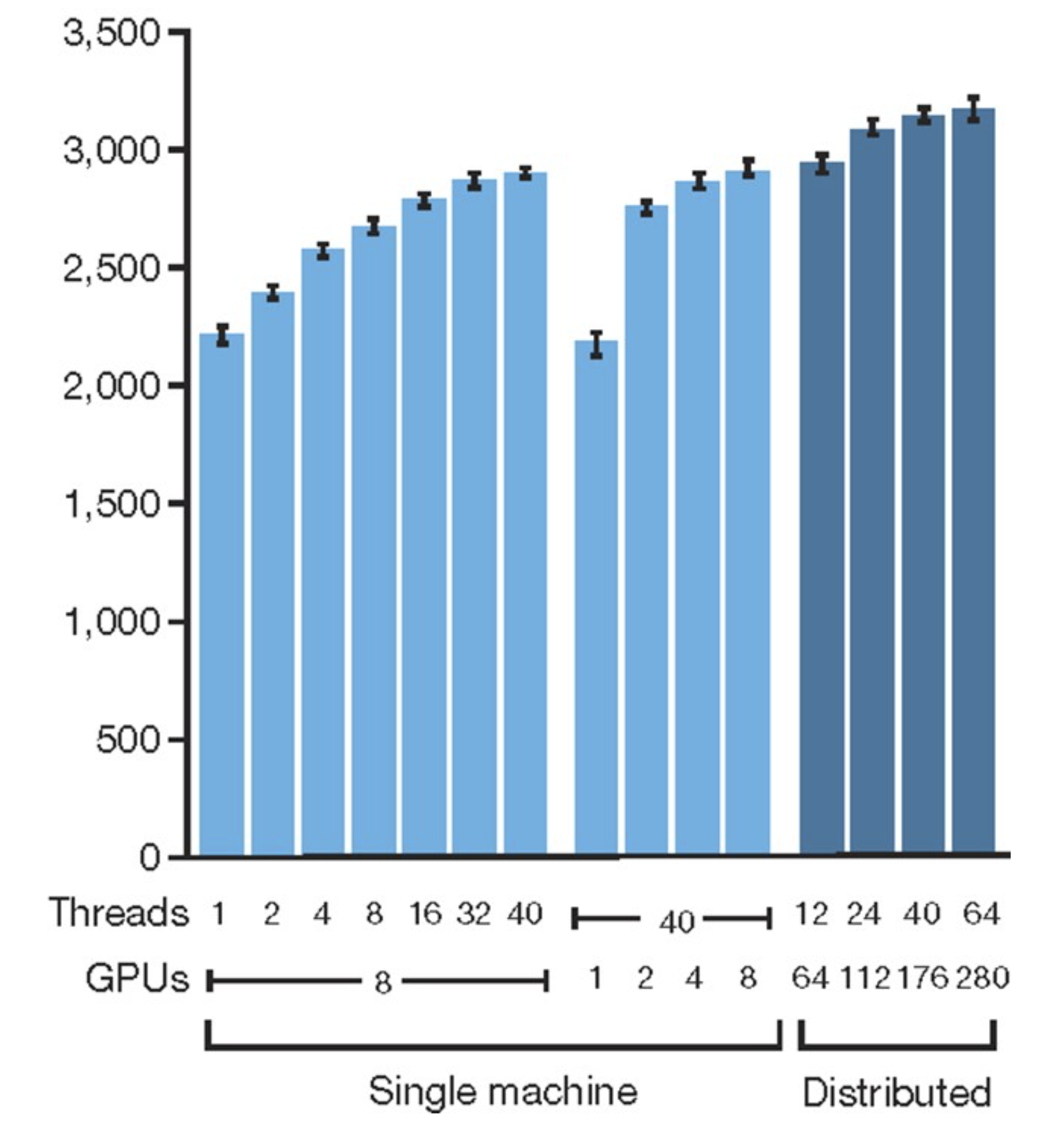}
     \caption{Elo rating of Alphago distributed versions \cite{silver2016mastering}}\label{fig:agDistr}}
\end{figure}

\begin{figure}[t]
    \centering{\includegraphics[width=9cm]{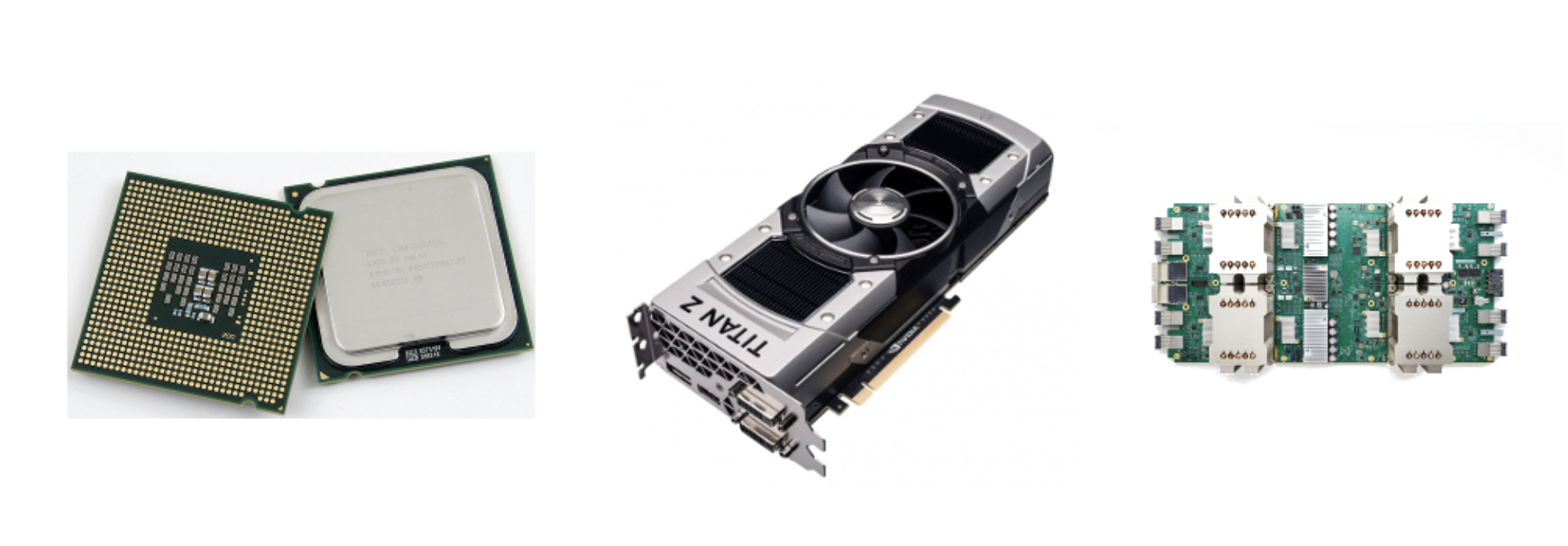}
     \caption{CPU, GPU, and TPU}\label{fig:cpugputpu}}
\end{figure}

AlphaGo  uses a significant amount of compute
power, for training, but also during tournament play time. Therefore, it is interesting to see how more compute
power influences playing strength. Figure~\ref{fig:agDistr} shows the impact
of  extra hardware for 
the distributed version of AlphaGo. Clearly the use of more than one
GPU is advantageous, as is the use of multi-threading. The use of a
cluster of machines (dark blue, distributed) gives an increase in
performance of around 200 Elo points. As preparation for the match
against Ke Jie, AlphaGo used TPUs, specially developed tensor
processing units from Google. TPUs are a kind of GPUs that are optimized for low
precision, high throughput matrix operations (where GPUs are optimized
for high-precision operations). Fig.~\ref{fig:cpugputpu} shows
pictures of a CPU, a GPU, and a TPU.

\begin{figure}[t]
    \centering{\includegraphics[width=\textwidth]{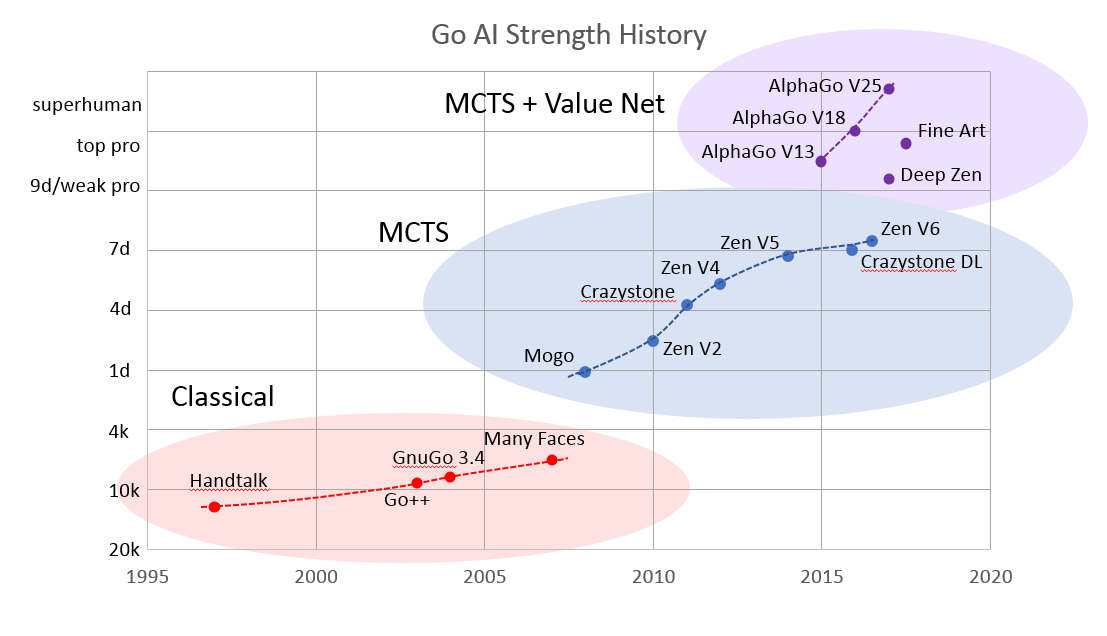}
     \caption{Go playing strength over the years~\cite{reddit2017}}\label{fig:gostrength}}
\end{figure}

\subsubsection*{Playing Strength in Go}
At this point it is interesting to compare programming paradigms  of the
different Go
programs that have been written over the years (Fig.~\ref{fig:gostrength}).  The programs fall into three
categories. First are the
programs based on heuristic planning, the Chess-style
programs. GNU Go is a well-known example of this group of
programs. The heuristics in these programs are hand-coded. Their level of play was at
medium amateur level. Then come the adaptive sampling
MCTS-based programs, which reached strong amateur level. Examples of
these programs are MoGo and Crazy Stone. Finally come the AlphaGo
programs, of MCTS combined with neural nets. They reach super-human
performance. The figure also shows the program Fine Art, by Tencent.

The figure clearly shows how the performance improved with the
different paradigms.

\subsection{Impact of AlphaGo}
The AlphaGo results have caused quite an impact in AI and beyond.
Let us first take a brief look at how science and society reacted to
the AlphaGo breakthrough.

\subsubsection*{Popular Media}
The Lee Sedol match of AlphaGo has  had a large impact in
the popular media. The idea of computers being ``smarter'' than highly
intelligent human champions continued to attract attention, as much as
it did in the days of Deep Blue--Kasparov. Artificial intelligence shot to
prominence, and many AI researchers were interviewed on questions
ranging from when we would see self-driving cars
to the future of the human race.

Newspaper articles appeared on AlphaGo, a  movie\footnote{\url{https://www.alphagomovie.com}} was
made, and many blogs were written by enthusiasts on how to build your
own AlphaGo and AlphaZero. This, by the way, turns out to be quite doable on the
algorithmic side (see Appendix~\ref{sec:reimp} for a list of open-source self-play
systems inspired by AlphaGo Zero), but  acquiring the
necessary computational power to
train the function approximators is a  significant challenge.

\subsubsection*{Impact on Go}
It is interesting to see how human Go players respond to the matches,
and to see the effect on the style of human play. Of course, there was
sadness and melancholy
when the human champions lost against the machines at this
beautiful game, a hold-out of the  supremacy of human intelligence. However, most reactions were positive, and not just for
the  achievement in artificial intelligence. Go 
players were enthusiastic about the innovations and new style of play
that AlphaGo exhibited. See also the commentary by Fan Hui and Lee
Sedol in Sect.~\ref{sec:alphago2}. Players were eager to learn more about how
they could improve their beloved
game even more. Books have been written about the new Go theory that was
created (see, for example, Zhou~\cite{zhou2018}). Furthermore, the
publicity around the 
matches created much new interest in Go. 

\subsubsection*{Impact on Science}\label{sec:science}
The AlphaGo matches have also had quite an impact on society and science. The interest
from society in artificial intelligence has increased significantly, with companies and
governments investing in research labs. Also within
science there has been much interest in multidisciplinary
collaborations, to see how machine learning can transfer to fields
ranging from astronomy
to the arts.
Many collaborations between artificial intelligence and these  fields are emerging.

\subsubsection*{Impact on AI}
On the field of artificial intelligence itself the results
have had a large impact as well. Machine learning has attracted many
new talents, and especially reinforcement learning has
benefited. Attendance at scientific conferences has increased
manyfold, to the extent that some conferences started limiting attendance.

The use of the game of Go as a benchmark for intelligence has led to
an increased interest in benchmarking and
reproducibility~\cite{henderson2018deep,islam2017reproducibility,nagarajan2018deterministic,hutson2018artificial}\index{reproducibility}. Popular suites are
ALE~\cite{bellemare2013arcade},\index{ALE}\index{Arcade Learning Environment}\index{Gym} 
Gym~\cite{brockman2016openai}, ELF~\cite{ELFOpenGo2018},\index{ELF}
StarCraft~\cite{uriarte2015benchmark,koeman2018starcraft,vinyals2017starcraft},
and Mario~\cite{karakovskiy2012mario}.\index{StarCraft}\index{Mario}
See also the list of frameworks in Appendix~\ref{ch:env}.
There is  much interest in  further challenges of self-play.
The computational demands for learning by self-play are large. Wu
reports on a list of improvements that speed up learning
significantly~\cite{wu2019accelerating}.

\subsubsection*{Weak, Strong, and Ultrastrong AI}\index{strong
  AI}\index{Donald Michie}\index{explainable AI}\index{XAI}\label{sec:strong}
At this point it is appropriate to put the AlphaGo achievements in a
more historical perspective. In 1988  Donald Michie defined three
types of AI: weak, strong, and  ultrastrong AI
\cite{michie1988machine}. Michie's criteria were meant to discern
various qualities of machine learning, beyond predictive
performance. Michie's criteria stressed understanding, or
comprehensibility, of learned knowledge. In weak AI the agent
produces improved predictive performance as the amount of data
increases. In strong AI the agent must also provide hypotheses, and
explanations. Ultrastrong AI  should be able to teach the
hypothesis to a human.

Clearly, most of modern machine learning has only achieved weak AI
according to
Michie's criteria, AlphaGo included. For AI to be truly useful in
society, it must achieve more. The current interest in
explainable AI aims to achieve strong AI (see also Sect.~\ref{sec:xai}).

\subsection{AlphaGo Zero Results}
After having discussed the AlphaGo results, let us now turn to the
playing strength of its successor, AlphaGo Zero.

\subsubsection*{Playing Strength}
\begin{figure}[t]
    \centering{\includegraphics[width=\textwidth]{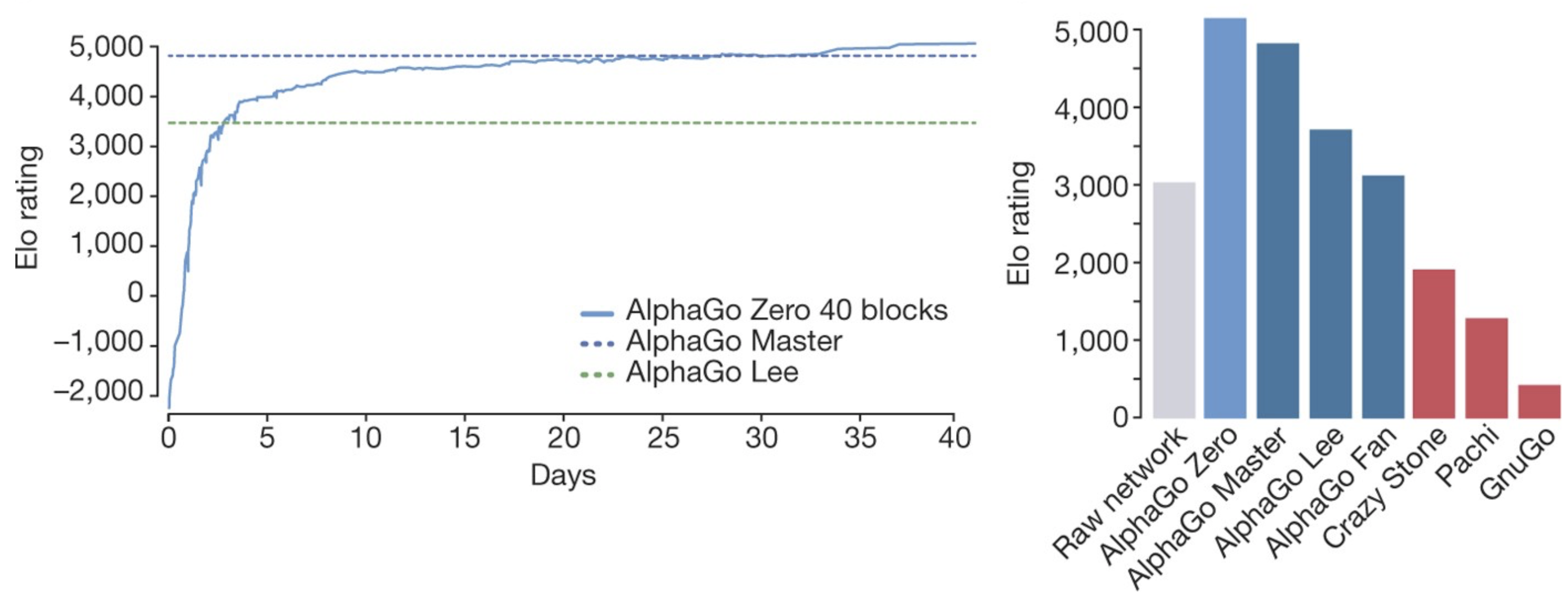}
     \caption{Performance of AlphaGo Zero~\cite{silver2017mastering}}\label{fig:ag0perf}}
\end{figure}

In their paper Silver et al.~\cite{silver2017mastering} describe that learning
progressed smoothly throughout the training. AlphaGo Zero
outperformed the original AlphaGo after just 36 hours. The training time for the
version of  AlphaGo that played Lee Sedol was  several
months. Furthermore AlphaGo Zero used a single machine with 4 tensor
processing units, whereas AlphaGo Lee was distributed over many
machines and used 48 TPUs. It is noted that the training did not suffer 
from the oscillations or catastrophic forgetting that have been
noted in previous
literature~\cite{foerster2017stabilising,heinrich2016deep}. Figure~\ref{fig:ag0perf} shows the
performance of AlphaGo Zero. Also shown is the performance of the raw
network, without MCTS search. The importance of MCTS is large, around
2000 Elo points.

\subsubsection*{Tabula Rasa}
AlphaGo Zero's reinforcement learning is truly learning from
scratch. The paper notes that
AlphaGo Zero discovered a remarkable level of Go knowledge during
its self-play training process. This knowledge included not only fundamental
elements of human Go knowledge, but also nonstandard strategies
beyond the scope of traditional Go knowledge.

\begin{figure}[t]
    \centering{\includegraphics[width=\textwidth]{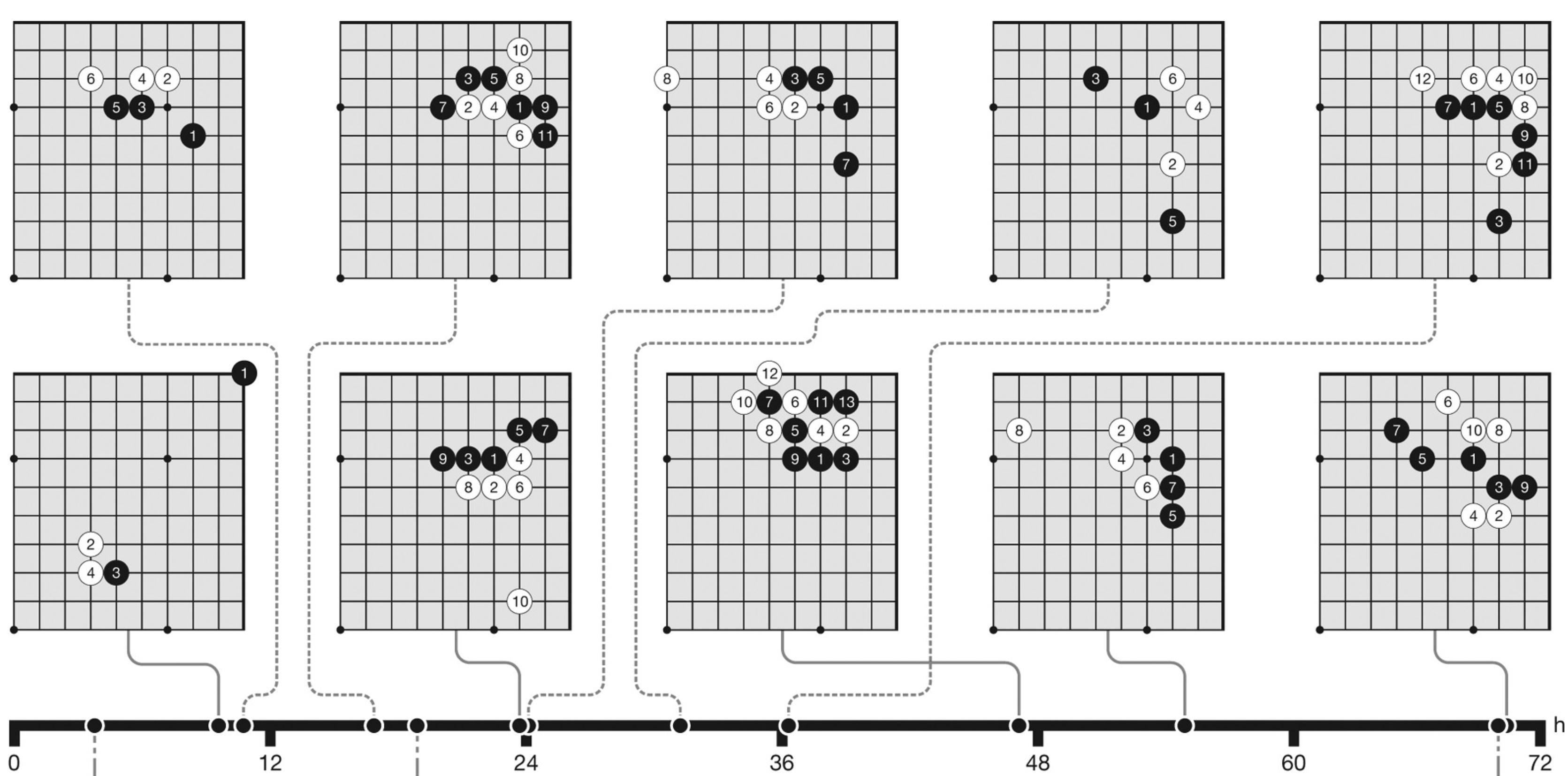}
     \caption{AlphaGo Zero is learning joseki~\cite{silver2017mastering}}\label{fig:joseki}}
\end{figure}

Joseki are standard corner
openings that all Go players become familiar with as they  learn
to play the game. There are
beginner's and advanced joseki. Over the course of its learning,
AlphaGo Zero did indeed learn beginner to advanced 
joseki. It is interesting to study how it did so, as it reveals the
progression of AlphaGo Zero's Go intelligence. Figure~\ref{fig:joseki}
shows how the program learned over the 
course of a few hours of self-play. Not to anthropomorphize too
much, but you can see the little program getting smarter.

The top row shows five joseki that AlphaGo Zero discovered. The first
joseki is one of the standard beginner's openings out of Go theory. As
we move to the right, more difficult joseki are learned, with 
stones being played in looser configurations.

The bottom row shows five joseki favored at different stages of the
self-play training. It starts with a preference for a weak corner
move. After 10 more hours of training, a better 3-3 corner sequence
is favored. More training reveals more, and better, variations.

For a human Go player, it is remarkable to see this kind of progression in
computer play, reminding them of the time when they themselves
discovered these joseki. With such evidence of the computer's
learning, it is hard not to anthropomorphize AlphaGo
Zero (think of the computer as a human being).


\label{sec:indepth}

\section{*Self-Play Enhancements}

We have  looked more broadly at self-play, and have seen earlier programs
that have tried self-play. AlphaGo Zero showed the large potential of
this method. We will now look at enhancements to self-play, further developments, and directions for future research.
First we will start with the third Alpha-$\star$ program: AlphaZero.

\subsection{General Play: AlphaZero}

\index{AlphaZero}\label{sec:alphazero}
After defeating the best human Go players, and after having created
the first tabula rasa self-learning Go computer, a third experiment
was created concerning 
another long-standing challenge in AI: constructing a general game
player. If AlphaGo 
Zero's architecture was able to learn Go from scratch, then perhaps it
would  be able to learn {\em any\/} game? A year after the AlphaGo Zero
publication, 
another paper was published, showing how AlphaGo Zero's design could
also learn to play
Chess and Shogi~\cite{silver2018general,silver2017shogi}.

This publication  showed that self-play was  able to teach the
same search-eval architecture (MCTS with 20
ResNet blocks) to learn Go, Chess, and Shogi, thus achieving a 
form of {\em general\/}
tabula rasa learning, albeit limited to three games. The program, named
AlphaZero, beat the strongest conventional 
players in Go, Chess, and Shogi (respectively AlphaGo Zero, the program
Stockfish,
and the program Elmo).

\begin{table}[t]
    \centering{\includegraphics[width=\textwidth]{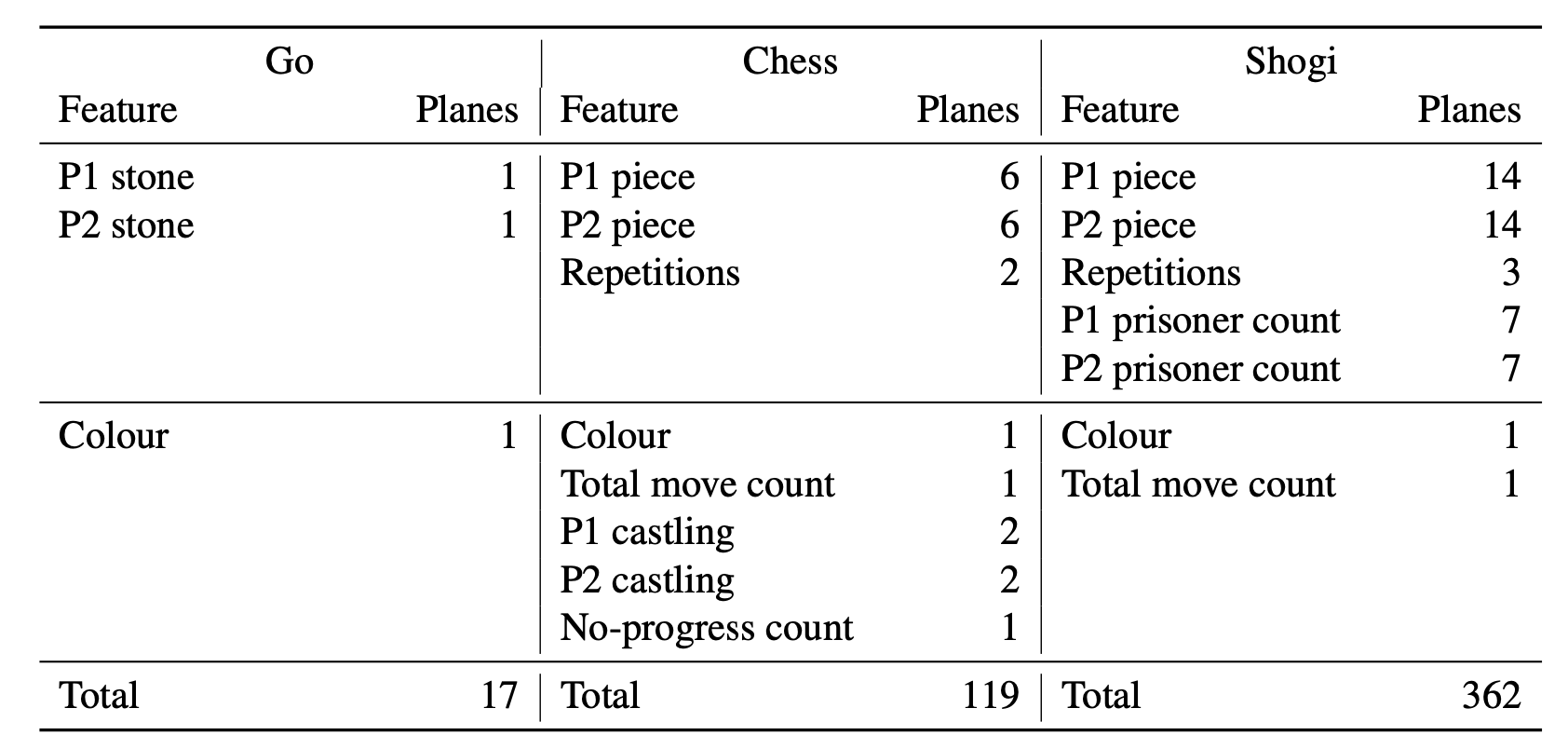}
     \caption[Input Planes]{Different input planes (first set repeated 8 times to
       capture history)~\cite{silver2018general}}\label{tab:planes}}
\end{table}

Note that  there are differences. The hand-crafted input and output layers of the networks are different
for each game. Table~\ref{tab:planes} shows these differences. The more
complicated move patterns of Chess and Shogi cause the input
planes to be significantly more elaborate than for
Go (see the paper for details).
Furthermore, the weights in the trained networks are
also different. Strict transfer learning was not achieved: a trained Shogi net
is not able to play Go well.

AlphaZero's achievement of a general game learning architecture is remarkable for more than one reason.
First, it is remarkable that self-play is so powerful that it can learn
from scratch {\em and\/} become so strong that it beats the current
world champions.
Second, it is remarkable that different games can be learned by the same
network architecture.
Perhaps, some kind of element of general learning must be present
in the structure of the ResNets of AlphaZero, although the networks
that are learned differ between games: a network trained for Chess
cannot play Go well.
Third, for Chess and Shogi all previous
strong programs  
had been based on the heuristic planning approach of alpha-beta with domain-specific heuristics, but  now they
are beaten by a general learning approach using self-play function
approximation (see also~\cite{ramanujan2010adversarial}).

\subsubsection*{Playing Strength}
\begin{figure}[t]
    \centering{\includegraphics[width=\textwidth]{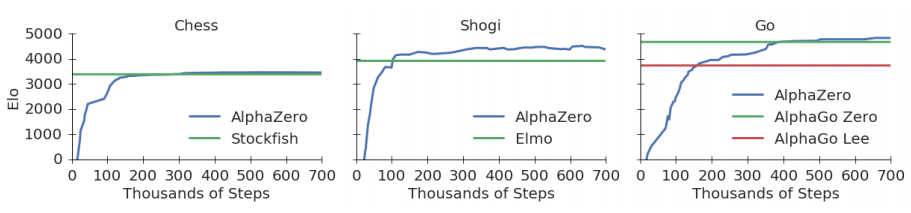}
     \caption{Elo rating of AlphaZero in Chess, Shogi, and Go~\cite{silver2018general}}\label{fig:azelo}}
\end{figure}

The Elo rating of AlphaZero in Chess, Shogi, and Go is shown in
Fig.~\ref{fig:azelo}. Also shown is the Elo rating of the programs
Stockfish, Elmo,  AlphaGo Zero~\cite{silver2017mastering}, and
AlphaGo Lee~\cite{silver2016mastering}.

We can see how AlphaZero is
stronger than the other programs, all through learning by self-play. In Chess, a field that has benefited from work
by a large community of researchers for as long as computers existed,
the difference is the smallest. For Shogi the difference in playing
strength is larger.

Interestingly, AlphaZero slightly outperforms AlphaGo Zero. There are small
differences between AlphaZero and AlphaGo Zero. AlphaZero updates the
neural network continually, whereas in AlphaGo Zero the new net is only
accepted if it wins 55\% of the time against the old
network. AlphaZero does not exploit Go symmetries. The paper does not
provide other reasons for the differences.

\subsubsection*{Go, Chess, and Shogi}\index{Shogi}\index{Elmo}\index{Stockfish}
That AlphaZero can play three different games with one architecture is surprising.
The three games, though all board games, are quite different. Go is a
static game of strategy. Stones, once placed, do not move, and are
rarely captured. Each stone played is therefore of strategic
importance. In Chess the pieces move, it is a  dynamic game
where tactics are important, and the possibility of sudden death
is an important element
of play: in Chess it is
possible to win in the middle of the game by capturing the king. No such
concept exists in Go. Indeed, one element of human Chess play is to
move the king to a safe position to reduce the level of stress in the game. Shogi,
sometimes called Japanese Chess, is
even more dynamic. Pieces, once captured, can be returned to the
game, changing loyalty, creating even more complex game dynamics.\footnote{Pieces
are of the same color, but are wedge-shaped, and change orientation
when switching sides.}

As all top Chess programs,  Stockfish follows the classic search-eval
design of alpha-beta  and a heuristic
evaluation function, that we have seen in
Sunfish (Sect.~\ref{sec:sunfish}) and Deep Blue
(Sect.~\ref{sec:deepblue}). Among the enhancements of Stockfish are 
forward pruning,
iterative deepening, transposition tables, quiescence, piece-square
tables, null window, null move, history heuristics, 
search extensions, search reductions, an
opening book, and
an end-game database. Chapter~\ref{chap:idtt} provides 
explanations on
most of these enhancements. In Stockfish, all these enhancements have been 
refined, tested, and tuned, for many years~\cite{stockfish}.

The state space complexity of Shogi is  substantially larger than
Chess: $10^{71}$ versus
$10^{47}$. The strongest
Shogi programs have recently started defeating human
champions~\cite{shogi}. Elmo, one of these programs, is
designed like  a Chess program, around alpha-beta search and a heuristic
evaluation function, with many  tuned and tested game-dependent enhancements.

It would seem that since the move patterns, the tactics, and the
strategies to be learned are very different, that it would require
a different kind of program to  play well. Indeed, this is the conventional 
approach: a  developer crafts a program specifically for a
certain game, incorporating all kinds of game-specific knowledge and
heuristics into the code.  For Go, the game is so subtle
and difficult that no 
scientist was able to find successful heuristics, and knowledge
had to be learned by machine learning from  game
databases (initially, in AlphaGo) or through self-play 
(eventually, in AlphaGo Zero).

The conventional game-specific
approach has worked well for Chess, Shogi, and  other games so
far.  AlphaZero
is the first to show that it is 
possible to be general {\em and\/} strong.\footnote{In
  Sect.~\ref{sec:ggp} another approach is introduced, general game
  playing, or GGP. The general programs created by GGP generally trade off
generality for playing strength.}
Indeed, AlphaZero has shown that the differences in move patterns are
not such that  a different kind of program is needed to  learn
to play well. 


\subsubsection*{One General Architecture}
Although it would seem that the different kinds of games
require a different program design, AlphaZero's results have proven this
conventional wisdom wrong, at least for the  learning of three board games. All that needed to be changed were the
input and output layers of the function approximator. The self-play training
algorithms, network architecture, search algorithm, most training
hyperparameters,\footnote{The learning rate decay scheme was slightly
  different for Go than for Chess and Shogi.} and compute hardware 
are all the same. The self-play
program was then able to learn to play the game beyond world class
level. Perhaps the game dynamics of Go, Chess, and Shogi are less
different than they seemed, but clearly the structure of self-play with ResNets is general
enough to allow it to  learn very different nonlinear value/policy
functions. 

It is this last aspect that makes the results so inspiring from an AI
point of view, since it
reminds us of how humans learn to play.


\subsection{Training as Generalization}
\index{transfer learning}\label{sec:learning}
We will now look at other generalization methods, in particular,
transfer learning. 

The goal of machine learning is to build an approximation function based on 
examples. Many methods exist to train the approximator.
Generalization is the process minimizing an error  function  over input examples.
In some sense,  all training is generalization.

\subsubsection*{Sample Efficiency}\index{sample efficiency}
The canonical deep learning case is training from zero knowledge, when
the network parameters are initialized to random numbers. Training
methods have been developed for efficient training yielding the lowest
error function in as few training steps as possible. Since in
reinforcement learning each training example is a costly sample from the
environment,  achieving high sample efficiency is important.

Model-based methods, such as planning, are more sample efficient than
model-free methods. All they need to do to find a value of a state is to follow
the model and lookup the value of the actions.
Compared to planning, deep learning is not very sample efficient.
Model-free methods have no model to follow and 
need to sample many states before the value  approximates the true
value.

Deep reinforcement learning is a model-free method 
that takes many samples to learn the features. Furthermore, the learning rate is
set low for good convergence and generalization. In contrast to
planning, however, deep learning is able to learn an evaluation
function, which is something that planning cannot do. And although planning is sample efficient, it is inadequate
for traversing exponentially large state spaces. For that, it needs
a good evaluation function. An evaluation that is better than a
heuristic, and better than averaged sampling. Function approximation
can learn such a good function, and self-play can use planning to generate
samples of increasing quality to continuously improve the learning evaluation
function.

Self-play is thus not very sample efficient, because of the model-free
evaluation function approximation, and because the model-based planning is
stretched to perform many state space traversals to improve the
quality of the learning targets. Without curriculum learning it would
be worse.

\subsubsection*{Pretraining}\index{pretraining}
Pre-training the network with another related task may increase sample
efficiency. The network weights are no longer randomly initialized,
but are such that the error is lower and training can progress with
fewer samples.

Curriculum learning performs a sequence of pretraining steps. In each
curriculum iteration the network is pretrained by the former
iteration, creating an easier learning task with a low error.

\subsubsection*{Lifelong Machine Learning}\index{lifelong learning}
It has been  observed that most machine learning tasks currently start from
zeroknowledge, and subsequently take a long time to learn. In
contrast, in human learning experiments, subjects do not start from zero knowledge,
and typically learn a new task quickly with few samples.
Humans already have knowledge of related tasks. 
Lifelong machine learning studies methods to retain learned knowledge
tasks that are trained from different domains over the lifetime of the
system~\cite{thrun2012explanation}.  The
goal is to  selectively transfer that knowledge when learning a new
task~\cite{silver2013lifelong}.

\subsubsection*{Transfer Learning}\index{multi-task learning}\label{sec:transfer}
Transfer learning aims at improving the process of learning new tasks
using the experience gained by solving similar predecessor problems~\cite{pratt1993discriminability,thrun2012learning}. As humans, we make use of our past
experience by not only repeating the same task in the future but also for
learning completely new tasks. That is, if the new problem that
we try to solve is similar to a few of our past experiences, it
becomes easier for us.  Transfer learning aims
 to transfer past experience of one or more source tasks and
 use  it to boost learning in a related target task.
 
When we want to transfer knowledge, it can be transferred by initializing part of the network by
other weights, for example, by transfering the entire network, and then
retraining, or  transferring some of the lower layers of a
network.

\subsubsection*{Multi-task learning}\index{multi-task learning}
Transfer
learning is related to multi-task learning, where one network is
optimized for multiple tasks that are  related. The
major task can be learned better through using the experience gained
by other tasks. This approach is effective when the tasks have some
commonality. In multi-task learning, related learning
tasks are learned at the same time, whereas in transfer learning they
are learned in sequence by different networks. Multi-task learning improves
regularization by  requiring the algorithm to perform well on a
related learning task instead of penalizing all overfitting
uniformly~\cite{evgeniou2004regularized,argyriou2007multi}. 
%
The two-headed AlphaGo Zero
network optimizes for value and
for policy at the same time in the same network~\cite{pan2010survey,caruana1997multitask}. 
Multi-task
learning has also been applied to Atari
games~\cite{kelly2018emergent,kelly2017multi}.

\subsubsection*{Meta-learning}\index{meta-learning}\index{class embedding}\index{embeddings}\index{zero-shot learning}
There are many forms of transfer learning~\cite{caruana1997multitask,pan2010survey,taylor2009transfer}. Going beyond transfer of network parameter values,
knowledge transfer  can also  include hyperparameter values, 
network architecture, or learning algorithms. Transfer learning is a form of meta-learning where the extent of
domain adaptation is typically restricted to weight parameters, where
in meta-learning other parts of the algorithm are also involved~\cite{brazdil2008metalearning}.

Meta-learning is sometimes described as ``learning to learn'' or
selecting the best learning algorithm for a task. It is related to
algorithm selection~\cite{rice1976algorithm,kerschke2019automated} and hyperparameter optimization, and also to
transfer learning. Although meta-learning work goes back to the
1980s~\cite{schaul2010metalearning,dickmanns1987genetische,bengio1990learning,brazdil2008metalearning}
there has been  recent interest in the field~\cite{hospedales2020meta,vanschoren2018meta,huisman2020deep}, also because of work on
zero-shot learning~\cite{lake2015human,graves2014neural,xian2017zero}. 
In few-shot or zero-shot learning new concepts are recognized after
seeing only
a few problem instances~\cite{xian2017zero,santoro2016meta,rezende2016one}.  

Normally a classifier is expected to correctly classify at test time new samples
from classes that have been seen during training. A zero-shot
classifier has seen no samples from the classes  at training
time. To be able to correctly classify the samples the classifier is
given some information about the classes. This can be a natural text
or structured description~\cite{hospedales2020meta,romera2015embarrassingly}. RNN
and LSTM are often used~\cite{ravi2016optimization}. The information
can also be a similarity measure between the classes.

This brings us to the question of similarity  metrics between
problems.
High-dimensional discrete classes can be embedded in  a
lower-dimensional continuous vector. The zero-shot classifier can then 
predict a sample's position in that space, and the nearest embedded
class is used as the predicted class~\cite{frome2013devise}.
In
addition to being a similarity measure, low-dimensional
representations also allow visualization, providing insight 
into the often high-dimensional state spaces of machine learning
problems~\cite{pezzotti2017deepeyes,montavon2018methods,maaten2008visualizing}.

Model-agnostic meta-learning (MAML) is a recent meta-learning approach
for few-shot learning~\cite{finn2017model}.\index{MAML}
MAML uses optimization-based meta-learning, and  learns  initial
network representations that can then be fine-tuned using a few class
examples. This work has spawned more
interest in 
meta-learning~\cite{antoniou2018train,behl2019alpha,finn2018probabilistic,zhang2018metagan,vinyals2016matching,rakelly2019efficient}. New
 benchmark suites for research on multi-task and meta
 learning have been introduced, Meta
World~\cite{yu2019meta} and Meta-dataset~\cite{triantafillou2019meta}.

In meta-reinforcement learning Duan et al.~\cite{duan2016rl} introduce
RL$^2$, an approach in which the  
reinforcement learning algorithm itself is a target of a training
process, and Wang et al.~\cite{wang2016learning} explore the link between
meta-learning and reinforcement learning, by having one reinforcement
learning algorithm implement a second reinforcement learning
algorithm.

Let us now look at hyperparameter optimization methods.

\subsection{Hyperparameter Optimization}\index{hyperparameter optimization}\label{sec:hyper}

Deep learning and self-play systems typically have many
hyperparameters, to control various aspects of the learning process. The
AlphaGo Zero paper lists those presented in
Table~\ref{tab:hyp}.

\begin{table}[t]
\begin{center}
\begin{tabular}{ll}
{\bf Hyperparameter} & {\bf Value}\\ \hline\hline
self play games per iteration & 25000\\ 
MCTS simulations & 1600\\ 
exploration temperature & 1 for first 30 moves, 0
afterwards \\ 
Dirichlet noise & $\eta=0.03, \epsilon=0.25$ \\
resignation threshold & 5\%\\ 
regularization parameter & $c=10^{0.4}$\\ 
mini batch size & 2048, 32 per worker \\ 
self play positions & 500,000 \\ 
learning rate&annealed with momentum  0.9\\ 
CE loss/MSE  weighting&  equal in the loss
function\\ 
optimization checkpoint & 1000\\
self play evaluator &  400\\ 
win margin for acceptance & 55\% \\ 
network architecture & 19 residual blocks\\ \hline
\end{tabular}
\caption{AlphaGo hyperparameters~\cite{silver2017mastering}}\label{tab:hyp}
\end{center}
\end{table}

The effects of some of these hyperparameters may interfere with
each other.
Finding the
optimal setting for all parameters is  a daunting task, best left to
machine optimization.\footnote{Just as finding the best coefficients for an
evaluation function of hand-crafted heuristics is best left to machine
learning, and, for that matter, crafting the evaluation features
themselves  is also best left to machine learning, and crafting a curriculum of
training positions is best left to self-play. Some would even say that
crafting machine learning algorithms is best left to machine learning~\cite{oh2020discovering},
but that is a topic for another book~\cite{harman2007current}.} Many optimization algorithms
have a large set of hyperparameters, and hyperparameter optimization
is an active field of research. Well-known methods are exhaustive
search~\cite{hoos2004stochastic}, random search~\cite{bergstra2012random,wang2016bayesian},
Bayesian optimization~\cite{bergstra2011algorithms,hutter2011sequential,thornton2013auto,snoek2012practical}, and
evolutionary
strategies~\cite{back1993overview,burke2013hyper,pappa2014contrasting,van2015optimally}. 
Some
software packages for hyperparameter optimization are
\href{https://github.com/automl/SMAC3}{SMAC}\footnote{\url{https://github.com/automl/SMAC3}}~\cite{hutter2011sequential},
\href{https://automl.github.io/auto-sklearn/stable/}{auto-sklearn}\footnote{\url{https://automl.github.io/auto-sklearn/stable/}}~\cite{feurer2015efficient},
\href{http://www.cs.ubc.ca/labs/beta/Projects/autoweka/}{Auto-WEKA}\footnote{\url{http://www.cs.ubc.ca/labs/beta/Projects/autoweka/}}~\cite{kotthoff2017auto},
\href{https://scikit-learn.org/stable/}{scikit-learn}\footnote{\url{https://scikit-learn.org/stable/}}~\cite{pedregosa2011scikit},
  \href{http://iridia.ulb.ac.be/irace/}{irace}\footnote{\url{http://iridia.ulb.ac.be/irace/}}~\cite{lopez2016irace},
  and
  \href{https://code.fb.com/ai-research/nevergrad/}{nevergrad}\footnote{\url{https://code.fb.com/ai-research/nevergrad/}}~\cite{nevergrad}.\index{nevergrad}\index{irace}\index{scikit-learn}\index{AutoML}\index{AutoWeka}\index{auto-sklearn}\index{SMAC}

   A parameter sweep of
AlphaZeroGeneral (Sect.~\ref{sec:a0g}) is Wang et al.~\cite{wang2019hyper,wang2020analysis}.
AlphaGo Zero optimized its hyperparameters with Bayesian optimization,
AlphaZero used the same set of hyperparameters.

\subsubsection*{Network Architecture}\label{sec:nas}
The  architecture of the network can also be regarded as a
hyperparameter to be optimized. Among the decisions to make are the number
and types of the layers, and the number and types of the neurons, whether to
have ResNet cells, and how they are connected. Getting the architecture 
right for your problem can be a time-consuming problem, perhaps also
best left to machine learning.

Neural Architecture Search is an active field of research, using
methods that are related to
hyperparameter optimization. Two surveys
  are~\cite{yao1999evolving} and \cite{elsken2018neural}. Random search, Bayesian
optimization~\cite{yoshinari2018probabilistic}, 
evolutionary
algorithms~\cite{kitano1990designing,van2018automatic,jaderberg2017population,miikkulainen2019evolving,wang2019on,back1996evolutionary}, and
reinforcement
learning~\cite{zoph2016neural,pham2018efficient,baker2016designing}
are popular methods.\index{evolutionary strategies}

\subsubsection*{Network Optimization}
An important element of all neural network function approximators is
the optimization algorithm. Stochastic gradient descent has been the
optimization algorithm of
choice, although alternatives exist. Recently works
report on evolution
strategies as network optimizers, for example when derivates of the error
function are not available~\cite{chrabaszcz2018back,salimans2017evolution,wierstra2008natural}. An
advantage of evolution strategies is that they are well suited for
parallelization, providing possible speedups for the training phases
of self-play.




Another alternative to SGD network optimization are distributional methods, where not just the
expected value is optimized, but also the variance of the error~\cite{bellemare2017distributional,moerland2017efficient,moerland2018potential}
(Sect.~\ref{sec:distrdqn}).



\subsection{Parallelism}\index{parallelism}\label{sec:parself}
A major problem of self-play is the large computational requirement,
which translates to long training times. Self-play has three obvious
areas where parallelism can be used for speedup. First of all, 
network training can be sped up. Here GPUs and TPUs are already used, and
TensorFlow and other packages typically include support for parallel training.

Another option is to parallelize MCTS. We have already seen a few parallel
MCTS approaches, such as tree parallelism and root parallelism, in
Sect.~\ref{sec:par}. Even though their use is not trivial, 
implementation in 
current self-play can speed up training. 

Furthermore, the tournament games of the self-play setup are all
independent, and can be parallelized within the same self-play
iteration. This should give an important 
speedup of self-play, filling the example buffer in parallel. Advances
in this area will enable more experimentation
and more  insights into self-play processes.


\subsection{Further Research}\label{sec:further}

Self-play has been an active area of research in two-agent
zero-sum games for many years. The
AlphaZero results have reinforced the power of the paradigm, showing
that TD-Gammon's success was not an exception. The
successes have created much research interest in fundamental
questions and applications of self-play. 

However, there remain open
questions to be answered.
Among them are:
\begin{itemize}
\item The learning time for AlphaGo Zero is long; many games are
  generated to achieve good coverage, (de-)correlation, and
  convergence. How can self-play be made faster?
  
\item Self-play works for two-agent board games.
For which other domains can self-play work? Does self-play also work for
other Markov decision problems, or for non-Markovian problems?
\item
Did AlphaGo/DQN self-play kill the deadly triad? How are overfitting
and specialization related in reinforcement learning?

\item
Why do ResNets perform so well in self-play? Are there other
architectures that work even better?

\item
Does self-play work well in imperfect-information games such as
Bridge, Poker, and StarCraft? 
\item
Are self-play results easy to transfer? How can 
concepts that are learned in one game be transferred to other games? (Note that in AlphaZero each
game  starts training from scratch.)
\item
Evolutionary strategies and related population-based methods have been successful in some studies. Will they
play a role in (general) self-play?
\end{itemize}
%
The
successes have prompted researchers to look at other domains, beyond
two-agent zero-sum, to see
if and how these questions can be answered and
 similar progress can be made. We will list some future
research directions, noting relevant papers. We start with different kinds of  problems,
such as single-agent games,  multi-agent problems, and then we continue with different
ideas.

\subsubsection*{Single-Agent Games}\label{sec:single}
Reinforcement learning is a natural training paradigm for two-agent games.
Reinforcement learning is also used in multi-agent and imperfect-information games,
such as StarCraft and 
Poker~\cite{heinrich2016deep}.

However, many optimization problems in science
are single-agent optimization problems, such as navigation, the Traveling
Salesperson Problem, the 15 puzzle, and the
Satisfiablity Problem~\cite{russell2016artificial}.
MCTS and neural
networks have been used  to
solve single-agent combinatorial problems~\cite{moerland2018a0c,ruijl2014hepgame,vinyals2015pointer,smith1999neural}. Single-agent optimization
problems, where the shortest solution must be found, can have
especially sparse state
spaces in which few reward signals are present. Learning in such
sparse spaces is hard. A solution can be to introduce intermediate
goals.

The advantage of self-play is that it creates a curriculum learning
situation, which speeds up learning.
Approximate reinforcement learning has been applied to
TSP~\cite{kool2018attention,bello2016neural,gambardella1995ant}.
Xu and Lieberherr~\cite{xu2019learning} also used self-play
in combinatorial optimization.
It can be expected that more research into single-agent self-play
approximators will  appear, attempting to benefit from the
curriculum learning effect that is 
inherent in self-play approximators~\cite{matiisen2017teacher}.

\subsubsection*{Multi-agent and Population-Based Methods}
\label{sec:multi}

Two-agent games are typically competitive (zero-sum) games. 
Multi-agent games are often (partly) cooperative. They have
been the subject of study for some time
~\cite{littman1994markov,tan1993multi,claus1998dynamics,busoniu2008comprehensive,panait2005cooperative,heinrich2016deep,lowe2017multi,charlesworth2018application}.

Multi-agent games, real-time strategy games, and
multiplayer online battle arenas
model more aspects of real life. They are a logical next step beyond
two-person zero-sum perfect-information games. Real-time strategy
games also feature movement aspects that 
are closer to robotics than board games do.

Applying 
the recent progress in self-play with multi-agent cooperative games
will be quite interesting, as we will see in the next
chapter.\index{multi-agent reinforcement learning}
Combinations of multi-agent and population-based approaches have
shown great progress in StarCraft II and Defense of the Ancients
2 (DOTA). See  Sect.~\ref{sec:rts} for more.

\subsubsection*{Sparse Rewards}
Many real-world  reinforcement learning  problems are sparse problems: the reward information in  state spaces is sparse, and 
credit assignment can be  delayed for a long time. Furthermore, the
action space may be large as well.  An example of a game with sparse rewards is Montezuma's Revenge,
an Atari game in which many actions can be taken without encountering a
 reward state or a change in value. This game requires more advanced strategic reasoning,
presenting a challenging problem for many reinforcement learning methods (Sect.~\ref{sec:montezuma})~\cite{painter2012greedy,qin2014sparse,abbeel2007application}.\index{sparse
reward} The  Go explore algorithm is specifically designed
for exploration of sparse or deceptive reward environment. It uses a
combination of remembering previously visited states, look-ahead, and exploring
from promising states. It is reported to
perform well in Montezuma's Revenge and in Pitfall~\cite{ecoffet2019go}.\index{Go-Explore}

\subsubsection*{Continuous Action Spaces}
Many applications have large, continuous action spaces, possibly
combined with imperfect information (see Poker and StarCraft in
Sect.~\ref{sec:poker}). We often encounter this in real
life~\cite{lillicrap2015continuous,moerland2018a0c,doya2000reinforcement,santamaria1997experiments}.\index{continuous  action space}
Large discrete or continuous action spaces can be segmented or
discretized~\cite{dulac2015deep}. Examples are  prevalent in
robotics, where an arm movement can be 
over a continuous angle; a good review of this field
is~\cite{kober2013reinforcement}.\index{robotics}


\subsubsection*{Hierarchical Reinforcement
  Learning}\label{sec:hier}\index{sample efficiency}
One of the main challenges in reinforcement learning is scaling up classic
methods to problems with large action or state spaces. 
A problem of model-free reinforcement learning is low  sample
efficiency; sample generation is slow, or samples are not used
efficiently (see also Sect.~\ref{sec:sample}).

After having seen how different parameters of the
training process can be optimized, we will look at hierarchical
reinforcement learning, a more principled
idea to scale reinforcement learning to larger problem classes, by
decomposing large problems into smaller ones.

Most reinforcement learning problems are specified at a low level of
detail. The steps that we can make are small (move a single
piece, perform a single movement of a robot arm). The advantage is
that this allows for precision and optimality
. The disadvantage is the large size of the state space.

In contrast, consider the following problem:
In the real world, when we plan a trip from A to B, we use abstraction to reduce the
state space, to be able to reason at a higher level. We
do not reason at the level of footsteps we need to take, but we first
decide on the mode of transportation. 
Hierarchical reinforcement learning tries to mimic this idea:
conventional reinforcement learning  works at the level of a single
state; {\em hierarchical\/} reinforcement 
learning performs  abstraction, solving subproblems in
sequence. Temporal abstraction is described in a  paper by
Sutton et al.~\cite{sutton1999between}. Despite the appealing
intuition, work remains to be done, such as learning hierarchies
automatically, and finding good benchmarks to show clear progress.

Flet-Berliac~\cite{flet2019promise} provides a recent
  overview of hierarchical reinforcement learning, noting that
  planning research into 
hierarchical methods showed that exponential improvements in
computation cost can be achieved with methods such as
hierarchical task networks~\cite{currie1991plan}, macro
actions~\cite{fikes1972learning},  and state abstraction
methods~\cite{knoblock1990learning}. The
promise of hierarchical reinforcement learning is summarized as: (1) to achieve long-term
credit assignment through faster learning and better generalization,
(2) to allow structured exploration by exploring with sub-policies
rather than primitive actions, and (3) to perform transfer learning
because different levels of hierarchy can encompass different
knowledge.

Rafati and
Noelle~\cite{rafati2018learning} learn subgoals in Montezuma's Revenge.
Schaul et al.'s~\cite{schaul2015universal} universal value function
approach suggests the use of function approximation for hierarchical
reinforcement learning.  Pang et al.~\cite{pang2019reinforcement}
describe experiments with macro actions in StarCraft.
%



 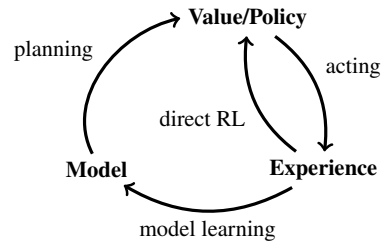
\begin{figure}[t]
 \begin{center}
   \begin{tikzpicture}[->,scale=1]
   \node (i) at (2,2)  {\bf Value/Policy};
   \node (j) at (3,0) {\bf Experience};
   \node (k) at (0,0) {\bf Model};


   \draw[bend left,->,very thick]  (i) to node [auto] {acting} (j);
   \draw[bend left,->,very thick]  (j) to node [auto] {model learning} (k);
   \draw[bend left=60,->,very thick,in=135]  (k) to node [auto] {planning} (i);
   \draw[bend left,->,very thick]  (j) to node [auto] {direct RL} (i);
\end{tikzpicture}
 \caption{Model-free and model-based methods~\cite{sutton2018introduction}}\label{fig:modelfreemodelbased11}
 \end{center}
 \end{figure}

\subsubsection*{Model-Free and Model-Based Reinforcement Learning}\label{sec:explicit}\index{implicit planning}\index{MuZero}\index{SimPLe}\index{PlaNet}
\index{model-based}\index{model-free}
Section~\ref{sec:modelbased} introduced model-free and model-based
reinforcement learning, the inner and the outer loop of
Fig.~\ref{fig:modelfreemodelbased11}~\cite{sutton2018introduction}. 
Model-free reinforcement learning is quite successful. Most work in
reinforcement learning is model-free (or {\em flat\/} reinforcement learning)
without a transition model, tree, self-play, or planning (see, e.g.,
\cite{anthony2019policy}). However, model-free reinforcement learning has low sample
efficiency.

Long-range credit assignment and hierarchical structures  are especially
difficult to fold into flat (model-free) function approximators; some
form of model,
abstraction, and strategic reasoning is likely to be necessary.
The option to use  exact planning methods
will also depend on the dimensionality of the problem. For 
high-dimensional problems approximation is a necessity;  for
low-dimensional problems exact methods will yield better answers. Can 
these be combined into a single architecture?

Network
architectures can be created that  combine planning and
learning.
RNN and LSTM
networks are well suited for modeling sequential problems. 
There are  studies on how look-ahead can be performed by neural
networks, obviating the need for the often slow MCTS planning
part~\cite{guez2018learning,guez2019investigation}. Schrittwieser et
al.~\cite{schrittwieser2019mastering} report on MuZero, a system that
combines model-based planning and model-free learning. MuZero learns
a latent model that is used during planning on-line, for board games as
well as Atari games.   Kaiser et al.~\cite{kaiser2019model} learn a model
for Atari games called SimPLe. Hafner et al.~\cite{hafner2018learning}
use planning in latent space to learn dynamics in a system called
PlaNet. This is an exciting research direction where more
can be expected~\cite{wang2019benchmarking,plaat2020model}.




\subsubsection*{Explainable AI}\index{explainable AI}\index{XAI}\label{sec:xai}
An important emerging research area in machine learning is explainable AI, or
XAI. Explainable AI is closely related to the topics of planning and
learning that we discuss in this book.  When a human expert gives a
recommendation, the expert can be questioned, and can explain the
reasoning behind the advice. This is a very desirable property of an agent. Most
clients receiving advice, be it financial, medical, or otherwise,
prefer to receive a well-reasoned explanation with their advice, to
see if they feel that they can trust the advice.

Most decision support systems that are based on symbolic AI can be
made to provide such reasoning. Decision trees~\cite{quinlan1986induction}, graphical models~\cite{jordan1998learning,lauritzen1996graphical}, and
search trees are eminently suited for following the decision points
of  advice.

Connectionist approaches such as deep learning  are more
opaque. However, the accuracy of deep learning approaches has assured
their popularity. The area of explainable AI aims to combine the ease of
reasoning of symbolic AI with the accuracy of connectionist
approaches~\cite{gunning2017explainable,doran2017does,browne2019strategic}.
Hybrid approaches combining
aspects of planning and learning are among the techniques that hold
promise. 
Soft decision
trees~\cite{frosst2017distilling,hinton2015distilling} and adaptive neural
trees~\cite{tanno2018adaptive} are ways to distill knowledge from
neural nets (that generalize well) into decision trees (that are highly
interpretable). These works build in part on research on model
compression~\cite{cheng2017survey,bucilua2006model}.\index{model
  compression}\index{adaptive neural trees}\index{soft decision trees}

Another area of interest is the intersection of knowledge representation and
reinforcement learning, also driven by the desire for decision support
where the reasoning behind a 
recommendation is also provided. This field is related to learning Bayesian
networks and belief
networks~\cite{heckerman1995learning,neapolitan2004learning,scutari2009learning,de2002ant,bellemare2013bayesian,teyssier2012ordering,brochu2010tutorial,asmuth2009bayesian}.\index{knowledge
representation}

\section{Practice}

We have now come to the end of a chapter that covered many
advanced topics, and provided pointers to even more research. Let us
turn to practice now, to implementations.

First are  questions to check your understanding of this
chapter. Each question is a closed question where a simple, one
sentence answer is possible.
\subsubsection*{Questions}
\begin{enumerate}
\item What are the differences between AlphaGo, AlphaGo Zero, and
  AlphaZero?
\item What is the dan-rating of Fan Hui, Lee Sedol, and Ke Jie?
\item Which networks does AlphaGo have, and which training methods?
\item Which networks does AlphaGo Zero have, and which training
  methods?
\item What is a double-headed network?
\item What is the error function that is trained in a double-headed
  value/policy net?
\item What is a TPU, and how is it different from a GPU?
\item What is the Elo rating of AlphaGo in 2016 (approximately: around
  2000, around 2500, or around 3000)?
\item Which three elements make up the self-play loop? (You may draw a
  picture.)
\item How is self-play recursive?
\item What is the network architecture of AlphaGo?
\item What is the network architecture of AlphaGo Zero?
\item What is tabula rasa learning?
\item How can tabula rasa learning be faster than reinforcement
  learning on top of supervised learning of grandmaster games?
\item What is the main result of the AlphaZero paper?
\item Name a top traditional Chess program.
\item Name a top traditional Shogi program.
\item Was the style of play of AlphaGo described by the Go
  community  as boring or as exciting?
\item What decorrelation efforts does AlphaGo use?
\item What decorrelation efforts does AlphaGo Zero use?
\item What decorrelation efforts does AlphaZero use?
\item Explain similarities between planning, learning, system 1, and
  system 2.

\item What is curriculum learning?
\item What is transfer learning?
\item What is multi-task learning? Does AlphaGo have multi-task
  learning? Does AlphaGo Zero have multi-task learning?
  \item What is meta-learning?
\item Describe three kinds of recursion in AlphaGo Zero-like self-play?
\item What is hyperparameter optimization, and how does it apply to
  reinforcement learning?
\item What is neural architecture search, and how does it apply to
  reinforcement learning?
\item What is hierarchical reinforcement learning? Name three approaches.
\item Why are games with sparse rewards difficult for reinforcement learning?
\item Why are domains with continuous action spaces difficult for
  reinforcement learning?
  \item What is explainable AI?
\end{enumerate}

\subsection{Exercises}
The AlphaZero General (A0G) code base is well suited for experimentation with
self-play. We use it in the exercises of this chapter.  A0G can be found 
at \href{https://github.com/suragnair/alpha-zero-general}{A0G}.\footnote{\url{https://github.com/suragnair/alpha-zero-general}} Note
that you also have to install TensorFlow and Keras. They are
at \href{https://www.tensorflow.org}{tensorflow.org}.

\begin{enumerate}
\item Install and run A0G Othello 6x6. Play against the computer to
  see if everything works. Look at the source code, especially at
  \verb|Coach.py|. Can you see a relation with the code in
  Listing~\ref{lst:selfplay}? What is an episode? 
\item Observe the learning processes of the different games Tic Tac
  Toe, Connect4, and Gobang. See how self-learning progresses.
\item Vary learning rate, vary MCTS Nsims, vary iterations. How is
  performance impacted? How is learning time impacted?
\item A stable self-play process in A0G depends in part on the amount of
  exploration in MCTS. Vary the Cpuct constant, and observe if
  training is impacted adversely.
\item Try different net architectures. How is performance impacted?

\item Take a traditional alpha-beta Othello program, for
example \href{http://dhconnelly.com/paip-python/docs/paip/othello.html}{Othello}.\footnote{\url{http://dhconnelly.com/paip-python/docs/paip/othello.html}} See
if A0G can learn a strong enough player to beat it.
\item Choose a simple game of your own preference, and implement it in A0G. Does it learn quickly?
\item *Implement Chess input and output layers in A0G, learn it tabula rasa, and play
  against Sunfish. You may want to have a look at LeelaZero Chess or ELF OpenGo for inspiration.

\item Play with tensorboard to look inside the brain/training process
  and try to understand the training process.
\item Implement hyperparameter tuning of A0G parameters.

    \item *Write a hyperparameter optimizer for A0G, trying different
    net architectures. This is challenging and quite computationally
    intensive. How can you reduce the computational demands?
\item Can single-agent optimization problems benefit from self-play?
Implement the 15 Puzzle in A0G, and Travelling Salesperson. See
also~\cite{laterre2018ranked,xu2019learning}.
\item See if transfer learning of 4x4 Othello, to 6x6 to 8x8 works, and
works faster. Are you seeing a curriculum learning effect?
\end{enumerate}

\section*{Summary}
The challenge of this chapter was to combine planning and learning, to
create a self-playing self-learning system.
We described the performance of AlphaGo,
AlphaGo Zero, and AlphaZero in depth. 
There was great interest in Go and AI from
science and society.

Issues of coverage, correlation, and convergence (the deadly triad)
were mostly overcome by the AlphaGo team through  science and
engineering, for which they received many accolades.

The main conceptual innovation of this
chapter is 
self-play,  a combination
of  planning and a training loop,  a combination of MCTS and
deep reinforcement learning. Self-play (Fig.~\ref{fig:selfplay}) takes evaluation (system 1) and planning
(system 2) and adds a third item, a learning loop, creating a
self-learning search-eval system. 

The self-play loop uses MCTS to generate high-quality examples,
which are used to train the neural net. This new neural net is then used in a
further self-play iteration to refine the network further (and again,
and again, and again).
Alpha(Go) Zero thus learns starting at zero knowledge, tabula rasa. And, just
as impressively, the same self-play architecture can learn tabula rasa
Chess and Shogi as well, defeating decades of heuristic refinement.

Self-play makes use of many reinforcement learning techniques. In
order to ensure stable, high performance learning, exploration is
important. MCTS is used for high-quality  sampling. Exploration
parameters in MCTS are set high, and convergent training is achieved by
a low training rate. Because of these parameter settings many games
have to be played. The computational demands of self-play are large.

MCTS has been changed significantly in the self-play setting. Gone are the
random play outs that gave MCTS the name Monte Carlo, and much of the
performance is due to a high-quality policy and value approximation
residual network. (The name Deep Q Tree Search might be more appropriate.)

Reproducibility in science is of great importance to
progress~\cite{henderson2018deep}. The impact of benchmarks and open-source algorithms in this respect is important. The reimplementations
of the self-play results in A0G, ELF OpenGo, PhoenixGo, and Leela will stimulate
experimentation and are
of  importance for progress towards general learning systems.

In this chapter we discussed enhancements of self-play, to understand
why it works, and how its performance can be enhanced further.
Self-play learns faster than a database approach. Self-play
creates a natural curriculum learning situation, where the examples are
ordered from easy to hard. We discussed the links between curriculum
learning and transfer learning. The self-play success 
stimulates research interest in curriculum and transfer learning.

We also discussed further enhancements to self-play. Optimization of
the network architecture and the training hyperparameters are 
active fields of research.
Another field in which planning and learning come together is
explainable AI.

\section*{Historical and Bibliographical Notes}

Self-learning systems are exciting. Self-learning is a holy grail of artificial intelligence:
the emergence of intelligent behavior out of basic
interactions~\cite{leibo2019autocurricula}.  
Machines that teach themselves how to play have intrigued researchers since the start of
game playing research. 

The work on AlphaGo is a landmark achievement in AI. Every AI
researcher is encouraged to study it well.
The primary sources of information for AlphaGo are the three AlphaGo
papers by Silver et
al.~\cite{silver2016mastering,silver2017mastering,silver2018general}. The
papers contain a lot of information, and
can be a bit overwhelming at times, especially the detailed methods sections. Many blogs and popular press
articles have been written about AlphaGo that might be  more
accessible.

Do not forget to watch the AlphaGo movie. Although it does not go too
deep into the details, it is a fascinating account of the match, and
its significance in AI. It is
at \href{https://www.alphagomovie.com}{AlphaGo
  Movie}.\footnote{\url{https://www.alphagomovie.com}}
There are also  explanations on YouTube, for example at
\href{https://www.youtube.com/watch?v=MgowR4pq3e8}{AlphaZero
  Explanation}.\footnote{\url{https://www.youtube.com/watch?v=MgowR4pq3e8}}

A book devoted to building your own state-of-the-art self
learning Go bot is \emph{Deep Learning and the Game of Go} by Pumperla and Ferguson~\cite{pumperla2019}.

AlphaGo was not created in a vacuum. There is a large body of research
of which we will now discuss some of the more notable papers.
Pre-MCTS, there are works on shape and move
prediction~\cite{stern2006bayesian,berlekamp1994mathematical}.
In the contexts of MCTS, many researchers worked on combining MCTS with
learned patterns, especially to improve the random roll outs of MCTS. Supervised learning on grandmaster games
was used to improve playouts and also to improve UCT selection. Gelly and Silver  et 
al.\ published notable works in this
area~\cite{gelly2006modification,silver2007reinforcement,gelly2008achieving},
and others~\cite{benson1976life}.

Patterns are also important in Othello. Around 1995 very good results were achieved by
Logistello~\cite{buro1995logistello}\index{Othello}\index{Logistello}. Logistello
used logistic regression on patterns. The author also successfully
introduced probabilistic depth reduction methods~\cite{buro1997experiments}.
Graf et al.~\cite{graf2015adaptive} describe experiments with adaptive
playouts in MCTS with deep learning.
Convolutional neural nets were also used in Go by Clark and
Storkey~\cite{clark2014teaching,clark2015training}, who had used a CNN for supervised learning
from a database of human professional games, showing that it
outperformed GNU Go and scored wins against Fuego.
In Chess, the program DeepChess\index{DeepChess} by
David et al.~\cite{david2016deepchess} performed end-to-end deep
learning in Chess, achieving results on par with
two  alpha-beta-based programs.
In Checkers,
Blondie24~\cite{chellapilla1999evolving,fogel2001blondie24} used a
network to evolve board features, also achieving good results.

Tesauro's success inspired many others to try temporal difference learning.
Wiering et al. and Van der
Ree~\cite{wiering2010self,van2013reinforcement} report on self-play
and TD learning in Othello and Backgammon.
The program Knightcap~\cite{baxter1999knightcap,baxter2000learning}\index{Knightcap} and
Beal et al.~\cite{beal2000temporal} also use temporal difference learning on evaluation
function features. Veness et al.~\cite{veness2009bootstrapping} use
self-play to learn evaluation function weights. They name their
approach {\em search bootstrapping\/} and, referencing Samuel's Checkers program, describe it as follows: ``The idea of search
bootstrapping is to adjust the parameters of a heuristic evaluation
function towards the value of a deep search.'' This sentence captures
the idea of self-play, as used in AlphaZero, quite well.
Arenz~\cite{arenz2012monte} applied MCTS to Chess.

Heinz also reported on self-play experiments in
Chess~\cite{heinz2000new}. 
In other games, Guo et al.~\cite{guo2014deep} report on the use of MCTS to play Atari
games; their combination of MCTS and DQN outperforms DQN on Atari.

In Poker\index{Poker} impressive results have been
published~\cite{bowling2015heads,moravvcik2017deepstack} with
combinations of planning and learning algorithms, most notably
counterfactual regret minimization and self-play.

Since the AlphaGo results many  other applications of machine learning
have been
shown to be successful.
Complex combinatorics is of
interest to theoretical physics~\cite{ruijl2014hepgame,paparo2014quantum}, chemistry, and pharmacology, specifically for
retrosynthetic molecular design~\cite{segler2018planning} and drug design~\cite{van2011compound}.   Image
recognition is of interest to experimental physics~\cite{baldi2014searching}, biology~\cite{nezhinsky2010pattern}, and medicine~\cite{de2018clinically}.
Decision support is of interest to medical decision
making~\cite{hsieh2018comparison}. Reasoning and image analysis are of
interest to smart industry~\cite{giotis2001low,jin2014subgroup}, linguistics, and humanities~\cite{stork2018priming}, and
strategic reasoning is of interest to  law,
international relations, and policy analysis~\cite{baarslag2013evaluating,lin2010can,van2011non,castelfranchi1999deliberative}. Finally, the
applications of robotics and reinforcement
learning are of great interest to the cognitive and social
sciences~\cite{johnson2011beyond,broekens2009assistive}. Of particular
interest are cheminformatics including drug design~\cite{segler2018planning}, theoretical
physics, cognitive robotics, and law and negotiation.

\index{Pieter Abbeel}
For more information on hyperparameter
optimization,
see~\cite{bergstra2011algorithms,hutter2011sequential}. A reference
work on 
evolutionary strategies  is~\cite{back1993overview}.
To learn more about 
curriculum learning, see~\cite{weng2020curriculum,bengio2009curriculum,matiisen2017teacher}.
For more information on transfer learning,
see~\cite{pratt1993discriminability,thrun2012learning}.


Ever since his paper on reinforcement learning for aerobatic
helicopter flight~\cite{abbeel2007application}---there are videos on the web---Pieter Abbeel's
work in robotics  and in reinforcement learning has been  
influential.

The study of model-based reinforcement learning is an active
area of research. Recent experiments are reported
in~\cite{silver2017predictron,tamar2016value,farquhar2018treeqn,racaniere2017imagination,kaiser2019model,hafner2018learning,plaat2020model}.

Meta-learning and transfer learning are related, and active areas of
research~\cite{frans2017meta,schweighofer2003meta,vilalta2002perspective,gupta2018meta,gupta2018unsupervised,raina2007self,soekhoe2016impact,toubman2015transfer,huisman2020deep}.

Hierarchical reinforcement learning is an active field of research
with a long history. See for further reading for example~\cite{kaelbling1993hierarchical,perkins1999using,frans2017meta,makar2001hierarchical,barto2003recent,bakker2004hierarchical,singh1992reinforcement,kulkarni2016hierarchical,levy2018hierarchical,stulp2011hierarchical,kaelbling2010hierarchical,vezhnevets2016strategic}.\index{hierarchical
  reinforcement learning} Applications of hierarchical
reinforcement learning to self-play remain to be explored, see also~\cite{sukhbaatar2018learning}.

Readable and insightful introductions into the field of abstraction in
deep representation learning are~\cite{bengio2013representation,lecun2015deep}.
A good place to start for network visualization is the ZFnet paper~\cite{zeiler2014visualizing}.
Wang et al.~\cite{wang2019alternative} study the optimization
target of a dual-headed self-play network in AlphaZeroGeneral.


\chapter{Conclusion}\label{chap:conclusions}\label{chap:agi}\label{chap:fast}

Computers are now faster at learning Go than humans: it takes AlphaGo Zero
 days  to reach world champion level, for humans this takes years.

This chapter is about the future. We summarize the lessons learned
in this book, to reflect upon where AI may go next.

In the previous chapters we have discussed in depth the methods that
have been used to achieve artificial intelligence in games. The focus has
been on single games, on narrow
intelligence.

The purpose of
this chapter is to see how narrow intelligence  can be extended to
general intelligence, or at least to some extent. We will
approach this goal in three steps. First, in
Sect.~\ref{sec:genmeth}, 
we will analyze the methods of the previous chapters, to find common
threads. Second, in Sect.~\ref{sec:future}, we
will discuss new problem domains that expand our horizon beyond zero-sum
perfect-information games. Then, in Sect.~\ref{sec:humanintel}, we
will compare human and artificial 
intelligence, and look into future methods of artificial intelligence.


\section*{Core Problems}
\begin{itemize}
\item How are artificial and human intelligence related?
\item What games can be used in our quest towards general intelligence?
\item What are  essential methods of intelligent decision making in
  games?
\end{itemize}

\section*{Core Concepts}
\begin{itemize}
\item Specialized intelligence and  general  intelligence 
\item Sample efficiency, transfer learning, and population-based methods
\item Multi-agent, imperfect-information games
\end{itemize}

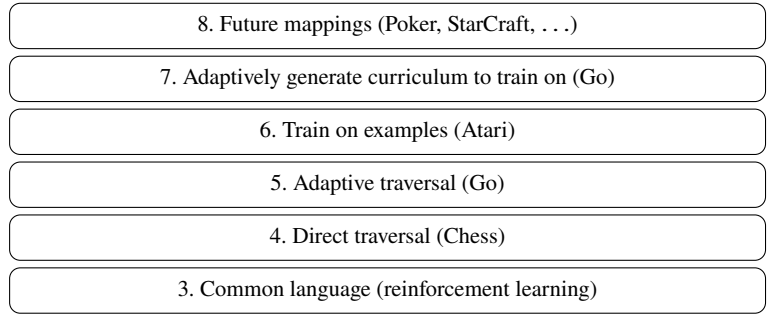
\begin{figure}[t]
\begin{center}
\begin{tikzpicture}[>=triangle 45,
  desc/.style={
		scale=1.0,
		rectangle,
		rounded corners,
		draw=black, 
		}]

  \node [desc,minimum width=10cm,minimum height=0.6cm] (rl) at
  (0,0.7) {3. Common language (reinforcement learning)};
  \node [desc,minimum width=10cm,minimum height=0.6cm] (mm) at
  (0,1.4) {4. Direct traversal (Chess)};
  \node [desc,minimum width=10cm,minimum height=0.6cm] (mcts) at
  (0,2.1) {5. Adaptive traversal (Go)};
  \node [desc,minimum width=10cm,minimum height=0.6cm] (dqn) at
  (0,2.8) {6. Train on examples (Atari)};
  \node [desc,minimum width=10cm,minimum height=0.6cm] (self) at
  (0,3.5) {7. Adaptively generate curriculum to train on (Go)};
  \node [desc,minimum width=10cm,minimum height=0.6cm] (future) at
  (0,4.2) {8. Future  mappings (Poker, StarCraft, $\ldots$)};

\end{tikzpicture}
\caption[Stack of chapters]{Stack of chapters, with (from bottom to top) increasingly advanced mappings from
problems to paradigms}\label{fig:stack12}
  \end{center}
\end{figure}


\section{Artificial Intelligence in Games}\label{sec:genmeth}

In this section we will review the methods that we have seen in the
book so far. We will analyze them to see how they created an
artificial kind of intelligence. We will look for common aspects in
the hope that we can extend them to create new methods, to create an
even more human kind of artificial intelligence, allowing better human-computer
interaction and better human-computer understanding.

\subsection{Summary of Paradigms}
Figure~\ref{fig:stack12} shows the structure of  this
book  as a stack of chapters, with the paradigms building on top of each
other. At the bottom is the reinforcement learning layer (3) that
provides a common language and formalism for  this book.

\subsubsection*{Planning}
The paradigms start with being closely related to the state space. The
methods in the second
layer (4) in Fig.~\ref{fig:stack12}
directly traverse the state space, enumerating the states one by
one. Basic undergraduate  planning methods such as depth-first and
breadth-first search are used~\cite{russell2016artificial}.

\subsubsection*{Heuristics}
When the state space is larger, the time complexity of full
enumeration becomes prohibitive, and  manually constructed features
are introduced, based on heuristics. The heuristics are used in both 
backward and  forward fashion: in evaluation functions
(backward propagation) and in transition functions or move generation
(forward pruning). Through these heuristics, depth-first and breadth-first search are
transformed into best-first search: the heuristics guide node expansion
towards parts of the tree that the heuristics deem best.

Heuristic planning combines the traversal algorithms with these
heuristics and has been quite successful. This field of artificial
intelligence has also been called heuristic search, or simply
heuristics~\cite{pearl1984heuristics}. The Deep  
Blue--Kasparov match in 1997 marked the success of this approach.

\subsubsection*{Adaptive}
Already with the heuristics  enhancements arrived that softened the rigid depth-first
fixed-depth expansion, when transpositions and search extensions were
introduced, amongst many, many others. Going towards the next layer
(5)  in Fig.~\ref{fig:stack12}, we arrive at the adaptive paradigm.
 Node selection rules such as
UCT based on  exploration   and exploitation
provide a formal basis and greatly improve on the  ad hoc forward
pruning idea in many new domains.

\subsubsection*{Sampling}
Furthermore, an alternative, non-minimax, backup rule is
introduced. Averaging, together with a sampling evaluation function
and UCT selection, provides a new search paradigm. The new paradigm works in
domains with large branching factors where no efficient heuristic
features have been found.

Together these methods constitute MCTS, which took the field of
heuristic search into the realm of reinforcement learning, where
methods are adaptive (i.e., they learn) and where methods balance exploration with exploitation.
Through MCTS the application domain of  combinatorial
methods has expanded greatly.

\subsubsection*{Supervised Generalization}
Still, the states that were explored in the state space are treated
separately by MCTS. Commonalities between states, such as shared features, are
not exploited explicitly. Even worse, in MCTS, in its most basic form, features
of states are deliberately ignored, since node evaluation is solely based on
random playouts.  Moving to the next layer (6) in
Fig.~\ref{fig:stack12}, this fact already suggests a further
paradigm. This new paradigm, function approximation, opens up the states,
looks at individual features, and uses methods to learn common
features by training on  examples of states.

\subsubsection*{Reinforcement Learning Generalization}
The field of deep
learning has introduced methods to learn features from examples,
balancing generalization and overfitting. In supervised learning a
fixed database of examples exists from which features must be
learned. In reinforcement learning the training process generates the
new examples as part of the training process. Because of this, there is a high degree
of correlation between subsequent examples. In reinforcement learning special care must be
taken to explore a sufficiently large and diverse part of the state
space, since otherwise the training gets stuck in a local optimum, or
the training process itself may even diverge (due to the deadly triad).

\begin{figure}[t]
    \centering{\begin{tikzpicture}[>=triangle 45,
  desc/.style={
                scale=0.9,
		rectangle,
		rounded corners,
		draw=black,
		}]
  \node [desc] (search) at    (0,2) {{\bf search}};
  \node [desc] (eval) at    (3,2) { {\bf eval}};
  \node [desc] (train) at       (1.5,0.5) {{\bf train}};

  \draw (search.south) edge[semithick,->,out=270,in=180,looseness=1] node[left] {\small examples}  (train.west);
  
  \path (eval.south) edge[loop,semithick,<-,in=0,out=270,looseness=1]
  node[right] {\small  \ new  net}  (train.east);
  \path[desc] (search.east) edge[semithick,loop above,<-,min distance=7mm,in=160,out=20,looseness=1] node[above] {\small \em reward}  (eval.west);

\end{tikzpicture}
     \caption{Simplified self-play loop}\label{fig:selfplay12}}
\end{figure}
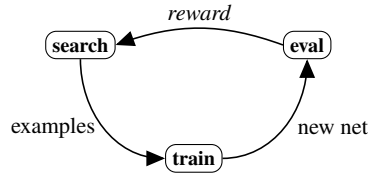

\subsubsection*{Self-Play}
In the next layer (7) in Fig.~\ref{fig:stack12}, self-play is shown,
which uses the generalization methods in a  self-play
setting. Here, the neural network is combined with MCTS. It is used
for node selection, incorporating it in UCT, and for evaluation,
replacing random playouts. In addition, the game playing program is
used to generate the examples for 
the generalization process, which is used to train the evaluation
features of the neural network, which is used by the program to
generate the next (better) set of examples, etc.

Self-play is a highly recursive approach in which many aspects of the
algorithm feed upon themselves (see also Fig.~\ref{fig:selfplay12}). Stability issues are plentiful, and
various exploration methods must be used (and tuned well) for
self-play to work. When it works, we find that it is quite an
efficient way of training features, because of curriculum learning.

Other application domains and learning methods are shown in the next layer (8).


\subsection{Summary of Methods}
There are a few methods, or principles, that occur frequently in the paradigms of reinforcement
learning  that we have seen 
in different shapes and forms in this book.
Let us now summarize these common principles.

\subsubsection*{Recursion and Bootstrapping}
First of all is recursion. Recursion is the calling of a  function by
the same function with different parameters.
It is used to traverse trees and graphs, it is the basis of Bellman's
dynamic programming, and it is also behind the idea of self-play:
reusing parts of the algorithm in a slightly different way to achieve
your goal. Recursion is a standard way of
traversing a state space, for example, to search for some optimal
value, or for traversing the weights in a network to update them
according to an error gradient.

Bootstrapping determines a state's value by making use of its previous
values. Bootstrapping is often
implemented using recursion, and dynamic programming makes use of this
principle. Bootstrapping is at the basis of reinforcement learning, in Bellman's
equation, defining the value of a state based on the discounted value of future
actions. Minimax, the principle of using the same model  for the
player and the  opponent, also finds the value at the root of the game
tree by bootstrapping on the tree of child values.

Finally, bootstrapping and  recursion feature three times in
AlphaZero self-play, in the move selection of MCTS, in the generalization
of the network  for example states, and in the curriculum of games that are
generated in a tournament against an identical  copy of the player.

\subsubsection*{Optimization}
This brings us to the next method that is used in all paradigms: optimization. Optimization is,
again, everywhere. Optimization is used to find the best value, to find the
best action, to find the best weights, and to find a global optimum. 

Optimization is the principle of choosing the best item out of a set,
preferably using some smart means or extra information to do so
efficiently. All reinforcement learners and all game playing programs are nothing more than
advanced optimizers, using many times the principle of optimization
to select best actions, best states, best examples, best network
architectures, and best training hyperparameters.


The principles of optimization and of
bootstrapping work together in the search-eval architecture.
Other
examples exist, many exact algorithms use the principle of
optimization, such as SARSA, A*, and depth-first search, and many
other algorithms exist for the other three cells as well.

Optimization methods can also be used to approximate, for example, by
using the features of  states.





\subsubsection*{Exploration}
Exploration is essential for stable deep reinforcement learning, to ensure broad
coverage of the state space, to break through correlated subsequent states
in reinforcement learning loops, and to ensure convergent training
methods. Exploration is of crucial importance in MCTS, DQN, and PPO,
and it is at the core of efficient simple reinforcement
learning (such as Q-learning)  and
of stable deep reinforcement learning and  self-play. 

\subsubsection*{Search-Eval}
The search-eval architecture provides a framework for the trade-off
between exact planning
algorithms and approximate learning algorithms. In two-player games the use of
a combination of exact and approximate methods has proven to be essential for
success, and all programs follow this model-based architecture. In self-play an
extra learning loop is added to search-eval, to allow for the evaluation network to
bootstrap itself on the search-eval examples.

Deep learning is a model-free method 
that takes many samples to learn the features. It is not very sample
efficient because it lacks a model, although curriculum learning
helps since examples are easy to learn. Deep learning is important
because it is, in
contrast to heuristic planning, able to learn an evaluation
function. Planning and learning can be sequenced together to make a
powerful model-based/model-free 
self-play system (which is not very sample efficient).


\subsubsection*{End-to-End}
A final principle of this book is the progression of compute power,
allowing larger and larger problems to be addressed. Where 60 years
ago Samuel
could only train coefficients of hand-crafted features, 30 years ago TD-Gammon
could train end-to-end temporal difference learning on a small network,
 now  AlphaGo can do tabula rasa end-to-end learning with a deep
network. Being able to train end-to-end allows us to forgo
hand-crafted features, with all  their inherent errors and biases, achieving
high performance and true (but still narrow) artificial intelligence.

\subsection{On the Importance of Enhancements}
The search-eval architecture provides a conceptual framework for game
playing programs. To achieve a high
level of performance, the
enhancements, however, are as important. 
High-performing game playing programs from 1950-2006 were based on
alpha-beta. Without transposition tables and
singular extensions there would be no Deep Blue. Without fast playout
patterns there would be no AlphaGo, without specific input layers no
AlphaZero, without ReLU, dropouts, 
and GPUs, no deep learning Turing award, and without the
replay buffer there would be no success in Atari. The
enhancements are the result of deep understanding by the researchers of their
domains. Understanding that is more specialized, and less general, than
abstract principled frameworks such as minimax, reinforcement learning, or function
approximation.

In addition to the elegant conceptual frameworks, deep,
dirty, domain-specific understanding is necessary for progress in this
field~\cite{schaeffer2001unifying}. 

\subsection{Towards General Intelligence}\label{sec:agi}


Let us revisit the  problem statement from the
Introduction. 
\begin{quotation}
{\em   What are the machine learning methods that are
used in Chess and Go to achieve a level of play stronger than the strongest humans?}
\end{quotation}
Various reinforcement learning methods  have beaten human champions in
Chess and Go,
ranging from heuristic planning, adaptive sampling, function
approximation, to self-play. All methods fit in the classic
search-eval architecture, and all use variants of the principles of
bootstrapping and optimization. Heuristic
planning  can achieve levels of play far
surpassing human ability in Chess and Checkers. We have also seen how
to combine planning 
and learning into a self-play system that learns to play  from scratch,
achieving even {\em higher\/} levels 
in
Chess, Shogi, and Go.

Did we achieve intelligence? For these three games, we
certainly passed the Turing test.\footnote{Actually, since we far
  surpassed human play, an argument can be made that in doing so we
  {\em failed\/} the Turing test, since no human can  play at this level.}
Systems were created that behave at a level
for which a human would need super human intelligence. The
intelligence is, however, single-domain intelligence. AlphaGo cannot
play Shogi, and it cannot think of an entertaining joke.


Joining the two fields of symbolic AI and connectionist AI, AlphaZero showed that the methods
 generalize to three games, hinting at artificial general
intelligence. However, it is more precise to speak of training
three almost identical systems to become three different systems specialized
in their own kind of special intelligence (since the  net, once  trained for
Go, cannot play Chess). (GGP systems can play general games, but
do not achieve performance close to  AlphaZero;  Sect.~\ref{sec:ggp}.)


\section{Future \emph{Drosophila}s}\label{sec:future}\index{\emph{Drosophila Melanogaster}}
The reinforcement learning results in  Atari and Go have
inspired much further research. The use of concrete games, the 
{\em Drosophila}s, has stimulated progress in game playing
greatly. In addition to the usual two-person zero-sum perfect-information games,
researchers have looked for 
games that capture other elements of real life, such as 
non-zero-sum (collaboration) games, imperfect
information,  and multi-agent games. In this section we 
will review some of these aspects.

\subsection{Real-Time Strategy Games}\index{imperfect information}\label{sec:rts}
Imperfect information games have been studied extensively in reinforcement
learning~\cite{sondik1978optimal,kaelbling1998planning}. 
Multi-agent imperfect-information \emph{Drosophila}s can be card games (such as  
Bridge and Poker) and real-time strategy games (such as Defense of
the Ancients (DOTA) 2 and
StarCraft). 
StarCraft has been studied for some
time\index{StarCraft}~\cite{vinyals2017starcraft,togelius2010multiobjective,uriarte2015benchmark},
as have DOTA 2~\cite{drachen2014skill,bansal2017emergent,OpenAI_dota} and other
RTS games \cite{preuss2010towards,lara2013review,tekofsky2015past,jaderberg2019human}.\index{DOTA 2}
Easy-to-use Python interfaces have been created for StarCraft and DOTA
2. These games are highly challenging for current methods. They are multi-player,
imperfect information, have large and continuous action spaces and very
large and sparse 
states spaces, and  long-time  credit assignment challenges. Impressive
results are already being achieved. Leibo et
al.~\cite{leibo2019autocurricula} have written a manifesto for further
research into curriculum learning in multi-agent games.

Vinyals et el.\ have success  in StarCraft II with their program AlphaStar
~\cite{vinyals2019grandmaster} which uses multi-agent
reinforcement learning in combination with supervised learning for
subparts of the game. The notion of self-play is extended to a group
of agents, the league. Using population-based methods agents try to improve the strategies of others in
their own league, so that the collective becomes stronger.

Jaderberg et al.~\cite{jaderberg2018human,jaderberg2019human} describe an impressive
result of a multi-level learning system. Their system uses
a variant of reinforcement learning that learns a population of
agents, in a multiplayer real-time strategy capture-the-flag game,
Quake III. The computational demands of their approach are mitigated by
not using an MCTS search and using relatively simple and fast RNNs
(see also Sect.~\ref{sec:rts}). 


OpenAI is  performing research on cooperation with OpenAI Five in
DOTA~2,
again demonstrating that self-play reinforcement learing can achieve
super-human performance in complex environments~\cite{berner2019dota}. 

\subsection{Negotiation and Collaboration}\index{Diplomacy}
Non-zero-sum games are close  to everyday life situations. Many  real-world
situations include negotiation and collaboration and  offer win/win
situations. AI has studied negotiation and collaboration extensively.

Negotiation and collaboration   also occur in card games. Poker and
Bridge have large action spaces and a large state 
space. Poker\index{Poker} has been researched for some
time,
and for simpler (two-player) versions of the game successful programs
have been
written~\cite{bowling2015heads,moravvcik2017deepstack,brown2018superhuman,koller1997representations}. Recently,
success has also been reported in multi-player Poker.   
Cooperation and negotiation are central elements of  Bridge\index{Bridge}, and research is
continuing~\cite{smith1998computer,ando2003cooperation}. 
Negotiation
has also been studied for some
time~\cite{jonker2017automated,kraus1994negotiation,jonker2007agent},
including negotiation in games such as
Diplomacy
~\cite{paquette2019no,anthony2020learning,de2015negotiations,de2017automated,de2017d,de2018challenge,kraus1988diplomat,ribeiro2009darkblade}. 

Let us now have a closer look at two games, Poker and StarCraft.

\subsection{Poker}\label{sec:poker}\index{Poker}
Poker is an excellent game to show what progress has been made in 
imperfect-information games.

Poker has been studied for some time in artificial intelligence, and
computer Poker championships 
have been conducted
regularly~\cite{billings2002challenge,billings2004game,gilpin2006competitive,rubin2011computer,bard2013annual,bowling2015heads,billings2004game,gilpin2006competitive,bowling2009demonstration,sandholm2010state,moravvcik2017deepstack}. 
No-limit Texas hold’em is
the most popular form of poker. The two-player 
variant (called Heads Up) prevents opponent collusion.  Heads-up
no-limit Texas hold’em (HUNL) has been the primary benchmark for 
imperfect-information game play for several years.

Poker features face-down cards, whose value is unknown to the other players.  The face-down cards constitute hidden
information. This hidden information  creates a large number of possible states, making Poker a game that is far more complex than Chess or
Checkers. There are many variants of Poker; the state space of HUNL is
reported to be 
$10^{161}$, so simply enumerating the state
space to find an optimal policy is infeasible~\cite{brown2018superhuman}. A further complication
is that a player reveals information during the course of
play. Therefore, an AI  must  balance 
actions, so that the opponent does not find out too much 
about the cards that the AI has. Players may be misled through bluffing, i.e., deliberately
betting a large sum to make the opponent believe that one has stronger
cards than one has. 

In 2018, one of the top programs, Libratus, 
defeated top human professionals in HUNL in a 20-day,
120,000-hand competition featuring a \$200,000 prize
pool. We will now describe this program in more details, based in~\cite{brown2018superhuman}. Libratus consists of three main modules.

The first module computes a smaller version of the game, and then computes game-theoretic
strategies. The algorithm used is a variant of counterfactual regret
minimization (CFR). CFR is an algorithm  to find
 Nash equilibrium strategies that is widely used in computer Poker~\cite{zinkevich2008regret}. CFR is based on multi-armed bandit theory.
CFR is a recursive algorithm that  starts with a random strategy.\index{CFR}  It then
simulates playing games against itself.  After every game, it improves
its decisions.  It repeats this process many times, constantly
improving the strategy.  As it plays, it 
comes closer to an optimal strategy, a
strategy that can do (on average) no worse than tie against any opponent.
CFR generates strategies that minimize exploitability, which means
that it plays conservatively. CFR algorithms do not exploit imperfect
play from opponents in a way that a human would.
From a game-theoretic  point of view, a perfect
strategy is one that cannot expect to lose to any other player's
strategy. A perfect strategy in the presence of
suboptimal players who have weaknesses would be one that exploits
those weaknesses.

 When a later part of the game is reached
during play, the second module of Libratus  constructs a finer-grained
abstraction for that subgame and solves it in real time.

 The third module  is the self-improver, which enhances the
first policy. It 
fills in missing branches in the first version of the game and computes
a game-theoretic policy for those branches. Computing this in advance
is infeasible.
Libratus uses the opponents’ actual moves to suggest where in the game
tree this  is worthwhile.

In the experiment against 
human players Libratus analyzed the bet
sizes that were most heavily used by its
opponents  during each day of the competition. On the
basis of the frequency of the opponent bet, the program  chose some bet
sizes for which it would try to calculate a response overnight. Each
of those bet sizes  was then added to the  strategy, 
together with the newly computed policy following that bet
size. In this way, the program was able to  narrow its gaps
as the competition proceeded.

Interestingly, Libratus was originally based on heuristics,
abstraction, and game-theoretical  algorithms, and not on deep
reinforcement learning. However, later work introduced deep 
counterfactual regret minimization~\cite{brown2018deep}, a form of CFR
that approximates CFR equilibria in large games, using a 
7 layer neural network. This algorithm
is also the basis for Pluribus, the later multi-player 
program, defeating top players in six-player
Poker~\cite{brown2019superhuman}.\index{Pluribus}  
Pluribus uses a form of self-play in combination with search.

Another top program, DeepStack, does
use randomly generated games to train 
its deep value function network~\cite{moravvcik2017deepstack}. In
Poker we see
two  different approaches achieving success.


\subsection{StarCraft}\label{sec:starcraft}\index{StarCraft}\index{AlphaStar}
StarCraft is a multi-player real-time strategy game of even larger
complexity~\cite{ontanon2013survey}. It is  a good test bed for AI
research, since it
features decision making under uncertainty, spatial and temporal
reasoning, collaboration and competition, opponent modeling, and
real-time planning. The companies DeepMind and Blizzard have released a Python
interface to stimulate research in
StarCraft~\cite{vinyals2017starcraft,vinyals2019grandmaster}.  

The state space of real-time strategy games is larger
than traditional board games such as Chess, Checkers, or
Go. The state space has been estimated to be on the order of
$10^{1685}$, a very large number, even for this book~\cite{ontanon2013survey}. The number of actions in each state  is also large. Standard 
exact planning methods are not feasible, and need to be augmented with approximate methods to be effective.
Planning in real-time strategy games has multiple levels of
abstraction (see hierarchical reinforcement learning in
Sect.~\ref{sec:hier}). At a high level of abstraction, long-term
planning is needed; at a low level of abstraction, tactical battles
must be fought and individual units must be moved.

StarCraft  offers opportunities to explore many challenging new
methods in reinforcement learning. First, it is an
imperfect-information game. The terrain map is partially visible. A
local 
  camera  must be  moved  by the player to
  acquire information. There is a ``fog-of-war'' that obscures
  unvisited regions. The map must be explored proactively  to
  determine the opponents’ state.
  Second, StarCraft is a multi-agent game in which several
  players compete for influence and resources. StarCraft is also multi-agent
  at another level. Each player controls many units that need
  to cooperate to achieve the player's goal.
  Third, the action space is large and diverse. There are around 108
  actions for a player, which can be selected with a point-and-click
  interface. There are many 
  different types of units and buildings, each type with its own local
  actions. Furthermore, the legal actions vary during play as it
  advances through possible technologies. Finally, games
  are long. They last for thousands of states and actions. Decisions
  early in the game, such as which units to build, have 
  consequences that will be seen much later in the game when
  armies meet in battle. There is a long distance in credit
  assignment, and exploration is important in StarCraft~\cite{ontanon2013survey}.

In a series of test matches held in December 2018, DeepMind's AlphaStar
 beat two top players in  two-player single-map matches, using a different
 user interface. The neural
network was initially trained by supervised learning from anonymized
human games, that were then further trained by playing against other
AlphaStar agents, using a  population-based version of self-play
reinforcement learning~\cite{vinyals2019grandmaster}.

\subsection{General Game
  Playing}\label{sec:ggp}\index{GGP}\index{General Game
  Playing}\index{\emph{Drosophila Melanogaster}}
As has already been mentioned a few times, another approach to studying generality in game intelligence is 
general game playing (GGP). GGP takes a  different approach, not using
concrete existing games, but using abstract generative games.

GGP started after Deep Blue  won from Kasparov, and the field was looking for
their new \emph{Drosophila}. 
One of the criticisms of
the field of game playing was that all programs had to be specifically
developed for each game anew. Human intelligence, on the other hand,
is general, and can just as easily be applied to different tasks.

A general game playing system  should
be able to play any game for which it is given the rules. It was hoped
that the general methods coming out of GGP research 
would also be able to help in other areas,  such as in assisting
search and rescue missions~\cite{pell1992metagame,genesereth2005general}.
The Logic group at Stanford University currently hosts
 the General Game Playing
project~\cite{genesereth2005general,thielscher2011general} and
initiates competitions. 
\index{Michael Genesereth}

The GGP
approach is to create a system and a language in which many games can
be expressed. The challenge is then to create a single game player
that can play any game that occurs in a tournament.  A GGP program must be
able to accept any model that can be expressed in the game description
language. The game description language is similar to PROLOG.
In GGP a parser for the game description must be written, which implements
the state/action transition function, and termination rules. The
program has to come up with an efficient search procedure, and a good
evaluation function.
It is hard to preprogram heuristics when the domain is unknown. It was
hoped that GGP would stimulate the creation of general problem solving
methods. 
It is therefore  curious to note that a method developed in Go,
the other \emph{Drosophila}, has been  successful in GGP. Monte Carlo
Tree Search is a general search procedure that does not need domain-specific heuristics, and turned out to be   well suited for GGP.


Arguably GGP is a less spectacular \emph{Drosophila}, since there is no human
world champion to beat, and there is no active human gaming
community. In another sense GGP is  more spectacular 
since it tries to find solutions that work for all games. 

GGP has yielded a range of interesting advances towards
more general AI. Much research in GGP is being performed 
\cite{pell1996strategic,thielscher2009answer,schiffel2007fluxplayer,thielscher2010general,thielscher2011general,schofield2012hyperplay,michulke2009neural,schofield2015lifting,finnsson2010learning,bjornsson2009cadiaplayer,sironi2016comparison,sironi2017line,wang2018search,wang2018assessing}. It
will be interesting to see if more techniques from the two fields (i.e., general and special game playing)
will cross over.

There are other general game playing systems, which use their own
languages for defining the game rules. In 1992,  Pell introduced 
Meta-Game Playing, one of the earliest programs to use
automated game generation. Pell's Metagamer  was able to play Chess-like
games, given game rules definition in a  language called the Game
Description
Language~\cite{pell1992metagame,pell1993logic,genesereth2005general}.
Following the GGP idea a General
Video Game Playing competition has been
started\index{GVGP}~\cite{perez20162014,perez2015open,kunanusont2017general}.\footnote{\url{http://www.gvgai.net}}

Both AlphaZero and GGP show how general game play can be achieved. The
AlphaZero approach to general game playing is
 less general, requiring hand-designed input and output
layers, where GGP programs generate these automatically from a
description of the rules. 
The game complexity of Go, Chess, and Shogi is much larger
than that of typical GGP challenges. 



\section{A Computer's Look at Human Intelligence}\label{sec:humanintel}
Now that we have discussed the reinforcement learning methods, and
discussed future challenges in reinforcement 
learning in games, it is time to compare the artificial approach to
intelligence with the other approach, the human kind. 

\subsection{Human Intelligence}
As we already mentioned in Chap.~\ref{chap:games}, human
intelligence is a complex, multifaceted concept. Many different 
definitions of intelligence have been proposed. 
Intelligence involves recognition,
reasoning,  memory,  learning, 
problem  solving, and creativity. Human intelligence is usually also
assumed to include understanding,  
emotion, self-awareness, and purpose (volition). Intelligence is  the
ability to infer knowledge from information, and to apply that
knowledge  towards adaptive behavior within a context. See, for
example,  
Legg et al.~\cite{legg2007collection} or Neisser et
al.~\cite{neisser1996intelligence}. 

We will now discuss these elements of intelligence in more detail, and
analyze how human and artificial intelligence differ in their approach.


\subsubsection*{Memory and Recognition}
One of the basic elements of intelligence is the ability to recognize
objects or features that have been encountered before. This requires
sensors, a recognition system, and a
functioning memory. All have been studied widely in human
perception. In Kahneman's terminology, system 1 is
the part of our thinking that recognizes
objects. 
Human recognition works
fast and associatively.

Computers have always had memories, and over the years the amount of
memory of computers has grown greatly. Computers have long struggled
with the kind of recognition that humans do so easily. Only after CNNs
were introduced (fashioned after an animal visual  cortex) and after enough
training data became available in MNIST and ImageNet, and after enough
computing power in the form of GPUs became available,  did
computer recognition for images start to achieve impressive results.

\subsubsection*{Reasoning}
Reasoning is another  core element of intelligence. 
In Kahneman's terminology, reasoning is system 2, or thinking
slow. Reasoning is considered a positive trait, and many humans and
scientists 
pride themselves on their capacity for elaborate and deep logic
reasoning. It is something that children are trained  at in school
topics such as calculus, mathematics, and history. It is the reason
why Chess classes are 
present at  schools.

Indeed, one of the precepts of
classical economics is the assumption of rational behavior of the homo
economicus.  One of the interesting aspects of Kahneman's book is that
it stresses the importance of system 1, the fast, intuitive, type of
thinking. The non rational part of our thinking is guiding much of our
daily behavior. (Why do people buy lottery tickets, if we know 
that the expected value of a lottery is negative?)  Indeed, Kahneman's work on prospect theory, together with
many other important insights from psychology, 
has been instrumental for the field of behavioral
economics. Thaler and Sunstein have written a popular book on
nudging~\cite{thaler2008nudge}, applying insights from behavioral
economics to influence behavior of people.\index{nudging}\index{Homo Economicus}

Computers are good at reasoning, and are typically considered to be faster
at reasoning 
than humans. 
In artificial intelligence, systems
for logic reasoning were one of the first systems built, right up there with
Chess and Checkers
programs~\cite{wos1984automated,laird1987soar,newell1958elements}.
Some of the many successful
results of this research are the programming language
PROLOG~\cite{clocksin2012programming,bratko2001prolog},  research into
semantic
reasoning~\cite{antoniou2004semantic,holt2005medical}, and
negotiation~\cite{lin2014genius}.

\subsubsection*{Learning}\index{Johan Huizinga}\index{Homo Ludens}
One of the central aspects of intelligence is the ability to adapt
one's behavior based on information from the environment. The
ability to learn is certainly a crucial aspect of
intelligence. When children are young, learning and playing are closely related. Johan Huizinga, an eminent Dutch  historian, argues in \emph{Homo Ludens}~\cite{huizinga1938homo} that play may be the primary educational element in  human culture. Small children
learn through playing, through interaction with their
environments. Later, they learn in school, in a less-playful
regimented curriculum. Some of our learning is by repeated exposure to
examples. In  supervised learning we teach ourselves associations and
reflexes. Children who have been taught the tables of multiplication
can immediately answer the question: ``What is six times seven?''
using their system 1. Others have to use system 2 to answer this
question.

In computers, machine learning also uses memory and recognition capacities  to build up
associations.  If we teach (supervise) ourselves, we use our rational reasoning
capacities to do so. Learning was recognized early on as a second
central element of artificial intelligence. Computers have become
quite good at learning, in image recognition, and now also in Go. Due
to curriculum learning AlphaGo Zero learns Go in days, where for humans
this takes years of study.

Indeed,  since
the impressive results with deep learning in image recognition, the
term  machine learning has become almost synonymous with artificial
intelligence, sometimes forgetting about symbolic reasoning
altogether (our societal memory is imperfect).

\subsubsection*{Creativity}
Creativity is the process by which something new and valuable is
created~\cite{plucker1999psychometric}. Creativity is 
strongly associated with human
intelligence.

There have been quite some studies into artificial creativity. A
computational view has 
emerged, in which creativity emerges out of computational
processes~\cite{schmidhuber2006developmental}. Indeed, in addition to
automated theorem
proving~\cite{wos1984automated,loveland2016automated,bibel2013automated},
fields such as artificial painting~\cite{colton2012painting} and
artificial music~\cite{fiebrink2011real,tzanetakis2002musical} have a
long history in  
AI.\footnote{See, for example, the International Computer Music
  Association at \url{http://www.computermusic.org}.} 
A formal theory of creativity has been put forward by
Schmidhuber~\cite{schmidhuber2010formal}. If we think back to Fan
Hui's description of AlphaGo's moves (``speechless'') we see words
describing beauty and creativity (Sect.~\ref{sec:beauty}).

\subsubsection*{Understanding}
Understanding is a concept that is related to human intelligence.
Clearly, humans
can lay claim to understanding something. Insofar as computers are not
self-aware or have an identity, computers cannot be said to understand
something. Computers can be made to analyze a concept and explain the
reasoning that supports a certain conclusion, but it would be harder
to claim that an explanation is equivalent to 
understanding~\cite{chaitin2006limits}.

A Chess end-game analysis can produce the exact subtree proving why a
certain move is the best move in a position. That does not mean that
the computer has an understanding of what it analyzed, even if the
explanation is identical to one given by a human. 

\subsubsection*{Emotion}
A similar reasoning applies to emotion. Emotion is an important aspect
that determines much of human thought and behavior. Much of our evolutionary
survival skills are encoded in our emotional reactions, and much of
our daily behavior is
too~\cite{kahneman2011thinking,lisetti1998affective,broekens2007emotion}.

In artificial intelligence, the field of affective computing studies
the formalization, recognition, and reasoning with
emotions~\cite{lisetti1998affective,bosse2008formalisation}. As with
understanding, a  
computer may be able to recognize emotions, it can be made to behave
as if it has emotions, but few people would say that it has emotions.

Affective computing, and the models of emotion, will   yield better
human--computer interactions, and interactions with robots that appear
more natural. 

\subsubsection*{Self-Awareness, Purpose}
Most people would argue that computers do not possess self-awareness,
but humans do. Together with 
our free will, we believe that it guides much of our intelligent
behavior. Intelligent computer behavior is guided by one or more goals
that it tries to achieve. Humans decide on their own goal; for
computers the goal is externally specified.

Note that the strict behavioral approach of the Turing test could
allow a situation where a machine is so good at answering questions
concerning self-awareness and introspection that we must conclude that
it is indistinguishable from a sentient human. The question of
self-aware machines raises many interesting neurological and
philosophical question. See, for example,
\cite{koch2004quest,graziano2011human}.

\subsubsection*{System 1 and System 2}\label{sec:sys3}
Kahneman describes two kinds of human thinking: thinking fast and thinking
slow~\cite{kahneman2011thinking}, also known as system 1 and system
2. In this book, we have seen how artificial
intelligence in games is created out of planning 
and learning. The  exact and approximate methods in the search-eval
architecture all have parallels to thinking fast and slow
(Table~\ref{tab:exact}).

In addition to these parallels, there are important differences
between human and artificial intelligence. Human intelligence is
genera; artificial intelligence is mostly specific to one domain. Human
intelligence is self-aware; artificial intelligence is judged by
behavior, to pass the Turing test. 
\begin{table}[t]
  \begin{center}
    \begin{tabular}{cc}
      {\bf System 1} &{\bf System 2}\\ \hline\hline
      fast&slow\\ 
      recognition&reasoning\\ 
      reflex&rational\\ 
      approximate&exact\\ 
      train&plan\\ 
      eval&search\\ \hline
    \end{tabular}
    \caption{Fast and slow, eval and search}\label{tab:exact}
  \end{center}
\end{table}


Let us first look at thinking fast.
Thinking fast is reflex-like recognition. It is not so much like
the training of the function approximator, which is off-line, and
takes days, as it is the usage of the approximator, which is on-line,
and takes milliseconds. 

This brings us to revisit a point made earlier. We
want to  understand Kahneman's scheme of two kinds of
thinking, in which system 1 is thinking fast, and system 2 is
thinking 
slow.  Now self-play has an analogon in Kahneman's world. Self-play consists of exact planning methods being used to 
generate examples to train the function approximator  (Fig.~\ref{fig:selfplay12}). System 2 (slow) can be used not only to
reason and solve complicated problems such as what is the product of
6 and 7, but can also be used to train
system 1   by repeatedly exposing system 1 to examples that
system 2 generates. In this way we create a  mode of thinking that
can be called conscious {\em learning\/} or {\em training}. 
Anyone who has
learned words of a different language, or has recited the tables of
multiplication at school  is familiar with this kind of
learning. Practice makes perfect. Learning by self-play exists
both in AI and in human intelligence. In the next section we will
discuss a consequence of the fact that  artificial intelligence is now  able to
do this kind of training.




Now that we have seen some similarities between human and artificial
intelligence, let us have a look at some of the differences. 

\subsection{General Intelligence}
We have superficially explored  the limits of artificial intelligence; let
us also look at its strengths. Some of the  aspects of human
intelligence can readily be 
simulated by a computer, such as recognition, memory, reasoning, and
learning. 

\subsubsection*{Artificial Intelligence}
For a computer to behave artificially intelligent it should behave  the
same as human intelligence. An intelligent computer must be able to remember, to
recognize, to reason, and to learn, in order to pass the Turing test.

As we have seen in this book, there is much research on techniques for
recognition, memory, reasoning, and learning. In describing 
the behavior of Deep Blue and AlphaGo, some Chess and Go players have
described it in terms of beauty,
divine intervention, and other language usually used for creative
works.

In the fields of game playing it can reasonably be argued that genuine artificial
intelligence exists. In Backgammon, Checkers,
Chess, Atari, and Go, the best games are played by silicon.

\subsubsection*{Game Intelligence}
Exceptional intelligence in a single field is valued highly among
humans. Exceptional Soccer intelligence, Tennis intelligence, violin
intelligence, novelist intelligence, or Chess intelligence is scarce
(and can be monetized through prize 
money and sponsor deals).

Computers have such  specialized
intelligence. In fact, that is all they have. Computers can do one thing wel; humans can do many
things. Evolution has endowed humans 
with intelligence that allows them to do many things, in order to
survive in the world in which they live. The intelligence of game
playing programs has not evolved. Game playing
programs have been programmed to (learn to) excel at one game. They have specific
intelligence.

When looking carefully at the history of game intelligence,  an
interesting pattern  emerges.
The specific game intelligence started with search algorithms that
exploit the game structure (two-player zero-sum adversarial games
following the minimax rule). Game-specific heuristic search enhancements
made the programs even more game specific, and the eval algorithms are
based on game-specific heuristics as well. In some cases the
evaluation function is tuned with a (general) learning algorithm.

The minimax algorithm itself is general in the sense that it applies
to all two-player zero-sum games.

A subsequent search algorithm is MCTS. It is more general than minimax,
since it also works for single-agent  problems. It also
is more general because it works without a domain-specific evaluation
function.

To achieve high performance, domain-specific heuristic
methods were added, such as pattern-based playouts, and selection
rules that can incorporate priors.

Next are even more general learning algorithms, such as supervised and
reinforcement learning neural network function approximators. They are
 more general, in theory, although  tuning a network architecture
is a long and tedious process in which different network architectures
and hyperparameters are tried and tested on the specific game for
which it should work.

Self-play takes this a step further, forgoing the games databases of
grandmaster games, and learning itself in a curriculum learning
fashion to play a game.

The pattern that emerges is that the methods in reinforcement learning
in games have progressed from domain specific to general (which is
then partly undone by domain-specific performance tuning). The
 trend is towards more generality in methods, 
and this generality has now reached the point where it is  starting to
spill over to general application domains. As we saw in
Sect.~\ref{sec:future}, more and more applications
emerge where the self-play AI methods work, and a single architecture
may even work in a few different games. 

\subsubsection*{General Intelligence}
For humans,  intelligence is general. The intelligence
of humans applies to many different domains. A human being with little
general intelligence and high intelligence in one special
domain is unusual.  However, most individual humans do have
certain specialities in which they excel.

Some game playing programs have been designed to exhibit general
intelligence. The AlphaZero architecture can teach itself to play
three games exceedingly well, provided that the right input and output layer of
the neural network are specialized.
General game playing programs  are designed to be general enough to
interpret all games that can be specified by the Game Description
Language.

\subsubsection*{Practicing, or the 10,000 Hour Rule}\label{sec:10000}
How does one achieve high levels of intelligence? 
To reach the top in Tennis, soccer, violin playing, writing, or Chess
  requires talent and
practice. Indeed, it is sometimes argued that ``all'' it takes to
achieve success is a total of 10,000 hours of focused
practice~\cite{gladwell2008outliers}. Scientific studies  dispute this point of view and
the number of 10,000, and argue that the focus on practice is an
oversimplification. However, it is clear that, in addition to
talent, practice is indeed important in
games, music, and sports (see~\cite{macnamara2014deliberate}).

To learn new behavior, deliberate attention of system 2 is
necessary. During training we perform the actions repeatedly, under
the conscious guidance of system 2. At some point, system 1 will have
picked up enough to perform the behavior semi-automatically. Less
attention of system 2 is needed. After enough practice, the behavior
will be performed without any deliberate attention. System 1 has
learned to do it unattended. $6\times 7 = 42$.

\subsection{Next Steps: Sample Efficiency and Collaboration}
Although artificial intelligence achieves impressive results in
single domains, and methods are becoming more general, the difference with human intelligence remains
large. The steps that will be made on the path to achieve general, human,
intelligence will be small.

Two important challenges  towards general
intelligence are sample efficiency and collaboration.
The training time for current AI is still quite long. For each new
task a very large number
of samples  is needed for current reinforcement learning
methods.

Collaboration is the second challenge. The essence of
zero-sum games is competition. Multi-agent games feature
collaboration, and allow modeling of teamwork and forms of man/machine interaction.

\subsubsection*{Curriculum Learning}
Following a suitable curriculum improves
training results. A good curriculum speeds up training, and achieves
better end results~\cite{elman1993learning}. 
Human formal teaching (in schools) follows a curriculum, and the same
improvements have been found in 
supervised and reinforcement learning in artificial intelligence.

Self-play in games creates
a curriculum-like setting, where training examples are generated from
easy to hard. This allows for quick and effective training.
Indeed, as was mentioned in Sect.~\ref{sec:days}, AlphaZero learns
to play at a level beyond that of human world champions in days, where
human champions take years of dedicated training to reach that level.

AlphaZero is  not only better at playing, but also learns much faster than
 human champions.

\subsubsection*{Sample Efficiency for AGI}
However, the sample efficiency for each task is still low, since for each
new task learning is starting all over, from scratch. Even AlphaZero
did not achieve transfer learning, where (a part of) a trained network could be
transfered for (pre)training of another task.

The field of transfer learning studies how sample efficiency in new
tasks can be
increased through the reuse of trained knowledge between related
tasks. Transfer learning can thus contribute to more general AI. Knowledge transfer can be achieved in various ways: through network
parameters, or through hyperparameters.

Transfer learning is
related to pretraining, few-shot learning, and life-long
learning~\cite{thrun2012explanation}. Section~\ref{sec:transfer} discussed transfer learning.
Related to transfer learning is meta-learning. Meta-learning
studies mechanisms that learn to learn. It is related to algorithm
selection and to transfer
learning in the sense that its aim is to adapt to new environments
with few training examples. It is a very active area of research~\cite{finn2017model,frans2017meta,huisman2020deep}.

Two other methods that aim at reducing the sample efficiency problem
are hierarchical learning and model-based learning. Hierarchical
learning aims to model problems hierarchically so that a
subproblem can be solved independently, in sequence. Model-based learning aims to
learn a model of the problem first, and then use the model with other
methods, such as planning~\cite{plaat2020model}. In doing so model-based learning is more
sample efficient than model-free learning. The AlphaGo approach is a
model-based approach. More recently MuZero is a model-based approach
that learns a latent model of the problem
end-to-end~\cite{schrittwieser2019mastering}: AlphaGo Zero needed to
be provided with the rules of the game, but MuZero infers the rules
from the environment, and then learns to play at a high level. 

The promise of life-long learning,  transfer learning, and
meta-learning is faster learning, by reusing knowledge that
was learned in one domain for a new domain. These methods are
inspired by human learning, just
as the original inspiration of reinforcement learning is in
conditioning experiments with humans and animals. 

\subsubsection*{Collaboration}

Artificial
intelligence has achieved very strong results in specific domains, and
the challenge is now to extend these results to more general
domains, to different games, and beyond games, to fields such as
negotiation and human-robot interaction.

As our computational abilities increase, collaboration is becoming more important in reinforcement learning and
 games. Recent advances in Poker~\cite{brown2019superhuman},
StarCraft~\cite{vinyals2019grandmaster}, Capture the Flag~\cite{jaderberg2019human}, and DOTA 2~\cite{OpenAI_dota} show that
reinforcement learning versions of
population-based methods are able to achieve a high level of
collaboration in leagues consisting of humans and computers.

Researchers are arguing for more research into curriculum learning in
multi-agent games~\cite{leibo2019autocurricula}. More can be expected
in AI for negotiation and team collaboration.

\subsubsection*{Faster and Together}
We have come to the end of this book. We have discussed a
wide array of artificial intelligence technologies, and we have
seen programs reach amazing achievements. In reinforcement learning
agents can train  to transcend the level of supervisors or
teachers. The search-eval architecture 
has been the basis for all technological advances that
we covered in this book---from  planning, sampling, and function
approximation to self-play. 

This book has shown how game playing programs have progressed from game-specific
heuristic methods to general learning methods that teach themselves to
play (and beat world champions).
In this chapter we have
looked at the link between artificial and human intelligence, in
general terms of recognition, reasoning, and self-awareness. Although
  progress has been made, the gap with human intelligence remains large. So far, AI has achieved success in narrow
fields. In describing this progress,  we have
noted parallels between planning and learning,
symbolic and connectionst, and thinking fast and slow. 

With progress in transfer learning, we can expect AI to  become more
general in the near future.
With self-play, computers fashion their own training
curriculums and training progresses faster than before. For the games of Chess, Shogi,
and Go, a computer was able to teach itself to play at
world champion level and beyond in a matter of days. Humans take years of
dedicated study to reach
this level of play. In these games, computers not only play better,
but also learn  faster. 

Collaboration and negotiation are becoming increasingly important in
AI research. 
Research in  games such as Poker and StarCraft  has shown 
fast and powerful learning in team play. Games research is continuing
towards faster learning and  towards  teams in which
computers and humans collaborate.
\\

The future of AI is human.

\newpage
\section{Practice}

Below are some questions to check your understanding of this
chapter. Each question is a closed question where a simple, one
sentence answer is possible.
\subsubsection*{Questions}
\begin{enumerate}
\item Explain the search-eval architecture briefly.
\item Describe bootstrapping and optimization.
\item Why is diversity important in learning?
\item Explain triple-nested recursive optimization.
  \item How does automation lead to generalization?
\item Why are enhancements important?
\item Describe the relation between recursion, optimization, features,
  generalization, learning, and intelligence.
\item Why is Poker an interesting research topic?
\item What is general game playing?
  \item How does AlphaGo Zero join symbolic AI and connectionist AI?
\item Describe system 1, system 2, and self-play learning.
  \item Describe general and special intelligence.
\end{enumerate}
At this point, you have become quite proficient at implementing reinforcement
learning in games technology. Below are ideas for larger projects, to
help you on the path to further research. These are challenging
projects, perhaps suitable for a thesis.
\subsubsection*{Exercises}
\begin{enumerate}
\item Design a simple imperfect-information game and implement it  in
  A0G. For example: Blackjack. Does it learn? Does self-play work in
  imperfect information?
\item Write a self-play single-agent player for your single-agent
  optimization problem of choice. [challenging]
\item Interface A0G with the StarCraft interface, and create a
  multi-player self-play system. [challenging]
\item Design a a cooperative multi-player game. For example, merging
  traffic on a highway on-ramp, with or without
  signalling/negotiation. [challenging]
  \item Write a negotiation or Diplomacy game. See
    \cite{baarslag2013evaluating} for negotiation competitions.
    [highly challenging]
\end{enumerate}

\section*{Summary}
In order to look to the future, in this chapter we have looked back at
what we have covered in this book. We have 
looked at the methods that form the search-eval 
architecture. The search-eval architecture can also be seen as an
architecture with several nested levels of optimization, of which self-play is
just one. We see how artificial
intelligence methods  are used  to
achieve higher  levels of intelligence, or more general intelligence.

We have also briefly discussed other
application domains, beyond two-player zero-sum perfect
information. Multi-agent imperfect-information games approach more
aspects of the real world. Our techniques have progressed to the
point where creating players for Poker and StarCraft are realistic
challenges that are succeeding.

We discussed an alternative \emph{Drosophila}, general game playing
(GGP). Where the challenge of Go lies primarily in the size of the
state space, in GGP the challenge lies in being able to play a game
for which the rules are not known beforehand.


We have compared artificial intelligence with human
intelligence. We have looked at Kahneman's two systems, system 1 for
thinking fast, and system 2 for thinking slow. Thinking
fast resembles a 
function approximator, and thinking slow resembles planning. Self-play uses
planning to generate examples for the function approximator to learn
from.

Self-play learns relatively fast, due to curriculum learning.
We have  noted that AlphaGo Zero needed days to teach itself to play
Go, whereas humans need years of dedicated study to learn this level 
of Go, if they succeed at all.

We have looked at the elements that compose human intelligence, and
noted again that artificial intelligence is special.  At least in the
narrow domain of zero-sum board games, we have the situation that
computers play better and learn faster then humans. However,  human
intelligence is more general.

Future research will try to enlarge the domains in which artificial
intelligence works.  Sample efficiency and collaboration will be
important topics. The techniques covered in this book
will  contribute to this research.






\appendix

\chapter{Deep Reinforcement Learning Environments}\label{ch:env}

Artificial intelligence is an open field of research. Many researchers
publish their code, allowing for easy reproduction of experiments, and
allowing researchers to build upon each other's progress. This appendix contains
pointers to software environments that are meant to help you further
in your own research.

We list three types of environments: general programming and learning
environments, deep learning environments, and self-learning environments.

We start with general learning environments.

\section*{General Learning  Environments}

\begin{table}
  \begin{center}\small
  \begin{tabular}{lllc}
    {\bf Name} & {\bf Type} & {\bf URL} & {\bf Ref.}\\
    \hline\hline
    Python & General-purpose language & \url{https://www.python.org} & \cite{rossum1995python} \\ 
    Weka & Machine learning env. &
                                      \url{https://www.cs.waikato.ac.nz/ml/weka}
                                        & \cite{witten2016data}  \\ 
    Scikit-Learn & Machine learning env. & \url{https://scikit-learn.org} & \cite{pedregosa2011scikit}  \\ 
    Caffe & Deep learning env. &
                                      \url{http://caffe.berkeleyvision.org}
                                        & \cite{jia2014caffe}  \\ 
    PyTorch & Deep learning env. & \url{https://pytorch.org}
                                        &  \cite{paszke2019pytorch} \\ 
    TensorFlow & Deep learning env. &
                                            \url{https://www.tensorflow.org}
                                        & \cite{abadi2016tensorflow} \\ 
    Keras & Deep learning library & \url{https://keras.io} & \cite{chollet2015keras} \\ \hline
  \end{tabular}
  \caption{General learning environments}\label{tab:gen-env}
\end{center}
\end{table}

General-purpose programming languages such as Python allow us to
program the experiments that we wish to perform. One step further
are machine learning environments such as
Weka~\cite{witten2016data,hall2009weka} and Scikit-Learn~\cite{pedregosa2011scikit}.

For deep learning, well-known environments are 
Caffe~\cite{jia2014caffe}, Facebook's
PyTorch~\cite{paszke2019pytorch}, Theano~\cite{bergstra2011theano}
which has been subsumed by Google's TensorFlow~\cite{tensorflow2015whitepaper,abadi2016tensorflow}, and
its user friendly add-on Keras~\cite{chollet2017deep}. Keras is
perhaps the easiest way to start with deep learning. PyTorch offers
great flexibility. TensorFlow may well be the most popular
environment. All environments offer seamless integration with CPU and
GPU backends; no specialized GPU programming knowledge is needed.

Table~\ref{tab:gen-env} summarizes the general learning environments.

\section*{Deep Reinforcement Learning  Environments}

\begin{table}
  \begin{center}\small
  \begin{tabular}{l l p{6.5cm}c}
    {\bf Name} & {\bf Type} & {\bf URL} & {\bf Ref.}\\
    \hline\hline
    ALE & Atari Games & \url{https://github.com/mgbellemare/Arcade-Learning-Environment} &
                                                               \cite{bellemare2013arcade} \\
    Gym & RL environments & \url{https://gym.openai.com} & \cite{brockman2016openai}  \\ 
    Stable  & RL algorithms &
                                       \url{https://stable-baselines.readthedocs.io} &  \cite{hill2018stable} \\
    Baselines & &  & \\ 
    Dopamine & Deep RL env. & \url{https://github.com/google/dopamine}
                                        & \cite{castro2018dopamine} \\

    RLlib & Distributed RL &
                             \url{https://docs.ray.io/en/latest/rllib.html}&
                                                                                    \cite{liang2017rllib}\\ 
    Bsuite & RL environments & \url{https://github.com/deepmind/bsuite} &
                                                              \cite{osband2019behaviour} \\ 
    OpenSpiel & RL in games env. &
                                  \url{https://github.com/deepmind/open_spiel} & \cite{lanctot2019openspiel} \\ 
    Meta World & Meta-learning  & \url{https://meta-world.github.io} & \cite{yu2019meta} \\ \hline    
  \end{tabular}
  \caption{Deep reinforcement learning environments}\label{tab:rl-env}
\end{center}
\end{table}
      
Reinforcement learning has seen much interest from researchers. The
availability of environments has stimulated this research
greatly. Bellemare et al.\ introduced the Atari Learning Environment~\cite{bellemare2013arcade}
that has subsequently been incorporated into OpenAI Gym. 
The Gym Github page can be found  \href{https://github.com/openai/gym}{here}.\footnote{\url{https://github.com/openai/gym}}
OpenAI Gym has, in addition to ALE, classic reinforcement learning
examples such as  Cartpole and Mountain
Car. OpenAI also provides baseline implementations of reinforcement
learning algorithms~\cite{baselines}.
A great introduction on their use is at Spinning up: OpenAIs
    \href{https://spinningup.openai.com/en/latest/user/introduction.html#what-this-is}{Spinning
      Up}, and the code repo, which is
    \href{https://github.com/openai/spinningup}{here}.\footnote{\url{https://github.com/openai/spinningup}}

Subsequently,  refactored
versions of baselines are presented as the Stable Baselines. Note:
Stable Baselines supports TensorFlow
      versions from 1.8.0 to 1.14.0, and may not yet work on TensorFlow
      versions 2.0.0 and above. Support for TensorFlow 2 API is
      planned.

      The Baseline Zoo contains trained models for the stable baselines.

      Dopamine~\cite{castro2018dopamine} is a framework by DeepMind
      for deep reinforcement learning.

      RLlib~\cite{liang2017rllib} provides abstractions for
      distributing reinforcement learning on large-scale clusters of machines.

      Bsuite~\cite{osband2019behaviour} is a behavior suite for
      reinforcement learning.

      OpenSpiel~\cite{lanctot2019openspiel} is a framework for
      reinforcement learning in games, implementing many of the
      algorithms that are discussed in this book.

      Meta World~\cite{yu2019meta} is a benchmark and evaluation suite
      for multi-task and meta reinforcement learning.

Table~\ref{tab:rl-env} summarizes the deep reinforcement learning environments.

\section*{Open Reimplementations of AlphaZero Self-Play}\label{sec:reimp}

\begin{table}
  \begin{center}\small
  \begin{tabular}{p{1.4cm}lp{6cm}c}
    {\bf Name} & {\bf Type} & {\bf URL} & {\bf Ref.}\\
    \hline\hline
    AlphaZero General& AlphaZero in Python &
                                              \url{https://github.com/suragnair/alpha-zero-general}
                                        &  \cite{nair2017learning} \\ 
    ELF & Game framework & \url{https://github.com/pytorch/ELF} & \cite{tian2017elf}  \\ 
    Leela & AlphaZero for Chess, Go &
                                     \url{https://github.com/LeelaChessZero/lczero} 
                                        &  \cite{pascutto2017leela} \\ 
    PhoenixGo &  AlphaZero-based Go prog. & \url{https://github.com/Tencent/PhoenixGo}
                                        & \cite{PhoenixGo2018} \\ 
    PolyGames &  Env. for Zero learning & \url{https://github.com/facebookincubator/Polygames} & \cite{cazenave2020polygames} \\ \hline
  \end{tabular}
  \caption{Self-learning environments}\label{tab:self-env}
\end{center}
\end{table}

The publication of the AlphaGo papers created great
interest in the research community in self-learning self-play approaches. The DeepMind team has not yet published
their code, but the
publications~\cite{silver2016mastering,silver2017shogi}
(and especially~\cite{silver2017mastering}) provide enough detail for 
researchers to create their own versions, some of which are open
sourced on GitHub. We will describe some of them.

\subsubsection*{A0G: AlphaZero General}\label{sec:a0g}
Thakoor et al.~\cite{nair2017learning} created\index{A0G}\index{AlphaZero General}
a self-play system as part of a course project.  The system is called AlphaZero
General (A0G), and is
on \href{https://github.com/suragnair/alpha-zero-general}{GitHub}. It
is implemented in Python and TensorFlow, Keras, and PyTorch, and
suitably scaled down for smaller computational resources. It has
implementations for $6\times 6$ Othello, Tic Tac Toe, Gobang, and
Connect4, all small games of significantly less complexity than
Go. Its main network architecture is a four layer CNN followed by 
two fully connected layers. The code is  easy to understand
in an afternoon of study, and is well suited for educational purposes.
The course project write-up provides some \href{https://github.com/suragnair/alpha-zero-general/raw/master/pretrained\_models/writeup.pdf}{documentation}~\cite{nair2017learning}.

\subsubsection*{Facebook ELF}
ELF stands for Extensible Lightweight Framework. It is a framework for game
research  in C++ and Python~\cite{tian2017elf}. Originally developed
for real-time strategy games by Facebook, it includes the Arcade Learning Environment and
the \href{https://github.com/facebookresearch/darkforestGo}{Darkforest}\footnote{\url{https://github.com/facebookresearch/darkforestGo}}
Go program~\cite{tian2015better}. ELF can be found
at \href{https://github.com/pytorch/ELF}{GitHub ELF}.
ELF  also contains the self-play program OpenGo~\cite{ELFOpenGo2018}, a reimplementation of  
AlphaGo Zero (in C++).\index{ELF}\index{OpenGo}

\subsubsection*{Leela}
Another reimplementation of AlphaZero is
 Leela. Both a Chess and a Go version of Leela
 exist. The \href{https://github.com/LeelaChessZero/lczero}{Chess}
 version is based on Chess engine
 Sjeng. The \href{https://github.com/gcp/leela-zero}{Go}\footnote{\url{https://github.com/gcp/leela-zero}}
 version is based on Go engine Leela.  Leela does not come with
 trained weights of the network. Part of Leela  is a community effort
 to compute these weights.\index{Leela}

\subsubsection*{PhoenixGo}\index{PhoenixGo}
PhoenixGo is a strong self-play Go program by Tencent~\cite{PhoenixGo2018}. It
is based on the AlphaGo Zero architecture. It can be
found at \href{https://github.com/Tencent/PhoenixGo}{Github
PhoenixGo}. A
trained network is available as well.

\subsubsection*{PolyGames}\index{PolyGames}
PolyGames~\cite{cazenave2020polygames} is an environment for Zero-based learning (MCTS with deep
reinforcement learning) inspired by AlphaGo Zero. Relevant learning
methods are implemented, and bots for Hex, Othello, and Havannah have
been implemented. PolyGames can be found at \href{https://github.com/facebookincubator/Polygames}{PolyGames}.

Table~\ref{tab:self-env} summarizes the self-learning environments.

\chapter{*AlphaGo Technical Details}\label{chap:godetails}
This appendix contains more technical details for the AlphaGo, AlphaGo
Zero, and AlphaZero programs, that were too technical to fit in the
main text of Chap.~\ref{chap:a0g}, but are important to truly
understand how things work. All information in this appendix is from
the three papers by  Silver et
al.~\cite{silver2016mastering,silver2017mastering,silver2018general}.

\section{AlphaGo}
We start with AlphaGo.
The AlphaGo program is a complicated program. All details are
described in Silver et al.~\cite{silver2016mastering}. In this
appendix we summarize some of the more technical details from this publication.

\subsubsection*{Search Algorithm}
The search algorithm of AlphaGo is MCTS (see also
Sect.~\ref{sec:mcts} and~\cite{browne2012survey}). The four MCTS
operations are implemented as follows. 

{\em Selection}
For selection, a variant of P-UCT~\cite{rosin2011multi,matsuzaki2018empirical}\index{P-UCT} is used, which
allows integration of prior information into the UCT selection
formula. This is especially important for new, unexplored children, to
be able to use the 
information of the policy net. With the policy net, AlphaGo does not employ the all-moves-as-first or
rapid action value estimation heuristics used in most other programs,
since the policy net provides  priors of better quality (Sect.~\ref{sec:rave}). In addition AlphaGo does not use progressive
widening~\cite{chaslot2008progressive}, dynamic komi~\cite{baudivs2011balancing}, or an opening book, which most other Go
programs do.

The constant  $C_p$ determines the exploration/exploitation trade-off.\index{$C_p$}
 A high value means more exploration. AlphaGo uses 
$C_p=5$. This  is high compared with most Go programs, where values of $C_p<0.75$ or even
$C_p\approx 0$ were
used for programs that use RAVE~\cite{gelly2006modification,chaslot2010monte}.\label{sec:cp} AlphaGo  favors a
considerable amount of exploration compared with earlier Go
programs.\index{$C_p$} Section~\ref{sec:explexpl} discusses the performance
sensitivity of $C_p$.

{\em Expansion} Children of leaves of the MCTS tree are expanded  when
the visit count exceeds a threshold.  New children get prior
probabilities from the policy network, and  are put in the
queue for evaluation by the GPU with both the policy and the value
network. The threshold is determined by 
the length of the work queue for the GPU.

{\em Rollout}
Leaves in AlphaGo are evaluated both by the value network (GPU queue)
and  rollouts with the fast rollout network. Results are
combined with a parameter $\lambda=0.5$ (half rollout, half value
network).\index{$\lambda$, AlphaGo mixing parameter}

{\em Backup}\index{virtual loss}\index{lock free}
The rollout statistics are updated as usual in MCTS (visit count and
win count). AlphaGo is a tree parallel program~\cite{chaslot2008parallel}, which
means that multiple processes can update the MCTS tree
concurrently. AlphaGo uses standard techniques for ensuring efficient
parallel selection (virtual loss~\cite{segal2010scalability,mirsoleimani2017analysis}) and
correctness of the tree (lock free
updates~\cite{enzenberger2009lock,mirsoleimani2018lock}).

\subsubsection*{Network Architecture}
For the fast rollout policy network, the function approximator is  based on a hash
table. The fast rollout policy is based on small ($3\times 3$) patterns that
are also found in other Go
programs~\cite{gelly2006modification,gelly2007combining}. It is based
on a  hash table of the {\em last good reply\/}
heuristic~\cite{baier2010power}.

The function approximator for the selection policy is based on a 13-layer
convolutional neural network. It has 12 convolutional hidden layers (plus one input
layer and one fully connected output layer). The first convolutional
layers uses a kernel size of $5\times 5$; the other hidden layers use
$3\times 3$. A ReLU unit is applied at each layer.

The network for the value network has one convolutional layer more
than the policy network, for a total of 14 layers.

\subsubsection*{Input/Output Features}
The input layer is  $19\times 19\times 48$ with $48$ feature
planes. The inputs to the AlphaGo nets are more than just the stones
on the board. The features of each point include stone color, how many
times ago a turn 
was played, the number of liberties, how many opponent stones would be
captured, how many own stones would be captured, the number of
liberties after the move is played, ladder status, ladder escape, if a
point fills an own eye, and color to play. 

The output layer is a fully connected layer with a soft-max function
for the policy network and a single $\tanh$ unit for the value
network.

\subsubsection*{Performance of the Three Networks}
We will now discuss the three  networks in more detail.

{\em Fast Rollout Policy Network}\index{AlphaGo!fast rollout policy
  network} The  
fast rollout policy network is based on $3\times 3$ patterns. Its
function is to replace (or enhance) the random rollouts of MCTS. It
achieves only 24\% 
accuracy on a test set, but it replies with an answer in $2 \mu$s,
which is 1500 times faster than the slow policy network, on the
hardware used by the AlphaGo team.

{\em Selection Policy Network}\index{AlphaGo!slow selection policy network}
The network trained by supervised learning is trained on 29 million
positions for classification of expert moves on a dataset of internet
games (split into a test set of one million positions, and a training
set of 28 million positions). This policy network achieves 57\%
accuracy of  correct actions
after training. Its time to compute  an answer is 3 ms.
It is trained by asynchronous stochastic gradient descent.
The supervised training took three weeks; 340 million training steps
were performed.

The policy network is further refined by
reinforcement learning with self-play positions. It was trained for one full day of
self-play of 10,000 mini-batches of 128
games, randomized to prevent correlations using positions from
a pool of opponents. 
The training algorithm is policy gradient descent, with the REINFORCE\index{REINFORCE}
algorithm~\cite{sutton2018introduction}. 
This network performs quite well. Without any search at all, it won 85\% of
games against Pachi, 
a strong open-source MCTS
program.\footnote{\href{http://pachi.or.cz}{Pachi code} is at \url{http://pachi.or.cz}} The reinforcement learning is
important, since a no-search network based on
supervised learning won only 11\% against Pachi.\index{Pachi}
Curiously, it is the weaker supervised net that was used in MCTS
selection in AlphaGo.  The paper speculates that this is because
of greater variation in the supervised network, it chooses from a wider
repertoire of ``best'' moves, whereas the reinforcement
network chooses always the single best move~\cite{silver2016mastering}. Still, it
is surprising that a weaker network is used in favor of a stronger
network, and more work is needed to fully understand the reasons.

{\em Value Network}\index{AlphaGo!value network}
The value network is trained as a  regression  of the policy
network. The value represents the probability of winning. For the
value network 30 million training positions were used from self-play
games. To avoid overfitting the training positions were created 
so as not to be correlated, using a small self-play search based on the
positions that were used for the supervised selection policy
network.
The training algorithm is stochastic gradient descent as in the
supervised training phase, based on the mean squared error between the
predicted values and the observed rewards.
This training phase took 50 GPUs one week. 

Many more  details of the various training
methods can be 
found in  the original
publication~\cite{silver2016mastering}.

\subsubsection*{Compute Hardware}\label{sec:tpu}
Training the networks occurs off-line, before a match is played, and large
amounts of computation power and time are available. During a match
the time to compute is limited. Function approximation does not come cheap.  Position
evaluation requires evaluating multiple networks, which costs
orders of magnitude  more computation than traditional heuristic
programs. Combining MCTS
with deep neural networks has required quite some software engineering. 
AlphaGo uses an asynchronous multi-threaded search that executes
simulations on CPUs, 
and computes policy and value networks in parallel on GPUs. AlphaGo
used 40 search threads, 48 CPU cores, and 8 GPUs. A distributed
version was also developed, to make use of a cluster of machines. It
has been run on  1202 CPUs and 176 GPUs.

Google developed a special kind of GPU  optimized
for tensor computations in 
TensorFlow, named TPU,\index{TPU} for tensor processing unit.
DeepMind started using TPUs for AlphaGo in 2016. TPUs are designed for
high-throughput low-precision computations (as low as 8-bit precision)
whereas GPUs are typically optimized for 32- or 64-bit precision
operations, and memory throughput is often a bottleneck~\cite{jouppi2017datacenter}.

\section{AlphaGo Zero}\label{sec:ag0-a}
A year later, AlphaGo Zero was introduced. The paper by Silver et
al.~\cite{silver2017mastering} contains all the details. We highlight some
 here.

\subsubsection*{Detailed Self-Play Training}
Let us look at how self-play is implemented. The neural network is trained as follows. A self-play
reinforcement learning algorithm uses MCTS to play each move. First, the neural network
is initialized to random weights $\theta_0$. At each subsequent iteration $i$,
games of self-play, or episodes, are generated, consisting of a sequence of
moves. For each move of each episode  $t$, an MCTS
search $\pi_t$ is performed using the previous neural network
$\theta_{i-1}$ and a move is played sampled from $\pi_t$.
The game $t$ terminates when both players pass at step $T$ at which
point the game is scored $z=\{-1,+1\}$. The data are stored in a
state-policy-distribution-reward triple $(s_t, \pi_t, z_t)$ as examples
in the example buffer for training the network parameters
$\theta_i$. The network $(p,v)=f_{\theta_i}(s)$ is trained by sampling uniformly over the
example triples of all time steps (episodes) of the last iterations of
self-play. A loss function is chosen to minimize the error between
predicted value $v$ and self-play outcome $z$, and to maximize the
similarity of predicted move probabilities $p$ and search
probabilities $\pi$.
Policy and value were used as a combined loss function that sums over
the mean squared value error and cross entropy policy
losses:\index{loss function AlphaGo Zero}
$l=(z-v)^2-\pi^T\log p + c\|\theta\|^2 $,
where $c$ is the controlling parameter for L2 weight regularization to
prevent overfitting~\cite{goodfellow2016deep,silver2017mastering}.

\subsubsection*{Network Architecture}
The new function approximation architecture is  simpler than in AlphaGo. It
consists of one residual network (Sect.~\ref{sec:resnet}) instead of a value net, a slow selection
policy net, and a 
fast rollout policy net. The ResNet has one input and two outputs: a
policy head and a value head. The ResNet is trained by self-play; the
databases of positions for supervised and reinforcement learning have
disappeared. MCTS no longer does rollouts, so the fast
rollout net has gone as well.

It is reported that combining policy and value  into a single network slightly
reduces move prediction accuracy, but also reduces the value error. It
improved playing performance in AlphaGo Zero by 600 Elo points, in
part because the dual objective regularizes the network better.

\begin{figure}[t]
    \centering{\includegraphics[width=5cm]{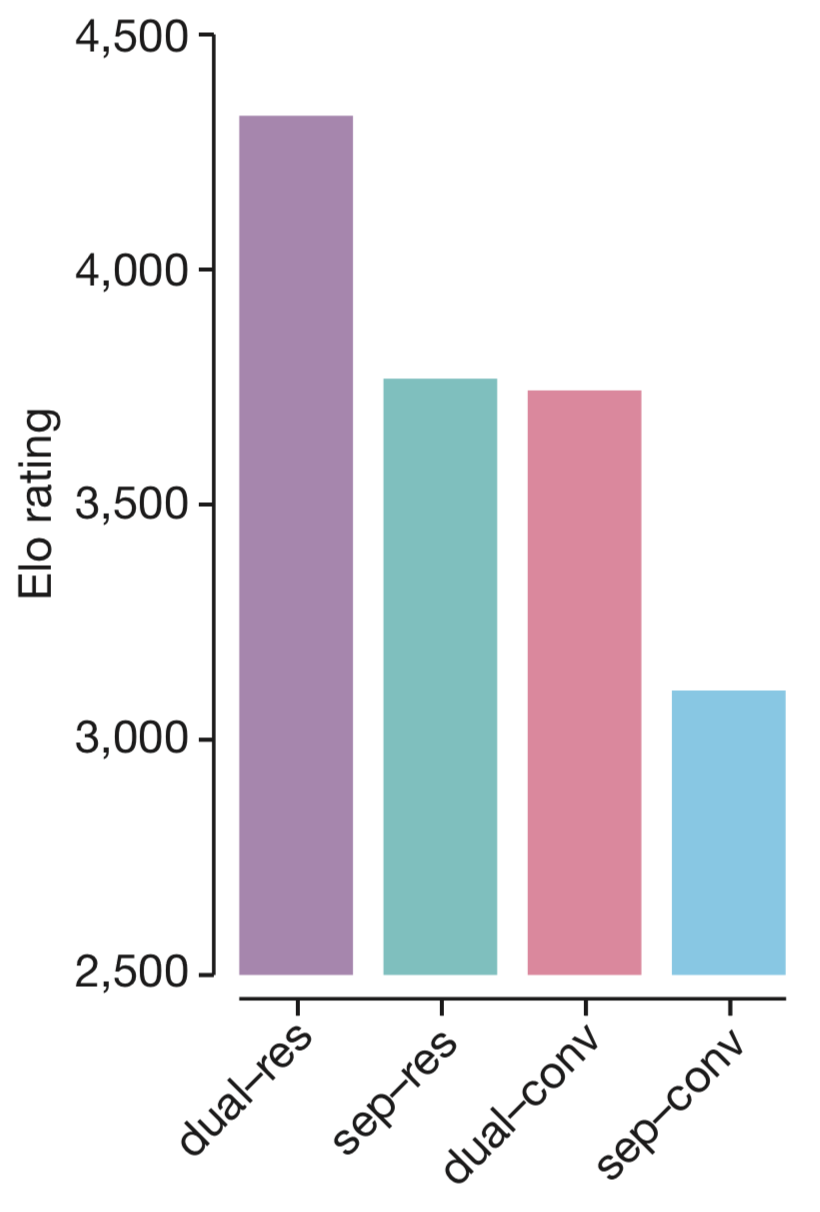}
     \caption[Dual/separate, convolution/residual]{Comparison of dual-headed and separate, and of convolutional
     and residual networks~\cite{silver2017mastering}}\label{fig:convnetresnet}}
\end{figure}

Figure~\ref{fig:convnetresnet} shows the difference in performance
between the combined dual policy/value network and the separate
networks, and the convolutional and residual networks~\cite{silver2017mastering}.

The ResNet consists of blocks of convolutions~\cite{he2016deep} with batch
normalization~\cite{ioffe2017batch} and ReLU units. Interestingly,
after experimenting with 40 block 
ResNets against 20 block ResNets, the team decided to use the 20 block
net in the tournament version of the player. The added complexity of 40 block nets did not increase
performance enough to outweigh computational cost. 

As before, neural network parameters are optimized by stochastic gradient descent
with momentum and learning rate annealing. 
In a comparison against AlphaGo the ResNet was more accurate, achieved lower errors and
improved performance in AlphaGo by over 600 Elo~\cite{silver2017mastering}.

\begin{figure}[t]
    \centering{\includegraphics[width=4cm]{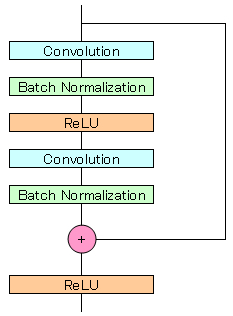}
     \caption{AlphaGo Zero residual block~\cite{silver2017mastering}}\label{fig:resnetblock}}
\end{figure}

The input to the neural network is a $19 \times 19 \times 17$ image
stack comprising 17 binary feature planes. The 17 feature planes
contain the current board and a history of the 8 previous moves, both
for black and for white, and the color to play. The input features 
are processed by a residual network tower. The tower consists of a single
convolutional block followed by either 19 or 39 residual blocks, one
of which is shown in Fig.~\ref{fig:resnetblock}. The total network
depth, in the 20 or 40 block network, is thus 39 or 79 parameterized
layers (since a ResNet block has 2 convolutional layers) plus  2 layers for the policy head and 3 layers for the value
head, consisting of a  block ending in a fully connected layer. See
the paper for details~\cite{silver2017mastering}.

\subsubsection*{Compute Hardware}
\begin{figure}[t]
    \centering{\includegraphics[width=10cm]{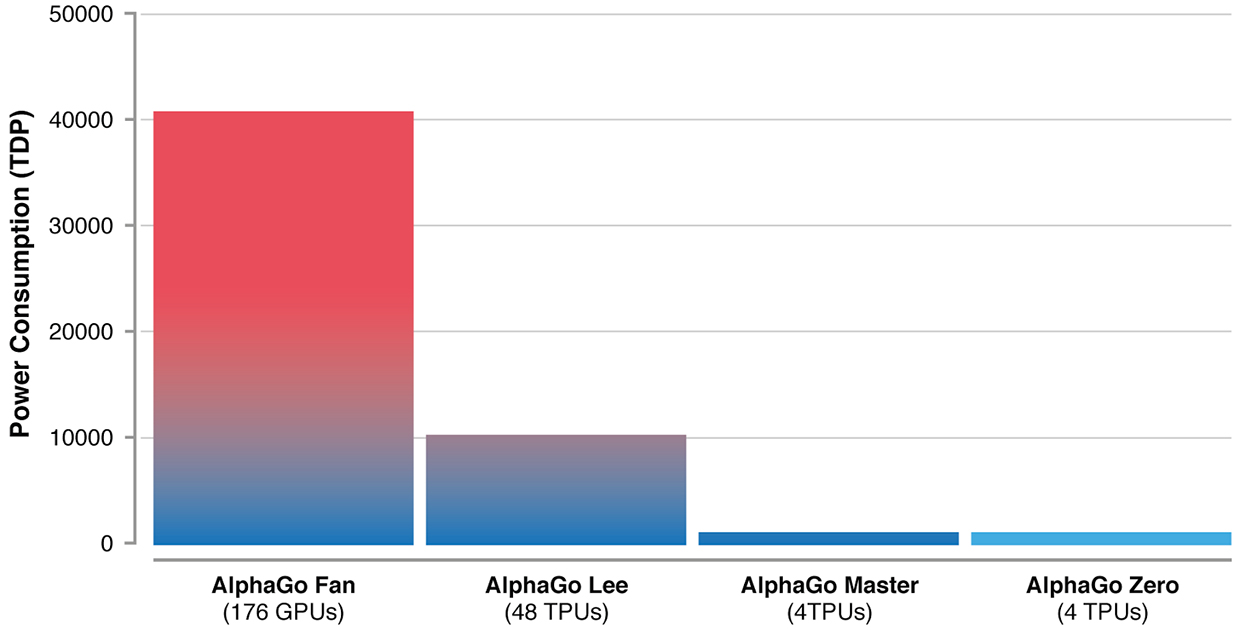}
     \caption{Comparison of power consumption for training~\cite{googlecloud2018}}\label{fig:power}}
\end{figure}

Originally, each neural network was trained with TensorFlow, with 64 GPU workers
and 19 CPU parameter servers. Later, TPUs were used.
Figure~\ref{fig:power} shows the
difference in total dissipated power consumption between the different versions of
AlphaGo.

\subsubsection*{Search Algorithm}
The search algorithm of AlphaGo Zero is MCTS. For selection, the same
P-UCT~\cite{rosin2011multi} variant was used, which incorporates priors from the policy head,
and not RAVE. This MCTS did not use rollouts at the leaves of the
MCTS tree, but value head lookups. In the self-play, a total of 25,000
games  were played in each self-play iteration. MCTS performs 1600
simulations for each move. 
MCTS hyperparameters were selected by Gaussian process
optimization~\cite{shahriari2016taking}.

\subsubsection*{Input/Output Features}
The input planes of AlphaGo Zero are simpler than in AlphaGo. Only the
board information is used; no additional information such as ko, or
liberties, is used. Even the basic heuristic of not playing moves that
would fill the player’s own
eyes is excluded. AlphaGo Zero learns from scratch. No heuristic
information is given to the program.

\section{AlphaZero}
Again a year later Silver et al.~\cite{silver2018general} published
their AlphaZero paper, showing how the simpler architecture of AlphaGo
Zero was also a more general architecture.
This appendix describes some of the more technical elements.

\subsubsection*{Input/Output Features}
The input and output
layers are different for each game.
For Go, the same $19\times
19\times 17$ planes as in AlphaGo Zero were used in AlphaZero, where 8
moves of history is encoded in addition to the stones on the
board (Table~\ref{tab:planes}). Move encoding in Go is simply a coordinate of a stone.

For Chess, a similar encoding scheme is used, although more elaborate
using many more planes. The scheme encodes all possible moves from
a position in different binary planes. In Chess there is a greater
variety of possible moves, and the encoding scheme therefore uses more planes
than in Go, $8\times 8\times 73$.\footnote{Silver et al.~\cite{silver2018general} contains the
details of their elaborate encoding. The encoding scheme is quite complex, although it is noted that simpler
encoding schemes also work.}

Shogi has an even greater variety of
moves, leading to a stack of $9\times 9\times 139$ planes as
input. The paper mentions that experiments with different coding
schemes worked as well, although training efficiency was affected somewhat.

\subsubsection*{Compute Hardware}
Both Stockfish (Chess) and Elmo (Shogi) ran with 44 threads on 44 cores (2 Intel Xeon
Broadwell CPUs with 22 cores) with 32 GB transposition tables and 3 hour
per match time controls with 15 seconds per move extra. AlphaZero ran
on a single machine with 44 CPU cores, and 4 first-generation TPUs.

\chapter{Matches of Fan Hui, Lee Sedol, and Ke Jie}\label{chap:matches}

This appendix contains the games of the matches played by AlphaGo
against Fan Hui, Lee Sedol, and Ke Jie.

Note that games can be played online  at
\href{https://deepmind.com/research/alphago/match-archive/alphago-games-english/}{DeepMind}.\footnote{\url{https://deepmind.com/research/alphago/match-archive/alphago-games-english/}}
You can also find match commmentary at this site.
SGF files of the games are easily found online, allowing
  interactive replay with an SGF viewer. The match diagrams below are
  based on~\cite{silver2016mastering,silver2017mastering}.

\section*{Fan Hui}
\begin{figure}[p]
    \begin{minipage}{0.48\textwidth}
    \begin{tabular}{c}
      \includegraphics[width=\textwidth]{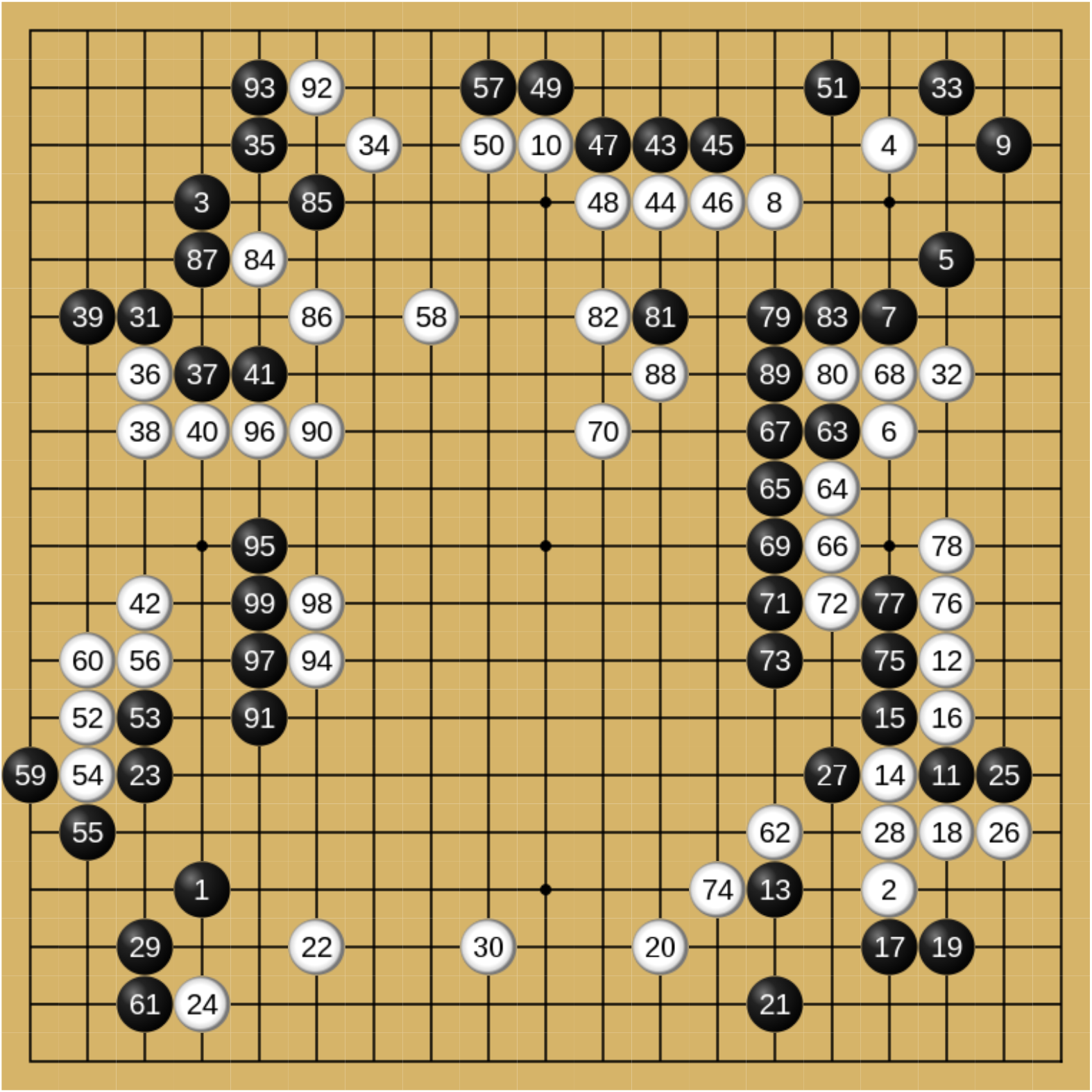}\\
      \includegraphics[width=\textwidth]{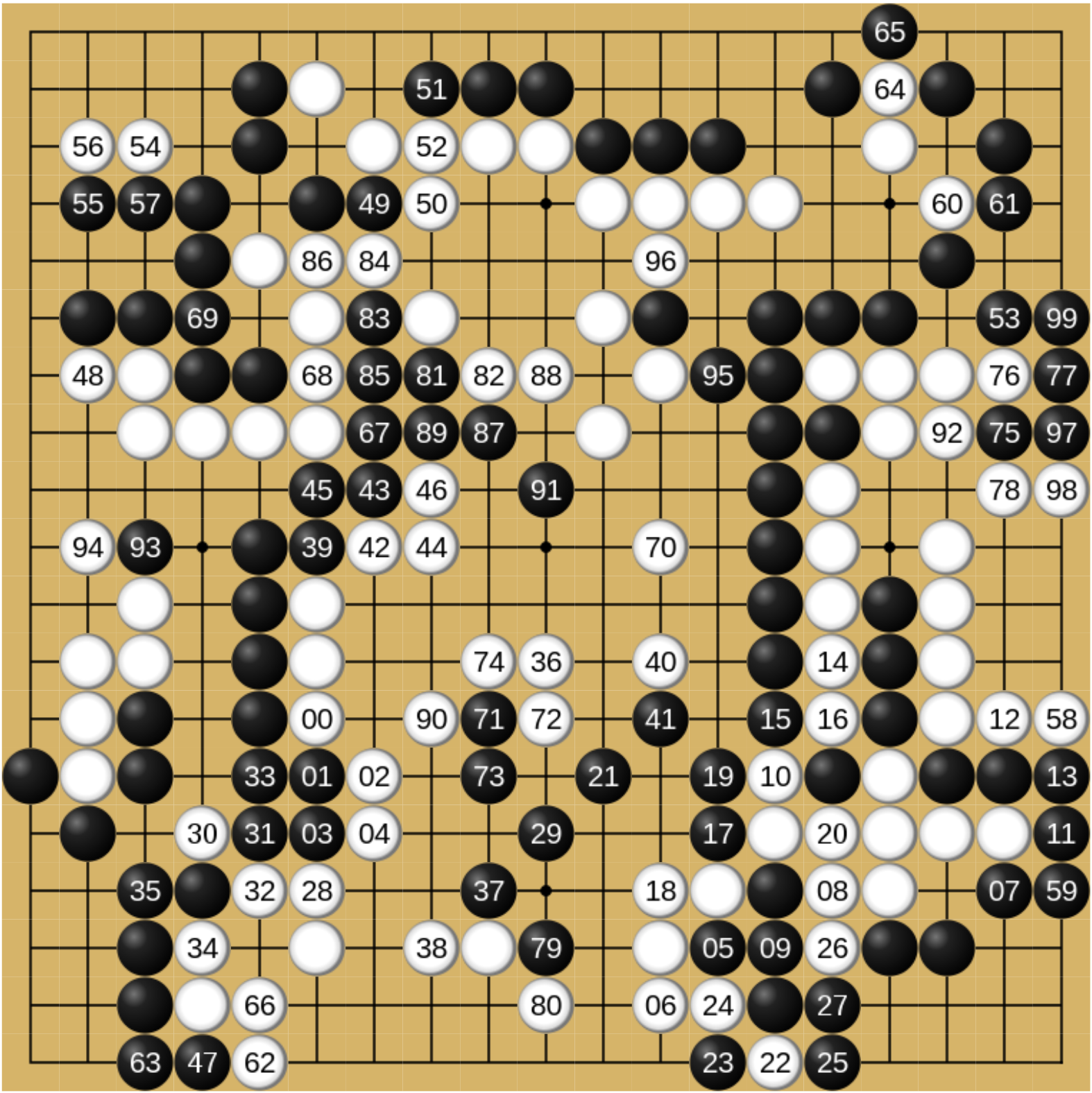}\\
      \includegraphics[width=\textwidth]{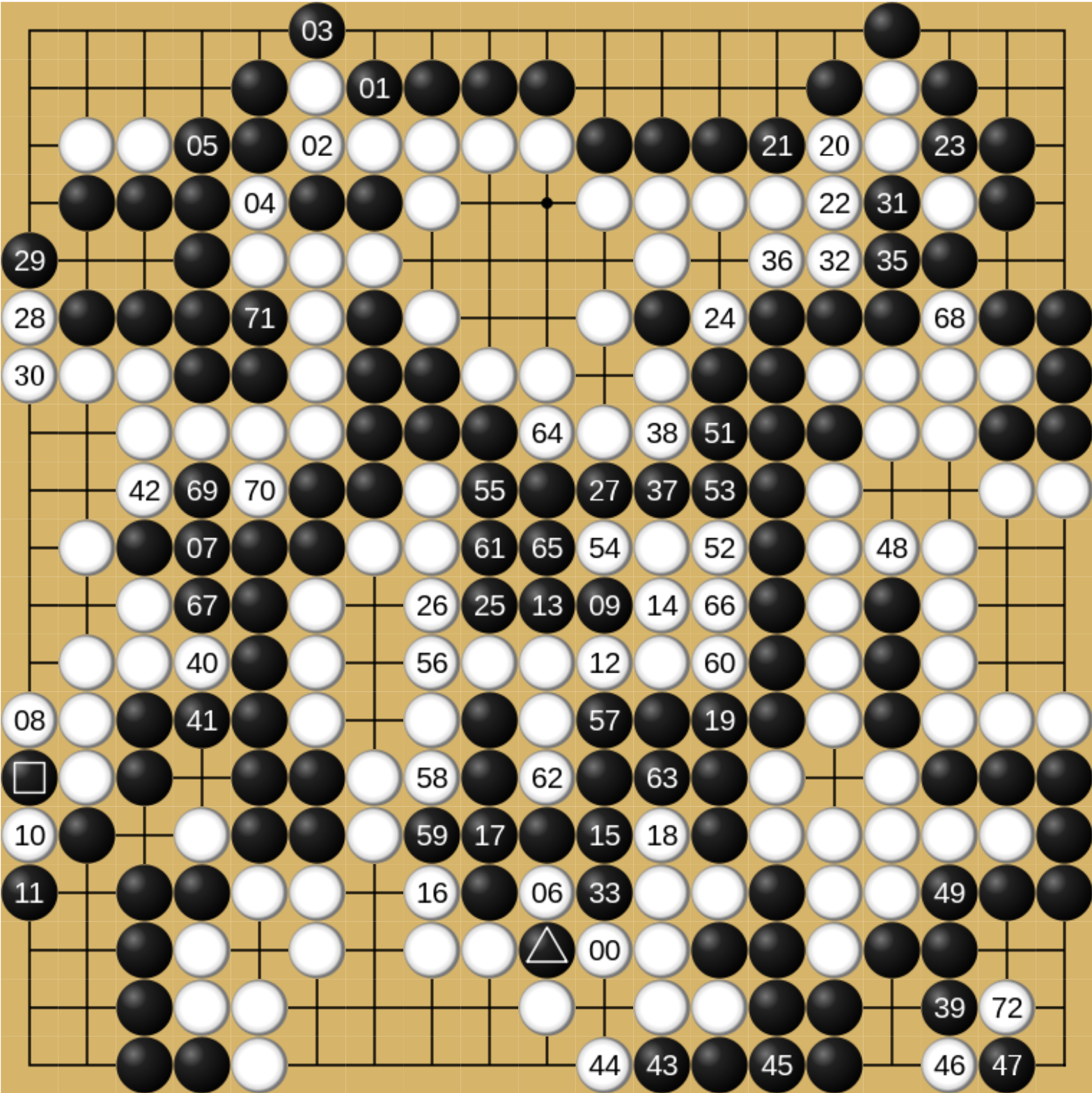}
    \end{tabular}
     \caption[Fan Hui game 1]{Game 1 Fan Hui (black) vs.\ AlphaGo (white): move 1--99,
       move 100--199, and move 200--272. Move 234 is played at triangle,
       move 250 at square.  AlphaGo won by 2.5 points.}\label{fig:gamesfanhui1}
  \end{minipage}
  \qquad
  \begin{minipage}{0.48\textwidth}
    \begin{tabular}{c}
      \includegraphics[width=\textwidth]{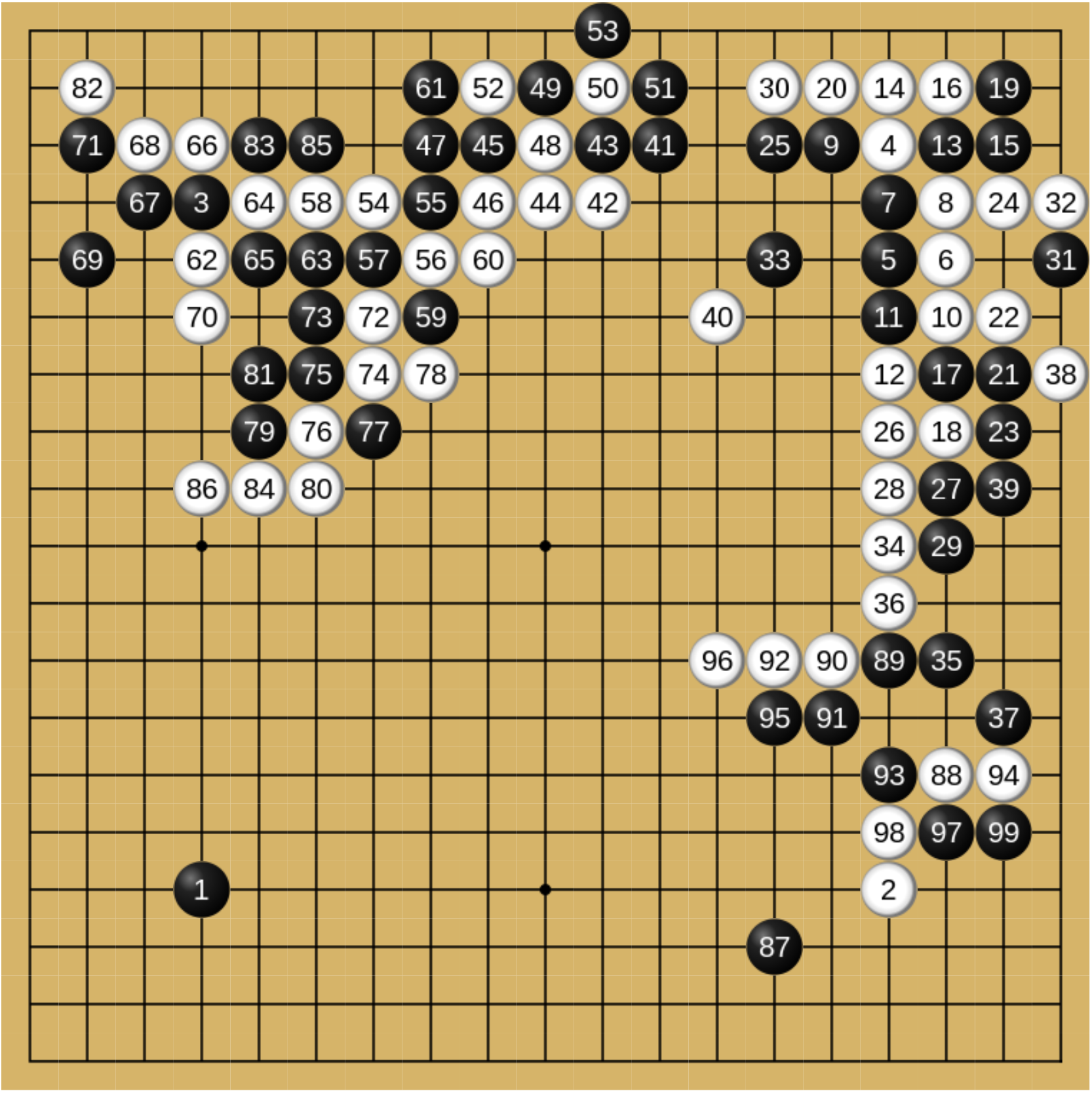}\\
      \includegraphics[width=\textwidth]{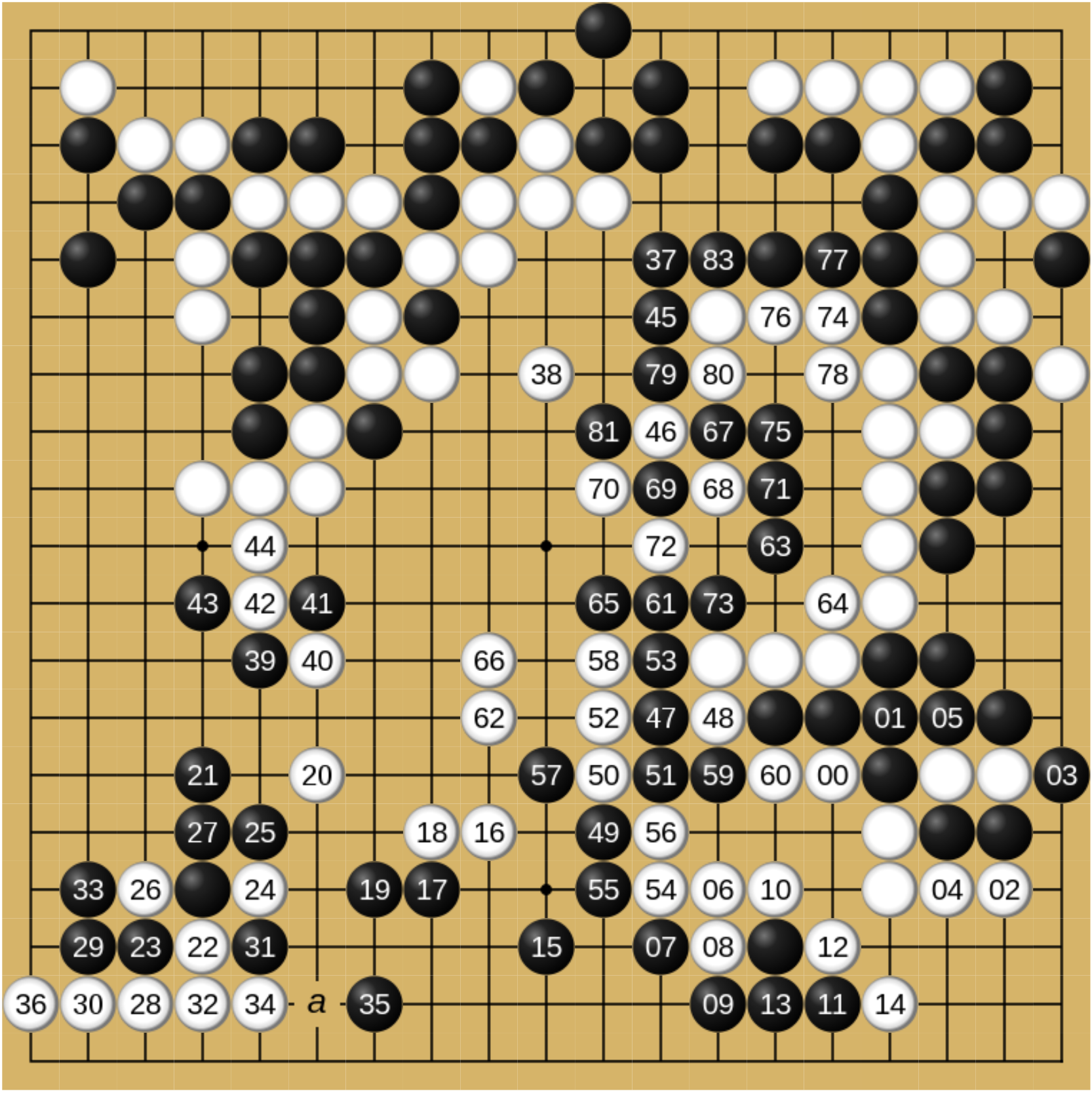}\\
      \vspace{0.9\textwidth}
    \end{tabular}
     \caption[Fan Hui game 2]{Game 2  AlphaGo (black) vs.\ Fan Hui (white): move
       1--99, and
       move 100--183 (move 182 at 169). AlphaGo did not play 135 at
       ``a,'' but played a safer move to win. AlphaGo won by resignation.}\label{fig:gamesfanhui2}
  \end{minipage}
\end{figure}

\begin{figure}[p]
    \begin{minipage}{0.48\textwidth}
    \begin{tabular}{c}
      \includegraphics[width=\textwidth]{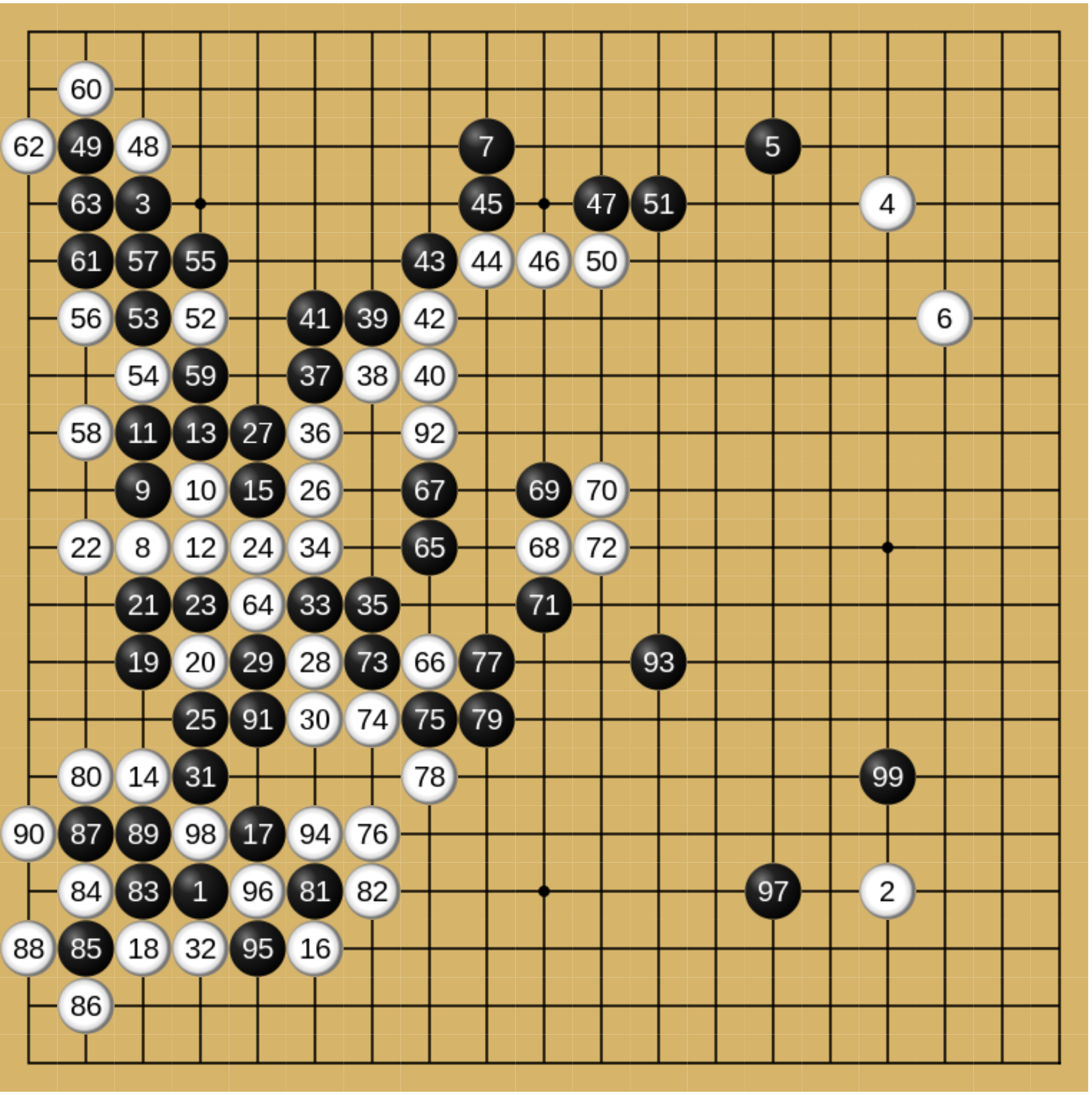}\\
      \includegraphics[width=\textwidth]{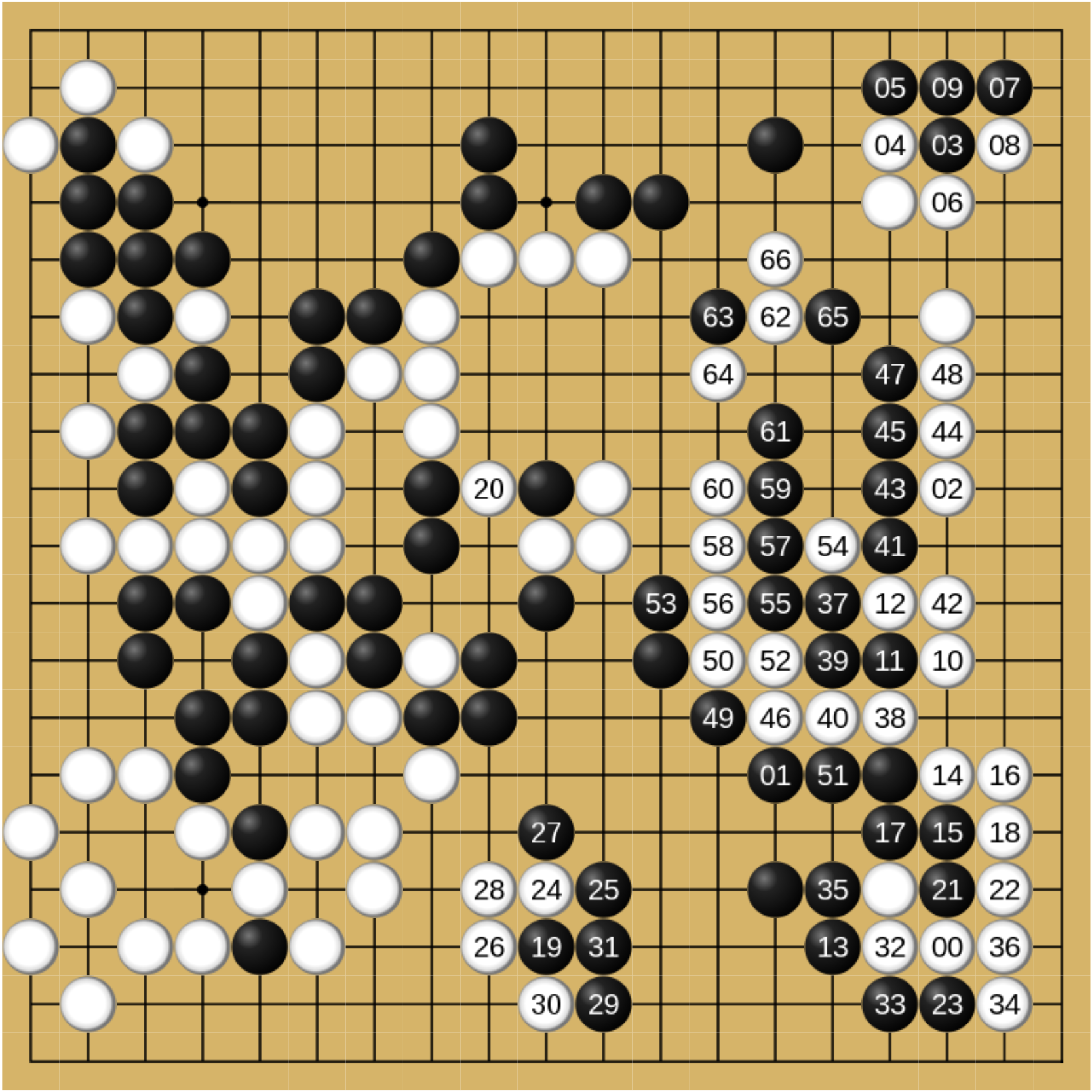}\\
      \vspace{\textwidth}
    \end{tabular}
     \caption[Fan Hui game 3]{Game 3 Fan Hui (black) vs.\ AlphaGo (white): move 1--99, and
       move 100--166. AlphaGo won by resignation.}\label{fig:gamesfanhui3}
  \end{minipage}
  \qquad
  \begin{minipage}{0.48\textwidth}
    \begin{tabular}{c}
      \includegraphics[width=\textwidth]{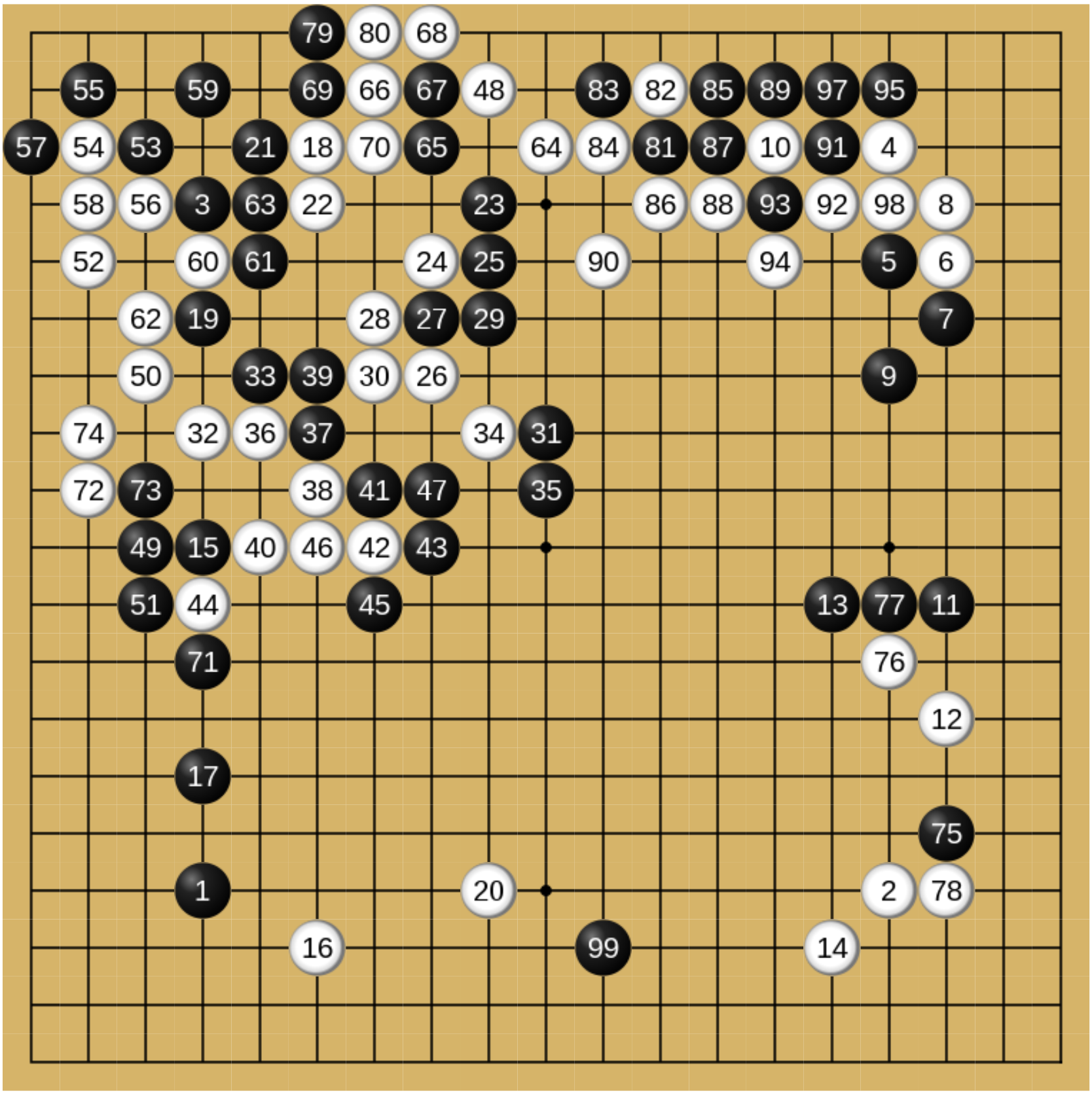}\\
      \includegraphics[width=\textwidth]{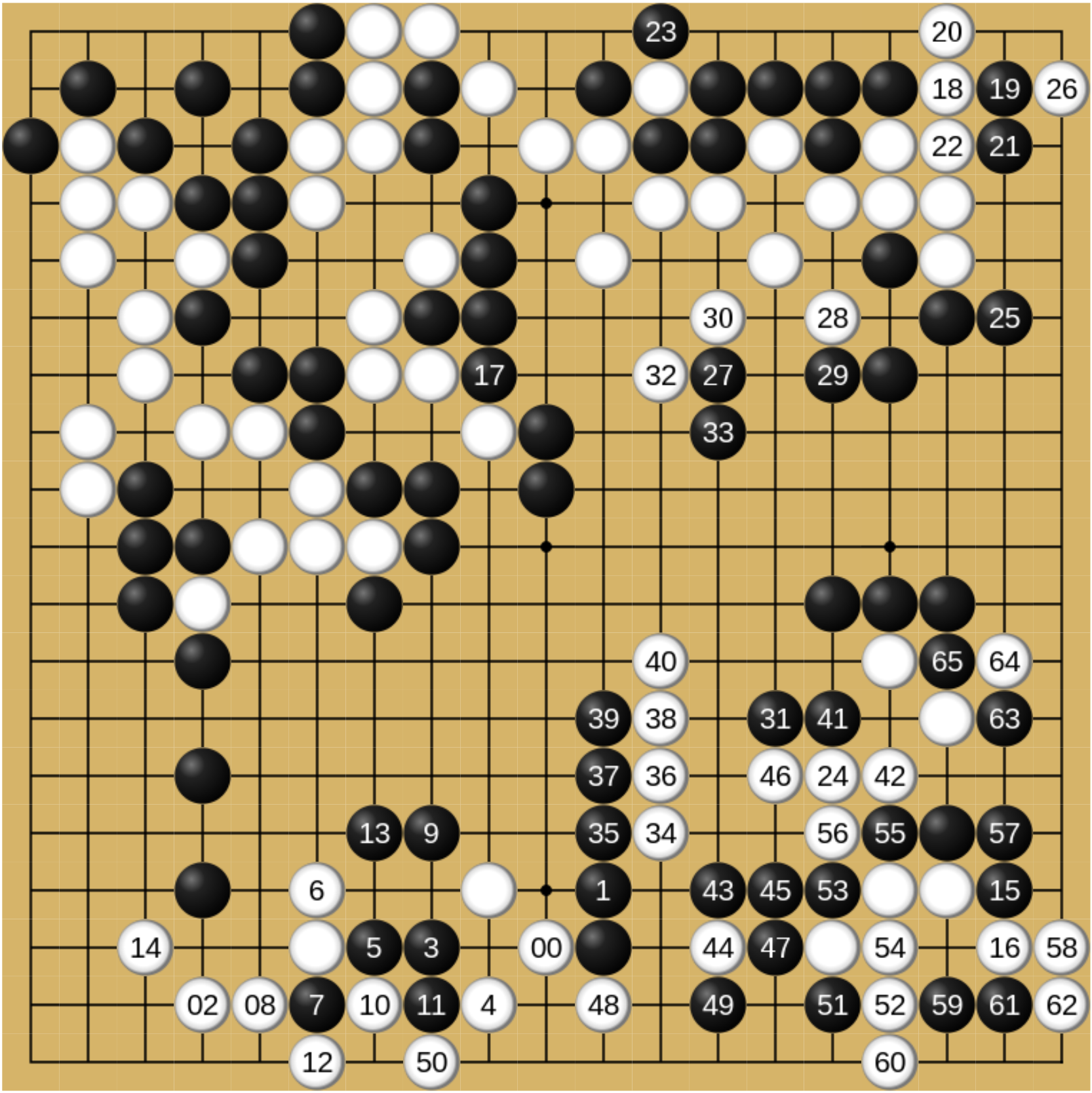}\\
      \vspace{\textwidth}
    \end{tabular}
     \caption[Fan Hui game 4]{Game 4  AlphaGo (black) vs.\ Fan Hui (white): move
       1--99 (move 96 at 10), and
       move 100--165.  AlphaGo won by resignation.}\label{fig:gamesfanhui4}
  \end{minipage}
\end{figure}

\begin{figure}[p]
    \begin{minipage}{0.48\textwidth}
    \begin{tabular}{c}
      \includegraphics[width=\textwidth]{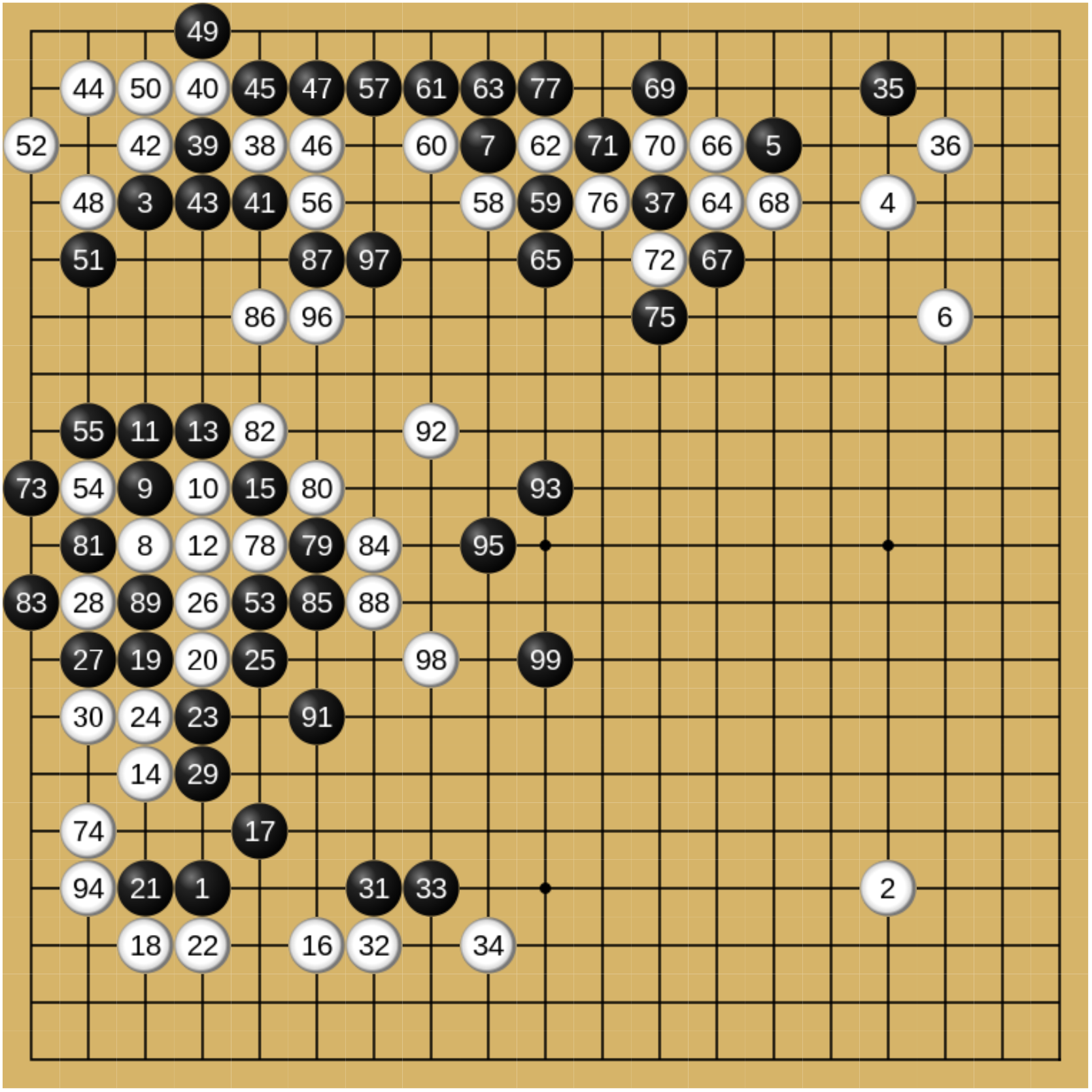}\\
      \includegraphics[width=\textwidth]{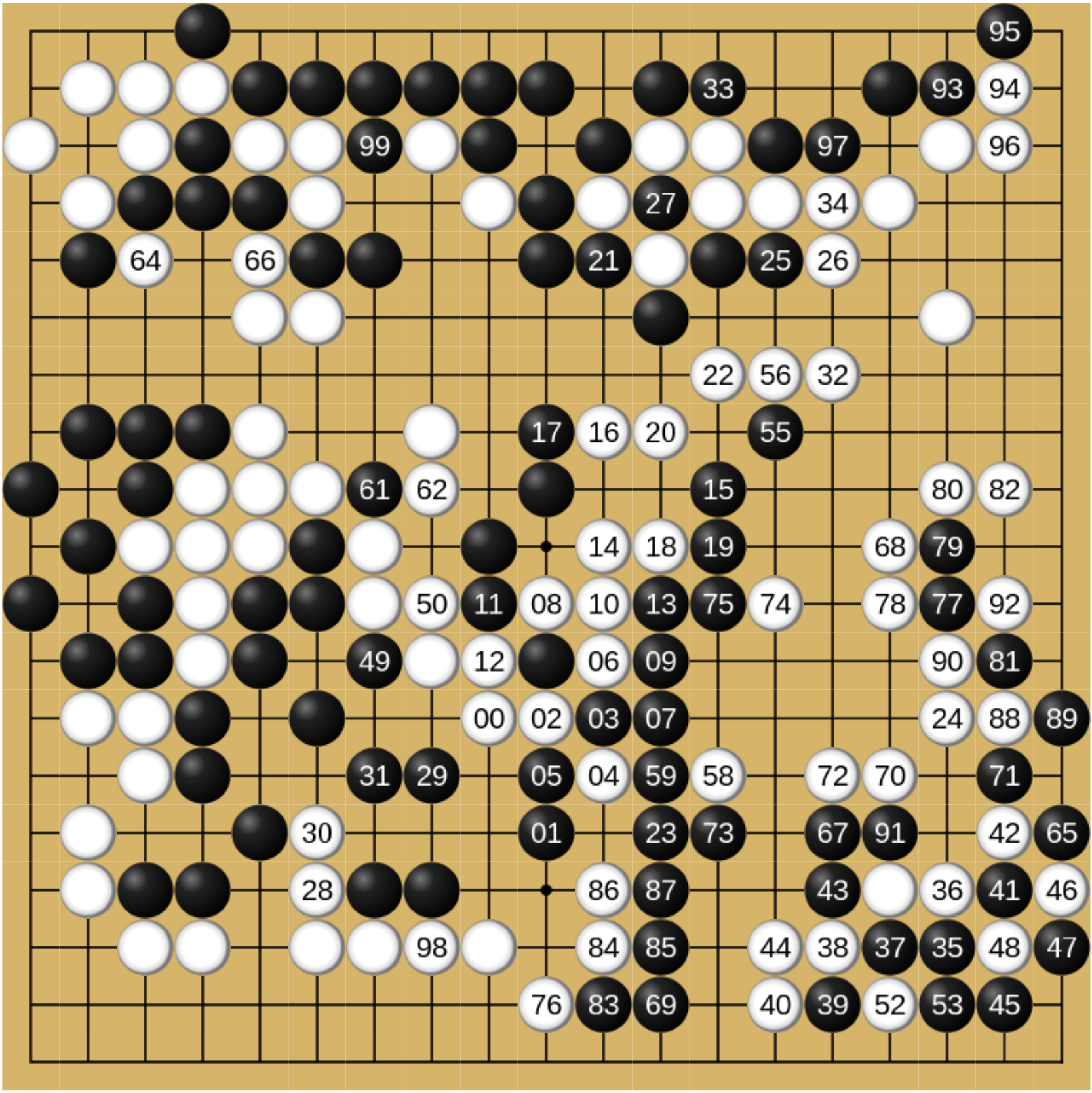}\\
      \includegraphics[width=\textwidth]{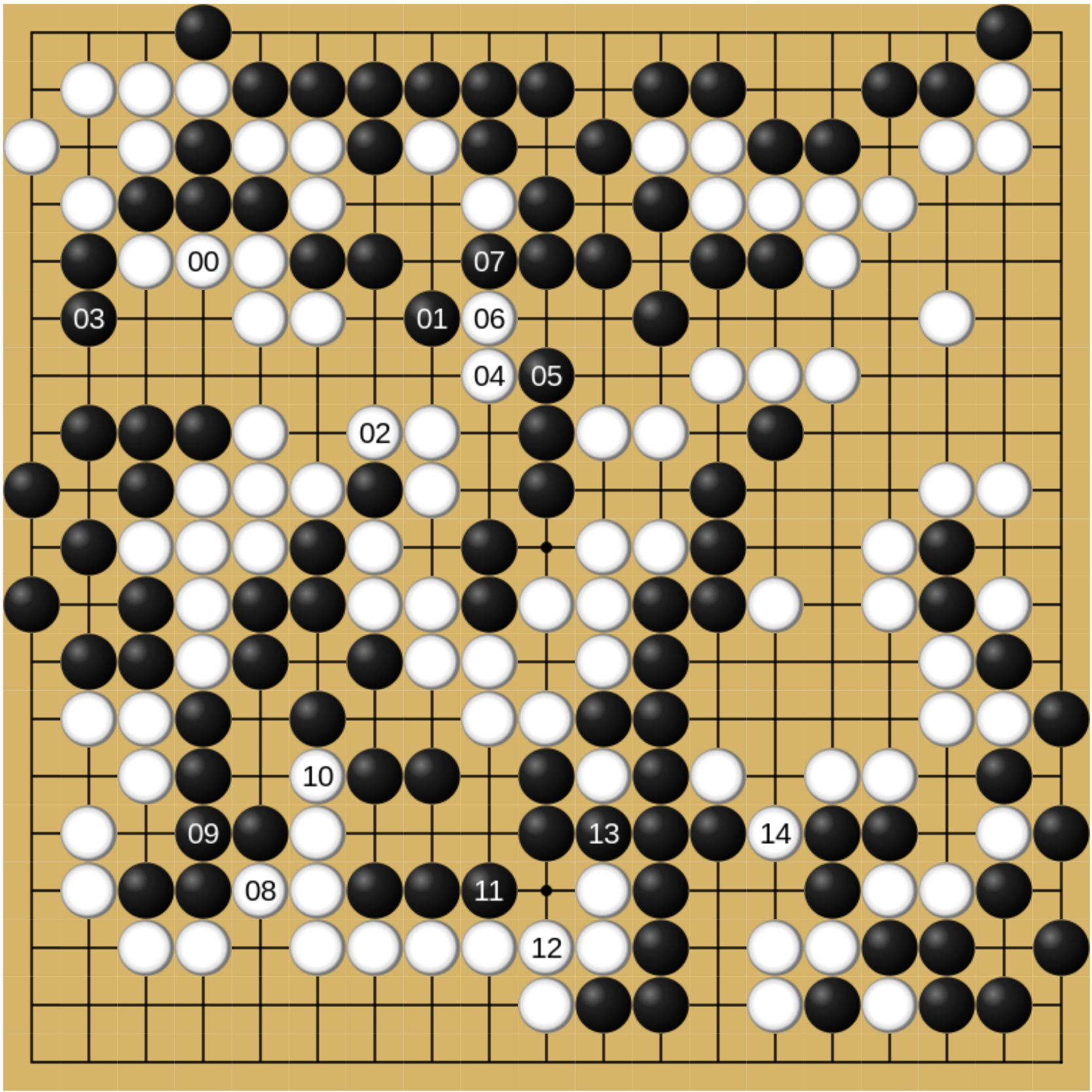}\\
    \end{tabular}
     \caption[Fan Hui game 5]{Game 5 Fan Hui (black) vs.\ AlphaGo (white): move 1--99,
       move 100--199 (moves 151/157/163 at 141, 154/160 at 148), and move 200--214.  AlphaGo won by resignation.}\label{fig:gamesfanhui5}
  \end{minipage}
  \qquad
  \begin{minipage}{0.48\textwidth}
    \begin{tabular}{c}
      \vspace{\textwidth}\\
      \vspace{\textwidth}\\
      \vspace{\textwidth}
    \end{tabular}
  \end{minipage}
\end{figure}

In October 2015 European Champion  Fan Hui 2p played the following games against
AlphaGo in London. AlphaGo won 5-0. See the games in
Figs.~\ref{fig:gamesfanhui1}--\ref{fig:gamesfanhui5}.

\section*{Lee Sedol}
\begin{figure}[p]
    \begin{minipage}{0.48\textwidth}
    \begin{tabular}{c}
      \includegraphics[width=\textwidth]{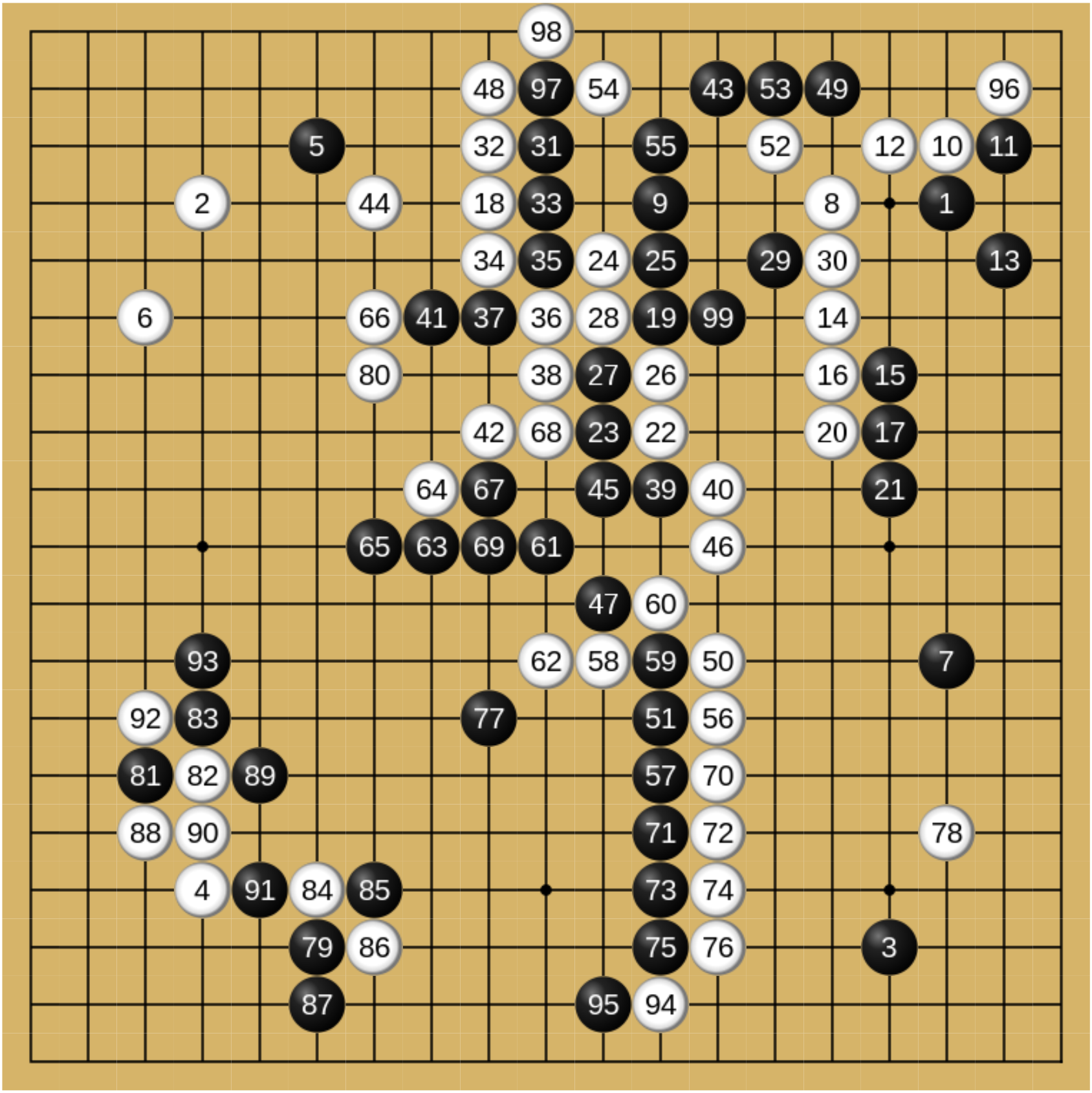}\\
      \includegraphics[width=\textwidth]{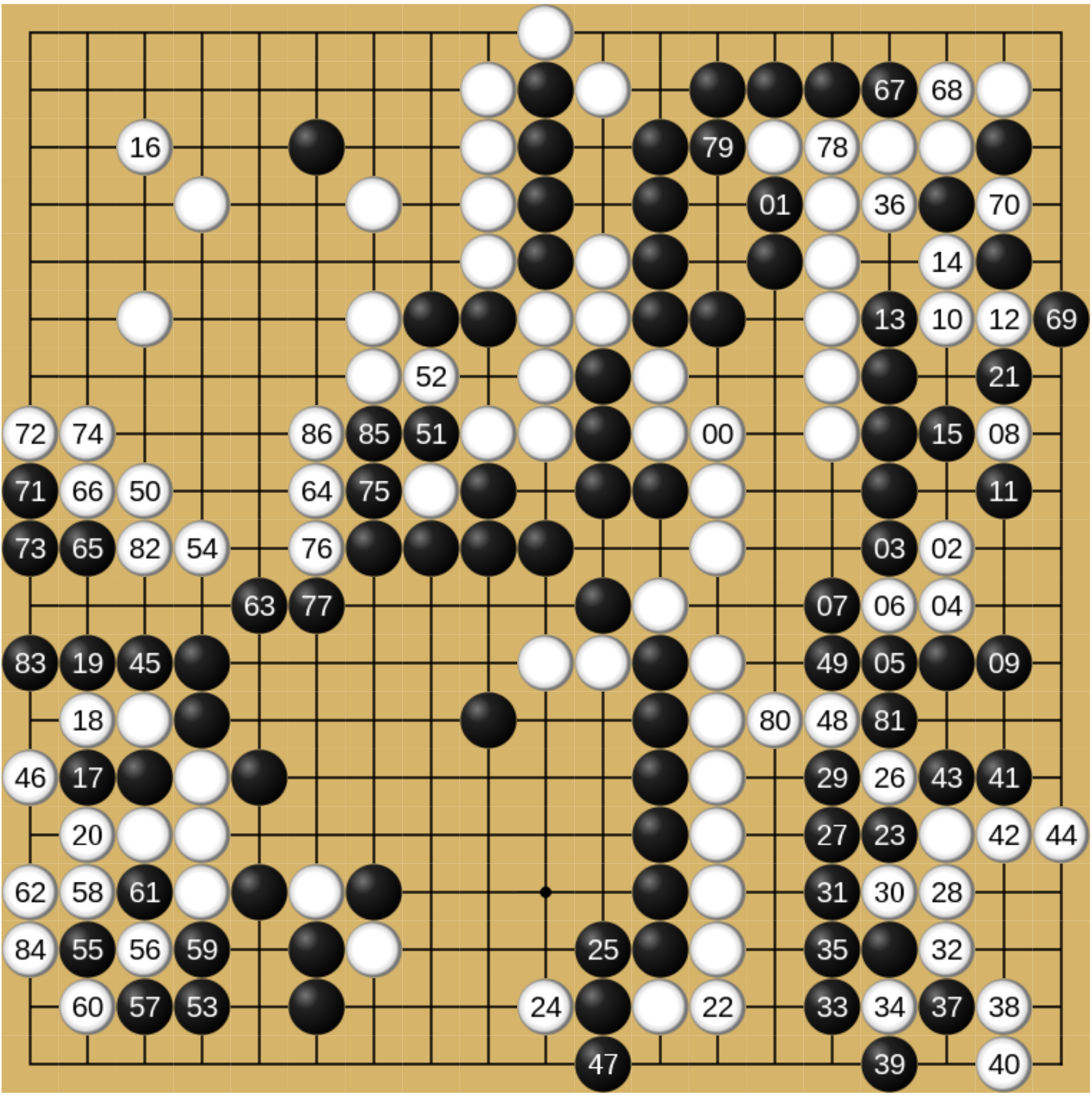}\\
      \vspace{\textwidth}
    \end{tabular}
     \caption[Lee Sedol game 1]{Game 1 Lee Sedol (black) vs.\ AlphaGo (white): move
       1--99, and
       move 100--186.   AlphaGo won by resignation.}\label{fig:gamesleesedol1}
  \end{minipage}
  \qquad
  \begin{minipage}{0.48\textwidth}
    \begin{tabular}{c}
      \includegraphics[width=\textwidth]{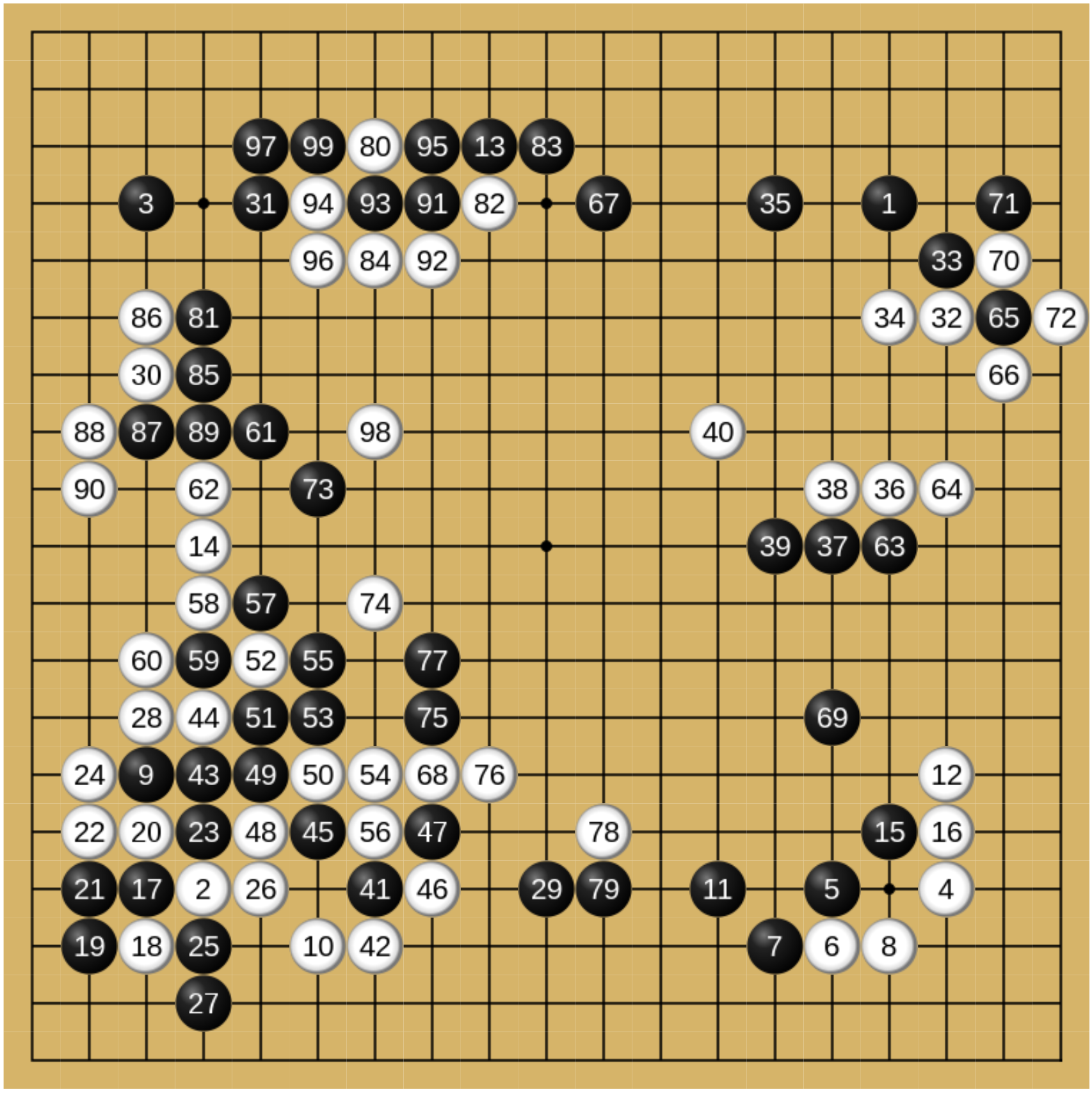}\\
      \includegraphics[width=\textwidth]{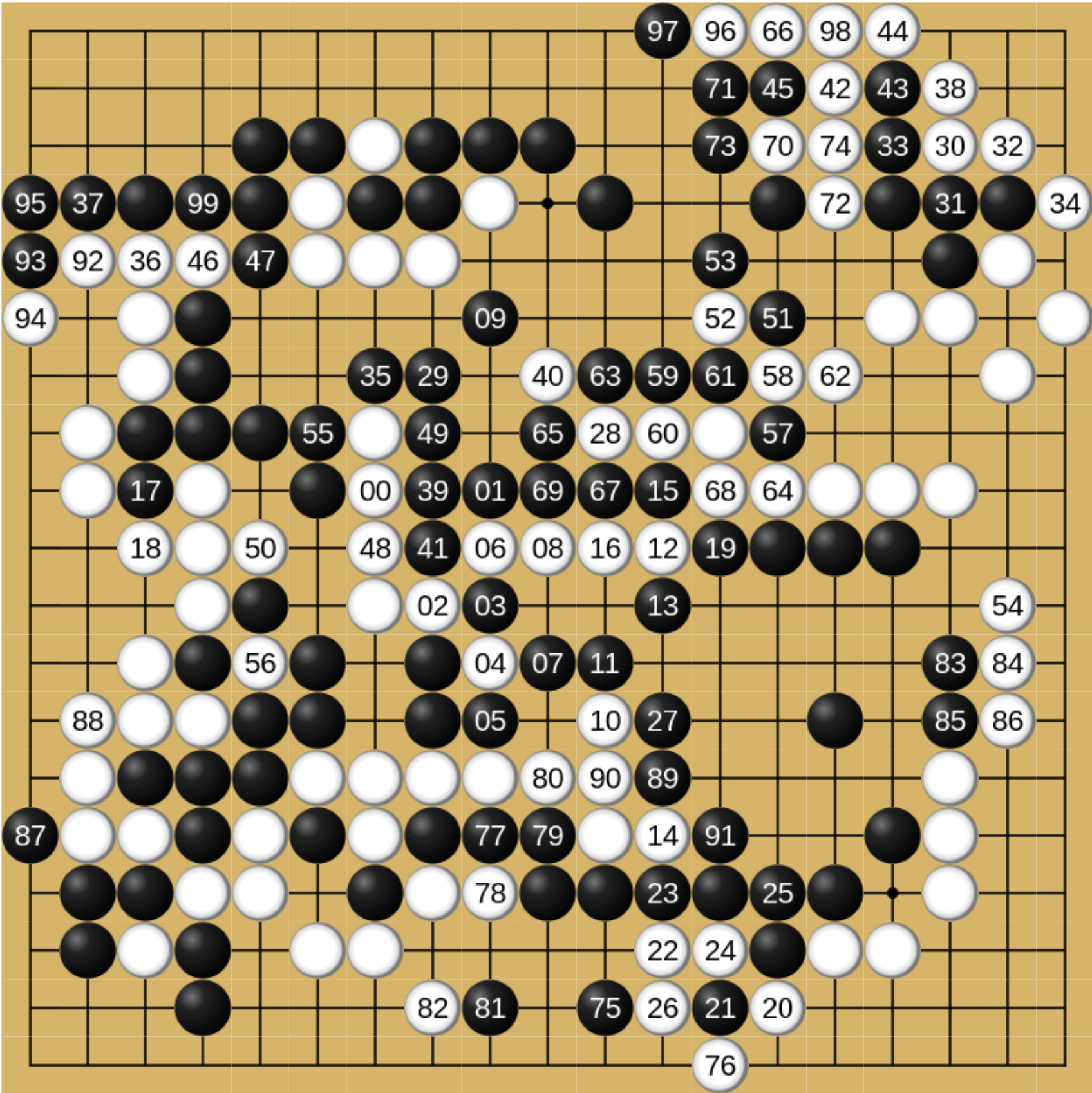}\\
      \includegraphics[width=\textwidth]{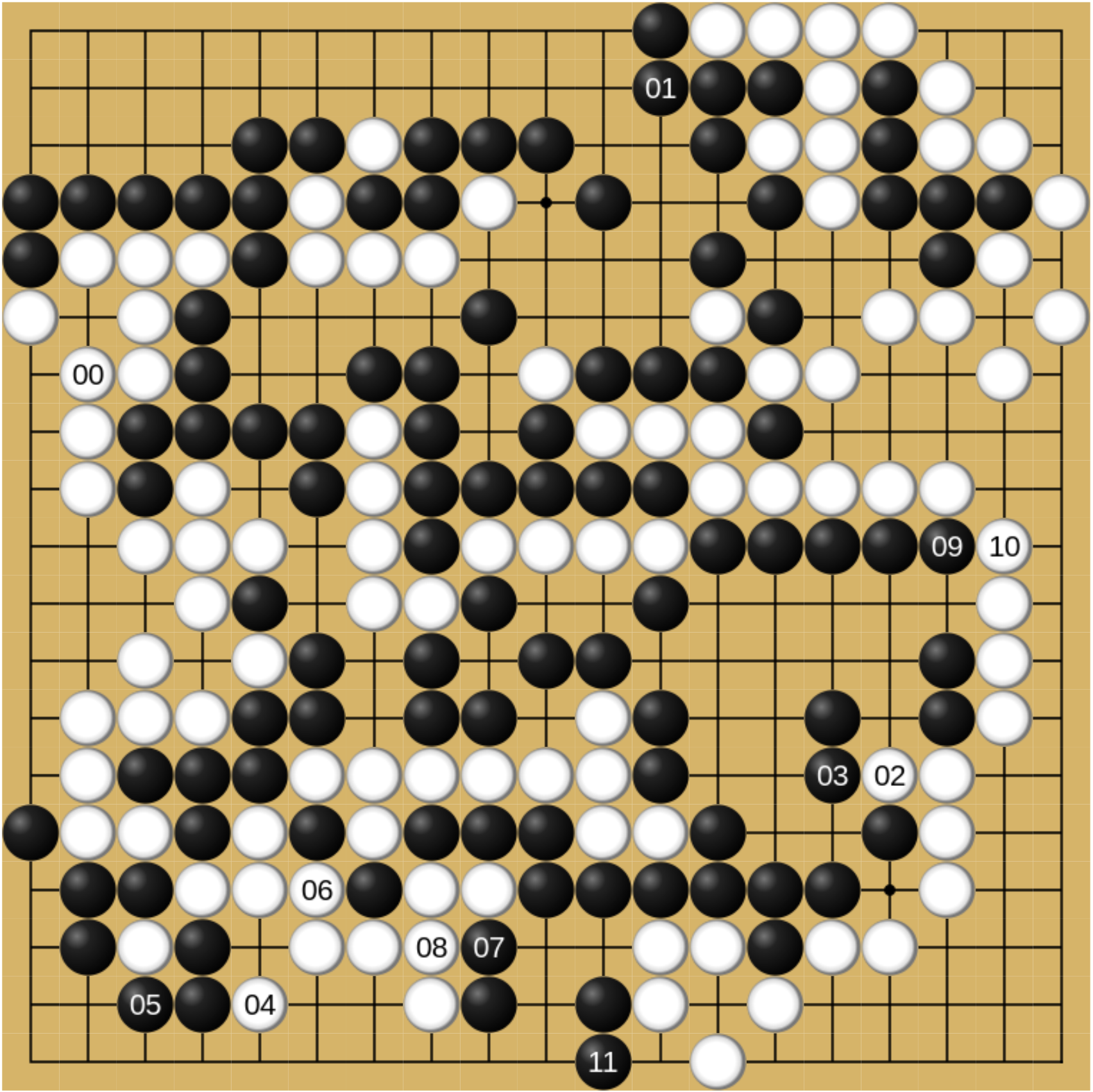}\\
    \end{tabular}
 \caption[Lee Sedol game 2]{Game 2  AlphaGo (black) vs.\ Lee Sedol (white): move
       1--99,
       move 100--199, and move 200--211. Move 37 is a famous
      move by AlphaGo. AlphaGo won by resignation.}\label{fig:gamesleesedol2}
  \end{minipage}
\end{figure}

\begin{figure}[p]
    \begin{minipage}{0.48\textwidth}
    \begin{tabular}{c}
      \includegraphics[width=\textwidth]{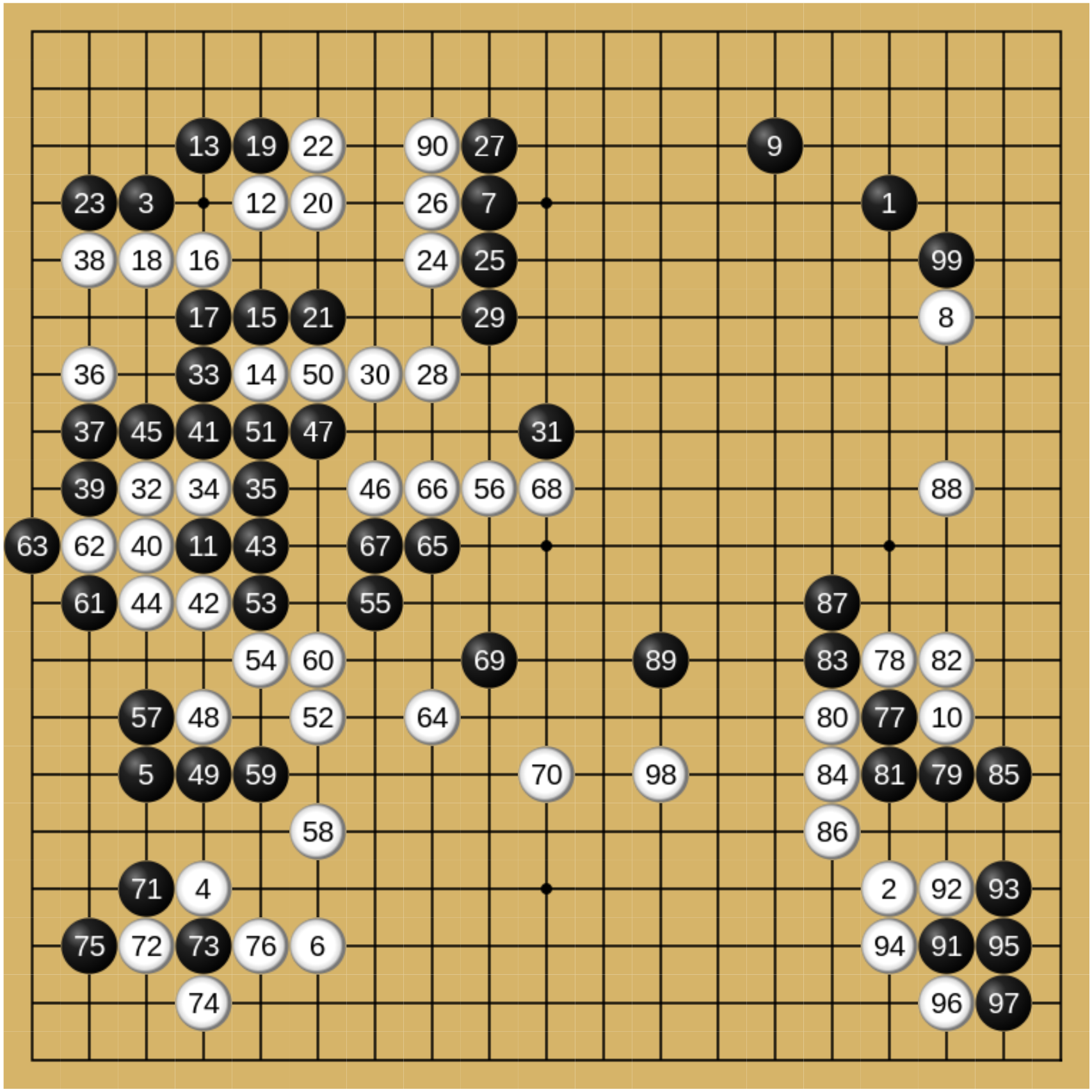}\\
      \includegraphics[width=\textwidth]{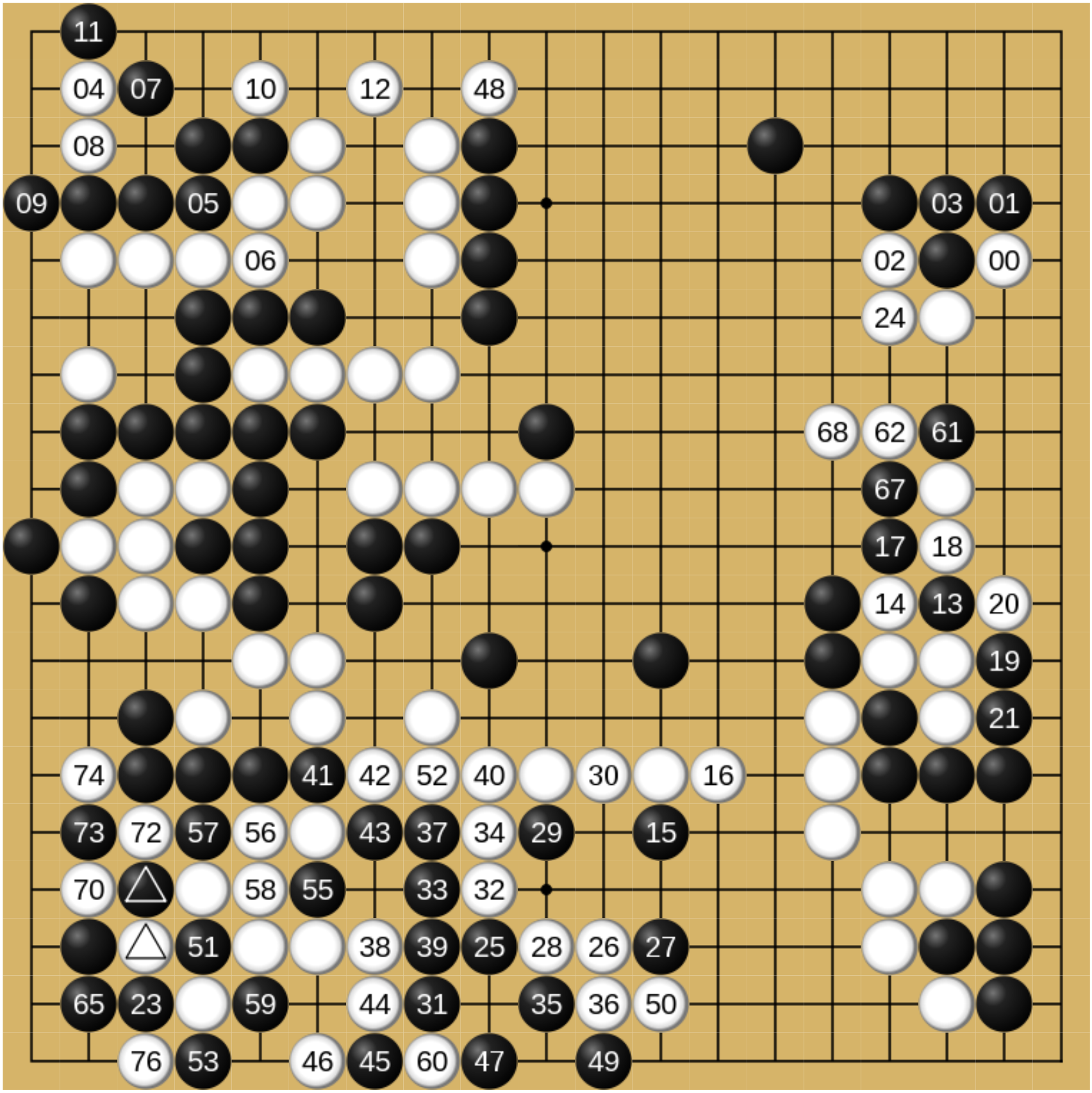}\\
      \vspace{\textwidth}
    \end{tabular}
       \caption[Lee Sedol game 3]{Game 3 Lee Sedol (black) vs.\ AlphaGo (white): move 1--99, and
       move 100--176 (move 122 at 113, 154 at white triangle, 163 at 145, 164 at 151,
166 and 171 at 160, 169 at 145, 175 at black triangle). AlphaGo won by
resignation.}\label{fig:gamesleesedol3}
  \end{minipage}
  \qquad
  \begin{minipage}{0.48\textwidth}
    \begin{tabular}{c}
      \includegraphics[width=\textwidth]{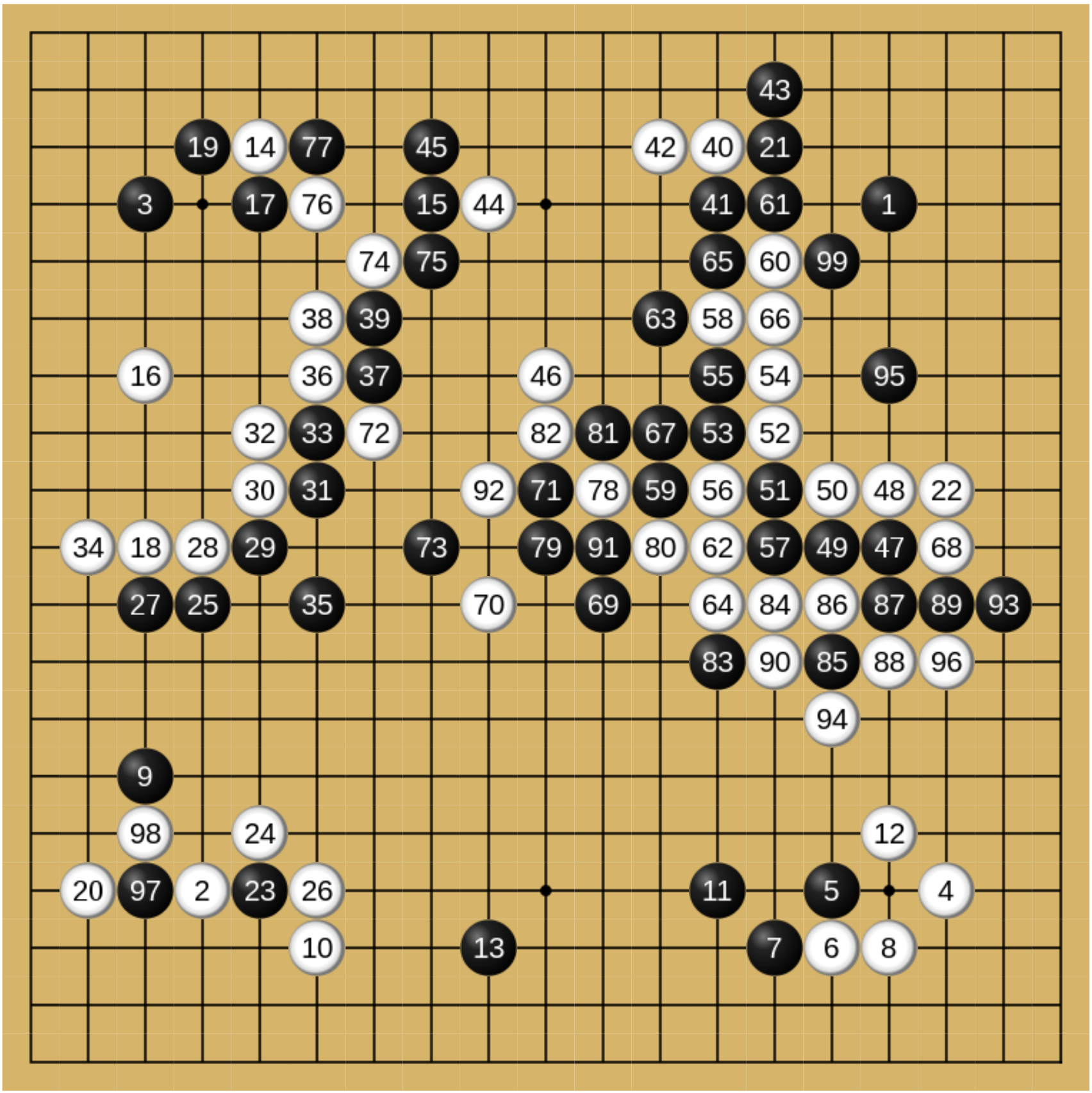}\\
      \includegraphics[width=\textwidth]{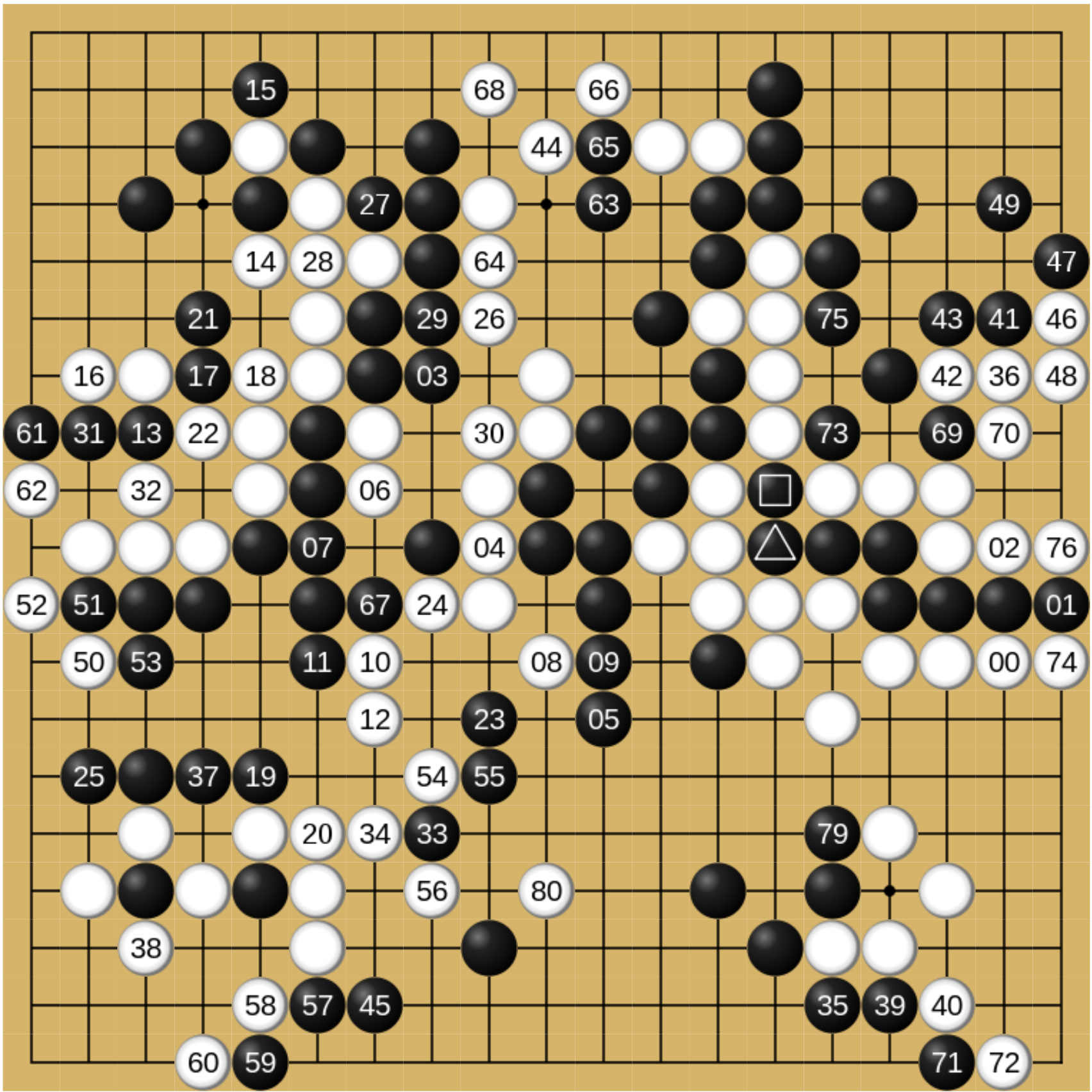}\\
      \vspace{\textwidth}
    \end{tabular}
     \caption[Lee Sedol game 4]{Game 4  AlphaGo (black) vs.\ Lee Sedol (white): move
       1--99, and
       move 100--180 (move 177 at square, 178 at triangle). Move 78 by
       Lee Sedol is considered a brilliant move. Lee Sedol won by resignation.}\label{fig:gamesleesedol4}
  \end{minipage}
\end{figure}

\begin{figure}[p]
    \begin{minipage}{0.48\textwidth}
    \begin{tabular}{c}
      \includegraphics[width=\textwidth]{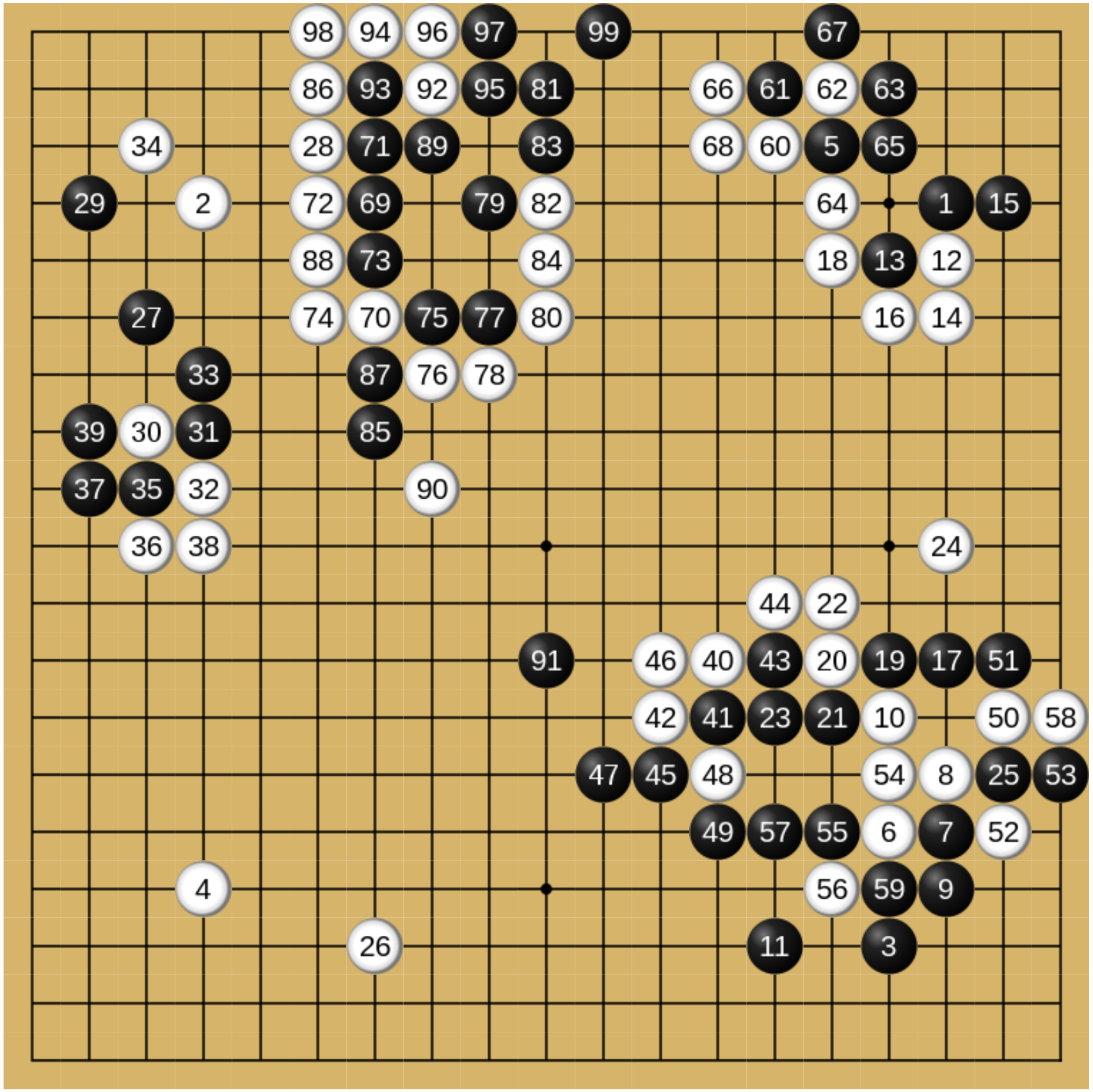}\\
      \includegraphics[width=\textwidth]{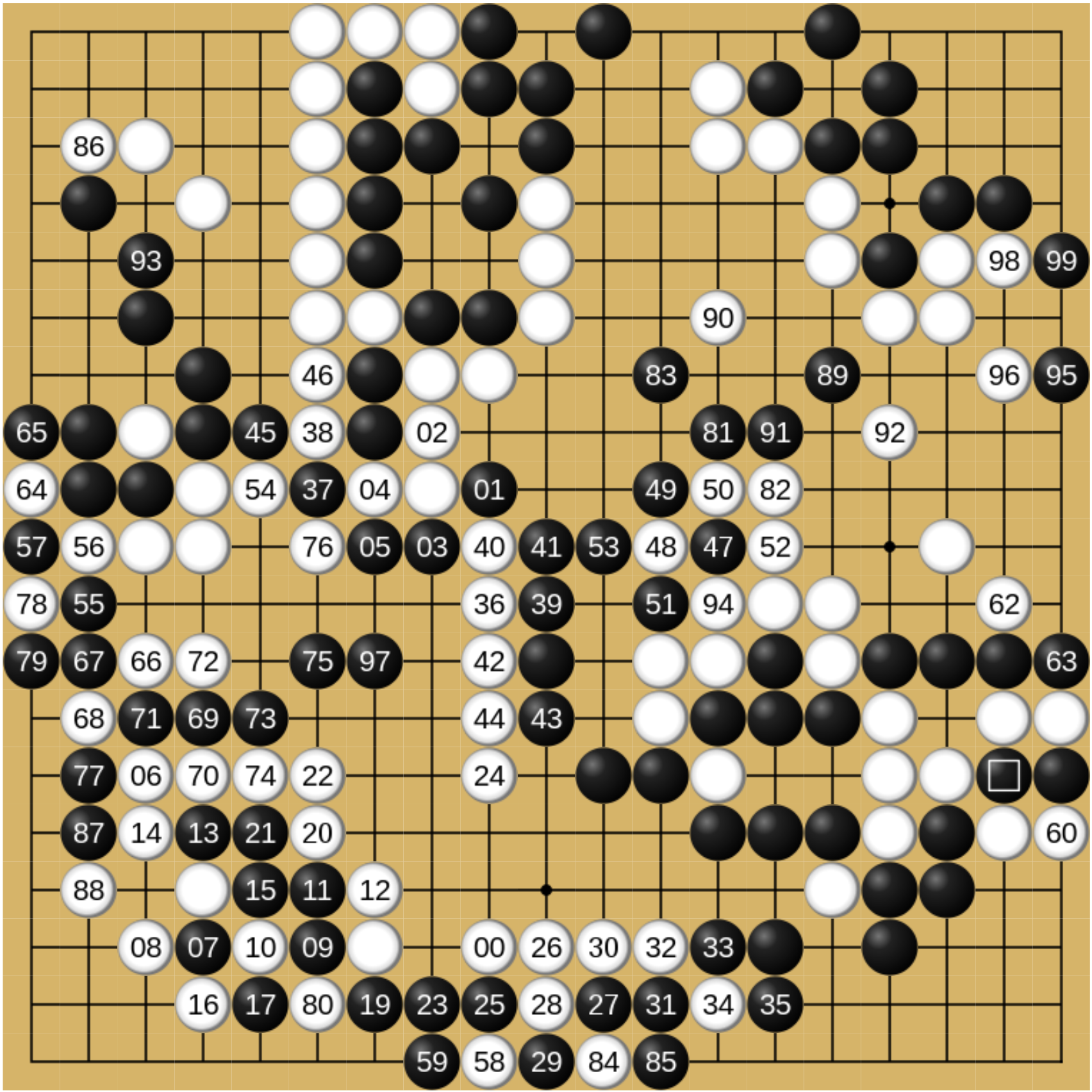}\\
      \includegraphics[width=\textwidth]{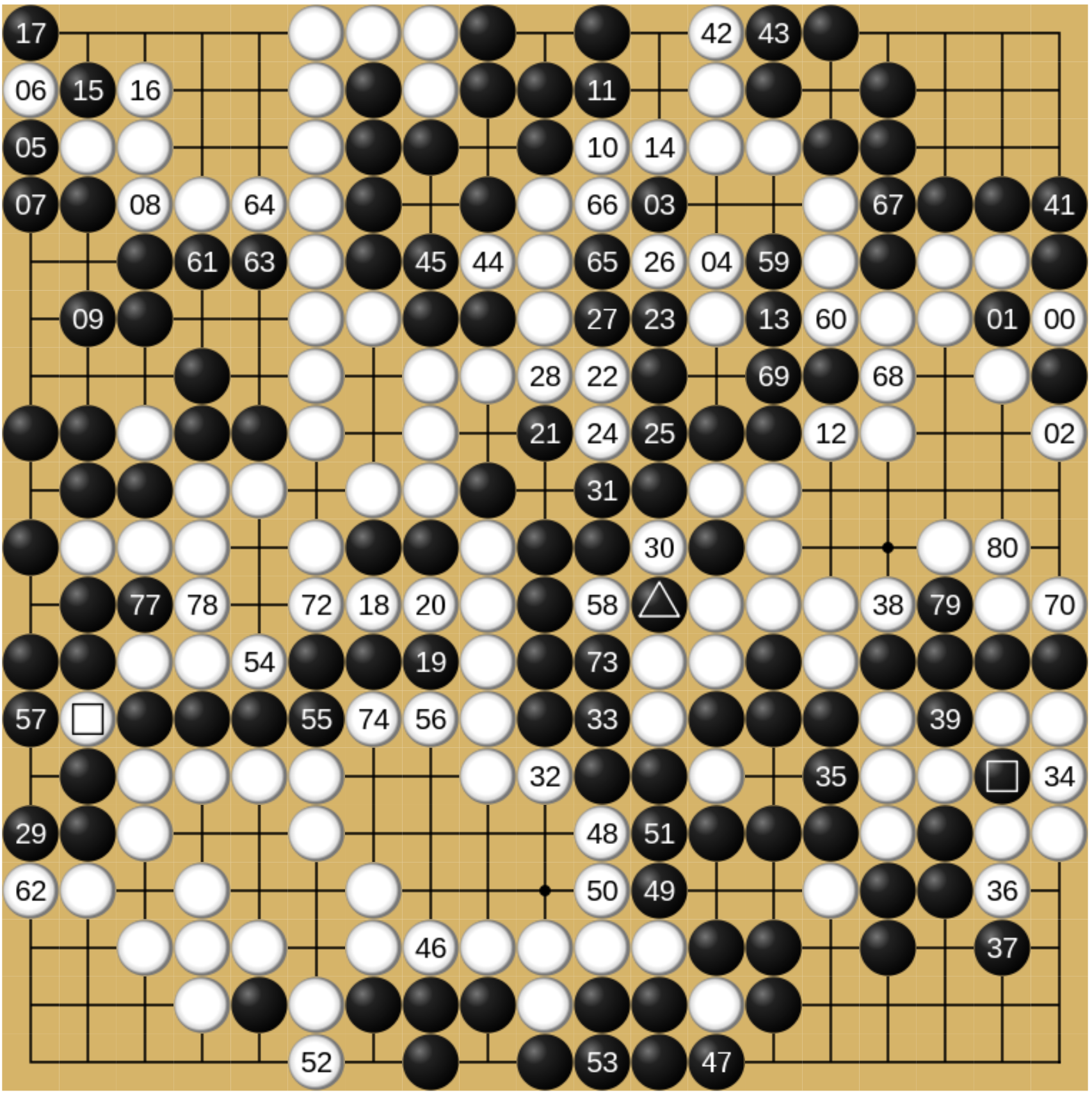}\\
    \end{tabular}
 \caption[Lee Sedol game 5]{Game 5 Lee Sedol (black) vs.\ AlphaGo (white): move 1-99,
       move 100-199 (118 at 107, 161 at square), and move 200-280
       (240 at 200, 271 at black square, 275 at white square, 276 at black triangle).  AlphaGo won by resignation.}\label{fig:gamesleesedol5}  
  \end{minipage}
  \qquad
  \begin{minipage}{0.48\textwidth}
    \begin{tabular}{c}
      \vspace{\textwidth}\\
      \vspace{\textwidth}\\
      \vspace{\textwidth}
    \end{tabular}
  \end{minipage}
\end{figure}

In March 2016 Lee Sedol 9p played the following games against
AlphaGo in Seoul. AlphaGo won 4-1.  See
Figs.~\ref{fig:gamesleesedol1}--\ref{fig:gamesleesedol5}.

\section*{Ke Jie}
\begin{figure}[p]
    \begin{minipage}{0.48\textwidth}
    \begin{tabular}{c}
      \includegraphics[width=\textwidth]{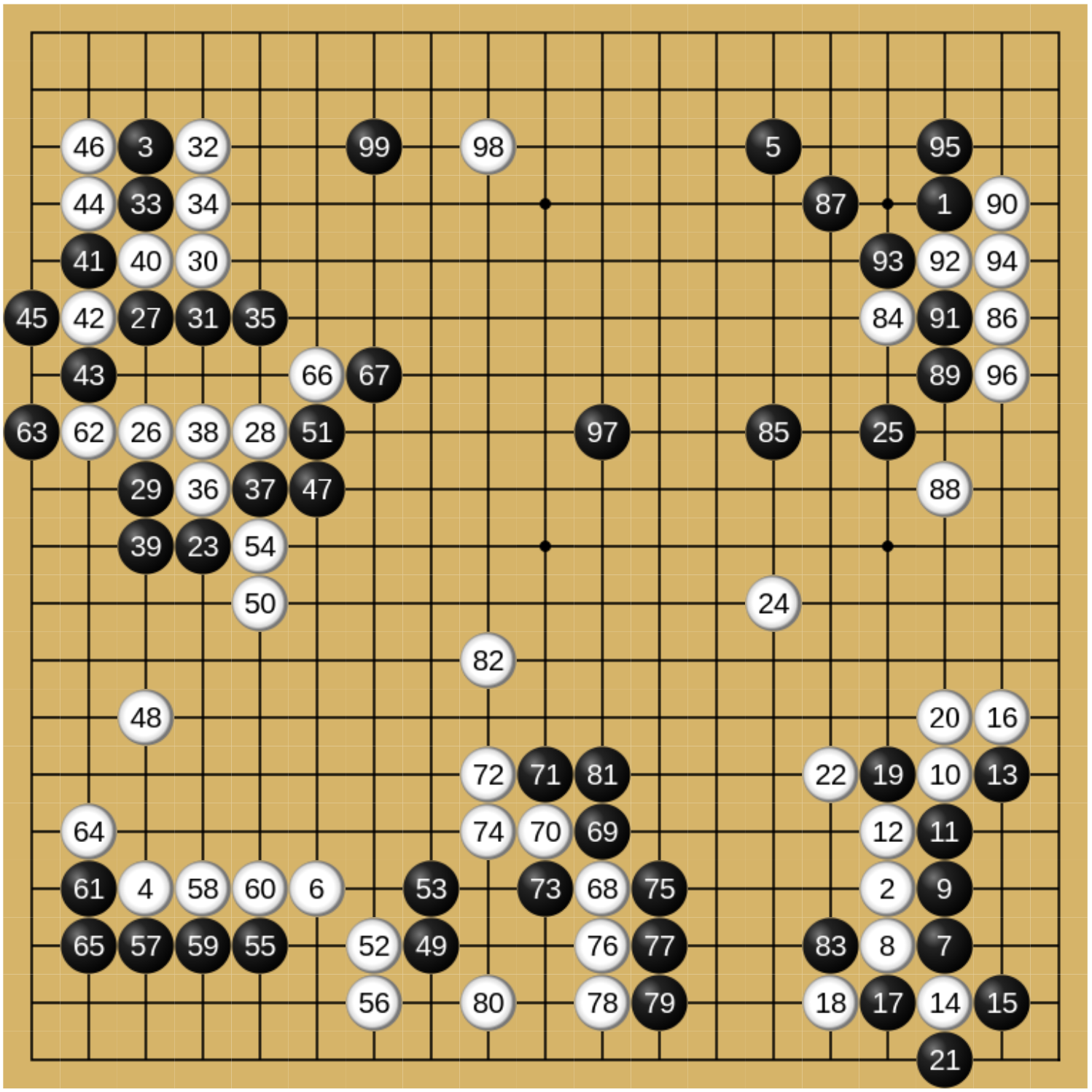}\\
      \includegraphics[width=\textwidth]{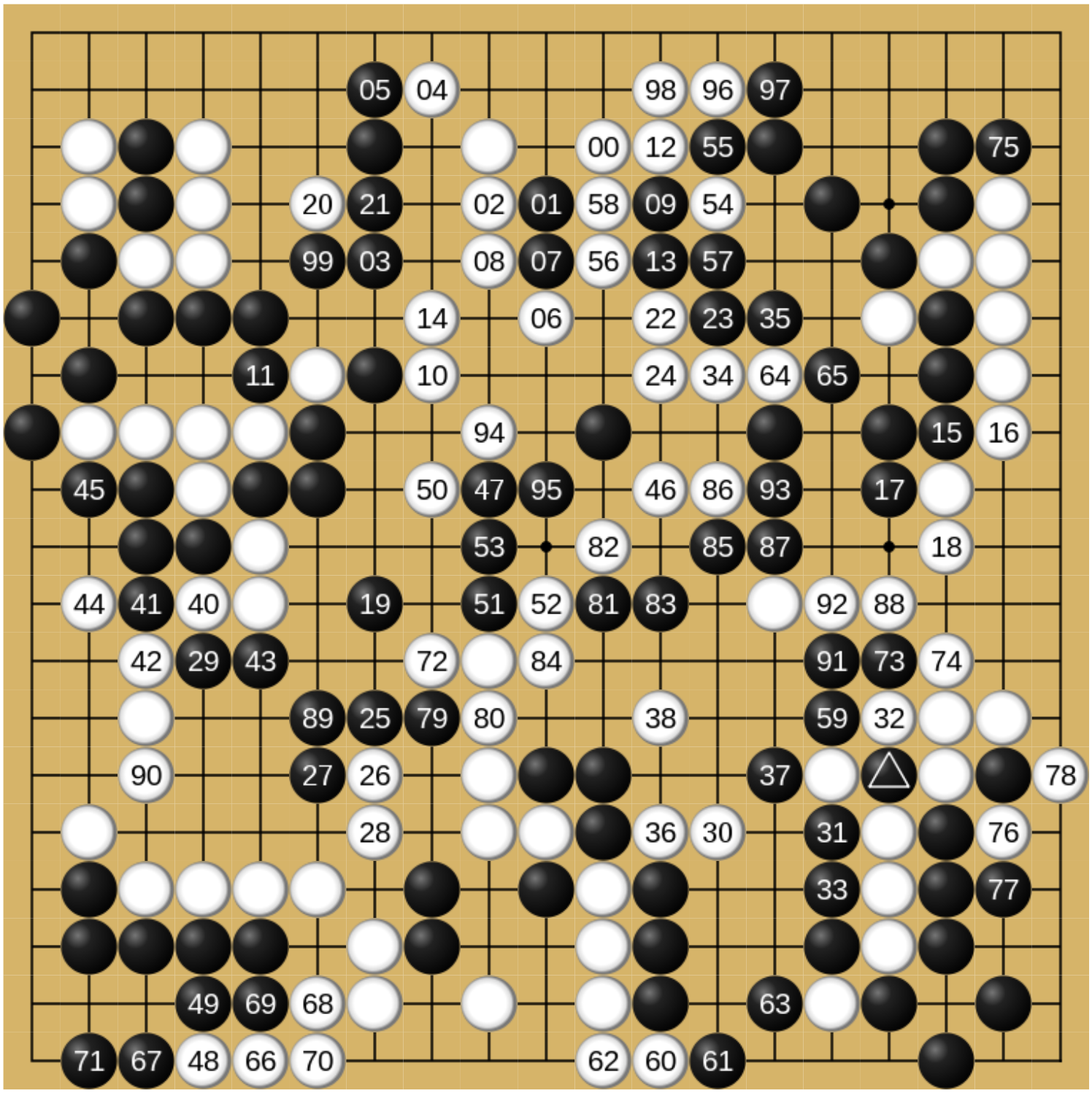}\\
      \includegraphics[width=\textwidth]{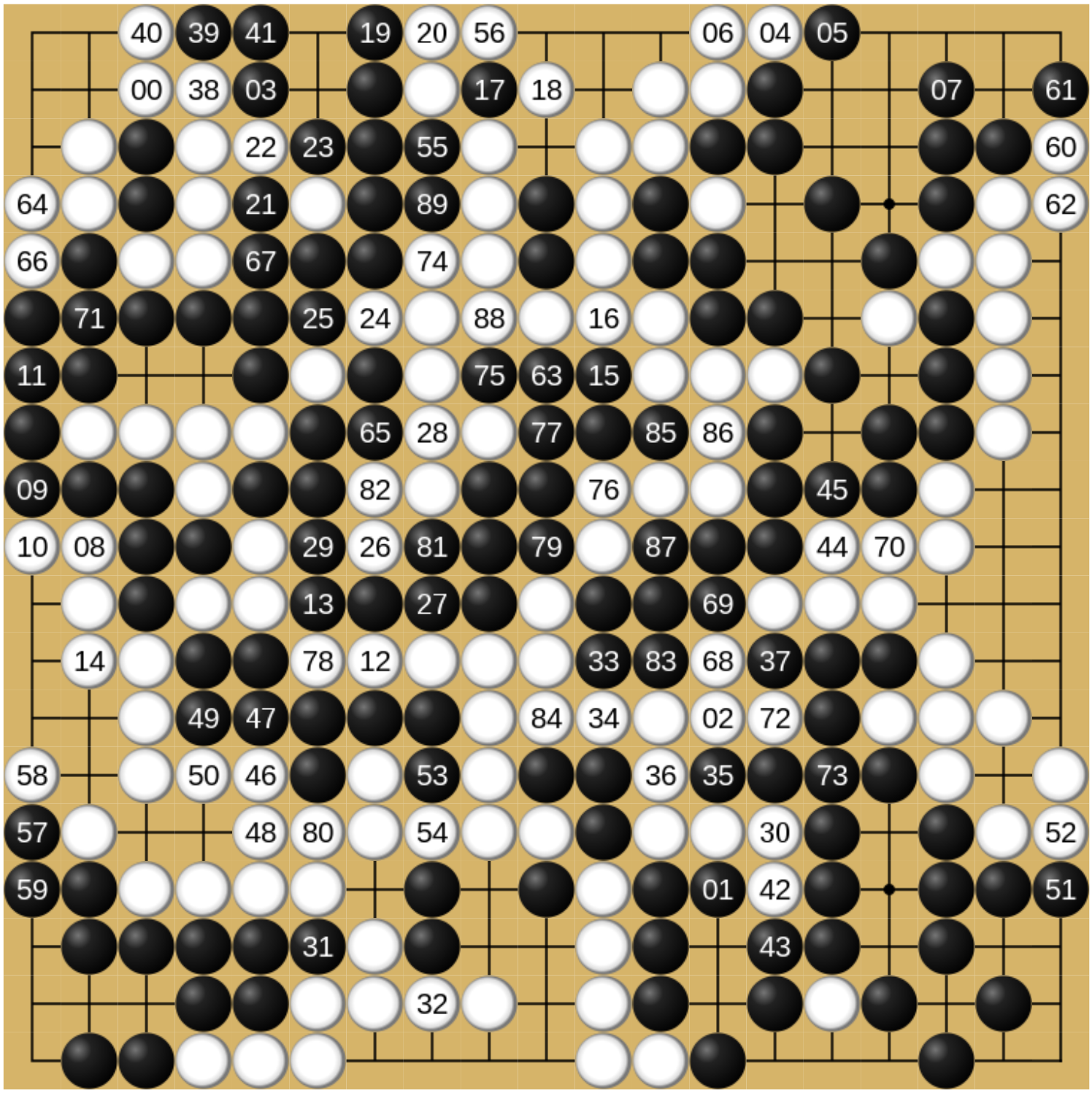}\\
    \end{tabular}
     \caption[Ke Jie game 1]{Game 1 Ke Jie (black) vs.\ AlphaGo (white): move 1-99,
       move 100-199 (139 at triangle), and move 200-289.  AlphaGo won by resignation.}\label{fig:gameskejie1}
  \end{minipage}
  \qquad
  \begin{minipage}{0.48\textwidth}
    \begin{tabular}{c}
      \includegraphics[width=\textwidth]{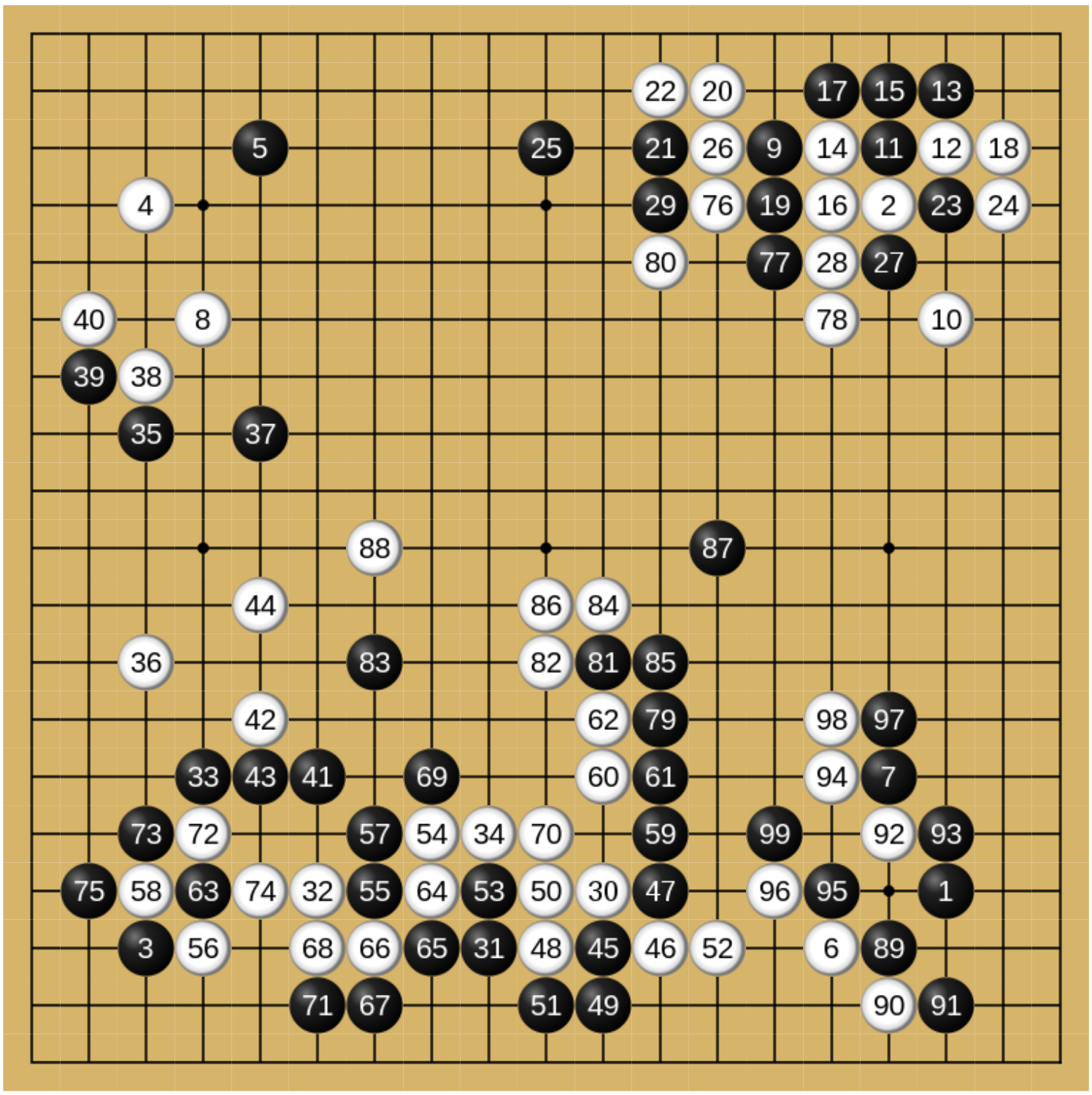}\\
      \includegraphics[width=\textwidth]{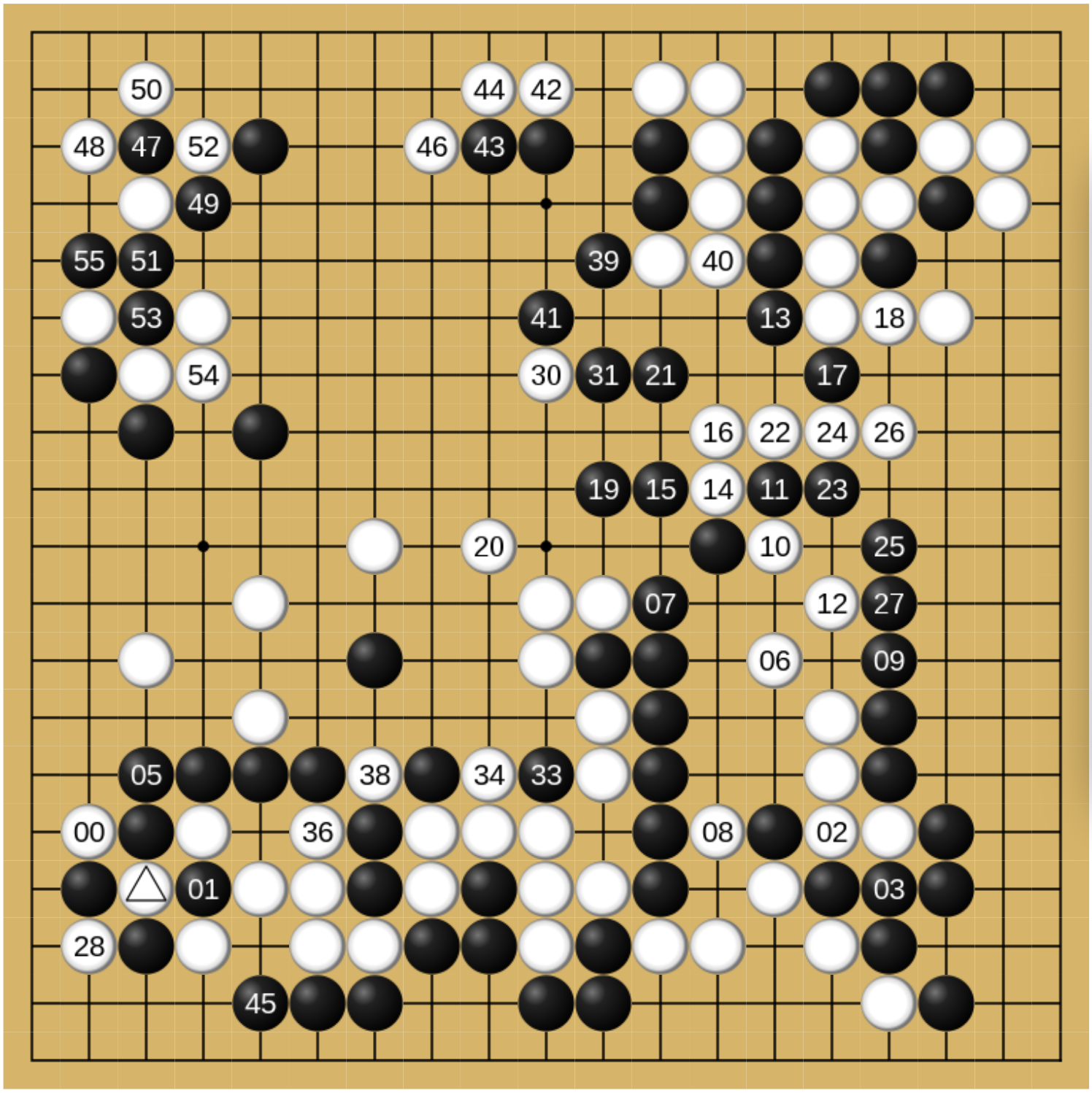}\\
      \vspace{0.96\textwidth}
    \end{tabular}
     \caption[Ke Jie game 2]{Game 2 AlphaGo (black) vs.\ Ke Jie (white): move 1-99,
       and move 100-155 (104/132/137 at triangle, 129/135 at 101).
       AlphaGo won by resignation.}\label{fig:gameskejie2}
  \end{minipage}
\end{figure}

\begin{figure}[p]
    \begin{minipage}{0.48\textwidth}
    \begin{tabular}{c}
      \includegraphics[width=\textwidth]{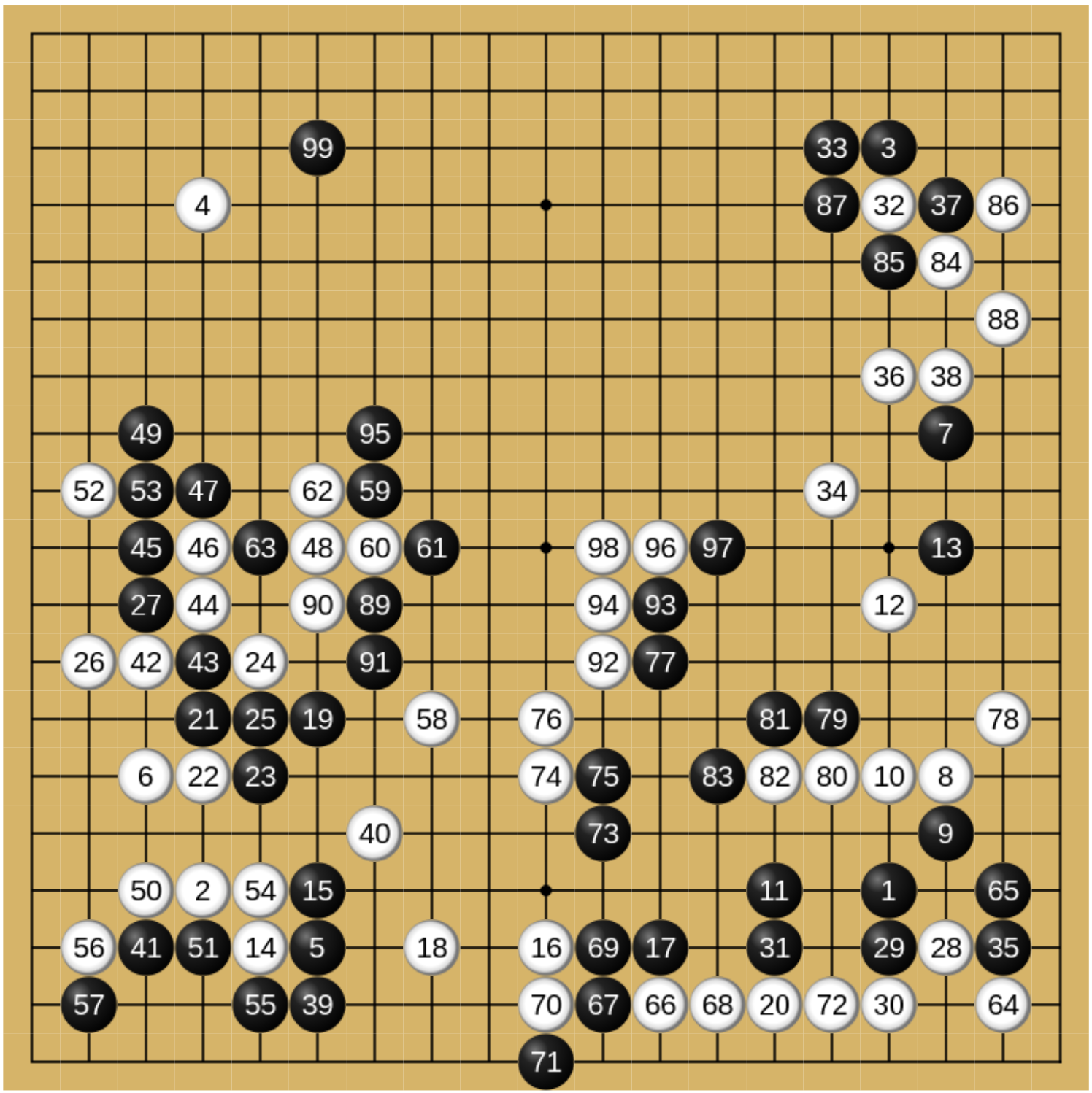}\\
      \includegraphics[width=\textwidth]{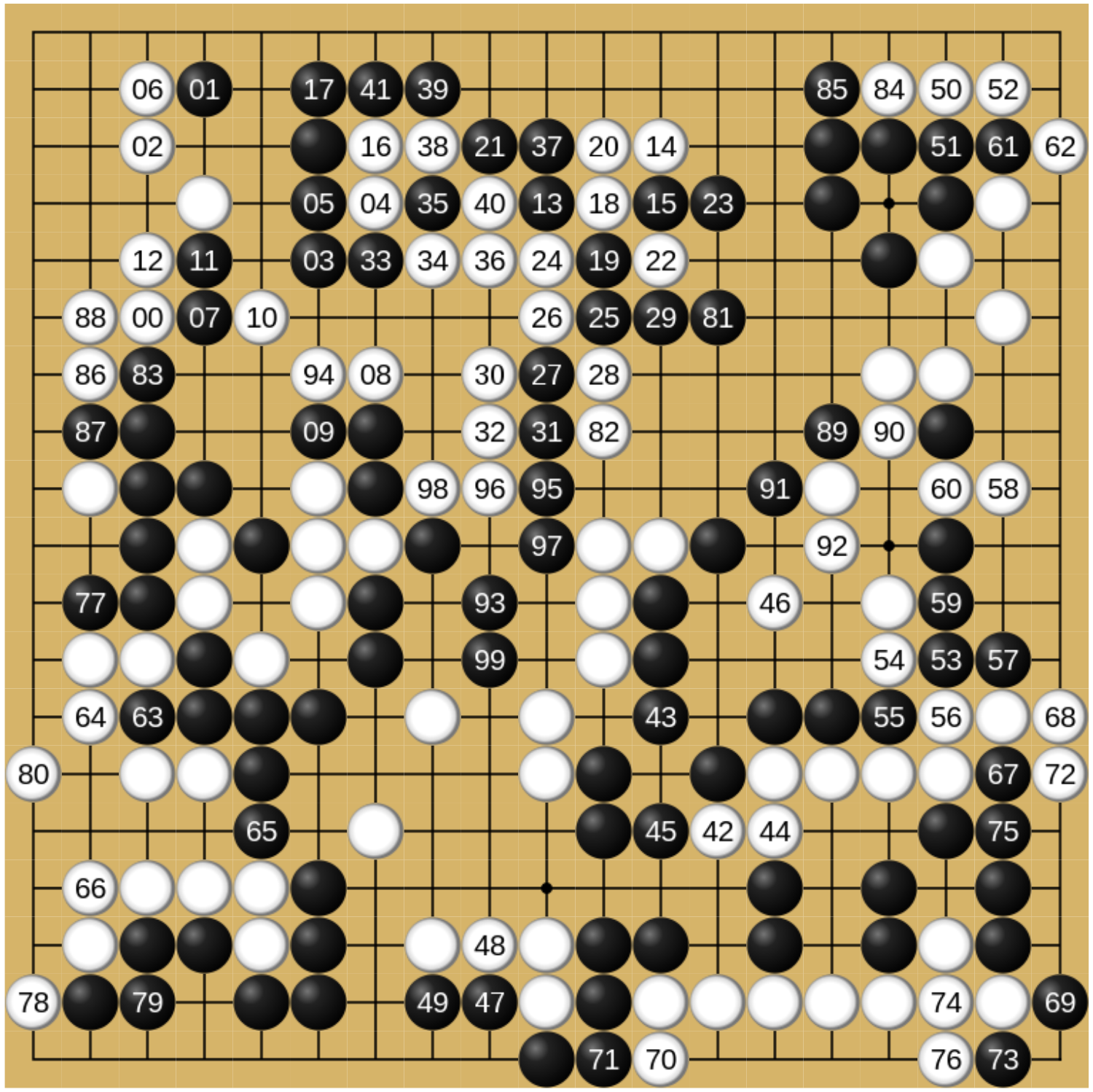}\\
      \includegraphics[width=\textwidth]{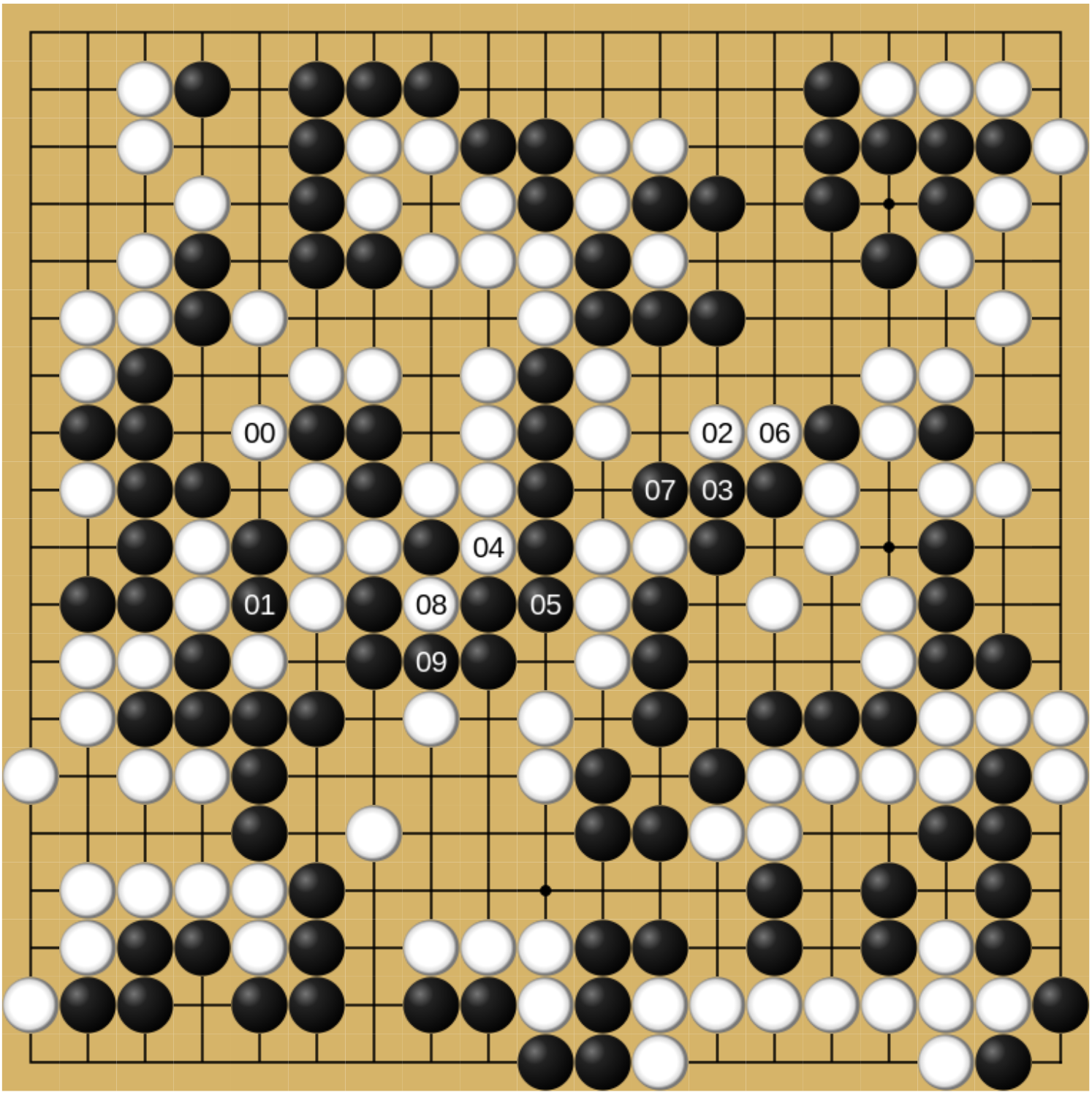}\\
    \end{tabular}
     \caption[Ke Jie game 3]{Game 3 AlphaGo (black) vs.\ Ke Jie (white): move 1-99,
       move 100-199, and move 200-209.
       AlphaGo won by resignation.}\label{fig:gameskejie3}
  \end{minipage}
  \qquad
  \begin{minipage}{0.48\textwidth}
    \begin{tabular}{c}
      \vspace{\textwidth}\\
      \vspace{\textwidth}\\
      \vspace{\textwidth}
    \end{tabular}
  \end{minipage}
\end{figure}

In May 2017 Ke Jie 9p played the following games against
AlphaGo in Wuzhen. AlphaGo won 3-0. See
Figs.~\ref{fig:gameskejie1}--\ref{fig:gameskejie3}. AlphaGo played
white in game  1 and black in game 2 and 3.

\chapter{Learning to Play and Program Go and Chess}\label{chap:resources}\index{Go}\index{Chess}
Go and Chess are challenging games. They challenge computers as well
as humans, and they provide
endless fun and enjoyment. 
If you would like to learn to play  Go or Chess,
then this appendix provides links to help you get started in the world of
game playing. If you would like to learn to program these games, then
you will find pointers here as well.

\section{Learning to Play}
\subsection{Go}
The best and most fun way to learn to play Go is by playing against
people.  Find a local club and go. Go is played by millions of people,
and there are many clubs around the world. Your school may have a Go club.
Search for ``Go Club'' on the internet and you will find
nearby clubs where you will find people to play Go with, and you will start
to learn this fascinating game.
Once you make progress, you may even want to join one of the national Go
associations. All addresses can be found easily online.

There are also many online possibilities to play
Go. Table~\ref{tab:servers} lists a few of the better known online Go
servers. Just click on the link, maybe download a client,  follow
some games, and start playing.
There are also tutorials and books on learning to play online
(Table~\ref{tab:tutorial}), and in your library and book store.

\begin{table}
    \begin{tabular}{lll}
      {\bf Server} &{\bf Name}&{\bf Link}\\ \hline\hline
      OGS&Online Go Server& \url{https://online-go.com}\\
      KGS&Kiseido Go Server& \url{http://www.gokgs.com}\\
      IGS&Pandanet Internet Go
           Server&\url{http://www.pandanet-igs.com}\\
      Tygem&Tygem Baduk &\url{http://www.tygembaduk.com}\\
      Fox&Fox Weiqi (Chinese)&\url{https://www.foxwq.com}\\
      Dragon& Dragon Go Server for
              correspondence&\url{https://www.dragongoserver.net}\\
      \hline \\
    \end{tabular}
    \caption{Internet Go Servers}\label{tab:servers}
\end{table}

\begin{table}
    \begin{tabular}{lp{10cm}}
      {\bf Tutorial}&{\bf Link}\\ \hline \hline
      Kiseido&\url{https://www.kiseido.com/ff.htm}\\
      Nordic&\url{https://www.nordicgodojo.eu/post/212/a-simple-beginners-guide-to-go}\\
      Online Go& \url{https://online-go.com/learn-to-play-go}\\
      Wikipedia&\url{https://en.wikipedia.org/wiki/Go_(game)}\\ \hline \\
    \end{tabular}
    \caption{Go Tutorials}\label{tab:tutorial}
\end{table}

\subsection{Chess}
Chess is a different game, with more types of pieces than Go, and a more
tactical kind of play. Like Go, it is   a game
requiring a high level of intelligence to play well (and to program well). Chess has a long and
distinguished history of well-known players and champions.  Chances
are that your school has a Chess club. Many books
have been written, for beginners and  for
advanced players.

There are many servers on the Internet where you can play online Chess.
Table~\ref{tab:servers_ch} lists a few of the better known online Chess
servers. They also provide tutorial material on how to learn to play.

\begin{table}
    \begin{tabular}{ll}
      {\bf Server} &{\bf Link}\\ \hline\hline
      Chess.com& \url{https://chess.com}\\
      Internet Chess Club& \url{http://chessclub.com}\\
      Chessbase&\url{https://en.chessbase.com}\\
      Chess24&\url{https://chess24.com/en/play/chess}\\
      Lichess&\url{https://lichess.org}\\
      Wikipedia&\url{https://en.wikipedia.org/wiki/Chess}\\                 
      \hline \\
    \end{tabular}
    \caption{Internet Chess Resources}\label{tab:servers_ch}
\end{table}

\section{Learning to Program}
Software for Go and Chess is widely available.
On your computer or phone you can probably find a program or app for
playing Go or Chess.

There are also many resources avaliable if you want to learn how to program.

\subsection{Go}
For Go, there are three basic approaches: heuristic planning, MCTS, and
self-play. Software for hand-held devices is typically of the first
two categories, because of the computational demands of self-play. At
Senseis.net\footnote{\url{https://senseis.xmp.net/?ComputerGoProgramming}}
many articles can be found on computer Go for all three approaches. The Computer Go
Server\footnote{\url{http://www.yss-aya.com/cgos/19x19/standings.html}} allows Go programs to play against each other over the
internet using the Go Text
Protocol.\footnote{\url{http://www.lysator.liu.se/~gunnar/gtp/}}. A
mailing list for computer Go can be found
here.\footnote{\url{http://computer-go.org}}

Self-play environments
were listed in a previous appendix in Table~\ref{tab:self-env}. The environments are
open-source, so you can learn from the code and experiment with your
own improvements. A book devoted to  building your own
Alpha(Go) Zero clone is~\cite{pumperla2019}.
  
\subsection{Chess}
For Chess, there are also many open-source programs available. Most
programs follow the classic and highly successful heuristic planning
approach, see Table~\ref{tab:freechess}. A valuable resource with a wealth of information is the Chess
Programming
Wiki.\footnote{\url{https://www.chessprogramming.org/Main_Page}} Many
of the concepts and programming tricks that are used in the programs
are explained in the wiki.

\begin{table}
    \begin{tabular}{ll}
      {\bf Engine} &{\bf Link}\\ \hline\hline
      Stockfish& \url{https://stockfishchess.org}\\
      Komodo& \url{http://komodochess.com}\\
      Houdini&\url{http://www.cruxis.com/chess/houdini.htm}\\
      Critter&\url{http://www.vlasak.biz/critter/}\\
      Ethereal&\url{https://github.com/AndyGrant/Ethereal}\\
      Gull&\url{https://sourceforge.net/projects/gullchess/}\\
      Leela Chess Zero&\url{https://github.com/LeelaChessZero/lc0}\\                 
      \hline \\
    \end{tabular}
    \caption{Open-source Chess Programs}\label{tab:freechess}
\end{table}

\section{Scientific Journals and Conferences}
From an AI perspective the question whether a program can 
 play games for which humans need intelligence, is a highly relevant
question. Many of the papers on computer game programming have been
published in AI journals (Table~\ref{tab:jour} lists a few relevant
journals).
Games papers are often
      published at AI conferences (Table~\ref{tab:conf} lists a few relevant
      conferences).       Your  library and the internet
      should have the issues and proceedings.

      Most authors make their
      work available as preprint at the arXiv preprint server for early
      access before a paper is accepted for
      publication.\footnote{\url{https://arxiv.org/list/cs.AI/recent}} The latest news
      can be found at this server, unfiltered.
      
\begin{table}
    \begin{tabular}{ll}
      {\bf Abbreviation}&{\bf Journal}\\ \hline\hline
PAMI & IEEE Transactions on Pattern Analysis and Machine
      Intelligence\\
AIJ &    Artificial Intelligence Journal\\
JAIR & Journal of AI Research\\
      ToG & IEEE Transactions on Games\\
      ICGAJ& ICCA/ICGA Journal\\
      Science & Science\\
      Nature & Nature\\ \hline
    \end{tabular}
    \caption{Journals}\label{tab:jour}
\end{table}

\begin{table}
    \begin{tabular}{ll}
      {\bf Abbreviation}&{\bf Conference}\\ \hline\hline
      IJCAI&International Joint Conference on AI\\
      AAAI&Association for the Advancement of AI\\
      ECAI& European Conference on AI\\
      FDG&Foundations of Digital Games\\
      CoG& Conference on  Games\\
      NeurIPS&Neural Information Processing Systems\\
      ICML&International Conference on Machine Learning\\
      ICLR&International Conference on Learning Representations\\
      ACC/ACG&Advances in Computer Chess/Games\\
      CG&Conference on Computers and Games\\ \hline
    \end{tabular}
    \caption{Conferences}\label{tab:conf}
\end{table}

\chapter{Running Python}\label{chap:python}\index{Python}
Python is a modern programming language that is quite popular in
AI. Many machine learning packages have Python interfaces, such as TensorFlow and
Keras. This appendix shows how to start working with Python. Python
was developed by Guido van Rossum while he was a researcher at the
national research center for mathematics and computer science CWI in Amsterdam~\cite{lutz1996programming}. 

Before you install Python you should check if it is already available
on your computer. Start a terminal, or a command prompt, and, at the
prompt \verb|$ |, type \verb|python|. Now one of two things will have
happened. If you see an error message then Python is not installed
correctly on your
computer. If you see a response of the form
\begin{verbatim}
Python 3.6.0 (v3.6.0:41df79263a11, Dec 22 2016, 17:23:13) 
[GCC 4.2.1 (Apple Inc. build 5666) (dot 3)] on darwin
Type "help", "copyright", "credits" or "license" for more 
                                                 information.
>>> 
\end{verbatim}
then Python is working already on your computer. You can get out of
Python by typing \verb|Ctrl-D| or \verb|quit()|. Interactive help is
available by typing \verb|help()|. You can skip the rest of this
appendix, and start playing around with the examples and
exercises. How to install TensorFlow and Keras is discussed in the
chapters (Sect.~\ref{sec:keras}). 

To install Python, go to the \verb|python.org|
\href{https://www.python.org}{website}.\footnote{\url{https://www.python.org}}
Go to Downloads, click on it, and you should see the latest version
of Python being offered for your current operating system (Linux,
macOS, or Windows). Click the button to start the download, and when
the download is finished after a few minutes, launch the installer.
While you are waiting, browse around on the website to the
Documentation section. There are great reference works and tutorials
on learning Python. Please consult your  library or bookstore
 or the Internet. Try Google or Stackoverflow.

One of the  advantages of Python is the wealth of high-quality
software packages that are available, especially for data science and
artificial intelligence. Among them are popular packages such as
scikit-learn, numpy, matplotlib, TensorFlow, PyTorch, and Gym.  These packages
can be  installed easily by
typing \verb|pip install numpy| for installing the numpy numerical
package.

\subsubsection*{Python 2 and Python 3}
There are two version of Python that you may encounter, version 2 and
version 3. There are small but critical differences in the
language. For example, the \verb|print| statement in Python 2 does not
use parentheses, but in Python 3 it  follows the ordinary function
call convention and it does.

We use Python 3 in this book, and I suggest you do the same. So if you
start up \verb|python| on your machine you may want to check for the
version number that it prints. And perhaps install Python 3.

You may also want to consider using an integrated development
environment of graphical editor, interpreter, and debugger to ease
your development. PyCharm, Netbeans, Spyder, and Visual Studio are popular
choices, but feel free to look around. Most are free or have free
student editions.

Python is a highly popular programming language, because it is
powerful and elegant, but also because it is easy and fun to learn to
program in Python.
Please go ahead and enjoy programming in Python!

\backmatter

\bibliographystyle{plain}
               {\footnotesize
\bibliography{\string~/Dropbox/BibTex/plaat}}

\begin{thebibliography}{100}

\bibitem{tensorflow2015whitepaper}
Mart\'{\i}n Abadi, Ashish Agarwal, Paul Barham, Eugene Brevdo, Zhifeng Chen,
  Craig Citro, Greg~S. Corrado, Andy Davis, Jeffrey Dean, Matthieu Devin,
  Sanjay Ghemawat, Ian Goodfellow, Andrew Harp, Geoffrey Irving, Michael Isard,
  Yangqing Jia, Rafal Jozefowicz, Lukasz Kaiser, Manjunath Kudlur, Josh
  Levenberg, Dandelion Man\'{e}, Rajat Monga, Sherry Moore, Derek Murray, Chris
  Olah, Mike Schuster, Jonathon Shlens, Benoit Steiner, Ilya Sutskever, Kunal
  Talwar, Paul Tucker, Vincent Vanhoucke, Vijay Vasudevan, Fernanda Vi\'{e}gas,
  Oriol Vinyals, Pete Warden, Martin Wattenberg, Martin Wicke, Yuan Yu, and
  Xiaoqiang Zheng.
\newblock {TensorFlow}: Large-scale machine learning on heterogeneous systems,
  2015.
\newblock Software available from tensorflow.org.

\bibitem{abadi2016tensorflow}
Mart{\'\i}n Abadi, Paul Barham, Jianmin Chen, Zhifeng Chen, Andy Davis, Jeffrey
  Dean, Matthieu Devin, Sanjay Ghemawat, Geoffrey Irving, Michael Isard, et~al.
\newblock Tensorflow: A system for large-scale machine learning.
\newblock In {\em 12th {USENIX} Symposium on Operating Systems Design and
  Implementation ({OSDI} 16)}, pages 265--283, 2016.

\bibitem{abbeel2007application}
Pieter Abbeel, Adam Coates, Morgan Quigley, and Andrew~Y Ng.
\newblock An application of reinforcement learning to aerobatic helicopter
  flight.
\newblock In {\em Advances in Neural Information Processing Systems}, pages
  1--8, 2007.

\bibitem{abelson1996structure}
Harold Abelson, Gerald~Jay Sussman, and Julie Sussman.
\newblock {\em Structure and interpretation of computer programs}.
\newblock MIT Press, 1996.

\bibitem{abramson1990expected}
Bruce Abramson.
\newblock Expected-outcome: A general model of static evaluation.
\newblock {\em IEEE Transactions on Pattern Analysis and Machine Intelligence},
  12(2):182--193, 1990.

\bibitem{allis1994searching}
L~Victor Allis.
\newblock {\em Searching for solutions in games and artificial intelligence}.
\newblock PhD thesis, Maastricht University, 1994.

\bibitem{allis1994proof}
L~Victor Allis, Maarten van~der Meulen, and H~Jaap Van Den~Herik.
\newblock Proof-number search.
\newblock {\em Artificial Intelligence}, 66(1):91--124, 1994.

\bibitem{alpaydin2009introduction}
Ethem Alpaydin.
\newblock {\em Introduction to machine learning}.
\newblock MIT press, 2009.

\bibitem{althofer1988root}
Ingo Alth{\"o}fer.
\newblock Root evaluation errors: How they arise and propagate.
\newblock {\em ICGA Journal}, 11(2-3):55--63, 1988.

\bibitem{anantharaman1990singular}
Thomas Anantharaman, Murray~S Campbell, and Feng-hsiung Hsu.
\newblock Singular extensions: Adding selectivity to brute-force searching.
\newblock {\em Artificial Intelligence}, 43(1):99--109, 1990.

\bibitem{ando2003cooperation}
Takahisa Ando, Noriyuki Kobayashi, and Takao Uehara.
\newblock Cooperation and competition of agents in the auction of computer
  bridge.
\newblock {\em Electronics and Communications in Japan (Part III: Fundamental
  Electronic Science)}, 86(12):76--86, 2003.

\bibitem{reddit2017}
Anonymous.
\newblock Go {AI} strength vs. time.
\newblock {\em Reddit post}, 2017.

\bibitem{anthony2020learning}
Thomas Anthony, Tom Eccles, Andrea Tacchetti, J{\'a}nos Kram{\'a}r, Ian Gemp,
  Thomas~C Hudson, Nicolas Porcel, Marc Lanctot, Julien P{\'e}rolat, Richard
  Everett, et~al.
\newblock Learning to play no-press diplomacy with best response policy
  iteration.
\newblock {\em arXiv preprint arXiv:2006.04635}, 2020.

\bibitem{anthony2019policy}
Thomas Anthony, Robert Nishihara, Philipp Moritz, Tim Salimans, and John
  Schulman.
\newblock Policy gradient search: Online planning and expert iteration without
  search trees.
\newblock {\em arXiv preprint arXiv:1904.03646}, 2019.

\bibitem{anthony2017thinking}
Thomas Anthony, Zheng Tian, and David Barber.
\newblock Thinking fast and slow with deep learning and tree search.
\newblock In {\em Advances in Neural Information Processing Systems}, pages
  5360--5370, 2017.

\bibitem{antoniou2018train}
Antreas Antoniou, Harrison Edwards, and Amos Storkey.
\newblock How to train your {MAML}.
\newblock {\em arXiv preprint arXiv:1810.09502}, 2018.

\bibitem{antoniou2004semantic}
Grigoris Antoniou and Frank Van~Harmelen.
\newblock {\em A semantic web primer}.
\newblock 2004.

\bibitem{arenz2012monte}
Oleg Arenz.
\newblock {Monte Carlo Chess}.
\newblock Master's thesis, Universit\"at Darmstadt, 2012.

\bibitem{argyriou2007multi}
Andreas Argyriou, Theodoros Evgeniou, and Massimiliano Pontil.
\newblock Multi-task feature learning.
\newblock In {\em Advances in Neural Information Processing Systems}, pages
  41--48, 2007.

\bibitem{arneson2010monte}
Broderick Arneson, Ryan~B Hayward, and Philip Henderson.
\newblock {Monte Carlo Tree Search} in {Hex}.
\newblock {\em IEEE Transactions on Computational Intelligence and AI in
  Games}, 2(4):251--258, 2010.

\bibitem{asmuth2009bayesian}
John Asmuth, Lihong Li, Michael~L Littman, Ali Nouri, and David Wingate.
\newblock A {Bayesian} sampling approach to exploration in reinforcement
  learning.
\newblock In {\em Proceedings of the Twenty-Fifth Conference on Uncertainty in
  Artificial Intelligence}, pages 19--26. AUAI Press, 2009.

\bibitem{auer2002using}
Peter Auer.
\newblock Using confidence bounds for exploitation-exploration trade-offs.
\newblock {\em Journal of Machine Learning Research}, 3(Nov):397--422, 2002.

\bibitem{auer2002finite}
Peter Auer, Nicolo Cesa-Bianchi, and Paul Fischer.
\newblock Finite-time analysis of the multiarmed bandit problem.
\newblock {\em {Machine Learning}}, 47(2-3):235--256, 2002.

\bibitem{auer2010ucb}
Peter Auer and Ronald Ortner.
\newblock {UCB} revisited: Improved regret bounds for the stochastic
  multi-armed bandit problem.
\newblock {\em Periodica Mathematica Hungarica}, 61(1-2):55--65, 2010.

\bibitem{baarslag2013evaluating}
Tim Baarslag, Katsuhide Fujita, Enrico~H Gerding, Koen Hindriks, Takayuki Ito,
  Nicholas~R Jennings, Catholijn Jonker, Sarit Kraus, Raz Lin, Valentin Robu,
  et~al.
\newblock Evaluating practical negotiating agents: Results and analysis of the
  2011 international competition.
\newblock {\em Artificial Intelligence}, 198:73--103, 2013.

\bibitem{back1996evolutionary}
Thomas B{\"a}ck.
\newblock {\em Evolutionary algorithms in theory and practice: evolutionary
  strategies, evolutionary programming, genetic algorithms}.
\newblock Oxford University Press, 1996.

\bibitem{back1993overview}
Thomas B{\"a}ck and Hans-Paul Schwefel.
\newblock An overview of evolutionary algorithms for parameter optimization.
\newblock {\em Evolutionary Computation}, 1(1):1--23, 1993.

\bibitem{badia2020agent57}
Adri{\`a}~Puigdom{\`e}nech Badia, Bilal Piot, Steven Kapturowski, Pablo
  Sprechmann, Alex Vitvitskyi, Daniel Guo, and Charles Blundell.
\newblock Agent57: Outperforming the {Atari} human benchmark.
\newblock {\em arXiv preprint arXiv:2003.13350}, 2020.

\bibitem{baier2010power}
Hendrik Baier and Peter~D Drake.
\newblock The power of forgetting: Improving the last-good-reply policy in
  {Monte} {Carlo} {Go}.
\newblock {\em IEEE Transactions on Computational Intelligence and AI in
  Games}, 2(4):303--309, 2010.

\bibitem{baker2016designing}
Bowen Baker, Otkrist Gupta, Nikhil Naik, and Ramesh Raskar.
\newblock Designing neural network architectures using reinforcement learning.
\newblock {\em arXiv preprint arXiv:1611.02167}, 2016.

\bibitem{bakker2004hierarchical}
Bram Bakker and J{\"u}rgen Schmidhuber.
\newblock Hierarchical reinforcement learning based on subgoal discovery and
  subpolicy specialization.
\newblock In {\em Proc. of the 8-th Conf. on Intelligent Autonomous Systems},
  pages 438--445, 2004.

\bibitem{baldi2014searching}
Pierre Baldi, Peter Sadowski, and Daniel Whiteson.
\newblock Searching for exotic particles in high-energy physics with deep
  learning.
\newblock {\em Nature Communications}, 5:4308, 2014.

\bibitem{ballard1983minimax}
Bruce~W Ballard.
\newblock The *-minimax search procedure for trees containing chance nodes.
\newblock {\em Artificial Intelligence}, 21(3):327--350, 1983.

\bibitem{bansal2017emergent}
Trapit Bansal, Jakub Pachocki, Szymon Sidor, Ilya Sutskever, and Igor Mordatch.
\newblock Emergent complexity via multi-agent competition.
\newblock {\em arXiv preprint arXiv:1710.03748}, 2017.

\bibitem{bard2013annual}
Nolan Bard, John Hawkin, Jonathan Rubin, and Martin Zinkevich.
\newblock The annual computer poker competition.
\newblock {\em AI Magazine}, 34(2):112, 2013.

\bibitem{barto2003recent}
Andrew~G Barto and Sridhar Mahadevan.
\newblock Recent advances in hierarchical reinforcement learning.
\newblock {\em Discrete Event Dynamic Systems}, 13(1-2):41--77, 2003.

\bibitem{bartz2010experimental}
Thomas Bartz-Beielstein, Marco Chiarandini, Lu{\'\i}s Paquete, and Mike Preuss.
\newblock {\em Experimental methods for the analysis of optimization
  algorithms}.
\newblock Springer, 2010.

\bibitem{openaibaseline2017}
OpenAI Baselines.
\newblock \url{https://openai.com/blog/openai-baselines-dqn/}, 2017.

\bibitem{baudivs2011balancing}
Petr Baudi{\v{s}}.
\newblock Balancing {MCTS} by dynamically adjusting the komi value.
\newblock {\em ICGA Journal}, 34(3):131--139, 2011.

\bibitem{baudivs2011pachi}
Petr Baudi{\v{s}} and Jean-loup Gailly.
\newblock Pachi: State of the art open source {Go} program.
\newblock In {\em Advances in Computer Games}, pages 24--38. Springer, 2011.

\bibitem{baum2017survey}
Seth Baum.
\newblock A survey of artificial general intelligence projects for ethics,
  risk, and policy.
\newblock 2017.

\bibitem{baxter1999knightcap}
Jonathan Baxter, Andrew Tridgell, and Lex Weaver.
\newblock Knightcap: a chess program that learns by combining {TD} $(\lambda)$
  with game-tree search.
\newblock {\em arXiv preprint cs/9901002}, 1999.

\bibitem{baxter2000learning}
Jonathan Baxter, Andrew Tridgell, and Lex Weaver.
\newblock Learning to play chess using temporal differences.
\newblock {\em Machine Learning}, 40(3):243--263, 2000.

\bibitem{beal1983recent}
Don Beal.
\newblock Recent progress in understanding minimax search.
\newblock In {\em Proceedings of the 1983 Annual Conference on Computers:
  Extending the Human Resource}, pages 164--169. ACM, 1983.

\bibitem{beal1989experiments}
Don Beal.
\newblock Experiments with the null move.
\newblock {\em Advances in Computer Chess}, 5:65--79, 1989.

\bibitem{beal1990generalised}
Don Beal.
\newblock A generalised quiescence search algorithm.
\newblock {\em Artificial Intelligence}, 43(1):85--98, 1990.

\bibitem{beal1999learning}
Don Beal and Martin~C Smith.
\newblock Learning piece-square values using temporal differences.
\newblock {\em ICGA Journal}, 22(4):223--235, 1999.

\bibitem{beal2000temporal}
Don Beal and Martin~C. Smith.
\newblock Temporal difference learning for heuristic search and game playing.
\newblock {\em Information Sciences}, 122(1):3--21, 2000.

\bibitem{bear2007neuroscience}
Mark~F Bear, Barry~W Connors, and Michael~A Paradiso.
\newblock {\em Neuroscience}, volume~2.
\newblock Lippincott Williams \& Wilkins, 2007.

\bibitem{behl2019alpha}
Harkirat~Singh Behl, At{\i}l{\i}m~G{\"u}ne{\c{s}} Baydin, and Philip~HS Torr.
\newblock Alpha {MAML}: {Adaptive Model-Agnostic Meta-Learning}.
\newblock {\em arXiv preprint arXiv:1905.07435}, 2019.

\bibitem{bellemare2013bayesian}
Marc Bellemare, Joel Veness, and Michael Bowling.
\newblock Bayesian learning of recursively factored environments.
\newblock In {\em International Conference on Machine Learning}, pages
  1211--1219, 2013.

\bibitem{bellemare2017distributional}
Marc~G Bellemare, Will Dabney, and R{\'e}mi Munos.
\newblock A distributional perspective on reinforcement learning.
\newblock {\em arXiv preprint arXiv:1707.06887}, 2017.

\bibitem{bellemare2013arcade}
Marc~G Bellemare, Yavar Naddaf, Joel Veness, and Michael Bowling.
\newblock The {Arcade Learning Environment}: An evaluation platform for general
  agents.
\newblock {\em Journal of Artificial Intelligence Research}, 47:253--279, 2013.

\bibitem{bellman1957dynamic}
Richard Bellman.
\newblock {\em Dynamic programming}.
\newblock Courier Corporation, 1957, 2013.

\bibitem{bellman1965application}
Richard Bellman.
\newblock On the application of dynamic programing to the determination of
  optimal play in chess and checkers.
\newblock {\em Proceedings of the National Academy of Sciences},
  53(2):244--247, 1965.

\bibitem{bello2016neural}
Irwan Bello, Hieu Pham, Quoc~V Le, Mohammad Norouzi, and Samy Bengio.
\newblock Neural combinatorial optimization with reinforcement learning.
\newblock {\em arXiv preprint arXiv:1611.09940}, 2016.

\bibitem{bengio2009learning}
Yoshua Bengio.
\newblock {\em Learning Deep Architectures for AI}.
\newblock Now Publishers Inc, 2009.

\bibitem{bengio2012practical}
Yoshua Bengio.
\newblock Practical recommendations for gradient-based training of deep
  architectures.
\newblock In {\em Neural networks: Tricks of the trade}, pages 437--478.
  Springer, 2012.

\bibitem{bengio1990learning}
Yoshua Bengio, Samy Bengio, and Jocelyn Cloutier.
\newblock Learning a synaptic learning rule.
\newblock Technical report, Montreal, 1990.

\bibitem{bengio2013representation}
Yoshua Bengio, Aaron Courville, and Pascal Vincent.
\newblock Representation learning: A review and new perspectives.
\newblock {\em IEEE Transactions on Pattern Analysis and Machine Intelligence},
  35(8):1798--1828, 2013.

\bibitem{bengio2007greedy}
Yoshua Bengio, Pascal Lamblin, Dan Popovici, and Hugo Larochelle.
\newblock Greedy layer-wise training of deep networks.
\newblock In {\em Advances in Neural Information Processing Systems}, pages
  153--160, 2007.

\bibitem{bengio2009curriculum}
Yoshua Bengio, J{\'e}r{\^o}me Louradour, Ronan Collobert, and Jason Weston.
\newblock Curriculum learning.
\newblock In {\em {Proceedings of the 26th Annual International Conference on
  Machine Learning}}, pages 41--48, 2009.

\bibitem{benson1976life}
David~B Benson.
\newblock Life in the game of {Go}.
\newblock {\em Information Sciences}, 10(1):17--29, 1976.

\bibitem{bergstra2011theano}
James Bergstra, Fr{\'e}d{\'e}ric Bastien, Olivier Breuleux, Pascal Lamblin,
  Razvan Pascanu, Olivier Delalleau, Guillaume Desjardins, David Warde-Farley,
  Ian Goodfellow, Arnaud Bergeron, et~al.
\newblock Theano: Deep learning on {GPUs} with {Python}.
\newblock In {\em NIPS 2011, BigLearning Workshop, Granada, Spain}, volume~3,
  pages 1--48. Citeseer, 2011.

\bibitem{bergstra2012random}
James Bergstra and Yoshua Bengio.
\newblock Random search for hyper-parameter optimization.
\newblock {\em Journal of Machine Learning Research}, 13(Feb):281--305, 2012.

\bibitem{bergstra2011algorithms}
James~S Bergstra, R{\'e}mi Bardenet, Yoshua Bengio, and Bal{\'a}zs K{\'e}gl.
\newblock Algorithms for hyper-parameter optimization.
\newblock In {\em Advances in Neural Information Processing Systems}, pages
  2546--2554, 2011.

\bibitem{berlekamp1994mathematical}
Elwyn Berlekamp and David Wolfe.
\newblock {\em Mathematical {Go}: Chilling gets the last point}.
\newblock AK Peters/CRC Press, 1994.

\bibitem{berliner1977experiences}
Hans~J Berliner.
\newblock Experiences in evaluation with {BKG}---a program that plays
  backgammon.
\newblock In {\em IJCAI}, pages 428--433, 1977.

\bibitem{berliner1980backgammon}
Hans~J Berliner.
\newblock Backgammon computer program beats world champion.
\newblock {\em Artificial Intelligence}, 14(2):205--220, 1980.

\bibitem{berner2019dota}
Christopher Berner, Greg Brockman, Brooke Chan, Vicki Cheung, Przemys{\l}aw
  Debiak, Christy Dennison, David Farhi, Quirin Fischer, Shariq Hashme, Chris
  Hesse, et~al.
\newblock Dota 2 with large scale deep reinforcement learning.
\newblock {\em arXiv preprint arXiv:1912.06680}, 2019.

\bibitem{berners2001semantic}
Tim Berners-Lee, James Hendler, Ora Lassila, et~al.
\newblock The semantic web.
\newblock {\em Scientific American}, 284(5):28--37, 2001.

\bibitem{bertolami2009novel}
R~Bertolami, H~Bunke, S~Fernandez, A~Graves, M~Liwicki, and J~Schmidhuber.
\newblock A novel connectionist system for improved unconstrained handwriting
  recognition.
\newblock {\em IEEE Transactions on Pattern Analysis and Machine Intelligence},
  31(5), 2009.

\bibitem{bertsekas2013rollout}
Dimitri~P Bertsekas.
\newblock Rollout algorithms for discrete optimization: A survey.
\newblock In {\em Handbook of combinatorial optimization}, pages 2989--3013.
  Springer, 2013.

\bibitem{bertsekas1996neuro}
Dimitri~P Bertsekas and John Tsitsiklis.
\newblock {\em Neuro-dynamic programming}.
\newblock MIT Press Cambridge, 1996.

\bibitem{bhatnagar2009convergent}
Shalabh Bhatnagar, Doina Precup, David Silver, Richard~S Sutton, Hamid~R Maei,
  and Csaba Szepesv{\'a}ri.
\newblock Convergent temporal-difference learning with arbitrary smooth
  function approximation.
\newblock In {\em Advances in Neural Information Processing Systems}, pages
  1204--1212, 2009.

\bibitem{bibel2013automated}
Wolfgang Bibel.
\newblock {\em Automated theorem proving}.
\newblock Springer Science \& Business Media, 1987.

\bibitem{billings2003approximating}
Darse Billings, Neil Burch, Aaron Davidson, Robert Holte, Jonathan Schaeffer,
  Terence Schauenberg, and Duane Szafron.
\newblock Approximating game-theoretic optimal strategies for full-scale poker.
\newblock In {\em IJCAI}, volume~3, page 661, 2003.

\bibitem{billings2002challenge}
Darse Billings, Aaron Davidson, Jonathan Schaeffer, and Duane Szafron.
\newblock The challenge of poker.
\newblock {\em Artificial Intelligence}, 134(1-2):201--240, 2002.

\bibitem{billings2004game}
Darse Billings, Aaron Davidson, Terence Schauenberg, Neil Burch, Michael
  Bowling, Robert Holte, Jonathan Schaeffer, and Duane Szafron.
\newblock Game-tree search with adaptation in stochastic imperfect-information
  games.
\newblock In {\em International Conference on Computers and Games}, pages
  21--34. Springer, 2004.

\bibitem{billings1998opponent}
Darse Billings, Denis Papp, Jonathan Schaeffer, and Duane Szafron.
\newblock Opponent modeling in poker.
\newblock {\em {AAAI/IAAI}}, 493:499, 1998.

\bibitem{bishop2006pattern}
Christopher~M Bishop.
\newblock {\em Pattern recognition and machine learning}.
\newblock Information science and statistics. Springer Verlag, Heidelberg,
  2006.

\bibitem{bishop1995neural}
Christopher~M Bishop et~al.
\newblock {\em Neural networks for pattern recognition}.
\newblock Oxford University Press, 1995.

\bibitem{bjornsson2009cadiaplayer}
Yngvi Bjornsson and Hilmar Finnsson.
\newblock Cadiaplayer: A simulation-based general game player.
\newblock {\em IEEE Transactions on Computational Intelligence and AI in
  Games}, 1(1):4--15, 2009.

\bibitem{bjornsson2001multi}
Yngvi Bj{\"o}rnsson and Tony~A Marsland.
\newblock Multi-cut $\alpha$$\beta$-pruning in game-tree search.
\newblock {\em Theoretical Computer Science}, 252(1-2):177--196, 2001.

\bibitem{bonabeau1999swarm}
Eric Bonabeau, Marco Dorigo, Guy Theraulaz, et~al.
\newblock {\em Swarm intelligence: from natural to artificial systems}.
\newblock Number~1. Oxford University Press, 1999.

\bibitem{bonnet2013complexity}
{\'E}douard Bonnet, Florian Jamain, and Abdallah Saffidine.
\newblock On the complexity of trick-taking card games.
\newblock In {\em IJCAI}, pages 482--488, 2013.

\bibitem{boon1991overzicht}
Mark Boon.
\newblock Overzicht van de ontwikkeling van een {G}o spelend programma.
\newblock 1991.

\bibitem{boser1992training}
Bernhard~E Boser, Isabelle~M Guyon, and Vladimir~N Vapnik.
\newblock A training algorithm for optimal margin classifiers.
\newblock In {\em Proceedings of the Fifth Annual Workshop on Computational
  learning theory}, pages 144--152. ACM, 1992.

\bibitem{bosse2008formalisation}
Tibor Bosse, Catholijn~M Jonker, and Jan Treur.
\newblock Formalisation of {Damasio's} theory of emotion, feeling and core
  consciousness.
\newblock {\em Consciousness and Cognition}, 17(1):94--113, 2008.

\bibitem{bouzy2004monte}
Bruno Bouzy and Bernard Helmstetter.
\newblock Monte {Carlo} {Go} developments.
\newblock In {\em Advances in Computer Games}, pages 159--174. Springer, 2004.

\bibitem{bouzy2011mcts}
Bruno Bouzy, Marc M{\'e}tivier, and Damien Pellier.
\newblock {MCTS} experiments on the {Voronoi} game.
\newblock In {\em Advances in Computer Games}, pages 96--107. Springer, 2011.

\bibitem{bowling2015heads}
Michael Bowling, Neil Burch, Michael Johanson, and Oskari Tammelin.
\newblock Heads-up limit hold'em poker is solved.
\newblock {\em Science}, 347(6218):145--149, 2015.

\bibitem{bowling2009demonstration}
Michael Bowling, Nicholas~Abou Risk, Nolan Bard, Darse Billings, Neil Burch,
  Joshua Davidson, John Hawkin, Robert Holte, Michael Johanson, Morgan Kan,
  et~al.
\newblock A demonstration of the polaris poker system.
\newblock In {\em Proceedings of The 8th International Conference on Autonomous
  Agents and Multiagent Systems-Volume 2}, pages 1391--1392. International
  Foundation for Autonomous Agents and Multiagent Systems, 2009.

\bibitem{bratko2001prolog}
Ivan Bratko.
\newblock {\em Prolog programming for artificial intelligence}.
\newblock Pearson Education, 2001.

\bibitem{brazdil2008metalearning}
Pavel Brazdil, Christophe~Giraud Carrier, Carlos Soares, and Ricardo Vilalta.
\newblock {\em Metalearning: Applications to data mining}.
\newblock Springer Science \& Business Media, 2008.

\bibitem{brochu2010tutorial}
Eric Brochu, Vlad~M Cora, and Nando De~Freitas.
\newblock A tutorial on {Bayesian} optimization of expensive cost functions,
  with application to active user modeling and hierarchical reinforcement
  learning.
\newblock {\em arXiv preprint arXiv:1012.2599}, 2010.

\bibitem{brockington1997keyano}
Mark~G Brockington.
\newblock Keyano unplugged---the construction of an {Othello} program.
\newblock Technical report, Technical Report TR 97-05, Department of Computing
  Science, University of Alberta, 1997.

\bibitem{brockman2016openai}
Greg Brockman, Vicki Cheung, Ludwig Pettersson, Jonas Schneider, John Schulman,
  Jie Tang, and Wojciech Zaremba.
\newblock {OpenAI} {G}ym.
\newblock {\em arXiv preprint arXiv:1606.01540}, 2016.

\bibitem{broekens2007emotion}
Joost Broekens.
\newblock Emotion and reinforcement: affective facial expressions facilitate
  robot learning.
\newblock In {\em Artifical Intelligence for Human Computing}, pages 113--132.
  Springer, 2007.

\bibitem{broekens2009assistive}
Joost Broekens, Marcel Heerink, Henk Rosendal, et~al.
\newblock Assistive social robots in elderly care: a review.
\newblock {\em Gerontechnology}, 8(2):94--103, 2009.

\bibitem{brooks1991intelligence}
Rodney~A Brooks.
\newblock Intelligence without representation.
\newblock {\em Artificial Intelligence}, 47(1-3):139--159, 1991.

\bibitem{brown2018deep}
Noam Brown, Adam Lerer, Sam Gross, and Tuomas Sandholm.
\newblock Deep counterfactual regret minimization.
\newblock {\em arXiv preprint arXiv:1811.00164}, 2018.

\bibitem{brown2018superhuman}
Noam Brown and Tuomas Sandholm.
\newblock Superhuman {AI} for heads-up no-limit poker: Libratus beats top
  professionals.
\newblock {\em Science}, 359(6374):418--424, 2018.

\bibitem{brown2019superhuman}
Noam Brown and Tuomas Sandholm.
\newblock Superhuman {AI} for multiplayer poker.
\newblock {\em Science}, 365(6456):885--890, 2019.

\bibitem{brown2018depth}
Noam Brown, Tuomas Sandholm, and Brandon Amos.
\newblock Depth-limited solving for imperfect-information games.
\newblock {\em arXiv preprint arXiv:1805.08195}, 2018.

\bibitem{browne2000hex}
Cameron Browne.
\newblock Hex strategy.
\newblock {\em AK Peters, Wellesley MA}, 2000.

\bibitem{browne2019strategic}
Cameron Browne, Dennis~JNJ Soemers, and Eric Piette.
\newblock Strategic features for general games.
\newblock In {\em KEG@ AAAI}, pages 70--75, 2019.

\bibitem{browne2012survey}
Cameron~B Browne, Edward Powley, Daniel Whitehouse, Simon~M Lucas, Peter~I
  Cowling, Philipp Rohlfshagen, Stephen Tavener, Diego Perez, Spyridon
  Samothrakis, and Simon Colton.
\newblock A survey of {Monte} {Carlo} {Tree} {Search} methods.
\newblock {\em IEEE Transactions on Computational Intelligence and AI in
  Games}, 4(1):1--43, 2012.

\bibitem{brugmann1993monte}
Bernd Br{\"u}gmann.
\newblock Monte {Carlo} {Go}.
\newblock Technical report, Syracuse University, 1993.

\bibitem{buchberger1982computer}
Bruno Buchberger, George~E Collins, R{\"u}diger Loos, and Rudolph Albrecht.
\newblock Computer algebra symbolic and algebraic computation.
\newblock {\em ACM SIGSAM Bulletin}, 16(4):5--5, 1982.

\bibitem{bucilua2006model}
Cristian Bucilu{\v a}, Rich Caruana, and Alexandru Niculescu-Mizil.
\newblock Model compression.
\newblock In {\em Proceedings of the 12th ACM SIGKDD International Conference
  on Knowledge Discovery and Data Mining}, pages 535--541, 2006.

\bibitem{burke2013hyper}
Edmund~K Burke, Michel Gendreau, Matthew Hyde, Graham Kendall, Gabriela Ochoa,
  Ender {\"O}zcan, and Rong Qu.
\newblock Hyper-heuristics: A survey of the state of the art.
\newblock {\em Journal of the Operational Research Society}, 64(12):1695--1724,
  2013.

\bibitem{buro1995logistello}
Michael Buro.
\newblock Logistello: A strong learning {Othello} program.
\newblock In {\em 19th Annual Conference Gesellschaft f{\"u}r Klassifikation
  eV}, volume~2. Citeseer, 1995.

\bibitem{buro1995statistical}
Michael Buro.
\newblock Statistical feature combination for the evaluation of game positions.
\newblock {\em Journal of Artificial Intelligence Research}, 3:373--382, 1995.

\bibitem{buro1997experiments}
Michael Buro.
\newblock Experiments with {Multi}-{ProbCut} and a new high-quality evaluation
  function for {Othello}.
\newblock {\em Games in AI Research}, pages 77--96, 1997.

\bibitem{buro1997othello}
Michael Buro.
\newblock The {Othello} match of the year: {Takeshi} {Murakami} vs.
  {Logistello}.
\newblock {\em ICGA Journal}, 20(3):189--193, 1997.

\bibitem{buro2003evolution}
Michael Buro.
\newblock The evolution of strong {Othello} programs.
\newblock In {\em Entertainment Computing}, pages 81--88. Springer, 2003.

\bibitem{busoniu2008comprehensive}
Lucian Busoniu, Robert Babuska, and Bart De~Schutter.
\newblock A comprehensive survey of multiagent reinforcement learning.
\newblock {\em IEEE Transactions on Systems, Man, And Cybernetics-Part C:
  Applications and Reviews, 38 (2), 2008}, 2008.

\bibitem{campbell2002deep}
Murray Campbell, A~Joseph Hoane~Jr, and Feng-hsiung Hsu.
\newblock Deep {Blue}.
\newblock {\em Artificial Intelligence}, 134(1-2):57--83, 2002.

\bibitem{caruana1997multitask}
Rich Caruana.
\newblock Multitask learning.
\newblock {\em Machine Learning}, 28(1):41--75, 1997.

\bibitem{caruana2001overfitting}
Rich Caruana, Steve Lawrence, and C~Lee Giles.
\newblock Overfitting in neural nets: Backpropagation, conjugate gradient, and
  early stopping.
\newblock In {\em Advances in Neural Information Processing Systems}, pages
  402--408, 2001.

\bibitem{castelfranchi1999deliberative}
Cristiano Castelfranchi, Frank Dignum, Catholijn~M Jonker, and Jan Treur.
\newblock Deliberative normative agents: Principles and architecture.
\newblock In {\em International Workshop on Agent Theories, Architectures, and
  Languages}, pages 364--378. Springer, 1999.

\bibitem{castro2018dopamine}
Pablo~Samuel Castro, Subhodeep Moitra, Carles Gelada, Saurabh Kumar, and Marc~G
  Bellemare.
\newblock Dopamine: A research framework for deep reinforcement learning.
\newblock {\em arXiv preprint arXiv:1812.06110}, 2018.

\bibitem{cazenave2020polygames}
Tristan Cazenave, Yen-Chi Chen, Guan-Wei Chen, Shi-Yu Chen, Xian-Dong Chiu,
  Julien Dehos, Maria Elsa, Qucheng Gong, Hengyuan Hu, Vasil Khalidov, et~al.
\newblock Polygames: Improved zero learning.
\newblock {\em arXiv preprint arXiv:2001.09832}, 2020.

\bibitem{cazenave2005combining}
Tristan Cazenave and Bernard Helmstetter.
\newblock Combining tactical search and {Monte-Carlo} in the game of {Go}.
\newblock {\em CIG}, 5:171--175, 2005.

\bibitem{chaitin2006limits}
Gregory Chaitin.
\newblock The limits of reason.
\newblock {\em Scientific American}, 294(3):74--81, 2006.

\bibitem{chang2005adaptive}
Hyeong~Soo Chang, Michael~C Fu, Jiaqiao Hu, and Steven~I Marcus.
\newblock An adaptive sampling algorithm for solving {Markov} decision
  processes.
\newblock {\em Operations Research}, 53(1):126--139, 2005.

\bibitem{charlesworth2018application}
Henry Charlesworth.
\newblock Application of self-play reinforcement learning to a four-player game
  of imperfect information.
\newblock {\em arXiv preprint arXiv:1808.10442}, 2018.

\bibitem{chaslot2010monte}
Guillaume Chaslot.
\newblock {\em {Monte-Carlo} tree search}.
\newblock PhD thesis, Maastricht University, 2010.

\bibitem{chaslot2008monte}
Guillaume Chaslot, Sander Bakkes, Istvan Szita, and Pieter Spronck.
\newblock {Monte-Carlo} tree search: A new framework for game {AI}.
\newblock In {\em AIIDE}, 2008.

\bibitem{chaslot2008parallel}
Guillaume Chaslot, Mark~HM Winands, and H~Jaap van~den Herik.
\newblock Parallel {Monte-Carlo} tree search.
\newblock In {\em International Conference on Computers and Games}, pages
  60--71. Springer, 2008.

\bibitem{chaslot2008progressive}
Guillaume~MJB Chaslot, Mark~HM Winands, H~Jaap van~den Herik, Jos~WHM
  Uiterwijk, and Bruno Bouzy.
\newblock Progressive strategies for {Monte-Carlo} tree search.
\newblock {\em New Mathematics and Natural Computation}, 4(03):343--357, 2008.

\bibitem{chellapilla1999evolving}
Kumar Chellapilla and David~B Fogel.
\newblock Evolving neural networks to play checkers without relying on expert
  knowledge.
\newblock {\em IEEE Transactions on Neural Networks}, 10(6):1382--1391, 1999.

\bibitem{chen1999static}
Ken Chen and Zhixing Chen.
\newblock Static analysis of life and death in the game of {Go}.
\newblock {\em Information Sciences}, 121(1-2):113--134, 1999.

\bibitem{chen2015microsoft}
Xinlei Chen, Hao Fang, Tsung-Yi Lin, Ramakrishna Vedantam, Saurabh Gupta, Piotr
  Doll{\'a}r, and C~Lawrence Zitnick.
\newblock Microsoft {COCO} captions: Data collection and evaluation server.
\newblock {\em arXiv preprint arXiv:1504.00325}, 2015.

\bibitem{cheng2017survey}
Yu~Cheng, Duo Wang, Pan Zhou, and Tao Zhang.
\newblock A survey of model compression and acceleration for deep neural
  networks.
\newblock {\em arXiv preprint arXiv:1710.09282}, 2017.

\bibitem{chi1988comparison}
Ping-Chung Chi and Dana~S Nau.
\newblock Comparison of the minimax and product back-up rules in a variety of
  games.
\newblock In {\em Search in artificial intelligence}, pages 450--471. Springer,
  1988.

\bibitem{choi1996pb}
Jaeyoung Choi, Jack~J Dongarra, and David~W Walker.
\newblock {PB-BLAS}: a set of parallel block basic linear algebra subprograms.
\newblock {\em Concurrency: Practice and Experience}, 8(7):517--535, 1996.

\bibitem{chollet2017deep}
Francois Chollet.
\newblock {\em Deep learning with Python}.
\newblock Manning Publications Co., 2017.

\bibitem{chollet2015keras}
Fran{\c{c}}ois Chollet et~al.
\newblock Keras.
\newblock \url{https://keras.io}, 2015.

\bibitem{chrabaszcz2018back}
Patryk Chrabaszcz, Ilya Loshchilov, and Frank Hutter.
\newblock Back to basics: Benchmarking canonical evolution strategies for
  playing {Atari}.
\newblock {\em arXiv preprint arXiv:1802.08842}, 2018.

\bibitem{cirecsan2012multi}
Dan Cire{\c{s}}an, Ueli Meier, and J{\"u}rgen Schmidhuber.
\newblock Multi-column deep neural networks for image classification.
\newblock {\em arXiv preprint arXiv:1202.2745}, 2012.

\bibitem{cirecsan2010deep}
Dan~Claudiu Cire{\c{s}}an, Ueli Meier, Luca~Maria Gambardella, and J{\"u}rgen
  Schmidhuber.
\newblock Deep, big, simple neural nets for handwritten digit recognition.
\newblock {\em Neural Computation}, 22(12):3207--3220, 2010.

\bibitem{clark2014teaching}
Christopher Clark and Amos Storkey.
\newblock Teaching deep convolutional neural networks to play {Go}. arxiv
  preprint.
\newblock {\em arXiv preprint arXiv:1412.3409}, 1, 2014.

\bibitem{clark2015training}
Christopher Clark and Amos Storkey.
\newblock Training deep convolutional neural networks to play {Go}.
\newblock In {\em International Conference on Machine Learning}, pages
  1766--1774, 2015.

\bibitem{claus1998dynamics}
Caroline Claus and Craig Boutilier.
\newblock The dynamics of reinforcement learning in cooperative multiagent
  systems.
\newblock {\em AAAI/IAAI}, 1998:746--752, 1998.

\bibitem{clocksin2012programming}
William~F Clocksin and Christopher~S Mellish.
\newblock {\em Programming in Prolog: Using the ISO standard}.
\newblock Springer Science \& Business Media, 1981.

\bibitem{colton2012painting}
Simon Colton.
\newblock The painting fool: Stories from building an automated painter.
\newblock In {\em Computers and creativity}, pages 3--38. Springer, 2012.

\bibitem{coplan1982special}
Kevin Coplan.
\newblock A special-purpose machine for an improved search algorithm for deep
  chess combinations.
\newblock In {\em Advances in Computer Chess}, pages 25--43. Elsevier, 1982.

\bibitem{goratings}
R\'emi Coulom.
\newblock Go ratings. \url{https://www.goratings.org/en/history/}.

\bibitem{coulom2006efficient}
R{\'e}mi Coulom.
\newblock Efficient selectivity and backup operators in {Monte-Carlo} {Tree
  Search}.
\newblock In {\em International Conference on Computers and Games}, pages
  72--83. Springer, 2006.

\bibitem{coulom2007monte}
R{\'e}mi Coulom.
\newblock {Monte-Carlo} tree search in {Crazy Stone}.
\newblock In {\em Proc. Game Prog. Workshop, Tokyo, Japan}, pages 74--75, 2007.

\bibitem{coulom2009monte}
R{\'e}mi Coulom.
\newblock The {Monte-Carlo} revolution in {Go}.
\newblock In {\em The Japanese-French Frontiers of Science Symposium (JFFoS
  2008), Roscoff, France}, 2009.

\bibitem{culberson1998pattern}
Joseph~C Culberson and Jonathan Schaeffer.
\newblock Pattern databases.
\newblock {\em Computational Intelligence}, 14(3):318--334, 1998.

\bibitem{currie1991plan}
Ken Currie and Austin Tate.
\newblock O-plan: the open planning architecture.
\newblock {\em Artificial Intelligence}, 52(1):49--86, 1991.

\bibitem{cybenko1989approximations}
George Cybenko.
\newblock Approximations by superpositions of a sigmoidal function.
\newblock {\em Mathematics of Control, Signals and Systems}, 2:183--192, 1989.

\bibitem{czarnogorski2018monte}
Kamil Czarnog{\'o}rski.
\newblock {Monte Carlo Tree Search} beginners guide
  \url{https://int8.io/monte-carlo-tree-search-beginners-guide/}, 2018.

\bibitem{dabney2020distributional}
Will Dabney, Zeb Kurth-Nelson, Naoshige Uchida, Clara~Kwon Starkweather, Demis
  Hassabis, R{\'e}mi Munos, and Matthew Botvinick.
\newblock A distributional code for value in dopamine-based reinforcement
  learning.
\newblock {\em Nature}, pages 1--5, 2020.

\bibitem{das2016incorporating}
Shubhomoy Das, Weng-Keen Wong, Thomas Dietterich, Alan Fern, and Andrew Emmott.
\newblock Incorporating expert feedback into active anomaly discovery.
\newblock In {\em 2016 IEEE 16th International Conference on Data Mining
  (ICDM)}, pages 853--858. IEEE, 2016.

\bibitem{david2016deepchess}
Omid~E David, Nathan~S Netanyahu, and Lior Wolf.
\newblock Deepchess: End-to-end deep neural network for automatic learning in
  chess.
\newblock In {\em International Conference on Artificial Neural Networks},
  pages 88--96. Springer, 2016.

\bibitem{david2014genetic}
Omid~E David, H~Jaap van~den Herik, Moshe Koppel, and Nathan~S Netanyahu.
\newblock Genetic algorithms for evolving computer chess programs.
\newblock {\em IEEE Transactions on Evolutionary Computation}, 18(5):779--789,
  2014.

\bibitem{david2008genetic}
Omid David-Tabibi, Moshe Koppel, and Nathan~S Netanyahu.
\newblock Genetic algorithms for mentor-assisted evaluation function
  optimization.
\newblock In {\em Proceedings of the 10th Annual Conference on Genetic and
  Evolutionary Computation}, pages 1469--1476. ACM, 2008.

\bibitem{davis2012game}
Morton~D Davis.
\newblock {\em Game theory: a nontechnical introduction}.
\newblock Courier Corporation, 2012.

\bibitem{de1994solution}
Arie De~Bruin, Wim Pijls, and Aske Plaat.
\newblock Solution trees as a basis for game-tree search.
\newblock {\em ICCA Journal}, 17(4):207--219, 1994.

\bibitem{de2002ant}
Luis~M De~Campos, Juan~M Fernandez-Luna, Jos{\'e}~A G{\'a}mez, and Jos{\'e}~M
  Puerta.
\newblock Ant colony optimization for learning {Bayesian} networks.
\newblock {\em International Journal of Approximate Reasoning}, 31(3):291--311,
  2002.

\bibitem{de2018clinically}
Jeffrey De~Fauw, Joseph~R Ledsam, Bernardino Romera-Paredes, Stanislav Nikolov,
  Nenad Tomasev, Sam Blackwell, Harry Askham, Xavier Glorot, Brendan
  O'Donoghue, Daniel Visentin, et~al.
\newblock Clinically applicable deep learning for diagnosis and referral in
  retinal disease.
\newblock {\em Nature Medicine}, 24(9):1342, 2018.

\bibitem{de2015negotiations}
Dave De~Jonge.
\newblock {\em Negotiations over large agreement spaces}.
\newblock PhD thesis, Universitat Aut{\`o}noma de Barcelona, 2015.

\bibitem{de2018challenge}
Dave De~Jonge, Tim Baarslag, Reyhan Aydo{\u{g}}an, Catholijn Jonker, Katsuhide
  Fujita, and Takayuki Ito.
\newblock The challenge of negotiation in the game of diplomacy.
\newblock In {\em International Conference on Agreement Technologies}, pages
  100--114. Springer, 2018.

\bibitem{de2017d}
Dave De~Jonge and Carles Sierra.
\newblock D-brane: a {Diplomacy} playing agent for automated negotiations
  research.
\newblock {\em Applied Intelligence}, 47(1):158--177, 2017.

\bibitem{de2017automated}
Dave De~Jonge and Dongmo Zhang.
\newblock Automated negotiations for general game playing.
\newblock In {\em Proceedings of the 16th Conference on Autonomous Agents and
  MultiAgent Systems}, pages 371--379. International Foundation for Autonomous
  Agents and Multiagent Systems, 2017.

\bibitem{dechter1986learning}
Rina Dechter.
\newblock {\em Learning while searching in constraint-satisfaction problems}.
\newblock University of California, Computer Science Department, 1986.

\bibitem{deng2009imagenet}
Jia Deng, Wei Dong, Richard Socher, Li-Jia Li, Kai Li, and Li~Fei-Fei.
\newblock Imagenet: A large-scale hierarchical image database.
\newblock In {\em 2009 IEEE Conference on Computer Vision and Pattern
  Recognition}, pages 248--255. Ieee, 2009.

\bibitem{baselines}
Prafulla Dhariwal, Christopher Hesse, Oleg Klimov, Alex Nichol, Matthias
  Plappert, Alec Radford, John Schulman, Szymon Sidor, Yuhuai Wu, and Peter
  Zhokhov.
\newblock {OpenAI} baselines.
\newblock \url{https://github.com/openai/baselines}, 2017.

\bibitem{dick1996androids}
Philip~K Dick.
\newblock Do androids dream of electric sheep? 1968.
\newblock {\em New York: Del Rey}, 1996.

\bibitem{dickmanns1987genetische}
Dirk Dickmanns, J{\"u}rgen Schmidhuber, and Andreas Winklhofer.
\newblock {Der genetische Algorithmus: Eine Implementierung in Prolog}.
\newblock {\em Fortgeschrittenenpraktikum, Institut f{\"u}r Informatik,
  Lehrstuhl Prof. Radig, Technische Universit at M{\"u}nchen}, 1987.

\bibitem{dietterich2000hierarchical}
Thomas~G Dietterich.
\newblock Hierarchical reinforcement learning with the {MAXQ} value function
  decomposition.
\newblock {\em Journal of Artificial Intelligence Research}, 13:227--303, 2000.

\bibitem{doan2019line}
Thang Doan, Joao Monteiro, Isabela Albuquerque, Bogdan Mazoure, Audrey Durand,
  Joelle Pineau, and R~Devon Hjelm.
\newblock On-line adaptative curriculum learning for {GANs}.
\newblock In {\em Proceedings of the AAAI Conference on Artificial
  Intelligence}, volume~33, pages 3470--3477, 2019.

\bibitem{dodge2017study}
Samuel Dodge and Lina Karam.
\newblock A study and comparison of human and deep learning recognition
  performance under visual distortions.
\newblock In {\em Computer Communication and Networks (ICCCN), 2017 26th
  International Conference on}, pages 1--7. IEEE, 2017.

\bibitem{donahue2015long}
Jeffrey Donahue, Lisa Anne~Hendricks, Sergio Guadarrama, Marcus Rohrbach,
  Subhashini Venugopalan, Kate Saenko, and Trevor Darrell.
\newblock Long-term recurrent convolutional networks for visual recognition and
  description.
\newblock In {\em Proceedings of the IEEE Conference on Computer Vision and
  Pattern Recognition}, pages 2625--2634, 2015.

\bibitem{dongarra1988extended}
Jack~J Dongarra, Jeremy Du~Croz, Sven Hammarling, and Richard~J Hanson.
\newblock An extended set of {FORTRAN} basic linear algebra subprograms.
\newblock {\em ACM Transactions on Mathematical Software (TOMS)}, 14(1):1--17,
  1988.

\bibitem{donninger1993null}
Christian Donninger.
\newblock Null move and deep search.
\newblock {\em ICGA Journal}, 16(3):137--143, 1993.

\bibitem{doran2017does}
Derek Doran, Sarah Schulz, and Tarek~R Besold.
\newblock What does explainable {AI} really mean? a new conceptualization of
  perspectives.
\newblock {\em arXiv preprint arXiv:1710.00794}, 2017.

\bibitem{dorigo1997ant}
Marco Dorigo and Luca~Maria Gambardella.
\newblock Ant colony system: a cooperative learning approach to the traveling
  salesman problem.
\newblock {\em IEEE Transactions on Evolutionary Computation}, 1(1):53--66,
  1997.

\bibitem{doya2000reinforcement}
Kenji Doya.
\newblock Reinforcement learning in continuous time and space.
\newblock {\em Neural Computation}, 12(1):219--245, 2000.

\bibitem{drachen2014skill}
Anders Drachen, Matthew Yancey, John Maguire, Derrek Chu, Iris~Yuhui Wang,
  Tobias Mahlmann, Matthias Schubert, and Diego Klabajan.
\newblock Skill-based differences in spatio-temporal team behaviour in {Defence
  of the Ancients 2 (DOTA 2)}.
\newblock In {\em 2014 IEEE Games Media Entertainment}, pages 1--8. IEEE, 2014.

\bibitem{droste2008learning}
Sacha Droste and Johannes F{\"u}rnkranz.
\newblock Learning of piece values for chess variants.
\newblock Technical report, Tech. Rep. TUD--KE--2008-07, Knowledge Engineering
  Group, TU Darmstadt, 2008.

\bibitem{duan2016benchmarking}
Yan Duan, Xi~Chen, Rein Houthooft, John Schulman, and Pieter Abbeel.
\newblock Benchmarking deep reinforcement learning for continuous control.
\newblock In {\em International Conference on Machine Learning}, pages
  1329--1338, 2016.

\bibitem{duan2016rl}
Yan Duan, John Schulman, Xi~Chen, Peter~L Bartlett, Ilya Sutskever, and Pieter
  Abbeel.
\newblock {RL}${}^{2}$: Fast reinforcement learning via slow reinforcement
  learning.
\newblock {\em arXiv preprint arXiv:1611.02779}, 2016.

\bibitem{duchi2011adaptive}
John Duchi, Elad Hazan, and Yoram Singer.
\newblock Adaptive subgradient methods for online learning and stochastic
  optimization.
\newblock {\em Journal of Machine Learning Research}, 12(Jul):2121--2159, 2011.

\bibitem{duke2004policy}
Richard~D Duke and Jac Geurts.
\newblock {\em Policy games for strategic management}.
\newblock Rozenberg Publishers, 2004.

\bibitem{dulac2015deep}
Gabriel Dulac-Arnold, Richard Evans, Hado van Hasselt, Peter Sunehag, Timothy
  Lillicrap, Jonathan Hunt, Timothy Mann, Theophane Weber, Thomas Degris, and
  Ben Coppin.
\newblock Deep reinforcement learning in large discrete action spaces.
\newblock {\em arXiv preprint arXiv:1512.07679}, 2015.

\bibitem{ecoffet2019go}
Adrien Ecoffet, Joost Huizinga, Joel Lehman, Kenneth~O Stanley, and Jeff Clune.
\newblock Go-explore: a new approach for hard-exploration problems.
\newblock {\em arXiv preprint arXiv:1901.10995}, 2019.

\bibitem{edwards1961alpha}
Daniel~James Edwards and TP~Hart.
\newblock The alpha-beta heuristic.
\newblock Technical Report AI Memo 030, MIT, 1961.

\bibitem{elman1993learning}
Jeffrey~L Elman.
\newblock Learning and development in neural networks: The importance of
  starting small.
\newblock {\em Cognition}, 48(1):71--99, 1993.

\bibitem{elnaggar2014comparative}
Ahmed~A Elnaggar, Mahmoud Gadallah, Mostafa~Abdel Aziem, and Hesham El-Deeb.
\newblock A comparative study of game tree searching methods.
\newblock {\em International Journal of Advanced Computer Science and
  Applications}, 5(5):68--77, 2014.

\bibitem{elo1978rating}
Arpad~E Elo.
\newblock {\em The rating of chessplayers, past and present}.
\newblock Arco Pub., 1978.

\bibitem{elsken2018neural}
Thomas Elsken, Jan~Hendrik Metzen, and Frank Hutter.
\newblock Neural architecture search: A survey.
\newblock {\em arXiv preprint arXiv:1808.05377}, 2018.

\bibitem{enderton1991golem}
Herbert~D Enderton.
\newblock The {Golem} {Go} program.
\newblock Technical report, Carnegie Mellon University, 1991.

\bibitem{enzenberger1996integration}
Markus Enzenberger.
\newblock The integration of a priori knowledge into a {Go} playing neural
  network.
\newblock 1996.

\bibitem{enzenberger2009lock}
Markus Enzenberger and Martin M{\"u}ller.
\newblock A lock-free multithreaded {Monte-Carlo} tree search algorithm.
\newblock In {\em Advances in Computer Games}, pages 14--20. Springer, 2009.

\bibitem{enzenberger2010fuego}
Markus Enzenberger, Martin Muller, Broderick Arneson, and Richard Segal.
\newblock Fuego---an open-source framework for board games and {Go} engine
  based on {Monte Carlo} tree search.
\newblock {\em IEEE Transactions on Computational Intelligence and AI in
  Games}, 2(4):259--270, 2010.

\bibitem{erhan2010does}
Dumitru Erhan, Yoshua Bengio, Aaron Courville, Pierre-Antoine Manzagol, Pascal
  Vincent, and Samy Bengio.
\newblock Why does unsupervised pre-training help deep learning?
\newblock {\em Journal of Machine Learning Research}, 11(Feb):625--660, 2010.

\bibitem{ernst2005tree}
Damien Ernst, Pierre Geurts, and Louis Wehenkel.
\newblock Tree-based batch mode reinforcement learning.
\newblock {\em Journal of Machine Learning Research}, 6(Apr):503--556, 2005.

\bibitem{evgeniou2004regularized}
Theodoros Evgeniou and Massimiliano Pontil.
\newblock Regularized multi-task learning.
\newblock In {\em Proceedings of the tenth ACM SIGKDD International Conference
  on Knowledge Discovery and Data Mining}, pages 109--117. ACM, 2004.

\bibitem{fang2000construction}
Haw-ren Fang, Tsan-sheng Hsu, and Shun-chin Hsu.
\newblock Construction of {Chinese} chess endgame databases by retrograde
  analysis.
\newblock In {\em International Conference on Computers and Games}, pages
  96--114. Springer, 2000.

\bibitem{farquhar2018treeqn}
Gregory Farquhar, Tim Rockt{\"a}schel, Maximilian Igl, and SA~Whiteson.
\newblock {TreeQN and ATreeC}: Differentiable tree planning for deep
  reinforcement learning.
\newblock International Conference on Learning Representations, 2018.

\bibitem{fei2009imagenet}
Li~Fei-Fei, Jia Deng, and Kai Li.
\newblock Imagenet: Constructing a large-scale image database.
\newblock {\em Journal of Vision}, 9(8):1037--1037, 2009.

\bibitem{fern2011ensemble}
Alan Fern and Paul Lewis.
\newblock Ensemble {Monte-Carlo} planning: An empirical study.
\newblock In {\em ICAPS}, 2011.

\bibitem{fernandez2007application}
Santiago Fern{\'a}ndez, Alex Graves, and J{\"u}rgen Schmidhuber.
\newblock An application of recurrent neural networks to discriminative keyword
  spotting.
\newblock In {\em International Conference on Artificial Neural Networks},
  pages 220--229. Springer, 2007.

\bibitem{feurer2015efficient}
Matthias Feurer, Aaron Klein, Katharina Eggensperger, Jost Springenberg, Manuel
  Blum, and Frank Hutter.
\newblock Efficient and robust automated machine learning.
\newblock In {\em Advances in Neural Information Processing Systems}, pages
  2962--2970, 2015.

\bibitem{fiebrink2011real}
Rebecca~Anne Fiebrink.
\newblock {\em Real-time human interaction with supervised learning algorithms
  for music composition and performance}.
\newblock Citeseer, 2011.

\bibitem{fikes1972learning}
Richard~E Fikes, Peter~E Hart, and Nils~J Nilsson.
\newblock Learning and executing generalized robot plans.
\newblock {\em Artificial Intelligence}, 3:251--288, 1972.

\bibitem{fikes1971strips}
Richard~E Fikes and Nils~J Nilsson.
\newblock {STRIPS: A new approach to the application of theorem proving to
  problem solving}.
\newblock {\em Artificial Intelligence}, 2(3-4):189--208, 1971.

\bibitem{finn2017model}
Chelsea Finn, Pieter Abbeel, and Sergey Levine.
\newblock {Model-Agnostic Meta-Learning} for fast adaptation of deep networks.
\newblock {\em arXiv preprint arXiv:1703.03400}, 2017.

\bibitem{finn2018probabilistic}
Chelsea Finn, Kelvin Xu, and Sergey Levine.
\newblock Probabilistic {Model-Agnostic Meta-Learning}.
\newblock In {\em Advances in Neural Information Processing Systems}, pages
  9516--9527, 2018.

\bibitem{finnsson2010learning}
Hilmar Finnsson and Yngvi Bj{\"o}rnsson.
\newblock Learning simulation control in general game-playing agents.
\newblock In {\em AAAI}, volume~10, pages 954--959, 2010.

\bibitem{fishburn1982analysis}
John~P Fishburn.
\newblock Analysis of speedup in distributed algorithms.
\newblock 1982.

\bibitem{flet2019promise}
Yannis Flet-Berliac.
\newblock The promise of hierarchical reinforcement learning.
\newblock
  \url{https://thegradient.pub/the-promise-of-hierarchical-reinforcement-learning/},
  March 2019.

\bibitem{foerster2017stabilising}
Jakob Foerster, Nantas Nardelli, Gregory Farquhar, Triantafyllos Afouras,
  Philip~HS Torr, Pushmeet Kohli, and Shimon Whiteson.
\newblock Stabilising experience replay for deep multi-agent reinforcement
  learning.
\newblock {\em arXiv preprint arXiv:1702.08887}, 2017.

\bibitem{fogel1994introduction}
David~B Fogel.
\newblock An introduction to simulated evolutionary optimization.
\newblock {\em IEEE Transactions on Neural Networks}, 5(1):3--14, 1994.

\bibitem{fogel2001blondie24}
David~B Fogel.
\newblock {\em Blondie24: Playing at the Edge of AI}.
\newblock Elsevier, 2001.

\bibitem{fogel2005further}
David~B Fogel, Timothy~J Hays, Sarah~L Hahn, and James Quon.
\newblock Further evolution of a self-learning chess program.
\newblock In {\em CIG}, 2005.

\bibitem{fogel2006blondie25}
David~B Fogel, Timothy~J Hays, Sarah~L Hahn, and James Quon.
\newblock The {Blondie25} chess program competes against {Fritz} 8.0 and a
  human chess master.
\newblock In {\em IEEE Symposium on Computational Intelligence and Games},
  pages 230--235. Citeseer, 2006.

\bibitem{fortunato2017noisy}
Meire Fortunato, Mohammad~Gheshlaghi Azar, Bilal Piot, Jacob Menick, Ian
  Osband, Alex Graves, Vlad Mnih, Remi Munos, Demis Hassabis, Olivier Pietquin,
  et~al.
\newblock Noisy networks for exploration.
\newblock {\em arXiv preprint arXiv:1706.10295}, 2017.

\bibitem{fraenkel1981computing}
Aviezri~S Fraenkel and David Lichtenstein.
\newblock Computing a perfect strategy for $n \times n$ chess requires time
  exponential in $n$.
\newblock In {\em International Colloquium on Automata, Languages, and
  Programming}, pages 278--293. Springer, 1981.

\bibitem{frans2017meta}
Kevin Frans, Jonathan Ho, Xi~Chen, Pieter Abbeel, and John Schulman.
\newblock Meta learning shared hierarchies.
\newblock {\em arXiv preprint arXiv:1710.09767}, 2017.

\bibitem{friedman2001elements}
Jerome Friedman, Trevor Hastie, and Robert Tibshirani.
\newblock {\em The elements of statistical learning}, volume~1.
\newblock Springer series in statistics New York, NY, USA:, 2001.

\bibitem{frome2013devise}
Andrea Frome, Greg~S Corrado, Jon Shlens, Samy Bengio, Jeff Dean, Marc'Aurelio
  Ranzato, and Tomas Mikolov.
\newblock Devise: A deep visual-semantic embedding model.
\newblock In {\em Advances in Neural Information Processing Systems}, pages
  2121--2129, 2013.

\bibitem{frosst2017distilling}
Nicholas Frosst and Geoffrey Hinton.
\newblock Distilling a neural network into a soft decision tree.
\newblock {\em arXiv preprint arXiv:1711.09784}, 2017.

\bibitem{furnkranz1996machine}
Johannes F{\"u}rnkranz.
\newblock Machine learning in computer chess: The next generation.
\newblock {\em ICGA Journal}, 19(3):147--161, 1996.

\bibitem{gambardella1995ant}
Luca~M Gambardella and Marco Dorigo.
\newblock {Ant-Q: A} reinforcement learning approach to the traveling salesman
  problem.
\newblock In {\em Machine Learning Proceedings 1995}, pages 252--260. Elsevier,
  1995.

\bibitem{ganzfried2011game}
Sam Ganzfried and Tuomas Sandholm.
\newblock Game theory-based opponent modeling in large imperfect-information
  games.
\newblock In {\em The 10th International Conference on Autonomous Agents and
  Multiagent Systems-Volume 2}, pages 533--540. International Foundation for
  Autonomous Agents and Multiagent Systems, 2011.

\bibitem{garcez2015neural}
Artur~d'Avila Garcez, Tarek~R Besold, Luc De~Raedt, Peter F{\"o}ldiak, Pascal
  Hitzler, Thomas Icard, Kai-Uwe K{\"u}hnberger, Luis~C Lamb, Risto
  Miikkulainen, and Daniel~L Silver.
\newblock Neural-symbolic learning and reasoning: contributions and challenges.
\newblock In {\em 2015 AAAI Spring Symposium Series}, 2015.

\bibitem{gasser1996solving}
Ralph Gasser.
\newblock Solving nine men's morris.
\newblock {\em Computational Intelligence}, 12(1):24--41, 1996.

\bibitem{gelly2012grand}
Sylvain Gelly, Levente Kocsis, Marc Schoenauer, Michele Sebag, David Silver,
  Csaba Szepesv{\'a}ri, and Olivier Teytaud.
\newblock The grand challenge of computer {Go}: {Monte Carlo} tree search and
  extensions.
\newblock {\em Communications of the ACM}, 55(3):106--113, 2012.

\bibitem{gelly2007combining}
Sylvain Gelly and David Silver.
\newblock Combining online and offline knowledge in {UCT}.
\newblock In {\em Proceedings of the 24th International Conference on Machine
  learning}, pages 273--280. ACM, 2007.

\bibitem{gelly2008achieving}
Sylvain Gelly and David Silver.
\newblock Achieving master level play in $9 \times 9$ computer {Go}.
\newblock In {\em AAAI}, volume~8, pages 1537--1540, 2008.

\bibitem{gelly2011monte}
Sylvain Gelly and David Silver.
\newblock {Monte-Carlo} tree search and rapid action value estimation in
  computer go.
\newblock {\em Artificial Intelligence}, 175(11):1856--1875, 2011.

\bibitem{gelly2006modification}
Sylvain Gelly, Yizao Wang, and Olivier Teytaud.
\newblock Modification of {UCT} with patterns in {Monte-Carlo} {Go}.
\newblock Technical Report RR-6062, INRIA, 2006.

\bibitem{genesereth2005general}
Michael Genesereth, Nathaniel Love, and Barney Pell.
\newblock General game playing: Overview of the {AAAI} competition.
\newblock {\em AI magazine}, 26(2):62, 2005.

\bibitem{geron2017hands}
Aur{\'e}lien G{\'e}ron.
\newblock {\em Hands-on machine learning with Scikit-Learn and TensorFlow:
  concepts, tools, and techniques to build intelligent systems}.
\newblock {O'Reilly Media, Inc.}, 2019.

\bibitem{gers1999learning}
Felix~A Gers, J{\"u}rgen Schmidhuber, and Fred Cummins.
\newblock Learning to forget: Continual prediction with {LSTM}.
\newblock In {\em 1999 Ninth International Conference on Artificial Neural
  Networks ICANN 99. (Conf. Publ. No. 470)}. IET, 1999.

\bibitem{gilpin2006competitive}
Andrew Gilpin and Tuomas Sandholm.
\newblock A competitive {Texas Hold'em} poker player via automated abstraction
  and real-time equilibrium computation.
\newblock In {\em Proceedings of the National Conference on Artificial
  Intelligence}, volume~21, page 1007. Menlo Park, CA; Cambridge, MA; London;
  AAAI Press; MIT Press; 1999, 2006.

\bibitem{giotis2001low}
Alexios Giotis, Michael Emmerich, Boris Naujoks, Kyriakos Giannakoglou, and
  Thomas B{\"a}ck.
\newblock Low-cost stochastic optimization for engineering applications.

\bibitem{gladwell2008outliers}
Malcolm Gladwell.
\newblock {\em Outliers: The story of success}.
\newblock Hachette UK, 2008.

\bibitem{gleave2019adversarial}
Adam Gleave, Michael Dennis, Neel Kant, Cody Wild, Sergey Levine, and Stuart
  Russell.
\newblock Adversarial policies: Attacking deep reinforcement learning.
\newblock {\em arXiv preprint arXiv:1905.10615}, 2019.

\bibitem{glynn1989importance}
Peter~W Glynn and Donald~L Iglehart.
\newblock Importance sampling for stochastic simulations.
\newblock {\em Management Science}, 35(11):1367--1392, 1989.

\bibitem{goetsch1990experiments}
Gordon Goetsch and Murray~S Campbell.
\newblock Experiments with the null-move heuristic.
\newblock In {\em Computers, Chess, and Cognition}, pages 159--168. Springer,
  1990.

\bibitem{good1968five}
Irving~John Good.
\newblock A five-year plan for automatic chess.
\newblock {\em Machine Intelligence 2}, pages 89--118, 1968.

\bibitem{goodfellow2016deep}
Ian Goodfellow, Yoshua Bengio, and Aaron Courville.
\newblock {\em Deep learning}.
\newblock MIT Press, Cambridge, 2016.

\bibitem{goodfellow2014generative}
Ian Goodfellow, Jean Pouget-Abadie, Mehdi Mirza, Bing Xu, David Warde-Farley,
  Sherjil Ozair, Aaron Courville, and Yoshua Bengio.
\newblock Generative adversarial nets.
\newblock In {\em Advances in Neural Information Processing Systems}, pages
  2672--2680, 2014.

\bibitem{gordon1995stable}
Geoffrey~J Gordon.
\newblock Stable function approximation in dynamic programming.
\newblock In {\em Machine Learning Proceedings 1995}, pages 261--268. Elsevier,
  1995.

\bibitem{graf2015adaptive}
Tobias Graf and Marco Platzner.
\newblock Adaptive playouts in {Monte-Carlo} tree search with policy-gradient
  reinforcement learning.
\newblock In {\em Advances in Computer Games}, pages 1--11. Springer, 2015.

\bibitem{graves2012supervised}
Alex Graves.
\newblock Supervised sequence labelling.
\newblock In {\em Supervised sequence labelling with recurrent neural
  networks}, pages 5--13. Springer, 2012.

\bibitem{graves2017automated}
Alex Graves, Marc~G Bellemare, Jacob Menick, Remi Munos, and Koray Kavukcuoglu.
\newblock Automated curriculum learning for neural networks.
\newblock {\em arXiv preprint arXiv:1704.03003}, 2017.

\bibitem{graves2005bidirectional}
Alex Graves, Santiago Fern{\'a}ndez, and J{\"u}rgen Schmidhuber.
\newblock Bidirectional {LSTM} networks for improved phoneme classification and
  recognition.
\newblock In {\em International Conference on Artificial Neural Networks},
  pages 799--804. Springer, 2005.

\bibitem{graves2014towards}
Alex Graves and Navdeep Jaitly.
\newblock Towards end-to-end speech recognition with recurrent neural networks.
\newblock In {\em International Conference on Machine Learning}, pages
  1764--1772, 2014.

\bibitem{graves2014neural}
Alex Graves, Greg Wayne, and Ivo Danihelka.
\newblock Neural {Turing} machines.
\newblock {\em arXiv preprint arXiv:1410.5401}, 2014.

\bibitem{graziano2011human}
Michael~SA Graziano and Sabine Kastner.
\newblock Human consciousness and its relationship to social neuroscience: a
  novel hypothesis.
\newblock {\em Cognitive Neuroscience}, 2(2):98--113, 2011.

\bibitem{greenblatt1967greenblatt}
Richard~D Greenblatt, Donald~E Eastlake~III, and Stephen~D Crocker.
\newblock {The Greenblatt chess program}.
\newblock In {\em Proceedings of the November 14-16, 1967, Fall Joint Computer
  Conference}, pages 801--810, 1967.

\bibitem{greff2017lstm}
Klaus Greff, Rupesh~K Srivastava, Jan Koutn{\'\i}k, Bas~R Steunebrink, and
  J{\"u}rgen Schmidhuber.
\newblock {LSTM: A} search space odyssey.
\newblock {\em IEEE Transactions on Neural Networks and Learning Systems},
  28(10):2222--2232, 2017.

\bibitem{grondman2012survey}
Ivo Grondman, Lucian Busoniu, Gabriel~AD Lopes, and Robert Babuska.
\newblock A survey of actor-critic reinforcement learning: Standard and natural
  policy gradients.
\newblock {\em IEEE Transactions on Systems, Man, and Cybernetics, Part C
  (Applications and Reviews)}, 42(6):1291--1307, 2012.

\bibitem{gruslys2017reactor}
Audrunas Gruslys, Will Dabney, Mohammad~Gheshlaghi Azar, Bilal Piot, Marc
  Bellemare, and Remi Munos.
\newblock The reactor: A fast and sample-efficient actor-critic agent for
  reinforcement learning.
\newblock {\em arXiv preprint arXiv:1704.04651}, 2017.

\bibitem{guez2019investigation}
Arthur Guez, Mehdi Mirza, Karol Gregor, Rishabh Kabra, S{\'e}bastien
  Racani{\`e}re, Th{\'e}ophane Weber, David Raposo, Adam Santoro, Laurent
  Orseau, Tom Eccles, et~al.
\newblock An investigation of model-free planning.
\newblock {\em arXiv preprint arXiv:1901.03559}, 2019.

\bibitem{guez2018learning}
Arthur Guez, Th{\'e}ophane Weber, Ioannis Antonoglou, Karen Simonyan, Oriol
  Vinyals, Daan Wierstra, R{\'e}mi Munos, and David Silver.
\newblock Learning to search with {MCTSnets}.
\newblock {\em arXiv preprint arXiv:1802.04697}, 2018.

\bibitem{gunning2017explainable}
David Gunning.
\newblock Explainable artificial intelligence {(XAI)}.
\newblock {\em Defense Advanced Research Projects Agency (DARPA), nd Web}, 2,
  2017.

\bibitem{guo2014deep}
Xiaoxiao Guo, Satinder Singh, Honglak Lee, Richard~L Lewis, and Xiaoshi Wang.
\newblock Deep learning for real-time {Atari} game play using offline
  {Monte-Carlo} tree search planning.
\newblock In {\em Advances in Neural Information Processing Systems}, pages
  3338--3346, 2014.

\bibitem{guo2016deep}
Yanming Guo, Yu~Liu, Ard Oerlemans, Songyang Lao, Song Wu, and Michael~S Lew.
\newblock Deep learning for visual understanding: A review.
\newblock {\em Neurocomputing}, 187:27--48, 2016.

\bibitem{gupta2018unsupervised}
Abhishek Gupta, Benjamin Eysenbach, Chelsea Finn, and Sergey Levine.
\newblock Unsupervised meta-learning for reinforcement learning.
\newblock {\em arXiv preprint arXiv:1806.04640}, 2018.

\bibitem{gupta2018meta}
Abhishek Gupta, Russell Mendonca, YuXuan Liu, Pieter Abbeel, and Sergey Levine.
\newblock Meta-reinforcement learning of structured exploration strategies.
\newblock {\em arXiv preprint arXiv:1802.07245}, 2018.

\bibitem{hafner2018learning}
Danijar Hafner, Timothy Lillicrap, Ian Fischer, Ruben Villegas, David Ha,
  Honglak Lee, and James Davidson.
\newblock Learning latent dynamics for planning from pixels.
\newblock {\em arXiv preprint arXiv:1811.04551}, 2018.

\bibitem{hall2009weka}
Mark Hall, Eibe Frank, Geoffrey Holmes, Bernhard Pfahringer, Peter Reutemann,
  and Ian~H Witten.
\newblock The {WEKA} data mining software: an update.
\newblock {\em ACM SIGKDD Explorations Newsletter}, 11(1):10--18, 2009.

\bibitem{hansen2007anytime}
Eric~A Hansen and Rong Zhou.
\newblock Anytime heuristic search.
\newblock {\em Journal of Artificial Intelligence Research}, 28:267--297, 2007.

\bibitem{harman2007current}
Mark Harman.
\newblock The current state and future of search based software engineering.
\newblock In {\em 2007 Future of Software Engineering}, pages 342--357. IEEE
  Computer Society, 2007.

\bibitem{hasselt2010double}
Hado~V Hasselt.
\newblock {Double Q-learning}.
\newblock In {\em Advances in Neural Information Processing Systems}, pages
  2613--2621, 2010.

\bibitem{haastad1991power}
Johan H{\aa}stad and Mikael Goldmann.
\newblock On the power of small-depth threshold circuits.
\newblock {\em Computational Complexity}, 1(2):113--129, 1991.

\bibitem{hauk2004rediscovering}
Thomas Hauk, Michael Buro, and Jonathan Schaeffer.
\newblock Rediscovering *-minimax search.
\newblock In {\em International Conference on Computers and Games}, pages
  35--50. Springer, 2004.

\bibitem{haykin1994neural}
Simon Haykin.
\newblock {\em Neural networks: a comprehensive foundation}.
\newblock Prentice Hall PTR, 1994.

\bibitem{haykin2009neural}
Simon~S Haykin et~al.
\newblock {\em Neural networks and learning machines/Simon Haykin.}
\newblock New York: Prentice Hall,, 2009.

\bibitem{hayward2019hex}
Ryan~B Hayward and Bjarne Toft.
\newblock {\em Hex: The Full Story}.
\newblock CRC Press, 2019.

\bibitem{he2016deep}
Kaiming He, Xiangyu Zhang, Shaoqing Ren, and Jian Sun.
\newblock Deep residual learning for image recognition.
\newblock In {\em Proceedings of the IEEE Conference on Computer Vision and
  Pattern Recognition}, pages 770--778, 2016.

\bibitem{heckerman1995learning}
David Heckerman, Dan Geiger, and David~M Chickering.
\newblock Learning {Bayesian} networks: The combination of knowledge and
  statistical data.
\newblock {\em Machine Learning}, 20(3):197--243, 1995.

\bibitem{heess2012actor}
Nicolas Heess, David Silver, and Yee~Whye Teh.
\newblock Actor-critic reinforcement learning with energy-based policies.
\newblock In {\em EWRL}, pages 43--58, 2012.

\bibitem{heess2017emergence}
Nicolas Heess, Dhruva TB, Srinivasan Sriram, Jay Lemmon, Josh Merel, Greg
  Wayne, Yuval Tassa, Tom Erez, Ziyu Wang, SM~Eslami, et~al.
\newblock Emergence of locomotion behaviours in rich environments.
\newblock {\em arXiv preprint arXiv:1707.02286}, 2017.

\bibitem{heinrich2016deep}
Johannes Heinrich and David Silver.
\newblock Deep reinforcement learning from self-play in imperfect-information
  games.
\newblock {\em arXiv preprint arXiv:1603.01121}, 2016.

\bibitem{heinz1999adaptive}
Ernst~A Heinz.
\newblock Adaptive null-move pruning.
\newblock {\em ICGA Journal}, 22(3):123--132, 1999.

\bibitem{heinz2000new}
Ernst~A Heinz.
\newblock New self-play results in computer chess.
\newblock In {\em International Conference on Computers and Games}, pages
  262--276. Springer, 2000.

\bibitem{helmbold2009all}
David~P Helmbold and Aleatha Parker-Wood.
\newblock {All-Moves-As-First} heuristics in {Monte-Carlo} {Go}.
\newblock In {\em IC-AI}, pages 605--610, 2009.

\bibitem{henderson2018deep}
Peter Henderson, Riashat Islam, Philip Bachman, Joelle Pineau, Doina Precup,
  and David Meger.
\newblock Deep reinforcement learning that matters.
\newblock In {\em Thirty-Second AAAI Conference on Artificial Intelligence},
  2018.

\bibitem{hessel2017rainbow}
Matteo Hessel, Joseph Modayil, Hado Van~Hasselt, Tom Schaul, Georg Ostrovski,
  Will Dabney, Dan Horgan, Bilal Piot, Mohammad Azar, and David Silver.
\newblock Rainbow: Combining improvements in deep reinforcement learning.
\newblock {\em arXiv preprint arXiv:1710.02298}, 2017.

\bibitem{hill2018stable}
Ashley Hill, Antonin Raffin, Maximilian Ernestus, Adam Gleave, Anssi
  Kanervisto, Rene Traore, Prafulla Dhariwal, Christopher Hesse, Oleg Klimov,
  Alex Nichol, Matthias Plappert, Alec Radford, John Schulman, Szymon Sidor,
  and Yuhuai Wu.
\newblock Stable baselines.
\newblock \url{https://github.com/hill-a/stable-baselines}, 2018.

\bibitem{hinton2015distilling}
Geoffrey Hinton, Oriol Vinyals, and Jeff Dean.
\newblock Distilling the knowledge in a neural network.
\newblock {\em arXiv preprint arXiv:1503.02531}, 2015.

\bibitem{hinton2006fast}
Geoffrey~E Hinton, Simon Osindero, and Yee-Whye Teh.
\newblock A fast learning algorithm for deep belief nets.
\newblock {\em Neural Computation}, 18(7):1527--1554, 2006.

\bibitem{hinton2006reducing}
Geoffrey~E Hinton and Ruslan~R Salakhutdinov.
\newblock Reducing the dimensionality of data with neural networks.
\newblock {\em Science}, 313(5786):504--507, 2006.

\bibitem{hinton2012improving}
Geoffrey~E Hinton, Nitish Srivastava, Alex Krizhevsky, Ilya Sutskever, and
  Ruslan~R Salakhutdinov.
\newblock Improving neural networks by preventing co-adaptation of feature
  detectors.
\newblock {\em arXiv preprint arXiv:1207.0580}, 2012.

\bibitem{hochreiter1991untersuchungen}
Sepp Hochreiter.
\newblock Untersuchungen zu dynamischen neuronalen {Netzen}.
\newblock {\em Diploma, Technische Universit{\"a}t M{\"u}nchen}, 91(1), 1991.

\bibitem{hochreiter2001gradient}
Sepp Hochreiter, Yoshua Bengio, Paolo Frasconi, and J{\"u}rgen Schmidhuber.
\newblock Gradient flow in recurrent nets: the difficulty of learning long-term
  dependencies.
\newblock In {\em A Field Guide to Dynamical Recurrent Neural Networks}. IEEE
  Press, 2001.

\bibitem{hochreiter1997long}
Sepp Hochreiter and J{\"u}rgen Schmidhuber.
\newblock Long short-term memory.
\newblock {\em Neural Computation}, 9(8):1735--1780, 1997.

\bibitem{holland1975adaptation}
John Holland.
\newblock Adaptation in natural and artificial systems: an introductory
  analysis with application to biology.
\newblock {\em Control and Artificial Intelligence}, 1975.

\bibitem{holland1992genetic}
John~H Holland.
\newblock Genetic algorithms.
\newblock {\em Scientific American}, 267(1):66--73, 1992.

\bibitem{holt2005medical}
Alec Holt, Isabelle Bichindaritz, Rainer Schmidt, and Petra Perner.
\newblock Medical applications in case-based reasoning.
\newblock {\em The Knowledge Engineering Review}, 20(3):289--292, 2005.

\bibitem{hoos2004stochastic}
Holger~H Hoos and Thomas St{\"u}tzle.
\newblock {\em Stochastic local search: Foundations and applications}.
\newblock Elsevier, 2004.

\bibitem{hopfield1982neural}
John~J Hopfield.
\newblock Neural networks and physical systems with emergent collective
  computational abilities.
\newblock {\em Proceedings of the National Academy of Sciences},
  79(8):2554--2558, 1982.

\bibitem{hordern1986sliding}
Edward Hordern.
\newblock {\em Sliding piece puzzles}.
\newblock Number~4. Oxford University Press, USA, 1986.

\bibitem{hospedales2020meta}
Timothy Hospedales, Antreas Antoniou, Paul Micaelli, and Amos Storkey.
\newblock Meta-learning in neural networks: A survey.
\newblock {\em arXiv preprint arXiv:2004.05439}, 2020.

\bibitem{howard1964dynamic}
Ronald~A Howard.
\newblock {\em Dynamic programming and {Markov} processes}.
\newblock New York: John Wiley, 1964.

\bibitem{hsieh2018comparison}
Meng~Hsuen Hsieh, Meng~Ju Hsieh, Chin-Ming Chen, Chia-Chang Hsieh, Chien-Ming
  Chao, and Chih-Cheng Lai.
\newblock Comparison of machine learning models for the prediction of mortality
  of patients with unplanned extubation in intensive care units.
\newblock {\em Scientific Reports}, 8(1):17116, 2018.

\bibitem{hsieh2007fairness}
Ming~Yu Hsieh and Shi-Chun Tsai.
\newblock On the fairness and complexity of generalized k-in-a-row games.
\newblock {\em Theoretical Computer Science}, 385(1-3):88--100, 2007.

\bibitem{hsu2004behind}
Feng-Hsiung Hsu.
\newblock {\em Behind {Deep Blue}: Building the computer that defeated the
  world chess champion}.
\newblock Princeton University Press, 2004.

\bibitem{hsu1990grandmaster}
Feng-hsiung Hsu, Thomas Anantharaman, Murray Campbell, and Andreas Nowatzyk.
\newblock A grandmaster chess machine.
\newblock {\em Scientific American}, 263(4):44--51, 1990.

\bibitem{huang2015pruning}
Bojun Huang.
\newblock Pruning game tree by rollouts.
\newblock In {\em AAAI}, pages 1165--1173, 2015.

\bibitem{huang2017densely}
Gao Huang, Zhuang Liu, Laurens Van Der~Maaten, and Kilian~Q Weinberger.
\newblock Densely connected convolutional networks.
\newblock In {\em Proceedings of the IEEE Conference on Computer Vision and
  Pattern Recognition}, pages 4700--4708, 2017.

\bibitem{huang2017adversarial}
Sandy Huang, Nicolas Papernot, Ian Goodfellow, Yan Duan, and Pieter Abbeel.
\newblock Adversarial attacks on neural network policies.
\newblock {\em arXiv preprint arXiv:1702.02284}, 2017.

\bibitem{huang2013mohex}
Shih-Chieh Huang, Broderick Arneson, Ryan~B Hayward, Martin M{\"u}ller, and
  Jakub Pawlewicz.
\newblock Mohex 2.0: a pattern-based {MCTS Hex} player.
\newblock In {\em International Conference on Computers and Games}, pages
  60--71. Springer, 2013.

\bibitem{huang2003learning}
Timothy Huang, Graeme Connell, and Bryan McQuade.
\newblock Learning opening strategy in the game of {Go}.
\newblock In {\em FLAIRS Conference}, pages 434--438, 2003.

\bibitem{hubel1963shape}
David~H Hubel and TN~Wiesel.
\newblock Shape and arrangement of columns in cat's striate cortex.
\newblock {\em The Journal of Physiology}, 165(3):559--568, 1963.

\bibitem{hubel1968receptive}
David~H Hubel and Torsten~N Wiesel.
\newblock Receptive fields and functional architecture of monkey striate
  cortex.
\newblock {\em The Journal of Physiology}, 195(1):215--243, 1968.

\bibitem{huberman1968program}
Barbara J~(Liskov) Huberman.
\newblock A program to play chess end games.
\newblock Technical report, Stanford University Department of Computer Science,
  1968.

\bibitem{fanhui2016deepmind}
Fan Hui.
\newblock {Fan Hui} commentary game 2 move 37 {Lee Sedol}.
  \url{https://deepmind.com/research/alphago/match-archive/alphago-games-english/},
  2018.

\bibitem{hui2018rl}
Jonathan Hui.
\newblock {RL---DQN Deep Q-network}
  \url{https://medium.com/@jonathan_hui/rl-dqn-deep-q-network-e207751f7ae4}.
\newblock Medium post.

\bibitem{huisman2020deep}
Mike Huisman, Jan van Rijn, and Aske Plaat. In: Pavel~Brazdil et~al.
\newblock {\em Metalearning for deep neural networks}.
\newblock Springer, 2021.

\bibitem{huizinga1938homo}
Johan Huizinga.
\newblock {\em Homo ludens: proeve eener bepaling van het spel-element der
  cultuur. English: Homo Ludens: A Study of the Play-Element in Culture}.
\newblock Amsterdam University Press, London: Routledge \& Kegan Paul, 1938,
  2008.

\bibitem{hutson2018artificial}
Matthew Hutson.
\newblock {Artificial Intelligence} faces reproducibility crisis, 2018.

\bibitem{hutter2011sequential}
Frank Hutter, Holger~H Hoos, and Kevin Leyton-Brown.
\newblock Sequential model-based optimization for general algorithm
  configuration.
\newblock In {\em International Conference on Learning and Intelligent
  Optimization}, pages 507--523. Springer, 2011.

\bibitem{hyatt1997crafty}
Robert~M Hyatt and Monty Newborn.
\newblock Crafty goes deep.
\newblock {\em ICGA Journal}, 20(2):79--86, 1997.

\bibitem{iida2002computer}
Hiroyuki Iida, Makoto Sakuta, and Jeff Rollason.
\newblock Computer shogi.
\newblock {\em Artificial Intelligence}, 134(1-2):121--144, 2002.

\bibitem{queensland2018brain}
Queensland~Brain Institute.
\newblock History of artificial intelligence
  \url{https://qbi.uq.edu.au/brain/intelligent-machines/history-artificial-intelligence},
  2018.

\bibitem{ioffe2017batch}
Sergey Ioffe.
\newblock Batch renormalization: Towards reducing minibatch dependence in
  batch-normalized models.
\newblock In {\em Advances in Neural Information Processing Systems}, pages
  1945--1953, 2017.

\bibitem{islam2017reproducibility}
Riashat Islam, Peter Henderson, Maziar Gomrokchi, and Doina Precup.
\newblock Reproducibility of benchmarked deep reinforcement learning tasks for
  continuous control.
\newblock {\em arXiv preprint arXiv:1708.04133}, 2017.

\bibitem{iwata1994othello}
Shigeki Iwata and Takumi Kasai.
\newblock The othello game on an $n\times n$ board is {PSPACE}-complete.
\newblock {\em Theoretical Computer Science}, 123(2):329--340, 1994.

\bibitem{jaderberg2018human}
Max Jaderberg, Wojciech~M Czarnecki, Iain Dunning, Luke Marris, Guy Lever,
  Antonio~Garcia Castaneda, Charles Beattie, Neil~C Rabinowitz, Ari~S Morcos,
  Avraham Ruderman, et~al.
\newblock Human-level performance in first-person multiplayer games with
  population-based deep reinforcement learning.
\newblock {\em arXiv preprint arXiv:1807.01281}, 2018.

\bibitem{jaderberg2019human}
Max Jaderberg, Wojciech~M Czarnecki, Iain Dunning, Luke Marris, Guy Lever,
  Antonio~Garcia Castaneda, Charles Beattie, Neil~C Rabinowitz, Ari~S Morcos,
  Avraham Ruderman, et~al.
\newblock Human-level performance in {3D} multiplayer games with
  population-based reinforcement learning.
\newblock {\em Science}, 364(6443):859--865, 2019.

\bibitem{jaderberg2017population}
Max Jaderberg, Valentin Dalibard, Simon Osindero, Wojciech~M Czarnecki, Jeff
  Donahue, Ali Razavi, Oriol Vinyals, Tim Green, Iain Dunning, Karen Simonyan,
  et~al.
\newblock Population based training of neural networks.
\newblock {\em arXiv preprint arXiv:1711.09846}, 2017.

\bibitem{jia2014caffe}
Yangqing Jia, Evan Shelhamer, Jeff Donahue, Sergey Karayev, Jonathan Long, Ross
  Girshick, Sergio Guadarrama, and Trevor Darrell.
\newblock Caffe: Convolutional architecture for fast feature embedding.
\newblock In {\em Proceedings of the 22nd ACM International Conference on
  Multimedia}, pages 675--678. ACM, 2014.

\bibitem{jin2014subgroup}
Nanlin Jin, Peter Flach, Tom Wilcox, Royston Sellman, Joshua Thumim, and Arno
  Knobbe.
\newblock Subgroup discovery in smart electricity meter data.
\newblock {\em IEEE Transactions on Industrial Informatics}, 10(2):1327--1336,
  2014.

\bibitem{johanson2013measuring}
Michael Johanson.
\newblock Measuring the size of large no-limit poker games.
\newblock {\em arXiv preprint arXiv:1302.7008}, 2013.

\bibitem{johnson2011beyond}
Matthew Johnson, Jeffrey~M Bradshaw, Paul~J Feltovich, Robert~R Hoffman,
  Catholijn Jonker, Birna van Riemsdijk, and Maarten Sierhuis.
\newblock Beyond cooperative robotics: The central role of interdependence in
  coactive design.
\newblock {\em IEEE Intelligent Systems}, 26(3):81--88, 2011.

\bibitem{jonker2017automated}
Catholijn~M Jonker, Reyhan Aydogan, Tim Baarslag, Katsuhide Fujita, Takayuki
  Ito, and Koen~V Hindriks.
\newblock Automated negotiating agents competition {(ANAC)}.
\newblock In {\em AAAI}, pages 5070--5072, 2017.

\bibitem{jonker2007agent}
Catholijn~M Jonker, Valentin Robu, and Jan Treur.
\newblock An agent architecture for multi-attribute negotiation using
  incomplete preference information.
\newblock {\em Autonomous Agents and Multi-Agent Systems}, 15(2):221--252,
  2007.

\bibitem{jordan1998learning}
Michael~Irwin Jordan.
\newblock {\em Learning in graphical models}, volume~89.
\newblock Springer Science \& Business Media, 1998.

\bibitem{jouppi2017datacenter}
Norman~P Jouppi, Cliff Young, Nishant Patil, David Patterson, Gaurav Agrawal,
  Raminder Bajwa, Sarah Bates, Suresh Bhatia, Nan Boden, Al~Borchers, et~al.
\newblock In-datacenter performance analysis of a tensor processing unit.
\newblock In {\em Computer Architecture (ISCA), 2017 ACM/IEEE 44th Annual
  International Symposium on}, pages 1--12. IEEE, 2017.

\bibitem{junghanns1997search}
Andreas Junghanns and Jonathan Schaeffer.
\newblock Search versus knowledge in game-playing programs revisited.
\newblock In {\em IJCAI (1)}, pages 692--697, 1997.

\bibitem{kaelbling1993hierarchical}
Leslie~Pack Kaelbling.
\newblock Hierarchical learning in stochastic domains: Preliminary results.
\newblock In {\em Proceedings of the tenth International Conference on Machine
  Learning}, volume 951, pages 167--173, 1993.

\bibitem{kaelbling1998planning}
Leslie~Pack Kaelbling, Michael~L Littman, and Anthony~R Cassandra.
\newblock Planning and acting in partially observable stochastic domains.
\newblock {\em Artificial Intelligence}, 101(1-2):99--134, 1998.

\bibitem{kaelbling1996reinforcement}
Leslie~Pack Kaelbling, Michael~L Littman, and Andrew~W Moore.
\newblock Reinforcement learning: A survey.
\newblock {\em Journal of Artificial Intelligence Research}, 4:237--285, 1996.

\bibitem{kaelbling2010hierarchical}
Leslie~Pack Kaelbling and Tom{\'a}s Lozano-P{\'e}rez.
\newblock Hierarchical planning in the now.
\newblock In {\em Workshops at the Twenty-Fourth AAAI Conference on Artificial
  Intelligence}, 2010.

\bibitem{kahneman2011thinking}
Daniel Kahneman.
\newblock {\em Thinking, fast and slow}.
\newblock Farrar, Straus and Giroux, 2011.

\bibitem{kaindl1982dynamic}
H~Kaindl.
\newblock Dynamic control of the quiescence search in computer chess.
\newblock {\em Cybernetics and Systems Research}, pages 973--977, 1982.

\bibitem{kaindl1988minimaxing}
Hermann Kaindl.
\newblock Minimaxing: Theory and practice.
\newblock {\em AI Magazine}, 9(3):69, 1988.

\bibitem{kaiser2019model}
Lukasz Kaiser, Mohammad Babaeizadeh, Piotr Milos, Blazej Osinski, Roy~H
  Campbell, Konrad Czechowski, Dumitru Erhan, Chelsea Finn, Piotr Kozakowski,
  Sergey Levine, et~al.
\newblock Model-based reinforcement learning for {Atari}.
\newblock {\em arXiv preprint arXiv:1903.00374}, 2019.

\bibitem{learn}
Satwik Kansal and Brendan Martin.
\newblock Learn data science webpage., 2018.

\bibitem{kapturowski2018recurrent}
Steven Kapturowski, Georg Ostrovski, John Quan, Remi Munos, and Will Dabney.
\newblock Recurrent experience replay in distributed reinforcement learning.
\newblock In {\em International conference on learning representations}, 2018.

\bibitem{karakovskiy2012mario}
Sergey Karakovskiy and Julian Togelius.
\newblock The {Mario AI} benchmark and competitions.
\newblock {\em IEEE Transactions on Computational Intelligence and AI in
  Games}, 4(1):55--67, 2012.

\bibitem{karpathy2015}
Andrej Karpathy.
\newblock The unreasonable effectiveness of recurrent neural networks.
  \url{http://karpathy.github.io/2015/05/21/rnn-effectiveness/}, 2015.

\bibitem{karpathy2016pong}
Andrej Karpathy.
\newblock Deep reinforcement learning: Pong from pixels
  \url{http://karpathy.github.io/2016/05/31/rl/}.
\newblock Andrej Karpathy Blog, 2016.

\bibitem{karras2017progressive}
Tero Karras, Timo Aila, Samuli Laine, and Jaakko Lehtinen.
\newblock Progressive growing of {GANs} for improved quality, stability, and
  variation.
\newblock {\em arXiv preprint arXiv:1710.10196}, 2017.

\bibitem{katehakis1987multi}
Michael~N Katehakis and Arthur~F Veinott~Jr.
\newblock The multi-armed bandit problem: decomposition and computation.
\newblock {\em Mathematics of Operations Research}, 12(2):262--268, 1987.

\bibitem{kelleher2015fundamentals}
John~D Kelleher, Brian Mac~Namee, and Aoife D'Arcy.
\newblock {\em Fundamentals of machine learning for predictive data analytics:
  algorithms, worked examples, and case studies}.
\newblock MIT Press, 2015.

\bibitem{kelly2017multi}
Stephen Kelly and Malcolm~I Heywood.
\newblock Multi-task learning in atari video games with emergent tangled
  program graphs.
\newblock In {\em Proceedings of the Genetic and Evolutionary Computation
  Conference}, pages 195--202. ACM, 2017.

\bibitem{kelly2018emergent}
Stephen Kelly and Malcolm~I Heywood.
\newblock Emergent tangled program graphs in multi-task learning.
\newblock In {\em IJCAI}, pages 5294--5298, 2018.

\bibitem{kennedy2006swarm}
James Kennedy.
\newblock Swarm intelligence.
\newblock In {\em Handbook of nature-inspired and innovative computing}, pages
  187--219. Springer, 2006.

\bibitem{kerschke2019automated}
Pascal Kerschke, Holger~H Hoos, Frank Neumann, and Heike Trautmann.
\newblock Automated algorithm selection: Survey and perspectives.
\newblock {\em Evolutionary Computation}, 27(1):3--45, 2019.

\bibitem{kierulf1990smart}
Anders Kierulf, Ken Chen, and Jurg Nievergelt.
\newblock Smart game board and {Go} explorer: a study in software and knowledge
  engineering.
\newblock {\em Communications of the ACM}, 33(2):152--166, 1990.

\bibitem{kimura2000analysis}
Hajime Kimura, Shigenobu Kobayashi, et~al.
\newblock An analysis of actor-critic algorithms using eligibility traces:
  reinforcement learning with imperfect value functions.
\newblock {\em Journal of Japanese Society for Artificial Intelligence},
  15(2):267--275, 2000.

\bibitem{kinga2015method}
D~Kinga and J~Ba.
\newblock Adam: method for stochastic optimization.
\newblock In {\em International Conference on Learning Representations (ICLR)},
  volume~5, 2015.

\bibitem{kitano1990designing}
Hiroaki Kitano.
\newblock Designing neural networks using genetic algorithms with graph
  generation system.
\newblock {\em Complex Systems}, 4(4):461--476, 1990.

\bibitem{kloetzer2010monte}
Julien Kloetzer.
\newblock {Monte-Carlo} opening books for amazons.
\newblock In {\em International Conference on Computers and Games}, pages
  124--135. Springer, 2010.

\bibitem{knoblock1990learning}
Craig~A Knoblock.
\newblock Learning abstraction hierarchies for problem solving.
\newblock In {\em AAAI}, pages 923--928, 1990.

\bibitem{knuth1975analysis}
Donald~E Knuth and Ronald~W Moore.
\newblock An analysis of alpha-beta pruning.
\newblock {\em Artificial Intelligence}, 6(4):293--326, 1975.

\bibitem{knuth1997art}
Donald~Ervin Knuth.
\newblock {\em The art of computer programming: sorting and searching},
  volume~3.
\newblock Pearson Education, 1997.

\bibitem{kober2013reinforcement}
Jens Kober, J~Andrew Bagnell, and Jan Peters.
\newblock Reinforcement learning in robotics: A survey.
\newblock {\em The International Journal of Robotics Research},
  32(11):1238--1274, 2013.

\bibitem{koch2004quest}
Christof Koch.
\newblock {\em The quest for consciousness: A neurobiological approach}.
\newblock Roberts and Company Englewood, CO, 2004.

\bibitem{kocsis2006bandit}
Levente Kocsis and Csaba Szepesv{\'a}ri.
\newblock Bandit based {Monte-Carlo} planning.
\newblock In {\em European Conference on Machine Learning}, pages 282--293.
  Springer, 2006.

\bibitem{koeman2018starcraft}
Vincent~J Koeman, Harm~J Griffioen, Danny~C Plenge, and Koen~V Hindriks.
\newblock Starcraft as a testbed for engineering complex distributed systems
  using cognitive agent technology.
\newblock In {\em Proceedings of the 17th International Conference on
  Autonomous Agents and MultiAgent Systems}, pages 1983--1985. International
  Foundation for Autonomous Agents and Multiagent Systems, 2018.

\bibitem{koetsier2001prehistory}
Teun Koetsier.
\newblock On the prehistory of programmable machines: musical automata, looms,
  calculators.
\newblock {\em Mechanism and Machine Theory}, 36(5):589--603, 2001.

\bibitem{koller1997representations}
Daphne Koller and Avi Pfeffer.
\newblock Representations and solutions for game-theoretic problems.
\newblock {\em Artificial Intelligence}, 94(1-2):167--215, 1997.

\bibitem{konda2000actor}
Vijay~R Konda and John~N Tsitsiklis.
\newblock Actor--critic algorithms.
\newblock In {\em Advances in Neural Information Processing Systems}, pages
  1008--1014, 2000.

\bibitem{konda1999actor}
Vijaymohan~R Konda and Vivek~S Borkar.
\newblock {Actor--Critic}-type learning algorithms for {Markov Decision
  Processes}.
\newblock {\em SIAM Journal on Control and Optimization}, 38(1):94--123, 1999.

\bibitem{kool2018attention}
Wouter Kool, Herke Van~Hoof, and Max Welling.
\newblock Attention, learn to solve routing problems!
\newblock {\em arXiv preprint arXiv:1803.08475}, 2018.

\bibitem{korf2000recent}
Richard~E Korf.
\newblock Recent progress in the design and analysis of admissible heuristic
  functions.
\newblock In {\em International Symposium on Abstraction, Reformulation, and
  Approximation}, pages 45--55. Springer, 2000.

\bibitem{korf1996finding}
Richard~E Korf and Larry~A Taylor.
\newblock Finding optimal solutions to the twenty-four puzzle.
\newblock In {\em Proceedings of the national conference on Artificial
  Intelligence}, pages 1202--1207, 1996.

\bibitem{kotthoff2017auto}
Lars Kotthoff, Chris Thornton, Holger~H Hoos, Frank Hutter, and Kevin
  Leyton-Brown.
\newblock {Auto-WEKA} 2.0: Automatic model selection and hyperparameter
  optimization in {WEKA}.
\newblock {\em The Journal of Machine Learning Research}, 18(1):826--830, 2017.

\bibitem{kraus1994negotiation}
Sarit Kraus, Eithan Ephrati, and Daniel Lehmann.
\newblock Negotiation in a non-cooperative environment.
\newblock {\em Journal of Experimental \& Theoretical Artificial Intelligence},
  3(4):255--281, 1994.

\bibitem{kraus1988diplomat}
Sarit Kraus and Daniel Lehmann.
\newblock Diplomat, an agent in a multi agent environment: An overview.
\newblock In {\em Computers and Communications, 1988. Conference Proceedings.,
  Seventh Annual International Phoenix Conference on}, pages 434--438. IEEE,
  1988.

\bibitem{krizhevsky2009learning}
Alex Krizhevsky and Geoffrey Hinton.
\newblock Learning multiple layers of features from tiny images.
\newblock Technical report, Citeseer, 2009.

\bibitem{krizhevsky2012imagenet}
Alex Krizhevsky, Ilya Sutskever, and Geoffrey~E Hinton.
\newblock Imagenet classification with deep convolutional neural networks.
\newblock In {\em Advances in Neural Information Processing Systems}, pages
  1097--1105, 2012.

\bibitem{krueger2009flexible}
Kai~A Krueger and Peter Dayan.
\newblock Flexible shaping: How learning in small steps helps.
\newblock {\em Cognition}, 110(3):380--394, 2009.

\bibitem{kuipers2013improving}
Jan Kuipers, Aske Plaat, JAM Vermaseren, and H~Jaap van~den Herik.
\newblock Improving multivariate {Horner} schemes with {Monte Carlo} tree
  search.
\newblock {\em Computer Physics Communications}, 184(11):2391--2395, 2013.

\bibitem{kulkarni2016hierarchical}
Tejas~D Kulkarni, Karthik Narasimhan, Ardavan Saeedi, and Josh Tenenbaum.
\newblock Hierarchical deep reinforcement learning: Integrating temporal
  abstraction and intrinsic motivation.
\newblock In {\em Advances in Neural Information Processing Systems}, pages
  3675--3683, 2016.

\bibitem{kumar1988general}
Vipin Kumar, Dana~S Nau, and Laveen~N Kanal.
\newblock A general branch-and-bound formulation for and/or graph and game tree
  search.
\newblock In {\em Search in Artificial Intelligence}, pages 91--130. Springer,
  1988.

\bibitem{kunanusont2017general}
Kamolwan Kunanusont, Simon~M Lucas, and Diego P{\'e}rez-Li{\'e}bana.
\newblock General video game {AI}: learning from screen capture.
\newblock In {\em Evolutionary Computation (CEC), 2017 IEEE Congress on}, pages
  2078--2085. IEEE, 2017.

\bibitem{lagoudakis2003least}
Michail~G Lagoudakis and Ronald Parr.
\newblock Least-squares policy iteration.
\newblock {\em Journal of Machine Learning Research}, 4(Dec):1107--1149, 2003.

\bibitem{lai2015giraffe}
Matthew Lai.
\newblock Giraffe: Using deep reinforcement learning to play chess.
\newblock {\em arXiv preprint arXiv:1509.01549}, 2015.

\bibitem{lai1985asymptotically}
Tze~Leung Lai and Herbert Robbins.
\newblock Asymptotically efficient adaptive allocation rules.
\newblock {\em Advances in Applied Mathematics}, 6(1):4--22, 1985.

\bibitem{lai1995machine}
Tze-Leung Lai and Sidney Yakowitz.
\newblock Machine learning and nonparametric bandit theory.
\newblock {\em IEEE Transactions on Automatic Control}, 40(7):1199--1209, 1995.

\bibitem{laird1987soar}
John~E Laird, Allen Newell, and Paul~S Rosenbloom.
\newblock Soar: An architecture for general intelligence.
\newblock {\em Artificial Intelligence}, 33(1):1--64, 1987.

\bibitem{lake2015human}
Brenden~M Lake, Ruslan Salakhutdinov, and Joshua~B Tenenbaum.
\newblock Human-level concept learning through probabilistic program induction.
\newblock {\em Science}, 350(6266):1332--1338, 2015.

\bibitem{lake1993solving}
Robert Lake, Jonathan Schaeffer, and Paul Lu.
\newblock {\em Solving large retrograde analysis problems using a network of
  workstations}.
\newblock Department of Computing Science, University of Alberta, 1993.

\bibitem{lanctot2019openspiel}
Marc Lanctot, Edward Lockhart, Jean-Baptiste Lespiau, Vinicius Zambaldi,
  Satyaki Upadhyay, Julien P{\'e}rolat, Sriram Srinivasan, Finbarr Timbers,
  Karl Tuyls, Shayegan Omidshafiei, et~al.
\newblock Openspiel: A framework for reinforcement learning in games.
\newblock {\em arXiv preprint arXiv:1908.09453}, 2019.

\bibitem{lanctot2013monte}
Marc Lanctot, Abdallah Saffidine, Joel Veness, Christopher Archibald, and
  Mark~HM Winands.
\newblock {Monte Carlo*}-minimax search.
\newblock In {\em IJCAI}, pages 580--586, 2013.

\bibitem{landis2001aleksandr}
Evgenii~Mikhailovich Landis and Isaak~Moiseevich Yaglom.
\newblock About {Aleksandr Semenovich Kronrod}.
\newblock {\em Russian Mathematical Surveys}, 56(5):993--1007, 2001.

\bibitem{lara2013review}
Ra{\'u}l Lara-Cabrera, Carlos Cotta, and Antonio~J Fern{\'a}ndez-Leiva.
\newblock A review of computational intelligence in {RTS} games.
\newblock In {\em 2013 IEEE Symposium on Foundations of Computational
  Intelligence (FOCI)}, pages 114--121. IEEE, 2013.

\bibitem{laterre2018ranked}
Alexandre Laterre, Yunguan Fu, Mohamed~Khalil Jabri, Alain-Sam Cohen, David
  Kas, Karl Hajjar, Torbjorn~S Dahl, Amine Kerkeni, and Karim Beguir.
\newblock Ranked reward: Enabling self-play reinforcement learning for
  combinatorial optimization.
\newblock {\em arXiv preprint arXiv:1807.01672}, 2018.

\bibitem{lauritzen1996graphical}
Steffen~L Lauritzen.
\newblock {\em Graphical models}, volume~17.
\newblock Clarendon Press, 1996.

\bibitem{lecun2015deep}
Yann LeCun, Yoshua Bengio, and Geoffrey Hinton.
\newblock Deep learning.
\newblock {\em Nature}, 521(7553):436, 2015.

\bibitem{lecun1989backpropagation}
Yann LeCun, Bernhard Boser, John~S Denker, Donnie Henderson, Richard~E Howard,
  Wayne Hubbard, and Lawrence~D Jackel.
\newblock Backpropagation applied to handwritten zip code recognition.
\newblock {\em Neural Computation}, 1(4):541--551, 1989.

\bibitem{lecun1998gradient}
Yann LeCun, L{\'e}on Bottou, Yoshua Bengio, and Patrick Haffner.
\newblock Gradient-based learning applied to document recognition.
\newblock {\em Proceedings of the IEEE}, 86(11):2278--2324, 1998.

\bibitem{lee2010special}
Chang-Shing Lee, Martin M{\"u}ller, and Olivier Teytaud.
\newblock Special issue on {Monte Carlo} techniques and computer {Go}.
\newblock {\em IEEE Transactions on Computational Intelligence and AI in
  Games}, (4):225--228, 2010.

\bibitem{lee2009convolutional}
Honglak Lee, Roger Grosse, Rajesh Ranganath, and Andrew~Y Ng.
\newblock Convolutional deep belief networks for scalable unsupervised learning
  of hierarchical representations.
\newblock In {\em Proceedings of the 26th Annual International Conference on
  Machine Learning}, pages 609--616. ACM, 2009.

\bibitem{legg2007collection}
Shane Legg, Marcus Hutter, et~al.
\newblock A collection of definitions of intelligence.
\newblock {\em Frontiers in Artificial Intelligence and Applications}, 157:17,
  2007.

\bibitem{leibo2019autocurricula}
Joel~Z Leibo, Edward Hughes, Marc Lanctot, and Thore Graepel.
\newblock Autocurricula and the emergence of innovation from social
  interaction: A manifesto for multi-agent intelligence research.
\newblock {\em arXiv preprint arXiv:1903.00742}, 2019.

\bibitem{leiserson1998programming}
Charles Leiserson and Aske Plaat.
\newblock Programming parallel applications in {Cilk}.
\newblock {\em SINEWS: SIAM News}, 31(4):6--7, 1998.

\bibitem{fruit}
Fabien Letouzey.
\newblock Fruit chess program. \url{http://www.fruitchess.com/about.htm}.

\bibitem{levine2013guided}
Sergey Levine and Vladlen Koltun.
\newblock Guided policy search.
\newblock In {\em International Conference on Machine Learning}, pages 1--9,
  2013.

\bibitem{levy2018hierarchical}
Andrew Levy, Robert Platt, and Kate Saenko.
\newblock Hierarchical reinforcement learning with hindsight.
\newblock {\em arXiv preprint arXiv:1805.08180}, 2018.

\bibitem{levy1991computers}
David Levy and Monty Newborn.
\newblock {\em How computers play chess}, volume~8.
\newblock Computer Science Press, New York, 1991.

\bibitem{liang2017rllib}
Eric Liang, Richard Liaw, Philipp Moritz, Robert Nishihara, Roy Fox, Ken
  Goldberg, Joseph~E Gonzalez, Michael~I Jordan, and Ion Stoica.
\newblock {RLlib:} abstractions for distributed reinforcement learning.
\newblock {\em arXiv preprint arXiv:1712.09381}, 2017.

\bibitem{lillicrap2015continuous}
Timothy~P Lillicrap, Jonathan~J Hunt, Alexander Pritzel, Nicolas Heess, Tom
  Erez, Yuval Tassa, David Silver, and Daan Wierstra.
\newblock Continuous control with deep reinforcement learning.
\newblock {\em arXiv preprint arXiv:1509.02971}, 2015.

\bibitem{lim2018}
Milton Lim.
\newblock History of {AI Winters}.
  \url{https://www.actuaries.digital/2018/09/05/history-of-ai-winters/}, 2018.

\bibitem{lin1992self}
Long-Ji Lin.
\newblock Self-improving reactive agents based on reinforcement learning,
  planning and teaching.
\newblock {\em Machine Learning}, 8(3-4):293--321, 1992.

\bibitem{lin1993reinforcement}
Long-Ji Lin.
\newblock Reinforcement learning for robots using neural networks.
\newblock Technical report, Carnegie-Mellon Univ Pittsburgh PA School of
  Computer Science, 1993.

\bibitem{lin2010can}
Raz Lin and Sarit Kraus.
\newblock Can automated agents proficiently negotiate with humans?
\newblock {\em Communications of the ACM}, 53(1):78--88, 2010.

\bibitem{lin2014genius}
Raz Lin, Sarit Kraus, Tim Baarslag, Dmytro Tykhonov, Koen Hindriks, and
  Catholijn~M Jonker.
\newblock Genius: An integrated environment for supporting the design of
  generic automated negotiators.
\newblock {\em Computational Intelligence}, 30(1):48--70, 2014.

\bibitem{lisetti1998affective}
Christine~L Lisetti.
\newblock {\em Affective computing}.
\newblock Springer, 1998.

\bibitem{littman1994markov}
Michael~L Littman.
\newblock {Markov} games as a framework for multi-agent reinforcement learning.
\newblock In {\em Machine Learning Proceedings 1994}, pages 157--163. Elsevier,
  1994.

\bibitem{liu2012sparse}
Hui Liu, Song Yu, Zhangxin Chen, Ben Hsieh, and Lei Shao.
\newblock Sparse matrix-vector multiplication on {NVIDIA GPU}.
\newblock {\em International Journal of Numerical Analysis \& Modeling, Series
  B}, 3(2):185--191, 2012.

\bibitem{lopez2016irace}
Manuel L{\'o}pez-Ib{\'a}{\~n}ez, J{\'e}r{\'e}mie Dubois-Lacoste,
  Leslie~P{\'e}rez C{\'a}ceres, Mauro Birattari, and Thomas St{\"u}tzle.
\newblock The irace package: Iterated racing for automatic algorithm
  configuration.
\newblock {\em Operations Research Perspectives}, 3:43--58, 2016.

\bibitem{lorentz2011experiments}
Richard~J Lorentz.
\newblock Experiments with {Monte-Carlo} tree search in the game of {Havannah}.
\newblock {\em ICGA Journal}, 34(3):140--149, 2011.

\bibitem{lorentz2011mcts}
Richard~J Lorentz.
\newblock An {MCTS} program to play {EinStein} w{\"u}rfelt nicht!
\newblock In {\em Advances in Computer Games}, pages 52--59. Springer, 2011.

\bibitem{loveland2016automated}
Donald~W Loveland.
\newblock {\em Automated Theorem Proving: a logical basis}.
\newblock Elsevier, 2016.

\bibitem{lowe2017multi}
Ryan Lowe, Yi~Wu, Aviv Tamar, Jean Harb, Pieter Abbeel, and Igor Mordatch.
\newblock Multi-agent {Actor--Critic} for mixed cooperative-competitive
  environments.
\newblock In {\em Advances in Neural Information Processing Systems}, pages
  6379--6390, 2017.

\bibitem{lutz1996programming}
Mark Lutz.
\newblock {\em Programming {Python}}, volume~8.
\newblock O'Reilly, 1996.

\bibitem{maaten2008visualizing}
Laurens van~der Maaten and Geoffrey Hinton.
\newblock Visualizing data using {t-SNE}.
\newblock {\em Journal of Machine Learning Research}, 9(Nov):2579--2605, 2008.

\bibitem{macnamara2014deliberate}
Brooke~N Macnamara, David~Z Hambrick, and Frederick~L Oswald.
\newblock Deliberate practice and performance in music, games, sports,
  education, and professions: A meta-analysis.
\newblock {\em Psychological Science}, 25(8):1608--1618, 2014.

\bibitem{maddison2014move}
Chris~J Maddison, Aja Huang, Ilya Sutskever, and David Silver.
\newblock Move evaluation in {Go} using deep convolutional neural networks.
\newblock {\em arXiv preprint arXiv:1412.6564}, 2014.

\bibitem{madrigalcheckers}
A~Madrigal.
\newblock How checkers was solved, the story of a duel between two men, one who
  dies, and the nature of the quest to build artificial intelligence (2017).
\newblock {\em The Atlantic}, July 2017.

\bibitem{makar2001hierarchical}
Rajbala Makar, Sridhar Mahadevan, and Mohammad Ghavamzadeh.
\newblock Hierarchical multi-agent reinforcement learning.
\newblock In {\em Proceedings of the Fifth International Conference on
  Autonomous Agents}, pages 246--253. ACM, 2001.

\bibitem{matiisen2017teacher}
Tambet Matiisen, Avital Oliver, Taco Cohen, and John Schulman.
\newblock Teacher-student curriculum learning.
\newblock {\em arXiv preprint arXiv:1707.00183}, 2017.

\bibitem{matsugu2003subject}
Masakazu Matsugu, Katsuhiko Mori, Yusuke Mitari, and Yuji Kaneda.
\newblock Subject independent facial expression recognition with robust face
  detection using a convolutional neural network.
\newblock {\em Neural Networks}, 16(5-6):555--559, 2003.

\bibitem{matsuzaki2018empirical}
Kiminori Matsuzaki.
\newblock Empirical analysis of puct algorithm with evaluation functions of
  different quality.
\newblock In {\em 2018 Conference on Technologies and Applications of
  Artificial Intelligence (TAAI)}, pages 142--147. IEEE, 2018.

\bibitem{mcallester1988conspiracy}
David~Allen McAllester.
\newblock Conspiracy numbers for min-max search.
\newblock {\em Artificial Intelligence}, 35(3):287--310, 1988.

\bibitem{mcclelland1995there}
James~L McClelland, Bruce~L McNaughton, and Randall~C O'Reilly.
\newblock Why there are complementary learning systems in the hippocampus and
  neocortex: insights from the successes and failures of connectionist models
  of learning and memory.
\newblock {\em Psychological Review}, 102(3):419, 1995.

\bibitem{mccorduck2009machines}
Pamela McCorduck.
\newblock {\em Machines who think: A personal inquiry into the history and
  prospects of artificial intelligence}.
\newblock AK Peters/CRC Press, 2009.

\bibitem{mcculloch1943logical}
Warren~S McCulloch and Walter Pitts.
\newblock A logical calculus of the ideas immanent in nervous activity.
\newblock {\em The Bulletin of Mathematical Biophysics}, 5(4):115--133, 1943.

\bibitem{michie1966game}
Donald Michie.
\newblock Game-playing and game-learning automata.
\newblock In {\em Advances in programming and non-numerical computation}, pages
  183--200. Elsevier, 1966.

\bibitem{michie1988machine}
Donald Michie.
\newblock Machine learning in the next five years.
\newblock In {\em Proceedings of the 3rd European Conference on European
  Working Session on Learning}, pages 107--122. Pitman Publishing, Inc., 1988.

\bibitem{michulke2009neural}
Daniel Michulke and Michael Thielscher.
\newblock Neural networks for state evaluation in general game playing.
\newblock In {\em Joint European Conference on Machine Learning and Knowledge
  Discovery in Databases}, pages 95--110. Springer, 2009.

\bibitem{miikkulainen2019evolving}
Risto Miikkulainen, Jason Liang, Elliot Meyerson, Aditya Rawal, Daniel Fink,
  Olivier Francon, Bala Raju, Hormoz Shahrzad, Arshak Navruzyan, Nigel Duffy,
  et~al.
\newblock Evolving deep neural networks.
\newblock In {\em Artificial Intelligence in the Age of Neural Networks and
  Brain Computing}, pages 293--312. Elsevier, 2019.

\bibitem{mikolov2012statistical}
Tom{\'a}{\v{s}} Mikolov.
\newblock Statistical language models based on neural networks.
\newblock {\em Presentation at Google, Mountain View, 2nd April}, 80, 2012.

\bibitem{millen1981programming}
Jonathan~K Millen.
\newblock Programming the game of {Go}.
\newblock {\em Byte Magazine}, 1981.

\bibitem{mirsoleimani2015ensemble}
S~Ali Mirsoleimani, Aske Plaat, Jaap van~den Herik, and Jos Vermaseren.
\newblock Ensemble {UCT} needs high exploitation.
\newblock In {\em ICAART2016, 8th International Conference on Agents and
  Artificial Intelligence}, 2016.

\bibitem{mirsoleimani2017structured}
S~Ali Mirsoleimani, Aske Plaat, Jaap van~den Herik, and Jos Vermaseren.
\newblock Structured parallel programming for {Monte Carlo} tree search.
\newblock Technical report, arXiv preprint arXiv:1704.00325, 2017.

\bibitem{mirsoleimani2017analysis}
S~Ali Mirsoleimani, Aske Plaat, H~Jaap van~den Herik, and Jos Vermaseren.
\newblock An analysis of virtual loss in parallel {MCTS}.
\newblock In {\em ICAART (2)}, pages 648--652, 2017.

\bibitem{mirsoleimani2015scaling}
S~Ali Mirsoleimani, Aske Plaat, Jaap Van Den~Herik, and Jos Vermaseren.
\newblock Scaling {Monte Carlo} tree search on {Intel} {Xeon} {Phi}.
\newblock In {\em Parallel and Distributed Systems (ICPADS), 2015 IEEE 21st
  International Conference on}, pages 666--673. IEEE, 2015.

\bibitem{mirsoleimani2018lock}
S~Ali Mirsoleimani, Jaap van~den Herik, Aske Plaat, and Jos Vermaseren.
\newblock A lock-free algorithm for parallel {MCTS}.
\newblock In {\em ICAART (2)}, pages 589--598, 2018.

\bibitem{mirsoleimani2018pipeline}
S~Ali Mirsoleimani, Jaap van~den Herik, Aske Plaat, and Jos Vermaseren.
\newblock Pipeline pattern for parallel {MCTS}.
\newblock In {\em ICAART (2)}, pages 614--621, 2018.

\bibitem{mitchell1980need}
Tom~M Mitchell.
\newblock {\em The need for biases in learning generalizations}.
\newblock Department of Computer Science, Carnegie Mellon University, 1980.

\bibitem{mitchell2006discipline}
Tom~M Mitchell.
\newblock {\em The discipline of machine learning}, volume~9.
\newblock Carnegie Mellon University, School of Computer Science, Machine
  Learning, 2006.

\bibitem{mnih2016asynchronous}
Volodymyr Mnih, Adria~Puigdomenech Badia, Mehdi Mirza, Alex Graves, Timothy
  Lillicrap, Tim Harley, David Silver, and Koray Kavukcuoglu.
\newblock Asynchronous methods for deep reinforcement learning.
\newblock In {\em International Conference on Machine Learning}, pages
  1928--1937, 2016.

\bibitem{mnih2014recurrent}
Volodymyr Mnih, Nicolas Heess, Alex Graves, et~al.
\newblock Recurrent models of visual attention.
\newblock In {\em Advances in Neural Information Processing Systems}, pages
  2204--2212, 2014.

\bibitem{mnih2013playing}
Volodymyr Mnih, Koray Kavukcuoglu, David Silver, Alex Graves, Ioannis
  Antonoglou, Daan Wierstra, and Martin Riedmiller.
\newblock Playing {A}tari with deep reinforcement learning.
\newblock {\em arXiv preprint arXiv:1312.5602}, 2013.

\bibitem{mnih2015human}
Volodymyr Mnih, Koray Kavukcuoglu, David Silver, Andrei~A Rusu, Joel Veness,
  Marc~G Bellemare, Alex Graves, Martin Riedmiller, Andreas~K Fidjeland, Georg
  Ostrovski, et~al.
\newblock Human-level control through deep reinforcement learning.
\newblock {\em Nature}, 518(7540):529, 2015.

\bibitem{moerland2017efficient}
Thomas~M Moerland, Joost Broekens, and Catholijn~M Jonker.
\newblock Efficient exploration with double uncertain value networks.
\newblock {\em arXiv preprint arXiv:1711.10789}, 2017.

\bibitem{moerland2018potential}
Thomas~M Moerland, Joost Broekens, and Catholijn~M Jonker.
\newblock The potential of the return distribution for exploration in {RL}.
\newblock {\em arXiv preprint arXiv:1806.04242}, 2018.

\bibitem{moerland2020model}
Thomas~M Moerland, Joost Broekens, and Catholijn~M Jonker.
\newblock Model-based reinforcement learning: A survey.
\newblock {\em arXiv preprint arXiv:2006.16712}, 2020.

\bibitem{moerland2018a0c}
Thomas~M Moerland, Joost Broekens, Aske Plaat, and Catholijn~M Jonker.
\newblock A0c: Alpha zero in continuous action space.
\newblock {\em arXiv preprint arXiv:1805.09613}, 2018.

\bibitem{moerland2018monte}
Thomas~M Moerland, Joost Broekens, Aske Plaat, and Catholijn~M Jonker.
\newblock {Monte Carlo} tree search for asymmetric trees.
\newblock {\em arXiv preprint arXiv:1805.09218}, 2018.

\bibitem{montavon2018methods}
Gr{\'e}goire Montavon, Wojciech Samek, and Klaus-Robert M{\"u}ller.
\newblock Methods for interpreting and understanding deep neural networks.
\newblock {\em Digital Signal Processing}, 73:1--15, 2018.

\bibitem{moor2006dartmouth}
James Moor.
\newblock The {Dartmouth College} artificial intelligence conference: The next
  fifty years.
\newblock {\em AI Magazine}, 27(4):87, 2006.

\bibitem{moravvcik2017deepstack}
Matej Morav{\v{c}}{\'\i}k, Martin Schmid, Neil Burch, Viliam Lis{\`y}, Dustin
  Morrill, Nolan Bard, Trevor Davis, Kevin Waugh, Michael Johanson, and Michael
  Bowling.
\newblock Deepstack: Expert-level artificial intelligence in heads-up no-limit
  poker.
\newblock {\em Science}, 356(6337):508--513, 2017.

\bibitem{muller2016introduction}
Andreas~C M{\"u}ller, Sarah Guido, et~al.
\newblock {\em Introduction to machine learning with Python: a guide for data
  scientists}.
\newblock {O'Reilly Media, Inc.}, 2016.

\bibitem{muller2002computer}
Martin M{\"u}ller.
\newblock Computer {Go}.
\newblock {\em Artificial Intelligence}, 134(1-2):145--179, 2002.

\bibitem{muller2009fuego}
Martin M{\"u}ller.
\newblock Fuego at the {Computer} {Olympiad} in {Pamplona} 2009: A tournament
  report.
\newblock {\em University of Alberta, TR}, pages 09--09, 2009.

\bibitem{munos2016safe}
R{\'e}mi Munos, Tom Stepleton, Anna Harutyunyan, and Marc Bellemare.
\newblock Safe and efficient off-policy reinforcement learning.
\newblock In {\em Advances in Neural Information Processing Systems}, pages
  1054--1062, 2016.

\bibitem{nachum2017bridging}
Ofir Nachum, Mohammad Norouzi, Kelvin Xu, and Dale Schuurmans.
\newblock Bridging the gap between value and policy based reinforcement
  learning.
\newblock In {\em Advances in Neural Information Processing Systems}, pages
  2775--2785, 2017.

\bibitem{nagarajan2018deterministic}
Prabhat Nagarajan, Garrett Warnell, and Peter Stone.
\newblock Deterministic implementations for reproducibility in deep
  reinforcement learning.
\newblock {\em arXiv preprint arXiv:1809.05676}, 2018.

\bibitem{narvekar2020curriculum}
Sanmit Narvekar, Bei Peng, Matteo Leonetti, Jivko Sinapov, Matthew~E Taylor,
  and Peter Stone.
\newblock Curriculum learning for reinforcement learning domains: A framework
  and survey.
\newblock {\em arXiv preprint arXiv:2003.04960}, 2020.

\bibitem{nau1986experiments}
Dana Nau, Paul Purdom, and Chun-Hung Tzeng.
\newblock Experiments on alternatives to minimax.
\newblock {\em International Journal of Parallel Programming}, 15(2):163--183,
  1986.

\bibitem{nau1980pathology}
Dana~S Nau.
\newblock Pathology on game trees: A summary of results.
\newblock In {\em AAAI}, pages 102--104, 1980.

\bibitem{neapolitan2004learning}
Richard~E Neapolitan et~al.
\newblock {\em Learning {Bayesian} networks}, volume~38.
\newblock Pearson Prentice Hall, Upper Saddle River, NJ, 2004.

\bibitem{needham1965science}
Joseph Needham.
\newblock Science and civilisation in china, vol. 4: Physics and physical
  technology. part iii.
\newblock {\em Civil Engineering and Nautics (d) Building Technology, Cambridge
  University Press}, pages 60--66, 1965.

\bibitem{neisser1996intelligence}
Ulric Neisser, Gwyneth Boodoo, Thomas~J Bouchard~Jr, A~Wade Boykin, Nathan
  Brody, Stephen~J Ceci, Diane~F Halpern, John~C Loehlin, Robert Perloff,
  Robert~J Sternberg, et~al.
\newblock Intelligence: Knowns and unknowns.
\newblock {\em American Psychologist}, 51(2):77, 1996.

\bibitem{newborn2013deep}
Monty Newborn.
\newblock {\em Deep {Blue}: an artificial intelligence milestone}.
\newblock Springer Science \& Business Media, 2013.

\bibitem{newell1958elements}
Allen Newell, John~Calman Shaw, and Herbert~A Simon.
\newblock Elements of a theory of human problem solving.
\newblock {\em Psychological Review}, 65(3):151, 1958.

\bibitem{newellcomputer}
Allen Newell and Herbert Simon.
\newblock Computer science as empirical enquiry: Symbols and search.
\newblock {\em CACM}, 1976.

\bibitem{nezhinsky2010pattern}
Alexander~E Nezhinsky and Fons~J Verbeek.
\newblock Pattern recognition for high throughput zebrafish imaging using
  genetic algorithm optimization.
\newblock In {\em IAPR International Conference on Pattern Recognition in
  Bioinformatics}, pages 301--312. Springer, 2010.

\bibitem{ng2004feature}
Andrew~Y Ng.
\newblock Feature selection, {L 1 vs. L 2} regularization, and rotational
  invariance.
\newblock In {\em Proceedings of the Twenty-first International Conference on
  Machine Learning}, page~78. ACM, 2004.

\bibitem{nilsson2009quest}
Nils~J Nilsson.
\newblock {\em The quest for artificial intelligence}.
\newblock Cambridge University Press, 2009.

\bibitem{nitsche1982learning}
T~Nitsche.
\newblock A learning chess program.
\newblock In {\em Advances in Computer Chess}, pages 113--120. Elsevier, 1982.

\bibitem{nussinov2019computational}
Ruth Nussinov, Chung-Jung Tsai, Amarda Shehu, and Hyunbum Jang.
\newblock Computational structural biology: Successes, future directions, and
  challenges.
\newblock {\em Molecules}, 24(3):637, 2019.

\bibitem{o2016combining}
Brendan O'Donoghue, Remi Munos, Koray Kavukcuoglu, and Volodymyr Mnih.
\newblock Combining policy gradient and {Q}-learning.
\newblock {\em arXiv preprint arXiv:1611.01626}, 2016.

\bibitem{oh2020discovering}
Junhyuk Oh, Matteo Hessel, Wojciech~M Czarnecki, Zhongwen Xu, Hado van Hasselt,
  Satinder Singh, and David Silver.
\newblock Discovering reinforcement learning algorithms.
\newblock {\em arXiv preprint arXiv:2007.08794}, 2020.

\bibitem{oh2004gpu}
Kyoung-Su Oh and Keechul Jung.
\newblock {GPU} implementation of neural networks.
\newblock {\em Pattern Recognition}, 37(6):1311--1314, 2004.

\bibitem{colah2015}
Chris Olah.
\newblock Understanding {LSTM} networks.
  \url{http://colah.github.io/posts/2015-08-Understanding-LSTMs/}, 2015.

\bibitem{o2010play}
Joseph O'Neill, Barty Pleydell-Bouverie, David Dupret, and Jozsef Csicsvari.
\newblock Play it again: reactivation of waking experience and memory.
\newblock {\em Trends in Neurosciences}, 33(5):220--229, 2010.

\bibitem{ontanon2013survey}
Santiago Ontan{\'o}n, Gabriel Synnaeve, Alberto Uriarte, Florian Richoux, David
  Churchill, and Mike Preuss.
\newblock A survey of real-time strategy game {AI} research and competition in
  {StarCraft}.
\newblock {\em IEEE Transactions on Computational Intelligence and AI in
  Games}, 5(4):293--311, 2013.

\bibitem{OpenAI_dota}
OpenAI.
\newblock {OpenAI} five.
\newblock \url{https://blog.openai.com/openai-five/}, 2018.

\bibitem{osband2019behaviour}
Ian Osband, Yotam Doron, Matteo Hessel, John Aslanides, Eren Sezener, Andre
  Saraiva, Katrina McKinney, Tor Lattimore, Csaba Szepezvari, Satinder Singh,
  et~al.
\newblock Behaviour suite for reinforcement learning.
\newblock {\em arXiv preprint arXiv:1908.03568}, 2019.

\bibitem{painter2012greedy}
Christopher Painter-Wakefield and Ronald Parr.
\newblock Greedy algorithms for sparse reinforcement learning.
\newblock {\em arXiv preprint arXiv:1206.6485}, 2012.

\bibitem{pan2010survey}
Sinno~Jialin Pan, Qiang Yang, et~al.
\newblock A survey on transfer learning.
\newblock {\em IEEE Transactions on Knowledge and Data Engineering},
  22(10):1345--1359, 2010.

\bibitem{panait2005cooperative}
Liviu Panait and Sean Luke.
\newblock Cooperative multi-agent learning: The state of the art.
\newblock {\em Autonomous Agents and Multi-Agent Systems}, 11(3):387--434,
  2005.

\bibitem{pang2019reinforcement}
Zhen-Jia Pang, Ruo-Ze Liu, Zhou-Yu Meng, Yi~Zhang, Yang Yu, and Tong Lu.
\newblock On reinforcement learning for full-length game of {StarCraft}.
\newblock In {\em Proceedings of the AAAI Conference on Artificial
  Intelligence}, volume~33, pages 4691--4698, 2019.

\bibitem{paparo2014quantum}
Giuseppe~Davide Paparo, Vedran Dunjko, Adi Makmal, Miguel~Angel Martin-Delgado,
  and Hans~J Briegel.
\newblock Quantum speedup for active learning agents.
\newblock {\em Physical Review X}, 4(3):031002, 2014.

\bibitem{pappa2014contrasting}
Gisele~L Pappa, Gabriela Ochoa, Matthew~R Hyde, Alex~A Freitas, John Woodward,
  and Jerry Swan.
\newblock Contrasting meta-learning and hyper-heuristic research: the role of
  evolutionary algorithms.
\newblock {\em Genetic Programming and Evolvable Machines}, 15(1):3--35, 2014.

\bibitem{paquette2019no}
Philip Paquette, Yuchen Lu, Steven Bocco, Max Smith, O-G Satya, Jonathan~K
  Kummerfeld, Joelle Pineau, Satinder Singh, and Aaron~C Courville.
\newblock No-press diplomacy: Modeling multi-agent gameplay.
\newblock In {\em Advances in Neural Information Processing Systems}, pages
  4476--4487, 2019.

\bibitem{pascanu2017learning}
Razvan Pascanu, Yujia Li, Oriol Vinyals, Nicolas Heess, Lars Buesing, Sebastien
  Racani{\`e}re, David Reichert, Th{\'e}ophane Weber, Daan Wierstra, and Peter
  Battaglia.
\newblock Learning model-based planning from scratch.
\newblock {\em arXiv preprint arXiv:1707.06170}, 2017.

\bibitem{pascutto2017leela}
Gian-Carlo Pascutto.
\newblock Leela zero. \url{https://github.com/leela-zero/leela-zero}, 2017.

\bibitem{paszke2019pytorch}
Adam Paszke, Sam Gross, Francisco Massa, Adam Lerer, James Bradbury, Gregory
  Chanan, Trevor Killeen, Zeming Lin, Natalia Gimelshein, Luca Antiga, et~al.
\newblock Pytorch: An imperative style, high-performance deep learning library.
\newblock In {\em Advances in Neural Information Processing Systems}, pages
  8024--8035, 2019.

\bibitem{pearl1980scout}
Judea Pearl.
\newblock Scout: A simple game-searching algorithm with proven optimal
  properties.
\newblock In {\em AAAI}, pages 143--145, 1980.

\bibitem{pearl1983nature}
Judea Pearl.
\newblock On the nature of pathology in game searching.
\newblock {\em Artificial Intelligence}, 20(4):427--453, 1983.

\bibitem{pearl1984heuristics}
Judea Pearl.
\newblock {\em Heuristics: Intelligent Search Strategies for Computer Problem
  Solving}.
\newblock Addison-Wesley, Reading, MA, 1984.

\bibitem{pearl2018book}
Judea Pearl and Dana Mackenzie.
\newblock {\em The book of why: the new science of cause and effect}.
\newblock Basic Books, 2018.

\bibitem{pedregosa2011scikit}
Fabian Pedregosa, Ga{\"e}l Varoquaux, Alexandre Gramfort, Vincent Michel,
  Bertrand Thirion, Olivier Grisel, Mathieu Blondel, Peter Prettenhofer, Ron
  Weiss, Vincent Dubourg, et~al.
\newblock Scikit-learn: Machine learning in {Python}.
\newblock {\em Journal of Machine Learning Research}, 12(Oct):2825--2830, 2011.

\bibitem{pell1992metagame}
Barney Pell.
\newblock Metagame: A new challenge for games and learning.
\newblock In {\em Heuristic Programming in Artificial Intelligence 3 -- The
  Third Computer Olympiad}. Ellis Horwood, 1992.

\bibitem{pell1993logic}
Barney Pell.
\newblock Logic programming for general game-playing.
\newblock Technical report, University of Cambridge, Computer Laboratory, 1993.

\bibitem{pell1996strategic}
Barney Pell.
\newblock A strategic metagame player for general chess-like games.
\newblock {\em Computational Intelligence}, 12(1):177--198, 1996.

\bibitem{perez2015open}
Diego Perez~Liebana, Jens Dieskau, Martin Hunermund, Sanaz Mostaghim, and Simon
  Lucas.
\newblock Open loop search for general video game playing.
\newblock In {\em Proceedings of the 2015 Annual Conference on Genetic and
  Evolutionary Computation}, pages 337--344. ACM, 2015.

\bibitem{perez20162014}
Diego Perez-Liebana, Spyridon Samothrakis, Julian Togelius, Tom Schaul, Simon~M
  Lucas, Adrien Cou{\"e}toux, Jerry Lee, Chong-U Lim, and Tommy Thompson.
\newblock The 2014 general video game playing competition.
\newblock {\em IEEE Transactions on Computational Intelligence and AI in
  Games}, 8(3):229--243, 2016.

\bibitem{perkins1999using}
Theodore~J Perkins, Doina Precup, et~al.
\newblock Using options for knowledge transfer in reinforcement learning.
\newblock {\em University of Massachusetts, Amherst, MA, USA, Tech. Rep}, 1999.

\bibitem{pezzotti2017deepeyes}
Nicola Pezzotti, Thomas H{\"o}llt, Jan Van~Gemert, Boudewijn~PF Lelieveldt,
  Elmar Eisemann, and Anna Vilanova.
\newblock Deepeyes: Progressive visual analytics for designing deep neural
  networks.
\newblock {\em IEEE Transactions on Visualization and Computer Graphics},
  24(1):98--108, 2017.

\bibitem{pham2018efficient}
Hieu Pham, Melody~Y Guan, Barret Zoph, Quoc~V Le, and Jeff Dean.
\newblock Efficient neural architecture search via parameter sharing.
\newblock {\em arXiv preprint arXiv:1802.03268}, 2018.

\bibitem{pijls1990another}
Wim Pijls and Arie de~Bruin.
\newblock Another view on the {SSS*} algorithm.
\newblock In {\em Algorithms}, pages 211--220. Springer, 1990.

\bibitem{plaat1996research}
Aske Plaat.
\newblock {\em Research re: search and re-search}.
\newblock PhD thesis, Erasmus University Rotterdam, 1996.

\bibitem{plaat2020model}
Aske Plaat, Walter Kosters, and Mike Preuss.
\newblock Model-based deep reinforcement learning for high-dimensional
  problems: A survey.
\newblock Technical report, preprint arXiv:2008.05598, 2020.

\bibitem{plaat1994new}
Aske Plaat, Jonathan Schaeffer, Wim Pijls, and Arie De~Bruin.
\newblock A new paradigm for minimax search.
\newblock Technical Report arXiv 1404.1515, University of Alberta, 1994.

\bibitem{plaat1996best}
Aske Plaat, Jonathan Schaeffer, Wim Pijls, and Arie De~Bruin.
\newblock Best-first fixed-depth minimax algorithms.
\newblock {\em Artificial Intelligence}, 87(1-2):255--293, 1996.

\bibitem{plaat1996exploiting}
Aske Plaat, Jonathan Schaeffer, Wim Pijls, and Arie De~Bruin.
\newblock Exploiting graph properties of game trees.
\newblock In {\em AAAI/IAAI, Vol. 1}, pages 234--239, 1996.

\bibitem{plappert2016kerasrl}
Matthias Plappert.
\newblock {Keras-RL}.
\newblock \url{https://github.com/keras-rl/keras-rl}, 2016.

\bibitem{plucker1999psychometric}
Jonathan~A Plucker and Joseph~S Renzulli.
\newblock Psychometric approaches to the study of human creativity.
\newblock {\em Handbook of creativity}, pages 35--61, 1999.

\bibitem{pollack1997did}
Jordan~B Pollack and Alan~D Blair.
\newblock Why did {TD-gammon} work?
\newblock In {\em Advances in Neural Information Processing Systems}, pages
  10--16, 1997.

\bibitem{prasaf2018lessons}
Aditya Prasad.
\newblock Lessons from implementing alphazero
  \url{https://medium.com/oracledevs/lessons-from-implementing-alphazero-7e36e9054191},
  2018.

\bibitem{pratt1993discriminability}
Lorien~Y Pratt.
\newblock Discriminability-based transfer between neural networks.
\newblock In {\em Advances in Neural Information Processing Systems}, pages
  204--211, 1993.

\bibitem{prechelt1998automatic}
Lutz Prechelt.
\newblock Automatic early stopping using cross validation: quantifying the
  criteria.
\newblock {\em Neural Networks}, 11(4):761--767, 1998.

\bibitem{prechelt1998early}
Lutz Prechelt.
\newblock Early stopping-but when?
\newblock In {\em Neural Networks: Tricks of the trade}, pages 55--69.
  Springer, 1998.

\bibitem{preuss2010towards}
Mike Preuss, Nicola Beume, Holger Danielsiek, Tobias Hein, Boris Naujoks, Nico
  Piatkowski, Raphael St{\"u}er, Andreas Thom, and Simon Wessing.
\newblock Towards intelligent team composition and maneuvering in real-time
  strategy games.
\newblock {\em IEEE Transactions on Computational Intelligence and AI in
  Games}, 2(2):82--98, 2010.

\bibitem{puigdomenech2020agent57}
Adri{\`a} Puigdom{\`e}nech~Badia, Bilal Piot, Steven Kapturowski, Pablo
  Sprechmann, Alex Vitvitskyi, Daniel Guo, and Charles Blundell.
\newblock Agent57: Outperforming the {Atari} human benchmark.
\newblock {\em arXiv}, pages arXiv--2003, 2020.

\bibitem{pumperla2019}
Max Pumperla and Kevin Ferguson.
\newblock {\em Deep Learning and the Game of {Go}}.
\newblock Manning, 2019.

\bibitem{qin2014sparse}
Zhiwei Qin, Weichang Li, and Firdaus Janoos.
\newblock Sparse reinforcement learning via convex optimization.
\newblock In {\em International Conference on Machine Learning}, pages
  424--432, 2014.

\bibitem{quinlan1983learning}
J~Ross Quinlan.
\newblock Learning efficient classification procedures and their application to
  chess end games.
\newblock In {\em Machine Learning}, pages 463--482. Springer, 1983.

\bibitem{quinlan1986induction}
J.~Ross Quinlan.
\newblock Induction of decision trees.
\newblock {\em Machine Learning}, 1(1):81--106, 1986.

\bibitem{rafati2018learning}
Jacob Rafati and David~C Noelle.
\newblock Learning representations in model-free hierarchical reinforcement
  learning.
\newblock {\em arXiv preprint arXiv:1810.10096}, 2018.

\bibitem{raina2007self}
Rajat Raina, Alexis Battle, Honglak Lee, Benjamin Packer, and Andrew~Y Ng.
\newblock Self-taught learning: transfer learning from unlabeled data.
\newblock In {\em Proceedings of the 24th International Conference on Machine
  learning}, pages 759--766. ACM, 2007.

\bibitem{rakelly2019efficient}
Kate Rakelly, Aurick Zhou, Chelsea Finn, Sergey Levine, and Deirdre Quillen.
\newblock Efficient off-policy meta-reinforcement learning via probabilistic
  context variables.
\newblock In {\em International conference on machine learning}, pages
  5331--5340. PMLR, 2019.

\bibitem{ramanujan2010adversarial}
Raghuram Ramanujan, Ashish Sabharwal, and Bart Selman.
\newblock On adversarial search spaces and sampling-based planning.
\newblock In {\em ICAPS}, volume~10, pages 242--245, 2010.

\bibitem{nevergrad}
J.~Rapin and O.~Teytaud.
\newblock {Nevergrad - A gradient-free optimization platform}.
\newblock \url{https://GitHub.com/FacebookResearch/Nevergrad}, 2018.

\bibitem{ravi2016optimization}
Sachin Ravi and Hugo Larochelle.
\newblock Optimization as a model for few-shot learning.
\newblock In {\em International Conference on Learning Representations (ICLR)},
  2017.

\bibitem{reibman1983non}
Andrew~L Reibman and Bruce~W Ballard.
\newblock Non-minimax search strategies for use against fallible opponents.
\newblock In {\em AAAI}, pages 338--342, 1983.

\bibitem{reinefeld1989spielbaum}
Alexander Reinefeld.
\newblock {\em Spielbaum-Suchverfahren}, volume 200.
\newblock Springer-Verlag, 1989.

\bibitem{rezende2016one}
Danilo~Jimenez Rezende, Shakir Mohamed, Ivo Danihelka, Karol Gregor, and Daan
  Wierstra.
\newblock One-shot generalization in deep generative models.
\newblock {\em preprint arXiv:1603.05106}, 2016.

\bibitem{ribeiro2009darkblade}
Joao Ribeiro, Pedro Mariano, and Lu{\'\i}s~Seabra Lopes.
\newblock Darkblade: A program that plays {Diplomacy}.
\newblock In {\em Portuguese Conference on Artificial Intelligence}, pages
  485--496. Springer, 2009.

\bibitem{rice1976algorithm}
John~R. Rice.
\newblock The algorithm selection problem.
\newblock {\em Advances in Computers}, 15(65-118):5, 1976.

\bibitem{riedmiller2005neural}
Martin Riedmiller.
\newblock Neural fitted q iteration--first experiences with a data efficient
  neural reinforcement learning method.
\newblock In {\em European Conference on Machine Learning}, pages 317--328.
  Springer, 2005.

\bibitem{rimmel2010current}
Arpad Rimmel, Olivier Teytaud, Chang-Shing Lee, Shi-Jim Yen, Mei-Hui Wang, and
  Shang-Rong Tsai.
\newblock Current frontiers in computer {Go}.
\newblock {\em IEEE Transactions on Computational Intelligence and AI in
  Games}, 2(4):229--238, 2010.

\bibitem{risi2020chess}
Sebastian Risi and Mike Preuss.
\newblock {From Chess and Atari to StarCraft and Beyond: How Game AI is Driving
  the World of AI}.
\newblock {\em KI-K{\"u}nstliche Intelligenz}, pages 1--11, 2020.

\bibitem{rivest1987game}
Ronald~L Rivest.
\newblock Game tree searching by min/max approximation.
\newblock {\em Artificial Intelligence}, 34(1):77--96, 1987.

\bibitem{rocki2011large}
Kamil Rocki and Reiji Suda.
\newblock Large-scale parallel {Monte Carlo Tree Search on GPU}.
\newblock In {\em 2011 IEEE International Symposium on Parallel and Distributed
  Processing Workshops and Phd Forum}, pages 2034--2037. IEEE, 2011.

\bibitem{romein2003solving}
John~W Romein and Henri~E Bal.
\newblock Solving awari with parallel retrograde analysis.
\newblock {\em Computer}, 36(10):26--33, 2003.

\bibitem{romera2015embarrassingly}
Bernardino Romera-Paredes and Philip Torr.
\newblock An embarrassingly simple approach to zero-shot learning.
\newblock In {\em International Conference on Machine Learning}, pages
  2152--2161, 2015.

\bibitem{stockfish}
Tord Romstad, Marco Costalba, and Joona~Kiiski.
  \url{https://stockfishchess.org} Source code~at
  \url{https://github.com/official-stockfish/Stockfish}.
\newblock Stockfish open source chess program.

\bibitem{rosenblatt1958perceptron}
Frank Rosenblatt.
\newblock The perceptron: a probabilistic model for information storage and
  organization in the brain.
\newblock {\em Psychological Review}, 65(6):386, 1958.

\bibitem{rosin2011multi}
Christopher~D Rosin.
\newblock Multi-armed bandits with episode context.
\newblock {\em Annals of Mathematics and Artificial Intelligence},
  61(3):203--230, 2011.

\bibitem{rossum1995python}
Guido Rossum.
\newblock Python reference manual.
\newblock 1995.

\bibitem{rost1993prediction}
Burkhard Rost and Chris Sander.
\newblock Prediction of protein secondary structure at better than 70\%
  accuracy.
\newblock {\em Journal of Molecular Biology}, 232(2):584--599, 1993.

\bibitem{roweis2000nonlinear}
Sam~T Roweis and Lawrence~K Saul.
\newblock Nonlinear dimensionality reduction by locally linear embedding.
\newblock {\em Science}, 290(5500):2323--2326, 2000.

\bibitem{rubens2015active}
Neil Rubens, Mehdi Elahi, Masashi Sugiyama, and Dain Kaplan.
\newblock Active learning in recommender systems.
\newblock In {\em Recommender systems handbook}, pages 809--846. Springer,
  2015.

\bibitem{rubin2011computer}
Jonathan Rubin and Ian Watson.
\newblock Computer poker: A review.
\newblock {\em Artificial intelligence}, 175(5-6):958--987, 2011.

\bibitem{Ruijl_2018}
Ben Ruijl.
\newblock Games and loop integrals.
\newblock {\em Journal of Physics: Conference Series}, 1085:022007, Sep 2018.

\bibitem{ruijl2013combining}
Ben Ruijl, Jos Vermaseren, Aske Plaat, and Jaap van~den Herik.
\newblock Combining simulated annealing and {Monte Carlo} tree search for
  expression simplification.
\newblock {\em arXiv preprint arXiv:1312.0841}, 2013.

\bibitem{ruijl2014hepgame}
Ben Ruijl, Jos Vermaseren, Aske Plaat, and Jaap van~den Herik.
\newblock Hepgame and the simplification of expressions.
\newblock {\em arXiv preprint arXiv:1405.6369}, 2014.

\bibitem{rumelhart1987parallel}
David~E Rumelhart, James~L McClelland, PDP~Research Group, et~al.
\newblock {\em Parallel distributed processing}, volume~1.
\newblock MIT Press, Cambridge, MA, 1987.

\bibitem{russakovsky2015imagenet}
Olga Russakovsky, Jia Deng, Hao Su, Jonathan Krause, Sanjeev Satheesh, Sean Ma,
  Zhiheng Huang, Andrej Karpathy, Aditya Khosla, Michael Bernstein, et~al.
\newblock Imagenet large scale visual recognition challenge.
\newblock {\em International journal of computer vision}, 115(3):211--252,
  2015.

\bibitem{russell2016artificial}
Stuart~J Russell and Peter Norvig.
\newblock {\em Artificial intelligence: a modern approach}.
\newblock Pearson Education Limited, Malaysia, 2016.

\bibitem{sadikov2005bias}
Aleksander Sadikov, Ivan Bratko, and Igor Kononenko.
\newblock Bias and pathology in minimax search.
\newblock {\em Theoretical Computer Science}, 349(2):268--281, 2005.

\bibitem{saha2018}
Sumit Saha.
\newblock A comprehensive guide to convolutional neural networks --- the eli5
  way.
\newblock {\em Towards Data Science},
  (\url{https://towardsdatascience.com/a-comprehensive-guide-to-convolutional-neural-networks-the-eli5-way-3bd2b1164a53}),
  2018.

\bibitem{salimans2017evolution}
Tim Salimans, Jonathan Ho, Xi~Chen, Szymon Sidor, and Ilya Sutskever.
\newblock Evolution strategies as a scalable alternative to reinforcement
  learning.
\newblock {\em preprint arXiv:1703.03864}, 2017.

\bibitem{sallans2004reinforcement}
Brian Sallans and Geoffrey~E Hinton.
\newblock Reinforcement learning with factored states and actions.
\newblock {\em Journal of Machine Learning Research}, 5(Aug):1063--1088, 2004.

\bibitem{samuel1959some}
Arthur~L Samuel.
\newblock Some studies in machine learning using the game of checkers.
\newblock {\em IBM Journal of Research and Development}, 3(3):210--229, 1959.

\bibitem{sanders2010cuda}
Jason Sanders and Edward Kandrot.
\newblock {\em CUDA by example: an introduction to general-purpose GPU
  programming}.
\newblock Addison-Wesley Professional, 2010.

\bibitem{sandholm2010state}
Tuomas Sandholm.
\newblock The state of solving large incomplete-information games, and
  application to poker.
\newblock {\em AI Magazine}, 31(4):13--32, 2010.

\bibitem{santamaria1997experiments}
Juan~C Santamar{\'\i}a, Richard~S Sutton, and Ashwin Ram.
\newblock Experiments with reinforcement learning in problems with continuous
  state and action spaces.
\newblock {\em Adaptive Behavior}, 6(2):163--217, 1997.

\bibitem{santoro2016meta}
Adam Santoro, Sergey Bartunov, Matthew Botvinick, Daan Wierstra, and Timothy
  Lillicrap.
\newblock Meta-learning with memory-augmented neural networks.
\newblock In {\em International Conference on Machine Learning}, pages
  1842--1850, 2016.

\bibitem{googlecloud2018}
Kaz Sato, Cliff Young, and David Patterson.
\newblock An in-depth look at {Google's first Tensor Processing Unit (TPU)}
  \url{https://cloud.google.com/blog/products/gcp/an-in-depth-look-at-googles-first-tensor-processing-unit-tpu},
  2017.

\bibitem{schadd2009quiescence}
Maarten Schadd and Mark Winands.
\newblock Quiescence search for stratego.
\newblock In {\em Proceedings of the 21st Benelux Conference on Artificial
  Intelligence. Eindhoven, the Netherlands}, 2009.

\bibitem{schadd2008single}
Maarten~PD Schadd, Mark~HM Winands, H~Jaap Van Den~Herik, Guillaume MJ-B
  Chaslot, and Jos~WHM Uiterwijk.
\newblock Single-player {Monte-Carlo} tree search.
\newblock In {\em International Conference on Computers and Games}, pages
  1--12. Springer, 2008.

\bibitem{schaefer2002}
Steve Schaefer.
\newblock Mathematical recreations.
  \url{http://www.mathrec.org/old/2002jan/solutions.html}, 2002.

\bibitem{schaeffer1986experiments}
Jonathan Schaeffer.
\newblock {\em Experiments in search and knowledge}.
\newblock PhD thesis, Department of Computing Science, University of Alberta,
  1986.

\bibitem{schaeffer1989history}
Jonathan Schaeffer.
\newblock The history heuristic and alpha-beta search enhancements in practice.
\newblock {\em IEEE Trans. on Pattern Analysis and Machine Intelligence},
  11(11):1203--1212, 1989.

\bibitem{schaeffer2008one}
Jonathan Schaeffer.
\newblock {\em One jump ahead: computer perfection at checkers}.
\newblock Springer Science \& Business Media, 2008.

\bibitem{schaeffer2007checkers}
Jonathan Schaeffer, Neil Burch, Yngvi Bj{\"o}rnsson, Akihiro Kishimoto, Martin
  M{\"u}ller, Robert Lake, Paul Lu, and Steve Sutphen.
\newblock Checkers is solved.
\newblock {\em Science}, 317(5844):1518--1522, 2007.

\bibitem{schaeffer1992world}
Jonathan Schaeffer, Joseph Culberson, Norman Treloar, Brent Knight, Paul Lu,
  and Duane Szafron.
\newblock A world championship caliber checkers program.
\newblock {\em Artificial Intelligence}, 53(2-3):273--289, 1992.

\bibitem{schaeffer1996chinook}
Jonathan Schaeffer, Robert Lake, Paul Lu, and Martin Bryant.
\newblock Chinook, the world man-machine checkers champion.
\newblock {\em AI Magazine}, 17(1):21, 1996.

\bibitem{schaeffer1996new}
Jonathan Schaeffer and Aske Plaat.
\newblock New advances in alpha-beta searching.
\newblock In {\em Proceedings of the 1996 ACM 24th Annual Conference on
  Computer science}, pages 124--130. ACM, 1996.

\bibitem{schaeffer1997kasparov}
Jonathan Schaeffer and Aske Plaat.
\newblock Kasparov versus {Deep Blue}: The rematch.
\newblock {\em ICGA Journal}, 20(2):95--101, 1997.

\bibitem{schaeffer2001unifying}
Jonathan Schaeffer, Aske Plaat, and Andreas Junghanns.
\newblock Unifying single-agent and two-player search.
\newblock {\em Information Sciences}, 135(3-4):151--175, 2001.

\bibitem{schaul2015universal}
Tom Schaul, Daniel Horgan, Karol Gregor, and David Silver.
\newblock Universal value function approximators.
\newblock In {\em International Conf. on Machine Learning}, pages 1312--1320,
  2015.

\bibitem{schaul2015prioritized}
Tom Schaul, John Quan, Ioannis Antonoglou, and David Silver.
\newblock Prioritized experience replay.
\newblock {\em arXiv preprint arXiv:1511.05952}, 2015.

\bibitem{schaul2010metalearning}
Tom Schaul and J{\"u}rgen Schmidhuber.
\newblock Metalearning.
\newblock {\em Scholarpedia}, 5(6):4650, 2010.

\bibitem{schiffel2007fluxplayer}
Stephan Schiffel and Michael Thielscher.
\newblock Fluxplayer: A successful general game player.
\newblock In {\em AAAI}, volume~7, pages 1191--1196, 2007.

\bibitem{schmidhuber1992learning}
J{\"u}rgen Schmidhuber.
\newblock Learning complex, extended sequences using the principle of history
  compression.
\newblock {\em Neural Computation}, 4(2):234--242, 1992.

\bibitem{schmidhuber2006developmental}
J{\"u}rgen Schmidhuber.
\newblock Developmental robotics, optimal artificial curiosity, creativity,
  music, and the fine arts.
\newblock {\em Connection Science}, 18(2):173--187, 2006.

\bibitem{schmidhuber2010formal}
J{\"u}rgen Schmidhuber.
\newblock Formal theory of creativity, fun, and intrinsic motivation
  (1990--2010).
\newblock {\em IEEE Transactions on Autonomous Mental Development},
  2(3):230--247, 2010.

\bibitem{schmidhuber2015deep}
J{\"u}rgen Schmidhuber.
\newblock Deep learning in neural networks: An overview.
\newblock {\em Neural Networks}, 61:85--117, 2015.

\bibitem{schmidhuber2002learning}
J{\"u}rgen Schmidhuber, F~Gers, and Douglas Eck.
\newblock Learning nonregular languages: A comparison of simple recurrent
  networks and {LSTM}.
\newblock {\em Neural Computation}, 14(9):2039--2041, 2002.

\bibitem{schofield2012hyperplay}
Michael~John Schofield, Timothy~Joseph Cerexhe, and Michael Thielscher.
\newblock Hyperplay: A solution to general game playing with imperfect
  information.
\newblock In {\em AAAI}, 2012.

\bibitem{schofield2015lifting}
Michael~John Schofield and Michael Thielscher.
\newblock Lifting model sampling for general game playing to
  incomplete-information models.
\newblock In {\em AAAI}, volume~15, pages 3585--3591, 2015.

\bibitem{schraudolph1994temporal}
Nicol~N Schraudolph, Peter Dayan, and Terrence~J Sejnowski.
\newblock Temporal difference learning of position evaluation in the game of
  {Go}.
\newblock In {\em Advances in Neural Information Processing Systems}, pages
  817--824, 1994.

\bibitem{schrittwieser2019mastering}
Julian Schrittwieser, Ioannis Antonoglou, Thomas Hubert, Karen Simonyan,
  Laurent Sifre, Simon Schmitt, Arthur Guez, Edward Lockhart, Demis Hassabis,
  Thore Graepel, et~al.
\newblock Mastering {Atari, Go,} chess and shogi by planning with a learned
  model.
\newblock {\em arXiv preprint arXiv:1911.08265}, 2019.

\bibitem{schrufer1986presence}
G{\"u}nther Schr{\"u}fer.
\newblock Presence and absence of pathology on game trees.
\newblock In {\em Advances in Computer Chess}, pages 101--112. Pergamon Press,
  Inc., 1986.

\bibitem{schulman2015trust}
John Schulman, Sergey Levine, Pieter Abbeel, Michael Jordan, and Philipp
  Moritz.
\newblock Trust region policy optimization.
\newblock In {\em International Conference on Machine Learning}, pages
  1889--1897, 2015.

\bibitem{schulman2017proximal}
John Schulman, Filip Wolski, Prafulla Dhariwal, Alec Radford, and Oleg Klimov.
\newblock Proximal policy optimization algorithms.
\newblock {\em arXiv preprint arXiv:1707.06347}, 2017.

\bibitem{schweighofer2003meta}
Nicolas Schweighofer and Kenji Doya.
\newblock Meta-learning in reinforcement learning.
\newblock {\em Neural Networks}, 16(1):5--9, 2003.

\bibitem{scutari2009learning}
Marco Scutari et~al.
\newblock Learning {Bayesian} networks with the bnlearn {R} package.
\newblock {\em Journal of Statistical Software}, 35(i03), 2010.

\bibitem{searle2008mind}
John~R Searle.
\newblock {\em Mind, language and society: Philosophy in the real world}.
\newblock Basic Books, 2008.

\bibitem{segal2010scalability}
Richard~B Segal.
\newblock On the scalability of parallel {UCT}.
\newblock In {\em International Conference on Computers and Games}, pages
  36--47. Springer, 2010.

\bibitem{segler2018planning}
Marwin~HS Segler, Mike Preuss, and Mark~P Waller.
\newblock Planning chemical syntheses with deep neural networks and symbolic
  {AI}.
\newblock {\em Nature}, 555(7698):604, 2018.

\bibitem{settles2009active}
Burr Settles.
\newblock Active learning literature survey.
\newblock Technical report, University of Wisconsin-Madison Department of
  Computer Sciences, 2009.

\bibitem{shahriari2016taking}
Bobak Shahriari, Kevin Swersky, Ziyu Wang, Ryan~P Adams, and Nando De~Freitas.
\newblock Taking the human out of the loop: A review of {Bayesian}
  optimization.
\newblock {\em Proceedings of the IEEE}, 104(1):148--175, 2016.

\bibitem{shannon1988programming}
Claude~E Shannon.
\newblock Programming a computer for playing chess.
\newblock In {\em Computer chess compendium}, pages 2--13. Springer, 1988.

\bibitem{shapiro1992turing}
Stuart~C Shapiro.
\newblock The {Turing Test} and the economist.
\newblock {\em ACM SIGART Bulletin}, 3(4):10--11, 1992.

\bibitem{sharma2018improved}
Rishi Sharma, Shane Barratt, Stefano Ermon, and Vijay Pande.
\newblock Improved training with curriculum {GANs}.
\newblock {\em arXiv preprint arXiv:1807.09295}, 2018.

\bibitem{silver2013lifelong}
Daniel~L Silver, Qiang Yang, and Lianghao Li.
\newblock Lifelong machine learning systems: Beyond learning algorithms.
\newblock In {\em 2013 AAAI Spring Symposium Series}, 2013.

\bibitem{silver2009reinforcement}
David Silver.
\newblock {\em Reinforcement learning and simulation based search in the game
  of {Go}}.
\newblock PhD thesis, University of Alberta, 2009.

\bibitem{silver2016mastering}
David Silver, Aja Huang, Chris~J. Maddison, Arthur Guez, Laurent Sifre, George
  van~den Driessche, Julian Schrittwieser, Ioannis Antonoglou, Veda
  Panneershelvam, Marc Lanctot, Sander Dieleman, Dominik Grewe, John Nham, Nal
  Kalchbrenner, Ilya Sutskever, Timothy Lillicrap, Madeleine Leach, Koray
  Kavukcuoglu, Thore Graepel, and Demis Hassabis.
\newblock Mastering the game of {Go} with deep neural networks and tree search.
\newblock {\em Nature}, 529(7587):484, 2016.

\bibitem{silver2017shogi}
David Silver, Thomas Hubert, Julian Schrittwieser, Ioannis Antonoglou, Matthew
  Lai, Arthur Guez, Marc Lanctot, Laurent Sifre, Dharshan Kumaran, Thore
  Graepel, et~al.
\newblock Mastering chess and shogi by self-play with a general reinforcement
  learning algorithm.
\newblock {\em arXiv preprint arXiv:1712.01815}, 2017.

\bibitem{silver2018general}
David Silver, Thomas Hubert, Julian Schrittwieser, Ioannis Antonoglou, Matthew
  Lai, Arthur Guez, Marc Lanctot, Laurent Sifre, Dharshan Kumaran, Thore
  Graepel, Timothy Lillicrap, Karen Simonyan, and Demis Hassabis.
\newblock A general reinforcement learning algorithm that masters chess, shogi,
  and {Go} through self-play.
\newblock {\em Science}, 362(6419):1140--1144, 2018.

\bibitem{silver2017mastering}
David Silver, Julian Schrittwieser, Karen Simonyan, Ioannis Antonoglou, Aja
  Huang, Arthur Guez, Thomas Hubert, Lucas Baker, Matthew Lai, Adrian Bolton,
  Yutian Chen, Timothy Lillicrap, Fan Hui, Laurent Sifre, George van~den
  Driessche, Thore Graepel, and Demis Hassabis.
\newblock Mastering the game of {Go} without human knowledge.
\newblock {\em Nature}, 550(7676):354, 2017.

\bibitem{silver2007reinforcement}
David Silver, Richard~S Sutton, and Martin M{\"u}ller.
\newblock Reinforcement learning of local shape in the game of {Go}.
\newblock In {\em IJCAI}, volume~7, pages 1053--1058, 2007.

\bibitem{silver2012temporal}
David Silver, Richard~S Sutton, and Martin M{\"u}ller.
\newblock Temporal-difference search in computer {Go}.
\newblock {\em Machine Learning}, 87(2):183--219, 2012.

\bibitem{silver2017predictron}
David Silver, Hado van Hasselt, Matteo Hessel, Tom Schaul, Arthur Guez, Tim
  Harley, Gabriel Dulac-Arnold, David Reichert, Neil Rabinowitz, Andre Barreto,
  et~al.
\newblock The predictron: End-to-end learning and planning.
\newblock In {\em Proceedings of the 34th International Conference on Machine
  Learning}, pages 3191--3199, 2017.

\bibitem{silver2010monte}
David Silver and Joel Veness.
\newblock {Monte-Carlo} planning in large {POMDPs}.
\newblock In {\em Advances in Neural Information Processing Systems}, pages
  2164--2172, 2010.

\bibitem{simonyan2014very}
Karen Simonyan and Andrew Zisserman.
\newblock Very deep convolutional networks for large-scale image recognition.
\newblock {\em arXiv preprint arXiv:1409.1556}, 2014.

\bibitem{singh1992reinforcement}
Satinder~P Singh.
\newblock Reinforcement learning with a hierarchy of abstract models.
\newblock In {\em Proceedings of the National Conference on Artificial
  Intelligence}, number~10, page 202. John Wiley \& Sons, 1992.

\bibitem{sironi2016comparison}
Chiara~F Sironi and Mark~HM Winands.
\newblock Comparison of rapid action value estimation variants for general game
  playing.
\newblock In {\em Computational Intelligence and Games (CIG), 2016 IEEE
  Conference on}, pages 1--8. IEEE, 2016.

\bibitem{sironi2017line}
Chiara~F Sironi and Mark~HM Winands.
\newblock On-line parameter tuning for {Monte-Carlo} tree search in {General
  Game Playing}.
\newblock In {\em Workshop on Computer Games}, pages 75--95. Springer, 2017.

\bibitem{smith1999neural}
Kate~A Smith.
\newblock Neural networks for combinatorial optimization: a review of more than
  a decade of research.
\newblock {\em INFORMS Journal on Computing}, 11(1):15--34, 1999.

\bibitem{smith1998computer}
Stephen~J Smith, Dana Nau, and Tom Throop.
\newblock Computer bridge: A big win for {AI} planning.
\newblock {\em AI Magazine}, 19(2):93, 1998.

\bibitem{smola1998learning}
Alex~J Smola and Bernhard Sch{\"o}lkopf.
\newblock {\em Learning with kernels}, volume~4.
\newblock Citeseer, 1998.

\bibitem{snoek2012practical}
Jasper Snoek, Hugo Larochelle, and Ryan~P Adams.
\newblock Practical {Bayesian} optimization of machine learning algorithms.
\newblock In {\em Advances in Neural Information Processing Systems}, pages
  2951--2959, 2012.

\bibitem{soekhoe2016impact}
Deepak Soekhoe, Peter Van Der~Putten, and Aske Plaat.
\newblock On the impact of data set size in transfer learning using deep neural
  networks.
\newblock In {\em International Symposium on Intelligent Data Analysis}, pages
  50--60. Springer, 2016.

\bibitem{sondik1978optimal}
Edward~J Sondik.
\newblock The optimal control of partially observable {Markov} processes over
  the infinite horizon: Discounted costs.
\newblock {\em Operations Research}, 26(2):282--304, 1978.

\bibitem{song2014scaling}
Fengguang Song and Jack Dongarra.
\newblock Scaling up matrix computations on shared-memory manycore systems with
  1000 {CPU} cores.
\newblock In {\em Proceedings of the 28th ACM International Conference on
  Supercomputing}, pages 333--342. ACM, 2014.

\bibitem{song2018mean}
Mei Song, A~Montanari, and P~Nguyen.
\newblock A mean field view of the landscape of two-layers neural networks.
\newblock In {\em Proceedings of the National Academy of Sciences}, volume 115,
  pages E7665--E7671, 2018.

\bibitem{srivastava2014dropout}
Nitish Srivastava, Geoffrey Hinton, Alex Krizhevsky, Ilya Sutskever, and Ruslan
  Salakhutdinov.
\newblock Dropout: a simple way to prevent neural networks from overfitting.
\newblock {\em The Journal of Machine Learning Research}, 15(1):1929--1958,
  2014.

\bibitem{stern2006bayesian}
David Stern, Ralf Herbrich, and Thore Graepel.
\newblock Bayesian pattern ranking for move prediction in the game of {Go}.
\newblock In {\em Proceedings of the 23rd International Conference on Machine
  Learning}, pages 873--880. ACM, 2006.

\bibitem{stockman1979minimax}
George~C Stockman.
\newblock A minimax algorithm better than alpha-beta?
\newblock {\em Artificial Intelligence}, 12(2):179--196, 1979.

\bibitem{stork2018priming}
Lise Stork, Katherine Wolstencroft, Andreas Weber, Fons Verbeek, and Aske
  Plaat.
\newblock Priming digitisation: Learning the textual structure in field books.
\newblock In {\em ICT. OPEN 2018: The Conference for ICT-Research in the
  Netherlands}, 2018.

\bibitem{stoutamire1991machine}
David Stoutamire.
\newblock Machine learning applied to {Go}.
\newblock {\em MS thesis}, 1991.

\bibitem{strachey1952logical}
Christopher~S Strachey.
\newblock Logical or non-mathematical programmes.
\newblock In {\em Proceedings of the 1952 ACM National Meeting (Toronto)},
  pages 46--49. ACM, 1952.

\bibitem{straus2018alphazero}
Darin Straus.
\newblock Alphazero implementation and tutorial
  \url{https://towardsdatascience.com/alphazero-implementation-and-tutorial-f4324d65fdfc},
  2018.

\bibitem{stulp2011hierarchical}
Freek Stulp and Stefan Schaal.
\newblock Hierarchical reinforcement learning with movement primitives.
\newblock In {\em Humanoid Robots (Humanoids), 2011 11th IEEE-RAS International
  Conference on}, pages 231--238. IEEE, 2011.

\bibitem{su2019one}
Jiawei Su, Danilo~Vasconcellos Vargas, and Kouichi Sakurai.
\newblock One pixel attack for fooling deep neural networks.
\newblock {\em IEEE Transactions on Evolutionary Computation}, 2019.

\bibitem{sukhbaatar2018learning}
Sainbayar Sukhbaatar, Emily Denton, Arthur Szlam, and Rob Fergus.
\newblock Learning goal embeddings via self-play for hierarchical reinforcement
  learning.
\newblock {\em arXiv preprint arXiv:1811.09083}, 2018.

\bibitem{sutskever2013training}
Ilya Sutskever.
\newblock {\em Training recurrent neural networks}.
\newblock PhD thesis, 2013.

\bibitem{sutskever2011generating}
Ilya Sutskever, James Martens, and Geoffrey~E Hinton.
\newblock Generating text with recurrent neural networks.
\newblock In {\em Proceedings of the 28th International Conference on Machine
  Learning (ICML-11)}, pages 1017--1024, 2011.

\bibitem{sutskever2008mimicking}
Ilya Sutskever and Vinod Nair.
\newblock Mimicking {Go} experts with convolutional neural networks.
\newblock In {\em International Conf. on Artificial Neural Networks}, pages
  101--110. Springer, 2008.

\bibitem{sutskever2014sequence}
Ilya Sutskever, Oriol Vinyals, and Quoc~V Le.
\newblock Sequence to sequence learning with neural networks.
\newblock In {\em Advances in neural information processing systems}, pages
  3104--3112, 2014.

\bibitem{sutton1984temporal}
Richard~S Sutton.
\newblock {\em Temporal credit assignment in reinforcement learning}.
\newblock PhD thesis, Univ Mass Amherst, 1984.

\bibitem{sutton1988learning}
Richard~S Sutton.
\newblock Learning to predict by the methods of temporal differences.
\newblock {\em Machine Learning}, 3(1):9--44, 1988.

\bibitem{sutton2018introduction}
Richard~S Sutton and Andrew~G Barto.
\newblock {\em Reinforcement learning, An Introduction, Second Edition}.
\newblock MIT Press, 2018.

\bibitem{sutton2000policy}
Richard~S Sutton, David~A McAllester, Satinder~P Singh, and Yishay Mansour.
\newblock Policy gradient methods for reinforcement learning with function
  approximation.
\newblock In {\em Advances in Neural Information Processing Systems}, pages
  1057--1063, 2000.

\bibitem{sutton1999between}
Richard~S Sutton, Doina Precup, and Satinder Singh.
\newblock Between {MDPs} and {semi-MDPs:} a framework for temporal abstraction
  in reinforcement learning.
\newblock {\em Artificial intelligence}, 112(1-2):181--211, 1999.

\bibitem{sze2017efficient}
Vivienne Sze, Yu-Hsin Chen, Tien-Ju Yang, and Joel~S Emer.
\newblock Efficient processing of deep neural networks: A tutorial and survey.
\newblock {\em Proceedings of the IEEE}, 105(12):2295--2329, 2017.

\bibitem{szegedy2015going}
Christian Szegedy, Wei Liu, Yangqing Jia, Pierre Sermanet, Scott Reed, Dragomir
  Anguelov, Dumitru Erhan, Vincent Vanhoucke, and Andrew Rabinovich.
\newblock Going deeper with convolutions.
\newblock In {\em Proceedings of the IEEE Conference on Computer Vision and
  Pattern Recognition}, pages 1--9, 2015.

\bibitem{szegedy2016rethinking}
Christian Szegedy, Vincent Vanhoucke, Sergey Ioffe, Jon Shlens, and Zbigniew
  Wojna.
\newblock Rethinking the inception architecture for computer vision.
\newblock In {\em Proceedings of the IEEE Conference on Computer Vision and
  Pattern Recognition}, pages 2818--2826, 2016.

\bibitem{szegedy2013intriguing}
Christian Szegedy, Wojciech Zaremba, Ilya Sutskever, Joan Bruna, Dumitru Erhan,
  Ian Goodfellow, and Rob Fergus.
\newblock Intriguing properties of neural networks.
\newblock {\em arXiv preprint arXiv:1312.6199}, 2013.

\bibitem{szita2009monte}
Istv{\'a}n Szita, Guillaume Chaslot, and Pieter Spronck.
\newblock {Monte-Carlo} tree search in {Settlers of Catan}.
\newblock In {\em Advances in Computer Games}, pages 21--32. Springer, 2009.

\bibitem{tamar2016value}
Aviv Tamar, Yi~Wu, Garrett Thomas, Sergey Levine, and Pieter Abbeel.
\newblock Value iteration networks.
\newblock In {\em Adv. in Neural Information Processing Systems}, pages
  2154--2162, 2016.

\bibitem{tan1993multi}
Ming Tan.
\newblock Multi-agent reinforcement learning: Independent vs.\ cooperative
  agents.
\newblock In {\em Proceedings of the Tenth Intl. Conf. on Machine Learning},
  pages 330--337, 1993.

\bibitem{tanno2018adaptive}
Ryutaro Tanno, Kai Arulkumaran, Daniel~C Alexander, Antonio Criminisi, and
  Aditya Nori.
\newblock Adaptive neural trees.
\newblock {\em arXiv preprint arXiv:1807.06699}, 2018.

\bibitem{taylor2009transfer}
Matthew~E Taylor and Peter Stone.
\newblock Transfer learning for reinforcement learning domains: A survey.
\newblock {\em Journal of Machine Learning Research}, 10(Jul):1633--1685, 2009.

\bibitem{tekofsky2015past}
Shoshannah Tekofsky, Pieter Spronck, Martijn Goudbeek, Aske Plaat, and Jaap
  van~den Herik.
\newblock Past our prime: A study of age and play style development in
  {Battlefield 3}.
\newblock {\em IEEE Transactions on Computational Intelligence and AI in
  Games}, 7(3):292--303, 2015.

\bibitem{tenenbaum2000global}
Joshua~B Tenenbaum, Vin De~Silva, and John~C Langford.
\newblock A global geometric framework for nonlinear dimensionality reduction.
\newblock {\em Science}, 290(5500):2319--2323, 2000.

\bibitem{tesauro1989neurogammon}
Gerald Tesauro.
\newblock Neurogammon wins {Computer Olympiad}.
\newblock {\em Neural Computation}, 1(3):321--323, 1989.

\bibitem{tesauro1990neurogammon}
Gerald Tesauro.
\newblock Neurogammon: A neural-network backgammon program.
\newblock In {\em 1990 IJCNN Intl. Joint Conference on Neural Networks}, pages
  33--39. IEEE, 1990.

\bibitem{tesauro1995td}
Gerald Tesauro.
\newblock {TD}-gammon: A self-teaching backgammon program.
\newblock In {\em Applications of Neural Networks}, pages 267--285. Springer,
  1995.

\bibitem{tesauro1995temporal}
Gerald Tesauro.
\newblock Temporal difference learning and {TD}-{G}ammon.
\newblock {\em Communications of the ACM}, 38(3):58--68, 1995.

\bibitem{tesauro2002programming}
Gerald Tesauro.
\newblock Programming backgammon using self-teaching neural nets.
\newblock {\em Artificial Intelligence}, 134(1-2):181--199, 2002.

\bibitem{teyssier2012ordering}
Marc Teyssier and Daphne Koller.
\newblock Ordering-based search: A simple and effective algorithm for learning
  {Bayesian} networks.
\newblock {\em arXiv preprint arXiv:1207.1429}, 2012.

\bibitem{nair2017learning}
Shantanu Thakoor, Surag Nair, and Megha Jhunjhunwala.
\newblock Learning to play othello without human knowledge.
\newblock Stanford University CS238 Final Project Report, 2017.

\bibitem{thaler2008nudge}
Richard~H Thaler and Cass~R Sunstein.
\newblock {\em Nudge: Improving decisions about health, wealth, and happiness}.
\newblock Penguin, 2009.

\bibitem{thielscher2009answer}
Michael Thielscher.
\newblock Answer set programming for single-player games in general game
  playing.
\newblock In {\em International Conference on Logic Programming}, pages
  327--341. Springer, 2009.

\bibitem{thielscher2010general}
Michael Thielscher.
\newblock A general game description language for incomplete information games.
\newblock In {\em AAAI}, volume~10, pages 994--999, 2010.

\bibitem{thielscher2011general}
Michael Thielscher.
\newblock The general game playing description language is universal.
\newblock In {\em IJCAI Proceedings-International Joint Conference on
  Artificial Intelligence}, volume~22, page 1107, 2011.

\bibitem{thompson1986retrograde}
Ken Thompson.
\newblock Retrograde analysis of certain endgames.
\newblock {\em ICCA Journal}, 9(3):131--139, 1986.

\bibitem{thornton2013auto}
Chris Thornton, Frank Hutter, Holger~H Hoos, and Kevin Leyton-Brown.
\newblock {Auto-WEKA:} combined selection and hyperparameter optimization of
  classification algorithms.
\newblock In {\em Proceedings of the 19th ACM SIGKDD International Conference
  on Knowledge Discovery and Data Mining}, pages 847--855. ACM, 2013.

\bibitem{thrun1995learning}
Sebastian Thrun.
\newblock Learning to play the game of chess.
\newblock In {\em Advances in Neural Information Processing Systems}, pages
  1069--1076, 1995.

\bibitem{thrun2012explanation}
Sebastian Thrun.
\newblock {\em Explanation-based neural network learning: A lifelong learning
  approach}, volume 357.
\newblock Springer Science \& Business Media, 2012.

\bibitem{thrun2012learning}
Sebastian Thrun and Lorien Pratt.
\newblock {\em Learning to learn}.
\newblock Springer Science \& Business Media, 2012.

\bibitem{tian2017elf}
Yuandong Tian, Qucheng Gong, Wenling Shang, Yuxin Wu, and C~Lawrence Zitnick.
\newblock Elf: An extensive, lightweight and flexible research platform for
  real-time strategy games.
\newblock In {\em Advances in Neural Information Processing Systems}, pages
  2659--2669, 2017.

\bibitem{ELFOpenGo2018}
Yuandong Tian, {Jerry Ma}, {Qucheng Gong}, Shubho Sengupta, Zhuoyuan Chen, and
  C.~Lawrence Zitnick.
\newblock {ELF OpenGo}.
\newblock \url{https://github.com/pytorch/ELF}, 2018.

\bibitem{tian2015better}
Yuandong Tian and Yan Zhu.
\newblock Better computer {Go} player with neural network and long-term
  prediction.
\newblock {\em arXiv preprint arXiv:1511.06410}, 2015.

\bibitem{togelius2010multiobjective}
Julian Togelius, Mike Preuss, Nicola Beume, Simon Wessing, Johan Hagelb{\"a}ck,
  and Georgios~N Yannakakis.
\newblock Multiobjective exploration of the starcraft map space.
\newblock In {\em Proceedings of the 2010 IEEE Conf. on Computational
  Intelligence and Games}, pages 265--272. IEEE, 2010.

\bibitem{togelius2010search}
Julian Togelius, Georgios~N Yannakakis, Kenneth~O Stanley, and Cameron Browne.
\newblock Search-based procedural content generation.
\newblock In {\em European Conference on the Applications of Evolutionary
  Computation}, pages 141--150. Springer, 2010.

\bibitem{toubman2015transfer}
Armon Toubman, Jan~Joris Roessingh, Pieter Spronck, Aske Plaat, and Jaap
  van~den Herik.
\newblock Transfer learning of air combat behavior.
\newblock In {\em 2015 IEEE 14th International Conference on Machine Learning
  and Applications (ICMLA)}, pages 226--231. IEEE, 2015.

\bibitem{tran2017disentangled}
Luan Tran, Xi~Yin, and Xiaoming Liu.
\newblock Disentangled representation learning {GAN} for pose-invariant face
  recognition.
\newblock In {\em CVPR}, volume~3, page~7, 2017.

\bibitem{triantafillou2019meta}
Eleni Triantafillou, Tyler Zhu, Vincent Dumoulin, Pascal Lamblin, Utku Evci,
  Kelvin Xu, Ross Goroshin, Carles Gelada, Kevin Swersky, Pierre-Antoine
  Manzagol, et~al.
\newblock Meta-dataset: A dataset of datasets for learning to learn from few
  examples.
\newblock {\em arXiv preprint arXiv:1903.03096}, 2019.

\bibitem{tromp2016}
John Tromp.
\newblock Number of legal {Go} states, 2016.

\bibitem{tromp2006combinatorics}
John Tromp and Gunnar Farneb{\"a}ck.
\newblock Combinatorics of {Go}.
\newblock In {\em International Conference on Computers and Games}, pages
  84--99, \url{https://tromp.github.io/go/gostate.pdf}, 2006. Springer.

\bibitem{tsitsiklis1997analysis}
John~N Tsitsiklis and Benjamin Van~Roy.
\newblock Analysis of temporal-diffference learning with function
  approximation.
\newblock In {\em Advances in Neural Information Processing Systems}, pages
  1075--1081, 1997.

\bibitem{turing1950computing}
Alan~M Turing.
\newblock Computing machinery and intelligence.
\newblock In {\em Parsing the Turing Test}, pages 23--65. Springer, 1950, 2009.

\bibitem{turing1953digital}
Alan~M Turing.
\newblock Digital computers applied to games.
\newblock {\em Faster than thought}, 1953.

\bibitem{tzanetakis2002musical}
George Tzanetakis and Perry Cook.
\newblock Musical genre classification of audio signals.
\newblock {\em IEEE Transactions on Speech and Audio Processing},
  10(5):293--302, 2002.

\bibitem{uriarte2015benchmark}
Alberto Uriarte and Santiago Ontan{\'o}n.
\newblock A benchmark for starcraft intelligent agents.
\newblock In {\em 11th Artificial Intelligence and Interactive Digital
  Entertainment Conference}, 2015.

\bibitem{shogi}
Computer Shogi~Association. \url{http://www2.computer-shogi.org/index_e.html}.
\newblock Website.

\bibitem{valiant2008knowledge}
Leslie~G Valiant.
\newblock Knowledge infusion: In pursuit of robustness in artificial
  intelligence.
\newblock In {\em IARCS Annual Conference on Foundations of Software Technology
  and Theoretical Computer Science}. Schloss Dagstuhl-Leibniz-Zentrum f{\"u}r
  Informatik, 2008.

\bibitem{vamvoudakis2010online}
Kyriakos~G Vamvoudakis and Frank~L Lewis.
\newblock Online {Actor--Critic} algorithm to solve the continuous-time
  infinite horizon optimal control problem.
\newblock {\em Automatica}, 46(5):878--888, 2010.

\bibitem{van2013reinforcement}
Michiel Van Der~Ree and Marco Wiering.
\newblock Reinforcement learning in the game of othello: learning against a
  fixed opponent and learning from self-play.
\newblock In {\em 2013 IEEE Symposium on Adaptive Dynamic Programming and
  Reinforcement Learning (ADPRL)}, pages 108--115. IEEE, 2013.

\bibitem{van2004ai}
Erik Van Der~Werf.
\newblock {\em AI techniques for the game of {Go}}.
\newblock PhD thesis, Maastricht University, 2004.

\bibitem{van2002local}
Erik Van Der~Werf, Jos~WHM Uiterwijk, Eric Postma, and Jaap Van Den~Herik.
\newblock Local move prediction in {Go}.
\newblock In {\em International Conference on Computers and Games}, pages
  393--412. Springer, 2002.

\bibitem{van2018deep}
Hado Van~Hasselt, Yotam Doron, Florian Strub, Matteo Hessel, Nicolas Sonnerat,
  and Joseph Modayil.
\newblock Deep reinforcement learning and the deadly triad.
\newblock {\em preprint arXiv:1812.02648}, 2018.

\bibitem{van2016deep}
Hado Van~Hasselt, Arthur Guez, and David Silver.
\newblock Deep reinforcement learning with {Double Q-Learning}.
\newblock In {\em AAAI}, volume~2, page~5. Phoenix, AZ, 2016.

\bibitem{van2011non}
Matthijs Van~Leeuwen and Arno Knobbe.
\newblock Non-redundant subgroup discovery in large and complex data.
\newblock In {\em Joint European Conference on Machine Learning and Knowledge
  Discovery in Databases}, pages 459--474. Springer, 2011.

\bibitem{van2004concepts}
Peter Van-Roy, Seif Haridi, et~al.
\newblock {\em Concepts, techniques, and models of computer programming}.
\newblock MIT Press, 2004.

\bibitem{van2018automatic}
Bas van Stein, Hao Wang, and Thomas B{\"a}ck.
\newblock Automatic configuration of deep neural networks with ego.
\newblock {\em arXiv preprint arXiv:1810.05526}, 2018.

\bibitem{van2015optimally}
Bas van Stein, Hao Wang, Wojtek Kowalczyk, Thomas B{\"a}ck, and Michael
  Emmerich.
\newblock Optimally weighted cluster kriging for big data regression.
\newblock In {\em International Symposium on Intelligent Data Analysis}, pages
  310--321. Springer, 2015.

\bibitem{van2011compound}
Gerard~JP Van~Westen, J{\"o}rg~K Wegner, Peggy Geluykens, Leen Kwanten, Inge
  Vereycken, Anik Peeters, Adriaan~P IJzerman, Herman~WT van Vlijmen, and
  Andreas Bender.
\newblock Which compound to select in lead optimization? prospectively
  validated proteochemometric models guide preclinical development.
\newblock {\em PloS One}, 6(11):e27518, 2011.

\bibitem{vanschoren2018meta}
Joaquin Vanschoren.
\newblock Meta-learning: A survey.
\newblock {\em arXiv preprint arXiv:1810.03548}, 2018.

\bibitem{francisco1991embodied}
Francisco Varela, Evan Thompson, and Eleanor Rosch.
\newblock {\em The embodied mind: cognitive science and human experience}.
\newblock MIT Press, Cambridge, 1991.

\bibitem{veness2009bootstrapping}
Joel Veness, David Silver, Alan Blair, and William Uther.
\newblock Bootstrapping from game tree search.
\newblock In {\em Advances in Neural Information Processing Systems}, pages
  1937--1945, 2009.

\bibitem{venugopalan2015sequence}
Subhashini Venugopalan, Marcus Rohrbach, Jeffrey Donahue, Raymond Mooney,
  Trevor Darrell, and Kate Saenko.
\newblock Sequence to sequence-video to text.
\newblock In {\em Proceedings of the IEEE International Conference on Computer
  Vision}, pages 4534--4542, 2015.

\bibitem{vermaseren2000new}
Jos~AM Vermaseren.
\newblock New features of form.
\newblock {\em arXiv preprint math-ph/0010025}, 2000.

\bibitem{vermaseren2014investigations}
Jos~AM Vermaseren, Aske Plaat, Jan Kuiper, and Jaap van~den Herik.
\newblock Investigations with {Monte Carlo} tree search for finding better
  multivariate {Horner} schemes.
\newblock In {\em Agents and Artificial Intelligence: 5th International
  Conference, ICAART 2013, Barcelona, Spain, February 15-18, 2013. Revised
  Selected Papers}, volume 449, page~3. Springer, 2014.

\bibitem{vezhnevets2016strategic}
Alexander Vezhnevets, Volodymyr Mnih, Simon Osindero, Alex Graves, Oriol
  Vinyals, John Agapiou, et~al.
\newblock Strategic attentive writer for learning macro-actions.
\newblock In {\em Advances in Neural Information Processing Systems}, pages
  3486--3494, 2016.

\bibitem{vilalta2002perspective}
Ricardo Vilalta and Youssef Drissi.
\newblock A perspective view and survey of meta-learning.
\newblock {\em Artificial Intelligence Review}, 18(2):77--95, 2002.

\bibitem{vinyals2019grandmaster}
Oriol Vinyals, Igor Babuschkin, Wojciech~M Czarnecki, Micha{\"e}l Mathieu,
  Andrew Dudzik, Junyoung Chung, David~H Choi, Richard Powell, Timo Ewalds,
  Petko Georgiev, et~al.
\newblock Grandmaster level in {StarCraft II} using multi-agent reinforcement
  learning.
\newblock {\em Nature}, 575(7782):350--354, 2019.

\bibitem{vinyals2016matching}
Oriol Vinyals, Charles Blundell, Timothy Lillicrap, Daan Wierstra, et~al.
\newblock Matching networks for one shot learning.
\newblock In {\em Advances in Neural Information Processing Systems}, pages
  3630--3638, 2016.

\bibitem{vinyals2017starcraft}
Oriol Vinyals, Timo Ewalds, Sergey Bartunov, Petko Georgiev, Alexander~Sasha
  Vezhnevets, Michelle Yeo, Alireza Makhzani, Heinrich K{\"u}ttler, John
  Agapiou, Julian Schrittwieser, et~al.
\newblock {Starcraft II}: A new challenge for reinforcement learning.
\newblock {\em arXiv preprint arXiv:1708.04782}, 2017.

\bibitem{vinyals2015pointer}
Oriol Vinyals, Meire Fortunato, and Navdeep Jaitly.
\newblock Pointer networks.
\newblock In {\em Advances in Neural Information Processing Systems}, pages
  2692--2700, 2015.

\bibitem{vinyals2015show}
Oriol Vinyals, Alexander Toshev, Samy Bengio, and Dumitru Erhan.
\newblock Show and tell: A neural image caption generator.
\newblock In {\em Proceedings of the IEEE Conference on Computer Vision and
  Pattern Recognition}, pages 3156--3164, 2015.

\bibitem{von1944theory}
John Von~Neumann and Oskar Morgenstern.
\newblock {\em Theory of games and economic behavior}.
\newblock Princeton University Press, 1944.

\bibitem{von2007theory}
John Von~Neumann, Oskar Morgenstern, and Harold~William Kuhn.
\newblock {\em Theory of games and economic behavior (commemorative edition)}.
\newblock Princeton University Press, 2007.

\bibitem{quoc2018}
Loc Vu-Quoc.
\newblock Neuron and myelinated axon, 2018.

\bibitem{wahlstrom2015pixels}
Niklas Wahlstr{\"o}m, Thomas~B Sch{\"o}n, and Marc~Peter Deisenroth.
\newblock From pixels to torques: Policy learning with deep dynamical models.
\newblock {\em arXiv preprint arXiv:1502.02251}, 2015.

\bibitem{walraet2016googolplex}
Matthieu Walraet and John Tromp.
\newblock A googolplex of {Go} games.
\newblock In {\em International Conference on Computers and Games}, pages
  191--201. Springer, 2016.

\bibitem{wang2019on}
Hao Wang, Thomas B{\"a}ck, Aske Plaat, Michael Emmerich, and Mike Preuss.
\newblock On the potential of evolutionary algorithms for replacing
  backpropagation for network weight optimization.
\newblock In {\em GECCO}, 2019.

\bibitem{wang2018assessing}
Hui Wang, Michael Emmerich, and Aske Plaat.
\newblock Assessing the potential of {Classical Q-learning in General Game
  Playing}.
\newblock {\em arXiv preprint arXiv:1810.06078}, 2018.

\bibitem{wang2019alternative}
Hui Wang, Michael Emmerich, Mike Preuss, and Aske Plaat.
\newblock Alternative loss functions in {AlphaZero}-like self-play.
\newblock In {\em 2019 IEEE Symposium Series on Computational Intelligence
  (SSCI)}, pages 155--162, 2019.

\bibitem{wang2019hyper}
Hui Wang, Michael Emmerich, Mike Preuss, and Aske Plaat.
\newblock Hyper-parameter sweep on {AlphaZero General}.
\newblock arxiv, arXiv:1903.08129, 2019.

\bibitem{wang2020analysis}
Hui Wang, Michael Emmerich, Mike Preuss, and Aske Plaat.
\newblock Analysis of hyper-parameters for small games: Iterations or epochs in
  self-play?
\newblock {\em arXiv preprint arXiv:2003.05988}, 2020.

\bibitem{wang2018search}
Hui Wang, Yanni Tang, Jiamou Liu, and Wu~Chen.
\newblock A search optimization method for rule learning in board games.
\newblock In Xin Geng and Byeong-Ho Kang, editors, {\em PRICAI 2018: Trends in
  Artificial Intelligence}, pages 174--181, Cham, 2018. Springer International
  Publishing.

\bibitem{wang2016learning}
Jane~X Wang, Zeb Kurth-Nelson, Dhruva Tirumala, Hubert Soyer, Joel~Z Leibo,
  Remi Munos, Charles Blundell, Dharshan Kumaran, and Matt Botvinick.
\newblock Learning to reinforcement learn.
\newblock {\em arXiv preprint arXiv:1611.05763}, 2016.

\bibitem{wang2015basic}
Panqu Wang and Garrison~W Cottrell.
\newblock Basic level categorization facilitates visual object recognition.
\newblock {\em arXiv preprint arXiv:1511.04103}, 2015.

\bibitem{wang2019benchmarking}
Tingwu Wang, Xuchan Bao, Ignasi Clavera, Jerrick Hoang, Yeming Wen, Eric
  Langlois, Shunshi Zhang, Guodong Zhang, Pieter Abbeel, and Jimmy Ba.
\newblock Benchmarking model-based reinforcement learning.
\newblock {\em preprint arXiv:1907.02057}, 2019.

\bibitem{wang2020cnn}
Zijie~J Wang, Robert Turko, Omar Shaikh, Haekyu Park, Nilaksh Das, Fred Hohman,
  Minsuk Kahng, and Duen~Horng Chau.
\newblock Cnn explainer: Learning convolutional neural networks with
  interactive visualization.
\newblock {\em arXiv preprint arXiv:2004.15004}, 2020.

\bibitem{wang2016bayesian}
Ziyu Wang, Frank Hutter, Masrour Zoghi, David Matheson, and Nando de~Feitas.
\newblock Bayesian optimization in a billion dimensions via random embeddings.
\newblock {\em Journal of Artificial Intelligence Research}, 55:361--387, 2016.

\bibitem{wang2016dueling}
Ziyu Wang, Tom Schaul, Matteo Hessel, Hado Hasselt, Marc Lanctot, and Nando
  Freitas.
\newblock Dueling network architectures for deep reinforcement learning.
\newblock In {\em International conference on machine learning}, pages
  1995--2003, 2016.

\bibitem{wang2015dueling}
Ziyu Wang, Tom Schaul, Matteo Hessel, Hado Van~Hasselt, Marc Lanctot, and Nando
  De~Freitas.
\newblock Dueling network architectures for deep reinforcement learning.
\newblock {\em arXiv preprint arXiv:1511.06581}, 2015.

\bibitem{watkins1989learning}
Christopher~JCH Watkins.
\newblock {\em Learning from delayed rewards}.
\newblock PhD thesis, King's College, Cambridge, 1989.

\bibitem{watter2015embed}
Manuel Watter, Jost Springenberg, Joschka Boedecker, and Martin Riedmiller.
\newblock Embed to control: A locally linear latent dynamics model for control
  from raw images.
\newblock In {\em Advances in Neural Information Processing Systems}, pages
  2746--2754, 2015.

\bibitem{racaniere2017imagination}
Th{\'e}ophane Weber, S{\'e}bastien Racani{\`e}re, David Reichert, Lars Buesing,
  Arthur Guez, Danilo~Jimenez Rezende, Adria~Puigdomenech Badia, Oriol Vinyals,
  Nicolas Heess, Yujia Li, et~al.
\newblock Imagination-augmented agents for deep reinforcement learning.
\newblock In {\em Advances in Neural Information Processing Systems}, pages
  5690--5701, 2017.

\bibitem{weill}
Eddie Weill.
\newblock {LeNet} in {Keras} on {Github}.
  \url{https://github.com/eweill/keras-deepcv/tree/master/models/classification}.

\bibitem{weill1992negac}
Jean-Christophe Weill.
\newblock The {NegaC*} search.
\newblock {\em ICGA Journal}, 15(1):3--7, 1992.

\bibitem{weinshall2018curriculum}
Daphna Weinshall, Gad Cohen, and Dan Amir.
\newblock Curriculum learning by transfer learning: Theory and experiments with
  deep networks.
\newblock {\em preprint arXiv:1802.03796}, 2018.

\bibitem{weizenbaum1966eliza}
Joseph Weizenbaum et~al.
\newblock Eliza---a computer program for the study of natural language
  communication between man and machine.
\newblock {\em Comm. of the ACM}, 9(1):36--45, 1966.

\bibitem{weng2020curriculum}
Lilian Weng.
\newblock Curriculum for reinforcement learning
  \url{https://lilianweng.github.io/lil-log/2020/01/29/curriculum-for-reinforcement-learning.html}.
\newblock Lil'Log, January 2020.

\bibitem{werbos1974beyond}
Paul Werbos.
\newblock Beyond regression: New tools for prediction and analysis in the
  behavioral sciences.
\newblock {\em Ph. D. dissertation, Harvard University}, 1974.

\bibitem{wiering2010self}
Marco~A Wiering.
\newblock Self-play and using an expert to learn to play backgammon with
  temporal difference learning.
\newblock {\em JILSA}, 2(2):57--68, 2010.

\bibitem{wierstra2008natural}
Daan Wierstra, Tom Schaul, Jan Peters, and J\"urgen Schmidhuber.
\newblock Natural evolution strategies.
\newblock In {\em Evolutionary Computation, 2008. CEC 2008.(IEEE World Congress
  on Computational Intelligence). IEEE Congress on}, pages 3381--3387. IEEE,
  2008.

\bibitem{williams1992simple}
Ronald~J Williams.
\newblock Simple statistical gradient-following algorithms for connectionist
  reinforcement learning.
\newblock {\em Machine Learning}, 8(3-4):229--256, 1992.

\bibitem{wilson2009error}
Brandon Wilson, Austin Parker, and DS~Nau.
\newblock Error minimizing minimax: Avoiding search pathology in game trees.
\newblock In {\em Proceedings of International Symposium on Combinatorial
  Search (SoCS-09)}, 2009.

\bibitem{witten1976apparent}
Ian~H Witten.
\newblock The apparent conflict between estimation and control---a survey of
  the two-armed bandit problem.
\newblock {\em Journal of the Franklin Institute}, 301(1-2):161--189, 1976.

\bibitem{witten2016data}
Ian~H Witten, Eibe Frank, Mark~A Hall, and Christopher~J Pal.
\newblock {\em Data Mining: Practical machine learning tools and techniques}.
\newblock Morgan Kaufmann, 2016.

\bibitem{wong2017}
Anson Wong.
\newblock Solving the multi-armed bandit problem.
\newblock {\em Towards Data Science}, 2017
  \url{https://towardsdatascience.com/solving-the-multi-armed-bandit-problem-b72de40db97c}.

\bibitem{wos1984automated}
Larry Wos, Ross Overbeck, Ewing Lusk, and Jim Boyle.
\newblock {\em Automated reasoning: introduction and applications}.
\newblock Prentice Hall Inc., Old Tappan, NJ, 1984.

\bibitem{wu2019accelerating}
David~J Wu.
\newblock Accelerating self-play learning in {Go}.
\newblock {\em preprint arXiv:1902.10565}, 2019.

\bibitem{wujastyk2003roots}
Dominik Wujastyk.
\newblock {\em The roots of ayurveda: Selections from Sanskrit medical
  writings}.
\newblock Penguin, 2003.

\bibitem{xian2017zero}
Yongqin Xian, Bernt Schiele, and Zeynep Akata.
\newblock Zero-shot learning---the good, the bad and the ugly.
\newblock In {\em Proceedings of the IEEE Conference on Computer Vision and
  Pattern Recognition}, pages 4582--4591, 2017.

\bibitem{xu2019learning}
Ruiyang Xu and Karl Lieberherr.
\newblock Learning self-game-play agents for combinatorial optimization
  problems.
\newblock {\em arXiv preprint arXiv:1903.03674}, 2019.

\bibitem{yang2019exposing}
Xin Yang, Yuezun Li, and Siwei Lyu.
\newblock Exposing deep fakes using inconsistent head poses.
\newblock In {\em ICASSP 2019-2019 IEEE International Conference on Acoustics,
  Speech and Signal Processing}, pages 8261--8265. IEEE, 2019.

\bibitem{yannakakis2018artificial}
Georgios~N Yannakakis and Julian Togelius.
\newblock {\em Artificial intelligence and games}.
\newblock Springer, 2018.

\bibitem{yao1999evolving}
Xin Yao.
\newblock Evolving artificial neural networks.
\newblock {\em Proceedings of the IEEE}, 87(9):1423--1447, 1999.

\bibitem{yoshinari2018probabilistic}
Nozomu Yoshinari, Kento Uchida, Shota Saito, Shinichi Shirakawa, and Youhei
  Akimoto.
\newblock Probabilistic model-based dynamic architecture search.
\newblock In {\em International Conference on Learning Representations (ICLR)},
  2019.

\bibitem{yu2019meta}
Tianhe Yu, Deirdre Quillen, Zhanpeng He, Ryan Julian, Karol Hausman, Chelsea
  Finn, and Sergey Levine.
\newblock Meta-world: A benchmark and evaluation for multi-task and meta
  reinforcement learning.
\newblock {\em arXiv preprint arXiv:1910.10897}, 2019.

\bibitem{zeiler2014visualizing}
Matthew~D Zeiler and Rob Fergus.
\newblock Visualizing and understanding convolutional networks.
\newblock In {\em European Conference on Computer Vision}, pages 818--833.
  Springer, 2014.

\bibitem{PhoenixGo2018}
Qinsong Zeng, Jianchang Zhang, Zhanpeng Zeng, Yongsheng Li, Ming Chen, and
  Sifan Liu.
\newblock {PhoenixGo}.
\newblock \url{https://github.com/Tencent/PhoenixGo}, 2018.

\bibitem{zhang2018metagan}
Ruixiang Zhang, Tong Che, Zoubin Ghahramani, Yoshua Bengio, and Yangqiu Song.
\newblock {MetaGAN}: An adversarial approach to {Few-Shot} learning.
\newblock In {\em Advances in Neural Information Processing Systems}, pages
  2365--2374, 2018.

\bibitem{zhang2017deeper}
Shangtong Zhang and Richard~S Sutton.
\newblock A deeper look at experience replay.
\newblock {\em arXiv preprint arXiv:1712.01275}, 2017.

\bibitem{zhou2018}
Yuan Zhou.
\newblock {\em Rethinking Opening Strategy: {AlphaGo's} Impact on Pro Play}.
\newblock CreateSpace, 2018.

\bibitem{zinkevich2008regret}
Martin Zinkevich, Michael Johanson, Michael Bowling, and Carmelo Piccione.
\newblock Regret minimization in games with incomplete information.
\newblock In {\em Advances in Neural Information Processing Systems}, pages
  1729--1736, 2008.

\bibitem{zobrist1970new}
Albert~L Zobrist.
\newblock A new hashing method with application for game playing by.
\newblock Technical report, University of Wisconsin, Madison, 1970.

\bibitem{zoph2016neural}
Barret Zoph and Quoc~V Le.
\newblock Neural architecture search with reinforcement learning.
\newblock {\em arXiv preprint arXiv:1611.01578}, 2016.

\end{thebibliography}

\addcontentsline{toc}{chapter}{Figures}
\listoffigures
\addcontentsline{toc}{chapter}{Tables}
\listoftables
\addcontentsline{toc}{chapter}{Algorithms}
\lstlistoflistings

\printindex

\end{document}